%% file: paper.tex
\def\year{2021}\relax
\documentclass[letterpaper]{article} % DO NOT CHANGE THIS
\usepackage{aaai21}  % DO NOT CHANGE THIS
\usepackage{times}  % DO NOT CHANGE THIS
\usepackage{helvet} % DO NOT CHANGE THIS
\usepackage{courier}  % DO NOT CHANGE THIS
\usepackage[hyphens]{url}  % DO NOT CHANGE THIS
\usepackage{graphicx} % DO NOT CHANGE THIS
\usepackage{natbib}  % DO NOT CHANGE THIS AND DO NOT ADD ANY OPTIONS TO IT
\usepackage{caption} % DO NOT CHANGE THIS AND DO NOT ADD ANY OPTIONS TO IT
\newcommand{\indep}{\perp \!\!\! \perp}
\usepackage{algorithm}
\usepackage[noend]{algpseudocode}
\usepackage{amssymb}
\usepackage{amsmath}
\usepackage{tikz,pgfplots}
\usepackage{subcaption}
\usepackage{hyperref}
\hypersetup{
    colorlinks=true,
    linkcolor=blue,
    filecolor=blue,      
    urlcolor=blue,
    citecolor=blue,
    }
\usepackage[capitalise]{cleveref}

\title{Causal Identification with Additive Noise Models: Quantifying the Effect of Noise}
\author {
    Benjamin Kap, %\textsuperscript{\rm 1}, 
    Marharyta Aleksandrova, %\textsuperscript{\rm 1},
    Thomas Engel%\textsuperscript{\rm 1} 
    \\
}
\affiliations {
    University of Luxembourg, 2 avenue de l'Université, L-4365 Esch-sur-Alzette, Luxembourg \\
    \{benjamin.kap.001,
    marharyta.aleksandrova, thomas.engel\}@uni.lu
}

\begin{document}

\maketitle

\input{sections/00-abstract}

\input{sections/02-literature}

\input{sections/03-model}

\input{sections/04-experimental-setup}

\input{sections/05-experimental-results}

\input{sections/06-conclusions}

\input{sections/07-acknowledgement}

\bibliography{paper}

\end{document}

%% file: sections/00-abstract.tex
\begin{abstract}
In recent years, a lot of research has been conducted within the area
of causal inference and causal learning. Many methods have been developed
to identify the cause-effect pairs in models and have been successfully
applied to observational real-world data to determine
the direction of causal relationships. 
Yet in bivariate
situations, causal discovery problems remain challenging. One class of such
methods, that also allows tackling the bivariate case, is based on Additive Noise Models (ANMs). Unfortunately, one aspect
of these methods has not received much attention until now: what is the impact of different noise levels 
on the ability of these methods to identify the direction of the causal relationship.
This work aims to bridge this gap with the help of an empirical study. 
We test \textit{Regression with Subsequent Independence Test}
using an exhaustive range of models where the level of additive noise gradually changes
from 1\% to 10000\% of the causes' noise level (the latter remains fixed). 
Additionally, the experiments in this work 
consider several different types of distributions as well as linear and non-linear
models. The results of the experiments show that ANMs methods can fail to capture
the true causal direction for some levels of noise.
\end{abstract}

%% file: sections/02-literature.tex
\section{Introduction \& Related Work}
\label{sec:literature}

\textit{Causal identification} is the procedure of determining causal relationship direction
from observational data only and representing these as a (causal) graph. This problem is closely related to \textit{structure learning of Bayesian networks},
as static causal graphs are often represented and studied as
Bayesian networks.

The basic idea of structure learning 
emerged from \citet{wright1921correlation} as \textit{path analysis}. 
%This is used to describe directed dependencies among a set of variables and includes various models such as ANOVA, ANCOVA, etc. 
In his work, Wright made a distinction between three possible
types of causal substructures that were allowed in a directed acyclic graph:
%\begin{equation}
%    \label{eq:structure1}
%    X \to Y \to Z,
%\end{equation}
%\begin{equation}
%    \label{eq:structure2}
%    X \gets Y \to Z,
%\end{equation}
%\begin{equation}
%    \label{eq:structure3}
%    X \to Y \gets Z
%\end{equation}
$$X \to Y \to Z \text{, or } X \gets Y \to Z \text{, or }  X \to Y \gets Z.$$
Later, \citet{rebane} developed 
an algorithm to recover directed acyclic graphs from statistical data, which relied
on the distinction of these substructures. %In general, one can easily
%identify the skeleton of a graph (that is the graph without arrows on the edges) and then
%partially identify the arrows (partially, because the three substructures' skeletons are identical
%but only 3) is distinguishable from others).
\citet{spirtes2000causation} used Bayes networks to axiomatize the connection
between causal structure and probabilistic independence and formalized under what assumptions
one could draw causal knowledge from observational data only. Furthermore, they also formalized
how incomplete causal knowledge could be used for causal intervention.
Judea Pearl presented in his work \cite{judea2000causality} a comprehensive theory
of causality and unified the probabilistic, manipulative, counterfactual, and structural approaches
to causation. 
%Judea also introduced precise mathematical definitions
%of causal analysis for the standard curricula of statistics.
From the work of \citet{judea2000causality} we have the following key point:
if there is a statistical association,
e.g. two variables $X,Y$ are dependent, then one of the following is true:
\begin{enumerate}
    \item there is a causal relationship, either $X$ affects $Y$ or $Y$ affects $X$;
    \item there is a common cause (\textit{confounder}) that affects both $X$ and $Y$;
    \item there is a possibly unobserved common effect of $X$ and $Y$ that is conditioned upon data acquisition (\textit{selection bias});
    \item there can be a combination of these.
\end{enumerate}
From there on a lot of research has been conducted to develop
theoretical approaches and methods for identifying causal relationships from 
observational data. 
%Before we present some major works from the last two decades,
%we introduce the common concept behind all these approaches in a short formal manner.

In general, all these methods exploit the complexity of the marginal and conditional probability
distributions in some way (e.g., \citet{janzing2012information, sgouritsa2015inference}), and under 
certain assumptions these methods are then able to solve the task of causal identification.
Let $C$ denote the cause and $E$ the effect. In a system with two or more variables
we might have cause-effect pairs and then their joint density can be expressed with $p_{C,E}(c,e)$.
This joint density can be factorized in either of the following ways: 
\begin{equation}
    \label{eq:factorizationCE}
    p_{C,E}(c,e) = p_C(c)\cdot P_{E|C}(e|c) \text{, or}
\end{equation}
\begin{equation}
    \label{eq:factorizationEC}
    p_{C,E}(c,e) = p_E(e)\cdot P_{C|E}(c|e).
\end{equation}
%1) $p_C(c)\cdot P_{E|C}(e|c)$ or 2) $p_E(e)\cdot P_{C|E}(c|e)$.
The idea is then that \cref{eq:factorizationCE} gives models of lower total complexity than \cref{eq:factorizationEC}, and this allows
us to draw conclusions about the causal relationship direction. Intuitively this makes sense, because
the effect contains information from the cause but not vice-versa (of course under the assumption
that there are no cycles aka feedback loops). Therefore, \cref{eq:factorizationEC} has at least as much complexity as \cref{eq:factorizationCE}.
%However, the definition of complexity is ambiguous. For example, one can say that
%"\textit{$p_C$ contains no information about $P_{E|C}(e|c)$}" and then is able
%to draw partial conclusions about the causal direction of the given system.
This unequal distribution of complexity is often colloquially referred to as "\textit{breaking
the symmetry}", that is $p_C(c)\cdot P_{E|C}(e|c) \neq p_E(e)\cdot P_{C|E}(c|e)$.

In recent years, numerous approaches were proposed for structure learning.
\citet{DBLP:journals/corr/abs-1301-3857} addressed the problem of learning
the structure of a Bayesian network in domains that contain continuous
variables. 
%In their work, they showed that in probabilistic networks with continuous variables
%one can use Gaussian Process priors to compute marginal likelihoods for structure learning.
\citet{kano2003causal} developed a model for causal inference
using non-normality of observed data and improved path analysis proposed by \citet{wright1921correlation}.
%using non-normal data. 
\citet{shimizu2006linear} proposed a method
to determine the complete causal graph of continuous data under three
assumptions: the data generating process is linear, no unobserved confounders,
and noise variables have non-Gaussian distributions of non-zero variances.
This method was not scale-invariant, but later work by \citet{shimizu2014direct}
addressed this problem.
%was addressed and a new method was proposed which was
%guaranteed to converge to the right solution within a small fixed number of steps
%if the data strictly followed the model. 
\citet{sun2006causal} introduced a method
based on comparing the conditional distributions of variables given their direct
causes for all hypothetical causal directions and choosing the most plausible one (Markov kernels).
%Those Markov kernels which maximize the conditional entropies constrained by their observed
%expectation, variance, and covariance with its direct causes based on their given domain
%are considered as plausible kernels. 
\citet{sun2008causal} continued the work on kernels
by using the concept of reproducing kernel Hilbert spaces.

A group of well-known and well-established methods is based on the \textit{Additive Noise Models} (ANMs), that yield many good results \cite{kpotufe2014consistency}.
In these models, the effect is a function
of the cause and some random and non-observed additive noise term.
These methods received a lot of attention from researchers in the past years.
\citet{hoyer2008nonlinear} generalized the linear framework of additive noise models
to nonlinear models. 
%\citet{janzing2009telling} proposed a method for inferring linear causal
%relationships among multi-dimensional variables by factorizing the joint
%distribution into products with marginal and conditional distributions (as seen %above with \cref{eq:factorizationCE,eq:factorizationEC}).
%Then, in one of these products the factors (e.g., $P(E)$ and $P(C|E)$) satisfy %non-generic
%relations indicating that $E \to C$ is wrong.
\citet{mooij2009regression} introduced a method that minimizes the statistical
dependence between the regressors and residuals. This method does not need to 
assume a particular distribution of the noise because any form of regression
can be used (e.g., Linear Regression) and is well suited for the task
of causal inference in additive noise models.
%\citet{stegle2010probabilistic} created a method to model observed data by using
%probabilistic latent (hidden) variable models, which incorporate the effects of
%unobserved noise. To analyze the joint density of cause and effect, 
%the effect is modeled as a function of the cause and some independent noise (not necessarily additive).
%With general non-parametric priors on this function and on the distribution of the cause
%the causal direction is then determined by using standard Bayesian model selection.
\citet{mooij2011causal} introduced a method to determine the causal relationship
in cyclic additive noise models and stated that such models are generally
identifiable in the bivariate, Gaussian-noise case. Their method works
for continuous data and can be seen as a special case of nonlinear
independent component analysis.
%\citet{zhang2012identifiability} tested the stability of the identifiability
%of post-nonlinear models with two variables and listed all cases in which this 
%model is not identifiable anymore. Furthermore, they showed how to approach
%multivariate cases with post-nonlinear models.
%\citet{daniusis2012inferring} showed that even in deterministic cases (noise-free
%cases) some asymmetries can be exploited for causal inference.
%Their method is based on the idea that if $X \to Y$ then the distribution of
%$X$ and the function mapping $X$ to $Y$ must be independent since they correspond
%to independent mechanisms of nature.
\citet{hyvarinen2013pairwise} proposed a method which is based
on the likelihood ratio under the linear non-Gaussian acyclic
model known as LiNGAM \cite{shimizu2014lingam}. This method does not resort to
independent component analysis algorithm as previous methods did.

As indicated in the name, ANMs are heavily based on the presence of noise. 
However, despite all the research in the past years one small but nonetheless important
aspect of causal discovery methods with ANMs has not received much attention: \textit{can
different noise levels have an impact on the correctness of these methods?}
In the real world, observational data often differ in terms of the noise level. Usually,
these levels do not differ significantly but it can occur that noise levels
change drastically from cause to effect. For example, if the data collection
process has a lot of interference (e.g., in outer space) then the related noise levels
can differ a lot.
In this work, we aim to bridge this research gap. 
We perform an empirical study with a well-established method of the ANMs group
\textit{Regression with Subsequent Independence Test (RESIT)} \cite{peters2014causal}. This method yields good results and can be used even when variables have different distribution types. In our experimental evaluation, we aim to quantify the impact of different noise levels on the performance of RESIT.

The rest of the paper is organized as follows. In \cref{sec:model} we describe RESIT and discuss its functioning. \cref{sec:experimental-setup} presents experimental setup followed by results analysis in \cref{sec:experimental-results}. Finally, we conclude our work and summarize our findings in \cref{sec:conclusions}.

%% file: sections/03-model.tex
\section{Model}
\label{sec:model}

\subsection{RESIT}
\label{sec:model:resit}

%\textit{We are assuming that there is no confounding, no selection bias, and no feedback loop between $X$ and $Y$ and therefore $X$ and $U$ are independent, which is denoted as $X \indep U$.}

The RESIT method is based on the fact that for each
node $X_i$ the corresponding noise variable $N_i$ is independent of all non-descendants of $X_i$.
For example, if we have $Y = X_1 + N_1$ then $X_1 \indep N_1$.
RESIT works in both bivariate and multivariate cases, see \citet{peters2014causal}. We restrict our experiments to bivariate cases to reduce runtimes.
In our experiments, we have two variables, $X$ 
and $Y$, and the task is to determine whether $X$ causes $Y$ ($X \to Y$) or $Y$ causes $X$ ($Y \to
X$).

We apply the same algorithm as \cref{algo:resit} from \citet{mooij2016distinguishing} which requires inputs
$X$ and $Y$, a regression method, and a score estimator $\hat{C}: \mathbb{R}^N \times \mathbb{R}^N \to \mathbb{R}$.
The algorithm outputs $dir$ (casual relationship \textbf{dir}ection).
First, the data is split into training data 
%(80\%) 
and test data.
%(20\%). 
\citet{kpotufe2014consistency}
refers to this as \textit{decoupled estimation}\footnote{As opposed to \textit{decoupled estimation}, in
    \textit{coupled estimation} the data is not split into training and test data, see \citet{kpotufe2014consistency,mooij2016distinguishing}.
    }.
The training data is used to fit the regression model
and the test data is used to calculate the value of the estimator. The idea is to regress $Y$ on $X$ with the training data, 
predict $\hat{Y}$ with the
test data and then calculate residuals $Y_{res} = \hat{Y} - Y_{Test}$. $Y_{res}$ and
$X_{Test}$ are then used to calculate the score for the assumed case $X \to Y$: $\hat{C}_{X \to Y}$. Similarly, to test
the other case ($Y \to X$), we regress $X$ on $Y$, 
calculate residuals $X_{res} = \hat{X} - X_{Test}$ and estimate $\hat{C}_{Y \to X}$.
%In our test scenario our generated data always follows $X \to Y$. 
If only one direction in our data is correct (and not both), we can compare estimates directly. Otherwise, we need
to determine the value of $\alpha$ for the independence tests. 
%\cref{algo:resit} shows the pseudo-code of the procedure explained above.

\begin{algorithm}[h]
    \caption{General procedure to decide whether $p(x,y)$ satisfies Additive
    Noise Model $X \to Y$ or $Y \to X$ with decoupled estimation.}\label{algo:resit}
    \begin{algorithmic}[1]
        \State \textbf{Input:}
        \State \hspace{2.5mm} 1) I.i.d. sample data $X$ and $Y$
        \State \hspace{2.5mm} 2) Regression method
        \State \hspace{2.5mm} 3) Score estimator $\hat{C}: \mathbb{R}^N \times \mathbb{R}^N \to \mathbb{R}$
        \State \textbf{Output:}
        \State \hspace{2.5mm} \textit{dir}
        \State
        \State \textbf{Procedure}
        \State 1) Split data into training and test data:
        \State \hspace{2.5mm} $X_{Train}, X_{Test} \gets X$
        \State \hspace{2.5mm} $Y_{Train}, Y_{Test} \gets Y$
        \State
        \State 2) Train regression models
        \State \hspace{2.5mm} $reg_1 \gets$ Regress $Y_{Train} \text{ on } X_{Train}$
        \State \hspace{2.5mm} $reg_2 \gets$ Regress $X_{Train} \text{ on } Y_{Train}$
        \State
        \State 3) Calculate Residuals:
        \State \hspace{2.5mm} $Y_{res} = reg_1.predict(X_{Test}) - Y_{Test}$
        \State \hspace{2.5mm} $X_{res} = reg_2.predict(Y_{Test}) - X_{Test}$
        \State
        \State 4) Calculate Scores:
        \State \hspace{2.5mm} $\hat{C}_{X \to Y} = \hat{C}(X_{Test}, Y_{res})$
        \State \hspace{2.5mm} $\hat{C}_{Y \to X} = \hat{C}(Y_{Test}, X_{res})$
        \State
        \State 5) Output direction \textit{dir}:
            \begin{gather*}
                dir = \begin{cases}
	            X \to Y & \text{if } \hat{C}_{X \to Y} < \hat{C}_{Y \to X},\\
	            Y \to X & \text{if } \hat{C}_{X \to Y} > \hat{C}_{Y \to X},\\
	            ? & \text{if } \hat{C}_{X \to Y} = \hat{C}_{Y \to X}.
	            \end{cases}
            %\label{eqn:dir}
            \end{gather*}

    \end{algorithmic}
\end{algorithm}

\subsection{Estimators} 
\label{sec:model:estimators}

Both \textit{independence tests} and \textit{entropy measures} can be used to calculate the scores $\hat{C}_{X \to Y}$ and $\hat{C}_{Y \to X}$. 
In general, for the independence tests we have:
$$\hat{C}(X_{Test},Y_{res}) = I(X_{Test},Y_{res})$$ with $I(\cdot,\cdot)$ being any independence test.
In the case of entropy estimators, we have:
$$\hat{C}(X_{Test},Y_{res}) = H(X_{Test}) + H(Y_{res}),$$ with $H(\cdot)$ being any entropy measure. The estimator
score for entropy is derived from Lemma 1 from \citet{kpotufe2014consistency}.

The following 6 independence tests and 6 entropy measures were used as estimators in this work. The implementation of all estimators except \textbf{HSIC} was taken from
the \textit{information theoretical estimators} toolbox \cite{szabo14information}:\footnote{See the documentation
of the toolbox for more details.}

\begin{enumerate}
  \item \textbf{HSIC}: Hilbert-Schmidt Independence Criterion with RBF Kernel\footnote{Source: \url{https://github.com/amber0309/HSIC}}:\\
                       $$I_{HSIC}(x,y) := ||C_{xy}||^2_{HS},$$ where $C_{xy}$ is the cross-covariance
                       operator and $HS$ the squared Hilbert-Schmidt norm.
  \item \textbf{HSIC\_IC}: Hilbert-Schmidt Independence Criterion using incomplete
	                   Cholesky decomposition (low rank decomposition of the Gram matrices, 
	                   which permits an accurate approximation
	                   to HSIC as long as the kernel has a fast decaying spectrum) with $\eta = 1*10^{-6}$
	                   precision in the incomplete cholesky decomposition.
  \item \textbf{HSIC\_IC2}: Same as HSIC\_IC but with $\eta = 1*10^{-2}$.
  \item \textbf{DISTCOV}: Distance covariance estimator using pairwise distances. This is simply
  the $L^2_w$ norm of the characteristic functions $\varphi_{12}$ and $\varphi_1 \varphi_2$ of input $x,y$:
  $$\varphi_{12}(\boldsymbol{u}^1,\boldsymbol{u}^2) = \mathbb{E}[e^{i\langle \boldsymbol{u}^1, \boldsymbol{x} \rangle +
  i\langle \boldsymbol{u}^2, \boldsymbol{y} \rangle}],$$
  $$\varphi_1(\boldsymbol{u}^1) = \mathbb{E}[e^{i\langle \boldsymbol{u}^1, \boldsymbol{x} \rangle}],$$
  $$\varphi_2(\boldsymbol{u}^2) = \mathbb{E}[e^{i\langle \boldsymbol{u}^2, \boldsymbol{y} \rangle}].$$
  With $i = \sqrt{-1}$, $\langle \cdot, \cdot \rangle$ - the standard Euclidean inner product, and $\mathbb{E}$ - the 
  expectation. Finally, we have:
  $$I_{dCov}(x,y) = ||\varphi_{12} - \varphi_1 \varphi_2||_{L^2_w}$$
  
  \item \textbf{DISTCORR}: Distance correlation estimator using pairwise distances. It is simply the standardized
                        version of the distance covariance:
  $$I_{dCor}(x,y) = 
              \frac{I_{dCov}(x,y)}{\sqrt{I_{dVar}(x,x)I_{dVar}(y,y)}}$$ with
  $I_{dVar}(x,x) = ||\varphi_{11} - \varphi_1 \varphi_1||_{L^2_w},\: I_{dVar}(y,y) = ||\varphi_{22} - \varphi_2
  \varphi_2||_{L^2_w}$
  (see characteristic functions under DISTCOV). If $I_{dVar}(x,x)I_{dVar}(y,y) \leq 0$, then $I_{dCor}(x,y) = 0$.

  \item \textbf{HOEFFDING}: Hoeffding's Phi:
  $$I_{\Phi}(x,y) = I_{\Phi}(C) = \left(h_2(d) \int_{[0,1]^d} [C(\boldsymbol{u}) - \Pi (\boldsymbol{u})]^2d\boldsymbol{u}\right)^{\frac{1}{2}}$$ with
  $C$ standing for the copula of the input and $\Pi$ standing for the product copula.
  
  \item \textbf{SH\_KNN}: Shannon differential entropy estimator using kNNs ($k$-nearest neighbors)
  $$H(\boldsymbol{Y}_{1:T}) = log(T-1) - \psi(k) + log(V_d) + \frac{d}{T}\sum^T_{t=1}log(\rho_k(t))$$
  with $T$ standing for the number of samples, $\rho_k(t)$ -  the Euclidean distance of the $k^{th}$ nearest neighbour of $\boldsymbol{y}_t$
  in the sample $\boldsymbol{Y}_{1:T}\backslash\{\boldsymbol{y}_t\}$, and $V \subseteq \mathbb{R}^d$ - a finite set.
  
  \item \textbf{SH\_KNN\_2}: Shannon differential entropy estimator using kNNs
	                     with $k=3$ and $kd$-tree for quick nearest-neighbour lookup.
	                     
  \item \textbf{SH\_KNN\_3}: Shannon differential entropy estimator using kNNs with $k=5$.
  
  \item \textbf{SH\_MAXENT1}: Maximum entropy distribution-based Shannon entropy estimator:
  $H(\boldsymbol{Y}_{1:T}) = H(n) - \left[k_1 \left(\frac{1}{T}\sum^T_{t=1}G_1(y'_t)\right)^2
  + k_2 \left(\frac{1}{T}\sum^T_{t=1}G_2(y'_t)-\sqrt{\frac{2}{\pi}}\right)^2\right] + log(\hat{\sigma}),$
  with
  $\hat{\sigma} = \hat{\sigma}(\boldsymbol{Y}_{1:T}) = \sqrt{\frac{1}{T-1}\sum^T_{t=1}(y_t)^2},$
  $y'_t = \frac{y_t}{\hat{\sigma}}, (t= 1, \dots, T)$, 
  $G_1(z) = ze^{\frac{-z^2}{2}},$
  $G_2(z) = |z|,$
  $k_1 = \frac{36}{8\sqrt{3}-9},$
  $k_2 = \frac{1}{2-\frac{6}{\pi}}$.

  \item \textbf{SH\_MAXENT2}: Same as SH\_MAXENT1 with
  the following changes:
  $$G_2(z) = e^{\frac{-z^2}{2}},
  k_2 = \frac{24}{16\sqrt{3}-27}.$$
  
  \item \textbf{SH\_SPACING\_V}: Shannon entropy estimator using Vasicek's spacing method:
  $$H(\boldsymbol{Y}_{1:T}) = \frac{1}{T}\sum^T_{t=1}log\left(\frac{T}{2m}[y_{(t+m)}-y_{(t-m)}]\right),$$
  with $T$ standing for the number of samples. The convention that $y_{(t)} = y_{(1)}$ if $t < 1$ and $y_{(t)} = y_{(T)}$ if
  $t > T$ and $m = \lfloor \sqrt{T} \rfloor$.
\end{enumerate}

%% file: sections/04-experimental-setup.tex
\section{Experimental Setup}
\label{sec:experimental-setup}

For all experiments, we generate artificial data using linear and non-linear functions.
While both linear and non-linear data can be identifiable in causal models,
non-linearity helps in identifying the causal direction as was shown by \citet{hoyer2008nonlinear}.
In all
experiments we use the equation $Y = X + N_Y$ for the linear cases 
and $Y = X^3 + N_Y$ for the non-linear cases. These two structural causal models
have been selected arbitrarily for simplicity. For the consistency of the identifiability
of linear and non-linear data in additive noise models, the reader is referred to
\citet{kpotufe2014consistency,shimizu2006linear,hoyer2008nonlinear,zhang2012identifiability}.
80\% of the generated data is used for training a regression model, and the rest 20\% is used to calculate the values of estimators $\hat{C}_{X \to Y}$ and $\hat{C}_{Y \to X}$.

In all our tests, we assume $X$ to be a cause of $Y$, that is  $X \to Y$.
$X$ and $N_Y$ can be
drawn from one of the following distributions: the normal distribution denoted by $\mathcal{N}$, the uniform distribution denoted by $\mathcal{U}$, or the laplace distribution denoted by $\mathcal{L}$. The parameters of the distributions for $X$ and $N_Y$ are defined by the equations below:
$$    X \sim \begin{cases}
	    \mathcal{N}(0, 1)& \text{or} 
	    \\
	    \mathcal{U}(-1, 1)& \text{or}\\
	    \mathcal{L}(0, 1)
	 \end{cases}
$$
$$N_Y \sim \begin{cases}
            \mathcal{N}(0, 1 \cdot i)& \text{or} \\
            \mathcal{U}(-1 \cdot i, 1 \cdot i)& \text{or}\\
            \mathcal{L}(0, 1 \cdot i)
         \end{cases}    
$$
with $i$ being a scaling factor for the noise level in $N_Y$, $i$-factor for short.
By varying the value of $i$, we can analyze how different values of standard deviations (boundaries for the uniform case)
in the noise term $N_Y$ relative to the standard deviation (or boundaries for the uniform case) in the $X$ term impact the accuracy of RESIT method.
         
In our experiments, we consider 199 different $i$-factors:
$$i \in \{0.01, 0.02, \dots, 1.00\} \cup \{1, 2, \dots, 100\}.$$ The values $i < 1$ correspond to the cases when deviation of $N_Y$ is less than that of $X$, and the values $i > 1$ correspond to the cases when the deviation of $N_Y$ is larger than that of $X$. The deviation of noise ranges from 1\% (for $i=0.01$) to 10000\% (for $i=100$) of the deviation of $X$.
For each value of $i$, we have 18 different combination of models:
two general structures $Y = X + N_Y$ and $Y = X^3 + N_Y$ where $X \text{ and } N_Y$
are drawn from one of the three different distributions: $\mathcal{N}, \mathcal{U} \text{ or } \mathcal{L}$. To represent the models, we use the notations like $Y = {L}^3 + \mathcal{U}$, that signifies a nonlinear model with $X \sim {L}$ and $N_Y \sim \mathcal{U}$.
For each of the 18 combinations, for a single test, we generate 1000 samples from the relative distributions. Next, we perform causal identification according to the procedure described in \cref{sec:model:resit} using one of the estimators presented in \cref{sec:model:estimators}. These tests are repeated 100 times. Finally, we calculate the fraction of successful tests for each combination of a model and an estimator, and define this
ratio as our accuracy measure.

For the regression, we used Linear Regression with an appropriate coordinates transformation for the non-linear cases.

%% file: sections/05-experimental-results.tex
\section{Experimental Results}
\label{sec:experimental-results}

\cref{fig:acc-lin,fig:acc-Nlin} show the results for different estimators obtained for liner and nonlinear models respectively.
In these figures, the y-axis shows the accuracy ($\frac{\text{\#successful tests}}{100}$) for different estimators, and 
the x-axis shows the range of the $i$-factor. The results for independence estimators are presented with solid lines and the results for entropy estimators are shown with dashed lines.
The values of the estimators close to $0.5$ indicate that in 50\% of the 
tests the algorithm chose the correct direction and vice versa 50\% chose the wrong one.
Such cases are \textbf{unidentifiable}. The values of accuracy closer to $1$ mean
very good or consistent \textbf{identifiability}. 
Also, the plots for DISTCOV (dark green) and DISTCORR (medium purple) 
often overlap (more than in 90\% of cases), resulting in a dark purple line. 

Additionally, in \cref{tab:summary:linear} and \cref{tab:summary:nonlinear} we summarize our experimental result.
The values in the cells show on what range of $i$-factor the
estimators \textit{can} reach over 90\%. Estimators have some variance in the results and thus on some intervals, they fall below 90\% accuracy. The limits in the cells were chosen as follows: the lower limit shows
where estimators reach the first time 90\% or higher, and the upper limit shows the last time where it
reaches 90\% or higher. In between, most of the time estimators remain above 90\% or rarely fall below, but not
more than 10\% of the cases.
An empty cell in the tables means that for the relevant model and estimator the accuracy never reached 90\%.
An open range from one side, for 
example, "-- 5" or "5 --", or from both sides, such as " -- ", indicates an unbounded interval with one or two missing 
bounds.

\subsection{Linear Models: $Y = X + N_Y$}
\label{sec:experimental-results:linear}

\input{sections/tex-plots-tables/fig_lin}

\input{sections/tex-plots-tables/tab_lin}

We start with the analysis of the results for the linear models presented in \cref{fig:acc-lin,tab:summary:linear}. 
First, we consider the models with the independent variable distributed normally, $X \sim \mathcal{N}$.
\cref{fig:acc-lin:NN} shows the only case where we never achieve identifiability. This is the well-known
linear Gaussian structural causal model $Y=N+N$. Only recently it has been tackled successfully by 
\citet{chen2019causal,park2019identifiability}.
However, we do not consider their approach in this work.
\cref{fig:acc-lin:NU} shows the linear model with $Y= \mathcal{N} + \mathcal{U}$.
SH\_SPACING\_V performs the best with the accuracy of 100\% for $i \in [0.55; 7]$. 
HSIC\_IC and HSIC\_IC2 perform the worst here. The associated accuracy reaches 90\% only for $i \in [3;7]$. 
All other estimators perform mediocre with an accuracy above $80\%$ for $i \in [0.5; 7]$.
\cref{fig:acc-lin:NL} shows the linear model $Y= \mathcal{N} + \mathcal{L}$. The best estimators are SH\_MAXENT1 and SH\_MAXENT2
with accuracy around 90\% for $i \in [0.30; 3]$. HSIC also performs good with accuracy over 90\% for $i \in [0.38; 2]$.
The worst estimators are the three Shannon differential
entropy estimators using kNNs which never remain consistently above 80\% accuracy. The remaining estimators
lie within the range 90\% $\pm$ 8\% accuracy for $i \in [0.3; 3]$.

Next, we consider models with $X \sim U$.
\cref{fig:acc-lin:UN} shows the linear model $Y= \mathcal{U} + \mathcal{N}$. Here all estimators differ stronger than in the previous
cases. First, SH\_SPACING\_V performs the best with 100\% accuracy for $i \in [0.08; 2]$.
With $i = 1$ all other estimators remain above 90\%, expect HOEFFDING ($\sim$88\%) and HSIC\_IC and HSIC\_IC2 (both $\sim 75\%$). 
After $i$ becomes larger than 2, all estimators drop drastically towards 50\% accuracy except SH\_SPACING\_V which remains above 70\%.
For $i \in [0.2; 1]$ some estimators remain between 80\% and 95\% while HSIC is above 95\%. HSIC\_IC and HSIC\_IC2 perform worse
than all other estimators.
\cref{fig:acc-lin:UU}, shows the linear model $Y= \mathcal{U} + \mathcal{U}$.
In this case, HSIC\_IC and HSIC\_IC2 reach
accuracy above 90\% only around $i = 3$, see the 3d column in \cref{tab:summary:linear}. All other estimator perform quite good with $i \in [0.2; 3]$, however HOEFFDING and DISTCOV
drop slightly below 90\% accuracy for $i = 1$. SH\_SPACING\_V has 100\% accuracy for $i \in [0.12;10]$ and has
on the remaining values of $i$ better accuracy than all other estimators. In general, we can observe that identifiability is much better for $i < 1$, that is when the range for noise term is less than the range of $X$. For $i > 50$ the accuracy of most of the estimators is around 50\%, indicating that the predicted direction is wrong in half of the tests.
\cref{fig:acc-lin:UL} shows the model $Y= \mathcal{U} + \mathcal{L}$. For $i \in [0.3; 1]$ all estimators
perform well with 90\% or higher accuracy, except HSIC\_IC and HSIC\_IC2 which remain above 90\% accuracy only after $i= 0.45$. 
After $i = 1$ each estimator drops drastically and all converge towards 50\%
accuracy. The only exception is SH\_SPACING\_V which remains with a mean of 70\% accuracy longer than other estimators.
For $i \in [0.08; 1]$ SH\_SPACING\_V also has accuracy 100\%. For $i < 0.3$ all other
estimators drop fast towards 50\%.

Now we proceed to the analysis of the remaining cases where the independent variable $X$ is
distributed according to the Laplace distribution, $X \sim \mathcal{L}$.
\cref{fig:acc-lin:LN} shows the model $Y= \mathcal{L} + \mathcal{N}$.
For $i \in [0.3; 1]$ HSIC, SH\_MAXENT1 and SH\_MAXENT2 have accuracy greater than 90\%.
SH\_SPACING\_V, HSIC\_IC, HSIC\_IC2, DISTCOV and DISTCORR lie between 85\% and 95\% accuracy
for $i \in [0.4; 1]$ and HOEFFDING remains between 80\% and 90\%.
Again, the three Shannon kNN estimators never reach an accuracy higher than 80\%.
After $i$ reaches the value of 1, all estimators drop fairly fast towards unidentifiability.
\cref{fig:acc-lin:LU} shows the linear model $Y= \mathcal{L} + \mathcal{U}$. SH\_SPACING\_V performs the best of all
estimators and has an accuracy of 100\% for $i \in [0.5; 5]$. All other estimators slowly climb
towards good identifiability and for $i \in [0.7; 7]$ they remain above 90\% accuracy. Afterward,
all other estimators drop with a similar pace towards unidentifiability.
Fianlly, \cref{fig:acc-lin:LL} shows the linear model $Y= \mathcal{L} + \mathcal{L}$. Here we can observe the following. For $i \in [0.4;2]$ SH\_MAXENT1 and
SH\_MAXENT2 have accuracy close to 100\%. Next, for $i \in [0.4;1]$ HSIC, HSIC\_IC, HSIC\_IC2, SH\_SPACING\_V, DISTCOV and DISTCORR
remain above 90\% accuracy. HOEFFDING, and the three Shannon kNN estimators never reach an accuracy above 90\%.
After $i = 1$ all estimators drop fast towards 50\%.

\subsection{Nonlinear Models: $Y = X^3 + N_Y$}
\label{sec:experimental-results:nonlinear}

\input{sections/tex-plots-tables/fig_nonlin}

\input{sections/tex-plots-tables/tab_nonlin}

The results for nonlinear models are grouped in the same way as for linear models and are presented in \cref{fig:acc-Nlin,tab:summary:nonlinear}. 
In general, we can notice much better identifiability in the nonlinear case. 

Similar to the linear case, We start with the analysis of the modes with $X \sim \mathcal{N}$.
\cref{fig:acc-Nlin:NN} shows the nonlinear model $Y= \mathcal{N}^3 + \mathcal{N}$. Here all estimators perform very good
with $i \in [0.4; 25]$ having an accuracy of almost 100\%. With $i < 0.3$ most estimators drop fast below 90\% 
accuracy.
With $i \in [20;100]$ all estimators remain above 90\% accuracy, except for HSIC\_IC, HSIC\_IC2,
SH\_MAXENT1 and SH\_MAXENT2 which drop below 90\% after $i = 45$. DISTCOV, SH\_SPACING\_V and the three
Shannon kNN estimators remain close to 100\% in $i \in [0.01; 100]$.
\cref{fig:acc-Nlin:NU} shows the model $Y= \mathcal{N}^3 + \mathcal{U}$. In this case, for $i \in [0.45; 80]$ we have 90\% or higher accuracy for all estimators. DISTCOV, SH\_SPACING\_V and the three
Shannon kNN estimators remain close to 100\% in $i \in [0.01; 100]$.
\cref{fig:acc-Nlin:NL} shows the nonlinear model $Y= \mathcal{N}^3 + \mathcal{L}$. For this model, all estimators perform very good
with $i \in [0.4; 30]$ having an accuracy close to 100\%.
With $i < 0.25$ estimator drop rapidly and for $i > 30$ HSIC\_IC, HSIC\_IC2, SH\_MAXENT1 and SH\_MAXENT2 drop
below 90\% accuracy.
All others remain over 90\% accuracy while HSIC remains around 90\%.

Now we proceed to the analysis of the models with $X \sim \mathcal{U}$.
\cref{fig:acc-Nlin:UN} shows the model $Y= \mathcal{U}^3 + \mathcal{N}$. All estimators, except
HSIC\_IC and HSIC\_IC2, remain
above 95\% for $i \in [0.05; 1]$. For $i \in [1; 100]$ the estimators converge differently. All three
Shannon differential entropy measures with kNNs and SH\_SPACING\_V remain above 95\% accuracy.
DISTCOV, DISCORR and HOEFFDING keep a mean of $\sim$85\% accuracy. HSIC and SH\_MAXENT1 remain above 60\% accuracy. SH\_MAXENT2 is pretty much unidentifiable. Finally,  HSIC\_IC and HSIC\_IC2
are unidentifiable for all values of $i$.
\cref{fig:acc-Nlin:UU} shows the nonlinear model $Y= \mathcal{U}^3 + \mathcal{U}$. For $i \in [0.09; 1]$ all
estimators except HSIC\_IC and HSIC\_IC2 remain above 95\% accuracy, while SH\_KNN, SH\_KNN\_2, and SH\_SPACING\_V
continue to do so for $i \in [1; 100]$. DISTCOV, DISCORR and HOEFFDING remain between
80\% and 90\%. HSIC and SH\_MAXENT1 drop to $\approx 60$\% after $i = 20$ and remain above 60\%
for $i < 100$. SH\_MAXENT2, HSIC\_IC and HSIC\_IC2 drop to 50\% for $i \in [20; 100]$.
\cref{fig:acc-Nlin:UL} shows the model $Y= \mathcal{U}^3 + \mathcal{L}$. The behaviour of different estimators is almost the same
as for $Y= \mathcal{U}^3 + \mathcal{N}$. The only differences are that HSIC\_IC performs slightly better
for $i \in [0.2;1]$ and DISTCOV, DISTCORR, HSIC and SH\_MAXENT2 perform worse.

Lastly, we analyze the 3 remaining models with $X \sim L$.
\cref{fig:acc-Nlin:LN} shows the model $Y= \mathcal{L}^3 + \mathcal{N}$.
For $i \in [0.1;100]$ all estimators (except SH\_MAXENT1 and SH\_MAXENT2) have an accuracy of 90\% or higher,
SH\_SPACING\_V and the three Shannon kNN estimators have an accuracy of 100\% for all values of $i$-factor.
Only SH\_MAXENT1 and SH\_MAXENT2 perform badly at the beginning but still have an accuracy of 90\% or higher
for $i \in [0.35;100]$.
\cref{fig:acc-Nlin:LU} shows the nonlinear model $Y= \mathcal{L}^3 + \mathcal{U}$. It is very similar to the previous case.
For $i \in [0.15;100]$ all estimators (except SH\_MAXENT1 and SH\_MAXENT2) have an accuracy of 90\% or higher, and
SH\_SPACING\_V, and the three Shannon kNN estimators have an accuracy of 100\% for all values of $i$.
As in the previous case, SH\_MAXENT1 and SH\_MAXENT2 perform badly at the beginning but still have an accuracy of 90\% or higher
for $i \in [0.7;100]$. For $i \geq 1$ all estimators are very close to 100\% accuracy.
Finally, \cref{fig:acc-Nlin:LL} shows the nonlinear model $Y= \mathcal{L}^3 + \mathcal{L}$. This model allows the  best identifiability of all. For $i \geq 0.1$ all estimators except SH\_MAXENT1 and SH\_MAXENT2 have an accuracy 90\% or higher.
SH\_SPACING\_V and the three Shannon kNN estimators have an accuracy of 100\% for all values of $i$.
Only SH\_MAXENT1 and SH\_MAXENT2 perform badly at the beginning but still have an accuracy of 90\% or higher
for $i \geq 0.35$.

\subsection{Summary}
\label{sec:experimental-results:summary}
As the results show, different noise levels do have an impact on the identifiability performance 
in RESIT methods.
In general, the linear equation models are more fragile in RESIT than the nonlinear equation models
because nonlinear relationships tend to break the symmetry between the variables easier
\citep{hoyer2008nonlinear}. Furthermore, in all cases the test results themselves
have a standard deviation between 0.05 to 0.1 as one can see in the sharp wiggles in the
plots.

We can notice some similarities in the models depending on how distributed their components $X$ and $N_Y$. It is visually visible in a matrix of plots in \cref{fig:acc-lin,fig:acc-Nlin}. The plots on the main diagonals, \cref{fig:acc-lin:NN,fig:acc-lin:UU,fig:acc-lin:LL,fig:acc-Nlin:NN,fig:acc-Nlin:UU,fig:acc-Nlin:LL}, represent the models for which both $X$ and $N_Y$ are drawn from the same type of distribution. 
In the case of nonlinear models, the diagonal plots demonstrate 3 distinct behaviors of estimators, as presented in 
\cref{fig:acc-Nlin:NN,fig:acc-Nlin:UU,fig:acc-Nlin:LL}. We can also clearly see the 
similarity of plots in the same rows. It means that the models with the same type of 
distribution for the independent variable $X$ have common characteristics. We observe that 
all models with $X \sim \mathcal{L}$, see \cref{fig:acc-Nlin:LN,fig:acc-Nlin:LU,fig:acc-Nlin:LL} allow very good and consistent identifiability by all estimators for $i \geq 1$. For the values $i < 1$, many estimators, 
except SH\_MAXENT1 and SH\_MAXENT2, also perform well with accuracy $\approx 90\%$. 
The group of models with $X \sim \mathcal{N}$ allow all estimators achieve almost perfect 
identifiability for $0.8 < i < 20$. The accuracy then reduces for larger and smaller 
values of $i$. Finally, the group of models with $X \sim N$ allow the worst 
identifiability, see \cref{fig:acc-Nlin:UN,fig:acc-Nlin:UU,fig:acc-Nlin:UL}. For $i > 20$ several estimators have accuracy of $50\% - 60\%$. However, SH\_KNN estimators allow consistent identifiability for all values of $i$ even in this case, see 
\cref{tab:summary:nonlinear}. Similar but much less prominent row-wise similarity can 
be observed for linear models as well, see \cref{fig:acc-lin}. This indicates that the type of distribution of the independent variable $X$ impacts the accuracy of different estimators.

Looking now only at the best estimation function and assuming a strong identifiability of $\geq 90\%$ accuracy, we can observe that for linear models and $i \notin [0.5; 5]$ the accuracy is usually below 90\%.
This looks different for the nonlinear cases. Such estimators as SH\_KNN, SH\_KNN\_2, SH\_KNN\_3, and SH\_SPACING\_V allow consistent identifiability for all nonlinear modes. At the same time, the models with  $X \sim \mathcal{L}$ are identifiable with accuracy $\geq 90\%$ by all estimators on almost all range of values of $i$, see \cref{tab:summary:nonlinear}.

Some estimators perform differently depending on the setup. For example, for all nonlinear cases, the three 
Shannon differential entropy estimators with kNNs always perform above 90\% accuracy for all values of $i$, see \cref{tab:summary:nonlinear}. The associated
accuracy even reaches 100\% for all $i$ in the case of nonlinear models with $X \sim \mathcal{L}$, see \cref{fig:acc-Nlin:LN,fig:acc-Nlin:LU,fig:acc-Nlin:LL}. In case of linear models, these estimatros perform relatively poor, sometimes never reaching $90\%$ accuracy, see linear models with $X \sim N$, $Y=\mathcal{L} + \mathcal{N}$, and $Y=\mathcal{L} + \mathcal{L}$ in \cref{tab:summary:linear}.

Overall, SH\_SPACING\_V performs the best in almost all
cases, and is only outperformed by SH\_MAXENT1 and SH\_MAXENT2
for the following three linear models:
$Y=\mathcal{N} + \mathcal{L}$, $Y=\mathcal{L} + \mathcal{L}$, and $Y=\mathcal{L} + \mathcal{N}$, see \cref{fig:acc-lin:NL,fig:acc-lin:LL,fig:acc-lin:LN,tab:summary:linear}.
Some independence tests lose some
of the accuracy while entropy estimators retain accuracy over 90\%. This is observed for nonlinear models with $X \sim \mathcal{U}$, see \cref{fig:acc-Nlin:UN,fig:acc-Nlin:UU,fig:acc-Nlin:UL,tab:summary:nonlinear}.
Additionally, it is worth mentioning that entropy estimators are less computationally demanding than 
independence tests but can be quite sensitive to discretization effects \cite{mooij2016distinguishing}.
However, entropy estimators can only be used with the prior assumption we made:
\textit{there is only one causal direction and it is present in the model}.

%% file: sections/tex-plots-tables/fig_lin.tex
\begin{figure*}[]
\centering
\begin{subfigure}{.333\linewidth}
  \centering
  \input{plots/decoupled2/GAU}
  \vspace{-.5cm}
  \caption{$\mathcal{N}+\mathcal{N}$}
  \vspace{.5cm}
  \label{fig:acc-lin:NN}
\end{subfigure}%
\begin{subfigure}{.333\textwidth}
  \centering
  \input{plots/decoupled2/GAUxUNI}
  \vspace{-.5cm}
  \caption{$\mathcal{N}+\mathcal{U}$}
  \vspace{.5cm}
  \label{fig:acc-lin:NU}
\end{subfigure}%
\begin{subfigure}{.333\textwidth}
  \centering
  \input{plots/decoupled2/GAUxLAP}
  \vspace{-.5cm}
  \caption{$\mathcal{N}+\mathcal{L}$}
  \vspace{.5cm}
  \label{fig:acc-lin:NL}
\end{subfigure}
\begin{subfigure}{.333\textwidth}
  \centering
  \input{plots/decoupled2/UNIxGAU}
  \vspace{-.5cm}
  \caption{$\mathcal{U}+\mathcal{N}$}
  \vspace{.5cm}
  \label{fig:acc-lin:UN}
\end{subfigure}%
\begin{subfigure}{.333\textwidth}
  \centering
  \input{plots/decoupled2/UNI}
  \vspace{-.5cm}
  \caption{$\mathcal{U}+\mathcal{U}$}
  \vspace{.5cm}
  \label{fig:acc-lin:UU}
\end{subfigure}%
\begin{subfigure}{.333\textwidth}
  \centering
  \input{plots/decoupled2/UNIxLAP}
  \vspace{-.5cm}
  \caption{$\mathcal{U}+\mathcal{L}$}
  \vspace{.5cm}
  \label{fig:acc-lin:UL}
\end{subfigure}
\begin{subfigure}{.33\textwidth}
  \centering
  \input{plots/decoupled2/LAPxGAU}
  \vspace{-.5cm}
  \caption{$\mathcal{L}+\mathcal{N}$}
  \vspace{.5cm}
  \label{fig:acc-lin:LN}
\end{subfigure}%
\begin{subfigure}{.33\textwidth}
  \centering
  \input{plots/decoupled2/LAPxUNI}
  \vspace{-.5cm}
  \caption{$\mathcal{L}+\mathcal{U}$}
  \vspace{.5cm}
  \label{fig:acc-lin:LU}
\end{subfigure}
\begin{subfigure}{.33\textwidth}
  \centering
  \input{plots/decoupled2/LAP}
  \vspace{-.5cm}
  \caption{$\mathcal{L}+\mathcal{L}$}
  \vspace{.5cm}
  \label{fig:acc-lin:LL}
\end{subfigure}
\begin{subfigure}{.8\textwidth}
  \centering
  \includegraphics[scale=.6]{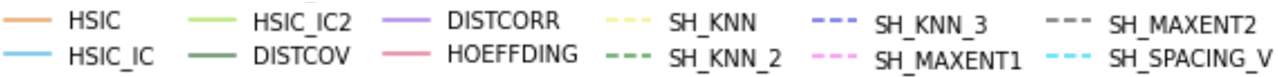}
\end{subfigure}
\caption{Accuracy of RESIT for linear models as a function of $i$-factor.}
\label{fig:acc-lin}
\vspace{5mm}
\end{figure*}

%% file: plots/decoupled2/GAU.tex
% This file was created by tikzplotlib v0.9.6.
\begin{tikzpicture}

\definecolor{color0}{rgb}{0.866666666666667,0.494117647058824,0.164705882352941}
\definecolor{color1}{rgb}{0.164705882352941,0.643137254901961,0.866666666666667}
\definecolor{color2}{rgb}{0.584313725490196,0.866666666666667,0.164705882352941}
\definecolor{color3}{rgb}{0.109803921568627,0.337254901960784,0.129411764705882}
\definecolor{color4}{rgb}{0.529411764705882,0.305882352941176,0.858823529411765}
\definecolor{color5}{rgb}{0.858823529411765,0.305882352941176,0.435294117647059}
\definecolor{color6}{rgb}{0.937254901960784,0.929411764705882,0.392156862745098}
\definecolor{color7}{rgb}{0.0901960784313725,0.486274509803922,0.0980392156862745}
\definecolor{color8}{rgb}{0.156862745098039,0.188235294117647,0.827450980392157}
\definecolor{color9}{rgb}{0.937254901960784,0.392156862745098,0.894117647058824}
\definecolor{color10}{rgb}{0.2,0.184313725490196,0.184313725490196}
\definecolor{color11}{rgb}{0.0156862745098039,0.803921568627451,0.976470588235294}

\begin{axis}[
tick align=outside,
tick pos=left,
x grid style={white!69.0196078431373!black},
xmajorgrids,
xmin=-0.0895, xmax=2.0995,
xtick style={color=black},
xtick={0,0.1,0.2,0.3,0.4,0.5,0.6,0.7,0.8,0.9,1,1.1,1.2,1.3,1.4,1.5,1.6,1.7,1.8,1.9,2},
xticklabels={0,,.2,,.4,,.6,,.8,,1,,20,,40,,60,,80,,100},
height=4.8cm,
width=6.5cm,
y grid style={white!69.0196078431373!black},
ymajorgrids,
ymin=0.2, ymax=1.05,
ytick style={color=black}
]
\addplot [semithick, color0, opacity=0.75]
table {%
0.01 0.5
0.02 0.52
0.03 0.42
0.04 0.41
0.05 0.53
0.06 0.41
0.07 0.54
0.08 0.55
0.09 0.51
0.1 0.49
0.11 0.47
0.12 0.45
0.13 0.51
0.14 0.5
0.15 0.53
0.16 0.48
0.17 0.41
0.18 0.51
0.19 0.5
0.2 0.49
0.21 0.46
0.22 0.45
0.23 0.51
0.24 0.47
0.25 0.42
0.26 0.57
0.27 0.54
0.28 0.51
0.29 0.48
0.3 0.37
0.31 0.49
0.32 0.54
0.33 0.49
0.34 0.5
0.35 0.49
0.36 0.5
0.37 0.49
0.38 0.48
0.39 0.48
0.4 0.51
0.41 0.52
0.42 0.49
0.43 0.4
0.44 0.46
0.45 0.55
0.46 0.47
0.47 0.53
0.48 0.43
0.49 0.49
0.5 0.53
0.51 0.59
0.52 0.46
0.53 0.43
0.54 0.5
0.55 0.58
0.56 0.53
0.57 0.54
0.58 0.53
0.59 0.51
0.6 0.5
0.61 0.55
0.62 0.51
0.63 0.47
0.64 0.54
0.65 0.39
0.66 0.42
0.67 0.49
0.68 0.5
0.69 0.51
0.7 0.51
0.71 0.58
0.72 0.44
0.73 0.51
0.74 0.49
0.75 0.46
0.76 0.52
0.77 0.45
0.78 0.49
0.79 0.54
0.8 0.51
0.81 0.49
0.82 0.46
0.83 0.45
0.84 0.51
0.85 0.46
0.86 0.54
0.87 0.6
0.88 0.45
0.89 0.52
0.9 0.51
0.91 0.48
0.92 0.55
0.93 0.53
0.94 0.53
0.95 0.53
0.96 0.47
0.97 0.49
0.98 0.47
0.99 0.48
1 0.5
1.02 0.57
1.03 0.48
1.04 0.55
1.05 0.6
1.06 0.45
1.07 0.51
1.08 0.46
1.09 0.55
1.1 0.51
1.11 0.53
1.12 0.47
1.13 0.48
1.14 0.49
1.15 0.62
1.16 0.46
1.17 0.5
1.18 0.48
1.19 0.49
1.2 0.4
1.21 0.51
1.22 0.44
1.23 0.51
1.24 0.56
1.25 0.45
1.26 0.46
1.27 0.54
1.28 0.5
1.29 0.58
1.3 0.44
1.31 0.53
1.32 0.49
1.33 0.58
1.34 0.55
1.35 0.51
1.36 0.48
1.37 0.45
1.38 0.41
1.39 0.45
1.4 0.4
1.41 0.53
1.42 0.55
1.43 0.42
1.44 0.52
1.45 0.52
1.46 0.48
1.47 0.49
1.48 0.46
1.49 0.44
1.5 0.47
1.51 0.44
1.52 0.5
1.53 0.52
1.54 0.52
1.55 0.52
1.56 0.55
1.57 0.54
1.58 0.56
1.59 0.48
1.6 0.44
1.61 0.37
1.62 0.51
1.63 0.45
1.64 0.55
1.65 0.56
1.66 0.56
1.67 0.55
1.68 0.45
1.69 0.47
1.7 0.55
1.71 0.51
1.72 0.51
1.73 0.39
1.74 0.49
1.75 0.45
1.76 0.49
1.77 0.47
1.78 0.61
1.79 0.44
1.8 0.56
1.81 0.49
1.82 0.47
1.83 0.44
1.84 0.49
1.85 0.46
1.86 0.55
1.87 0.45
1.88 0.45
1.89 0.48
1.9 0.44
1.91 0.42
1.92 0.52
1.93 0.46
1.94 0.59
1.95 0.56
1.96 0.42
1.97 0.47
1.98 0.52
1.99 0.5
2 0.51
};
\addplot [semithick, color1, opacity=0.75]
table {%
0.01 0.43
0.02 0.48
0.03 0.53
0.04 0.43
0.05 0.48
0.06 0.46
0.07 0.44
0.08 0.5
0.09 0.42
0.1 0.45
0.11 0.43
0.12 0.44
0.13 0.47
0.14 0.44
0.15 0.49
0.16 0.45
0.17 0.37
0.18 0.53
0.19 0.47
0.2 0.47
0.21 0.39
0.22 0.48
0.23 0.56
0.24 0.51
0.25 0.53
0.26 0.53
0.27 0.48
0.28 0.49
0.29 0.46
0.3 0.45
0.31 0.49
0.32 0.5
0.33 0.42
0.34 0.51
0.35 0.45
0.36 0.53
0.37 0.45
0.38 0.49
0.39 0.47
0.4 0.58
0.41 0.4
0.42 0.56
0.43 0.37
0.44 0.43
0.45 0.46
0.46 0.42
0.47 0.56
0.48 0.43
0.49 0.47
0.5 0.49
0.51 0.49
0.52 0.48
0.53 0.35
0.54 0.53
0.55 0.54
0.56 0.45
0.57 0.45
0.58 0.46
0.59 0.41
0.6 0.43
0.61 0.49
0.62 0.51
0.63 0.44
0.64 0.54
0.65 0.39
0.66 0.44
0.67 0.44
0.68 0.42
0.69 0.45
0.7 0.39
0.71 0.49
0.72 0.33
0.73 0.48
0.74 0.41
0.75 0.44
0.76 0.53
0.77 0.46
0.78 0.48
0.79 0.45
0.8 0.52
0.81 0.45
0.82 0.47
0.83 0.45
0.84 0.44
0.85 0.47
0.86 0.44
0.87 0.59
0.88 0.41
0.89 0.47
0.9 0.47
0.91 0.43
0.92 0.59
0.93 0.54
0.94 0.43
0.95 0.46
0.96 0.47
0.97 0.37
0.98 0.42
0.99 0.41
1 0.42
1.02 0.58
1.03 0.42
1.04 0.59
1.05 0.48
1.06 0.41
1.07 0.5
1.08 0.45
1.09 0.43
1.1 0.38
1.11 0.52
1.12 0.49
1.13 0.46
1.14 0.57
1.15 0.51
1.16 0.42
1.17 0.47
1.18 0.51
1.19 0.55
1.2 0.51
1.21 0.49
1.22 0.61
1.23 0.54
1.24 0.52
1.25 0.53
1.26 0.48
1.27 0.51
1.28 0.47
1.29 0.48
1.3 0.49
1.31 0.52
1.32 0.52
1.33 0.57
1.34 0.46
1.35 0.51
1.36 0.47
1.37 0.52
1.38 0.41
1.39 0.48
1.4 0.47
1.41 0.5
1.42 0.53
1.43 0.53
1.44 0.48
1.45 0.47
1.46 0.53
1.47 0.42
1.48 0.51
1.49 0.41
1.5 0.6
1.51 0.56
1.52 0.54
1.53 0.52
1.54 0.51
1.55 0.42
1.56 0.55
1.57 0.49
1.58 0.55
1.59 0.48
1.6 0.41
1.61 0.57
1.62 0.47
1.63 0.48
1.64 0.44
1.65 0.48
1.66 0.51
1.67 0.56
1.68 0.5
1.69 0.5
1.7 0.53
1.71 0.49
1.72 0.49
1.73 0.46
1.74 0.49
1.75 0.5
1.76 0.59
1.77 0.45
1.78 0.43
1.79 0.57
1.8 0.51
1.81 0.44
1.82 0.48
1.83 0.49
1.84 0.53
1.85 0.4
1.86 0.47
1.87 0.46
1.88 0.5
1.89 0.47
1.9 0.47
1.91 0.56
1.92 0.57
1.93 0.55
1.94 0.47
1.95 0.46
1.96 0.37
1.97 0.56
1.98 0.46
1.99 0.46
2 0.51
};
\addplot [semithick, color2, opacity=0.75]
table {%
0.01 0.43
0.02 0.5
0.03 0.54
0.04 0.45
0.05 0.5
0.06 0.42
0.07 0.5
0.08 0.48
0.09 0.39
0.1 0.44
0.11 0.43
0.12 0.46
0.13 0.48
0.14 0.42
0.15 0.49
0.16 0.43
0.17 0.41
0.18 0.53
0.19 0.46
0.2 0.49
0.21 0.42
0.22 0.49
0.23 0.55
0.24 0.5
0.25 0.51
0.26 0.53
0.27 0.47
0.28 0.5
0.29 0.45
0.3 0.46
0.31 0.49
0.32 0.49
0.33 0.41
0.34 0.52
0.35 0.46
0.36 0.53
0.37 0.46
0.38 0.49
0.39 0.47
0.4 0.59
0.41 0.4
0.42 0.56
0.43 0.36
0.44 0.44
0.45 0.48
0.46 0.42
0.47 0.55
0.48 0.43
0.49 0.49
0.5 0.51
0.51 0.49
0.52 0.48
0.53 0.34
0.54 0.53
0.55 0.55
0.56 0.44
0.57 0.43
0.58 0.46
0.59 0.42
0.6 0.42
0.61 0.48
0.62 0.51
0.63 0.44
0.64 0.54
0.65 0.39
0.66 0.43
0.67 0.44
0.68 0.43
0.69 0.45
0.7 0.4
0.71 0.47
0.72 0.34
0.73 0.49
0.74 0.41
0.75 0.43
0.76 0.54
0.77 0.47
0.78 0.49
0.79 0.45
0.8 0.53
0.81 0.45
0.82 0.45
0.83 0.44
0.84 0.45
0.85 0.46
0.86 0.45
0.87 0.6
0.88 0.41
0.89 0.48
0.9 0.47
0.91 0.43
0.92 0.59
0.93 0.55
0.94 0.43
0.95 0.46
0.96 0.46
0.97 0.37
0.98 0.42
0.99 0.42
1 0.42
1.02 0.59
1.03 0.42
1.04 0.58
1.05 0.49
1.06 0.42
1.07 0.49
1.08 0.47
1.09 0.43
1.1 0.38
1.11 0.52
1.12 0.49
1.13 0.48
1.14 0.55
1.15 0.5
1.16 0.41
1.17 0.47
1.18 0.52
1.19 0.54
1.2 0.51
1.21 0.48
1.22 0.6
1.23 0.54
1.24 0.55
1.25 0.53
1.26 0.46
1.27 0.52
1.28 0.47
1.29 0.47
1.3 0.49
1.31 0.52
1.32 0.52
1.33 0.56
1.34 0.47
1.35 0.47
1.36 0.47
1.37 0.54
1.38 0.42
1.39 0.49
1.4 0.48
1.41 0.5
1.42 0.52
1.43 0.53
1.44 0.52
1.45 0.44
1.46 0.53
1.47 0.41
1.48 0.47
1.49 0.39
1.5 0.6
1.51 0.55
1.52 0.56
1.53 0.55
1.54 0.5
1.55 0.42
1.56 0.57
1.57 0.5
1.58 0.55
1.59 0.47
1.6 0.39
1.61 0.55
1.62 0.48
1.63 0.48
1.64 0.45
1.65 0.5
1.66 0.51
1.67 0.57
1.68 0.5
1.69 0.52
1.7 0.5
1.71 0.49
1.72 0.5
1.73 0.46
1.74 0.49
1.75 0.48
1.76 0.61
1.77 0.46
1.78 0.43
1.79 0.58
1.8 0.5
1.81 0.44
1.82 0.47
1.83 0.48
1.84 0.52
1.85 0.4
1.86 0.49
1.87 0.49
1.88 0.5
1.89 0.47
1.9 0.47
1.91 0.57
1.92 0.56
1.93 0.54
1.94 0.44
1.95 0.44
1.96 0.4
1.97 0.58
1.98 0.49
1.99 0.46
2 0.5
};
\addplot [semithick, color3, opacity=0.75]
table {%
0.01 0.45
0.02 0.43
0.03 0.47
0.04 0.46
0.05 0.49
0.06 0.42
0.07 0.5
0.08 0.49
0.09 0.49
0.1 0.53
0.11 0.53
0.12 0.49
0.13 0.46
0.14 0.42
0.15 0.54
0.16 0.5
0.17 0.42
0.18 0.47
0.19 0.48
0.2 0.53
0.21 0.45
0.22 0.42
0.23 0.59
0.24 0.5
0.25 0.48
0.26 0.62
0.27 0.55
0.28 0.52
0.29 0.51
0.3 0.51
0.31 0.55
0.32 0.5
0.33 0.51
0.34 0.56
0.35 0.45
0.36 0.49
0.37 0.49
0.38 0.39
0.39 0.54
0.4 0.55
0.41 0.45
0.42 0.55
0.43 0.43
0.44 0.5
0.45 0.52
0.46 0.49
0.47 0.57
0.48 0.44
0.49 0.51
0.5 0.53
0.51 0.52
0.52 0.5
0.53 0.46
0.54 0.56
0.55 0.57
0.56 0.56
0.57 0.57
0.58 0.5
0.59 0.47
0.6 0.51
0.61 0.54
0.62 0.49
0.63 0.51
0.64 0.62
0.65 0.45
0.66 0.46
0.67 0.55
0.68 0.53
0.69 0.48
0.7 0.53
0.71 0.5
0.72 0.46
0.73 0.49
0.74 0.46
0.75 0.48
0.76 0.57
0.77 0.53
0.78 0.52
0.79 0.52
0.8 0.53
0.81 0.48
0.82 0.46
0.83 0.45
0.84 0.47
0.85 0.51
0.86 0.5
0.87 0.56
0.88 0.46
0.89 0.5
0.9 0.42
0.91 0.4
0.92 0.55
0.93 0.56
0.94 0.47
0.95 0.46
0.96 0.5
0.97 0.46
0.98 0.47
0.99 0.51
1 0.49
1.02 0.55
1.03 0.48
1.04 0.54
1.05 0.49
1.06 0.49
1.07 0.63
1.08 0.55
1.09 0.59
1.1 0.49
1.11 0.48
1.12 0.45
1.13 0.53
1.14 0.44
1.15 0.49
1.16 0.47
1.17 0.51
1.18 0.5
1.19 0.54
1.2 0.47
1.21 0.56
1.22 0.46
1.23 0.44
1.24 0.57
1.25 0.55
1.26 0.43
1.27 0.57
1.28 0.52
1.29 0.49
1.3 0.51
1.31 0.53
1.32 0.52
1.33 0.58
1.34 0.5
1.35 0.53
1.36 0.44
1.37 0.44
1.38 0.51
1.39 0.42
1.4 0.43
1.41 0.45
1.42 0.45
1.43 0.47
1.44 0.45
1.45 0.51
1.46 0.48
1.47 0.48
1.48 0.49
1.49 0.5
1.5 0.56
1.51 0.48
1.52 0.53
1.53 0.54
1.54 0.52
1.55 0.53
1.56 0.49
1.57 0.53
1.58 0.48
1.59 0.53
1.6 0.49
1.61 0.43
1.62 0.65
1.63 0.45
1.64 0.46
1.65 0.46
1.66 0.53
1.67 0.56
1.68 0.48
1.69 0.5
1.7 0.55
1.71 0.57
1.72 0.43
1.73 0.42
1.74 0.44
1.75 0.51
1.76 0.47
1.77 0.43
1.78 0.6
1.79 0.43
1.8 0.52
1.81 0.51
1.82 0.5
1.83 0.49
1.84 0.5
1.85 0.57
1.86 0.53
1.87 0.46
1.88 0.54
1.89 0.49
1.9 0.47
1.91 0.42
1.92 0.5
1.93 0.48
1.94 0.49
1.95 0.5
1.96 0.5
1.97 0.56
1.98 0.54
1.99 0.49
2 0.61
};
\addplot [semithick, color4, opacity=0.75]
table {%
0.01 0.44
0.02 0.42
0.03 0.47
0.04 0.47
0.05 0.49
0.06 0.43
0.07 0.49
0.08 0.49
0.09 0.49
0.1 0.52
0.11 0.53
0.12 0.47
0.13 0.44
0.14 0.41
0.15 0.54
0.16 0.5
0.17 0.41
0.18 0.47
0.19 0.48
0.2 0.52
0.21 0.42
0.22 0.44
0.23 0.6
0.24 0.51
0.25 0.47
0.26 0.62
0.27 0.55
0.28 0.55
0.29 0.5
0.3 0.5
0.31 0.56
0.32 0.51
0.33 0.53
0.34 0.57
0.35 0.43
0.36 0.49
0.37 0.47
0.38 0.4
0.39 0.52
0.4 0.56
0.41 0.42
0.42 0.55
0.43 0.4
0.44 0.48
0.45 0.52
0.46 0.5
0.47 0.57
0.48 0.44
0.49 0.53
0.5 0.57
0.51 0.53
0.52 0.5
0.53 0.44
0.54 0.55
0.55 0.56
0.56 0.57
0.57 0.56
0.58 0.51
0.59 0.46
0.6 0.52
0.61 0.54
0.62 0.51
0.63 0.51
0.64 0.62
0.65 0.44
0.66 0.46
0.67 0.57
0.68 0.53
0.69 0.47
0.7 0.53
0.71 0.52
0.72 0.43
0.73 0.49
0.74 0.44
0.75 0.47
0.76 0.57
0.77 0.55
0.78 0.52
0.79 0.5
0.8 0.54
0.81 0.52
0.82 0.46
0.83 0.46
0.84 0.46
0.85 0.5
0.86 0.5
0.87 0.57
0.88 0.46
0.89 0.51
0.9 0.42
0.91 0.43
0.92 0.55
0.93 0.57
0.94 0.47
0.95 0.46
0.96 0.51
0.97 0.45
0.98 0.48
0.99 0.52
1 0.5
1.02 0.54
1.03 0.5
1.04 0.55
1.05 0.48
1.06 0.52
1.07 0.62
1.08 0.54
1.09 0.59
1.1 0.46
1.11 0.51
1.12 0.47
1.13 0.52
1.14 0.44
1.15 0.47
1.16 0.47
1.17 0.48
1.18 0.51
1.19 0.55
1.2 0.48
1.21 0.54
1.22 0.45
1.23 0.44
1.24 0.57
1.25 0.5
1.26 0.44
1.27 0.55
1.28 0.54
1.29 0.49
1.3 0.52
1.31 0.51
1.32 0.53
1.33 0.56
1.34 0.5
1.35 0.54
1.36 0.43
1.37 0.44
1.38 0.5
1.39 0.43
1.4 0.44
1.41 0.47
1.42 0.47
1.43 0.46
1.44 0.44
1.45 0.52
1.46 0.49
1.47 0.47
1.48 0.49
1.49 0.5
1.5 0.56
1.51 0.48
1.52 0.54
1.53 0.54
1.54 0.52
1.55 0.52
1.56 0.51
1.57 0.55
1.58 0.49
1.59 0.54
1.6 0.49
1.61 0.44
1.62 0.65
1.63 0.45
1.64 0.47
1.65 0.45
1.66 0.54
1.67 0.57
1.68 0.48
1.69 0.51
1.7 0.56
1.71 0.55
1.72 0.45
1.73 0.4
1.74 0.41
1.75 0.49
1.76 0.47
1.77 0.42
1.78 0.57
1.79 0.45
1.8 0.55
1.81 0.51
1.82 0.5
1.83 0.49
1.84 0.49
1.85 0.56
1.86 0.5
1.87 0.46
1.88 0.5
1.89 0.48
1.9 0.5
1.91 0.43
1.92 0.5
1.93 0.49
1.94 0.48
1.95 0.52
1.96 0.49
1.97 0.55
1.98 0.53
1.99 0.47
2 0.6
};
\addplot [semithick, color5, opacity=0.75]
table {%
0.01 0.49
0.02 0.5
0.03 0.51
0.04 0.41
0.05 0.48
0.06 0.49
0.07 0.49
0.08 0.51
0.09 0.5
0.1 0.53
0.11 0.48
0.12 0.43
0.13 0.52
0.14 0.49
0.15 0.56
0.16 0.47
0.17 0.4
0.18 0.46
0.19 0.55
0.2 0.54
0.21 0.45
0.22 0.45
0.23 0.6
0.24 0.51
0.25 0.47
0.26 0.59
0.27 0.47
0.28 0.57
0.29 0.51
0.3 0.5
0.31 0.54
0.32 0.42
0.33 0.55
0.34 0.51
0.35 0.45
0.36 0.53
0.37 0.46
0.38 0.48
0.39 0.51
0.4 0.6
0.41 0.47
0.42 0.55
0.43 0.34
0.44 0.51
0.45 0.53
0.46 0.51
0.47 0.59
0.48 0.42
0.49 0.53
0.5 0.57
0.51 0.51
0.52 0.48
0.53 0.45
0.54 0.52
0.55 0.54
0.56 0.55
0.57 0.53
0.58 0.48
0.59 0.47
0.6 0.49
0.61 0.51
0.62 0.5
0.63 0.5
0.64 0.58
0.65 0.42
0.66 0.45
0.67 0.54
0.68 0.51
0.69 0.46
0.7 0.48
0.71 0.51
0.72 0.45
0.73 0.51
0.74 0.47
0.75 0.51
0.76 0.52
0.77 0.47
0.78 0.52
0.79 0.53
0.8 0.58
0.81 0.5
0.82 0.49
0.83 0.51
0.84 0.48
0.85 0.52
0.86 0.49
0.87 0.57
0.88 0.46
0.89 0.49
0.9 0.45
0.91 0.41
0.92 0.56
0.93 0.57
0.94 0.49
0.95 0.49
0.96 0.47
0.97 0.45
0.98 0.52
0.99 0.48
1 0.49
1.02 0.53
1.03 0.47
1.04 0.53
1.05 0.48
1.06 0.46
1.07 0.51
1.08 0.48
1.09 0.56
1.1 0.52
1.11 0.46
1.12 0.4
1.13 0.56
1.14 0.46
1.15 0.55
1.16 0.43
1.17 0.54
1.18 0.54
1.19 0.55
1.2 0.41
1.21 0.57
1.22 0.42
1.23 0.49
1.24 0.5
1.25 0.51
1.26 0.44
1.27 0.61
1.28 0.57
1.29 0.51
1.3 0.43
1.31 0.5
1.32 0.55
1.33 0.55
1.34 0.53
1.35 0.49
1.36 0.48
1.37 0.5
1.38 0.47
1.39 0.48
1.4 0.49
1.41 0.47
1.42 0.48
1.43 0.5
1.44 0.43
1.45 0.55
1.46 0.51
1.47 0.41
1.48 0.41
1.49 0.52
1.5 0.58
1.51 0.44
1.52 0.46
1.53 0.53
1.54 0.53
1.55 0.53
1.56 0.54
1.57 0.49
1.58 0.43
1.59 0.52
1.6 0.49
1.61 0.58
1.62 0.6
1.63 0.41
1.64 0.49
1.65 0.52
1.66 0.54
1.67 0.51
1.68 0.45
1.69 0.6
1.7 0.51
1.71 0.55
1.72 0.43
1.73 0.48
1.74 0.46
1.75 0.52
1.76 0.57
1.77 0.48
1.78 0.55
1.79 0.43
1.8 0.52
1.81 0.49
1.82 0.5
1.83 0.47
1.84 0.39
1.85 0.52
1.86 0.49
1.87 0.48
1.88 0.5
1.89 0.5
1.9 0.45
1.91 0.4
1.92 0.41
1.93 0.51
1.94 0.52
1.95 0.55
1.96 0.47
1.97 0.47
1.98 0.56
1.99 0.48
2 0.54
};
\addplot [semithick, color6, opacity=0.75, dashed]
table {%
0.01 0.49
0.02 0.48
0.03 0.51
0.04 0.56
0.05 0.49
0.06 0.42
0.07 0.52
0.08 0.51
0.09 0.49
0.1 0.51
0.11 0.53
0.12 0.43
0.13 0.51
0.14 0.47
0.15 0.54
0.16 0.5
0.17 0.43
0.18 0.63
0.19 0.48
0.2 0.53
0.21 0.5
0.22 0.57
0.23 0.4
0.24 0.51
0.25 0.53
0.26 0.53
0.27 0.49
0.28 0.53
0.29 0.43
0.3 0.49
0.31 0.54
0.32 0.51
0.33 0.54
0.34 0.46
0.35 0.49
0.36 0.51
0.37 0.52
0.38 0.4
0.39 0.53
0.4 0.47
0.41 0.49
0.42 0.39
0.43 0.5
0.44 0.52
0.45 0.58
0.46 0.54
0.47 0.46
0.48 0.5
0.49 0.52
0.5 0.5
0.51 0.49
0.52 0.52
0.53 0.5
0.54 0.48
0.55 0.51
0.56 0.53
0.57 0.46
0.58 0.5
0.59 0.5
0.6 0.51
0.61 0.48
0.62 0.46
0.63 0.45
0.64 0.55
0.65 0.49
0.66 0.42
0.67 0.55
0.68 0.49
0.69 0.45
0.7 0.47
0.71 0.47
0.72 0.42
0.73 0.49
0.74 0.38
0.75 0.47
0.76 0.54
0.77 0.48
0.78 0.55
0.79 0.51
0.8 0.5
0.81 0.5
0.82 0.52
0.83 0.52
0.84 0.46
0.85 0.44
0.86 0.52
0.87 0.49
0.88 0.53
0.89 0.57
0.9 0.41
0.91 0.51
0.92 0.54
0.93 0.58
0.94 0.42
0.95 0.4
0.96 0.48
0.97 0.45
0.98 0.47
0.99 0.47
1 0.51
1.02 0.49
1.03 0.53
1.04 0.52
1.05 0.43
1.06 0.54
1.07 0.53
1.08 0.41
1.09 0.46
1.1 0.44
1.11 0.56
1.12 0.47
1.13 0.47
1.14 0.49
1.15 0.46
1.16 0.48
1.17 0.51
1.18 0.58
1.19 0.45
1.2 0.53
1.21 0.48
1.22 0.47
1.23 0.52
1.24 0.53
1.25 0.44
1.26 0.45
1.27 0.55
1.28 0.46
1.29 0.41
1.3 0.45
1.31 0.48
1.32 0.57
1.33 0.44
1.34 0.62
1.35 0.5
1.36 0.57
1.37 0.51
1.38 0.48
1.39 0.52
1.4 0.55
1.41 0.47
1.42 0.56
1.43 0.56
1.44 0.5
1.45 0.47
1.46 0.44
1.47 0.51
1.48 0.52
1.49 0.48
1.5 0.39
1.51 0.47
1.52 0.56
1.53 0.44
1.54 0.5
1.55 0.47
1.56 0.61
1.57 0.54
1.58 0.6
1.59 0.62
1.6 0.49
1.61 0.39
1.62 0.5
1.63 0.42
1.64 0.5
1.65 0.49
1.66 0.48
1.67 0.58
1.68 0.42
1.69 0.49
1.7 0.46
1.71 0.5
1.72 0.53
1.73 0.46
1.74 0.39
1.75 0.51
1.76 0.57
1.77 0.5
1.78 0.46
1.79 0.52
1.8 0.53
1.81 0.48
1.82 0.58
1.83 0.49
1.84 0.48
1.85 0.58
1.86 0.47
1.87 0.56
1.88 0.49
1.89 0.59
1.9 0.56
1.91 0.38
1.92 0.51
1.93 0.63
1.94 0.54
1.95 0.46
1.96 0.54
1.97 0.42
1.98 0.45
1.99 0.48
2 0.49
};
\addplot [semithick, color7, opacity=0.75, dashed]
table {%
0.01 0.49
0.02 0.48
0.03 0.51
0.04 0.56
0.05 0.49
0.06 0.42
0.07 0.52
0.08 0.51
0.09 0.49
0.1 0.51
0.11 0.53
0.12 0.43
0.13 0.51
0.14 0.47
0.15 0.54
0.16 0.5
0.17 0.43
0.18 0.63
0.19 0.48
0.2 0.53
0.21 0.5
0.22 0.57
0.23 0.4
0.24 0.51
0.25 0.53
0.26 0.53
0.27 0.49
0.28 0.53
0.29 0.43
0.3 0.49
0.31 0.54
0.32 0.51
0.33 0.54
0.34 0.46
0.35 0.49
0.36 0.51
0.37 0.52
0.38 0.4
0.39 0.53
0.4 0.47
0.41 0.49
0.42 0.39
0.43 0.5
0.44 0.52
0.45 0.58
0.46 0.54
0.47 0.46
0.48 0.5
0.49 0.52
0.5 0.5
0.51 0.49
0.52 0.52
0.53 0.5
0.54 0.48
0.55 0.51
0.56 0.53
0.57 0.46
0.58 0.5
0.59 0.5
0.6 0.51
0.61 0.48
0.62 0.46
0.63 0.45
0.64 0.55
0.65 0.49
0.66 0.42
0.67 0.55
0.68 0.49
0.69 0.45
0.7 0.47
0.71 0.47
0.72 0.42
0.73 0.49
0.74 0.38
0.75 0.47
0.76 0.54
0.77 0.48
0.78 0.55
0.79 0.51
0.8 0.5
0.81 0.5
0.82 0.52
0.83 0.52
0.84 0.46
0.85 0.44
0.86 0.52
0.87 0.49
0.88 0.53
0.89 0.57
0.9 0.41
0.91 0.51
0.92 0.54
0.93 0.58
0.94 0.42
0.95 0.4
0.96 0.48
0.97 0.45
0.98 0.47
0.99 0.47
1 0.51
1.02 0.49
1.03 0.53
1.04 0.52
1.05 0.43
1.06 0.54
1.07 0.53
1.08 0.41
1.09 0.46
1.1 0.44
1.11 0.56
1.12 0.47
1.13 0.47
1.14 0.49
1.15 0.46
1.16 0.48
1.17 0.51
1.18 0.58
1.19 0.45
1.2 0.53
1.21 0.48
1.22 0.47
1.23 0.52
1.24 0.53
1.25 0.44
1.26 0.45
1.27 0.55
1.28 0.46
1.29 0.41
1.3 0.45
1.31 0.48
1.32 0.57
1.33 0.44
1.34 0.62
1.35 0.5
1.36 0.57
1.37 0.51
1.38 0.48
1.39 0.52
1.4 0.55
1.41 0.47
1.42 0.56
1.43 0.56
1.44 0.5
1.45 0.47
1.46 0.44
1.47 0.51
1.48 0.52
1.49 0.48
1.5 0.39
1.51 0.47
1.52 0.56
1.53 0.44
1.54 0.5
1.55 0.47
1.56 0.61
1.57 0.54
1.58 0.6
1.59 0.62
1.6 0.49
1.61 0.39
1.62 0.5
1.63 0.42
1.64 0.5
1.65 0.49
1.66 0.48
1.67 0.58
1.68 0.42
1.69 0.49
1.7 0.46
1.71 0.5
1.72 0.53
1.73 0.46
1.74 0.39
1.75 0.51
1.76 0.57
1.77 0.5
1.78 0.46
1.79 0.52
1.8 0.53
1.81 0.48
1.82 0.58
1.83 0.49
1.84 0.48
1.85 0.58
1.86 0.47
1.87 0.56
1.88 0.49
1.89 0.59
1.9 0.56
1.91 0.38
1.92 0.51
1.93 0.63
1.94 0.54
1.95 0.46
1.96 0.54
1.97 0.42
1.98 0.45
1.99 0.48
2 0.49
};
\addplot [semithick, color8, opacity=0.75, dashed]
table {%
0.01 0.48
0.02 0.38
0.03 0.52
0.04 0.6
0.05 0.54
0.06 0.43
0.07 0.58
0.08 0.58
0.09 0.55
0.1 0.38
0.11 0.5
0.12 0.46
0.13 0.55
0.14 0.53
0.15 0.53
0.16 0.57
0.17 0.47
0.18 0.54
0.19 0.5
0.2 0.46
0.21 0.47
0.22 0.55
0.23 0.42
0.24 0.45
0.25 0.54
0.26 0.46
0.27 0.49
0.28 0.55
0.29 0.49
0.3 0.45
0.31 0.47
0.32 0.53
0.33 0.51
0.34 0.52
0.35 0.46
0.36 0.59
0.37 0.44
0.38 0.42
0.39 0.45
0.4 0.43
0.41 0.5
0.42 0.42
0.43 0.39
0.44 0.51
0.45 0.51
0.46 0.47
0.47 0.49
0.48 0.49
0.49 0.49
0.5 0.5
0.51 0.58
0.52 0.57
0.53 0.49
0.54 0.48
0.55 0.49
0.56 0.49
0.57 0.52
0.58 0.51
0.59 0.49
0.6 0.54
0.61 0.5
0.62 0.52
0.63 0.49
0.64 0.52
0.65 0.5
0.66 0.41
0.67 0.46
0.68 0.48
0.69 0.53
0.7 0.48
0.71 0.47
0.72 0.47
0.73 0.58
0.74 0.45
0.75 0.47
0.76 0.5
0.77 0.49
0.78 0.56
0.79 0.47
0.8 0.6
0.81 0.48
0.82 0.47
0.83 0.48
0.84 0.53
0.85 0.52
0.86 0.48
0.87 0.56
0.88 0.53
0.89 0.52
0.9 0.36
0.91 0.47
0.92 0.57
0.93 0.55
0.94 0.44
0.95 0.5
0.96 0.45
0.97 0.54
0.98 0.55
0.99 0.42
1 0.51
1.02 0.51
1.03 0.63
1.04 0.45
1.05 0.53
1.06 0.51
1.07 0.56
1.08 0.48
1.09 0.52
1.1 0.48
1.11 0.54
1.12 0.53
1.13 0.45
1.14 0.47
1.15 0.53
1.16 0.47
1.17 0.54
1.18 0.54
1.19 0.52
1.2 0.56
1.21 0.55
1.22 0.43
1.23 0.47
1.24 0.52
1.25 0.54
1.26 0.52
1.27 0.55
1.28 0.51
1.29 0.5
1.3 0.47
1.31 0.56
1.32 0.53
1.33 0.48
1.34 0.53
1.35 0.5
1.36 0.48
1.37 0.44
1.38 0.47
1.39 0.49
1.4 0.46
1.41 0.53
1.42 0.56
1.43 0.59
1.44 0.51
1.45 0.53
1.46 0.55
1.47 0.46
1.48 0.56
1.49 0.49
1.5 0.44
1.51 0.45
1.52 0.49
1.53 0.47
1.54 0.49
1.55 0.54
1.56 0.61
1.57 0.42
1.58 0.5
1.59 0.6
1.6 0.45
1.61 0.49
1.62 0.46
1.63 0.48
1.64 0.49
1.65 0.48
1.66 0.61
1.67 0.49
1.68 0.43
1.69 0.49
1.7 0.49
1.71 0.51
1.72 0.51
1.73 0.44
1.74 0.5
1.75 0.51
1.76 0.55
1.77 0.52
1.78 0.48
1.79 0.47
1.8 0.56
1.81 0.53
1.82 0.51
1.83 0.54
1.84 0.51
1.85 0.51
1.86 0.56
1.87 0.48
1.88 0.4
1.89 0.49
1.9 0.53
1.91 0.44
1.92 0.63
1.93 0.57
1.94 0.51
1.95 0.45
1.96 0.47
1.97 0.48
1.98 0.49
1.99 0.5
2 0.42
};
\addplot [semithick, color9, opacity=0.75, dashed]
table {%
0.01 0.47
0.02 0.47
0.03 0.47
0.04 0.48
0.05 0.48
0.06 0.48
0.07 0.45
0.08 0.56
0.09 0.51
0.1 0.51
0.11 0.52
0.12 0.42
0.13 0.46
0.14 0.51
0.15 0.54
0.16 0.5
0.17 0.43
0.18 0.5
0.19 0.46
0.2 0.48
0.21 0.41
0.22 0.35
0.23 0.49
0.24 0.51
0.25 0.39
0.26 0.57
0.27 0.52
0.28 0.47
0.29 0.48
0.3 0.43
0.31 0.58
0.32 0.52
0.33 0.47
0.34 0.61
0.35 0.51
0.36 0.49
0.37 0.5
0.38 0.5
0.39 0.46
0.4 0.52
0.41 0.55
0.42 0.54
0.43 0.44
0.44 0.49
0.45 0.47
0.46 0.47
0.47 0.48
0.48 0.44
0.49 0.54
0.5 0.54
0.51 0.54
0.52 0.41
0.53 0.39
0.54 0.5
0.55 0.5
0.56 0.48
0.57 0.53
0.58 0.52
0.59 0.51
0.6 0.49
0.61 0.57
0.62 0.56
0.63 0.45
0.64 0.55
0.65 0.46
0.66 0.44
0.67 0.55
0.68 0.54
0.69 0.48
0.7 0.53
0.71 0.52
0.72 0.44
0.73 0.55
0.74 0.45
0.75 0.45
0.76 0.52
0.77 0.41
0.78 0.52
0.79 0.53
0.8 0.47
0.81 0.54
0.82 0.46
0.83 0.45
0.84 0.55
0.85 0.5
0.86 0.49
0.87 0.6
0.88 0.44
0.89 0.56
0.9 0.5
0.91 0.51
0.92 0.63
0.93 0.56
0.94 0.54
0.95 0.47
0.96 0.51
0.97 0.42
0.98 0.47
0.99 0.54
1 0.39
1.02 0.53
1.03 0.48
1.04 0.58
1.05 0.57
1.06 0.53
1.07 0.59
1.08 0.55
1.09 0.51
1.1 0.47
1.11 0.52
1.12 0.46
1.13 0.58
1.14 0.53
1.15 0.53
1.16 0.44
1.17 0.53
1.18 0.54
1.19 0.49
1.2 0.55
1.21 0.61
1.22 0.39
1.23 0.5
1.24 0.56
1.25 0.48
1.26 0.41
1.27 0.5
1.28 0.51
1.29 0.6
1.3 0.5
1.31 0.53
1.32 0.5
1.33 0.53
1.34 0.47
1.35 0.52
1.36 0.5
1.37 0.46
1.38 0.5
1.39 0.46
1.4 0.42
1.41 0.39
1.42 0.51
1.43 0.4
1.44 0.49
1.45 0.51
1.46 0.42
1.47 0.51
1.48 0.55
1.49 0.49
1.5 0.67
1.51 0.55
1.52 0.54
1.53 0.47
1.54 0.52
1.55 0.54
1.56 0.55
1.57 0.47
1.58 0.5
1.59 0.63
1.6 0.53
1.61 0.43
1.62 0.6
1.63 0.44
1.64 0.51
1.65 0.46
1.66 0.49
1.67 0.48
1.68 0.54
1.69 0.5
1.7 0.58
1.71 0.52
1.72 0.51
1.73 0.48
1.74 0.44
1.75 0.46
1.76 0.46
1.77 0.46
1.78 0.61
1.79 0.48
1.8 0.46
1.81 0.49
1.82 0.43
1.83 0.43
1.84 0.56
1.85 0.5
1.86 0.63
1.87 0.45
1.88 0.44
1.89 0.51
1.9 0.46
1.91 0.45
1.92 0.45
1.93 0.47
1.94 0.5
1.95 0.59
1.96 0.44
1.97 0.5
1.98 0.55
1.99 0.53
2 0.56
};
\addplot [semithick, color10, opacity=0.75, dashed]
table {%
0.01 0.4
0.02 0.43
0.03 0.5
0.04 0.46
0.05 0.49
0.06 0.47
0.07 0.48
0.08 0.49
0.09 0.54
0.1 0.57
0.11 0.49
0.12 0.4
0.13 0.47
0.14 0.52
0.15 0.49
0.16 0.44
0.17 0.49
0.18 0.49
0.19 0.48
0.2 0.44
0.21 0.36
0.22 0.37
0.23 0.51
0.24 0.47
0.25 0.45
0.26 0.61
0.27 0.53
0.28 0.47
0.29 0.45
0.3 0.42
0.31 0.54
0.32 0.48
0.33 0.42
0.34 0.56
0.35 0.47
0.36 0.54
0.37 0.42
0.38 0.43
0.39 0.44
0.4 0.54
0.41 0.46
0.42 0.54
0.43 0.46
0.44 0.49
0.45 0.52
0.46 0.48
0.47 0.51
0.48 0.44
0.49 0.57
0.5 0.53
0.51 0.55
0.52 0.5
0.53 0.43
0.54 0.52
0.55 0.46
0.56 0.52
0.57 0.55
0.58 0.54
0.59 0.51
0.6 0.42
0.61 0.51
0.62 0.55
0.63 0.46
0.64 0.52
0.65 0.41
0.66 0.4
0.67 0.53
0.68 0.53
0.69 0.51
0.7 0.52
0.71 0.5
0.72 0.44
0.73 0.57
0.74 0.41
0.75 0.41
0.76 0.58
0.77 0.48
0.78 0.52
0.79 0.5
0.8 0.49
0.81 0.51
0.82 0.51
0.83 0.47
0.84 0.52
0.85 0.46
0.86 0.5
0.87 0.59
0.88 0.46
0.89 0.57
0.9 0.49
0.91 0.49
0.92 0.59
0.93 0.57
0.94 0.53
0.95 0.5
0.96 0.47
0.97 0.43
0.98 0.47
0.99 0.55
1 0.4
1.02 0.58
1.03 0.48
1.04 0.55
1.05 0.57
1.06 0.52
1.07 0.53
1.08 0.54
1.09 0.5
1.1 0.43
1.11 0.57
1.12 0.42
1.13 0.54
1.14 0.46
1.15 0.5
1.16 0.45
1.17 0.53
1.18 0.48
1.19 0.49
1.2 0.57
1.21 0.58
1.22 0.43
1.23 0.52
1.24 0.58
1.25 0.46
1.26 0.52
1.27 0.48
1.28 0.51
1.29 0.54
1.3 0.5
1.31 0.53
1.32 0.42
1.33 0.55
1.34 0.47
1.35 0.55
1.36 0.52
1.37 0.5
1.38 0.51
1.39 0.42
1.4 0.39
1.41 0.42
1.42 0.53
1.43 0.47
1.44 0.49
1.45 0.56
1.46 0.4
1.47 0.44
1.48 0.47
1.49 0.45
1.5 0.64
1.51 0.49
1.52 0.52
1.53 0.52
1.54 0.56
1.55 0.52
1.56 0.51
1.57 0.44
1.58 0.47
1.59 0.54
1.6 0.47
1.61 0.47
1.62 0.59
1.63 0.41
1.64 0.53
1.65 0.45
1.66 0.48
1.67 0.49
1.68 0.48
1.69 0.49
1.7 0.56
1.71 0.47
1.72 0.55
1.73 0.47
1.74 0.4
1.75 0.45
1.76 0.49
1.77 0.49
1.78 0.62
1.79 0.5
1.8 0.52
1.81 0.55
1.82 0.47
1.83 0.44
1.84 0.52
1.85 0.51
1.86 0.54
1.87 0.49
1.88 0.48
1.89 0.55
1.9 0.47
1.91 0.42
1.92 0.44
1.93 0.45
1.94 0.47
1.95 0.59
1.96 0.49
1.97 0.53
1.98 0.57
1.99 0.58
2 0.57
};
\addplot [semithick, color11, opacity=0.75, dashed]
table {%
0.01 0.45
0.02 0.48
0.03 0.54
0.04 0.63
0.05 0.58
0.06 0.51
0.07 0.53
0.08 0.51
0.09 0.47
0.1 0.47
0.11 0.59
0.12 0.49
0.13 0.5
0.14 0.43
0.15 0.54
0.16 0.53
0.17 0.44
0.18 0.49
0.19 0.52
0.2 0.45
0.21 0.41
0.22 0.53
0.23 0.42
0.24 0.53
0.25 0.47
0.26 0.61
0.27 0.48
0.28 0.46
0.29 0.42
0.3 0.48
0.31 0.48
0.32 0.54
0.33 0.45
0.34 0.51
0.35 0.42
0.36 0.51
0.37 0.5
0.38 0.52
0.39 0.59
0.4 0.51
0.41 0.45
0.42 0.52
0.43 0.45
0.44 0.57
0.45 0.47
0.46 0.47
0.47 0.53
0.48 0.45
0.49 0.6
0.5 0.56
0.51 0.52
0.52 0.52
0.53 0.46
0.54 0.48
0.55 0.51
0.56 0.49
0.57 0.54
0.58 0.54
0.59 0.49
0.6 0.45
0.61 0.44
0.62 0.43
0.63 0.48
0.64 0.54
0.65 0.47
0.66 0.52
0.67 0.57
0.68 0.4
0.69 0.47
0.7 0.43
0.71 0.52
0.72 0.44
0.73 0.58
0.74 0.52
0.75 0.53
0.76 0.55
0.77 0.47
0.78 0.46
0.79 0.58
0.8 0.51
0.81 0.46
0.82 0.55
0.83 0.39
0.84 0.5
0.85 0.55
0.86 0.47
0.87 0.47
0.88 0.39
0.89 0.57
0.9 0.5
0.91 0.48
0.92 0.54
0.93 0.42
0.94 0.48
0.95 0.46
0.96 0.5
0.97 0.45
0.98 0.47
0.99 0.48
1 0.41
1.02 0.47
1.03 0.57
1.04 0.53
1.05 0.54
1.06 0.46
1.07 0.55
1.08 0.53
1.09 0.5
1.1 0.4
1.11 0.55
1.12 0.45
1.13 0.51
1.14 0.55
1.15 0.48
1.16 0.45
1.17 0.54
1.18 0.54
1.19 0.5
1.2 0.51
1.21 0.54
1.22 0.59
1.23 0.4
1.24 0.5
1.25 0.45
1.26 0.55
1.27 0.45
1.28 0.54
1.29 0.5
1.3 0.52
1.31 0.44
1.32 0.57
1.33 0.58
1.34 0.54
1.35 0.58
1.36 0.47
1.37 0.43
1.38 0.5
1.39 0.51
1.4 0.44
1.41 0.52
1.42 0.45
1.43 0.51
1.44 0.48
1.45 0.49
1.46 0.53
1.47 0.49
1.48 0.52
1.49 0.57
1.5 0.48
1.51 0.59
1.52 0.52
1.53 0.46
1.54 0.42
1.55 0.5
1.56 0.46
1.57 0.52
1.58 0.55
1.59 0.57
1.6 0.37
1.61 0.44
1.62 0.5
1.63 0.54
1.64 0.59
1.65 0.52
1.66 0.5
1.67 0.47
1.68 0.5
1.69 0.45
1.7 0.52
1.71 0.42
1.72 0.45
1.73 0.49
1.74 0.37
1.75 0.46
1.76 0.46
1.77 0.42
1.78 0.5
1.79 0.43
1.8 0.56
1.81 0.45
1.82 0.47
1.83 0.44
1.84 0.45
1.85 0.55
1.86 0.55
1.87 0.51
1.88 0.44
1.89 0.5
1.9 0.5
1.91 0.46
1.92 0.47
1.93 0.43
1.94 0.58
1.95 0.56
1.96 0.47
1.97 0.5
1.98 0.51
1.99 0.48
2 0.45
};
\addplot [semithick, black, opacity=1, dash pattern=on 1pt off 1pt]
table {%
-0.0895000000000001 0.5
2.0995 0.5
};
\addplot [semithick, black, opacity=1, dash pattern=on 1pt off 1pt]
table {%
-0.0895000000000001 1
2.0995 1
};
\addplot [semithick, black, opacity=1, dash pattern=on 1pt off 1pt]
table {%
1 0.2
1 1.05
};
\end{axis}

\end{tikzpicture}

%% file: plots/decoupled2/GAUxUNI.tex
% This file was created by tikzplotlib v0.9.6.
\begin{tikzpicture}

\definecolor{color0}{rgb}{0.866666666666667,0.494117647058824,0.164705882352941}
\definecolor{color1}{rgb}{0.164705882352941,0.643137254901961,0.866666666666667}
\definecolor{color2}{rgb}{0.584313725490196,0.866666666666667,0.164705882352941}
\definecolor{color3}{rgb}{0.109803921568627,0.337254901960784,0.129411764705882}
\definecolor{color4}{rgb}{0.529411764705882,0.305882352941176,0.858823529411765}
\definecolor{color5}{rgb}{0.858823529411765,0.305882352941176,0.435294117647059}
\definecolor{color6}{rgb}{0.937254901960784,0.929411764705882,0.392156862745098}
\definecolor{color7}{rgb}{0.0901960784313725,0.486274509803922,0.0980392156862745}
\definecolor{color8}{rgb}{0.156862745098039,0.188235294117647,0.827450980392157}
\definecolor{color9}{rgb}{0.937254901960784,0.392156862745098,0.894117647058824}
\definecolor{color10}{rgb}{0.2,0.184313725490196,0.184313725490196}
\definecolor{color11}{rgb}{0.0156862745098039,0.803921568627451,0.976470588235294}

\begin{axis}[
tick align=outside,
tick pos=left,
x grid style={white!69.0196078431373!black},
xmajorgrids,
xmin=-0.0895, xmax=2.0995,
xtick style={color=black},
xtick={0,0.1,0.2,0.3,0.4,0.5,0.6,0.7,0.8,0.9,1,1.1,1.2,1.3,1.4,1.5,1.6,1.7,1.8,1.9,2},
xticklabels={0,,.2,,.4,,.6,,.8,,1,,20,,40,,60,,80,,100},
height=4.8cm,
width=6.5cm,
y grid style={white!69.0196078431373!black},
ymajorgrids,
ymin=0.2, ymax=1.05,
ytick style={color=black}
]
\addplot [semithick, color0, opacity=0.75]
table {%
0.01 0.56
0.02 0.55
0.03 0.52
0.04 0.7
0.05 0.61
0.06 0.5
0.07 0.59
0.08 0.59
0.09 0.66
0.1 0.47
0.11 0.62
0.12 0.61
0.13 0.65
0.14 0.6
0.15 0.61
0.16 0.63
0.17 0.62
0.18 0.69
0.19 0.72
0.2 0.62
0.21 0.75
0.22 0.75
0.23 0.72
0.24 0.79
0.25 0.73
0.26 0.75
0.27 0.74
0.28 0.73
0.29 0.69
0.3 0.77
0.31 0.78
0.32 0.77
0.33 0.74
0.34 0.82
0.35 0.72
0.36 0.74
0.37 0.82
0.38 0.87
0.39 0.83
0.4 0.76
0.41 0.81
0.42 0.86
0.43 0.79
0.44 0.79
0.45 0.83
0.46 0.87
0.47 0.88
0.48 0.84
0.49 0.82
0.5 0.88
0.51 0.86
0.52 0.81
0.53 0.81
0.54 0.91
0.55 0.85
0.56 0.9
0.57 0.85
0.58 0.9
0.59 0.92
0.6 0.93
0.61 0.96
0.62 0.92
0.63 0.88
0.64 0.89
0.65 0.86
0.66 0.89
0.67 0.94
0.68 0.96
0.69 0.93
0.7 0.92
0.71 0.96
0.72 0.92
0.73 0.93
0.74 0.92
0.75 0.97
0.76 0.96
0.77 0.95
0.78 0.92
0.79 0.92
0.8 0.96
0.81 0.93
0.82 0.98
0.83 0.91
0.84 0.97
0.85 0.96
0.86 0.96
0.87 0.95
0.88 0.96
0.89 0.96
0.9 0.95
0.91 0.95
0.92 0.97
0.93 1
0.94 0.95
0.95 0.95
0.96 0.95
0.97 1
0.98 0.94
0.99 0.99
1 0.96
1.02 1
1.03 0.93
1.04 0.92
1.05 0.86
1.06 0.83
1.07 0.83
1.08 0.81
1.09 0.74
1.1 0.77
1.11 0.72
1.12 0.61
1.13 0.71
1.14 0.69
1.15 0.65
1.16 0.62
1.17 0.74
1.18 0.61
1.19 0.64
1.2 0.62
1.21 0.65
1.22 0.58
1.23 0.62
1.24 0.61
1.25 0.59
1.26 0.61
1.27 0.62
1.28 0.66
1.29 0.61
1.3 0.56
1.31 0.6
1.32 0.59
1.33 0.58
1.34 0.61
1.35 0.49
1.36 0.5
1.37 0.61
1.38 0.65
1.39 0.62
1.4 0.5
1.41 0.59
1.42 0.58
1.43 0.5
1.44 0.5
1.45 0.41
1.46 0.62
1.47 0.56
1.48 0.59
1.49 0.51
1.5 0.55
1.51 0.55
1.52 0.61
1.53 0.64
1.54 0.43
1.55 0.48
1.56 0.57
1.57 0.66
1.58 0.52
1.59 0.53
1.6 0.56
1.61 0.55
1.62 0.62
1.63 0.55
1.64 0.52
1.65 0.63
1.66 0.56
1.67 0.49
1.68 0.58
1.69 0.5
1.7 0.63
1.71 0.65
1.72 0.46
1.73 0.58
1.74 0.57
1.75 0.56
1.76 0.56
1.77 0.51
1.78 0.64
1.79 0.52
1.8 0.49
1.81 0.59
1.82 0.56
1.83 0.6
1.84 0.48
1.85 0.54
1.86 0.57
1.87 0.54
1.88 0.53
1.89 0.52
1.9 0.56
1.91 0.57
1.92 0.52
1.93 0.54
1.94 0.53
1.95 0.59
1.96 0.52
1.97 0.58
1.98 0.57
1.99 0.54
2 0.52
};
\addplot [semithick, color1, opacity=0.75]
table {%
0.01 0.46
0.02 0.5
0.03 0.46
0.04 0.52
0.05 0.58
0.06 0.45
0.07 0.6
0.08 0.51
0.09 0.53
0.1 0.51
0.11 0.43
0.12 0.55
0.13 0.45
0.14 0.5
0.15 0.61
0.16 0.52
0.17 0.42
0.18 0.48
0.19 0.49
0.2 0.41
0.21 0.59
0.22 0.49
0.23 0.59
0.24 0.43
0.25 0.42
0.26 0.47
0.27 0.58
0.28 0.45
0.29 0.47
0.3 0.53
0.31 0.5
0.32 0.46
0.33 0.39
0.34 0.53
0.35 0.46
0.36 0.4
0.37 0.51
0.38 0.44
0.39 0.36
0.4 0.46
0.41 0.41
0.42 0.52
0.43 0.46
0.44 0.49
0.45 0.45
0.46 0.53
0.47 0.5
0.48 0.43
0.49 0.45
0.5 0.6
0.51 0.5
0.52 0.4
0.53 0.45
0.54 0.53
0.55 0.49
0.56 0.49
0.57 0.46
0.58 0.46
0.59 0.53
0.6 0.55
0.61 0.56
0.62 0.59
0.63 0.49
0.64 0.55
0.65 0.43
0.66 0.52
0.67 0.51
0.68 0.62
0.69 0.59
0.7 0.46
0.71 0.56
0.72 0.53
0.73 0.67
0.74 0.58
0.75 0.54
0.76 0.58
0.77 0.6
0.78 0.6
0.79 0.58
0.8 0.58
0.81 0.68
0.82 0.67
0.83 0.71
0.84 0.63
0.85 0.66
0.86 0.54
0.87 0.6
0.88 0.63
0.89 0.64
0.9 0.64
0.91 0.71
0.92 0.66
0.93 0.75
0.94 0.59
0.95 0.73
0.96 0.67
0.97 0.7
0.98 0.63
0.99 0.77
1 0.75
1.02 0.94
1.03 0.92
1.04 0.95
1.05 0.92
1.06 0.93
1.07 0.93
1.08 0.86
1.09 0.88
1.1 0.78
1.11 0.8
1.12 0.8
1.13 0.77
1.14 0.81
1.15 0.76
1.16 0.74
1.17 0.74
1.18 0.66
1.19 0.67
1.2 0.71
1.21 0.64
1.22 0.54
1.23 0.64
1.24 0.7
1.25 0.59
1.26 0.68
1.27 0.63
1.28 0.67
1.29 0.56
1.3 0.61
1.31 0.54
1.32 0.6
1.33 0.65
1.34 0.6
1.35 0.56
1.36 0.61
1.37 0.5
1.38 0.55
1.39 0.57
1.4 0.46
1.41 0.65
1.42 0.59
1.43 0.53
1.44 0.52
1.45 0.47
1.46 0.52
1.47 0.54
1.48 0.52
1.49 0.54
1.5 0.55
1.51 0.56
1.52 0.51
1.53 0.56
1.54 0.53
1.55 0.45
1.56 0.57
1.57 0.5
1.58 0.56
1.59 0.45
1.6 0.57
1.61 0.52
1.62 0.54
1.63 0.5
1.64 0.46
1.65 0.45
1.66 0.55
1.67 0.47
1.68 0.56
1.69 0.41
1.7 0.49
1.71 0.52
1.72 0.54
1.73 0.39
1.74 0.53
1.75 0.48
1.76 0.5
1.77 0.48
1.78 0.41
1.79 0.55
1.8 0.49
1.81 0.49
1.82 0.46
1.83 0.51
1.84 0.56
1.85 0.48
1.86 0.44
1.87 0.39
1.88 0.5
1.89 0.53
1.9 0.52
1.91 0.44
1.92 0.58
1.93 0.41
1.94 0.5
1.95 0.44
1.96 0.48
1.97 0.52
1.98 0.4
1.99 0.48
2 0.5
};
\addplot [semithick, color2, opacity=0.75]
table {%
0.01 0.5
0.02 0.47
0.03 0.5
0.04 0.54
0.05 0.51
0.06 0.49
0.07 0.51
0.08 0.53
0.09 0.43
0.1 0.52
0.11 0.38
0.12 0.47
0.13 0.54
0.14 0.46
0.15 0.51
0.16 0.49
0.17 0.39
0.18 0.48
0.19 0.47
0.2 0.41
0.21 0.59
0.22 0.46
0.23 0.61
0.24 0.43
0.25 0.43
0.26 0.47
0.27 0.56
0.28 0.45
0.29 0.46
0.3 0.5
0.31 0.49
0.32 0.46
0.33 0.38
0.34 0.52
0.35 0.43
0.36 0.4
0.37 0.51
0.38 0.43
0.39 0.34
0.4 0.45
0.41 0.39
0.42 0.5
0.43 0.46
0.44 0.48
0.45 0.42
0.46 0.52
0.47 0.51
0.48 0.44
0.49 0.46
0.5 0.58
0.51 0.47
0.52 0.4
0.53 0.48
0.54 0.53
0.55 0.48
0.56 0.49
0.57 0.48
0.58 0.44
0.59 0.49
0.6 0.55
0.61 0.52
0.62 0.57
0.63 0.48
0.64 0.54
0.65 0.39
0.66 0.52
0.67 0.51
0.68 0.6
0.69 0.56
0.7 0.43
0.71 0.48
0.72 0.49
0.73 0.63
0.74 0.57
0.75 0.52
0.76 0.55
0.77 0.55
0.78 0.56
0.79 0.56
0.8 0.56
0.81 0.66
0.82 0.63
0.83 0.69
0.84 0.61
0.85 0.66
0.86 0.54
0.87 0.59
0.88 0.64
0.89 0.63
0.9 0.62
0.91 0.7
0.92 0.65
0.93 0.74
0.94 0.58
0.95 0.69
0.96 0.67
0.97 0.7
0.98 0.62
0.99 0.76
1 0.72
1.02 0.95
1.03 0.92
1.04 0.95
1.05 0.94
1.06 0.92
1.07 0.93
1.08 0.86
1.09 0.87
1.1 0.78
1.11 0.81
1.12 0.8
1.13 0.76
1.14 0.81
1.15 0.77
1.16 0.74
1.17 0.75
1.18 0.66
1.19 0.66
1.2 0.72
1.21 0.65
1.22 0.54
1.23 0.64
1.24 0.72
1.25 0.59
1.26 0.68
1.27 0.63
1.28 0.67
1.29 0.57
1.3 0.62
1.31 0.54
1.32 0.6
1.33 0.66
1.34 0.6
1.35 0.58
1.36 0.6
1.37 0.51
1.38 0.57
1.39 0.56
1.4 0.48
1.41 0.65
1.42 0.6
1.43 0.53
1.44 0.52
1.45 0.5
1.46 0.53
1.47 0.55
1.48 0.52
1.49 0.53
1.5 0.55
1.51 0.57
1.52 0.48
1.53 0.53
1.54 0.51
1.55 0.48
1.56 0.56
1.57 0.5
1.58 0.56
1.59 0.44
1.6 0.57
1.61 0.51
1.62 0.55
1.63 0.49
1.64 0.47
1.65 0.47
1.66 0.57
1.67 0.48
1.68 0.57
1.69 0.43
1.7 0.48
1.71 0.52
1.72 0.52
1.73 0.38
1.74 0.53
1.75 0.5
1.76 0.53
1.77 0.46
1.78 0.41
1.79 0.56
1.8 0.49
1.81 0.5
1.82 0.46
1.83 0.52
1.84 0.55
1.85 0.49
1.86 0.45
1.87 0.38
1.88 0.5
1.89 0.49
1.9 0.53
1.91 0.44
1.92 0.58
1.93 0.41
1.94 0.51
1.95 0.43
1.96 0.48
1.97 0.55
1.98 0.39
1.99 0.48
2 0.47
};
\addplot [semithick, color3, opacity=0.75]
table {%
0.01 0.58
0.02 0.64
0.03 0.56
0.04 0.7
0.05 0.59
0.06 0.54
0.07 0.61
0.08 0.58
0.09 0.68
0.1 0.55
0.11 0.59
0.12 0.67
0.13 0.67
0.14 0.55
0.15 0.63
0.16 0.65
0.17 0.62
0.18 0.68
0.19 0.67
0.2 0.68
0.21 0.73
0.22 0.68
0.23 0.71
0.24 0.77
0.25 0.73
0.26 0.74
0.27 0.78
0.28 0.7
0.29 0.68
0.3 0.75
0.31 0.75
0.32 0.75
0.33 0.69
0.34 0.78
0.35 0.73
0.36 0.78
0.37 0.79
0.38 0.86
0.39 0.8
0.4 0.76
0.41 0.85
0.42 0.89
0.43 0.78
0.44 0.8
0.45 0.83
0.46 0.86
0.47 0.86
0.48 0.81
0.49 0.83
0.5 0.82
0.51 0.86
0.52 0.79
0.53 0.79
0.54 0.83
0.55 0.8
0.56 0.81
0.57 0.85
0.58 0.82
0.59 0.89
0.6 0.9
0.61 0.92
0.62 0.89
0.63 0.83
0.64 0.87
0.65 0.86
0.66 0.83
0.67 0.88
0.68 0.91
0.69 0.89
0.7 0.79
0.71 0.87
0.72 0.87
0.73 0.89
0.74 0.84
0.75 0.92
0.76 0.87
0.77 0.87
0.78 0.83
0.79 0.88
0.8 0.86
0.81 0.85
0.82 0.91
0.83 0.87
0.84 0.92
0.85 0.85
0.86 0.83
0.87 0.92
0.88 0.9
0.89 0.9
0.9 0.86
0.91 0.89
0.92 0.88
0.93 0.92
0.94 0.87
0.95 0.85
0.96 0.87
0.97 0.82
0.98 0.79
0.99 0.93
1 0.9
1.02 0.88
1.03 0.83
1.04 0.86
1.05 0.82
1.06 0.8
1.07 0.83
1.08 0.79
1.09 0.69
1.1 0.7
1.11 0.7
1.12 0.68
1.13 0.7
1.14 0.67
1.15 0.64
1.16 0.7
1.17 0.69
1.18 0.64
1.19 0.62
1.2 0.65
1.21 0.61
1.22 0.57
1.23 0.61
1.24 0.66
1.25 0.58
1.26 0.62
1.27 0.52
1.28 0.66
1.29 0.53
1.3 0.63
1.31 0.62
1.32 0.59
1.33 0.6
1.34 0.58
1.35 0.56
1.36 0.54
1.37 0.59
1.38 0.61
1.39 0.69
1.4 0.38
1.41 0.59
1.42 0.55
1.43 0.48
1.44 0.48
1.45 0.38
1.46 0.61
1.47 0.58
1.48 0.62
1.49 0.54
1.5 0.51
1.51 0.55
1.52 0.62
1.53 0.63
1.54 0.51
1.55 0.5
1.56 0.56
1.57 0.62
1.58 0.56
1.59 0.53
1.6 0.55
1.61 0.55
1.62 0.59
1.63 0.64
1.64 0.54
1.65 0.62
1.66 0.58
1.67 0.51
1.68 0.64
1.69 0.55
1.7 0.64
1.71 0.64
1.72 0.52
1.73 0.58
1.74 0.56
1.75 0.53
1.76 0.57
1.77 0.59
1.78 0.62
1.79 0.54
1.8 0.5
1.81 0.6
1.82 0.58
1.83 0.56
1.84 0.48
1.85 0.58
1.86 0.6
1.87 0.5
1.88 0.58
1.89 0.48
1.9 0.47
1.91 0.52
1.92 0.56
1.93 0.48
1.94 0.54
1.95 0.57
1.96 0.51
1.97 0.57
1.98 0.55
1.99 0.53
2 0.52
};
\addplot [semithick, color4, opacity=0.75]
table {%
0.01 0.58
0.02 0.62
0.03 0.56
0.04 0.7
0.05 0.59
0.06 0.55
0.07 0.61
0.08 0.57
0.09 0.67
0.1 0.53
0.11 0.61
0.12 0.67
0.13 0.67
0.14 0.57
0.15 0.63
0.16 0.65
0.17 0.6
0.18 0.7
0.19 0.68
0.2 0.68
0.21 0.76
0.22 0.68
0.23 0.72
0.24 0.76
0.25 0.75
0.26 0.74
0.27 0.8
0.28 0.71
0.29 0.68
0.3 0.76
0.31 0.75
0.32 0.76
0.33 0.7
0.34 0.79
0.35 0.73
0.36 0.79
0.37 0.8
0.38 0.87
0.39 0.8
0.4 0.77
0.41 0.87
0.42 0.9
0.43 0.78
0.44 0.83
0.45 0.85
0.46 0.86
0.47 0.86
0.48 0.85
0.49 0.85
0.5 0.84
0.51 0.88
0.52 0.8
0.53 0.81
0.54 0.84
0.55 0.8
0.56 0.82
0.57 0.85
0.58 0.82
0.59 0.9
0.6 0.91
0.61 0.93
0.62 0.9
0.63 0.82
0.64 0.88
0.65 0.88
0.66 0.84
0.67 0.9
0.68 0.92
0.69 0.91
0.7 0.84
0.71 0.87
0.72 0.87
0.73 0.89
0.74 0.84
0.75 0.94
0.76 0.92
0.77 0.89
0.78 0.87
0.79 0.89
0.8 0.88
0.81 0.88
0.82 0.93
0.83 0.88
0.84 0.93
0.85 0.9
0.86 0.85
0.87 0.92
0.88 0.9
0.89 0.93
0.9 0.89
0.91 0.89
0.92 0.89
0.93 0.92
0.94 0.88
0.95 0.88
0.96 0.89
0.97 0.84
0.98 0.83
0.99 0.93
1 0.91
1.02 0.9
1.03 0.85
1.04 0.9
1.05 0.84
1.06 0.81
1.07 0.86
1.08 0.8
1.09 0.69
1.1 0.73
1.11 0.7
1.12 0.7
1.13 0.71
1.14 0.68
1.15 0.64
1.16 0.72
1.17 0.7
1.18 0.64
1.19 0.63
1.2 0.65
1.21 0.61
1.22 0.56
1.23 0.61
1.24 0.67
1.25 0.59
1.26 0.62
1.27 0.54
1.28 0.66
1.29 0.53
1.3 0.61
1.31 0.61
1.32 0.59
1.33 0.6
1.34 0.58
1.35 0.56
1.36 0.57
1.37 0.58
1.38 0.61
1.39 0.69
1.4 0.39
1.41 0.59
1.42 0.56
1.43 0.48
1.44 0.48
1.45 0.38
1.46 0.59
1.47 0.56
1.48 0.63
1.49 0.52
1.5 0.52
1.51 0.56
1.52 0.6
1.53 0.64
1.54 0.5
1.55 0.49
1.56 0.55
1.57 0.62
1.58 0.55
1.59 0.54
1.6 0.54
1.61 0.57
1.62 0.59
1.63 0.62
1.64 0.54
1.65 0.61
1.66 0.57
1.67 0.5
1.68 0.65
1.69 0.58
1.7 0.63
1.71 0.64
1.72 0.52
1.73 0.58
1.74 0.55
1.75 0.51
1.76 0.56
1.77 0.59
1.78 0.62
1.79 0.55
1.8 0.5
1.81 0.6
1.82 0.57
1.83 0.54
1.84 0.48
1.85 0.58
1.86 0.58
1.87 0.5
1.88 0.58
1.89 0.49
1.9 0.47
1.91 0.51
1.92 0.55
1.93 0.48
1.94 0.55
1.95 0.58
1.96 0.51
1.97 0.6
1.98 0.56
1.99 0.51
2 0.53
};
\addplot [semithick, color5, opacity=0.75]
table {%
0.01 0.43
0.02 0.54
0.03 0.49
0.04 0.63
0.05 0.54
0.06 0.55
0.07 0.64
0.08 0.57
0.09 0.65
0.1 0.55
0.11 0.63
0.12 0.64
0.13 0.64
0.14 0.54
0.15 0.65
0.16 0.62
0.17 0.61
0.18 0.67
0.19 0.71
0.2 0.64
0.21 0.65
0.22 0.65
0.23 0.72
0.24 0.74
0.25 0.68
0.26 0.72
0.27 0.72
0.28 0.68
0.29 0.67
0.3 0.73
0.31 0.74
0.32 0.73
0.33 0.69
0.34 0.72
0.35 0.69
0.36 0.72
0.37 0.74
0.38 0.82
0.39 0.78
0.4 0.72
0.41 0.83
0.42 0.86
0.43 0.71
0.44 0.79
0.45 0.78
0.46 0.79
0.47 0.77
0.48 0.76
0.49 0.78
0.5 0.78
0.51 0.78
0.52 0.76
0.53 0.75
0.54 0.78
0.55 0.75
0.56 0.77
0.57 0.8
0.58 0.79
0.59 0.87
0.6 0.87
0.61 0.88
0.62 0.83
0.63 0.77
0.64 0.84
0.65 0.79
0.66 0.78
0.67 0.85
0.68 0.88
0.69 0.88
0.7 0.74
0.71 0.82
0.72 0.76
0.73 0.87
0.74 0.82
0.75 0.89
0.76 0.84
0.77 0.86
0.78 0.79
0.79 0.84
0.8 0.81
0.81 0.84
0.82 0.85
0.83 0.85
0.84 0.85
0.85 0.79
0.86 0.76
0.87 0.89
0.88 0.83
0.89 0.88
0.9 0.81
0.91 0.85
0.92 0.86
0.93 0.87
0.94 0.81
0.95 0.79
0.96 0.83
0.97 0.81
0.98 0.78
0.99 0.85
1 0.86
1.02 0.85
1.03 0.8
1.04 0.87
1.05 0.75
1.06 0.77
1.07 0.81
1.08 0.73
1.09 0.66
1.1 0.69
1.11 0.71
1.12 0.68
1.13 0.7
1.14 0.67
1.15 0.66
1.16 0.63
1.17 0.67
1.18 0.57
1.19 0.65
1.2 0.6
1.21 0.6
1.22 0.55
1.23 0.63
1.24 0.7
1.25 0.56
1.26 0.61
1.27 0.56
1.28 0.72
1.29 0.51
1.3 0.67
1.31 0.64
1.32 0.64
1.33 0.55
1.34 0.57
1.35 0.54
1.36 0.56
1.37 0.6
1.38 0.54
1.39 0.6
1.4 0.42
1.41 0.58
1.42 0.5
1.43 0.5
1.44 0.51
1.45 0.48
1.46 0.65
1.47 0.49
1.48 0.62
1.49 0.55
1.5 0.58
1.51 0.55
1.52 0.64
1.53 0.57
1.54 0.53
1.55 0.59
1.56 0.51
1.57 0.6
1.58 0.56
1.59 0.57
1.6 0.6
1.61 0.53
1.62 0.64
1.63 0.6
1.64 0.54
1.65 0.59
1.66 0.53
1.67 0.55
1.68 0.68
1.69 0.48
1.7 0.59
1.71 0.54
1.72 0.5
1.73 0.58
1.74 0.57
1.75 0.53
1.76 0.51
1.77 0.56
1.78 0.58
1.79 0.52
1.8 0.53
1.81 0.58
1.82 0.52
1.83 0.52
1.84 0.45
1.85 0.55
1.86 0.56
1.87 0.52
1.88 0.54
1.89 0.55
1.9 0.48
1.91 0.5
1.92 0.52
1.93 0.43
1.94 0.46
1.95 0.59
1.96 0.5
1.97 0.51
1.98 0.6
1.99 0.52
2 0.57
};
\addplot [semithick, color6, opacity=0.75, dashed]
table {%
0.01 0.44
0.02 0.41
0.03 0.5
0.04 0.52
0.05 0.46
0.06 0.49
0.07 0.58
0.08 0.53
0.09 0.61
0.1 0.54
0.11 0.63
0.12 0.56
0.13 0.48
0.14 0.53
0.15 0.56
0.16 0.55
0.17 0.58
0.18 0.63
0.19 0.55
0.2 0.64
0.21 0.64
0.22 0.62
0.23 0.6
0.24 0.66
0.25 0.65
0.26 0.67
0.27 0.77
0.28 0.69
0.29 0.7
0.3 0.72
0.31 0.58
0.32 0.67
0.33 0.76
0.34 0.74
0.35 0.7
0.36 0.79
0.37 0.71
0.38 0.72
0.39 0.82
0.4 0.7
0.41 0.79
0.42 0.77
0.43 0.84
0.44 0.7
0.45 0.74
0.46 0.79
0.47 0.79
0.48 0.83
0.49 0.72
0.5 0.79
0.51 0.75
0.52 0.74
0.53 0.85
0.54 0.9
0.55 0.84
0.56 0.72
0.57 0.85
0.58 0.85
0.59 0.86
0.6 0.81
0.61 0.89
0.62 0.78
0.63 0.81
0.64 0.78
0.65 0.82
0.66 0.77
0.67 0.85
0.68 0.88
0.69 0.75
0.7 0.84
0.71 0.86
0.72 0.75
0.73 0.86
0.74 0.89
0.75 0.86
0.76 0.92
0.77 0.87
0.78 0.91
0.79 0.74
0.8 0.88
0.81 0.89
0.82 0.89
0.83 0.9
0.84 0.78
0.85 0.92
0.86 0.87
0.87 0.87
0.88 0.94
0.89 0.87
0.9 0.85
0.91 0.84
0.92 0.88
0.93 0.9
0.94 0.88
0.95 0.81
0.96 0.89
0.97 0.86
0.98 0.85
0.99 0.89
1 0.87
1.02 0.89
1.03 0.86
1.04 0.89
1.05 0.84
1.06 0.83
1.07 0.74
1.08 0.79
1.09 0.72
1.1 0.74
1.11 0.69
1.12 0.71
1.13 0.61
1.14 0.64
1.15 0.59
1.16 0.62
1.17 0.7
1.18 0.63
1.19 0.65
1.2 0.65
1.21 0.54
1.22 0.59
1.23 0.54
1.24 0.49
1.25 0.58
1.26 0.59
1.27 0.64
1.28 0.54
1.29 0.63
1.3 0.54
1.31 0.5
1.32 0.65
1.33 0.52
1.34 0.51
1.35 0.5
1.36 0.5
1.37 0.5
1.38 0.56
1.39 0.48
1.4 0.55
1.41 0.55
1.42 0.59
1.43 0.55
1.44 0.56
1.45 0.41
1.46 0.49
1.47 0.51
1.48 0.5
1.49 0.48
1.5 0.48
1.51 0.49
1.52 0.47
1.53 0.48
1.54 0.5
1.55 0.44
1.56 0.55
1.57 0.56
1.58 0.48
1.59 0.41
1.6 0.51
1.61 0.47
1.62 0.56
1.63 0.5
1.64 0.49
1.65 0.43
1.66 0.52
1.67 0.48
1.68 0.49
1.69 0.46
1.7 0.58
1.71 0.48
1.72 0.47
1.73 0.47
1.74 0.55
1.75 0.46
1.76 0.51
1.77 0.51
1.78 0.47
1.79 0.55
1.8 0.46
1.81 0.44
1.82 0.46
1.83 0.48
1.84 0.4
1.85 0.48
1.86 0.53
1.87 0.4
1.88 0.45
1.89 0.43
1.9 0.47
1.91 0.43
1.92 0.54
1.93 0.45
1.94 0.41
1.95 0.39
1.96 0.46
1.97 0.51
1.98 0.45
1.99 0.39
2 0.46
};
\addplot [semithick, color7, opacity=0.75, dashed]
table {%
0.01 0.44
0.02 0.41
0.03 0.5
0.04 0.52
0.05 0.46
0.06 0.49
0.07 0.58
0.08 0.53
0.09 0.61
0.1 0.54
0.11 0.63
0.12 0.56
0.13 0.48
0.14 0.53
0.15 0.56
0.16 0.55
0.17 0.58
0.18 0.63
0.19 0.55
0.2 0.64
0.21 0.64
0.22 0.62
0.23 0.6
0.24 0.66
0.25 0.65
0.26 0.67
0.27 0.77
0.28 0.69
0.29 0.7
0.3 0.72
0.31 0.58
0.32 0.67
0.33 0.76
0.34 0.74
0.35 0.7
0.36 0.79
0.37 0.71
0.38 0.72
0.39 0.82
0.4 0.7
0.41 0.79
0.42 0.77
0.43 0.84
0.44 0.7
0.45 0.74
0.46 0.79
0.47 0.79
0.48 0.83
0.49 0.72
0.5 0.79
0.51 0.75
0.52 0.74
0.53 0.85
0.54 0.9
0.55 0.84
0.56 0.72
0.57 0.85
0.58 0.85
0.59 0.86
0.6 0.81
0.61 0.89
0.62 0.78
0.63 0.81
0.64 0.78
0.65 0.82
0.66 0.77
0.67 0.85
0.68 0.88
0.69 0.75
0.7 0.84
0.71 0.86
0.72 0.75
0.73 0.86
0.74 0.89
0.75 0.86
0.76 0.92
0.77 0.87
0.78 0.91
0.79 0.74
0.8 0.88
0.81 0.89
0.82 0.89
0.83 0.9
0.84 0.78
0.85 0.92
0.86 0.87
0.87 0.87
0.88 0.94
0.89 0.87
0.9 0.85
0.91 0.84
0.92 0.88
0.93 0.9
0.94 0.88
0.95 0.81
0.96 0.89
0.97 0.86
0.98 0.85
0.99 0.89
1 0.87
1.02 0.89
1.03 0.86
1.04 0.89
1.05 0.84
1.06 0.83
1.07 0.74
1.08 0.79
1.09 0.72
1.1 0.74
1.11 0.69
1.12 0.71
1.13 0.61
1.14 0.64
1.15 0.59
1.16 0.62
1.17 0.7
1.18 0.63
1.19 0.65
1.2 0.65
1.21 0.54
1.22 0.59
1.23 0.54
1.24 0.49
1.25 0.58
1.26 0.59
1.27 0.64
1.28 0.54
1.29 0.63
1.3 0.54
1.31 0.5
1.32 0.65
1.33 0.52
1.34 0.51
1.35 0.5
1.36 0.5
1.37 0.5
1.38 0.56
1.39 0.48
1.4 0.55
1.41 0.55
1.42 0.59
1.43 0.55
1.44 0.56
1.45 0.41
1.46 0.49
1.47 0.51
1.48 0.5
1.49 0.48
1.5 0.48
1.51 0.49
1.52 0.47
1.53 0.48
1.54 0.5
1.55 0.44
1.56 0.55
1.57 0.56
1.58 0.48
1.59 0.41
1.6 0.51
1.61 0.47
1.62 0.56
1.63 0.5
1.64 0.49
1.65 0.43
1.66 0.52
1.67 0.48
1.68 0.49
1.69 0.46
1.7 0.58
1.71 0.48
1.72 0.47
1.73 0.47
1.74 0.55
1.75 0.46
1.76 0.51
1.77 0.51
1.78 0.47
1.79 0.55
1.8 0.46
1.81 0.44
1.82 0.46
1.83 0.48
1.84 0.4
1.85 0.48
1.86 0.53
1.87 0.4
1.88 0.45
1.89 0.43
1.9 0.47
1.91 0.43
1.92 0.54
1.93 0.45
1.94 0.41
1.95 0.39
1.96 0.46
1.97 0.51
1.98 0.45
1.99 0.39
2 0.46
};
\addplot [semithick, color8, opacity=0.75, dashed]
table {%
0.01 0.61
0.02 0.48
0.03 0.42
0.04 0.56
0.05 0.41
0.06 0.59
0.07 0.59
0.08 0.5
0.09 0.53
0.1 0.59
0.11 0.56
0.12 0.54
0.13 0.53
0.14 0.57
0.15 0.53
0.16 0.57
0.17 0.56
0.18 0.58
0.19 0.59
0.2 0.63
0.21 0.64
0.22 0.74
0.23 0.63
0.24 0.65
0.25 0.66
0.26 0.79
0.27 0.76
0.28 0.71
0.29 0.69
0.3 0.74
0.31 0.65
0.32 0.76
0.33 0.74
0.34 0.74
0.35 0.74
0.36 0.8
0.37 0.73
0.38 0.78
0.39 0.8
0.4 0.76
0.41 0.89
0.42 0.82
0.43 0.85
0.44 0.78
0.45 0.81
0.46 0.84
0.47 0.82
0.48 0.86
0.49 0.76
0.5 0.87
0.51 0.8
0.52 0.81
0.53 0.84
0.54 0.91
0.55 0.81
0.56 0.82
0.57 0.88
0.58 0.85
0.59 0.88
0.6 0.9
0.61 0.87
0.62 0.86
0.63 0.9
0.64 0.82
0.65 0.91
0.66 0.86
0.67 0.88
0.68 0.92
0.69 0.89
0.7 0.93
0.71 0.89
0.72 0.88
0.73 0.86
0.74 0.93
0.75 0.89
0.76 0.94
0.77 0.87
0.78 0.89
0.79 0.88
0.8 0.92
0.81 0.89
0.82 0.88
0.83 0.91
0.84 0.93
0.85 0.92
0.86 0.88
0.87 0.91
0.88 0.95
0.89 0.98
0.9 0.91
0.91 0.91
0.92 0.92
0.93 0.96
0.94 0.95
0.95 0.9
0.96 0.97
0.97 0.95
0.98 0.9
0.99 0.96
1 0.9
1.02 0.95
1.03 0.94
1.04 0.86
1.05 0.86
1.06 0.86
1.07 0.87
1.08 0.74
1.09 0.78
1.1 0.81
1.11 0.63
1.12 0.7
1.13 0.73
1.14 0.65
1.15 0.58
1.16 0.64
1.17 0.73
1.18 0.63
1.19 0.58
1.2 0.7
1.21 0.5
1.22 0.61
1.23 0.58
1.24 0.47
1.25 0.56
1.26 0.53
1.27 0.54
1.28 0.56
1.29 0.57
1.3 0.53
1.31 0.55
1.32 0.59
1.33 0.55
1.34 0.52
1.35 0.53
1.36 0.57
1.37 0.5
1.38 0.56
1.39 0.47
1.4 0.45
1.41 0.48
1.42 0.59
1.43 0.45
1.44 0.58
1.45 0.55
1.46 0.54
1.47 0.52
1.48 0.59
1.49 0.48
1.5 0.56
1.51 0.47
1.52 0.52
1.53 0.58
1.54 0.48
1.55 0.45
1.56 0.53
1.57 0.54
1.58 0.47
1.59 0.34
1.6 0.44
1.61 0.51
1.62 0.52
1.63 0.51
1.64 0.39
1.65 0.43
1.66 0.58
1.67 0.45
1.68 0.54
1.69 0.52
1.7 0.55
1.71 0.49
1.72 0.49
1.73 0.43
1.74 0.52
1.75 0.5
1.76 0.53
1.77 0.45
1.78 0.47
1.79 0.51
1.8 0.51
1.81 0.46
1.82 0.5
1.83 0.44
1.84 0.51
1.85 0.51
1.86 0.48
1.87 0.41
1.88 0.53
1.89 0.46
1.9 0.49
1.91 0.38
1.92 0.52
1.93 0.48
1.94 0.5
1.95 0.51
1.96 0.44
1.97 0.58
1.98 0.52
1.99 0.49
2 0.38
};
\addplot [semithick, color9, opacity=0.75, dashed]
table {%
0.01 0.54
0.02 0.47
0.03 0.48
0.04 0.61
0.05 0.48
0.06 0.52
0.07 0.57
0.08 0.53
0.09 0.55
0.1 0.57
0.11 0.52
0.12 0.59
0.13 0.6
0.14 0.63
0.15 0.65
0.16 0.69
0.17 0.55
0.18 0.65
0.19 0.59
0.2 0.67
0.21 0.7
0.22 0.73
0.23 0.7
0.24 0.69
0.25 0.65
0.26 0.76
0.27 0.77
0.28 0.68
0.29 0.74
0.3 0.73
0.31 0.76
0.32 0.79
0.33 0.79
0.34 0.8
0.35 0.74
0.36 0.72
0.37 0.79
0.38 0.87
0.39 0.77
0.4 0.74
0.41 0.82
0.42 0.78
0.43 0.83
0.44 0.83
0.45 0.79
0.46 0.9
0.47 0.8
0.48 0.84
0.49 0.85
0.5 0.83
0.51 0.86
0.52 0.9
0.53 0.88
0.54 0.92
0.55 0.82
0.56 0.87
0.57 0.86
0.58 0.9
0.59 0.93
0.6 0.92
0.61 0.94
0.62 0.91
0.63 0.91
0.64 0.94
0.65 0.94
0.66 0.85
0.67 0.93
0.68 0.96
0.69 0.95
0.7 0.96
0.71 0.96
0.72 0.95
0.73 0.95
0.74 0.95
0.75 0.99
0.76 0.95
0.77 0.93
0.78 0.97
0.79 0.95
0.8 0.94
0.81 0.96
0.82 0.98
0.83 0.98
0.84 0.94
0.85 0.98
0.86 0.94
0.87 0.95
0.88 0.96
0.89 0.97
0.9 0.95
0.91 0.92
0.92 0.96
0.93 1
0.94 0.97
0.95 0.97
0.96 0.94
0.97 0.94
0.98 0.99
0.99 0.96
1 0.98
1.02 0.99
1.03 0.94
1.04 0.92
1.05 0.94
1.06 0.83
1.07 0.86
1.08 0.8
1.09 0.66
1.1 0.68
1.11 0.76
1.12 0.63
1.13 0.7
1.14 0.71
1.15 0.67
1.16 0.68
1.17 0.62
1.18 0.68
1.19 0.64
1.2 0.67
1.21 0.69
1.22 0.6
1.23 0.56
1.24 0.7
1.25 0.54
1.26 0.59
1.27 0.49
1.28 0.49
1.29 0.61
1.3 0.53
1.31 0.55
1.32 0.67
1.33 0.58
1.34 0.49
1.35 0.57
1.36 0.59
1.37 0.53
1.38 0.63
1.39 0.52
1.4 0.5
1.41 0.55
1.42 0.49
1.43 0.43
1.44 0.53
1.45 0.45
1.46 0.55
1.47 0.49
1.48 0.62
1.49 0.52
1.5 0.43
1.51 0.48
1.52 0.51
1.53 0.53
1.54 0.46
1.55 0.51
1.56 0.45
1.57 0.56
1.58 0.55
1.59 0.43
1.6 0.51
1.61 0.55
1.62 0.54
1.63 0.5
1.64 0.52
1.65 0.49
1.66 0.46
1.67 0.41
1.68 0.53
1.69 0.52
1.7 0.51
1.71 0.53
1.72 0.49
1.73 0.53
1.74 0.43
1.75 0.46
1.76 0.57
1.77 0.46
1.78 0.58
1.79 0.47
1.8 0.41
1.81 0.5
1.82 0.5
1.83 0.53
1.84 0.45
1.85 0.53
1.86 0.49
1.87 0.44
1.88 0.45
1.89 0.48
1.9 0.54
1.91 0.49
1.92 0.48
1.93 0.4
1.94 0.48
1.95 0.39
1.96 0.44
1.97 0.5
1.98 0.44
1.99 0.52
2 0.5
};
\addplot [semithick, color10, opacity=0.75, dashed]
table {%
0.01 0.51
0.02 0.51
0.03 0.46
0.04 0.65
0.05 0.57
0.06 0.47
0.07 0.64
0.08 0.6
0.09 0.6
0.1 0.62
0.11 0.62
0.12 0.71
0.13 0.7
0.14 0.62
0.15 0.73
0.16 0.69
0.17 0.62
0.18 0.74
0.19 0.74
0.2 0.78
0.21 0.8
0.22 0.79
0.23 0.85
0.24 0.84
0.25 0.83
0.26 0.84
0.27 0.84
0.28 0.78
0.29 0.91
0.3 0.9
0.31 0.87
0.32 0.87
0.33 0.86
0.34 0.91
0.35 0.87
0.36 0.9
0.37 0.91
0.38 0.92
0.39 0.9
0.4 0.88
0.41 0.97
0.42 0.93
0.43 0.95
0.44 0.99
0.45 0.96
0.46 0.97
0.47 0.95
0.48 0.98
0.49 0.97
0.5 0.95
0.51 0.98
0.52 0.95
0.53 0.96
0.54 0.98
0.55 0.93
0.56 0.96
0.57 0.97
0.58 0.99
0.59 0.99
0.6 0.99
0.61 0.99
0.62 0.96
0.63 0.99
0.64 0.97
0.65 0.98
0.66 0.95
0.67 1
0.68 0.99
0.69 0.98
0.7 1
0.71 1
0.72 0.99
0.73 1
0.74 0.99
0.75 1
0.76 0.99
0.77 0.99
0.78 1
0.79 0.99
0.8 1
0.81 1
0.82 1
0.83 1
0.84 0.99
0.85 0.99
0.86 1
0.87 0.99
0.88 0.99
0.89 0.99
0.9 1
0.91 0.98
0.92 1
0.93 1
0.94 1
0.95 1
0.96 1
0.97 1
0.98 1
0.99 0.99
1 0.99
1.02 1
1.03 0.99
1.04 1
1.05 0.97
1.06 0.98
1.07 0.98
1.08 0.91
1.09 0.84
1.1 0.82
1.11 0.85
1.12 0.79
1.13 0.82
1.14 0.77
1.15 0.73
1.16 0.75
1.17 0.74
1.18 0.82
1.19 0.73
1.2 0.73
1.21 0.73
1.22 0.65
1.23 0.64
1.24 0.76
1.25 0.66
1.26 0.69
1.27 0.55
1.28 0.62
1.29 0.65
1.3 0.55
1.31 0.64
1.32 0.7
1.33 0.64
1.34 0.54
1.35 0.57
1.36 0.62
1.37 0.59
1.38 0.59
1.39 0.58
1.4 0.44
1.41 0.54
1.42 0.59
1.43 0.47
1.44 0.54
1.45 0.45
1.46 0.55
1.47 0.47
1.48 0.64
1.49 0.49
1.5 0.5
1.51 0.57
1.52 0.58
1.53 0.54
1.54 0.38
1.55 0.52
1.56 0.44
1.57 0.51
1.58 0.52
1.59 0.5
1.6 0.52
1.61 0.56
1.62 0.5
1.63 0.46
1.64 0.45
1.65 0.46
1.66 0.49
1.67 0.38
1.68 0.61
1.69 0.46
1.7 0.48
1.71 0.52
1.72 0.45
1.73 0.54
1.74 0.42
1.75 0.49
1.76 0.53
1.77 0.46
1.78 0.51
1.79 0.42
1.8 0.47
1.81 0.51
1.82 0.44
1.83 0.53
1.84 0.43
1.85 0.45
1.86 0.42
1.87 0.42
1.88 0.43
1.89 0.43
1.9 0.51
1.91 0.49
1.92 0.42
1.93 0.45
1.94 0.42
1.95 0.51
1.96 0.46
1.97 0.49
1.98 0.43
1.99 0.4
2 0.42
};
\addplot [semithick, color11, opacity=0.75, dashed]
table {%
0.01 0.56
0.02 0.47
0.03 0.63
0.04 0.68
0.05 0.6
0.06 0.61
0.07 0.68
0.08 0.72
0.09 0.74
0.1 0.73
0.11 0.75
0.12 0.77
0.13 0.82
0.14 0.74
0.15 0.9
0.16 0.86
0.17 0.9
0.18 0.87
0.19 0.9
0.2 0.9
0.21 0.92
0.22 0.92
0.23 0.96
0.24 0.97
0.25 0.94
0.26 0.97
0.27 0.98
0.28 0.94
0.29 0.97
0.3 0.99
0.31 0.95
0.32 0.98
0.33 1
0.34 0.98
0.35 0.97
0.36 0.98
0.37 0.99
0.38 1
0.39 1
0.4 1
0.41 1
0.42 0.98
0.43 1
0.44 0.99
0.45 0.99
0.46 0.99
0.47 1
0.48 1
0.49 1
0.5 1
0.51 1
0.52 1
0.53 1
0.54 1
0.55 0.98
0.56 1
0.57 1
0.58 1
0.59 1
0.6 1
0.61 1
0.62 0.98
0.63 1
0.64 1
0.65 1
0.66 1
0.67 1
0.68 1
0.69 1
0.7 1
0.71 1
0.72 1
0.73 1
0.74 1
0.75 1
0.76 1
0.77 1
0.78 1
0.79 1
0.8 1
0.81 1
0.82 1
0.83 1
0.84 0.99
0.85 1
0.86 1
0.87 1
0.88 1
0.89 1
0.9 1
0.91 1
0.92 1
0.93 1
0.94 1
0.95 1
0.96 1
0.97 1
0.98 1
0.99 1
1 1
1.02 1
1.03 1
1.04 1
1.05 1
1.06 1
1.07 1
1.08 1
1.09 1
1.1 0.99
1.11 0.99
1.12 0.99
1.13 1
1.14 0.97
1.15 0.97
1.16 0.94
1.17 0.95
1.18 0.95
1.19 0.96
1.2 0.92
1.21 0.91
1.22 0.93
1.23 0.88
1.24 0.84
1.25 0.92
1.26 0.89
1.27 0.82
1.28 0.87
1.29 0.77
1.3 0.73
1.31 0.7
1.32 0.83
1.33 0.74
1.34 0.72
1.35 0.7
1.36 0.75
1.37 0.73
1.38 0.66
1.39 0.73
1.4 0.68
1.41 0.58
1.42 0.73
1.43 0.61
1.44 0.74
1.45 0.61
1.46 0.69
1.47 0.67
1.48 0.74
1.49 0.54
1.5 0.61
1.51 0.61
1.52 0.6
1.53 0.64
1.54 0.51
1.55 0.51
1.56 0.55
1.57 0.51
1.58 0.58
1.59 0.54
1.6 0.49
1.61 0.52
1.62 0.5
1.63 0.5
1.64 0.44
1.65 0.5
1.66 0.51
1.67 0.42
1.68 0.61
1.69 0.45
1.7 0.43
1.71 0.5
1.72 0.43
1.73 0.52
1.74 0.44
1.75 0.42
1.76 0.49
1.77 0.41
1.78 0.5
1.79 0.48
1.8 0.49
1.81 0.56
1.82 0.38
1.83 0.51
1.84 0.45
1.85 0.39
1.86 0.45
1.87 0.4
1.88 0.52
1.89 0.41
1.9 0.45
1.91 0.37
1.92 0.53
1.93 0.4
1.94 0.37
1.95 0.44
1.96 0.39
1.97 0.53
1.98 0.37
1.99 0.36
2 0.44
};
\addplot [semithick, black, opacity=1, dash pattern=on 1pt off 1pt]
table {%
-0.0895000000000001 0.5
2.0995 0.5
};
\addplot [semithick, black, opacity=1, dash pattern=on 1pt off 1pt]
table {%
-0.0895000000000001 1
2.0995 1
};
\addplot [semithick, black, opacity=1, dash pattern=on 1pt off 1pt]
table {%
1 0.2
1 1.05
};
\end{axis}

\end{tikzpicture}

%% file: plots/decoupled2/GAUxLAP.tex
% This file was created by tikzplotlib v0.9.6.
\begin{tikzpicture}

\definecolor{color0}{rgb}{0.866666666666667,0.494117647058824,0.164705882352941}
\definecolor{color1}{rgb}{0.164705882352941,0.643137254901961,0.866666666666667}
\definecolor{color2}{rgb}{0.584313725490196,0.866666666666667,0.164705882352941}
\definecolor{color3}{rgb}{0.109803921568627,0.337254901960784,0.129411764705882}
\definecolor{color4}{rgb}{0.529411764705882,0.305882352941176,0.858823529411765}
\definecolor{color5}{rgb}{0.858823529411765,0.305882352941176,0.435294117647059}
\definecolor{color6}{rgb}{0.937254901960784,0.929411764705882,0.392156862745098}
\definecolor{color7}{rgb}{0.0901960784313725,0.486274509803922,0.0980392156862745}
\definecolor{color8}{rgb}{0.156862745098039,0.188235294117647,0.827450980392157}
\definecolor{color9}{rgb}{0.937254901960784,0.392156862745098,0.894117647058824}
\definecolor{color10}{rgb}{0.2,0.184313725490196,0.184313725490196}
\definecolor{color11}{rgb}{0.0156862745098039,0.803921568627451,0.976470588235294}

\begin{axis}[
tick align=outside,
tick pos=left,
x grid style={white!69.0196078431373!black},
xmajorgrids,
xmin=-0.0895, xmax=2.0995,
xtick style={color=black},
xtick={0,0.1,0.2,0.3,0.4,0.5,0.6,0.7,0.8,0.9,1,1.1,1.2,1.3,1.4,1.5,1.6,1.7,1.8,1.9,2},
xticklabels={0,,.2,,.4,,.6,,.8,,1,,20,,40,,60,,80,,100},
height=4.8cm,
width=6.5cm,
y grid style={white!69.0196078431373!black},
ymajorgrids,
ymin=0.2, ymax=1.05,
ytick style={color=black}
]
\addplot [semithick, color0, opacity=0.75]
table {%
0.01 0.47
0.02 0.48
0.03 0.57
0.04 0.59
0.05 0.54
0.06 0.58
0.07 0.48
0.08 0.67
0.09 0.49
0.1 0.64
0.11 0.64
0.12 0.64
0.13 0.66
0.14 0.61
0.15 0.69
0.16 0.63
0.17 0.71
0.18 0.82
0.19 0.75
0.2 0.76
0.21 0.77
0.22 0.79
0.23 0.78
0.24 0.77
0.25 0.78
0.26 0.85
0.27 0.89
0.28 0.85
0.29 0.82
0.3 0.8
0.31 0.88
0.32 0.85
0.33 0.85
0.34 0.86
0.35 0.86
0.36 0.82
0.37 0.89
0.38 0.94
0.39 0.92
0.4 0.92
0.41 0.93
0.42 0.93
0.43 0.89
0.44 0.91
0.45 0.89
0.46 0.95
0.47 0.96
0.48 0.94
0.49 0.94
0.5 0.93
0.51 0.9
0.52 0.99
0.53 0.95
0.54 0.95
0.55 0.95
0.56 0.92
0.57 0.97
0.58 0.98
0.59 0.98
0.6 0.97
0.61 0.94
0.62 0.93
0.63 0.94
0.64 0.99
0.65 0.93
0.66 0.95
0.67 0.94
0.68 0.96
0.69 0.97
0.7 0.97
0.71 0.96
0.72 0.97
0.73 0.98
0.74 0.96
0.75 0.99
0.76 0.96
0.77 0.99
0.78 0.99
0.79 0.98
0.8 0.98
0.81 0.96
0.82 0.99
0.83 0.98
0.84 0.97
0.85 0.96
0.86 1
0.87 0.98
0.88 1
0.89 0.99
0.9 0.98
0.91 0.98
0.92 0.98
0.93 0.99
0.94 0.97
0.95 0.96
0.96 0.99
0.97 1
0.98 0.99
0.99 0.97
1 0.99
1.02 0.9
1.03 0.96
1.04 0.81
1.05 0.79
1.06 0.72
1.07 0.76
1.08 0.72
1.09 0.66
1.1 0.6
1.11 0.64
1.12 0.58
1.13 0.59
1.14 0.6
1.15 0.55
1.16 0.48
1.17 0.56
1.18 0.59
1.19 0.55
1.2 0.57
1.21 0.47
1.22 0.46
1.23 0.62
1.24 0.48
1.25 0.57
1.26 0.47
1.27 0.47
1.28 0.44
1.29 0.47
1.3 0.52
1.31 0.47
1.32 0.47
1.33 0.45
1.34 0.46
1.35 0.45
1.36 0.4
1.37 0.53
1.38 0.44
1.39 0.5
1.4 0.37
1.41 0.38
1.42 0.42
1.43 0.43
1.44 0.38
1.45 0.41
1.46 0.45
1.47 0.44
1.48 0.43
1.49 0.41
1.5 0.53
1.51 0.4
1.52 0.46
1.53 0.41
1.54 0.39
1.55 0.48
1.56 0.39
1.57 0.45
1.58 0.37
1.59 0.53
1.6 0.44
1.61 0.38
1.62 0.41
1.63 0.36
1.64 0.32
1.65 0.48
1.66 0.32
1.67 0.43
1.68 0.46
1.69 0.46
1.7 0.46
1.71 0.39
1.72 0.38
1.73 0.42
1.74 0.38
1.75 0.4
1.76 0.46
1.77 0.51
1.78 0.48
1.79 0.41
1.8 0.4
1.81 0.46
1.82 0.46
1.83 0.35
1.84 0.34
1.85 0.34
1.86 0.41
1.87 0.4
1.88 0.32
1.89 0.51
1.9 0.45
1.91 0.45
1.92 0.39
1.93 0.37
1.94 0.45
1.95 0.42
1.96 0.36
1.97 0.43
1.98 0.49
1.99 0.39
2 0.44
};
\addplot [semithick, color1, opacity=0.75]
table {%
0.01 0.4
0.02 0.49
0.03 0.49
0.04 0.43
0.05 0.56
0.06 0.49
0.07 0.45
0.08 0.57
0.09 0.55
0.1 0.56
0.11 0.55
0.12 0.45
0.13 0.48
0.14 0.43
0.15 0.55
0.16 0.42
0.17 0.49
0.18 0.62
0.19 0.6
0.2 0.52
0.21 0.56
0.22 0.59
0.23 0.51
0.24 0.56
0.25 0.53
0.26 0.61
0.27 0.54
0.28 0.6
0.29 0.65
0.3 0.6
0.31 0.64
0.32 0.58
0.33 0.48
0.34 0.6
0.35 0.61
0.36 0.6
0.37 0.64
0.38 0.76
0.39 0.68
0.4 0.78
0.41 0.67
0.42 0.74
0.43 0.65
0.44 0.7
0.45 0.72
0.46 0.77
0.47 0.76
0.48 0.78
0.49 0.85
0.5 0.83
0.51 0.8
0.52 0.85
0.53 0.85
0.54 0.8
0.55 0.82
0.56 0.79
0.57 0.88
0.58 0.81
0.59 0.84
0.6 0.9
0.61 0.84
0.62 0.73
0.63 0.87
0.64 0.84
0.65 0.86
0.66 0.86
0.67 0.92
0.68 0.88
0.69 0.85
0.7 0.85
0.71 0.89
0.72 0.87
0.73 0.9
0.74 0.89
0.75 0.92
0.76 0.91
0.77 0.95
0.78 0.93
0.79 0.92
0.8 0.93
0.81 0.88
0.82 0.92
0.83 0.91
0.84 0.92
0.85 0.9
0.86 0.93
0.87 0.92
0.88 0.97
0.89 0.96
0.9 0.96
0.91 0.94
0.92 0.93
0.93 0.96
0.94 0.95
0.95 0.91
0.96 0.96
0.97 0.96
0.98 0.97
0.99 0.96
1 0.98
1.02 0.91
1.03 0.94
1.04 0.8
1.05 0.84
1.06 0.68
1.07 0.63
1.08 0.66
1.09 0.63
1.1 0.6
1.11 0.6
1.12 0.52
1.13 0.58
1.14 0.55
1.15 0.51
1.16 0.46
1.17 0.51
1.18 0.44
1.19 0.42
1.2 0.53
1.21 0.6
1.22 0.47
1.23 0.49
1.24 0.43
1.25 0.56
1.26 0.44
1.27 0.46
1.28 0.48
1.29 0.44
1.3 0.54
1.31 0.53
1.32 0.55
1.33 0.43
1.34 0.52
1.35 0.5
1.36 0.55
1.37 0.47
1.38 0.5
1.39 0.4
1.4 0.46
1.41 0.54
1.42 0.44
1.43 0.44
1.44 0.46
1.45 0.48
1.46 0.49
1.47 0.51
1.48 0.43
1.49 0.54
1.5 0.49
1.51 0.52
1.52 0.59
1.53 0.44
1.54 0.55
1.55 0.42
1.56 0.5
1.57 0.48
1.58 0.5
1.59 0.46
1.6 0.55
1.61 0.52
1.62 0.53
1.63 0.51
1.64 0.59
1.65 0.5
1.66 0.47
1.67 0.5
1.68 0.54
1.69 0.52
1.7 0.45
1.71 0.59
1.72 0.42
1.73 0.49
1.74 0.45
1.75 0.54
1.76 0.51
1.77 0.5
1.78 0.49
1.79 0.46
1.8 0.43
1.81 0.55
1.82 0.51
1.83 0.51
1.84 0.44
1.85 0.49
1.86 0.54
1.87 0.45
1.88 0.38
1.89 0.5
1.9 0.42
1.91 0.51
1.92 0.47
1.93 0.5
1.94 0.5
1.95 0.52
1.96 0.53
1.97 0.51
1.98 0.53
1.99 0.5
2 0.53
};
\addplot [semithick, color2, opacity=0.75]
table {%
0.01 0.42
0.02 0.49
0.03 0.47
0.04 0.52
0.05 0.51
0.06 0.52
0.07 0.55
0.08 0.55
0.09 0.45
0.1 0.54
0.11 0.54
0.12 0.45
0.13 0.47
0.14 0.42
0.15 0.55
0.16 0.39
0.17 0.52
0.18 0.61
0.19 0.6
0.2 0.57
0.21 0.54
0.22 0.56
0.23 0.54
0.24 0.54
0.25 0.53
0.26 0.61
0.27 0.54
0.28 0.59
0.29 0.64
0.3 0.59
0.31 0.64
0.32 0.56
0.33 0.49
0.34 0.58
0.35 0.6
0.36 0.62
0.37 0.62
0.38 0.75
0.39 0.69
0.4 0.77
0.41 0.67
0.42 0.72
0.43 0.66
0.44 0.7
0.45 0.72
0.46 0.76
0.47 0.77
0.48 0.78
0.49 0.85
0.5 0.83
0.51 0.78
0.52 0.85
0.53 0.83
0.54 0.8
0.55 0.82
0.56 0.79
0.57 0.88
0.58 0.81
0.59 0.84
0.6 0.89
0.61 0.85
0.62 0.72
0.63 0.86
0.64 0.83
0.65 0.86
0.66 0.85
0.67 0.92
0.68 0.88
0.69 0.85
0.7 0.85
0.71 0.87
0.72 0.86
0.73 0.9
0.74 0.89
0.75 0.91
0.76 0.91
0.77 0.95
0.78 0.93
0.79 0.9
0.8 0.93
0.81 0.88
0.82 0.91
0.83 0.9
0.84 0.92
0.85 0.9
0.86 0.93
0.87 0.92
0.88 0.97
0.89 0.96
0.9 0.96
0.91 0.94
0.92 0.93
0.93 0.96
0.94 0.95
0.95 0.91
0.96 0.96
0.97 0.96
0.98 0.96
0.99 0.96
1 0.98
1.02 0.91
1.03 0.92
1.04 0.8
1.05 0.84
1.06 0.69
1.07 0.65
1.08 0.66
1.09 0.63
1.1 0.6
1.11 0.6
1.12 0.51
1.13 0.58
1.14 0.56
1.15 0.52
1.16 0.45
1.17 0.54
1.18 0.44
1.19 0.44
1.2 0.51
1.21 0.55
1.22 0.48
1.23 0.49
1.24 0.42
1.25 0.55
1.26 0.46
1.27 0.45
1.28 0.5
1.29 0.45
1.3 0.56
1.31 0.54
1.32 0.53
1.33 0.46
1.34 0.49
1.35 0.46
1.36 0.56
1.37 0.46
1.38 0.52
1.39 0.41
1.4 0.47
1.41 0.51
1.42 0.42
1.43 0.44
1.44 0.45
1.45 0.49
1.46 0.51
1.47 0.5
1.48 0.41
1.49 0.56
1.5 0.49
1.51 0.5
1.52 0.59
1.53 0.43
1.54 0.57
1.55 0.44
1.56 0.5
1.57 0.48
1.58 0.47
1.59 0.45
1.6 0.54
1.61 0.54
1.62 0.52
1.63 0.51
1.64 0.57
1.65 0.49
1.66 0.49
1.67 0.51
1.68 0.54
1.69 0.57
1.7 0.47
1.71 0.61
1.72 0.43
1.73 0.49
1.74 0.48
1.75 0.56
1.76 0.52
1.77 0.48
1.78 0.48
1.79 0.46
1.8 0.43
1.81 0.55
1.82 0.54
1.83 0.51
1.84 0.46
1.85 0.46
1.86 0.53
1.87 0.47
1.88 0.38
1.89 0.49
1.9 0.43
1.91 0.51
1.92 0.48
1.93 0.51
1.94 0.47
1.95 0.53
1.96 0.53
1.97 0.49
1.98 0.54
1.99 0.5
2 0.55
};
\addplot [semithick, color3, opacity=0.75]
table {%
0.01 0.43
0.02 0.47
0.03 0.55
0.04 0.57
0.05 0.58
0.06 0.57
0.07 0.51
0.08 0.62
0.09 0.58
0.1 0.68
0.11 0.62
0.12 0.57
0.13 0.7
0.14 0.61
0.15 0.73
0.16 0.62
0.17 0.69
0.18 0.76
0.19 0.8
0.2 0.74
0.21 0.74
0.22 0.84
0.23 0.77
0.24 0.76
0.25 0.73
0.26 0.85
0.27 0.83
0.28 0.84
0.29 0.81
0.3 0.74
0.31 0.84
0.32 0.79
0.33 0.79
0.34 0.84
0.35 0.82
0.36 0.81
0.37 0.84
0.38 0.88
0.39 0.9
0.4 0.9
0.41 0.85
0.42 0.88
0.43 0.83
0.44 0.81
0.45 0.88
0.46 0.94
0.47 0.89
0.48 0.9
0.49 0.9
0.5 0.91
0.51 0.8
0.52 0.94
0.53 0.91
0.54 0.91
0.55 0.89
0.56 0.85
0.57 0.9
0.58 0.92
0.59 0.91
0.6 0.93
0.61 0.9
0.62 0.86
0.63 0.91
0.64 0.91
0.65 0.91
0.66 0.88
0.67 0.93
0.68 0.91
0.69 0.9
0.7 0.91
0.71 0.91
0.72 0.87
0.73 0.91
0.74 0.93
0.75 0.95
0.76 0.95
0.77 0.97
0.78 0.92
0.79 0.92
0.8 0.93
0.81 0.9
0.82 0.91
0.83 0.89
0.84 0.92
0.85 0.94
0.86 0.91
0.87 0.91
0.88 0.93
0.89 0.96
0.9 0.96
0.91 0.92
0.92 0.94
0.93 0.9
0.94 0.92
0.95 0.91
0.96 0.91
0.97 0.94
0.98 0.94
0.99 0.88
1 0.96
1.02 0.84
1.03 0.9
1.04 0.74
1.05 0.81
1.06 0.84
1.07 0.68
1.08 0.62
1.09 0.61
1.1 0.55
1.11 0.54
1.12 0.56
1.13 0.65
1.14 0.57
1.15 0.59
1.16 0.52
1.17 0.52
1.18 0.53
1.19 0.51
1.2 0.62
1.21 0.49
1.22 0.58
1.23 0.53
1.24 0.55
1.25 0.51
1.26 0.43
1.27 0.52
1.28 0.5
1.29 0.51
1.3 0.47
1.31 0.46
1.32 0.48
1.33 0.51
1.34 0.51
1.35 0.45
1.36 0.45
1.37 0.44
1.38 0.47
1.39 0.47
1.4 0.42
1.41 0.42
1.42 0.39
1.43 0.36
1.44 0.47
1.45 0.44
1.46 0.49
1.47 0.47
1.48 0.47
1.49 0.41
1.5 0.51
1.51 0.39
1.52 0.41
1.53 0.42
1.54 0.42
1.55 0.52
1.56 0.43
1.57 0.43
1.58 0.37
1.59 0.46
1.6 0.46
1.61 0.4
1.62 0.38
1.63 0.35
1.64 0.36
1.65 0.5
1.66 0.33
1.67 0.44
1.68 0.45
1.69 0.42
1.7 0.38
1.71 0.37
1.72 0.39
1.73 0.4
1.74 0.45
1.75 0.44
1.76 0.45
1.77 0.45
1.78 0.51
1.79 0.41
1.8 0.4
1.81 0.43
1.82 0.44
1.83 0.31
1.84 0.39
1.85 0.38
1.86 0.36
1.87 0.44
1.88 0.33
1.89 0.59
1.9 0.47
1.91 0.42
1.92 0.38
1.93 0.38
1.94 0.48
1.95 0.5
1.96 0.35
1.97 0.46
1.98 0.53
1.99 0.37
2 0.41
};
\addplot [semithick, color4, opacity=0.75]
table {%
0.01 0.44
0.02 0.46
0.03 0.57
0.04 0.58
0.05 0.59
0.06 0.57
0.07 0.54
0.08 0.61
0.09 0.57
0.1 0.67
0.11 0.62
0.12 0.58
0.13 0.7
0.14 0.58
0.15 0.72
0.16 0.6
0.17 0.69
0.18 0.74
0.19 0.81
0.2 0.73
0.21 0.76
0.22 0.84
0.23 0.74
0.24 0.76
0.25 0.72
0.26 0.84
0.27 0.84
0.28 0.85
0.29 0.81
0.3 0.74
0.31 0.82
0.32 0.77
0.33 0.79
0.34 0.84
0.35 0.8
0.36 0.79
0.37 0.84
0.38 0.87
0.39 0.88
0.4 0.9
0.41 0.81
0.42 0.87
0.43 0.8
0.44 0.81
0.45 0.88
0.46 0.9
0.47 0.89
0.48 0.9
0.49 0.91
0.5 0.91
0.51 0.8
0.52 0.93
0.53 0.92
0.54 0.91
0.55 0.88
0.56 0.85
0.57 0.89
0.58 0.9
0.59 0.91
0.6 0.92
0.61 0.88
0.62 0.86
0.63 0.9
0.64 0.9
0.65 0.91
0.66 0.82
0.67 0.92
0.68 0.9
0.69 0.87
0.7 0.9
0.71 0.9
0.72 0.87
0.73 0.9
0.74 0.93
0.75 0.95
0.76 0.93
0.77 0.97
0.78 0.91
0.79 0.91
0.8 0.92
0.81 0.89
0.82 0.89
0.83 0.88
0.84 0.88
0.85 0.92
0.86 0.89
0.87 0.9
0.88 0.92
0.89 0.96
0.9 0.92
0.91 0.9
0.92 0.9
0.93 0.88
0.94 0.91
0.95 0.9
0.96 0.89
0.97 0.94
0.98 0.9
0.99 0.84
1 0.94
1.02 0.83
1.03 0.92
1.04 0.73
1.05 0.79
1.06 0.84
1.07 0.67
1.08 0.62
1.09 0.61
1.1 0.54
1.11 0.54
1.12 0.56
1.13 0.64
1.14 0.57
1.15 0.57
1.16 0.52
1.17 0.53
1.18 0.56
1.19 0.51
1.2 0.61
1.21 0.49
1.22 0.57
1.23 0.54
1.24 0.55
1.25 0.51
1.26 0.43
1.27 0.52
1.28 0.48
1.29 0.53
1.3 0.48
1.31 0.46
1.32 0.46
1.33 0.5
1.34 0.51
1.35 0.46
1.36 0.44
1.37 0.44
1.38 0.47
1.39 0.48
1.4 0.43
1.41 0.42
1.42 0.39
1.43 0.38
1.44 0.46
1.45 0.44
1.46 0.48
1.47 0.47
1.48 0.47
1.49 0.41
1.5 0.51
1.51 0.4
1.52 0.41
1.53 0.43
1.54 0.41
1.55 0.5
1.56 0.42
1.57 0.44
1.58 0.36
1.59 0.46
1.6 0.47
1.61 0.4
1.62 0.38
1.63 0.35
1.64 0.36
1.65 0.5
1.66 0.34
1.67 0.43
1.68 0.44
1.69 0.43
1.7 0.39
1.71 0.37
1.72 0.38
1.73 0.42
1.74 0.45
1.75 0.43
1.76 0.44
1.77 0.46
1.78 0.51
1.79 0.42
1.8 0.4
1.81 0.43
1.82 0.45
1.83 0.32
1.84 0.38
1.85 0.38
1.86 0.37
1.87 0.45
1.88 0.33
1.89 0.56
1.9 0.46
1.91 0.42
1.92 0.38
1.93 0.38
1.94 0.52
1.95 0.51
1.96 0.36
1.97 0.45
1.98 0.53
1.99 0.38
2 0.41
};
\addplot [semithick, color5, opacity=0.75]
table {%
0.01 0.5
0.02 0.46
0.03 0.5
0.04 0.57
0.05 0.52
0.06 0.52
0.07 0.52
0.08 0.59
0.09 0.52
0.1 0.66
0.11 0.64
0.12 0.61
0.13 0.63
0.14 0.56
0.15 0.67
0.16 0.59
0.17 0.66
0.18 0.69
0.19 0.77
0.2 0.69
0.21 0.65
0.22 0.74
0.23 0.64
0.24 0.73
0.25 0.72
0.26 0.79
0.27 0.78
0.28 0.79
0.29 0.78
0.3 0.75
0.31 0.8
0.32 0.78
0.33 0.74
0.34 0.81
0.35 0.75
0.36 0.75
0.37 0.82
0.38 0.87
0.39 0.85
0.4 0.9
0.41 0.8
0.42 0.83
0.43 0.75
0.44 0.78
0.45 0.84
0.46 0.89
0.47 0.89
0.48 0.85
0.49 0.84
0.5 0.85
0.51 0.75
0.52 0.86
0.53 0.81
0.54 0.85
0.55 0.86
0.56 0.81
0.57 0.84
0.58 0.86
0.59 0.86
0.6 0.9
0.61 0.86
0.62 0.78
0.63 0.86
0.64 0.82
0.65 0.86
0.66 0.77
0.67 0.87
0.68 0.83
0.69 0.85
0.7 0.87
0.71 0.82
0.72 0.86
0.73 0.84
0.74 0.86
0.75 0.84
0.76 0.87
0.77 0.94
0.78 0.85
0.79 0.86
0.8 0.88
0.81 0.85
0.82 0.87
0.83 0.85
0.84 0.88
0.85 0.85
0.86 0.87
0.87 0.87
0.88 0.87
0.89 0.89
0.9 0.9
0.91 0.83
0.92 0.84
0.93 0.83
0.94 0.86
0.95 0.86
0.96 0.84
0.97 0.86
0.98 0.85
0.99 0.85
1 0.88
1.02 0.84
1.03 0.89
1.04 0.74
1.05 0.72
1.06 0.81
1.07 0.65
1.08 0.66
1.09 0.56
1.1 0.59
1.11 0.54
1.12 0.54
1.13 0.55
1.14 0.6
1.15 0.55
1.16 0.49
1.17 0.54
1.18 0.46
1.19 0.47
1.2 0.6
1.21 0.47
1.22 0.48
1.23 0.56
1.24 0.49
1.25 0.56
1.26 0.43
1.27 0.5
1.28 0.41
1.29 0.46
1.3 0.45
1.31 0.46
1.32 0.47
1.33 0.43
1.34 0.49
1.35 0.45
1.36 0.41
1.37 0.46
1.38 0.47
1.39 0.45
1.4 0.44
1.41 0.39
1.42 0.34
1.43 0.32
1.44 0.41
1.45 0.42
1.46 0.45
1.47 0.39
1.48 0.51
1.49 0.43
1.5 0.48
1.51 0.37
1.52 0.46
1.53 0.42
1.54 0.37
1.55 0.5
1.56 0.43
1.57 0.39
1.58 0.34
1.59 0.4
1.6 0.45
1.61 0.36
1.62 0.37
1.63 0.39
1.64 0.34
1.65 0.43
1.66 0.31
1.67 0.34
1.68 0.42
1.69 0.38
1.7 0.38
1.71 0.38
1.72 0.3
1.73 0.34
1.74 0.38
1.75 0.35
1.76 0.43
1.77 0.45
1.78 0.46
1.79 0.4
1.8 0.46
1.81 0.42
1.82 0.42
1.83 0.28
1.84 0.33
1.85 0.31
1.86 0.35
1.87 0.39
1.88 0.35
1.89 0.39
1.9 0.39
1.91 0.37
1.92 0.37
1.93 0.44
1.94 0.41
1.95 0.45
1.96 0.3
1.97 0.46
1.98 0.47
1.99 0.39
2 0.34
};
\addplot [semithick, color6, opacity=0.75, dashed]
table {%
0.01 0.51
0.02 0.52
0.03 0.51
0.04 0.55
0.05 0.57
0.06 0.6
0.07 0.53
0.08 0.47
0.09 0.49
0.1 0.48
0.11 0.48
0.12 0.52
0.13 0.44
0.14 0.56
0.15 0.55
0.16 0.59
0.17 0.58
0.18 0.43
0.19 0.59
0.2 0.46
0.21 0.59
0.22 0.61
0.23 0.57
0.24 0.54
0.25 0.51
0.26 0.51
0.27 0.58
0.28 0.56
0.29 0.53
0.3 0.65
0.31 0.53
0.32 0.58
0.33 0.53
0.34 0.59
0.35 0.67
0.36 0.55
0.37 0.54
0.38 0.63
0.39 0.65
0.4 0.76
0.41 0.64
0.42 0.61
0.43 0.68
0.44 0.56
0.45 0.58
0.46 0.69
0.47 0.65
0.48 0.64
0.49 0.71
0.5 0.64
0.51 0.67
0.52 0.63
0.53 0.7
0.54 0.67
0.55 0.61
0.56 0.68
0.57 0.78
0.58 0.59
0.59 0.55
0.6 0.73
0.61 0.67
0.62 0.65
0.63 0.67
0.64 0.64
0.65 0.74
0.66 0.7
0.67 0.69
0.68 0.66
0.69 0.69
0.7 0.77
0.71 0.74
0.72 0.66
0.73 0.73
0.74 0.66
0.75 0.57
0.76 0.68
0.77 0.67
0.78 0.77
0.79 0.7
0.8 0.71
0.81 0.72
0.82 0.6
0.83 0.74
0.84 0.62
0.85 0.72
0.86 0.65
0.87 0.6
0.88 0.7
0.89 0.72
0.9 0.64
0.91 0.67
0.92 0.73
0.93 0.68
0.94 0.74
0.95 0.61
0.96 0.65
0.97 0.72
0.98 0.71
0.99 0.73
1 0.68
1.02 0.6
1.03 0.61
1.04 0.61
1.05 0.5
1.06 0.59
1.07 0.55
1.08 0.53
1.09 0.54
1.1 0.58
1.11 0.55
1.12 0.56
1.13 0.52
1.14 0.6
1.15 0.58
1.16 0.53
1.17 0.51
1.18 0.48
1.19 0.47
1.2 0.52
1.21 0.55
1.22 0.49
1.23 0.62
1.24 0.59
1.25 0.52
1.26 0.43
1.27 0.53
1.28 0.48
1.29 0.43
1.3 0.55
1.31 0.53
1.32 0.42
1.33 0.49
1.34 0.44
1.35 0.5
1.36 0.42
1.37 0.51
1.38 0.51
1.39 0.45
1.4 0.49
1.41 0.54
1.42 0.49
1.43 0.48
1.44 0.49
1.45 0.44
1.46 0.49
1.47 0.41
1.48 0.55
1.49 0.57
1.5 0.54
1.51 0.57
1.52 0.56
1.53 0.56
1.54 0.45
1.55 0.5
1.56 0.55
1.57 0.4
1.58 0.5
1.59 0.54
1.6 0.49
1.61 0.53
1.62 0.44
1.63 0.59
1.64 0.51
1.65 0.5
1.66 0.52
1.67 0.48
1.68 0.4
1.69 0.56
1.7 0.44
1.71 0.52
1.72 0.55
1.73 0.45
1.74 0.52
1.75 0.41
1.76 0.48
1.77 0.51
1.78 0.57
1.79 0.53
1.8 0.58
1.81 0.52
1.82 0.58
1.83 0.47
1.84 0.45
1.85 0.44
1.86 0.43
1.87 0.47
1.88 0.47
1.89 0.51
1.9 0.53
1.91 0.56
1.92 0.51
1.93 0.55
1.94 0.52
1.95 0.48
1.96 0.56
1.97 0.53
1.98 0.32
1.99 0.51
2 0.55
};
\addplot [semithick, color7, opacity=0.75, dashed]
table {%
0.01 0.51
0.02 0.52
0.03 0.51
0.04 0.55
0.05 0.57
0.06 0.6
0.07 0.53
0.08 0.47
0.09 0.49
0.1 0.48
0.11 0.48
0.12 0.52
0.13 0.44
0.14 0.56
0.15 0.55
0.16 0.59
0.17 0.58
0.18 0.43
0.19 0.59
0.2 0.46
0.21 0.59
0.22 0.61
0.23 0.57
0.24 0.54
0.25 0.51
0.26 0.51
0.27 0.58
0.28 0.56
0.29 0.53
0.3 0.65
0.31 0.53
0.32 0.58
0.33 0.53
0.34 0.59
0.35 0.67
0.36 0.55
0.37 0.54
0.38 0.63
0.39 0.65
0.4 0.76
0.41 0.64
0.42 0.61
0.43 0.68
0.44 0.56
0.45 0.58
0.46 0.69
0.47 0.65
0.48 0.64
0.49 0.71
0.5 0.64
0.51 0.67
0.52 0.63
0.53 0.7
0.54 0.67
0.55 0.61
0.56 0.68
0.57 0.78
0.58 0.59
0.59 0.55
0.6 0.73
0.61 0.67
0.62 0.65
0.63 0.67
0.64 0.64
0.65 0.74
0.66 0.7
0.67 0.69
0.68 0.66
0.69 0.69
0.7 0.77
0.71 0.74
0.72 0.66
0.73 0.73
0.74 0.66
0.75 0.57
0.76 0.68
0.77 0.67
0.78 0.77
0.79 0.7
0.8 0.71
0.81 0.72
0.82 0.6
0.83 0.74
0.84 0.62
0.85 0.72
0.86 0.65
0.87 0.6
0.88 0.7
0.89 0.72
0.9 0.64
0.91 0.67
0.92 0.73
0.93 0.68
0.94 0.74
0.95 0.61
0.96 0.65
0.97 0.72
0.98 0.71
0.99 0.73
1 0.68
1.02 0.6
1.03 0.61
1.04 0.61
1.05 0.5
1.06 0.59
1.07 0.55
1.08 0.53
1.09 0.54
1.1 0.58
1.11 0.55
1.12 0.56
1.13 0.52
1.14 0.6
1.15 0.58
1.16 0.53
1.17 0.51
1.18 0.48
1.19 0.47
1.2 0.52
1.21 0.55
1.22 0.49
1.23 0.62
1.24 0.59
1.25 0.52
1.26 0.43
1.27 0.53
1.28 0.48
1.29 0.43
1.3 0.55
1.31 0.53
1.32 0.42
1.33 0.49
1.34 0.44
1.35 0.5
1.36 0.42
1.37 0.51
1.38 0.51
1.39 0.45
1.4 0.49
1.41 0.54
1.42 0.49
1.43 0.48
1.44 0.49
1.45 0.44
1.46 0.49
1.47 0.41
1.48 0.55
1.49 0.57
1.5 0.54
1.51 0.57
1.52 0.56
1.53 0.56
1.54 0.45
1.55 0.5
1.56 0.55
1.57 0.4
1.58 0.5
1.59 0.54
1.6 0.49
1.61 0.53
1.62 0.44
1.63 0.59
1.64 0.51
1.65 0.5
1.66 0.52
1.67 0.48
1.68 0.4
1.69 0.56
1.7 0.44
1.71 0.52
1.72 0.55
1.73 0.45
1.74 0.52
1.75 0.41
1.76 0.48
1.77 0.51
1.78 0.57
1.79 0.53
1.8 0.58
1.81 0.52
1.82 0.58
1.83 0.47
1.84 0.45
1.85 0.44
1.86 0.43
1.87 0.47
1.88 0.47
1.89 0.51
1.9 0.53
1.91 0.56
1.92 0.51
1.93 0.55
1.94 0.52
1.95 0.48
1.96 0.56
1.97 0.53
1.98 0.32
1.99 0.51
2 0.55
};
\addplot [semithick, color8, opacity=0.75, dashed]
table {%
0.01 0.45
0.02 0.46
0.03 0.45
0.04 0.51
0.05 0.59
0.06 0.49
0.07 0.57
0.08 0.45
0.09 0.52
0.1 0.54
0.11 0.64
0.12 0.56
0.13 0.44
0.14 0.65
0.15 0.49
0.16 0.57
0.17 0.56
0.18 0.55
0.19 0.6
0.2 0.57
0.21 0.55
0.22 0.57
0.23 0.65
0.24 0.55
0.25 0.53
0.26 0.63
0.27 0.6
0.28 0.54
0.29 0.62
0.3 0.66
0.31 0.59
0.32 0.63
0.33 0.61
0.34 0.61
0.35 0.65
0.36 0.59
0.37 0.54
0.38 0.68
0.39 0.63
0.4 0.71
0.41 0.63
0.42 0.59
0.43 0.74
0.44 0.66
0.45 0.65
0.46 0.69
0.47 0.67
0.48 0.65
0.49 0.71
0.5 0.73
0.51 0.73
0.52 0.67
0.53 0.73
0.54 0.72
0.55 0.66
0.56 0.73
0.57 0.74
0.58 0.7
0.59 0.61
0.6 0.71
0.61 0.72
0.62 0.69
0.63 0.7
0.64 0.68
0.65 0.79
0.66 0.69
0.67 0.74
0.68 0.77
0.69 0.71
0.7 0.81
0.71 0.77
0.72 0.74
0.73 0.68
0.74 0.76
0.75 0.64
0.76 0.72
0.77 0.75
0.78 0.8
0.79 0.84
0.8 0.85
0.81 0.7
0.82 0.73
0.83 0.73
0.84 0.74
0.85 0.75
0.86 0.71
0.87 0.65
0.88 0.78
0.89 0.74
0.9 0.77
0.91 0.74
0.92 0.77
0.93 0.78
0.94 0.78
0.95 0.79
0.96 0.74
0.97 0.71
0.98 0.74
0.99 0.8
1 0.77
1.02 0.7
1.03 0.67
1.04 0.61
1.05 0.52
1.06 0.58
1.07 0.58
1.08 0.45
1.09 0.56
1.1 0.53
1.11 0.53
1.12 0.46
1.13 0.42
1.14 0.56
1.15 0.5
1.16 0.46
1.17 0.52
1.18 0.51
1.19 0.5
1.2 0.53
1.21 0.5
1.22 0.56
1.23 0.57
1.24 0.58
1.25 0.58
1.26 0.52
1.27 0.51
1.28 0.46
1.29 0.49
1.3 0.47
1.31 0.51
1.32 0.6
1.33 0.54
1.34 0.39
1.35 0.47
1.36 0.47
1.37 0.49
1.38 0.56
1.39 0.42
1.4 0.57
1.41 0.48
1.42 0.44
1.43 0.46
1.44 0.51
1.45 0.43
1.46 0.44
1.47 0.48
1.48 0.48
1.49 0.53
1.5 0.57
1.51 0.49
1.52 0.54
1.53 0.52
1.54 0.44
1.55 0.5
1.56 0.56
1.57 0.46
1.58 0.5
1.59 0.51
1.6 0.47
1.61 0.47
1.62 0.46
1.63 0.49
1.64 0.48
1.65 0.49
1.66 0.44
1.67 0.54
1.68 0.49
1.69 0.58
1.7 0.52
1.71 0.48
1.72 0.46
1.73 0.51
1.74 0.54
1.75 0.51
1.76 0.53
1.77 0.5
1.78 0.58
1.79 0.55
1.8 0.55
1.81 0.56
1.82 0.54
1.83 0.51
1.84 0.45
1.85 0.4
1.86 0.46
1.87 0.54
1.88 0.43
1.89 0.46
1.9 0.52
1.91 0.46
1.92 0.48
1.93 0.51
1.94 0.55
1.95 0.54
1.96 0.49
1.97 0.49
1.98 0.43
1.99 0.49
2 0.53
};
\addplot [semithick, color9, opacity=0.75, dashed]
table {%
0.01 0.43
0.02 0.58
0.03 0.53
0.04 0.59
0.05 0.55
0.06 0.53
0.07 0.6
0.08 0.75
0.09 0.69
0.1 0.69
0.11 0.66
0.12 0.71
0.13 0.72
0.14 0.71
0.15 0.82
0.16 0.74
0.17 0.78
0.18 0.84
0.19 0.85
0.2 0.88
0.21 0.83
0.22 0.83
0.23 0.91
0.24 0.88
0.25 0.83
0.26 0.9
0.27 0.96
0.28 0.92
0.29 0.9
0.3 0.9
0.31 0.96
0.32 0.92
0.33 0.93
0.34 0.96
0.35 0.95
0.36 0.9
0.37 0.97
0.38 0.98
0.39 0.99
0.4 1
0.41 0.94
0.42 0.95
0.43 0.94
0.44 0.97
0.45 0.96
0.46 0.98
0.47 0.97
0.48 0.97
0.49 0.98
0.5 0.98
0.51 0.98
0.52 0.97
0.53 0.99
0.54 0.99
0.55 0.97
0.56 1
0.57 0.99
0.58 1
0.59 1
0.6 0.98
0.61 0.99
0.62 0.96
0.63 0.99
0.64 1
0.65 0.98
0.66 0.98
0.67 0.99
0.68 0.98
0.69 1
0.7 0.99
0.71 0.99
0.72 0.99
0.73 1
0.74 0.99
0.75 1
0.76 0.99
0.77 1
0.78 0.99
0.79 0.99
0.8 1
0.81 0.99
0.82 0.97
0.83 0.99
0.84 1
0.85 0.99
0.86 1
0.87 1
0.88 1
0.89 1
0.9 1
0.91 0.99
0.92 1
0.93 1
0.94 1
0.95 0.99
0.96 1
0.97 0.99
0.98 1
0.99 0.97
1 0.99
1.02 0.96
1.03 0.96
1.04 0.83
1.05 0.87
1.06 0.85
1.07 0.78
1.08 0.72
1.09 0.74
1.1 0.74
1.11 0.71
1.12 0.7
1.13 0.66
1.14 0.71
1.15 0.61
1.16 0.49
1.17 0.63
1.18 0.67
1.19 0.57
1.2 0.61
1.21 0.54
1.22 0.57
1.23 0.61
1.24 0.6
1.25 0.58
1.26 0.49
1.27 0.6
1.28 0.56
1.29 0.53
1.3 0.57
1.31 0.48
1.32 0.49
1.33 0.57
1.34 0.48
1.35 0.49
1.36 0.48
1.37 0.5
1.38 0.45
1.39 0.59
1.4 0.41
1.41 0.43
1.42 0.41
1.43 0.51
1.44 0.44
1.45 0.44
1.46 0.57
1.47 0.48
1.48 0.51
1.49 0.45
1.5 0.52
1.51 0.46
1.52 0.5
1.53 0.47
1.54 0.39
1.55 0.55
1.56 0.45
1.57 0.43
1.58 0.4
1.59 0.48
1.6 0.44
1.61 0.5
1.62 0.41
1.63 0.47
1.64 0.34
1.65 0.55
1.66 0.43
1.67 0.45
1.68 0.42
1.69 0.53
1.7 0.39
1.71 0.35
1.72 0.42
1.73 0.4
1.74 0.53
1.75 0.46
1.76 0.46
1.77 0.49
1.78 0.57
1.79 0.4
1.8 0.42
1.81 0.5
1.82 0.4
1.83 0.43
1.84 0.43
1.85 0.38
1.86 0.4
1.87 0.47
1.88 0.31
1.89 0.54
1.9 0.49
1.91 0.47
1.92 0.43
1.93 0.43
1.94 0.48
1.95 0.44
1.96 0.43
1.97 0.52
1.98 0.54
1.99 0.39
2 0.46
};
\addplot [semithick, color10, opacity=0.75, dashed]
table {%
0.01 0.48
0.02 0.52
0.03 0.49
0.04 0.6
0.05 0.63
0.06 0.59
0.07 0.66
0.08 0.65
0.09 0.6
0.1 0.68
0.11 0.6
0.12 0.67
0.13 0.61
0.14 0.65
0.15 0.77
0.16 0.76
0.17 0.67
0.18 0.84
0.19 0.84
0.2 0.81
0.21 0.85
0.22 0.84
0.23 0.87
0.24 0.85
0.25 0.81
0.26 0.85
0.27 0.92
0.28 0.89
0.29 0.89
0.3 0.86
0.31 0.92
0.32 0.91
0.33 0.87
0.34 0.95
0.35 0.93
0.36 0.91
0.37 0.95
0.38 0.94
0.39 0.98
0.4 0.98
0.41 0.94
0.42 0.92
0.43 0.93
0.44 0.96
0.45 0.98
0.46 0.97
0.47 0.97
0.48 0.96
0.49 0.97
0.5 0.95
0.51 0.98
0.52 0.98
0.53 0.98
0.54 0.98
0.55 0.97
0.56 0.98
0.57 0.99
0.58 0.99
0.59 1
0.6 0.98
0.61 0.97
0.62 0.95
0.63 0.97
0.64 0.97
0.65 0.97
0.66 0.98
0.67 1
0.68 0.98
0.69 1
0.7 0.99
0.71 0.97
0.72 0.99
0.73 1
0.74 0.99
0.75 0.98
0.76 0.98
0.77 1
0.78 0.99
0.79 0.99
0.8 0.98
0.81 0.97
0.82 0.97
0.83 0.99
0.84 0.99
0.85 0.99
0.86 0.98
0.87 1
0.88 1
0.89 1
0.9 0.99
0.91 0.99
0.92 1
0.93 0.99
0.94 0.99
0.95 0.99
0.96 0.97
0.97 1
0.98 0.99
0.99 0.97
1 0.99
1.02 0.97
1.03 0.95
1.04 0.85
1.05 0.86
1.06 0.76
1.07 0.78
1.08 0.7
1.09 0.73
1.1 0.64
1.11 0.69
1.12 0.65
1.13 0.62
1.14 0.67
1.15 0.6
1.16 0.57
1.17 0.57
1.18 0.6
1.19 0.52
1.2 0.6
1.21 0.58
1.22 0.53
1.23 0.54
1.24 0.59
1.25 0.49
1.26 0.5
1.27 0.51
1.28 0.54
1.29 0.52
1.3 0.56
1.31 0.47
1.32 0.42
1.33 0.47
1.34 0.51
1.35 0.49
1.36 0.54
1.37 0.5
1.38 0.5
1.39 0.56
1.4 0.43
1.41 0.46
1.42 0.43
1.43 0.46
1.44 0.46
1.45 0.42
1.46 0.5
1.47 0.47
1.48 0.41
1.49 0.43
1.5 0.45
1.51 0.4
1.52 0.48
1.53 0.45
1.54 0.45
1.55 0.55
1.56 0.43
1.57 0.49
1.58 0.46
1.59 0.47
1.6 0.45
1.61 0.41
1.62 0.45
1.63 0.41
1.64 0.32
1.65 0.54
1.66 0.48
1.67 0.48
1.68 0.57
1.69 0.49
1.7 0.44
1.71 0.38
1.72 0.46
1.73 0.42
1.74 0.58
1.75 0.44
1.76 0.49
1.77 0.47
1.78 0.52
1.79 0.4
1.8 0.46
1.81 0.49
1.82 0.45
1.83 0.41
1.84 0.47
1.85 0.4
1.86 0.53
1.87 0.38
1.88 0.38
1.89 0.54
1.9 0.5
1.91 0.55
1.92 0.44
1.93 0.39
1.94 0.45
1.95 0.46
1.96 0.42
1.97 0.46
1.98 0.5
1.99 0.39
2 0.47
};
\addplot [semithick, color11, opacity=0.75, dashed]
table {%
0.01 0.43
0.02 0.57
0.03 0.63
0.04 0.6
0.05 0.55
0.06 0.61
0.07 0.55
0.08 0.64
0.09 0.54
0.1 0.56
0.11 0.74
0.12 0.61
0.13 0.61
0.14 0.69
0.15 0.72
0.16 0.69
0.17 0.73
0.18 0.79
0.19 0.69
0.2 0.67
0.21 0.69
0.22 0.73
0.23 0.81
0.24 0.83
0.25 0.73
0.26 0.83
0.27 0.79
0.28 0.8
0.29 0.79
0.3 0.72
0.31 0.84
0.32 0.88
0.33 0.72
0.34 0.79
0.35 0.8
0.36 0.78
0.37 0.86
0.38 0.84
0.39 0.89
0.4 0.85
0.41 0.87
0.42 0.83
0.43 0.77
0.44 0.86
0.45 0.82
0.46 0.9
0.47 0.89
0.48 0.89
0.49 0.91
0.5 0.88
0.51 0.85
0.52 0.91
0.53 0.9
0.54 0.85
0.55 0.8
0.56 0.84
0.57 0.89
0.58 0.92
0.59 0.91
0.6 0.9
0.61 0.88
0.62 0.88
0.63 0.9
0.64 0.89
0.65 0.91
0.66 0.87
0.67 0.93
0.68 0.92
0.69 0.95
0.7 0.88
0.71 0.91
0.72 0.89
0.73 0.91
0.74 0.88
0.75 0.92
0.76 0.91
0.77 0.86
0.78 0.95
0.79 0.9
0.8 0.9
0.81 0.93
0.82 0.87
0.83 0.95
0.84 0.91
0.85 0.92
0.86 0.94
0.87 0.92
0.88 0.91
0.89 0.93
0.9 0.93
0.91 0.94
0.92 0.92
0.93 0.92
0.94 0.92
0.95 0.89
0.96 0.92
0.97 0.91
0.98 0.92
0.99 0.9
1 0.89
1.02 0.82
1.03 0.79
1.04 0.72
1.05 0.77
1.06 0.75
1.07 0.72
1.08 0.62
1.09 0.72
1.1 0.63
1.11 0.65
1.12 0.55
1.13 0.61
1.14 0.63
1.15 0.59
1.16 0.52
1.17 0.6
1.18 0.58
1.19 0.52
1.2 0.59
1.21 0.5
1.22 0.56
1.23 0.57
1.24 0.48
1.25 0.54
1.26 0.45
1.27 0.44
1.28 0.49
1.29 0.51
1.3 0.53
1.31 0.47
1.32 0.44
1.33 0.47
1.34 0.47
1.35 0.5
1.36 0.46
1.37 0.42
1.38 0.5
1.39 0.58
1.4 0.51
1.41 0.52
1.42 0.45
1.43 0.46
1.44 0.43
1.45 0.46
1.46 0.57
1.47 0.43
1.48 0.5
1.49 0.43
1.5 0.5
1.51 0.49
1.52 0.36
1.53 0.45
1.54 0.45
1.55 0.47
1.56 0.48
1.57 0.38
1.58 0.4
1.59 0.39
1.6 0.44
1.61 0.4
1.62 0.5
1.63 0.47
1.64 0.44
1.65 0.55
1.66 0.45
1.67 0.47
1.68 0.49
1.69 0.41
1.7 0.55
1.71 0.46
1.72 0.33
1.73 0.48
1.74 0.53
1.75 0.45
1.76 0.58
1.77 0.51
1.78 0.45
1.79 0.49
1.8 0.53
1.81 0.54
1.82 0.5
1.83 0.43
1.84 0.46
1.85 0.35
1.86 0.45
1.87 0.45
1.88 0.44
1.89 0.39
1.9 0.42
1.91 0.51
1.92 0.41
1.93 0.42
1.94 0.54
1.95 0.51
1.96 0.44
1.97 0.49
1.98 0.45
1.99 0.41
2 0.46
};
\addplot [semithick, black, opacity=1, dash pattern=on 1pt off 1pt]
table {%
-0.0895000000000001 0.5
2.0995 0.5
};
\addplot [semithick, black, opacity=1, dash pattern=on 1pt off 1pt]
table {%
-0.0895000000000001 1
2.0995 1
};
\addplot [semithick, black, opacity=1, dash pattern=on 1pt off 1pt]
table {%
1 0.2
1 1.05
};
\end{axis}

\end{tikzpicture}

%% file: plots/decoupled2/UNIxGAU.tex
% This file was created by tikzplotlib v0.9.6.
\begin{tikzpicture}

\definecolor{color0}{rgb}{0.866666666666667,0.494117647058824,0.164705882352941}
\definecolor{color1}{rgb}{0.164705882352941,0.643137254901961,0.866666666666667}
\definecolor{color2}{rgb}{0.584313725490196,0.866666666666667,0.164705882352941}
\definecolor{color3}{rgb}{0.109803921568627,0.337254901960784,0.129411764705882}
\definecolor{color4}{rgb}{0.529411764705882,0.305882352941176,0.858823529411765}
\definecolor{color5}{rgb}{0.858823529411765,0.305882352941176,0.435294117647059}
\definecolor{color6}{rgb}{0.937254901960784,0.929411764705882,0.392156862745098}
\definecolor{color7}{rgb}{0.0901960784313725,0.486274509803922,0.0980392156862745}
\definecolor{color8}{rgb}{0.156862745098039,0.188235294117647,0.827450980392157}
\definecolor{color9}{rgb}{0.937254901960784,0.392156862745098,0.894117647058824}
\definecolor{color10}{rgb}{0.2,0.184313725490196,0.184313725490196}
\definecolor{color11}{rgb}{0.0156862745098039,0.803921568627451,0.976470588235294}

\begin{axis}[
tick align=outside,
tick pos=left,
x grid style={white!69.0196078431373!black},
xmajorgrids,
xmin=-0.0895, xmax=2.0995,
xtick style={color=black},
xtick={0,0.1,0.2,0.3,0.4,0.5,0.6,0.7,0.8,0.9,1,1.1,1.2,1.3,1.4,1.5,1.6,1.7,1.8,1.9,2},
xticklabels={0,,.2,,.4,,.6,,.8,,1,,20,,40,,60,,80,,100},
height=4.8cm,
width=6.5cm,
y grid style={white!69.0196078431373!black},
ymajorgrids,
ymin=0.2, ymax=1.05,
ytick style={color=black}
]
\addplot [semithick, color0, opacity=0.75]
table {%
0.01 0.49
0.02 0.57
0.03 0.61
0.04 0.63
0.05 0.57
0.06 0.68
0.07 0.69
0.08 0.65
0.09 0.76
0.1 0.72
0.11 0.76
0.12 0.87
0.13 0.85
0.14 0.8
0.15 0.8
0.16 0.81
0.17 0.85
0.18 0.9
0.19 0.89
0.2 0.84
0.21 0.9
0.22 0.96
0.23 0.9
0.24 0.9
0.25 0.92
0.26 0.98
0.27 0.91
0.28 0.89
0.29 0.97
0.3 0.96
0.31 0.97
0.32 0.96
0.33 0.93
0.34 0.98
0.35 0.96
0.36 0.96
0.37 1
0.38 0.98
0.39 0.96
0.4 0.97
0.41 0.98
0.42 0.94
0.43 0.98
0.44 0.95
0.45 0.96
0.46 1
0.47 0.94
0.48 0.99
0.49 0.99
0.5 0.99
0.51 1
0.52 1
0.53 0.97
0.54 1
0.55 1
0.56 0.99
0.57 0.99
0.58 0.98
0.59 0.97
0.6 0.99
0.61 1
0.62 0.99
0.63 0.98
0.64 0.98
0.65 1
0.66 1
0.67 1
0.68 0.98
0.69 1
0.7 0.99
0.71 0.98
0.72 0.98
0.73 0.96
0.74 0.99
0.75 0.98
0.76 0.98
0.77 0.97
0.78 0.98
0.79 0.97
0.8 0.96
0.81 0.99
0.82 0.99
0.83 0.98
0.84 0.97
0.85 0.97
0.86 0.96
0.87 1
0.88 1
0.89 0.98
0.9 0.98
0.91 0.99
0.92 0.98
0.93 0.95
0.94 0.97
0.95 0.96
0.96 0.96
0.97 0.98
0.98 0.96
0.99 0.97
1 0.98
1.02 0.86
1.03 0.81
1.04 0.75
1.05 0.74
1.06 0.63
1.07 0.64
1.08 0.6
1.09 0.5
1.1 0.58
1.11 0.6
1.12 0.58
1.13 0.49
1.14 0.54
1.15 0.58
1.16 0.53
1.17 0.54
1.18 0.54
1.19 0.4
1.2 0.51
1.21 0.5
1.22 0.46
1.23 0.57
1.24 0.47
1.25 0.48
1.26 0.37
1.27 0.58
1.28 0.45
1.29 0.52
1.3 0.45
1.31 0.49
1.32 0.42
1.33 0.49
1.34 0.57
1.35 0.46
1.36 0.4
1.37 0.39
1.38 0.52
1.39 0.45
1.4 0.45
1.41 0.41
1.42 0.46
1.43 0.47
1.44 0.48
1.45 0.43
1.46 0.49
1.47 0.42
1.48 0.36
1.49 0.42
1.5 0.5
1.51 0.45
1.52 0.44
1.53 0.39
1.54 0.57
1.55 0.44
1.56 0.42
1.57 0.42
1.58 0.48
1.59 0.51
1.6 0.37
1.61 0.41
1.62 0.52
1.63 0.53
1.64 0.4
1.65 0.58
1.66 0.47
1.67 0.42
1.68 0.49
1.69 0.46
1.7 0.49
1.71 0.5
1.72 0.43
1.73 0.45
1.74 0.46
1.75 0.55
1.76 0.49
1.77 0.48
1.78 0.48
1.79 0.41
1.8 0.4
1.81 0.45
1.82 0.5
1.83 0.46
1.84 0.47
1.85 0.4
1.86 0.47
1.87 0.5
1.88 0.45
1.89 0.36
1.9 0.48
1.91 0.45
1.92 0.53
1.93 0.46
1.94 0.43
1.95 0.39
1.96 0.48
1.97 0.44
1.98 0.46
1.99 0.45
2 0.53
};
\addplot [semithick, color1, opacity=0.75]
table {%
0.01 0.4
0.02 0.51
0.03 0.5
0.04 0.55
0.05 0.56
0.06 0.56
0.07 0.51
0.08 0.53
0.09 0.61
0.1 0.4
0.11 0.51
0.12 0.66
0.13 0.65
0.14 0.72
0.15 0.69
0.16 0.68
0.17 0.71
0.18 0.69
0.19 0.75
0.2 0.65
0.21 0.66
0.22 0.74
0.23 0.66
0.24 0.76
0.25 0.7
0.26 0.75
0.27 0.64
0.28 0.67
0.29 0.76
0.3 0.76
0.31 0.71
0.32 0.73
0.33 0.63
0.34 0.76
0.35 0.73
0.36 0.78
0.37 0.76
0.38 0.8
0.39 0.8
0.4 0.74
0.41 0.67
0.42 0.68
0.43 0.81
0.44 0.73
0.45 0.76
0.46 0.77
0.47 0.7
0.48 0.71
0.49 0.69
0.5 0.84
0.51 0.8
0.52 0.81
0.53 0.76
0.54 0.73
0.55 0.76
0.56 0.8
0.57 0.86
0.58 0.77
0.59 0.69
0.6 0.78
0.61 0.73
0.62 0.79
0.63 0.73
0.64 0.68
0.65 0.78
0.66 0.86
0.67 0.77
0.68 0.83
0.69 0.82
0.7 0.84
0.71 0.78
0.72 0.77
0.73 0.77
0.74 0.76
0.75 0.73
0.76 0.74
0.77 0.73
0.78 0.64
0.79 0.76
0.8 0.69
0.81 0.77
0.82 0.75
0.83 0.7
0.84 0.73
0.85 0.79
0.86 0.72
0.87 0.68
0.88 0.76
0.89 0.8
0.9 0.69
0.91 0.68
0.92 0.7
0.93 0.59
0.94 0.76
0.95 0.67
0.96 0.7
0.97 0.66
0.98 0.69
0.99 0.81
1 0.76
1.02 0.56
1.03 0.47
1.04 0.43
1.05 0.5
1.06 0.46
1.07 0.51
1.08 0.55
1.09 0.52
1.1 0.47
1.11 0.54
1.12 0.48
1.13 0.5
1.14 0.46
1.15 0.47
1.16 0.52
1.17 0.59
1.18 0.45
1.19 0.53
1.2 0.53
1.21 0.51
1.22 0.43
1.23 0.42
1.24 0.46
1.25 0.44
1.26 0.5
1.27 0.39
1.28 0.47
1.29 0.49
1.3 0.53
1.31 0.46
1.32 0.44
1.33 0.5
1.34 0.51
1.35 0.52
1.36 0.6
1.37 0.53
1.38 0.52
1.39 0.41
1.4 0.49
1.41 0.48
1.42 0.47
1.43 0.46
1.44 0.59
1.45 0.49
1.46 0.46
1.47 0.47
1.48 0.48
1.49 0.54
1.5 0.54
1.51 0.54
1.52 0.53
1.53 0.53
1.54 0.51
1.55 0.51
1.56 0.46
1.57 0.52
1.58 0.52
1.59 0.48
1.6 0.53
1.61 0.57
1.62 0.63
1.63 0.55
1.64 0.54
1.65 0.5
1.66 0.54
1.67 0.49
1.68 0.61
1.69 0.41
1.7 0.6
1.71 0.44
1.72 0.38
1.73 0.54
1.74 0.53
1.75 0.48
1.76 0.49
1.77 0.41
1.78 0.46
1.79 0.47
1.8 0.49
1.81 0.56
1.82 0.46
1.83 0.55
1.84 0.5
1.85 0.53
1.86 0.46
1.87 0.51
1.88 0.53
1.89 0.49
1.9 0.46
1.91 0.52
1.92 0.45
1.93 0.39
1.94 0.52
1.95 0.45
1.96 0.5
1.97 0.49
1.98 0.5
1.99 0.47
2 0.53
};
\addplot [semithick, color2, opacity=0.75]
table {%
0.01 0.56
0.02 0.42
0.03 0.54
0.04 0.55
0.05 0.55
0.06 0.49
0.07 0.48
0.08 0.48
0.09 0.54
0.1 0.42
0.11 0.52
0.12 0.67
0.13 0.63
0.14 0.71
0.15 0.69
0.16 0.67
0.17 0.71
0.18 0.7
0.19 0.75
0.2 0.65
0.21 0.67
0.22 0.74
0.23 0.64
0.24 0.75
0.25 0.67
0.26 0.75
0.27 0.61
0.28 0.65
0.29 0.75
0.3 0.74
0.31 0.72
0.32 0.72
0.33 0.62
0.34 0.77
0.35 0.71
0.36 0.79
0.37 0.77
0.38 0.79
0.39 0.74
0.4 0.73
0.41 0.66
0.42 0.66
0.43 0.81
0.44 0.71
0.45 0.76
0.46 0.77
0.47 0.71
0.48 0.69
0.49 0.68
0.5 0.84
0.51 0.79
0.52 0.8
0.53 0.76
0.54 0.71
0.55 0.76
0.56 0.8
0.57 0.85
0.58 0.77
0.59 0.68
0.6 0.78
0.61 0.73
0.62 0.78
0.63 0.71
0.64 0.68
0.65 0.78
0.66 0.86
0.67 0.77
0.68 0.83
0.69 0.81
0.7 0.84
0.71 0.78
0.72 0.76
0.73 0.76
0.74 0.75
0.75 0.72
0.76 0.73
0.77 0.7
0.78 0.62
0.79 0.76
0.8 0.68
0.81 0.76
0.82 0.76
0.83 0.69
0.84 0.73
0.85 0.79
0.86 0.71
0.87 0.66
0.88 0.76
0.89 0.78
0.9 0.68
0.91 0.66
0.92 0.68
0.93 0.59
0.94 0.75
0.95 0.67
0.96 0.68
0.97 0.65
0.98 0.68
0.99 0.79
1 0.74
1.02 0.52
1.03 0.46
1.04 0.43
1.05 0.5
1.06 0.48
1.07 0.52
1.08 0.56
1.09 0.56
1.1 0.46
1.11 0.55
1.12 0.46
1.13 0.5
1.14 0.48
1.15 0.46
1.16 0.51
1.17 0.61
1.18 0.46
1.19 0.52
1.2 0.52
1.21 0.52
1.22 0.42
1.23 0.42
1.24 0.47
1.25 0.45
1.26 0.53
1.27 0.41
1.28 0.46
1.29 0.48
1.3 0.53
1.31 0.47
1.32 0.46
1.33 0.53
1.34 0.5
1.35 0.53
1.36 0.58
1.37 0.55
1.38 0.52
1.39 0.43
1.4 0.48
1.41 0.48
1.42 0.48
1.43 0.48
1.44 0.6
1.45 0.53
1.46 0.44
1.47 0.48
1.48 0.47
1.49 0.52
1.5 0.53
1.51 0.53
1.52 0.55
1.53 0.5
1.54 0.5
1.55 0.53
1.56 0.46
1.57 0.51
1.58 0.55
1.59 0.49
1.6 0.54
1.61 0.55
1.62 0.62
1.63 0.53
1.64 0.53
1.65 0.5
1.66 0.57
1.67 0.48
1.68 0.63
1.69 0.42
1.7 0.6
1.71 0.45
1.72 0.39
1.73 0.54
1.74 0.52
1.75 0.51
1.76 0.51
1.77 0.42
1.78 0.48
1.79 0.48
1.8 0.51
1.81 0.58
1.82 0.45
1.83 0.59
1.84 0.5
1.85 0.55
1.86 0.46
1.87 0.5
1.88 0.54
1.89 0.49
1.9 0.47
1.91 0.54
1.92 0.46
1.93 0.42
1.94 0.52
1.95 0.47
1.96 0.49
1.97 0.45
1.98 0.52
1.99 0.47
2 0.55
};
\addplot [semithick, color3, opacity=0.75]
table {%
0.01 0.46
0.02 0.57
0.03 0.59
0.04 0.6
0.05 0.62
0.06 0.67
0.07 0.64
0.08 0.66
0.09 0.77
0.1 0.66
0.11 0.69
0.12 0.83
0.13 0.8
0.14 0.78
0.15 0.81
0.16 0.87
0.17 0.84
0.18 0.84
0.19 0.89
0.2 0.76
0.21 0.86
0.22 0.88
0.23 0.85
0.24 0.84
0.25 0.87
0.26 0.93
0.27 0.78
0.28 0.82
0.29 0.92
0.3 0.91
0.31 0.92
0.32 0.88
0.33 0.81
0.34 0.88
0.35 0.92
0.36 0.9
0.37 0.94
0.38 0.92
0.39 0.89
0.4 0.89
0.41 0.8
0.42 0.84
0.43 0.92
0.44 0.84
0.45 0.9
0.46 0.85
0.47 0.83
0.48 0.87
0.49 0.89
0.5 0.92
0.51 0.91
0.52 0.92
0.53 0.92
0.54 0.92
0.55 0.87
0.56 0.91
0.57 0.92
0.58 0.9
0.59 0.78
0.6 0.85
0.61 0.81
0.62 0.88
0.63 0.82
0.64 0.9
0.65 0.88
0.66 0.94
0.67 0.89
0.68 0.87
0.69 0.91
0.7 0.91
0.71 0.88
0.72 0.89
0.73 0.89
0.74 0.88
0.75 0.87
0.76 0.89
0.77 0.87
0.78 0.79
0.79 0.84
0.8 0.84
0.81 0.91
0.82 0.9
0.83 0.87
0.84 0.86
0.85 0.9
0.86 0.84
0.87 0.82
0.88 0.9
0.89 0.9
0.9 0.81
0.91 0.89
0.92 0.83
0.93 0.79
0.94 0.86
0.95 0.86
0.96 0.83
0.97 0.89
0.98 0.88
0.99 0.94
1 0.88
1.02 0.86
1.03 0.77
1.04 0.73
1.05 0.74
1.06 0.61
1.07 0.65
1.08 0.57
1.09 0.61
1.1 0.59
1.11 0.65
1.12 0.55
1.13 0.48
1.14 0.51
1.15 0.66
1.16 0.5
1.17 0.56
1.18 0.45
1.19 0.46
1.2 0.45
1.21 0.51
1.22 0.49
1.23 0.59
1.24 0.45
1.25 0.44
1.26 0.45
1.27 0.52
1.28 0.46
1.29 0.47
1.3 0.46
1.31 0.48
1.32 0.46
1.33 0.51
1.34 0.53
1.35 0.54
1.36 0.39
1.37 0.46
1.38 0.54
1.39 0.55
1.4 0.49
1.41 0.44
1.42 0.4
1.43 0.51
1.44 0.58
1.45 0.5
1.46 0.53
1.47 0.47
1.48 0.45
1.49 0.5
1.5 0.53
1.51 0.46
1.52 0.53
1.53 0.42
1.54 0.54
1.55 0.47
1.56 0.45
1.57 0.48
1.58 0.51
1.59 0.49
1.6 0.43
1.61 0.45
1.62 0.53
1.63 0.62
1.64 0.45
1.65 0.57
1.66 0.49
1.67 0.46
1.68 0.47
1.69 0.44
1.7 0.51
1.71 0.53
1.72 0.44
1.73 0.45
1.74 0.56
1.75 0.56
1.76 0.55
1.77 0.47
1.78 0.45
1.79 0.44
1.8 0.45
1.81 0.46
1.82 0.52
1.83 0.48
1.84 0.48
1.85 0.41
1.86 0.48
1.87 0.52
1.88 0.49
1.89 0.37
1.9 0.46
1.91 0.51
1.92 0.44
1.93 0.4
1.94 0.44
1.95 0.44
1.96 0.52
1.97 0.43
1.98 0.49
1.99 0.47
2 0.5
};
\addplot [semithick, color4, opacity=0.75]
table {%
0.01 0.43
0.02 0.58
0.03 0.6
0.04 0.61
0.05 0.61
0.06 0.68
0.07 0.64
0.08 0.67
0.09 0.77
0.1 0.68
0.11 0.71
0.12 0.86
0.13 0.82
0.14 0.79
0.15 0.81
0.16 0.87
0.17 0.86
0.18 0.85
0.19 0.89
0.2 0.79
0.21 0.86
0.22 0.91
0.23 0.87
0.24 0.85
0.25 0.88
0.26 0.96
0.27 0.82
0.28 0.85
0.29 0.93
0.3 0.92
0.31 0.93
0.32 0.92
0.33 0.84
0.34 0.95
0.35 0.93
0.36 0.9
0.37 0.94
0.38 0.92
0.39 0.9
0.4 0.9
0.41 0.86
0.42 0.87
0.43 0.92
0.44 0.85
0.45 0.93
0.46 0.9
0.47 0.87
0.48 0.89
0.49 0.92
0.5 0.93
0.51 0.91
0.52 0.94
0.53 0.92
0.54 0.93
0.55 0.89
0.56 0.95
0.57 0.94
0.58 0.92
0.59 0.87
0.6 0.87
0.61 0.84
0.62 0.91
0.63 0.87
0.64 0.9
0.65 0.89
0.66 0.94
0.67 0.91
0.68 0.88
0.69 0.93
0.7 0.91
0.71 0.9
0.72 0.91
0.73 0.91
0.74 0.91
0.75 0.89
0.76 0.92
0.77 0.89
0.78 0.84
0.79 0.87
0.8 0.86
0.81 0.95
0.82 0.92
0.83 0.89
0.84 0.92
0.85 0.92
0.86 0.86
0.87 0.85
0.88 0.9
0.89 0.92
0.9 0.84
0.91 0.91
0.92 0.86
0.93 0.79
0.94 0.86
0.95 0.91
0.96 0.85
0.97 0.91
0.98 0.89
0.99 0.96
1 0.92
1.02 0.87
1.03 0.77
1.04 0.73
1.05 0.75
1.06 0.62
1.07 0.66
1.08 0.57
1.09 0.62
1.1 0.59
1.11 0.65
1.12 0.57
1.13 0.49
1.14 0.52
1.15 0.65
1.16 0.5
1.17 0.56
1.18 0.44
1.19 0.46
1.2 0.46
1.21 0.49
1.22 0.51
1.23 0.59
1.24 0.46
1.25 0.44
1.26 0.45
1.27 0.52
1.28 0.46
1.29 0.47
1.3 0.48
1.31 0.49
1.32 0.47
1.33 0.52
1.34 0.53
1.35 0.53
1.36 0.4
1.37 0.46
1.38 0.54
1.39 0.56
1.4 0.49
1.41 0.44
1.42 0.41
1.43 0.51
1.44 0.58
1.45 0.5
1.46 0.54
1.47 0.47
1.48 0.45
1.49 0.5
1.5 0.52
1.51 0.47
1.52 0.53
1.53 0.42
1.54 0.53
1.55 0.47
1.56 0.46
1.57 0.47
1.58 0.51
1.59 0.5
1.6 0.43
1.61 0.45
1.62 0.53
1.63 0.62
1.64 0.45
1.65 0.56
1.66 0.49
1.67 0.48
1.68 0.48
1.69 0.44
1.7 0.51
1.71 0.53
1.72 0.46
1.73 0.46
1.74 0.57
1.75 0.55
1.76 0.53
1.77 0.47
1.78 0.46
1.79 0.44
1.8 0.45
1.81 0.44
1.82 0.51
1.83 0.48
1.84 0.51
1.85 0.4
1.86 0.47
1.87 0.52
1.88 0.5
1.89 0.37
1.9 0.46
1.91 0.51
1.92 0.45
1.93 0.4
1.94 0.45
1.95 0.47
1.96 0.53
1.97 0.42
1.98 0.5
1.99 0.47
2 0.5
};
\addplot [semithick, color5, opacity=0.75]
table {%
0.01 0.41
0.02 0.57
0.03 0.63
0.04 0.56
0.05 0.56
0.06 0.7
0.07 0.63
0.08 0.63
0.09 0.74
0.1 0.61
0.11 0.73
0.12 0.82
0.13 0.75
0.14 0.76
0.15 0.73
0.16 0.83
0.17 0.79
0.18 0.81
0.19 0.85
0.2 0.71
0.21 0.85
0.22 0.8
0.23 0.87
0.24 0.84
0.25 0.81
0.26 0.91
0.27 0.79
0.28 0.76
0.29 0.87
0.3 0.89
0.31 0.87
0.32 0.87
0.33 0.77
0.34 0.85
0.35 0.84
0.36 0.84
0.37 0.89
0.38 0.87
0.39 0.86
0.4 0.85
0.41 0.79
0.42 0.81
0.43 0.87
0.44 0.79
0.45 0.85
0.46 0.84
0.47 0.84
0.48 0.81
0.49 0.84
0.5 0.86
0.51 0.88
0.52 0.92
0.53 0.91
0.54 0.85
0.55 0.85
0.56 0.88
0.57 0.9
0.58 0.87
0.59 0.75
0.6 0.82
0.61 0.81
0.62 0.86
0.63 0.81
0.64 0.86
0.65 0.87
0.66 0.88
0.67 0.84
0.68 0.86
0.69 0.87
0.7 0.86
0.71 0.88
0.72 0.87
0.73 0.87
0.74 0.85
0.75 0.82
0.76 0.89
0.77 0.82
0.78 0.73
0.79 0.83
0.8 0.8
0.81 0.87
0.82 0.89
0.83 0.84
0.84 0.83
0.85 0.88
0.86 0.82
0.87 0.81
0.88 0.86
0.89 0.87
0.9 0.75
0.91 0.81
0.92 0.82
0.93 0.77
0.94 0.84
0.95 0.84
0.96 0.8
0.97 0.85
0.98 0.83
0.99 0.94
1 0.84
1.02 0.79
1.03 0.76
1.04 0.66
1.05 0.66
1.06 0.61
1.07 0.62
1.08 0.59
1.09 0.57
1.1 0.61
1.11 0.58
1.12 0.59
1.13 0.48
1.14 0.56
1.15 0.62
1.16 0.52
1.17 0.51
1.18 0.48
1.19 0.42
1.2 0.51
1.21 0.47
1.22 0.52
1.23 0.62
1.24 0.46
1.25 0.51
1.26 0.4
1.27 0.47
1.28 0.48
1.29 0.49
1.3 0.46
1.31 0.43
1.32 0.59
1.33 0.54
1.34 0.56
1.35 0.51
1.36 0.39
1.37 0.46
1.38 0.49
1.39 0.57
1.4 0.44
1.41 0.45
1.42 0.42
1.43 0.51
1.44 0.59
1.45 0.43
1.46 0.53
1.47 0.47
1.48 0.45
1.49 0.41
1.5 0.52
1.51 0.46
1.52 0.49
1.53 0.42
1.54 0.54
1.55 0.42
1.56 0.46
1.57 0.51
1.58 0.55
1.59 0.49
1.6 0.46
1.61 0.57
1.62 0.55
1.63 0.58
1.64 0.46
1.65 0.57
1.66 0.43
1.67 0.47
1.68 0.46
1.69 0.4
1.7 0.59
1.71 0.41
1.72 0.46
1.73 0.5
1.74 0.48
1.75 0.48
1.76 0.55
1.77 0.52
1.78 0.42
1.79 0.4
1.8 0.4
1.81 0.43
1.82 0.49
1.83 0.46
1.84 0.49
1.85 0.48
1.86 0.5
1.87 0.56
1.88 0.48
1.89 0.38
1.9 0.47
1.91 0.5
1.92 0.42
1.93 0.37
1.94 0.42
1.95 0.5
1.96 0.48
1.97 0.5
1.98 0.45
1.99 0.54
2 0.46
};
\addplot [semithick, color6, opacity=0.75, dashed]
table {%
0.01 0.51
0.02 0.58
0.03 0.6
0.04 0.6
0.05 0.55
0.06 0.57
0.07 0.66
0.08 0.67
0.09 0.64
0.1 0.71
0.11 0.7
0.12 0.77
0.13 0.71
0.14 0.77
0.15 0.79
0.16 0.83
0.17 0.81
0.18 0.72
0.19 0.87
0.2 0.81
0.21 0.83
0.22 0.87
0.23 0.81
0.24 0.89
0.25 0.85
0.26 0.93
0.27 0.92
0.28 0.87
0.29 0.88
0.3 0.9
0.31 0.88
0.32 0.91
0.33 0.91
0.34 0.92
0.35 0.93
0.36 0.91
0.37 0.87
0.38 0.9
0.39 0.91
0.4 0.86
0.41 0.9
0.42 0.95
0.43 0.87
0.44 0.94
0.45 0.96
0.46 0.94
0.47 0.94
0.48 0.93
0.49 0.88
0.5 0.9
0.51 0.86
0.52 0.9
0.53 0.9
0.54 0.95
0.55 0.89
0.56 0.9
0.57 0.92
0.58 0.9
0.59 0.93
0.6 0.93
0.61 0.92
0.62 0.92
0.63 0.9
0.64 0.87
0.65 0.91
0.66 0.89
0.67 0.88
0.68 0.92
0.69 0.92
0.7 0.96
0.71 0.92
0.72 0.96
0.73 0.88
0.74 0.93
0.75 0.93
0.76 0.94
0.77 0.89
0.78 0.92
0.79 0.92
0.8 0.89
0.81 0.9
0.82 0.9
0.83 0.91
0.84 0.88
0.85 0.95
0.86 0.92
0.87 0.92
0.88 0.91
0.89 0.87
0.9 0.94
0.91 0.92
0.92 0.87
0.93 0.9
0.94 0.91
0.95 0.9
0.96 0.9
0.97 0.95
0.98 0.88
0.99 0.92
1 0.94
1.02 0.83
1.03 0.73
1.04 0.69
1.05 0.57
1.06 0.69
1.07 0.65
1.08 0.59
1.09 0.63
1.1 0.53
1.11 0.54
1.12 0.59
1.13 0.52
1.14 0.62
1.15 0.55
1.16 0.58
1.17 0.52
1.18 0.51
1.19 0.65
1.2 0.56
1.21 0.58
1.22 0.54
1.23 0.54
1.24 0.52
1.25 0.53
1.26 0.5
1.27 0.53
1.28 0.61
1.29 0.59
1.3 0.53
1.31 0.64
1.32 0.42
1.33 0.49
1.34 0.67
1.35 0.54
1.36 0.58
1.37 0.52
1.38 0.59
1.39 0.48
1.4 0.5
1.41 0.45
1.42 0.54
1.43 0.54
1.44 0.51
1.45 0.53
1.46 0.53
1.47 0.57
1.48 0.59
1.49 0.52
1.5 0.48
1.51 0.54
1.52 0.59
1.53 0.5
1.54 0.54
1.55 0.54
1.56 0.56
1.57 0.45
1.58 0.59
1.59 0.47
1.6 0.57
1.61 0.54
1.62 0.59
1.63 0.6
1.64 0.65
1.65 0.49
1.66 0.48
1.67 0.56
1.68 0.61
1.69 0.59
1.7 0.56
1.71 0.58
1.72 0.46
1.73 0.49
1.74 0.59
1.75 0.58
1.76 0.56
1.77 0.45
1.78 0.58
1.79 0.53
1.8 0.52
1.81 0.5
1.82 0.59
1.83 0.58
1.84 0.49
1.85 0.63
1.86 0.51
1.87 0.58
1.88 0.54
1.89 0.52
1.9 0.51
1.91 0.54
1.92 0.58
1.93 0.6
1.94 0.58
1.95 0.55
1.96 0.56
1.97 0.55
1.98 0.57
1.99 0.6
2 0.58
};
\addplot [semithick, color7, opacity=0.75, dashed]
table {%
0.01 0.51
0.02 0.58
0.03 0.6
0.04 0.6
0.05 0.55
0.06 0.57
0.07 0.66
0.08 0.67
0.09 0.64
0.1 0.71
0.11 0.7
0.12 0.77
0.13 0.71
0.14 0.77
0.15 0.79
0.16 0.83
0.17 0.81
0.18 0.72
0.19 0.87
0.2 0.81
0.21 0.83
0.22 0.87
0.23 0.81
0.24 0.89
0.25 0.85
0.26 0.93
0.27 0.92
0.28 0.87
0.29 0.88
0.3 0.9
0.31 0.88
0.32 0.91
0.33 0.91
0.34 0.92
0.35 0.93
0.36 0.91
0.37 0.87
0.38 0.9
0.39 0.91
0.4 0.86
0.41 0.9
0.42 0.95
0.43 0.87
0.44 0.94
0.45 0.96
0.46 0.94
0.47 0.94
0.48 0.93
0.49 0.88
0.5 0.9
0.51 0.86
0.52 0.9
0.53 0.9
0.54 0.95
0.55 0.89
0.56 0.9
0.57 0.92
0.58 0.9
0.59 0.93
0.6 0.93
0.61 0.92
0.62 0.92
0.63 0.9
0.64 0.87
0.65 0.91
0.66 0.89
0.67 0.88
0.68 0.92
0.69 0.92
0.7 0.96
0.71 0.92
0.72 0.96
0.73 0.88
0.74 0.93
0.75 0.93
0.76 0.94
0.77 0.89
0.78 0.92
0.79 0.92
0.8 0.89
0.81 0.9
0.82 0.9
0.83 0.91
0.84 0.88
0.85 0.95
0.86 0.92
0.87 0.92
0.88 0.91
0.89 0.87
0.9 0.94
0.91 0.92
0.92 0.87
0.93 0.9
0.94 0.91
0.95 0.9
0.96 0.9
0.97 0.95
0.98 0.88
0.99 0.92
1 0.94
1.02 0.83
1.03 0.73
1.04 0.69
1.05 0.57
1.06 0.69
1.07 0.65
1.08 0.59
1.09 0.63
1.1 0.53
1.11 0.54
1.12 0.59
1.13 0.52
1.14 0.62
1.15 0.55
1.16 0.58
1.17 0.52
1.18 0.51
1.19 0.65
1.2 0.56
1.21 0.58
1.22 0.54
1.23 0.54
1.24 0.52
1.25 0.53
1.26 0.5
1.27 0.53
1.28 0.61
1.29 0.59
1.3 0.53
1.31 0.64
1.32 0.42
1.33 0.49
1.34 0.67
1.35 0.54
1.36 0.58
1.37 0.52
1.38 0.59
1.39 0.48
1.4 0.5
1.41 0.45
1.42 0.54
1.43 0.54
1.44 0.51
1.45 0.53
1.46 0.53
1.47 0.57
1.48 0.59
1.49 0.52
1.5 0.48
1.51 0.54
1.52 0.59
1.53 0.5
1.54 0.54
1.55 0.54
1.56 0.56
1.57 0.45
1.58 0.59
1.59 0.47
1.6 0.57
1.61 0.54
1.62 0.59
1.63 0.6
1.64 0.65
1.65 0.49
1.66 0.48
1.67 0.56
1.68 0.61
1.69 0.59
1.7 0.56
1.71 0.58
1.72 0.46
1.73 0.49
1.74 0.59
1.75 0.58
1.76 0.56
1.77 0.45
1.78 0.58
1.79 0.53
1.8 0.52
1.81 0.5
1.82 0.59
1.83 0.58
1.84 0.49
1.85 0.63
1.86 0.51
1.87 0.58
1.88 0.54
1.89 0.52
1.9 0.51
1.91 0.54
1.92 0.58
1.93 0.6
1.94 0.58
1.95 0.55
1.96 0.56
1.97 0.55
1.98 0.57
1.99 0.6
2 0.58
};
\addplot [semithick, color8, opacity=0.75, dashed]
table {%
0.01 0.54
0.02 0.62
0.03 0.66
0.04 0.66
0.05 0.54
0.06 0.64
0.07 0.62
0.08 0.77
0.09 0.72
0.1 0.75
0.11 0.77
0.12 0.73
0.13 0.79
0.14 0.78
0.15 0.85
0.16 0.87
0.17 0.83
0.18 0.87
0.19 0.9
0.2 0.86
0.21 0.89
0.22 0.91
0.23 0.9
0.24 0.9
0.25 0.93
0.26 0.92
0.27 0.97
0.28 0.96
0.29 0.93
0.3 0.97
0.31 0.95
0.32 0.94
0.33 0.94
0.34 0.99
0.35 0.98
0.36 0.94
0.37 0.95
0.38 0.99
0.39 0.92
0.4 0.95
0.41 0.96
0.42 0.97
0.43 0.92
0.44 0.95
0.45 0.94
0.46 0.93
0.47 0.95
0.48 0.94
0.49 0.97
0.5 0.96
0.51 0.93
0.52 0.94
0.53 0.96
0.54 0.96
0.55 0.88
0.56 0.94
0.57 0.95
0.58 0.9
0.59 0.96
0.6 0.98
0.61 0.95
0.62 0.94
0.63 0.91
0.64 0.96
0.65 0.94
0.66 0.94
0.67 0.93
0.68 0.95
0.69 0.96
0.7 0.95
0.71 0.97
0.72 0.98
0.73 0.97
0.74 0.93
0.75 0.95
0.76 0.95
0.77 0.92
0.78 0.95
0.79 0.96
0.8 0.96
0.81 0.94
0.82 0.98
0.83 0.96
0.84 0.98
0.85 0.95
0.86 0.95
0.87 0.94
0.88 0.98
0.89 0.98
0.9 0.96
0.91 0.94
0.92 0.96
0.93 0.92
0.94 0.94
0.95 0.94
0.96 0.95
0.97 0.94
0.98 0.95
0.99 0.99
1 0.95
1.02 0.84
1.03 0.81
1.04 0.71
1.05 0.58
1.06 0.68
1.07 0.6
1.08 0.57
1.09 0.53
1.1 0.52
1.11 0.54
1.12 0.64
1.13 0.57
1.14 0.61
1.15 0.56
1.16 0.49
1.17 0.56
1.18 0.6
1.19 0.66
1.2 0.59
1.21 0.48
1.22 0.57
1.23 0.6
1.24 0.52
1.25 0.51
1.26 0.5
1.27 0.55
1.28 0.58
1.29 0.56
1.3 0.57
1.31 0.46
1.32 0.47
1.33 0.55
1.34 0.6
1.35 0.55
1.36 0.54
1.37 0.52
1.38 0.58
1.39 0.46
1.4 0.58
1.41 0.5
1.42 0.48
1.43 0.53
1.44 0.49
1.45 0.55
1.46 0.57
1.47 0.52
1.48 0.52
1.49 0.56
1.5 0.5
1.51 0.52
1.52 0.56
1.53 0.54
1.54 0.5
1.55 0.5
1.56 0.58
1.57 0.5
1.58 0.47
1.59 0.58
1.6 0.55
1.61 0.55
1.62 0.56
1.63 0.64
1.64 0.55
1.65 0.62
1.66 0.47
1.67 0.55
1.68 0.55
1.69 0.61
1.7 0.54
1.71 0.63
1.72 0.5
1.73 0.48
1.74 0.61
1.75 0.56
1.76 0.48
1.77 0.46
1.78 0.45
1.79 0.56
1.8 0.56
1.81 0.56
1.82 0.5
1.83 0.54
1.84 0.52
1.85 0.6
1.86 0.52
1.87 0.57
1.88 0.55
1.89 0.54
1.9 0.56
1.91 0.59
1.92 0.58
1.93 0.61
1.94 0.49
1.95 0.46
1.96 0.59
1.97 0.55
1.98 0.52
1.99 0.53
2 0.62
};
\addplot [semithick, color9, opacity=0.75, dashed]
table {%
0.01 0.54
0.02 0.56
0.03 0.58
0.04 0.52
0.05 0.66
0.06 0.65
0.07 0.63
0.08 0.67
0.09 0.82
0.1 0.74
0.11 0.72
0.12 0.72
0.13 0.79
0.14 0.77
0.15 0.8
0.16 0.83
0.17 0.89
0.18 0.89
0.19 0.87
0.2 0.88
0.21 0.94
0.22 0.93
0.23 0.96
0.24 0.94
0.25 0.97
0.26 0.98
0.27 0.94
0.28 0.93
0.29 0.98
0.3 0.98
0.31 0.99
0.32 0.98
0.33 0.96
0.34 0.99
0.35 1
0.36 0.97
0.37 0.97
0.38 0.99
0.39 0.98
0.4 0.96
0.41 1
0.42 0.99
0.43 1
0.44 0.98
0.45 0.99
0.46 1
0.47 0.99
0.48 0.99
0.49 1
0.5 1
0.51 0.99
0.52 1
0.53 1
0.54 0.99
0.55 1
0.56 0.99
0.57 0.99
0.58 0.99
0.59 0.98
0.6 0.99
0.61 0.98
0.62 0.99
0.63 0.98
0.64 0.99
0.65 0.99
0.66 1
0.67 1
0.68 0.99
0.69 1
0.7 0.99
0.71 0.97
0.72 1
0.73 0.99
0.74 0.96
0.75 1
0.76 1
0.77 0.98
0.78 0.99
0.79 0.99
0.8 0.98
0.81 1
0.82 0.98
0.83 0.99
0.84 0.98
0.85 0.99
0.86 0.97
0.87 1
0.88 0.99
0.89 0.98
0.9 0.98
0.91 0.99
0.92 0.97
0.93 0.98
0.94 0.96
0.95 0.98
0.96 0.99
0.97 0.99
0.98 0.96
0.99 0.98
1 0.98
1.02 0.9
1.03 0.78
1.04 0.6
1.05 0.74
1.06 0.66
1.07 0.61
1.08 0.67
1.09 0.58
1.1 0.53
1.11 0.58
1.12 0.64
1.13 0.58
1.14 0.6
1.15 0.6
1.16 0.56
1.17 0.6
1.18 0.54
1.19 0.45
1.2 0.51
1.21 0.52
1.22 0.58
1.23 0.6
1.24 0.61
1.25 0.57
1.26 0.49
1.27 0.57
1.28 0.45
1.29 0.59
1.3 0.55
1.31 0.5
1.32 0.48
1.33 0.51
1.34 0.55
1.35 0.56
1.36 0.49
1.37 0.55
1.38 0.55
1.39 0.63
1.4 0.54
1.41 0.45
1.42 0.55
1.43 0.51
1.44 0.56
1.45 0.55
1.46 0.56
1.47 0.53
1.48 0.47
1.49 0.51
1.5 0.55
1.51 0.53
1.52 0.5
1.53 0.47
1.54 0.63
1.55 0.56
1.56 0.52
1.57 0.44
1.58 0.59
1.59 0.48
1.6 0.6
1.61 0.53
1.62 0.55
1.63 0.63
1.64 0.52
1.65 0.64
1.66 0.52
1.67 0.55
1.68 0.54
1.69 0.55
1.7 0.55
1.71 0.58
1.72 0.6
1.73 0.59
1.74 0.6
1.75 0.58
1.76 0.58
1.77 0.44
1.78 0.55
1.79 0.57
1.8 0.56
1.81 0.54
1.82 0.5
1.83 0.6
1.84 0.6
1.85 0.53
1.86 0.52
1.87 0.59
1.88 0.51
1.89 0.48
1.9 0.66
1.91 0.52
1.92 0.61
1.93 0.58
1.94 0.51
1.95 0.65
1.96 0.58
1.97 0.58
1.98 0.62
1.99 0.57
2 0.65
};
\addplot [semithick, color10, opacity=0.75, dashed]
table {%
0.01 0.54
0.02 0.57
0.03 0.6
0.04 0.64
0.05 0.72
0.06 0.84
0.07 0.78
0.08 0.83
0.09 0.9
0.1 0.87
0.11 0.92
0.12 0.92
0.13 0.97
0.14 0.91
0.15 0.94
0.16 0.99
0.17 0.96
0.18 0.96
0.19 0.98
0.2 0.97
0.21 0.98
0.22 1
0.23 1
0.24 0.98
0.25 1
0.26 0.99
0.27 1
0.28 0.99
0.29 1
0.3 1
0.31 1
0.32 1
0.33 0.99
0.34 1
0.35 1
0.36 1
0.37 1
0.38 1
0.39 1
0.4 1
0.41 1
0.42 0.99
0.43 1
0.44 1
0.45 1
0.46 1
0.47 1
0.48 1
0.49 1
0.5 1
0.51 1
0.52 1
0.53 1
0.54 1
0.55 1
0.56 1
0.57 1
0.58 1
0.59 0.99
0.6 1
0.61 1
0.62 1
0.63 1
0.64 1
0.65 1
0.66 1
0.67 1
0.68 1
0.69 1
0.7 1
0.71 1
0.72 1
0.73 1
0.74 1
0.75 1
0.76 1
0.77 1
0.78 1
0.79 1
0.8 1
0.81 1
0.82 1
0.83 1
0.84 1
0.85 1
0.86 1
0.87 1
0.88 1
0.89 0.99
0.9 1
0.91 1
0.92 1
0.93 1
0.94 1
0.95 1
0.96 1
0.97 1
0.98 1
0.99 1
1 1
1.02 0.96
1.03 0.96
1.04 0.79
1.05 0.81
1.06 0.75
1.07 0.73
1.08 0.73
1.09 0.63
1.1 0.58
1.11 0.71
1.12 0.69
1.13 0.64
1.14 0.55
1.15 0.59
1.16 0.57
1.17 0.62
1.18 0.6
1.19 0.49
1.2 0.53
1.21 0.59
1.22 0.58
1.23 0.66
1.24 0.65
1.25 0.62
1.26 0.54
1.27 0.63
1.28 0.5
1.29 0.53
1.3 0.56
1.31 0.54
1.32 0.56
1.33 0.57
1.34 0.61
1.35 0.67
1.36 0.58
1.37 0.59
1.38 0.6
1.39 0.57
1.4 0.62
1.41 0.53
1.42 0.59
1.43 0.5
1.44 0.6
1.45 0.64
1.46 0.65
1.47 0.65
1.48 0.48
1.49 0.6
1.5 0.53
1.51 0.56
1.52 0.58
1.53 0.56
1.54 0.68
1.55 0.58
1.56 0.62
1.57 0.58
1.58 0.57
1.59 0.53
1.6 0.61
1.61 0.5
1.62 0.54
1.63 0.65
1.64 0.55
1.65 0.64
1.66 0.59
1.67 0.5
1.68 0.55
1.69 0.61
1.7 0.61
1.71 0.63
1.72 0.59
1.73 0.59
1.74 0.67
1.75 0.64
1.76 0.59
1.77 0.56
1.78 0.64
1.79 0.53
1.8 0.54
1.81 0.6
1.82 0.52
1.83 0.6
1.84 0.59
1.85 0.55
1.86 0.63
1.87 0.63
1.88 0.61
1.89 0.49
1.9 0.66
1.91 0.56
1.92 0.6
1.93 0.54
1.94 0.45
1.95 0.61
1.96 0.62
1.97 0.58
1.98 0.62
1.99 0.58
2 0.68
};
\addplot [semithick, color11, opacity=0.75, dashed]
table {%
0.01 0.74
0.02 0.81
0.03 0.93
0.04 0.91
0.05 0.97
0.06 0.99
0.07 0.99
0.08 1
0.09 1
0.1 1
0.11 1
0.12 0.99
0.13 1
0.14 1
0.15 1
0.16 1
0.17 1
0.18 1
0.19 1
0.2 1
0.21 1
0.22 1
0.23 1
0.24 1
0.25 1
0.26 1
0.27 1
0.28 1
0.29 1
0.3 1
0.31 1
0.32 1
0.33 1
0.34 1
0.35 1
0.36 1
0.37 1
0.38 1
0.39 1
0.4 1
0.41 1
0.42 1
0.43 1
0.44 1
0.45 1
0.46 1
0.47 1
0.48 1
0.49 1
0.5 1
0.51 1
0.52 1
0.53 1
0.54 1
0.55 1
0.56 1
0.57 1
0.58 1
0.59 1
0.6 1
0.61 1
0.62 1
0.63 1
0.64 1
0.65 1
0.66 1
0.67 1
0.68 1
0.69 1
0.7 1
0.71 1
0.72 1
0.73 1
0.74 1
0.75 1
0.76 1
0.77 1
0.78 1
0.79 1
0.8 1
0.81 1
0.82 1
0.83 1
0.84 1
0.85 1
0.86 1
0.87 1
0.88 1
0.89 1
0.9 1
0.91 1
0.92 1
0.93 1
0.94 1
0.95 1
0.96 1
0.97 1
0.98 1
0.99 1
1 1
1.02 1
1.03 0.99
1.04 0.98
1.05 0.94
1.06 0.94
1.07 0.95
1.08 0.88
1.09 0.92
1.1 0.85
1.11 0.84
1.12 0.84
1.13 0.82
1.14 0.82
1.15 0.81
1.16 0.82
1.17 0.82
1.18 0.79
1.19 0.79
1.2 0.75
1.21 0.83
1.22 0.77
1.23 0.83
1.24 0.74
1.25 0.72
1.26 0.81
1.27 0.81
1.28 0.78
1.29 0.76
1.3 0.79
1.31 0.77
1.32 0.72
1.33 0.76
1.34 0.77
1.35 0.79
1.36 0.73
1.37 0.68
1.38 0.78
1.39 0.75
1.4 0.87
1.41 0.74
1.42 0.73
1.43 0.72
1.44 0.77
1.45 0.84
1.46 0.77
1.47 0.75
1.48 0.75
1.49 0.72
1.5 0.73
1.51 0.72
1.52 0.76
1.53 0.7
1.54 0.75
1.55 0.77
1.56 0.76
1.57 0.78
1.58 0.78
1.59 0.68
1.6 0.79
1.61 0.79
1.62 0.76
1.63 0.8
1.64 0.71
1.65 0.72
1.66 0.71
1.67 0.74
1.68 0.71
1.69 0.71
1.7 0.7
1.71 0.79
1.72 0.71
1.73 0.71
1.74 0.75
1.75 0.82
1.76 0.86
1.77 0.72
1.78 0.77
1.79 0.73
1.8 0.71
1.81 0.75
1.82 0.77
1.83 0.82
1.84 0.78
1.85 0.7
1.86 0.72
1.87 0.79
1.88 0.78
1.89 0.72
1.9 0.75
1.91 0.76
1.92 0.73
1.93 0.72
1.94 0.73
1.95 0.71
1.96 0.78
1.97 0.72
1.98 0.79
1.99 0.79
2 0.77
};
\addplot [semithick, black, opacity=1, dash pattern=on 1pt off 1pt]
table {%
-0.0895000000000001 0.5
2.0995 0.5
};
\addplot [semithick, black, opacity=1, dash pattern=on 1pt off 1pt]
table {%
-0.0895000000000001 1
2.0995 1
};
\addplot [semithick, black, opacity=1, dash pattern=on 1pt off 1pt]
table {%
1 0.2
1 1.05
};
\end{axis}

\end{tikzpicture}

%% file: plots/decoupled2/UNI.tex
% This file was created by tikzplotlib v0.9.6.
\begin{tikzpicture}

\definecolor{color0}{rgb}{0.866666666666667,0.494117647058824,0.164705882352941}
\definecolor{color1}{rgb}{0.164705882352941,0.643137254901961,0.866666666666667}
\definecolor{color2}{rgb}{0.584313725490196,0.866666666666667,0.164705882352941}
\definecolor{color3}{rgb}{0.109803921568627,0.337254901960784,0.129411764705882}
\definecolor{color4}{rgb}{0.529411764705882,0.305882352941176,0.858823529411765}
\definecolor{color5}{rgb}{0.858823529411765,0.305882352941176,0.435294117647059}
\definecolor{color6}{rgb}{0.937254901960784,0.929411764705882,0.392156862745098}
\definecolor{color7}{rgb}{0.0901960784313725,0.486274509803922,0.0980392156862745}
\definecolor{color8}{rgb}{0.156862745098039,0.188235294117647,0.827450980392157}
\definecolor{color9}{rgb}{0.937254901960784,0.392156862745098,0.894117647058824}
\definecolor{color10}{rgb}{0.2,0.184313725490196,0.184313725490196}
\definecolor{color11}{rgb}{0.0156862745098039,0.803921568627451,0.976470588235294}

\begin{axis}[
tick align=outside,
tick pos=left,
x grid style={white!69.0196078431373!black},
xmajorgrids,
xmin=-0.0895, xmax=2.0995,
xtick style={color=black},
xtick={0,0.1,0.2,0.3,0.4,0.5,0.6,0.7,0.8,0.9,1,1.1,1.2,1.3,1.4,1.5,1.6,1.7,1.8,1.9,2},
xticklabels={0,,.2,,.4,,.6,,.8,,1,,20,,40,,60,,80,,100},
height=4.8cm,
width=6.5cm,
y grid style={white!69.0196078431373!black},
ymajorgrids,
ymin=0.2, ymax=1.05,
ytick style={color=black}
]
\addplot [semithick, color0, opacity=0.75]
table {%
0.01 0.54
0.02 0.56
0.03 0.46
0.04 0.69
0.05 0.64
0.06 0.73
0.07 0.66
0.08 0.77
0.09 0.81
0.1 0.85
0.11 0.8
0.12 0.85
0.13 0.91
0.14 0.86
0.15 0.87
0.16 0.87
0.17 0.91
0.18 0.84
0.19 0.92
0.2 0.95
0.21 0.94
0.22 0.94
0.23 0.94
0.24 0.97
0.25 0.97
0.26 0.96
0.27 0.99
0.28 1
0.29 0.97
0.3 0.97
0.31 1
0.32 0.98
0.33 0.99
0.34 1
0.35 1
0.36 1
0.37 1
0.38 1
0.39 0.99
0.4 0.99
0.41 0.99
0.42 1
0.43 1
0.44 1
0.45 0.99
0.46 1
0.47 0.99
0.48 1
0.49 1
0.5 1
0.51 0.99
0.52 0.99
0.53 0.99
0.54 1
0.55 0.99
0.56 1
0.57 1
0.58 1
0.59 1
0.6 1
0.61 1
0.62 1
0.63 1
0.64 1
0.65 1
0.66 1
0.67 1
0.68 1
0.69 1
0.7 1
0.71 1
0.72 1
0.73 1
0.74 1
0.75 1
0.76 1
0.77 1
0.78 1
0.79 1
0.8 1
0.81 1
0.82 1
0.83 1
0.84 1
0.85 1
0.86 1
0.87 1
0.88 1
0.89 1
0.9 1
0.91 1
0.92 1
0.93 0.99
0.94 1
0.95 1
0.96 1
0.97 1
0.98 1
0.99 1
1 1
1.02 0.99
1.03 0.99
1.04 0.97
1.05 0.92
1.06 0.91
1.07 0.86
1.08 0.9
1.09 0.82
1.1 0.79
1.11 0.76
1.12 0.81
1.13 0.75
1.14 0.69
1.15 0.74
1.16 0.71
1.17 0.68
1.18 0.56
1.19 0.61
1.2 0.71
1.21 0.73
1.22 0.59
1.23 0.69
1.24 0.57
1.25 0.6
1.26 0.68
1.27 0.55
1.28 0.62
1.29 0.6
1.3 0.54
1.31 0.49
1.32 0.55
1.33 0.44
1.34 0.66
1.35 0.6
1.36 0.54
1.37 0.48
1.38 0.58
1.39 0.56
1.4 0.56
1.41 0.56
1.42 0.58
1.43 0.6
1.44 0.51
1.45 0.61
1.46 0.49
1.47 0.6
1.48 0.48
1.49 0.52
1.5 0.54
1.51 0.41
1.52 0.54
1.53 0.48
1.54 0.53
1.55 0.37
1.56 0.57
1.57 0.52
1.58 0.57
1.59 0.55
1.6 0.56
1.61 0.55
1.62 0.56
1.63 0.49
1.64 0.5
1.65 0.47
1.66 0.6
1.67 0.59
1.68 0.47
1.69 0.47
1.7 0.47
1.71 0.52
1.72 0.53
1.73 0.51
1.74 0.48
1.75 0.51
1.76 0.51
1.77 0.61
1.78 0.49
1.79 0.44
1.8 0.47
1.81 0.45
1.82 0.46
1.83 0.58
1.84 0.57
1.85 0.5
1.86 0.56
1.87 0.48
1.88 0.54
1.89 0.56
1.9 0.49
1.91 0.46
1.92 0.47
1.93 0.46
1.94 0.54
1.95 0.43
1.96 0.64
1.97 0.49
1.98 0.54
1.99 0.51
2 0.49
};
\addplot [semithick, color1, opacity=0.75]
table {%
0.01 0.55
0.02 0.51
0.03 0.48
0.04 0.53
0.05 0.55
0.06 0.44
0.07 0.52
0.08 0.55
0.09 0.55
0.1 0.55
0.11 0.51
0.12 0.56
0.13 0.63
0.14 0.5
0.15 0.55
0.16 0.55
0.17 0.55
0.18 0.63
0.19 0.64
0.2 0.59
0.21 0.67
0.22 0.58
0.23 0.57
0.24 0.55
0.25 0.59
0.26 0.56
0.27 0.68
0.28 0.64
0.29 0.73
0.3 0.68
0.31 0.73
0.32 0.58
0.33 0.68
0.34 0.67
0.35 0.69
0.36 0.71
0.37 0.61
0.38 0.68
0.39 0.65
0.4 0.62
0.41 0.6
0.42 0.64
0.43 0.67
0.44 0.69
0.45 0.66
0.46 0.67
0.47 0.68
0.48 0.71
0.49 0.66
0.5 0.71
0.51 0.74
0.52 0.72
0.53 0.68
0.54 0.69
0.55 0.73
0.56 0.65
0.57 0.64
0.58 0.76
0.59 0.71
0.6 0.7
0.61 0.72
0.62 0.73
0.63 0.72
0.64 0.72
0.65 0.8
0.66 0.76
0.67 0.71
0.68 0.74
0.69 0.8
0.7 0.76
0.71 0.79
0.72 0.69
0.73 0.74
0.74 0.67
0.75 0.76
0.76 0.79
0.77 0.85
0.78 0.74
0.79 0.77
0.8 0.85
0.81 0.79
0.82 0.74
0.83 0.86
0.84 0.81
0.85 0.76
0.86 0.76
0.87 0.88
0.88 0.81
0.89 0.83
0.9 0.81
0.91 0.81
0.92 0.83
0.93 0.84
0.94 0.88
0.95 0.82
0.96 0.82
0.97 0.84
0.98 0.89
0.99 0.75
1 0.84
1.02 0.85
1.03 0.98
1.04 0.82
1.05 0.86
1.06 0.88
1.07 0.81
1.08 0.82
1.09 0.67
1.1 0.68
1.11 0.73
1.12 0.77
1.13 0.64
1.14 0.71
1.15 0.62
1.16 0.67
1.17 0.64
1.18 0.61
1.19 0.52
1.2 0.6
1.21 0.61
1.22 0.69
1.23 0.52
1.24 0.55
1.25 0.62
1.26 0.52
1.27 0.54
1.28 0.56
1.29 0.6
1.3 0.53
1.31 0.53
1.32 0.53
1.33 0.5
1.34 0.6
1.35 0.55
1.36 0.49
1.37 0.44
1.38 0.52
1.39 0.57
1.4 0.51
1.41 0.48
1.42 0.46
1.43 0.46
1.44 0.55
1.45 0.58
1.46 0.49
1.47 0.54
1.48 0.52
1.49 0.48
1.5 0.46
1.51 0.41
1.52 0.49
1.53 0.52
1.54 0.47
1.55 0.57
1.56 0.39
1.57 0.51
1.58 0.52
1.59 0.45
1.6 0.57
1.61 0.49
1.62 0.47
1.63 0.47
1.64 0.49
1.65 0.46
1.66 0.5
1.67 0.5
1.68 0.51
1.69 0.47
1.7 0.48
1.71 0.47
1.72 0.35
1.73 0.51
1.74 0.52
1.75 0.48
1.76 0.58
1.77 0.6
1.78 0.47
1.79 0.44
1.8 0.45
1.81 0.42
1.82 0.47
1.83 0.43
1.84 0.43
1.85 0.51
1.86 0.39
1.87 0.5
1.88 0.45
1.89 0.49
1.9 0.48
1.91 0.47
1.92 0.43
1.93 0.41
1.94 0.48
1.95 0.42
1.96 0.43
1.97 0.39
1.98 0.44
1.99 0.54
2 0.5
};
\addplot [semithick, color2, opacity=0.75]
table {%
0.01 0.54
0.02 0.46
0.03 0.52
0.04 0.52
0.05 0.44
0.06 0.51
0.07 0.48
0.08 0.59
0.09 0.43
0.1 0.52
0.11 0.55
0.12 0.55
0.13 0.56
0.14 0.43
0.15 0.5
0.16 0.56
0.17 0.58
0.18 0.61
0.19 0.61
0.2 0.55
0.21 0.64
0.22 0.56
0.23 0.56
0.24 0.5
0.25 0.57
0.26 0.53
0.27 0.67
0.28 0.62
0.29 0.72
0.3 0.66
0.31 0.71
0.32 0.55
0.33 0.65
0.34 0.63
0.35 0.65
0.36 0.68
0.37 0.6
0.38 0.67
0.39 0.6
0.4 0.6
0.41 0.58
0.42 0.64
0.43 0.64
0.44 0.66
0.45 0.62
0.46 0.6
0.47 0.65
0.48 0.64
0.49 0.6
0.5 0.65
0.51 0.69
0.52 0.7
0.53 0.66
0.54 0.66
0.55 0.7
0.56 0.62
0.57 0.62
0.58 0.73
0.59 0.7
0.6 0.69
0.61 0.72
0.62 0.72
0.63 0.71
0.64 0.72
0.65 0.81
0.66 0.74
0.67 0.69
0.68 0.73
0.69 0.77
0.7 0.75
0.71 0.78
0.72 0.66
0.73 0.73
0.74 0.66
0.75 0.76
0.76 0.78
0.77 0.84
0.78 0.74
0.79 0.75
0.8 0.83
0.81 0.77
0.82 0.74
0.83 0.86
0.84 0.81
0.85 0.75
0.86 0.78
0.87 0.87
0.88 0.81
0.89 0.84
0.9 0.81
0.91 0.81
0.92 0.83
0.93 0.84
0.94 0.88
0.95 0.82
0.96 0.82
0.97 0.84
0.98 0.88
0.99 0.76
1 0.84
1.02 0.85
1.03 0.98
1.04 0.82
1.05 0.87
1.06 0.88
1.07 0.82
1.08 0.81
1.09 0.66
1.1 0.68
1.11 0.73
1.12 0.75
1.13 0.64
1.14 0.71
1.15 0.64
1.16 0.68
1.17 0.64
1.18 0.61
1.19 0.53
1.2 0.6
1.21 0.62
1.22 0.72
1.23 0.55
1.24 0.56
1.25 0.63
1.26 0.52
1.27 0.52
1.28 0.57
1.29 0.59
1.3 0.52
1.31 0.54
1.32 0.56
1.33 0.5
1.34 0.6
1.35 0.54
1.36 0.51
1.37 0.46
1.38 0.55
1.39 0.57
1.4 0.52
1.41 0.5
1.42 0.43
1.43 0.5
1.44 0.54
1.45 0.61
1.46 0.5
1.47 0.54
1.48 0.53
1.49 0.48
1.5 0.46
1.51 0.42
1.52 0.49
1.53 0.53
1.54 0.48
1.55 0.58
1.56 0.39
1.57 0.51
1.58 0.51
1.59 0.46
1.6 0.59
1.61 0.5
1.62 0.47
1.63 0.45
1.64 0.51
1.65 0.48
1.66 0.49
1.67 0.51
1.68 0.54
1.69 0.46
1.7 0.48
1.71 0.48
1.72 0.37
1.73 0.51
1.74 0.52
1.75 0.48
1.76 0.56
1.77 0.59
1.78 0.45
1.79 0.45
1.8 0.47
1.81 0.41
1.82 0.46
1.83 0.43
1.84 0.42
1.85 0.5
1.86 0.4
1.87 0.51
1.88 0.45
1.89 0.49
1.9 0.48
1.91 0.49
1.92 0.45
1.93 0.43
1.94 0.48
1.95 0.41
1.96 0.45
1.97 0.38
1.98 0.49
1.99 0.52
2 0.5
};
\addplot [semithick, color3, opacity=0.75]
table {%
0.01 0.51
0.02 0.61
0.03 0.53
0.04 0.71
0.05 0.68
0.06 0.75
0.07 0.66
0.08 0.79
0.09 0.81
0.1 0.84
0.11 0.79
0.12 0.88
0.13 0.91
0.14 0.8
0.15 0.9
0.16 0.83
0.17 0.93
0.18 0.89
0.19 0.92
0.2 0.94
0.21 0.92
0.22 0.94
0.23 0.94
0.24 0.95
0.25 0.96
0.26 0.9
0.27 0.97
0.28 0.98
0.29 0.96
0.3 0.91
0.31 0.99
0.32 0.95
0.33 1
0.34 0.97
0.35 0.97
0.36 0.97
0.37 1
0.38 0.99
0.39 0.97
0.4 0.94
0.41 0.95
0.42 0.97
0.43 0.98
0.44 0.96
0.45 0.97
0.46 0.96
0.47 0.97
0.48 0.95
0.49 0.94
0.5 0.96
0.51 0.99
0.52 0.98
0.53 0.93
0.54 0.98
0.55 0.95
0.56 0.93
0.57 0.94
0.58 0.96
0.59 0.91
0.6 0.96
0.61 0.99
0.62 0.96
0.63 0.94
0.64 0.95
0.65 0.99
0.66 0.93
0.67 0.93
0.68 0.92
0.69 0.95
0.7 0.94
0.71 0.93
0.72 0.9
0.73 0.92
0.74 0.92
0.75 0.92
0.76 0.97
0.77 0.95
0.78 0.94
0.79 0.94
0.8 0.97
0.81 0.94
0.82 0.9
0.83 0.96
0.84 0.91
0.85 0.95
0.86 0.87
0.87 0.97
0.88 0.95
0.89 0.92
0.9 0.9
0.91 0.93
0.92 0.92
0.93 0.92
0.94 0.93
0.95 0.93
0.96 0.89
0.97 0.92
0.98 0.98
0.99 0.87
1 0.96
1.02 0.95
1.03 0.97
1.04 0.91
1.05 0.92
1.06 0.87
1.07 0.88
1.08 0.86
1.09 0.84
1.1 0.75
1.11 0.77
1.12 0.84
1.13 0.79
1.14 0.71
1.15 0.77
1.16 0.7
1.17 0.69
1.18 0.63
1.19 0.64
1.2 0.75
1.21 0.71
1.22 0.65
1.23 0.68
1.24 0.53
1.25 0.64
1.26 0.67
1.27 0.61
1.28 0.61
1.29 0.64
1.3 0.65
1.31 0.57
1.32 0.57
1.33 0.47
1.34 0.61
1.35 0.62
1.36 0.59
1.37 0.47
1.38 0.54
1.39 0.55
1.4 0.59
1.41 0.55
1.42 0.61
1.43 0.62
1.44 0.53
1.45 0.54
1.46 0.48
1.47 0.56
1.48 0.54
1.49 0.48
1.5 0.53
1.51 0.37
1.52 0.5
1.53 0.39
1.54 0.52
1.55 0.46
1.56 0.48
1.57 0.53
1.58 0.55
1.59 0.54
1.6 0.56
1.61 0.57
1.62 0.53
1.63 0.55
1.64 0.48
1.65 0.48
1.66 0.62
1.67 0.59
1.68 0.53
1.69 0.51
1.7 0.45
1.71 0.55
1.72 0.44
1.73 0.58
1.74 0.5
1.75 0.55
1.76 0.55
1.77 0.52
1.78 0.49
1.79 0.44
1.8 0.5
1.81 0.43
1.82 0.48
1.83 0.53
1.84 0.52
1.85 0.51
1.86 0.54
1.87 0.5
1.88 0.54
1.89 0.53
1.9 0.44
1.91 0.45
1.92 0.45
1.93 0.53
1.94 0.48
1.95 0.48
1.96 0.53
1.97 0.51
1.98 0.51
1.99 0.58
2 0.48
};
\addplot [semithick, color4, opacity=0.75]
table {%
0.01 0.51
0.02 0.59
0.03 0.53
0.04 0.71
0.05 0.68
0.06 0.75
0.07 0.67
0.08 0.79
0.09 0.82
0.1 0.85
0.11 0.8
0.12 0.88
0.13 0.91
0.14 0.85
0.15 0.91
0.16 0.83
0.17 0.93
0.18 0.9
0.19 0.92
0.2 0.94
0.21 0.92
0.22 0.94
0.23 0.97
0.24 0.96
0.25 0.96
0.26 0.91
0.27 0.97
0.28 0.98
0.29 0.96
0.3 0.91
0.31 0.99
0.32 0.98
0.33 1
0.34 0.98
0.35 0.98
0.36 0.98
0.37 1
0.38 0.99
0.39 0.97
0.4 0.97
0.41 0.97
0.42 0.98
0.43 0.99
0.44 0.99
0.45 0.97
0.46 0.97
0.47 0.98
0.48 0.97
0.49 0.95
0.5 0.98
0.51 0.99
0.52 0.98
0.53 0.95
0.54 0.99
0.55 0.97
0.56 0.94
0.57 0.96
0.58 0.97
0.59 0.96
0.6 1
0.61 1
0.62 1
0.63 0.97
0.64 0.97
0.65 1
0.66 0.96
0.67 0.96
0.68 0.98
0.69 0.99
0.7 0.96
0.71 0.96
0.72 0.95
0.73 0.96
0.74 0.95
0.75 0.97
0.76 0.99
0.77 0.98
0.78 0.97
0.79 0.96
0.8 0.97
0.81 0.97
0.82 0.92
0.83 0.99
0.84 0.94
0.85 0.98
0.86 0.96
0.87 0.98
0.88 0.97
0.89 0.93
0.9 0.93
0.91 0.96
0.92 0.98
0.93 0.97
0.94 0.94
0.95 0.94
0.96 0.92
0.97 0.96
0.98 0.99
0.99 0.93
1 0.96
1.02 0.96
1.03 0.98
1.04 0.91
1.05 0.92
1.06 0.89
1.07 0.91
1.08 0.88
1.09 0.85
1.1 0.76
1.11 0.79
1.12 0.86
1.13 0.79
1.14 0.74
1.15 0.77
1.16 0.71
1.17 0.69
1.18 0.64
1.19 0.65
1.2 0.75
1.21 0.72
1.22 0.65
1.23 0.68
1.24 0.53
1.25 0.66
1.26 0.69
1.27 0.61
1.28 0.61
1.29 0.64
1.3 0.65
1.31 0.56
1.32 0.58
1.33 0.48
1.34 0.62
1.35 0.64
1.36 0.59
1.37 0.47
1.38 0.56
1.39 0.55
1.4 0.59
1.41 0.56
1.42 0.59
1.43 0.62
1.44 0.54
1.45 0.56
1.46 0.47
1.47 0.57
1.48 0.54
1.49 0.47
1.5 0.53
1.51 0.37
1.52 0.52
1.53 0.4
1.54 0.53
1.55 0.45
1.56 0.47
1.57 0.52
1.58 0.55
1.59 0.54
1.6 0.56
1.61 0.57
1.62 0.54
1.63 0.54
1.64 0.48
1.65 0.49
1.66 0.62
1.67 0.58
1.68 0.52
1.69 0.5
1.7 0.45
1.71 0.55
1.72 0.44
1.73 0.56
1.74 0.5
1.75 0.55
1.76 0.54
1.77 0.52
1.78 0.47
1.79 0.46
1.8 0.51
1.81 0.43
1.82 0.48
1.83 0.52
1.84 0.52
1.85 0.52
1.86 0.54
1.87 0.5
1.88 0.55
1.89 0.52
1.9 0.44
1.91 0.46
1.92 0.45
1.93 0.53
1.94 0.48
1.95 0.47
1.96 0.56
1.97 0.51
1.98 0.52
1.99 0.58
2 0.48
};
\addplot [semithick, color5, opacity=0.75]
table {%
0.01 0.52
0.02 0.52
0.03 0.54
0.04 0.65
0.05 0.65
0.06 0.74
0.07 0.64
0.08 0.78
0.09 0.82
0.1 0.81
0.11 0.78
0.12 0.89
0.13 0.88
0.14 0.82
0.15 0.85
0.16 0.78
0.17 0.9
0.18 0.88
0.19 0.88
0.2 0.9
0.21 0.89
0.22 0.92
0.23 0.9
0.24 0.93
0.25 0.95
0.26 0.9
0.27 0.95
0.28 0.97
0.29 0.95
0.3 0.92
0.31 0.98
0.32 0.9
0.33 0.99
0.34 0.96
0.35 0.95
0.36 0.97
0.37 1
0.38 0.94
0.39 0.96
0.4 0.91
0.41 0.87
0.42 0.96
0.43 0.99
0.44 0.95
0.45 0.95
0.46 0.93
0.47 0.97
0.48 0.95
0.49 0.94
0.5 0.95
0.51 0.95
0.52 0.96
0.53 0.9
0.54 0.99
0.55 0.93
0.56 0.89
0.57 0.91
0.58 0.95
0.59 0.92
0.6 0.95
0.61 0.97
0.62 0.93
0.63 0.93
0.64 0.91
0.65 0.98
0.66 0.93
0.67 0.9
0.68 0.89
0.69 0.95
0.7 0.9
0.71 0.9
0.72 0.86
0.73 0.88
0.74 0.89
0.75 0.9
0.76 0.92
0.77 0.95
0.78 0.9
0.79 0.92
0.8 0.96
0.81 0.91
0.82 0.88
0.83 0.95
0.84 0.89
0.85 0.92
0.86 0.86
0.87 0.96
0.88 0.92
0.89 0.86
0.9 0.85
0.91 0.9
0.92 0.9
0.93 0.89
0.94 0.89
0.95 0.9
0.96 0.88
0.97 0.88
0.98 0.93
0.99 0.84
1 0.88
1.02 0.91
1.03 0.95
1.04 0.87
1.05 0.9
1.06 0.84
1.07 0.87
1.08 0.81
1.09 0.83
1.1 0.76
1.11 0.71
1.12 0.79
1.13 0.72
1.14 0.74
1.15 0.76
1.16 0.63
1.17 0.68
1.18 0.6
1.19 0.54
1.2 0.7
1.21 0.62
1.22 0.59
1.23 0.62
1.24 0.53
1.25 0.59
1.26 0.62
1.27 0.55
1.28 0.57
1.29 0.55
1.3 0.58
1.31 0.49
1.32 0.48
1.33 0.55
1.34 0.59
1.35 0.57
1.36 0.6
1.37 0.47
1.38 0.56
1.39 0.63
1.4 0.6
1.41 0.61
1.42 0.56
1.43 0.56
1.44 0.54
1.45 0.62
1.46 0.42
1.47 0.49
1.48 0.54
1.49 0.45
1.5 0.54
1.51 0.39
1.52 0.47
1.53 0.47
1.54 0.49
1.55 0.54
1.56 0.47
1.57 0.55
1.58 0.52
1.59 0.55
1.6 0.57
1.61 0.56
1.62 0.53
1.63 0.56
1.64 0.6
1.65 0.5
1.66 0.52
1.67 0.59
1.68 0.5
1.69 0.49
1.7 0.39
1.71 0.51
1.72 0.51
1.73 0.58
1.74 0.51
1.75 0.49
1.76 0.52
1.77 0.51
1.78 0.5
1.79 0.45
1.8 0.47
1.81 0.39
1.82 0.48
1.83 0.53
1.84 0.53
1.85 0.53
1.86 0.52
1.87 0.56
1.88 0.53
1.89 0.56
1.9 0.51
1.91 0.45
1.92 0.5
1.93 0.54
1.94 0.57
1.95 0.48
1.96 0.58
1.97 0.55
1.98 0.5
1.99 0.53
2 0.48
};
\addplot [semithick, color6, opacity=0.75, dashed]
table {%
0.01 0.48
0.02 0.47
0.03 0.52
0.04 0.59
0.05 0.65
0.06 0.65
0.07 0.76
0.08 0.67
0.09 0.66
0.1 0.71
0.11 0.72
0.12 0.76
0.13 0.76
0.14 0.78
0.15 0.82
0.16 0.87
0.17 0.83
0.18 0.86
0.19 0.87
0.2 0.93
0.21 0.92
0.22 0.93
0.23 0.87
0.24 0.84
0.25 0.95
0.26 0.94
0.27 0.99
0.28 0.94
0.29 0.88
0.3 0.96
0.31 0.93
0.32 0.95
0.33 0.97
0.34 0.97
0.35 0.99
0.36 0.99
0.37 0.95
0.38 0.95
0.39 0.96
0.4 0.98
0.41 1
0.42 0.98
0.43 0.97
0.44 0.98
0.45 0.94
0.46 0.99
0.47 0.97
0.48 0.99
0.49 1
0.5 0.96
0.51 1
0.52 0.99
0.53 0.97
0.54 1
0.55 1
0.56 0.99
0.57 0.99
0.58 0.98
0.59 0.98
0.6 0.99
0.61 1
0.62 1
0.63 0.99
0.64 0.98
0.65 0.99
0.66 0.98
0.67 0.98
0.68 0.99
0.69 1
0.7 0.98
0.71 0.99
0.72 0.99
0.73 1
0.74 0.99
0.75 1
0.76 1
0.77 0.98
0.78 1
0.79 0.98
0.8 1
0.81 1
0.82 0.98
0.83 1
0.84 1
0.85 1
0.86 0.99
0.87 0.99
0.88 1
0.89 0.98
0.9 0.99
0.91 1
0.92 0.97
0.93 0.97
0.94 0.97
0.95 1
0.96 1
0.97 1
0.98 0.98
0.99 1
1 1
1.02 0.98
1.03 1
1.04 0.85
1.05 0.86
1.06 0.84
1.07 0.8
1.08 0.74
1.09 0.73
1.1 0.7
1.11 0.69
1.12 0.77
1.13 0.65
1.14 0.72
1.15 0.61
1.16 0.61
1.17 0.57
1.18 0.61
1.19 0.64
1.2 0.64
1.21 0.65
1.22 0.55
1.23 0.64
1.24 0.58
1.25 0.62
1.26 0.56
1.27 0.53
1.28 0.64
1.29 0.45
1.3 0.51
1.31 0.58
1.32 0.57
1.33 0.59
1.34 0.57
1.35 0.5
1.36 0.51
1.37 0.56
1.38 0.53
1.39 0.49
1.4 0.54
1.41 0.52
1.42 0.55
1.43 0.56
1.44 0.47
1.45 0.57
1.46 0.49
1.47 0.47
1.48 0.55
1.49 0.51
1.5 0.55
1.51 0.45
1.52 0.59
1.53 0.48
1.54 0.51
1.55 0.55
1.56 0.55
1.57 0.48
1.58 0.46
1.59 0.52
1.6 0.47
1.61 0.46
1.62 0.47
1.63 0.51
1.64 0.48
1.65 0.52
1.66 0.51
1.67 0.5
1.68 0.45
1.69 0.46
1.7 0.53
1.71 0.47
1.72 0.51
1.73 0.55
1.74 0.45
1.75 0.55
1.76 0.5
1.77 0.54
1.78 0.43
1.79 0.5
1.8 0.54
1.81 0.5
1.82 0.51
1.83 0.51
1.84 0.54
1.85 0.55
1.86 0.46
1.87 0.55
1.88 0.51
1.89 0.49
1.9 0.51
1.91 0.53
1.92 0.42
1.93 0.5
1.94 0.5
1.95 0.48
1.96 0.42
1.97 0.5
1.98 0.49
1.99 0.52
2 0.54
};
\addplot [semithick, color7, opacity=0.75, dashed]
table {%
0.01 0.48
0.02 0.47
0.03 0.52
0.04 0.59
0.05 0.65
0.06 0.65
0.07 0.76
0.08 0.67
0.09 0.66
0.1 0.71
0.11 0.72
0.12 0.76
0.13 0.76
0.14 0.78
0.15 0.82
0.16 0.87
0.17 0.83
0.18 0.86
0.19 0.87
0.2 0.93
0.21 0.92
0.22 0.93
0.23 0.87
0.24 0.84
0.25 0.95
0.26 0.94
0.27 0.99
0.28 0.94
0.29 0.88
0.3 0.96
0.31 0.93
0.32 0.95
0.33 0.97
0.34 0.97
0.35 0.99
0.36 0.99
0.37 0.95
0.38 0.95
0.39 0.96
0.4 0.98
0.41 1
0.42 0.98
0.43 0.97
0.44 0.98
0.45 0.94
0.46 0.99
0.47 0.97
0.48 0.99
0.49 1
0.5 0.96
0.51 1
0.52 0.99
0.53 0.97
0.54 1
0.55 1
0.56 0.99
0.57 0.99
0.58 0.98
0.59 0.98
0.6 0.99
0.61 1
0.62 1
0.63 0.99
0.64 0.98
0.65 0.99
0.66 0.98
0.67 0.98
0.68 0.99
0.69 1
0.7 0.98
0.71 0.99
0.72 0.99
0.73 1
0.74 0.99
0.75 1
0.76 1
0.77 0.98
0.78 1
0.79 0.98
0.8 1
0.81 1
0.82 0.98
0.83 1
0.84 1
0.85 1
0.86 0.99
0.87 0.99
0.88 1
0.89 0.98
0.9 0.99
0.91 1
0.92 0.97
0.93 0.97
0.94 0.97
0.95 1
0.96 1
0.97 1
0.98 0.98
0.99 1
1 1
1.02 0.98
1.03 1
1.04 0.85
1.05 0.86
1.06 0.84
1.07 0.8
1.08 0.74
1.09 0.73
1.1 0.7
1.11 0.69
1.12 0.77
1.13 0.65
1.14 0.72
1.15 0.61
1.16 0.61
1.17 0.57
1.18 0.61
1.19 0.64
1.2 0.64
1.21 0.65
1.22 0.55
1.23 0.64
1.24 0.58
1.25 0.62
1.26 0.56
1.27 0.53
1.28 0.64
1.29 0.45
1.3 0.51
1.31 0.58
1.32 0.57
1.33 0.59
1.34 0.57
1.35 0.5
1.36 0.51
1.37 0.56
1.38 0.53
1.39 0.49
1.4 0.54
1.41 0.52
1.42 0.55
1.43 0.56
1.44 0.47
1.45 0.57
1.46 0.49
1.47 0.47
1.48 0.55
1.49 0.51
1.5 0.55
1.51 0.45
1.52 0.59
1.53 0.48
1.54 0.51
1.55 0.55
1.56 0.55
1.57 0.48
1.58 0.46
1.59 0.52
1.6 0.47
1.61 0.46
1.62 0.47
1.63 0.51
1.64 0.48
1.65 0.52
1.66 0.51
1.67 0.5
1.68 0.45
1.69 0.46
1.7 0.53
1.71 0.47
1.72 0.51
1.73 0.55
1.74 0.45
1.75 0.55
1.76 0.5
1.77 0.54
1.78 0.43
1.79 0.5
1.8 0.54
1.81 0.5
1.82 0.51
1.83 0.51
1.84 0.54
1.85 0.55
1.86 0.46
1.87 0.55
1.88 0.51
1.89 0.49
1.9 0.51
1.91 0.53
1.92 0.42
1.93 0.5
1.94 0.5
1.95 0.48
1.96 0.42
1.97 0.5
1.98 0.49
1.99 0.52
2 0.54
};
\addplot [semithick, color8, opacity=0.75, dashed]
table {%
0.01 0.48
0.02 0.54
0.03 0.64
0.04 0.65
0.05 0.67
0.06 0.73
0.07 0.78
0.08 0.75
0.09 0.7
0.1 0.76
0.11 0.73
0.12 0.81
0.13 0.8
0.14 0.87
0.15 0.84
0.16 0.85
0.17 0.94
0.18 0.92
0.19 0.9
0.2 0.89
0.21 0.96
0.22 0.94
0.23 0.9
0.24 0.94
0.25 0.97
0.26 0.94
0.27 1
0.28 0.97
0.29 0.98
0.3 0.97
0.31 0.99
0.32 0.98
0.33 0.97
0.34 0.99
0.35 0.97
0.36 0.98
0.37 0.98
0.38 0.98
0.39 0.99
0.4 0.98
0.41 1
0.42 1
0.43 0.99
0.44 0.98
0.45 0.99
0.46 1
0.47 1
0.48 1
0.49 1
0.5 1
0.51 0.99
0.52 1
0.53 1
0.54 0.99
0.55 1
0.56 1
0.57 0.99
0.58 1
0.59 1
0.6 1
0.61 1
0.62 1
0.63 1
0.64 1
0.65 1
0.66 0.98
0.67 1
0.68 1
0.69 1
0.7 1
0.71 0.99
0.72 1
0.73 1
0.74 1
0.75 1
0.76 1
0.77 1
0.78 1
0.79 0.99
0.8 1
0.81 1
0.82 1
0.83 1
0.84 1
0.85 1
0.86 1
0.87 1
0.88 1
0.89 1
0.9 0.98
0.91 1
0.92 1
0.93 1
0.94 1
0.95 1
0.96 1
0.97 1
0.98 0.99
0.99 1
1 1
1.02 0.99
1.03 1
1.04 0.94
1.05 0.95
1.06 0.89
1.07 0.84
1.08 0.79
1.09 0.8
1.1 0.73
1.11 0.7
1.12 0.68
1.13 0.7
1.14 0.7
1.15 0.66
1.16 0.64
1.17 0.56
1.18 0.64
1.19 0.59
1.2 0.54
1.21 0.6
1.22 0.61
1.23 0.59
1.24 0.55
1.25 0.59
1.26 0.51
1.27 0.54
1.28 0.65
1.29 0.57
1.3 0.51
1.31 0.53
1.32 0.62
1.33 0.63
1.34 0.58
1.35 0.61
1.36 0.6
1.37 0.52
1.38 0.56
1.39 0.54
1.4 0.52
1.41 0.55
1.42 0.65
1.43 0.51
1.44 0.46
1.45 0.51
1.46 0.5
1.47 0.57
1.48 0.49
1.49 0.41
1.5 0.53
1.51 0.5
1.52 0.52
1.53 0.6
1.54 0.48
1.55 0.57
1.56 0.65
1.57 0.54
1.58 0.55
1.59 0.45
1.6 0.49
1.61 0.57
1.62 0.59
1.63 0.54
1.64 0.48
1.65 0.57
1.66 0.45
1.67 0.56
1.68 0.54
1.69 0.51
1.7 0.48
1.71 0.48
1.72 0.5
1.73 0.54
1.74 0.52
1.75 0.48
1.76 0.57
1.77 0.41
1.78 0.53
1.79 0.53
1.8 0.54
1.81 0.54
1.82 0.56
1.83 0.41
1.84 0.57
1.85 0.59
1.86 0.51
1.87 0.45
1.88 0.47
1.89 0.48
1.9 0.49
1.91 0.56
1.92 0.39
1.93 0.53
1.94 0.5
1.95 0.48
1.96 0.49
1.97 0.42
1.98 0.5
1.99 0.5
2 0.48
};
\addplot [semithick, color9, opacity=0.75, dashed]
table {%
0.01 0.57
0.02 0.51
0.03 0.49
0.04 0.64
0.05 0.6
0.06 0.64
0.07 0.58
0.08 0.71
0.09 0.73
0.1 0.73
0.11 0.69
0.12 0.73
0.13 0.88
0.14 0.76
0.15 0.8
0.16 0.8
0.17 0.79
0.18 0.78
0.19 0.81
0.2 0.85
0.21 0.87
0.22 0.89
0.23 0.9
0.24 0.97
0.25 0.94
0.26 0.94
0.27 0.95
0.28 0.97
0.29 0.95
0.3 0.93
0.31 0.96
0.32 0.99
0.33 1
0.34 0.99
0.35 0.99
0.36 0.96
0.37 1
0.38 0.97
0.39 1
0.4 0.96
0.41 0.99
0.42 0.97
0.43 0.99
0.44 1
0.45 0.98
0.46 0.98
0.47 0.99
0.48 1
0.49 1
0.5 1
0.51 0.99
0.52 1
0.53 0.99
0.54 1
0.55 0.99
0.56 1
0.57 1
0.58 1
0.59 1
0.6 1
0.61 1
0.62 1
0.63 1
0.64 1
0.65 1
0.66 1
0.67 1
0.68 1
0.69 1
0.7 1
0.71 1
0.72 1
0.73 1
0.74 1
0.75 1
0.76 1
0.77 1
0.78 1
0.79 1
0.8 1
0.81 1
0.82 1
0.83 0.99
0.84 1
0.85 1
0.86 1
0.87 1
0.88 1
0.89 1
0.9 0.99
0.91 1
0.92 1
0.93 1
0.94 1
0.95 1
0.96 1
0.97 1
0.98 1
0.99 1
1 1
1.02 1
1.03 0.98
1.04 0.93
1.05 0.92
1.06 0.87
1.07 0.76
1.08 0.77
1.09 0.7
1.1 0.66
1.11 0.7
1.12 0.7
1.13 0.69
1.14 0.6
1.15 0.69
1.16 0.69
1.17 0.62
1.18 0.58
1.19 0.66
1.2 0.66
1.21 0.62
1.22 0.6
1.23 0.51
1.24 0.58
1.25 0.61
1.26 0.52
1.27 0.57
1.28 0.59
1.29 0.56
1.3 0.55
1.31 0.48
1.32 0.49
1.33 0.5
1.34 0.65
1.35 0.55
1.36 0.49
1.37 0.47
1.38 0.56
1.39 0.49
1.4 0.48
1.41 0.51
1.42 0.63
1.43 0.55
1.44 0.59
1.45 0.61
1.46 0.41
1.47 0.54
1.48 0.51
1.49 0.52
1.5 0.43
1.51 0.5
1.52 0.54
1.53 0.48
1.54 0.55
1.55 0.48
1.56 0.54
1.57 0.47
1.58 0.61
1.59 0.48
1.6 0.57
1.61 0.49
1.62 0.52
1.63 0.5
1.64 0.53
1.65 0.45
1.66 0.51
1.67 0.56
1.68 0.53
1.69 0.53
1.7 0.47
1.71 0.59
1.72 0.5
1.73 0.46
1.74 0.46
1.75 0.58
1.76 0.49
1.77 0.51
1.78 0.57
1.79 0.46
1.8 0.49
1.81 0.44
1.82 0.45
1.83 0.55
1.84 0.63
1.85 0.52
1.86 0.49
1.87 0.55
1.88 0.46
1.89 0.59
1.9 0.44
1.91 0.44
1.92 0.51
1.93 0.46
1.94 0.53
1.95 0.51
1.96 0.64
1.97 0.54
1.98 0.41
1.99 0.48
2 0.49
};
\addplot [semithick, color10, opacity=0.75, dashed]
table {%
0.01 0.55
0.02 0.57
0.03 0.52
0.04 0.73
0.05 0.68
0.06 0.78
0.07 0.73
0.08 0.82
0.09 0.87
0.1 0.87
0.11 0.84
0.12 0.92
0.13 0.97
0.14 0.91
0.15 0.93
0.16 0.93
0.17 0.93
0.18 0.94
0.19 0.97
0.2 0.99
0.21 0.93
0.22 0.98
0.23 1
0.24 1
0.25 1
0.26 0.98
0.27 1
0.28 1
0.29 1
0.3 0.99
0.31 1
0.32 1
0.33 1
0.34 1
0.35 1
0.36 1
0.37 1
0.38 1
0.39 1
0.4 0.99
0.41 1
0.42 1
0.43 1
0.44 1
0.45 1
0.46 1
0.47 1
0.48 1
0.49 1
0.5 1
0.51 1
0.52 1
0.53 1
0.54 1
0.55 1
0.56 1
0.57 1
0.58 1
0.59 1
0.6 1
0.61 1
0.62 1
0.63 1
0.64 1
0.65 1
0.66 1
0.67 1
0.68 1
0.69 1
0.7 1
0.71 1
0.72 1
0.73 1
0.74 1
0.75 1
0.76 1
0.77 1
0.78 1
0.79 1
0.8 1
0.81 1
0.82 1
0.83 1
0.84 1
0.85 1
0.86 1
0.87 1
0.88 1
0.89 1
0.9 1
0.91 1
0.92 1
0.93 1
0.94 1
0.95 1
0.96 1
0.97 1
0.98 1
0.99 1
1 1
1.02 1
1.03 0.99
1.04 0.99
1.05 0.95
1.06 0.95
1.07 0.92
1.08 0.94
1.09 0.88
1.1 0.89
1.11 0.82
1.12 0.87
1.13 0.84
1.14 0.74
1.15 0.8
1.16 0.72
1.17 0.75
1.18 0.65
1.19 0.67
1.2 0.68
1.21 0.7
1.22 0.64
1.23 0.7
1.24 0.62
1.25 0.64
1.26 0.7
1.27 0.65
1.28 0.6
1.29 0.6
1.3 0.59
1.31 0.48
1.32 0.63
1.33 0.51
1.34 0.71
1.35 0.58
1.36 0.56
1.37 0.51
1.38 0.55
1.39 0.57
1.4 0.55
1.41 0.57
1.42 0.6
1.43 0.65
1.44 0.57
1.45 0.6
1.46 0.48
1.47 0.6
1.48 0.49
1.49 0.49
1.5 0.5
1.51 0.41
1.52 0.49
1.53 0.44
1.54 0.62
1.55 0.46
1.56 0.53
1.57 0.47
1.58 0.57
1.59 0.56
1.6 0.56
1.61 0.49
1.62 0.52
1.63 0.47
1.64 0.49
1.65 0.49
1.66 0.61
1.67 0.62
1.68 0.46
1.69 0.46
1.7 0.5
1.71 0.59
1.72 0.48
1.73 0.54
1.74 0.46
1.75 0.54
1.76 0.5
1.77 0.61
1.78 0.48
1.79 0.43
1.8 0.52
1.81 0.43
1.82 0.53
1.83 0.57
1.84 0.5
1.85 0.44
1.86 0.58
1.87 0.54
1.88 0.53
1.89 0.57
1.9 0.48
1.91 0.43
1.92 0.44
1.93 0.52
1.94 0.55
1.95 0.45
1.96 0.58
1.97 0.47
1.98 0.53
1.99 0.52
2 0.47
};
\addplot [semithick, color11, opacity=0.75, dashed]
table {%
0.01 0.57
0.02 0.76
0.03 0.77
0.04 0.89
0.05 0.9
0.06 0.94
0.07 0.94
0.08 0.98
0.09 0.95
0.1 0.99
0.11 0.97
0.12 1
0.13 1
0.14 1
0.15 1
0.16 1
0.17 1
0.18 1
0.19 1
0.2 1
0.21 1
0.22 1
0.23 1
0.24 1
0.25 1
0.26 1
0.27 1
0.28 1
0.29 1
0.3 1
0.31 1
0.32 1
0.33 1
0.34 1
0.35 1
0.36 1
0.37 1
0.38 1
0.39 1
0.4 1
0.41 1
0.42 1
0.43 1
0.44 1
0.45 1
0.46 1
0.47 1
0.48 1
0.49 1
0.5 1
0.51 1
0.52 1
0.53 1
0.54 1
0.55 1
0.56 1
0.57 1
0.58 1
0.59 1
0.6 1
0.61 1
0.62 1
0.63 1
0.64 1
0.65 1
0.66 1
0.67 1
0.68 1
0.69 1
0.7 1
0.71 1
0.72 1
0.73 1
0.74 1
0.75 1
0.76 1
0.77 1
0.78 1
0.79 1
0.8 1
0.81 1
0.82 1
0.83 1
0.84 1
0.85 1
0.86 1
0.87 1
0.88 1
0.89 1
0.9 1
0.91 1
0.92 1
0.93 1
0.94 1
0.95 1
0.96 1
0.97 1
0.98 1
0.99 1
1 1
1.02 1
1.03 1
1.04 1
1.05 1
1.06 1
1.07 1
1.08 1
1.09 1
1.1 1
1.11 0.97
1.12 1
1.13 1
1.14 0.96
1.15 0.97
1.16 0.9
1.17 0.93
1.18 0.88
1.19 0.89
1.2 0.92
1.21 0.88
1.22 0.87
1.23 0.87
1.24 0.76
1.25 0.86
1.26 0.87
1.27 0.74
1.28 0.83
1.29 0.75
1.3 0.8
1.31 0.75
1.32 0.83
1.33 0.75
1.34 0.78
1.35 0.7
1.36 0.68
1.37 0.65
1.38 0.71
1.39 0.73
1.4 0.71
1.41 0.73
1.42 0.74
1.43 0.72
1.44 0.71
1.45 0.7
1.46 0.69
1.47 0.67
1.48 0.64
1.49 0.58
1.5 0.7
1.51 0.6
1.52 0.65
1.53 0.59
1.54 0.62
1.55 0.63
1.56 0.58
1.57 0.55
1.58 0.73
1.59 0.68
1.6 0.69
1.61 0.66
1.62 0.64
1.63 0.64
1.64 0.59
1.65 0.6
1.66 0.6
1.67 0.67
1.68 0.63
1.69 0.58
1.7 0.57
1.71 0.54
1.72 0.64
1.73 0.56
1.74 0.59
1.75 0.58
1.76 0.58
1.77 0.68
1.78 0.56
1.79 0.6
1.8 0.51
1.81 0.53
1.82 0.57
1.83 0.6
1.84 0.55
1.85 0.64
1.86 0.51
1.87 0.65
1.88 0.6
1.89 0.58
1.9 0.5
1.91 0.6
1.92 0.61
1.93 0.57
1.94 0.62
1.95 0.54
1.96 0.54
1.97 0.67
1.98 0.55
1.99 0.53
2 0.48
};
\addplot [semithick, black, opacity=1, dash pattern=on 1pt off 1pt]
table {%
-0.0895000000000001 0.5
2.0995 0.5
};
\addplot [semithick, black, opacity=1, dash pattern=on 1pt off 1pt]
table {%
-0.0895000000000001 1
2.0995 1
};
\addplot [semithick, black, opacity=1, dash pattern=on 1pt off 1pt]
table {%
1 0.2
1 1.05
};
\end{axis}

\end{tikzpicture}

%% file: plots/decoupled2/UNIxLAP.tex
% This file was created by tikzplotlib v0.9.6.
\begin{tikzpicture}

\definecolor{color0}{rgb}{0.866666666666667,0.494117647058824,0.164705882352941}
\definecolor{color1}{rgb}{0.164705882352941,0.643137254901961,0.866666666666667}
\definecolor{color2}{rgb}{0.584313725490196,0.866666666666667,0.164705882352941}
\definecolor{color3}{rgb}{0.109803921568627,0.337254901960784,0.129411764705882}
\definecolor{color4}{rgb}{0.529411764705882,0.305882352941176,0.858823529411765}
\definecolor{color5}{rgb}{0.858823529411765,0.305882352941176,0.435294117647059}
\definecolor{color6}{rgb}{0.937254901960784,0.929411764705882,0.392156862745098}
\definecolor{color7}{rgb}{0.0901960784313725,0.486274509803922,0.0980392156862745}
\definecolor{color8}{rgb}{0.156862745098039,0.188235294117647,0.827450980392157}
\definecolor{color9}{rgb}{0.937254901960784,0.392156862745098,0.894117647058824}
\definecolor{color10}{rgb}{0.2,0.184313725490196,0.184313725490196}
\definecolor{color11}{rgb}{0.0156862745098039,0.803921568627451,0.976470588235294}

\begin{axis}[
tick align=outside,
tick pos=left,
x grid style={white!69.0196078431373!black},
xmajorgrids,
xmin=-0.0895, xmax=2.0995,
xtick style={color=black},
xtick={0,0.1,0.2,0.3,0.4,0.5,0.6,0.7,0.8,0.9,1,1.1,1.2,1.3,1.4,1.5,1.6,1.7,1.8,1.9,2},
xticklabels={0,,.2,,.4,,.6,,.8,,1,,20,,40,,60,,80,,100},
height=4.8cm,
width=6.5cm,
y grid style={white!69.0196078431373!black},
ymajorgrids,
ymin=0.2, ymax=1.05,
ytick style={color=black}
]
\addplot [semithick, color0, opacity=0.75]
table {%
0.01 0.5
0.02 0.52
0.03 0.64
0.04 0.63
0.05 0.73
0.06 0.67
0.07 0.78
0.08 0.75
0.09 0.78
0.1 0.83
0.11 0.85
0.12 0.91
0.13 0.86
0.14 0.9
0.15 0.89
0.16 0.93
0.17 0.94
0.18 0.92
0.19 0.95
0.2 0.98
0.21 0.96
0.22 0.97
0.23 0.95
0.24 0.97
0.25 0.97
0.26 1
0.27 0.99
0.28 1
0.29 0.99
0.3 0.99
0.31 0.99
0.32 1
0.33 1
0.34 1
0.35 1
0.36 1
0.37 1
0.38 1
0.39 1
0.4 1
0.41 1
0.42 1
0.43 1
0.44 1
0.45 1
0.46 1
0.47 1
0.48 1
0.49 1
0.5 1
0.51 1
0.52 1
0.53 1
0.54 1
0.55 1
0.56 1
0.57 1
0.58 1
0.59 1
0.6 1
0.61 1
0.62 1
0.63 1
0.64 1
0.65 1
0.66 1
0.67 1
0.68 1
0.69 1
0.7 1
0.71 1
0.72 1
0.73 1
0.74 1
0.75 1
0.76 1
0.77 1
0.78 1
0.79 1
0.8 1
0.81 1
0.82 1
0.83 1
0.84 1
0.85 1
0.86 1
0.87 0.99
0.88 1
0.89 1
0.9 1
0.91 1
0.92 1
0.93 1
0.94 1
0.95 1
0.96 0.98
0.97 1
0.98 1
0.99 1
1 1
1.02 0.9
1.03 0.88
1.04 0.75
1.05 0.74
1.06 0.6
1.07 0.65
1.08 0.69
1.09 0.66
1.1 0.58
1.11 0.53
1.12 0.59
1.13 0.59
1.14 0.53
1.15 0.53
1.16 0.46
1.17 0.5
1.18 0.51
1.19 0.48
1.2 0.46
1.21 0.5
1.22 0.44
1.23 0.4
1.24 0.38
1.25 0.41
1.26 0.49
1.27 0.42
1.28 0.34
1.29 0.57
1.3 0.48
1.31 0.44
1.32 0.36
1.33 0.42
1.34 0.48
1.35 0.34
1.36 0.37
1.37 0.39
1.38 0.5
1.39 0.4
1.4 0.44
1.41 0.4
1.42 0.47
1.43 0.37
1.44 0.47
1.45 0.4
1.46 0.34
1.47 0.4
1.48 0.33
1.49 0.42
1.5 0.35
1.51 0.33
1.52 0.44
1.53 0.36
1.54 0.4
1.55 0.41
1.56 0.36
1.57 0.4
1.58 0.47
1.59 0.37
1.6 0.52
1.61 0.38
1.62 0.46
1.63 0.46
1.64 0.47
1.65 0.44
1.66 0.37
1.67 0.4
1.68 0.4
1.69 0.31
1.7 0.37
1.71 0.41
1.72 0.34
1.73 0.35
1.74 0.38
1.75 0.38
1.76 0.41
1.77 0.47
1.78 0.41
1.79 0.33
1.8 0.45
1.81 0.42
1.82 0.37
1.83 0.41
1.84 0.48
1.85 0.38
1.86 0.49
1.87 0.41
1.88 0.45
1.89 0.35
1.9 0.35
1.91 0.38
1.92 0.37
1.93 0.42
1.94 0.48
1.95 0.43
1.96 0.35
1.97 0.41
1.98 0.38
1.99 0.45
2 0.41
};
\addplot [semithick, color1, opacity=0.75]
table {%
0.01 0.47
0.02 0.54
0.03 0.61
0.04 0.56
0.05 0.61
0.06 0.64
0.07 0.64
0.08 0.55
0.09 0.58
0.1 0.65
0.11 0.63
0.12 0.69
0.13 0.7
0.14 0.66
0.15 0.65
0.16 0.67
0.17 0.83
0.18 0.69
0.19 0.76
0.2 0.79
0.21 0.75
0.22 0.79
0.23 0.81
0.24 0.8
0.25 0.8
0.26 0.81
0.27 0.79
0.28 0.85
0.29 0.84
0.3 0.84
0.31 0.88
0.32 0.95
0.33 0.84
0.34 0.87
0.35 0.89
0.36 0.88
0.37 0.93
0.38 0.91
0.39 0.89
0.4 0.92
0.41 0.92
0.42 0.96
0.43 0.92
0.44 0.87
0.45 0.96
0.46 0.91
0.47 0.96
0.48 0.92
0.49 0.95
0.5 0.93
0.51 0.96
0.52 0.96
0.53 0.94
0.54 0.97
0.55 1
0.56 0.99
0.57 0.95
0.58 0.96
0.59 0.98
0.6 0.97
0.61 0.98
0.62 0.96
0.63 0.95
0.64 0.97
0.65 0.98
0.66 0.98
0.67 0.99
0.68 0.98
0.69 0.99
0.7 0.98
0.71 0.98
0.72 0.97
0.73 0.97
0.74 0.97
0.75 0.99
0.76 1
0.77 0.98
0.78 0.97
0.79 0.97
0.8 0.97
0.81 0.98
0.82 0.98
0.83 0.97
0.84 0.98
0.85 0.97
0.86 0.98
0.87 0.98
0.88 0.96
0.89 0.94
0.9 0.97
0.91 0.98
0.92 1
0.93 0.99
0.94 0.96
0.95 1
0.96 0.94
0.97 0.96
0.98 0.96
0.99 0.98
1 0.95
1.02 0.86
1.03 0.84
1.04 0.72
1.05 0.62
1.06 0.67
1.07 0.55
1.08 0.58
1.09 0.59
1.1 0.48
1.11 0.48
1.12 0.51
1.13 0.57
1.14 0.46
1.15 0.58
1.16 0.47
1.17 0.5
1.18 0.51
1.19 0.43
1.2 0.48
1.21 0.5
1.22 0.36
1.23 0.48
1.24 0.4
1.25 0.57
1.26 0.44
1.27 0.5
1.28 0.47
1.29 0.59
1.3 0.52
1.31 0.51
1.32 0.47
1.33 0.49
1.34 0.49
1.35 0.47
1.36 0.49
1.37 0.51
1.38 0.48
1.39 0.43
1.4 0.47
1.41 0.45
1.42 0.46
1.43 0.46
1.44 0.52
1.45 0.56
1.46 0.55
1.47 0.55
1.48 0.5
1.49 0.44
1.5 0.48
1.51 0.43
1.52 0.51
1.53 0.46
1.54 0.46
1.55 0.5
1.56 0.55
1.57 0.51
1.58 0.49
1.59 0.5
1.6 0.49
1.61 0.42
1.62 0.53
1.63 0.46
1.64 0.54
1.65 0.66
1.66 0.52
1.67 0.39
1.68 0.52
1.69 0.52
1.7 0.46
1.71 0.47
1.72 0.53
1.73 0.48
1.74 0.51
1.75 0.53
1.76 0.53
1.77 0.55
1.78 0.57
1.79 0.43
1.8 0.49
1.81 0.47
1.82 0.5
1.83 0.46
1.84 0.42
1.85 0.52
1.86 0.39
1.87 0.47
1.88 0.45
1.89 0.48
1.9 0.44
1.91 0.42
1.92 0.47
1.93 0.49
1.94 0.57
1.95 0.48
1.96 0.4
1.97 0.6
1.98 0.53
1.99 0.58
2 0.47
};
\addplot [semithick, color2, opacity=0.75]
table {%
0.01 0.51
0.02 0.51
0.03 0.41
0.04 0.5
0.05 0.55
0.06 0.52
0.07 0.56
0.08 0.55
0.09 0.49
0.1 0.57
0.11 0.6
0.12 0.66
0.13 0.66
0.14 0.64
0.15 0.61
0.16 0.64
0.17 0.82
0.18 0.66
0.19 0.74
0.2 0.75
0.21 0.77
0.22 0.79
0.23 0.79
0.24 0.79
0.25 0.78
0.26 0.8
0.27 0.76
0.28 0.82
0.29 0.83
0.3 0.84
0.31 0.87
0.32 0.94
0.33 0.83
0.34 0.87
0.35 0.88
0.36 0.86
0.37 0.9
0.38 0.9
0.39 0.89
0.4 0.92
0.41 0.91
0.42 0.96
0.43 0.92
0.44 0.86
0.45 0.94
0.46 0.91
0.47 0.94
0.48 0.91
0.49 0.95
0.5 0.92
0.51 0.96
0.52 0.96
0.53 0.94
0.54 0.98
0.55 0.99
0.56 0.99
0.57 0.95
0.58 0.96
0.59 0.98
0.6 0.97
0.61 0.98
0.62 0.95
0.63 0.94
0.64 0.97
0.65 0.98
0.66 0.98
0.67 0.98
0.68 0.97
0.69 0.98
0.7 0.98
0.71 0.98
0.72 0.97
0.73 0.97
0.74 0.96
0.75 0.99
0.76 1
0.77 0.98
0.78 0.97
0.79 0.97
0.8 0.97
0.81 0.98
0.82 0.98
0.83 0.97
0.84 0.98
0.85 0.97
0.86 0.98
0.87 0.97
0.88 0.96
0.89 0.94
0.9 0.97
0.91 0.98
0.92 1
0.93 1
0.94 0.96
0.95 1
0.96 0.94
0.97 0.96
0.98 0.96
0.99 0.98
1 0.95
1.02 0.86
1.03 0.84
1.04 0.73
1.05 0.62
1.06 0.68
1.07 0.54
1.08 0.59
1.09 0.6
1.1 0.47
1.11 0.48
1.12 0.51
1.13 0.55
1.14 0.44
1.15 0.57
1.16 0.46
1.17 0.5
1.18 0.51
1.19 0.45
1.2 0.46
1.21 0.51
1.22 0.34
1.23 0.49
1.24 0.4
1.25 0.57
1.26 0.45
1.27 0.5
1.28 0.49
1.29 0.57
1.3 0.53
1.31 0.54
1.32 0.48
1.33 0.48
1.34 0.49
1.35 0.46
1.36 0.51
1.37 0.5
1.38 0.49
1.39 0.43
1.4 0.47
1.41 0.45
1.42 0.47
1.43 0.45
1.44 0.51
1.45 0.58
1.46 0.53
1.47 0.54
1.48 0.5
1.49 0.43
1.5 0.5
1.51 0.44
1.52 0.51
1.53 0.45
1.54 0.46
1.55 0.51
1.56 0.55
1.57 0.48
1.58 0.47
1.59 0.5
1.6 0.5
1.61 0.42
1.62 0.55
1.63 0.46
1.64 0.54
1.65 0.65
1.66 0.51
1.67 0.39
1.68 0.53
1.69 0.5
1.7 0.47
1.71 0.47
1.72 0.53
1.73 0.49
1.74 0.51
1.75 0.55
1.76 0.53
1.77 0.58
1.78 0.56
1.79 0.43
1.8 0.48
1.81 0.46
1.82 0.5
1.83 0.45
1.84 0.43
1.85 0.53
1.86 0.41
1.87 0.46
1.88 0.45
1.89 0.51
1.9 0.43
1.91 0.42
1.92 0.46
1.93 0.49
1.94 0.57
1.95 0.5
1.96 0.42
1.97 0.55
1.98 0.51
1.99 0.58
2 0.46
};
\addplot [semithick, color3, opacity=0.75]
table {%
0.01 0.56
0.02 0.55
0.03 0.7
0.04 0.66
0.05 0.67
0.06 0.67
0.07 0.82
0.08 0.76
0.09 0.74
0.1 0.81
0.11 0.8
0.12 0.86
0.13 0.82
0.14 0.86
0.15 0.85
0.16 0.86
0.17 0.97
0.18 0.86
0.19 0.89
0.2 0.97
0.21 0.91
0.22 0.94
0.23 0.95
0.24 0.94
0.25 0.95
0.26 0.97
0.27 0.97
0.28 0.96
0.29 0.97
0.3 0.96
0.31 0.99
0.32 0.98
0.33 1
0.34 0.97
0.35 0.99
0.36 0.97
0.37 0.98
0.38 1
0.39 0.98
0.4 1
0.41 0.98
0.42 1
0.43 0.98
0.44 0.99
0.45 0.98
0.46 1
0.47 0.99
0.48 0.99
0.49 0.98
0.5 0.99
0.51 0.99
0.52 0.98
0.53 0.98
0.54 0.98
0.55 1
0.56 1
0.57 0.99
0.58 1
0.59 0.99
0.6 0.98
0.61 1
0.62 0.98
0.63 0.99
0.64 0.98
0.65 1
0.66 0.98
0.67 1
0.68 0.98
0.69 1
0.7 0.99
0.71 0.99
0.72 1
0.73 1
0.74 0.98
0.75 1
0.76 1
0.77 1
0.78 1
0.79 0.98
0.8 1
0.81 0.99
0.82 0.99
0.83 0.98
0.84 0.98
0.85 1
0.86 0.98
0.87 0.99
0.88 0.98
0.89 0.95
0.9 0.98
0.91 0.98
0.92 1
0.93 1
0.94 0.99
0.95 1
0.96 0.99
0.97 0.98
0.98 0.99
0.99 0.98
1 0.96
1.02 0.88
1.03 0.89
1.04 0.7
1.05 0.72
1.06 0.62
1.07 0.68
1.08 0.66
1.09 0.67
1.1 0.56
1.11 0.56
1.12 0.59
1.13 0.59
1.14 0.47
1.15 0.5
1.16 0.46
1.17 0.53
1.18 0.57
1.19 0.49
1.2 0.49
1.21 0.53
1.22 0.44
1.23 0.49
1.24 0.35
1.25 0.43
1.26 0.48
1.27 0.47
1.28 0.39
1.29 0.49
1.3 0.42
1.31 0.4
1.32 0.38
1.33 0.45
1.34 0.5
1.35 0.37
1.36 0.39
1.37 0.46
1.38 0.48
1.39 0.39
1.4 0.41
1.41 0.42
1.42 0.48
1.43 0.36
1.44 0.42
1.45 0.41
1.46 0.34
1.47 0.4
1.48 0.35
1.49 0.42
1.5 0.39
1.51 0.41
1.52 0.41
1.53 0.31
1.54 0.41
1.55 0.45
1.56 0.35
1.57 0.46
1.58 0.42
1.59 0.33
1.6 0.44
1.61 0.39
1.62 0.42
1.63 0.44
1.64 0.5
1.65 0.41
1.66 0.43
1.67 0.39
1.68 0.37
1.69 0.32
1.7 0.45
1.71 0.41
1.72 0.34
1.73 0.34
1.74 0.34
1.75 0.36
1.76 0.4
1.77 0.45
1.78 0.42
1.79 0.25
1.8 0.43
1.81 0.43
1.82 0.33
1.83 0.43
1.84 0.46
1.85 0.37
1.86 0.45
1.87 0.45
1.88 0.41
1.89 0.35
1.9 0.32
1.91 0.29
1.92 0.35
1.93 0.35
1.94 0.42
1.95 0.42
1.96 0.32
1.97 0.42
1.98 0.37
1.99 0.48
2 0.41
};
\addplot [semithick, color4, opacity=0.75]
table {%
0.01 0.56
0.02 0.56
0.03 0.7
0.04 0.68
0.05 0.66
0.06 0.67
0.07 0.8
0.08 0.76
0.09 0.74
0.1 0.8
0.11 0.8
0.12 0.87
0.13 0.82
0.14 0.86
0.15 0.85
0.16 0.85
0.17 0.97
0.18 0.86
0.19 0.89
0.2 0.97
0.21 0.91
0.22 0.94
0.23 0.95
0.24 0.95
0.25 0.97
0.26 0.97
0.27 0.97
0.28 0.97
0.29 0.98
0.3 0.96
0.31 0.99
0.32 0.98
0.33 1
0.34 0.97
0.35 0.99
0.36 0.97
0.37 0.98
0.38 1
0.39 0.98
0.4 1
0.41 0.99
0.42 1
0.43 0.98
0.44 0.99
0.45 0.99
0.46 1
0.47 1
0.48 1
0.49 0.99
0.5 0.99
0.51 0.99
0.52 0.98
0.53 0.99
0.54 0.98
0.55 1
0.56 1
0.57 0.99
0.58 1
0.59 0.99
0.6 0.98
0.61 1
0.62 1
0.63 0.99
0.64 0.98
0.65 1
0.66 0.99
0.67 1
0.68 0.98
0.69 1
0.7 0.99
0.71 0.99
0.72 1
0.73 1
0.74 0.99
0.75 1
0.76 1
0.77 1
0.78 0.99
0.79 0.98
0.8 1
0.81 0.99
0.82 0.99
0.83 0.98
0.84 0.99
0.85 1
0.86 0.98
0.87 0.98
0.88 0.98
0.89 0.97
0.9 0.98
0.91 0.98
0.92 1
0.93 1
0.94 0.99
0.95 1
0.96 1
0.97 0.98
0.98 0.99
0.99 0.99
1 0.96
1.02 0.88
1.03 0.88
1.04 0.71
1.05 0.72
1.06 0.64
1.07 0.69
1.08 0.66
1.09 0.67
1.1 0.55
1.11 0.56
1.12 0.59
1.13 0.59
1.14 0.47
1.15 0.5
1.16 0.47
1.17 0.54
1.18 0.55
1.19 0.51
1.2 0.47
1.21 0.54
1.22 0.45
1.23 0.47
1.24 0.36
1.25 0.43
1.26 0.48
1.27 0.47
1.28 0.38
1.29 0.49
1.3 0.43
1.31 0.4
1.32 0.38
1.33 0.46
1.34 0.5
1.35 0.37
1.36 0.38
1.37 0.46
1.38 0.48
1.39 0.41
1.4 0.41
1.41 0.42
1.42 0.49
1.43 0.36
1.44 0.42
1.45 0.41
1.46 0.36
1.47 0.4
1.48 0.35
1.49 0.42
1.5 0.39
1.51 0.41
1.52 0.42
1.53 0.31
1.54 0.41
1.55 0.45
1.56 0.35
1.57 0.46
1.58 0.42
1.59 0.32
1.6 0.44
1.61 0.38
1.62 0.43
1.63 0.45
1.64 0.49
1.65 0.41
1.66 0.43
1.67 0.39
1.68 0.37
1.69 0.32
1.7 0.45
1.71 0.4
1.72 0.36
1.73 0.33
1.74 0.35
1.75 0.36
1.76 0.42
1.77 0.45
1.78 0.43
1.79 0.26
1.8 0.43
1.81 0.43
1.82 0.35
1.83 0.44
1.84 0.45
1.85 0.37
1.86 0.45
1.87 0.44
1.88 0.41
1.89 0.35
1.9 0.32
1.91 0.28
1.92 0.36
1.93 0.36
1.94 0.43
1.95 0.42
1.96 0.32
1.97 0.42
1.98 0.37
1.99 0.48
2 0.41
};
\addplot [semithick, color5, opacity=0.75]
table {%
0.01 0.47
0.02 0.55
0.03 0.65
0.04 0.61
0.05 0.64
0.06 0.67
0.07 0.77
0.08 0.72
0.09 0.71
0.1 0.82
0.11 0.82
0.12 0.83
0.13 0.79
0.14 0.8
0.15 0.82
0.16 0.82
0.17 0.95
0.18 0.84
0.19 0.85
0.2 0.93
0.21 0.88
0.22 0.92
0.23 0.9
0.24 0.91
0.25 0.93
0.26 0.97
0.27 0.96
0.28 0.97
0.29 0.94
0.3 0.93
0.31 0.98
0.32 0.98
0.33 0.98
0.34 0.94
0.35 0.98
0.36 0.97
0.37 0.97
0.38 0.98
0.39 0.97
0.4 1
0.41 0.97
0.42 0.98
0.43 0.97
0.44 0.95
0.45 0.99
0.46 0.99
0.47 1
0.48 0.98
0.49 0.99
0.5 0.97
0.51 0.99
0.52 0.97
0.53 0.96
0.54 0.98
0.55 1
0.56 1
0.57 0.99
0.58 0.99
0.59 1
0.6 0.99
0.61 1
0.62 0.97
0.63 0.99
0.64 0.98
0.65 0.98
0.66 0.96
0.67 0.99
0.68 0.97
0.69 0.99
0.7 0.99
0.71 0.98
0.72 0.98
0.73 1
0.74 0.97
0.75 0.98
0.76 1
0.77 0.99
0.78 0.98
0.79 0.98
0.8 0.98
0.81 0.98
0.82 0.99
0.83 0.96
0.84 0.98
0.85 1
0.86 0.96
0.87 0.97
0.88 0.96
0.89 0.95
0.9 0.98
0.91 0.97
0.92 0.99
0.93 1
0.94 0.96
0.95 1
0.96 0.95
0.97 0.96
0.98 0.97
0.99 0.98
1 0.93
1.02 0.85
1.03 0.82
1.04 0.64
1.05 0.7
1.06 0.66
1.07 0.61
1.08 0.64
1.09 0.66
1.1 0.51
1.11 0.56
1.12 0.46
1.13 0.58
1.14 0.47
1.15 0.5
1.16 0.42
1.17 0.5
1.18 0.49
1.19 0.46
1.2 0.41
1.21 0.42
1.22 0.44
1.23 0.45
1.24 0.34
1.25 0.45
1.26 0.48
1.27 0.4
1.28 0.33
1.29 0.5
1.3 0.36
1.31 0.41
1.32 0.34
1.33 0.42
1.34 0.5
1.35 0.34
1.36 0.39
1.37 0.38
1.38 0.52
1.39 0.41
1.4 0.4
1.41 0.36
1.42 0.43
1.43 0.34
1.44 0.39
1.45 0.4
1.46 0.33
1.47 0.42
1.48 0.3
1.49 0.4
1.5 0.38
1.51 0.3
1.52 0.41
1.53 0.26
1.54 0.34
1.55 0.42
1.56 0.32
1.57 0.43
1.58 0.39
1.59 0.33
1.6 0.43
1.61 0.36
1.62 0.4
1.63 0.44
1.64 0.44
1.65 0.4
1.66 0.39
1.67 0.35
1.68 0.43
1.69 0.32
1.7 0.4
1.71 0.35
1.72 0.36
1.73 0.31
1.74 0.34
1.75 0.37
1.76 0.39
1.77 0.45
1.78 0.36
1.79 0.28
1.8 0.43
1.81 0.43
1.82 0.41
1.83 0.42
1.84 0.49
1.85 0.34
1.86 0.46
1.87 0.38
1.88 0.37
1.89 0.34
1.9 0.39
1.91 0.25
1.92 0.36
1.93 0.33
1.94 0.4
1.95 0.43
1.96 0.33
1.97 0.39
1.98 0.43
1.99 0.43
2 0.44
};
\addplot [semithick, color6, opacity=0.75, dashed]
table {%
0.01 0.63
0.02 0.6
0.03 0.49
0.04 0.56
0.05 0.7
0.06 0.63
0.07 0.71
0.08 0.66
0.09 0.66
0.1 0.69
0.11 0.83
0.12 0.76
0.13 0.77
0.14 0.69
0.15 0.77
0.16 0.88
0.17 0.92
0.18 0.85
0.19 0.81
0.2 0.8
0.21 0.92
0.22 0.87
0.23 0.89
0.24 0.92
0.25 0.93
0.26 0.94
0.27 0.91
0.28 0.93
0.29 0.94
0.3 0.97
0.31 0.93
0.32 0.93
0.33 0.96
0.34 0.96
0.35 0.98
0.36 0.97
0.37 0.98
0.38 0.94
0.39 0.95
0.4 0.99
0.41 0.96
0.42 0.95
0.43 0.97
0.44 0.97
0.45 0.95
0.46 0.96
0.47 0.95
0.48 0.95
0.49 0.99
0.5 0.98
0.51 0.96
0.52 0.98
0.53 0.96
0.54 0.99
0.55 0.96
0.56 0.99
0.57 0.96
0.58 0.97
0.59 0.98
0.6 0.99
0.61 0.95
0.62 0.96
0.63 0.99
0.64 0.96
0.65 0.98
0.66 0.96
0.67 0.97
0.68 0.98
0.69 0.96
0.7 0.96
0.71 1
0.72 0.97
0.73 0.92
0.74 0.98
0.75 0.96
0.76 0.97
0.77 0.97
0.78 0.98
0.79 0.97
0.8 0.95
0.81 0.92
0.82 0.96
0.83 0.97
0.84 0.97
0.85 0.92
0.86 0.97
0.87 0.94
0.88 0.97
0.89 0.97
0.9 0.96
0.91 0.97
0.92 0.97
0.93 0.99
0.94 0.95
0.95 0.94
0.96 0.96
0.97 0.95
0.98 0.98
0.99 0.95
1 0.96
1.02 0.85
1.03 0.75
1.04 0.73
1.05 0.72
1.06 0.62
1.07 0.57
1.08 0.66
1.09 0.58
1.1 0.62
1.11 0.58
1.12 0.55
1.13 0.57
1.14 0.57
1.15 0.58
1.16 0.57
1.17 0.58
1.18 0.54
1.19 0.47
1.2 0.53
1.21 0.55
1.22 0.57
1.23 0.45
1.24 0.5
1.25 0.59
1.26 0.5
1.27 0.58
1.28 0.56
1.29 0.5
1.3 0.57
1.31 0.56
1.32 0.5
1.33 0.59
1.34 0.52
1.35 0.58
1.36 0.52
1.37 0.59
1.38 0.47
1.39 0.39
1.4 0.54
1.41 0.51
1.42 0.52
1.43 0.57
1.44 0.62
1.45 0.58
1.46 0.51
1.47 0.52
1.48 0.58
1.49 0.53
1.5 0.59
1.51 0.45
1.52 0.5
1.53 0.57
1.54 0.51
1.55 0.53
1.56 0.48
1.57 0.49
1.58 0.55
1.59 0.54
1.6 0.55
1.61 0.61
1.62 0.56
1.63 0.54
1.64 0.52
1.65 0.58
1.66 0.56
1.67 0.5
1.68 0.61
1.69 0.53
1.7 0.54
1.71 0.62
1.72 0.5
1.73 0.58
1.74 0.57
1.75 0.57
1.76 0.52
1.77 0.61
1.78 0.51
1.79 0.48
1.8 0.51
1.81 0.54
1.82 0.54
1.83 0.52
1.84 0.52
1.85 0.66
1.86 0.56
1.87 0.45
1.88 0.56
1.89 0.56
1.9 0.5
1.91 0.57
1.92 0.51
1.93 0.52
1.94 0.49
1.95 0.64
1.96 0.44
1.97 0.6
1.98 0.47
1.99 0.6
2 0.56
};
\addplot [semithick, color7, opacity=0.75, dashed]
table {%
0.01 0.63
0.02 0.6
0.03 0.49
0.04 0.56
0.05 0.7
0.06 0.63
0.07 0.71
0.08 0.66
0.09 0.66
0.1 0.69
0.11 0.83
0.12 0.76
0.13 0.77
0.14 0.69
0.15 0.77
0.16 0.88
0.17 0.92
0.18 0.85
0.19 0.81
0.2 0.8
0.21 0.92
0.22 0.87
0.23 0.89
0.24 0.92
0.25 0.93
0.26 0.94
0.27 0.91
0.28 0.93
0.29 0.94
0.3 0.97
0.31 0.93
0.32 0.93
0.33 0.96
0.34 0.96
0.35 0.98
0.36 0.97
0.37 0.98
0.38 0.94
0.39 0.95
0.4 0.99
0.41 0.96
0.42 0.95
0.43 0.97
0.44 0.97
0.45 0.95
0.46 0.96
0.47 0.95
0.48 0.95
0.49 0.99
0.5 0.98
0.51 0.96
0.52 0.98
0.53 0.96
0.54 0.99
0.55 0.96
0.56 0.99
0.57 0.96
0.58 0.97
0.59 0.98
0.6 0.99
0.61 0.95
0.62 0.96
0.63 0.99
0.64 0.96
0.65 0.98
0.66 0.96
0.67 0.97
0.68 0.98
0.69 0.96
0.7 0.96
0.71 1
0.72 0.97
0.73 0.92
0.74 0.98
0.75 0.96
0.76 0.97
0.77 0.97
0.78 0.98
0.79 0.97
0.8 0.95
0.81 0.92
0.82 0.96
0.83 0.97
0.84 0.97
0.85 0.92
0.86 0.97
0.87 0.94
0.88 0.97
0.89 0.97
0.9 0.96
0.91 0.97
0.92 0.97
0.93 0.99
0.94 0.95
0.95 0.94
0.96 0.96
0.97 0.95
0.98 0.98
0.99 0.95
1 0.96
1.02 0.85
1.03 0.75
1.04 0.73
1.05 0.72
1.06 0.62
1.07 0.57
1.08 0.66
1.09 0.58
1.1 0.62
1.11 0.58
1.12 0.55
1.13 0.57
1.14 0.57
1.15 0.58
1.16 0.57
1.17 0.58
1.18 0.54
1.19 0.47
1.2 0.53
1.21 0.55
1.22 0.57
1.23 0.45
1.24 0.5
1.25 0.59
1.26 0.5
1.27 0.58
1.28 0.56
1.29 0.5
1.3 0.57
1.31 0.56
1.32 0.5
1.33 0.59
1.34 0.52
1.35 0.58
1.36 0.52
1.37 0.59
1.38 0.47
1.39 0.39
1.4 0.54
1.41 0.51
1.42 0.52
1.43 0.57
1.44 0.62
1.45 0.58
1.46 0.51
1.47 0.52
1.48 0.58
1.49 0.53
1.5 0.59
1.51 0.45
1.52 0.5
1.53 0.57
1.54 0.51
1.55 0.53
1.56 0.48
1.57 0.49
1.58 0.55
1.59 0.54
1.6 0.55
1.61 0.61
1.62 0.56
1.63 0.54
1.64 0.52
1.65 0.58
1.66 0.56
1.67 0.5
1.68 0.61
1.69 0.53
1.7 0.54
1.71 0.62
1.72 0.5
1.73 0.58
1.74 0.57
1.75 0.57
1.76 0.52
1.77 0.61
1.78 0.51
1.79 0.48
1.8 0.51
1.81 0.54
1.82 0.54
1.83 0.52
1.84 0.52
1.85 0.66
1.86 0.56
1.87 0.45
1.88 0.56
1.89 0.56
1.9 0.5
1.91 0.57
1.92 0.51
1.93 0.52
1.94 0.49
1.95 0.64
1.96 0.44
1.97 0.6
1.98 0.47
1.99 0.6
2 0.56
};
\addplot [semithick, color8, opacity=0.75, dashed]
table {%
0.01 0.55
0.02 0.6
0.03 0.56
0.04 0.58
0.05 0.64
0.06 0.7
0.07 0.67
0.08 0.74
0.09 0.76
0.1 0.72
0.11 0.8
0.12 0.81
0.13 0.78
0.14 0.8
0.15 0.89
0.16 0.93
0.17 0.9
0.18 0.88
0.19 0.92
0.2 0.93
0.21 0.95
0.22 0.94
0.23 0.93
0.24 0.96
0.25 0.95
0.26 0.96
0.27 0.95
0.28 0.96
0.29 0.98
0.3 0.99
0.31 0.95
0.32 1
0.33 0.95
0.34 0.98
0.35 0.99
0.36 1
0.37 0.99
0.38 0.96
0.39 0.98
0.4 0.99
0.41 0.99
0.42 0.98
0.43 0.99
0.44 1
0.45 0.98
0.46 0.97
0.47 0.99
0.48 0.99
0.49 1
0.5 0.99
0.51 0.97
0.52 0.98
0.53 1
0.54 0.97
0.55 1
0.56 0.97
0.57 0.99
0.58 0.99
0.59 0.97
0.6 1
0.61 0.98
0.62 0.99
0.63 0.98
0.64 1
0.65 0.99
0.66 0.97
0.67 0.99
0.68 0.97
0.69 0.97
0.7 0.99
0.71 1
0.72 0.97
0.73 0.98
0.74 1
0.75 0.96
0.76 1
0.77 0.99
0.78 1
0.79 1
0.8 0.98
0.81 0.96
0.82 1
0.83 1
0.84 1
0.85 0.99
0.86 0.99
0.87 0.99
0.88 0.98
0.89 0.99
0.9 0.99
0.91 0.96
0.92 0.97
0.93 1
0.94 0.97
0.95 0.98
0.96 0.97
0.97 0.98
0.98 0.97
0.99 0.98
1 0.95
1.02 0.92
1.03 0.77
1.04 0.72
1.05 0.71
1.06 0.66
1.07 0.54
1.08 0.57
1.09 0.6
1.1 0.61
1.11 0.51
1.12 0.57
1.13 0.49
1.14 0.59
1.15 0.6
1.16 0.51
1.17 0.53
1.18 0.52
1.19 0.62
1.2 0.48
1.21 0.53
1.22 0.46
1.23 0.53
1.24 0.58
1.25 0.57
1.26 0.52
1.27 0.54
1.28 0.61
1.29 0.57
1.3 0.54
1.31 0.57
1.32 0.5
1.33 0.53
1.34 0.5
1.35 0.57
1.36 0.52
1.37 0.53
1.38 0.54
1.39 0.47
1.4 0.56
1.41 0.47
1.42 0.51
1.43 0.44
1.44 0.46
1.45 0.57
1.46 0.55
1.47 0.46
1.48 0.49
1.49 0.47
1.5 0.55
1.51 0.42
1.52 0.53
1.53 0.57
1.54 0.52
1.55 0.52
1.56 0.59
1.57 0.52
1.58 0.58
1.59 0.58
1.6 0.58
1.61 0.58
1.62 0.47
1.63 0.52
1.64 0.58
1.65 0.52
1.66 0.53
1.67 0.48
1.68 0.61
1.69 0.46
1.7 0.5
1.71 0.52
1.72 0.52
1.73 0.54
1.74 0.52
1.75 0.61
1.76 0.63
1.77 0.62
1.78 0.51
1.79 0.48
1.8 0.52
1.81 0.61
1.82 0.54
1.83 0.46
1.84 0.47
1.85 0.6
1.86 0.52
1.87 0.54
1.88 0.6
1.89 0.47
1.9 0.54
1.91 0.6
1.92 0.55
1.93 0.46
1.94 0.63
1.95 0.61
1.96 0.47
1.97 0.63
1.98 0.53
1.99 0.55
2 0.56
};
\addplot [semithick, color9, opacity=0.75, dashed]
table {%
0.01 0.52
0.02 0.59
0.03 0.61
0.04 0.73
0.05 0.67
0.06 0.77
0.07 0.8
0.08 0.85
0.09 0.81
0.1 0.9
0.11 0.86
0.12 0.89
0.13 0.92
0.14 0.93
0.15 0.96
0.16 0.94
0.17 0.95
0.18 0.98
0.19 0.97
0.2 0.99
0.21 1
0.22 1
0.23 0.99
0.24 0.99
0.25 1
0.26 1
0.27 1
0.28 1
0.29 1
0.3 1
0.31 1
0.32 1
0.33 1
0.34 1
0.35 1
0.36 1
0.37 1
0.38 1
0.39 1
0.4 1
0.41 1
0.42 1
0.43 1
0.44 1
0.45 1
0.46 1
0.47 1
0.48 1
0.49 1
0.5 1
0.51 1
0.52 1
0.53 1
0.54 1
0.55 1
0.56 1
0.57 1
0.58 1
0.59 1
0.6 1
0.61 1
0.62 1
0.63 1
0.64 1
0.65 1
0.66 1
0.67 1
0.68 1
0.69 1
0.7 1
0.71 1
0.72 1
0.73 1
0.74 1
0.75 1
0.76 1
0.77 1
0.78 1
0.79 1
0.8 1
0.81 1
0.82 1
0.83 1
0.84 1
0.85 1
0.86 1
0.87 0.99
0.88 1
0.89 1
0.9 1
0.91 1
0.92 1
0.93 1
0.94 1
0.95 1
0.96 1
0.97 1
0.98 1
0.99 1
1 1
1.02 0.97
1.03 0.88
1.04 0.9
1.05 0.81
1.06 0.75
1.07 0.69
1.08 0.66
1.09 0.72
1.1 0.66
1.11 0.54
1.12 0.62
1.13 0.72
1.14 0.46
1.15 0.6
1.16 0.41
1.17 0.53
1.18 0.54
1.19 0.61
1.2 0.52
1.21 0.52
1.22 0.54
1.23 0.45
1.24 0.42
1.25 0.48
1.26 0.52
1.27 0.48
1.28 0.42
1.29 0.45
1.3 0.51
1.31 0.52
1.32 0.44
1.33 0.53
1.34 0.54
1.35 0.43
1.36 0.38
1.37 0.57
1.38 0.52
1.39 0.53
1.4 0.48
1.41 0.52
1.42 0.6
1.43 0.36
1.44 0.41
1.45 0.52
1.46 0.48
1.47 0.55
1.48 0.39
1.49 0.42
1.5 0.42
1.51 0.37
1.52 0.44
1.53 0.45
1.54 0.41
1.55 0.46
1.56 0.47
1.57 0.46
1.58 0.52
1.59 0.42
1.6 0.6
1.61 0.43
1.62 0.51
1.63 0.43
1.64 0.44
1.65 0.43
1.66 0.4
1.67 0.51
1.68 0.43
1.69 0.46
1.7 0.44
1.71 0.53
1.72 0.38
1.73 0.47
1.74 0.47
1.75 0.45
1.76 0.46
1.77 0.52
1.78 0.42
1.79 0.32
1.8 0.48
1.81 0.44
1.82 0.5
1.83 0.42
1.84 0.47
1.85 0.4
1.86 0.5
1.87 0.47
1.88 0.57
1.89 0.44
1.9 0.45
1.91 0.51
1.92 0.31
1.93 0.49
1.94 0.52
1.95 0.49
1.96 0.5
1.97 0.5
1.98 0.5
1.99 0.49
2 0.5
};
\addplot [semithick, color10, opacity=0.75, dashed]
table {%
0.01 0.61
0.02 0.59
0.03 0.65
0.04 0.76
0.05 0.74
0.06 0.82
0.07 0.87
0.08 0.85
0.09 0.85
0.1 0.91
0.11 0.92
0.12 0.95
0.13 0.96
0.14 0.94
0.15 0.95
0.16 0.97
0.17 0.98
0.18 0.99
0.19 0.99
0.2 1
0.21 0.98
0.22 1
0.23 0.99
0.24 0.99
0.25 1
0.26 1
0.27 1
0.28 1
0.29 1
0.3 1
0.31 1
0.32 1
0.33 1
0.34 1
0.35 1
0.36 1
0.37 1
0.38 1
0.39 1
0.4 1
0.41 1
0.42 1
0.43 1
0.44 1
0.45 1
0.46 1
0.47 1
0.48 1
0.49 1
0.5 1
0.51 1
0.52 1
0.53 1
0.54 1
0.55 1
0.56 1
0.57 1
0.58 1
0.59 1
0.6 1
0.61 1
0.62 1
0.63 1
0.64 1
0.65 1
0.66 1
0.67 1
0.68 1
0.69 1
0.7 1
0.71 1
0.72 1
0.73 1
0.74 1
0.75 1
0.76 1
0.77 1
0.78 1
0.79 1
0.8 1
0.81 1
0.82 1
0.83 1
0.84 1
0.85 1
0.86 1
0.87 0.99
0.88 1
0.89 1
0.9 1
0.91 1
0.92 1
0.93 1
0.94 1
0.95 1
0.96 1
0.97 1
0.98 1
0.99 1
1 1
1.02 0.98
1.03 0.93
1.04 0.86
1.05 0.82
1.06 0.71
1.07 0.74
1.08 0.67
1.09 0.77
1.1 0.59
1.11 0.67
1.12 0.53
1.13 0.65
1.14 0.53
1.15 0.63
1.16 0.51
1.17 0.55
1.18 0.58
1.19 0.57
1.2 0.56
1.21 0.56
1.22 0.55
1.23 0.53
1.24 0.5
1.25 0.45
1.26 0.52
1.27 0.53
1.28 0.5
1.29 0.49
1.3 0.54
1.31 0.56
1.32 0.49
1.33 0.54
1.34 0.56
1.35 0.46
1.36 0.42
1.37 0.57
1.38 0.53
1.39 0.47
1.4 0.46
1.41 0.58
1.42 0.6
1.43 0.47
1.44 0.56
1.45 0.47
1.46 0.59
1.47 0.55
1.48 0.46
1.49 0.44
1.5 0.47
1.51 0.4
1.52 0.48
1.53 0.47
1.54 0.45
1.55 0.47
1.56 0.52
1.57 0.48
1.58 0.53
1.59 0.44
1.6 0.6
1.61 0.45
1.62 0.59
1.63 0.47
1.64 0.51
1.65 0.51
1.66 0.41
1.67 0.42
1.68 0.46
1.69 0.5
1.7 0.52
1.71 0.5
1.72 0.4
1.73 0.47
1.74 0.42
1.75 0.44
1.76 0.48
1.77 0.53
1.78 0.46
1.79 0.39
1.8 0.56
1.81 0.41
1.82 0.48
1.83 0.5
1.84 0.45
1.85 0.43
1.86 0.51
1.87 0.57
1.88 0.53
1.89 0.48
1.9 0.51
1.91 0.48
1.92 0.4
1.93 0.5
1.94 0.56
1.95 0.57
1.96 0.43
1.97 0.46
1.98 0.54
1.99 0.5
2 0.51
};
\addplot [semithick, color11, opacity=0.75, dashed]
table {%
0.01 0.74
0.02 0.74
0.03 0.89
0.04 0.93
0.05 0.97
0.06 0.97
0.07 0.98
0.08 1
0.09 0.97
0.1 0.99
0.11 1
0.12 1
0.13 1
0.14 1
0.15 1
0.16 1
0.17 1
0.18 1
0.19 1
0.2 1
0.21 1
0.22 1
0.23 1
0.24 1
0.25 1
0.26 1
0.27 1
0.28 1
0.29 1
0.3 1
0.31 1
0.32 1
0.33 1
0.34 1
0.35 1
0.36 1
0.37 1
0.38 1
0.39 1
0.4 1
0.41 1
0.42 1
0.43 1
0.44 1
0.45 1
0.46 1
0.47 1
0.48 1
0.49 1
0.5 1
0.51 1
0.52 1
0.53 1
0.54 1
0.55 1
0.56 1
0.57 1
0.58 1
0.59 1
0.6 1
0.61 1
0.62 1
0.63 1
0.64 1
0.65 1
0.66 1
0.67 1
0.68 1
0.69 1
0.7 1
0.71 1
0.72 1
0.73 1
0.74 1
0.75 1
0.76 1
0.77 1
0.78 1
0.79 1
0.8 1
0.81 1
0.82 1
0.83 1
0.84 1
0.85 1
0.86 1
0.87 1
0.88 1
0.89 1
0.9 1
0.91 1
0.92 1
0.93 1
0.94 1
0.95 1
0.96 1
0.97 1
0.98 1
0.99 1
1 1
1.02 1
1.03 1
1.04 0.99
1.05 0.97
1.06 0.95
1.07 0.93
1.08 0.83
1.09 0.93
1.1 0.85
1.11 0.9
1.12 0.84
1.13 0.88
1.14 0.82
1.15 0.84
1.16 0.79
1.17 0.72
1.18 0.81
1.19 0.79
1.2 0.75
1.21 0.82
1.22 0.76
1.23 0.72
1.24 0.71
1.25 0.74
1.26 0.75
1.27 0.77
1.28 0.63
1.29 0.66
1.3 0.78
1.31 0.72
1.32 0.76
1.33 0.68
1.34 0.8
1.35 0.7
1.36 0.68
1.37 0.8
1.38 0.72
1.39 0.69
1.4 0.73
1.41 0.73
1.42 0.74
1.43 0.75
1.44 0.73
1.45 0.72
1.46 0.65
1.47 0.76
1.48 0.79
1.49 0.75
1.5 0.72
1.51 0.69
1.52 0.73
1.53 0.75
1.54 0.69
1.55 0.71
1.56 0.74
1.57 0.67
1.58 0.71
1.59 0.65
1.6 0.78
1.61 0.74
1.62 0.64
1.63 0.72
1.64 0.77
1.65 0.72
1.66 0.76
1.67 0.75
1.68 0.75
1.69 0.7
1.7 0.7
1.71 0.76
1.72 0.72
1.73 0.71
1.74 0.69
1.75 0.68
1.76 0.68
1.77 0.74
1.78 0.69
1.79 0.66
1.8 0.75
1.81 0.73
1.82 0.7
1.83 0.74
1.84 0.72
1.85 0.69
1.86 0.8
1.87 0.77
1.88 0.65
1.89 0.72
1.9 0.7
1.91 0.71
1.92 0.68
1.93 0.68
1.94 0.83
1.95 0.8
1.96 0.65
1.97 0.74
1.98 0.67
1.99 0.73
2 0.76
};
\addplot [semithick, black, opacity=1, dash pattern=on 1pt off 1pt]
table {%
-0.0895000000000001 0.5
2.0995 0.5
};
\addplot [semithick, black, opacity=1, dash pattern=on 1pt off 1pt]
table {%
-0.0895000000000001 1
2.0995 1
};
\addplot [semithick, black, opacity=1, dash pattern=on 1pt off 1pt]
table {%
1 0.2
1 1.05
};
\end{axis}

\end{tikzpicture}

%% file: plots/decoupled2/LAPxGAU.tex
% This file was created by tikzplotlib v0.9.6.
\begin{tikzpicture}

\definecolor{color0}{rgb}{0.866666666666667,0.494117647058824,0.164705882352941}
\definecolor{color1}{rgb}{0.164705882352941,0.643137254901961,0.866666666666667}
\definecolor{color2}{rgb}{0.584313725490196,0.866666666666667,0.164705882352941}
\definecolor{color3}{rgb}{0.109803921568627,0.337254901960784,0.129411764705882}
\definecolor{color4}{rgb}{0.529411764705882,0.305882352941176,0.858823529411765}
\definecolor{color5}{rgb}{0.858823529411765,0.305882352941176,0.435294117647059}
\definecolor{color6}{rgb}{0.937254901960784,0.929411764705882,0.392156862745098}
\definecolor{color7}{rgb}{0.0901960784313725,0.486274509803922,0.0980392156862745}
\definecolor{color8}{rgb}{0.156862745098039,0.188235294117647,0.827450980392157}
\definecolor{color9}{rgb}{0.937254901960784,0.392156862745098,0.894117647058824}
\definecolor{color10}{rgb}{0.2,0.184313725490196,0.184313725490196}
\definecolor{color11}{rgb}{0.0156862745098039,0.803921568627451,0.976470588235294}

\begin{axis}[
tick align=outside,
tick pos=left,
x grid style={white!69.0196078431373!black},
xmajorgrids,
xmin=-0.0895, xmax=2.0995,
xtick style={color=black},
xtick={0,0.1,0.2,0.3,0.4,0.5,0.6,0.7,0.8,0.9,1,1.1,1.2,1.3,1.4,1.5,1.6,1.7,1.8,1.9,2},
xticklabels={0,,.2,,.4,,.6,,.8,,1,,20,,40,,60,,80,,100},
height=4.8cm,
width=6.5cm,
y grid style={white!69.0196078431373!black},
ymajorgrids,
ymin=0.2, ymax=1.05,
ytick style={color=black}
]
\addplot [semithick, color0, opacity=0.75]
table {%
0.01 0.46
0.02 0.6
0.03 0.6
0.04 0.5
0.05 0.57
0.06 0.51
0.07 0.64
0.08 0.59
0.09 0.59
0.1 0.72
0.11 0.7
0.12 0.69
0.13 0.67
0.14 0.75
0.15 0.66
0.16 0.72
0.17 0.77
0.18 0.78
0.19 0.79
0.2 0.67
0.21 0.77
0.22 0.84
0.23 0.77
0.24 0.85
0.25 0.84
0.26 0.81
0.27 0.81
0.28 0.9
0.29 0.87
0.3 0.88
0.31 0.87
0.32 0.91
0.33 0.93
0.34 0.91
0.35 0.94
0.36 0.92
0.37 0.94
0.38 0.92
0.39 0.91
0.4 0.92
0.41 0.95
0.42 0.93
0.43 0.91
0.44 0.92
0.45 0.96
0.46 0.94
0.47 0.87
0.48 0.93
0.49 0.91
0.5 0.96
0.51 0.95
0.52 0.98
0.53 0.96
0.54 0.97
0.55 0.98
0.56 0.97
0.57 0.97
0.58 0.97
0.59 0.92
0.6 0.97
0.61 0.98
0.62 0.93
0.63 0.96
0.64 0.99
0.65 0.97
0.66 0.96
0.67 0.98
0.68 1
0.69 0.99
0.7 0.99
0.71 0.94
0.72 0.99
0.73 0.96
0.74 0.95
0.75 0.97
0.76 0.99
0.77 0.98
0.78 0.97
0.79 0.98
0.8 0.95
0.81 0.97
0.82 0.98
0.83 0.98
0.84 0.98
0.85 0.98
0.86 0.98
0.87 1
0.88 0.99
0.89 0.98
0.9 0.98
0.91 0.99
0.92 0.98
0.93 0.99
0.94 1
0.95 0.98
0.96 0.97
0.97 0.97
0.98 0.96
0.99 0.99
1 0.98
1.02 0.92
1.03 0.88
1.04 0.82
1.05 0.79
1.06 0.73
1.07 0.68
1.08 0.7
1.09 0.65
1.1 0.67
1.11 0.58
1.12 0.68
1.13 0.64
1.14 0.62
1.15 0.66
1.16 0.6
1.17 0.69
1.18 0.62
1.19 0.61
1.2 0.64
1.21 0.51
1.22 0.65
1.23 0.55
1.24 0.57
1.25 0.58
1.26 0.62
1.27 0.49
1.28 0.6
1.29 0.55
1.3 0.54
1.31 0.61
1.32 0.57
1.33 0.59
1.34 0.67
1.35 0.58
1.36 0.57
1.37 0.62
1.38 0.61
1.39 0.51
1.4 0.63
1.41 0.61
1.42 0.68
1.43 0.6
1.44 0.66
1.45 0.66
1.46 0.56
1.47 0.55
1.48 0.67
1.49 0.45
1.5 0.51
1.51 0.59
1.52 0.54
1.53 0.54
1.54 0.56
1.55 0.6
1.56 0.62
1.57 0.53
1.58 0.66
1.59 0.53
1.6 0.62
1.61 0.58
1.62 0.6
1.63 0.62
1.64 0.53
1.65 0.61
1.66 0.55
1.67 0.47
1.68 0.62
1.69 0.6
1.7 0.68
1.71 0.57
1.72 0.66
1.73 0.62
1.74 0.53
1.75 0.53
1.76 0.57
1.77 0.63
1.78 0.6
1.79 0.7
1.8 0.54
1.81 0.55
1.82 0.65
1.83 0.59
1.84 0.64
1.85 0.64
1.86 0.61
1.87 0.52
1.88 0.57
1.89 0.62
1.9 0.6
1.91 0.58
1.92 0.57
1.93 0.6
1.94 0.64
1.95 0.51
1.96 0.58
1.97 0.64
1.98 0.57
1.99 0.61
2 0.65
};
\addplot [semithick, color1, opacity=0.75]
table {%
0.01 0.51
0.02 0.59
0.03 0.57
0.04 0.48
0.05 0.57
0.06 0.49
0.07 0.65
0.08 0.55
0.09 0.69
0.1 0.74
0.11 0.65
0.12 0.71
0.13 0.7
0.14 0.65
0.15 0.66
0.16 0.7
0.17 0.7
0.18 0.75
0.19 0.77
0.2 0.67
0.21 0.71
0.22 0.78
0.23 0.78
0.24 0.78
0.25 0.74
0.26 0.82
0.27 0.84
0.28 0.9
0.29 0.85
0.3 0.78
0.31 0.82
0.32 0.83
0.33 0.83
0.34 0.85
0.35 0.8
0.36 0.87
0.37 0.91
0.38 0.92
0.39 0.84
0.4 0.85
0.41 0.89
0.42 0.9
0.43 0.88
0.44 0.91
0.45 0.92
0.46 0.91
0.47 0.88
0.48 0.9
0.49 0.93
0.5 0.92
0.51 0.89
0.52 0.93
0.53 0.94
0.54 0.96
0.55 0.93
0.56 0.94
0.57 0.94
0.58 0.97
0.59 0.88
0.6 0.95
0.61 0.9
0.62 0.94
0.63 0.93
0.64 0.93
0.65 0.98
0.66 0.92
0.67 0.94
0.68 0.93
0.69 0.96
0.7 0.92
0.71 0.91
0.72 0.96
0.73 0.91
0.74 0.88
0.75 0.97
0.76 0.98
0.77 0.97
0.78 0.93
0.79 0.95
0.8 0.94
0.81 0.9
0.82 0.98
0.83 0.95
0.84 0.99
0.85 0.96
0.86 0.97
0.87 0.94
0.88 0.98
0.89 0.95
0.9 0.94
0.91 0.94
0.92 0.98
0.93 0.95
0.94 0.97
0.95 0.94
0.96 0.92
0.97 0.95
0.98 0.95
0.99 0.97
1 0.94
1.02 0.89
1.03 0.76
1.04 0.69
1.05 0.55
1.06 0.55
1.07 0.46
1.08 0.48
1.09 0.56
1.1 0.58
1.11 0.46
1.12 0.64
1.13 0.56
1.14 0.47
1.15 0.55
1.16 0.52
1.17 0.49
1.18 0.57
1.19 0.49
1.2 0.51
1.21 0.43
1.22 0.43
1.23 0.49
1.24 0.58
1.25 0.5
1.26 0.54
1.27 0.53
1.28 0.53
1.29 0.54
1.3 0.51
1.31 0.47
1.32 0.57
1.33 0.52
1.34 0.57
1.35 0.48
1.36 0.61
1.37 0.55
1.38 0.41
1.39 0.52
1.4 0.57
1.41 0.54
1.42 0.56
1.43 0.57
1.44 0.52
1.45 0.54
1.46 0.46
1.47 0.52
1.48 0.57
1.49 0.5
1.5 0.44
1.51 0.49
1.52 0.48
1.53 0.5
1.54 0.49
1.55 0.53
1.56 0.43
1.57 0.46
1.58 0.55
1.59 0.5
1.6 0.54
1.61 0.47
1.62 0.47
1.63 0.51
1.64 0.54
1.65 0.51
1.66 0.49
1.67 0.51
1.68 0.51
1.69 0.48
1.7 0.49
1.71 0.51
1.72 0.44
1.73 0.55
1.74 0.54
1.75 0.49
1.76 0.46
1.77 0.65
1.78 0.59
1.79 0.49
1.8 0.51
1.81 0.53
1.82 0.58
1.83 0.49
1.84 0.5
1.85 0.48
1.86 0.44
1.87 0.51
1.88 0.54
1.89 0.47
1.9 0.49
1.91 0.52
1.92 0.57
1.93 0.53
1.94 0.5
1.95 0.48
1.96 0.51
1.97 0.44
1.98 0.52
1.99 0.47
2 0.47
};
\addplot [semithick, color2, opacity=0.75]
table {%
0.01 0.52
0.02 0.51
0.03 0.47
0.04 0.52
0.05 0.57
0.06 0.54
0.07 0.55
0.08 0.57
0.09 0.62
0.1 0.66
0.11 0.65
0.12 0.7
0.13 0.7
0.14 0.65
0.15 0.65
0.16 0.69
0.17 0.7
0.18 0.75
0.19 0.77
0.2 0.67
0.21 0.68
0.22 0.8
0.23 0.77
0.24 0.77
0.25 0.72
0.26 0.82
0.27 0.84
0.28 0.9
0.29 0.85
0.3 0.78
0.31 0.82
0.32 0.84
0.33 0.82
0.34 0.85
0.35 0.8
0.36 0.87
0.37 0.91
0.38 0.93
0.39 0.84
0.4 0.85
0.41 0.89
0.42 0.9
0.43 0.88
0.44 0.91
0.45 0.92
0.46 0.91
0.47 0.88
0.48 0.9
0.49 0.93
0.5 0.92
0.51 0.89
0.52 0.93
0.53 0.94
0.54 0.95
0.55 0.94
0.56 0.93
0.57 0.95
0.58 0.97
0.59 0.88
0.6 0.94
0.61 0.9
0.62 0.94
0.63 0.93
0.64 0.93
0.65 0.98
0.66 0.92
0.67 0.94
0.68 0.94
0.69 0.97
0.7 0.92
0.71 0.91
0.72 0.96
0.73 0.92
0.74 0.88
0.75 0.97
0.76 0.98
0.77 0.97
0.78 0.93
0.79 0.95
0.8 0.94
0.81 0.9
0.82 0.98
0.83 0.96
0.84 0.99
0.85 0.96
0.86 0.97
0.87 0.95
0.88 0.98
0.89 0.95
0.9 0.94
0.91 0.94
0.92 0.98
0.93 0.95
0.94 0.97
0.95 0.94
0.96 0.92
0.97 0.95
0.98 0.93
0.99 0.97
1 0.94
1.02 0.9
1.03 0.76
1.04 0.66
1.05 0.56
1.06 0.54
1.07 0.45
1.08 0.48
1.09 0.56
1.1 0.57
1.11 0.46
1.12 0.63
1.13 0.55
1.14 0.47
1.15 0.52
1.16 0.52
1.17 0.51
1.18 0.58
1.19 0.5
1.2 0.52
1.21 0.44
1.22 0.43
1.23 0.5
1.24 0.58
1.25 0.49
1.26 0.53
1.27 0.55
1.28 0.54
1.29 0.53
1.3 0.5
1.31 0.49
1.32 0.56
1.33 0.48
1.34 0.56
1.35 0.5
1.36 0.6
1.37 0.55
1.38 0.4
1.39 0.51
1.4 0.57
1.41 0.53
1.42 0.56
1.43 0.56
1.44 0.54
1.45 0.54
1.46 0.49
1.47 0.52
1.48 0.56
1.49 0.49
1.5 0.43
1.51 0.54
1.52 0.46
1.53 0.53
1.54 0.46
1.55 0.53
1.56 0.45
1.57 0.47
1.58 0.54
1.59 0.52
1.6 0.52
1.61 0.45
1.62 0.5
1.63 0.52
1.64 0.55
1.65 0.52
1.66 0.47
1.67 0.51
1.68 0.52
1.69 0.47
1.7 0.51
1.71 0.54
1.72 0.46
1.73 0.55
1.74 0.53
1.75 0.48
1.76 0.46
1.77 0.64
1.78 0.6
1.79 0.5
1.8 0.5
1.81 0.51
1.82 0.58
1.83 0.47
1.84 0.49
1.85 0.49
1.86 0.43
1.87 0.53
1.88 0.52
1.89 0.5
1.9 0.51
1.91 0.53
1.92 0.59
1.93 0.49
1.94 0.52
1.95 0.5
1.96 0.51
1.97 0.44
1.98 0.51
1.99 0.5
2 0.46
};
\addplot [semithick, color3, opacity=0.75]
table {%
0.01 0.47
0.02 0.54
0.03 0.55
0.04 0.57
0.05 0.64
0.06 0.56
0.07 0.66
0.08 0.6
0.09 0.72
0.1 0.69
0.11 0.67
0.12 0.73
0.13 0.74
0.14 0.72
0.15 0.66
0.16 0.71
0.17 0.78
0.18 0.78
0.19 0.78
0.2 0.6
0.21 0.71
0.22 0.8
0.23 0.75
0.24 0.86
0.25 0.76
0.26 0.82
0.27 0.83
0.28 0.84
0.29 0.86
0.3 0.83
0.31 0.82
0.32 0.91
0.33 0.84
0.34 0.85
0.35 0.8
0.36 0.87
0.37 0.9
0.38 0.9
0.39 0.86
0.4 0.87
0.41 0.88
0.42 0.85
0.43 0.87
0.44 0.86
0.45 0.94
0.46 0.93
0.47 0.89
0.48 0.89
0.49 0.88
0.5 0.9
0.51 0.91
0.52 0.91
0.53 0.91
0.54 0.94
0.55 0.96
0.56 0.89
0.57 0.94
0.58 0.97
0.59 0.88
0.6 0.93
0.61 0.9
0.62 0.92
0.63 0.93
0.64 0.91
0.65 0.9
0.66 0.93
0.67 0.9
0.68 0.9
0.69 0.94
0.7 0.92
0.71 0.86
0.72 0.92
0.73 0.87
0.74 0.87
0.75 0.91
0.76 0.96
0.77 0.95
0.78 0.91
0.79 0.94
0.8 0.9
0.81 0.9
0.82 0.93
0.83 0.92
0.84 0.93
0.85 0.9
0.86 0.95
0.87 0.91
0.88 0.96
0.89 0.92
0.9 0.9
0.91 0.93
0.92 0.95
0.93 0.95
0.94 0.96
0.95 0.9
0.96 0.9
0.97 0.91
0.98 0.9
0.99 0.94
1 0.93
1.02 0.88
1.03 0.85
1.04 0.82
1.05 0.76
1.06 0.72
1.07 0.69
1.08 0.67
1.09 0.62
1.1 0.64
1.11 0.59
1.12 0.7
1.13 0.68
1.14 0.59
1.15 0.62
1.16 0.62
1.17 0.62
1.18 0.65
1.19 0.59
1.2 0.61
1.21 0.61
1.22 0.61
1.23 0.56
1.24 0.54
1.25 0.56
1.26 0.7
1.27 0.59
1.28 0.64
1.29 0.53
1.3 0.48
1.31 0.69
1.32 0.61
1.33 0.58
1.34 0.61
1.35 0.64
1.36 0.63
1.37 0.65
1.38 0.67
1.39 0.57
1.4 0.61
1.41 0.67
1.42 0.64
1.43 0.55
1.44 0.62
1.45 0.68
1.46 0.47
1.47 0.51
1.48 0.66
1.49 0.49
1.5 0.56
1.51 0.62
1.52 0.57
1.53 0.58
1.54 0.65
1.55 0.61
1.56 0.56
1.57 0.52
1.58 0.54
1.59 0.6
1.6 0.64
1.61 0.59
1.62 0.55
1.63 0.62
1.64 0.47
1.65 0.67
1.66 0.59
1.67 0.45
1.68 0.58
1.69 0.64
1.7 0.67
1.71 0.62
1.72 0.57
1.73 0.62
1.74 0.57
1.75 0.59
1.76 0.62
1.77 0.64
1.78 0.56
1.79 0.65
1.8 0.55
1.81 0.5
1.82 0.65
1.83 0.63
1.84 0.61
1.85 0.64
1.86 0.56
1.87 0.57
1.88 0.54
1.89 0.63
1.9 0.63
1.91 0.66
1.92 0.57
1.93 0.57
1.94 0.59
1.95 0.5
1.96 0.57
1.97 0.63
1.98 0.63
1.99 0.56
2 0.58
};
\addplot [semithick, color4, opacity=0.75]
table {%
0.01 0.47
0.02 0.55
0.03 0.55
0.04 0.57
0.05 0.64
0.06 0.56
0.07 0.65
0.08 0.6
0.09 0.71
0.1 0.7
0.11 0.67
0.12 0.73
0.13 0.74
0.14 0.71
0.15 0.64
0.16 0.71
0.17 0.77
0.18 0.78
0.19 0.79
0.2 0.62
0.21 0.7
0.22 0.76
0.23 0.75
0.24 0.84
0.25 0.74
0.26 0.81
0.27 0.83
0.28 0.84
0.29 0.84
0.3 0.82
0.31 0.81
0.32 0.89
0.33 0.83
0.34 0.86
0.35 0.78
0.36 0.86
0.37 0.87
0.38 0.89
0.39 0.85
0.4 0.84
0.41 0.88
0.42 0.85
0.43 0.84
0.44 0.86
0.45 0.94
0.46 0.91
0.47 0.9
0.48 0.89
0.49 0.88
0.5 0.88
0.51 0.88
0.52 0.89
0.53 0.92
0.54 0.93
0.55 0.96
0.56 0.89
0.57 0.94
0.58 0.96
0.59 0.86
0.6 0.93
0.61 0.88
0.62 0.92
0.63 0.92
0.64 0.9
0.65 0.9
0.66 0.93
0.67 0.9
0.68 0.89
0.69 0.92
0.7 0.9
0.71 0.84
0.72 0.92
0.73 0.85
0.74 0.86
0.75 0.9
0.76 0.95
0.77 0.94
0.78 0.9
0.79 0.93
0.8 0.9
0.81 0.9
0.82 0.92
0.83 0.9
0.84 0.91
0.85 0.89
0.86 0.93
0.87 0.9
0.88 0.92
0.89 0.89
0.9 0.87
0.91 0.92
0.92 0.95
0.93 0.92
0.94 0.95
0.95 0.88
0.96 0.88
0.97 0.87
0.98 0.88
0.99 0.92
1 0.91
1.02 0.87
1.03 0.84
1.04 0.8
1.05 0.75
1.06 0.72
1.07 0.68
1.08 0.68
1.09 0.62
1.1 0.64
1.11 0.57
1.12 0.7
1.13 0.66
1.14 0.61
1.15 0.62
1.16 0.61
1.17 0.61
1.18 0.66
1.19 0.58
1.2 0.61
1.21 0.6
1.22 0.59
1.23 0.57
1.24 0.54
1.25 0.55
1.26 0.7
1.27 0.6
1.28 0.63
1.29 0.54
1.3 0.5
1.31 0.69
1.32 0.62
1.33 0.58
1.34 0.6
1.35 0.64
1.36 0.62
1.37 0.65
1.38 0.65
1.39 0.58
1.4 0.6
1.41 0.67
1.42 0.63
1.43 0.56
1.44 0.62
1.45 0.68
1.46 0.49
1.47 0.53
1.48 0.68
1.49 0.49
1.5 0.56
1.51 0.61
1.52 0.57
1.53 0.6
1.54 0.64
1.55 0.61
1.56 0.57
1.57 0.53
1.58 0.55
1.59 0.61
1.6 0.65
1.61 0.59
1.62 0.56
1.63 0.63
1.64 0.48
1.65 0.68
1.66 0.59
1.67 0.47
1.68 0.58
1.69 0.65
1.7 0.67
1.71 0.62
1.72 0.54
1.73 0.63
1.74 0.57
1.75 0.57
1.76 0.61
1.77 0.63
1.78 0.55
1.79 0.66
1.8 0.54
1.81 0.5
1.82 0.65
1.83 0.62
1.84 0.6
1.85 0.63
1.86 0.55
1.87 0.58
1.88 0.53
1.89 0.61
1.9 0.63
1.91 0.66
1.92 0.57
1.93 0.57
1.94 0.6
1.95 0.5
1.96 0.57
1.97 0.62
1.98 0.64
1.99 0.59
2 0.6
};
\addplot [semithick, color5, opacity=0.75]
table {%
0.01 0.52
0.02 0.56
0.03 0.55
0.04 0.56
0.05 0.62
0.06 0.54
0.07 0.65
0.08 0.54
0.09 0.68
0.1 0.62
0.11 0.64
0.12 0.64
0.13 0.7
0.14 0.69
0.15 0.64
0.16 0.71
0.17 0.71
0.18 0.77
0.19 0.71
0.2 0.63
0.21 0.7
0.22 0.75
0.23 0.72
0.24 0.83
0.25 0.73
0.26 0.77
0.27 0.8
0.28 0.82
0.29 0.81
0.3 0.72
0.31 0.76
0.32 0.79
0.33 0.78
0.34 0.81
0.35 0.75
0.36 0.81
0.37 0.81
0.38 0.88
0.39 0.8
0.4 0.8
0.41 0.84
0.42 0.84
0.43 0.82
0.44 0.82
0.45 0.87
0.46 0.87
0.47 0.8
0.48 0.85
0.49 0.85
0.5 0.84
0.51 0.8
0.52 0.88
0.53 0.86
0.54 0.86
0.55 0.85
0.56 0.83
0.57 0.87
0.58 0.92
0.59 0.85
0.6 0.85
0.61 0.8
0.62 0.88
0.63 0.88
0.64 0.87
0.65 0.86
0.66 0.85
0.67 0.83
0.68 0.84
0.69 0.88
0.7 0.85
0.71 0.84
0.72 0.87
0.73 0.86
0.74 0.79
0.75 0.88
0.76 0.92
0.77 0.87
0.78 0.89
0.79 0.9
0.8 0.89
0.81 0.85
0.82 0.88
0.83 0.85
0.84 0.89
0.85 0.86
0.86 0.86
0.87 0.82
0.88 0.91
0.89 0.87
0.9 0.79
0.91 0.88
0.92 0.92
0.93 0.9
0.94 0.89
0.95 0.86
0.96 0.84
0.97 0.84
0.98 0.87
0.99 0.89
1 0.9
1.02 0.86
1.03 0.82
1.04 0.69
1.05 0.66
1.06 0.69
1.07 0.66
1.08 0.69
1.09 0.58
1.1 0.67
1.11 0.57
1.12 0.69
1.13 0.67
1.14 0.6
1.15 0.54
1.16 0.58
1.17 0.64
1.18 0.62
1.19 0.64
1.2 0.63
1.21 0.55
1.22 0.59
1.23 0.58
1.24 0.54
1.25 0.58
1.26 0.69
1.27 0.62
1.28 0.65
1.29 0.61
1.3 0.61
1.31 0.66
1.32 0.6
1.33 0.58
1.34 0.64
1.35 0.66
1.36 0.64
1.37 0.66
1.38 0.58
1.39 0.6
1.4 0.61
1.41 0.61
1.42 0.66
1.43 0.53
1.44 0.64
1.45 0.79
1.46 0.56
1.47 0.56
1.48 0.67
1.49 0.59
1.5 0.6
1.51 0.67
1.52 0.56
1.53 0.59
1.54 0.67
1.55 0.62
1.56 0.61
1.57 0.56
1.58 0.6
1.59 0.6
1.6 0.72
1.61 0.62
1.62 0.6
1.63 0.61
1.64 0.55
1.65 0.68
1.66 0.63
1.67 0.62
1.68 0.71
1.69 0.59
1.7 0.72
1.71 0.61
1.72 0.65
1.73 0.62
1.74 0.63
1.75 0.53
1.76 0.64
1.77 0.64
1.78 0.57
1.79 0.59
1.8 0.58
1.81 0.55
1.82 0.63
1.83 0.6
1.84 0.65
1.85 0.69
1.86 0.57
1.87 0.61
1.88 0.62
1.89 0.6
1.9 0.64
1.91 0.57
1.92 0.57
1.93 0.58
1.94 0.58
1.95 0.48
1.96 0.59
1.97 0.59
1.98 0.58
1.99 0.58
2 0.61
};
\addplot [semithick, color6, opacity=0.75, dashed]
table {%
0.01 0.58
0.02 0.46
0.03 0.48
0.04 0.51
0.05 0.54
0.06 0.44
0.07 0.61
0.08 0.53
0.09 0.49
0.1 0.52
0.11 0.49
0.12 0.51
0.13 0.46
0.14 0.51
0.15 0.52
0.16 0.57
0.17 0.5
0.18 0.51
0.19 0.58
0.2 0.65
0.21 0.56
0.22 0.66
0.23 0.52
0.24 0.46
0.25 0.45
0.26 0.53
0.27 0.55
0.28 0.62
0.29 0.53
0.3 0.58
0.31 0.57
0.32 0.59
0.33 0.57
0.34 0.68
0.35 0.65
0.36 0.58
0.37 0.67
0.38 0.62
0.39 0.65
0.4 0.66
0.41 0.6
0.42 0.63
0.43 0.68
0.44 0.59
0.45 0.66
0.46 0.62
0.47 0.65
0.48 0.63
0.49 0.68
0.5 0.67
0.51 0.61
0.52 0.71
0.53 0.72
0.54 0.73
0.55 0.65
0.56 0.6
0.57 0.68
0.58 0.7
0.59 0.67
0.6 0.69
0.61 0.68
0.62 0.69
0.63 0.71
0.64 0.66
0.65 0.67
0.66 0.76
0.67 0.76
0.68 0.64
0.69 0.67
0.7 0.77
0.71 0.66
0.72 0.73
0.73 0.65
0.74 0.74
0.75 0.75
0.76 0.67
0.77 0.72
0.78 0.7
0.79 0.67
0.8 0.65
0.81 0.73
0.82 0.71
0.83 0.74
0.84 0.67
0.85 0.62
0.86 0.69
0.87 0.69
0.88 0.67
0.89 0.69
0.9 0.66
0.91 0.7
0.92 0.71
0.93 0.7
0.94 0.67
0.95 0.75
0.96 0.68
0.97 0.73
0.98 0.64
0.99 0.64
1 0.69
1.02 0.66
1.03 0.66
1.04 0.55
1.05 0.62
1.06 0.48
1.07 0.53
1.08 0.5
1.09 0.49
1.1 0.59
1.11 0.56
1.12 0.54
1.13 0.46
1.14 0.55
1.15 0.5
1.16 0.56
1.17 0.46
1.18 0.54
1.19 0.55
1.2 0.53
1.21 0.51
1.22 0.52
1.23 0.47
1.24 0.43
1.25 0.51
1.26 0.51
1.27 0.58
1.28 0.48
1.29 0.45
1.3 0.45
1.31 0.46
1.32 0.56
1.33 0.48
1.34 0.55
1.35 0.46
1.36 0.54
1.37 0.54
1.38 0.55
1.39 0.45
1.4 0.5
1.41 0.46
1.42 0.46
1.43 0.5
1.44 0.47
1.45 0.44
1.46 0.38
1.47 0.46
1.48 0.47
1.49 0.56
1.5 0.54
1.51 0.43
1.52 0.51
1.53 0.52
1.54 0.49
1.55 0.55
1.56 0.51
1.57 0.54
1.58 0.49
1.59 0.61
1.6 0.54
1.61 0.49
1.62 0.49
1.63 0.54
1.64 0.48
1.65 0.51
1.66 0.44
1.67 0.49
1.68 0.48
1.69 0.52
1.7 0.58
1.71 0.56
1.72 0.48
1.73 0.49
1.74 0.49
1.75 0.53
1.76 0.53
1.77 0.5
1.78 0.55
1.79 0.54
1.8 0.57
1.81 0.49
1.82 0.49
1.83 0.45
1.84 0.39
1.85 0.56
1.86 0.4
1.87 0.52
1.88 0.51
1.89 0.48
1.9 0.43
1.91 0.52
1.92 0.46
1.93 0.45
1.94 0.53
1.95 0.63
1.96 0.51
1.97 0.5
1.98 0.47
1.99 0.61
2 0.51
};
\addplot [semithick, color7, opacity=0.75, dashed]
table {%
0.01 0.58
0.02 0.46
0.03 0.48
0.04 0.51
0.05 0.54
0.06 0.44
0.07 0.61
0.08 0.53
0.09 0.49
0.1 0.52
0.11 0.49
0.12 0.51
0.13 0.46
0.14 0.51
0.15 0.52
0.16 0.57
0.17 0.5
0.18 0.51
0.19 0.58
0.2 0.65
0.21 0.56
0.22 0.66
0.23 0.52
0.24 0.46
0.25 0.45
0.26 0.53
0.27 0.55
0.28 0.62
0.29 0.53
0.3 0.58
0.31 0.57
0.32 0.59
0.33 0.57
0.34 0.68
0.35 0.65
0.36 0.58
0.37 0.67
0.38 0.62
0.39 0.65
0.4 0.66
0.41 0.6
0.42 0.63
0.43 0.68
0.44 0.59
0.45 0.66
0.46 0.62
0.47 0.65
0.48 0.63
0.49 0.68
0.5 0.67
0.51 0.61
0.52 0.71
0.53 0.72
0.54 0.73
0.55 0.65
0.56 0.6
0.57 0.68
0.58 0.7
0.59 0.67
0.6 0.69
0.61 0.68
0.62 0.69
0.63 0.71
0.64 0.66
0.65 0.67
0.66 0.76
0.67 0.76
0.68 0.64
0.69 0.67
0.7 0.77
0.71 0.66
0.72 0.73
0.73 0.65
0.74 0.74
0.75 0.75
0.76 0.67
0.77 0.72
0.78 0.7
0.79 0.67
0.8 0.65
0.81 0.73
0.82 0.71
0.83 0.74
0.84 0.67
0.85 0.62
0.86 0.69
0.87 0.69
0.88 0.67
0.89 0.69
0.9 0.66
0.91 0.7
0.92 0.71
0.93 0.7
0.94 0.67
0.95 0.75
0.96 0.68
0.97 0.73
0.98 0.64
0.99 0.64
1 0.69
1.02 0.66
1.03 0.66
1.04 0.55
1.05 0.62
1.06 0.48
1.07 0.53
1.08 0.5
1.09 0.49
1.1 0.59
1.11 0.56
1.12 0.54
1.13 0.46
1.14 0.55
1.15 0.5
1.16 0.56
1.17 0.46
1.18 0.54
1.19 0.55
1.2 0.53
1.21 0.51
1.22 0.52
1.23 0.47
1.24 0.43
1.25 0.51
1.26 0.51
1.27 0.58
1.28 0.48
1.29 0.45
1.3 0.45
1.31 0.46
1.32 0.56
1.33 0.48
1.34 0.55
1.35 0.46
1.36 0.54
1.37 0.54
1.38 0.55
1.39 0.45
1.4 0.5
1.41 0.46
1.42 0.46
1.43 0.5
1.44 0.47
1.45 0.44
1.46 0.38
1.47 0.46
1.48 0.47
1.49 0.56
1.5 0.54
1.51 0.43
1.52 0.51
1.53 0.52
1.54 0.49
1.55 0.55
1.56 0.51
1.57 0.54
1.58 0.49
1.59 0.61
1.6 0.54
1.61 0.49
1.62 0.49
1.63 0.54
1.64 0.48
1.65 0.51
1.66 0.44
1.67 0.49
1.68 0.48
1.69 0.52
1.7 0.58
1.71 0.56
1.72 0.48
1.73 0.49
1.74 0.49
1.75 0.53
1.76 0.53
1.77 0.5
1.78 0.55
1.79 0.54
1.8 0.57
1.81 0.49
1.82 0.49
1.83 0.45
1.84 0.39
1.85 0.56
1.86 0.4
1.87 0.52
1.88 0.51
1.89 0.48
1.9 0.43
1.91 0.52
1.92 0.46
1.93 0.45
1.94 0.53
1.95 0.63
1.96 0.51
1.97 0.5
1.98 0.47
1.99 0.61
2 0.51
};
\addplot [semithick, color8, opacity=0.75, dashed]
table {%
0.01 0.54
0.02 0.5
0.03 0.46
0.04 0.51
0.05 0.54
0.06 0.44
0.07 0.6
0.08 0.47
0.09 0.44
0.1 0.47
0.11 0.44
0.12 0.58
0.13 0.53
0.14 0.59
0.15 0.49
0.16 0.49
0.17 0.41
0.18 0.55
0.19 0.63
0.2 0.62
0.21 0.52
0.22 0.66
0.23 0.54
0.24 0.59
0.25 0.54
0.26 0.59
0.27 0.56
0.28 0.62
0.29 0.61
0.3 0.67
0.31 0.67
0.32 0.54
0.33 0.65
0.34 0.7
0.35 0.65
0.36 0.65
0.37 0.69
0.38 0.67
0.39 0.73
0.4 0.74
0.41 0.61
0.42 0.57
0.43 0.68
0.44 0.67
0.45 0.69
0.46 0.72
0.47 0.69
0.48 0.71
0.49 0.68
0.5 0.76
0.51 0.71
0.52 0.71
0.53 0.68
0.54 0.78
0.55 0.64
0.56 0.7
0.57 0.75
0.58 0.76
0.59 0.7
0.6 0.67
0.61 0.67
0.62 0.73
0.63 0.67
0.64 0.69
0.65 0.74
0.66 0.79
0.67 0.8
0.68 0.65
0.69 0.75
0.7 0.72
0.71 0.75
0.72 0.79
0.73 0.71
0.74 0.76
0.75 0.76
0.76 0.73
0.77 0.76
0.78 0.65
0.79 0.73
0.8 0.67
0.81 0.73
0.82 0.77
0.83 0.84
0.84 0.71
0.85 0.78
0.86 0.79
0.87 0.7
0.88 0.72
0.89 0.72
0.9 0.69
0.91 0.71
0.92 0.82
0.93 0.82
0.94 0.64
0.95 0.74
0.96 0.75
0.97 0.73
0.98 0.81
0.99 0.68
1 0.74
1.02 0.67
1.03 0.62
1.04 0.52
1.05 0.58
1.06 0.58
1.07 0.56
1.08 0.56
1.09 0.47
1.1 0.64
1.11 0.48
1.12 0.51
1.13 0.57
1.14 0.54
1.15 0.43
1.16 0.49
1.17 0.44
1.18 0.54
1.19 0.47
1.2 0.57
1.21 0.43
1.22 0.56
1.23 0.53
1.24 0.32
1.25 0.52
1.26 0.48
1.27 0.47
1.28 0.52
1.29 0.53
1.3 0.51
1.31 0.54
1.32 0.48
1.33 0.44
1.34 0.44
1.35 0.46
1.36 0.46
1.37 0.44
1.38 0.49
1.39 0.43
1.4 0.43
1.41 0.53
1.42 0.43
1.43 0.52
1.44 0.44
1.45 0.55
1.46 0.52
1.47 0.44
1.48 0.54
1.49 0.51
1.5 0.5
1.51 0.49
1.52 0.51
1.53 0.5
1.54 0.49
1.55 0.56
1.56 0.48
1.57 0.55
1.58 0.47
1.59 0.6
1.6 0.45
1.61 0.52
1.62 0.46
1.63 0.5
1.64 0.6
1.65 0.58
1.66 0.54
1.67 0.48
1.68 0.47
1.69 0.58
1.7 0.52
1.71 0.57
1.72 0.55
1.73 0.52
1.74 0.47
1.75 0.5
1.76 0.53
1.77 0.46
1.78 0.51
1.79 0.53
1.8 0.48
1.81 0.5
1.82 0.54
1.83 0.54
1.84 0.54
1.85 0.47
1.86 0.42
1.87 0.53
1.88 0.59
1.89 0.53
1.9 0.41
1.91 0.51
1.92 0.46
1.93 0.5
1.94 0.5
1.95 0.57
1.96 0.47
1.97 0.45
1.98 0.48
1.99 0.47
2 0.54
};
\addplot [semithick, color9, opacity=0.75, dashed]
table {%
0.01 0.53
0.02 0.53
0.03 0.58
0.04 0.63
0.05 0.61
0.06 0.61
0.07 0.7
0.08 0.68
0.09 0.7
0.1 0.75
0.11 0.78
0.12 0.79
0.13 0.79
0.14 0.85
0.15 0.72
0.16 0.8
0.17 0.83
0.18 0.87
0.19 0.83
0.2 0.84
0.21 0.89
0.22 0.96
0.23 0.88
0.24 0.93
0.25 0.86
0.26 0.89
0.27 0.91
0.28 0.93
0.29 0.96
0.3 0.93
0.31 0.97
0.32 0.96
0.33 0.97
0.34 0.95
0.35 0.96
0.36 0.95
0.37 0.98
0.38 0.97
0.39 0.97
0.4 0.95
0.41 0.99
0.42 0.95
0.43 0.97
0.44 0.98
0.45 0.98
0.46 1
0.47 0.96
0.48 0.97
0.49 0.99
0.5 0.99
0.51 0.99
0.52 1
0.53 0.99
0.54 0.99
0.55 0.99
0.56 0.97
0.57 0.99
0.58 0.99
0.59 0.99
0.6 0.99
0.61 0.98
0.62 0.99
0.63 1
0.64 1
0.65 1
0.66 0.99
0.67 1
0.68 0.99
0.69 1
0.7 1
0.71 0.98
0.72 1
0.73 0.96
0.74 0.99
0.75 0.98
0.76 1
0.77 0.99
0.78 0.99
0.79 1
0.8 0.99
0.81 0.99
0.82 0.98
0.83 1
0.84 1
0.85 0.99
0.86 0.98
0.87 0.99
0.88 1
0.89 0.99
0.9 1
0.91 0.99
0.92 0.98
0.93 1
0.94 1
0.95 1
0.96 0.99
0.97 0.99
0.98 1
0.99 1
1 1
1.02 0.98
1.03 0.97
1.04 0.88
1.05 0.85
1.06 0.84
1.07 0.84
1.08 0.75
1.09 0.74
1.1 0.74
1.11 0.63
1.12 0.73
1.13 0.69
1.14 0.68
1.15 0.7
1.16 0.69
1.17 0.64
1.18 0.61
1.19 0.65
1.2 0.67
1.21 0.58
1.22 0.61
1.23 0.65
1.24 0.63
1.25 0.65
1.26 0.71
1.27 0.49
1.28 0.55
1.29 0.52
1.3 0.53
1.31 0.64
1.32 0.55
1.33 0.53
1.34 0.54
1.35 0.53
1.36 0.57
1.37 0.62
1.38 0.61
1.39 0.58
1.4 0.55
1.41 0.6
1.42 0.59
1.43 0.61
1.44 0.62
1.45 0.58
1.46 0.54
1.47 0.52
1.48 0.6
1.49 0.48
1.5 0.5
1.51 0.58
1.52 0.57
1.53 0.52
1.54 0.66
1.55 0.6
1.56 0.62
1.57 0.47
1.58 0.54
1.59 0.62
1.6 0.57
1.61 0.53
1.62 0.55
1.63 0.63
1.64 0.61
1.65 0.63
1.66 0.59
1.67 0.57
1.68 0.56
1.69 0.62
1.7 0.59
1.71 0.58
1.72 0.55
1.73 0.58
1.74 0.6
1.75 0.49
1.76 0.55
1.77 0.67
1.78 0.58
1.79 0.59
1.8 0.45
1.81 0.5
1.82 0.6
1.83 0.58
1.84 0.56
1.85 0.61
1.86 0.45
1.87 0.58
1.88 0.54
1.89 0.57
1.9 0.53
1.91 0.63
1.92 0.59
1.93 0.54
1.94 0.61
1.95 0.5
1.96 0.51
1.97 0.63
1.98 0.64
1.99 0.59
2 0.53
};
\addplot [semithick, color10, opacity=0.75, dashed]
table {%
0.01 0.49
0.02 0.58
0.03 0.54
0.04 0.57
0.05 0.59
0.06 0.53
0.07 0.65
0.08 0.7
0.09 0.67
0.1 0.74
0.11 0.72
0.12 0.8
0.13 0.76
0.14 0.78
0.15 0.73
0.16 0.78
0.17 0.84
0.18 0.8
0.19 0.81
0.2 0.79
0.21 0.84
0.22 0.93
0.23 0.81
0.24 0.94
0.25 0.8
0.26 0.87
0.27 0.86
0.28 0.93
0.29 0.92
0.3 0.94
0.31 0.95
0.32 0.96
0.33 0.96
0.34 0.94
0.35 0.94
0.36 0.95
0.37 0.97
0.38 0.94
0.39 0.95
0.4 0.96
0.41 1
0.42 0.92
0.43 0.96
0.44 0.96
0.45 0.98
0.46 0.99
0.47 0.92
0.48 0.98
0.49 0.97
0.5 0.99
0.51 0.97
0.52 1
0.53 0.97
0.54 0.98
0.55 0.98
0.56 0.97
0.57 0.99
0.58 0.99
0.59 0.97
0.6 0.98
0.61 0.98
0.62 0.98
0.63 1
0.64 1
0.65 1
0.66 0.99
0.67 0.99
0.68 0.99
0.69 1
0.7 0.99
0.71 0.97
0.72 1
0.73 0.95
0.74 0.97
0.75 0.98
0.76 1
0.77 0.99
0.78 0.98
0.79 0.98
0.8 0.99
0.81 0.99
0.82 0.97
0.83 0.99
0.84 0.99
0.85 0.99
0.86 0.98
0.87 0.99
0.88 0.99
0.89 0.98
0.9 1
0.91 0.98
0.92 0.99
0.93 0.99
0.94 1
0.95 1
0.96 0.99
0.97 0.98
0.98 0.99
0.99 1
1 1
1.02 0.98
1.03 0.95
1.04 0.87
1.05 0.83
1.06 0.79
1.07 0.77
1.08 0.69
1.09 0.67
1.1 0.75
1.11 0.63
1.12 0.74
1.13 0.62
1.14 0.59
1.15 0.71
1.16 0.7
1.17 0.65
1.18 0.62
1.19 0.65
1.2 0.63
1.21 0.61
1.22 0.63
1.23 0.58
1.24 0.56
1.25 0.58
1.26 0.62
1.27 0.5
1.28 0.62
1.29 0.57
1.3 0.52
1.31 0.65
1.32 0.63
1.33 0.55
1.34 0.55
1.35 0.54
1.36 0.59
1.37 0.53
1.38 0.61
1.39 0.53
1.4 0.56
1.41 0.61
1.42 0.61
1.43 0.56
1.44 0.66
1.45 0.58
1.46 0.52
1.47 0.49
1.48 0.66
1.49 0.49
1.5 0.53
1.51 0.57
1.52 0.52
1.53 0.53
1.54 0.6
1.55 0.58
1.56 0.59
1.57 0.45
1.58 0.49
1.59 0.51
1.6 0.62
1.61 0.52
1.62 0.55
1.63 0.51
1.64 0.57
1.65 0.67
1.66 0.53
1.67 0.46
1.68 0.57
1.69 0.63
1.7 0.56
1.71 0.5
1.72 0.51
1.73 0.57
1.74 0.57
1.75 0.53
1.76 0.54
1.77 0.59
1.78 0.53
1.79 0.66
1.8 0.52
1.81 0.54
1.82 0.6
1.83 0.54
1.84 0.53
1.85 0.53
1.86 0.55
1.87 0.56
1.88 0.47
1.89 0.58
1.9 0.56
1.91 0.58
1.92 0.57
1.93 0.54
1.94 0.69
1.95 0.54
1.96 0.52
1.97 0.59
1.98 0.6
1.99 0.56
2 0.6
};
\addplot [semithick, color11, opacity=0.75, dashed]
table {%
0.01 0.48
0.02 0.53
0.03 0.52
0.04 0.53
0.05 0.58
0.06 0.52
0.07 0.68
0.08 0.56
0.09 0.55
0.1 0.64
0.11 0.56
0.12 0.75
0.13 0.66
0.14 0.7
0.15 0.67
0.16 0.72
0.17 0.72
0.18 0.75
0.19 0.76
0.2 0.73
0.21 0.71
0.22 0.84
0.23 0.73
0.24 0.85
0.25 0.8
0.26 0.8
0.27 0.77
0.28 0.86
0.29 0.82
0.3 0.79
0.31 0.83
0.32 0.82
0.33 0.81
0.34 0.86
0.35 0.86
0.36 0.82
0.37 0.82
0.38 0.84
0.39 0.86
0.4 0.88
0.41 0.86
0.42 0.88
0.43 0.88
0.44 0.88
0.45 0.9
0.46 0.87
0.47 0.88
0.48 0.87
0.49 0.91
0.5 0.85
0.51 0.88
0.52 0.88
0.53 0.87
0.54 0.94
0.55 0.92
0.56 0.93
0.57 0.93
0.58 0.87
0.59 0.85
0.6 0.9
0.61 0.87
0.62 0.9
0.63 0.86
0.64 0.96
0.65 0.88
0.66 0.89
0.67 0.93
0.68 0.9
0.69 0.86
0.7 0.91
0.71 0.92
0.72 0.88
0.73 0.86
0.74 0.91
0.75 0.94
0.76 0.89
0.77 0.91
0.78 0.88
0.79 0.9
0.8 0.85
0.81 0.85
0.82 0.94
0.83 0.91
0.84 0.92
0.85 0.88
0.86 0.9
0.87 0.93
0.88 0.94
0.89 0.9
0.9 0.9
0.91 0.94
0.92 0.94
0.93 0.98
0.94 0.94
0.95 0.87
0.96 0.89
0.97 0.94
0.98 0.93
0.99 0.91
1 0.96
1.02 0.91
1.03 0.91
1.04 0.72
1.05 0.75
1.06 0.74
1.07 0.69
1.08 0.63
1.09 0.57
1.1 0.62
1.11 0.6
1.12 0.68
1.13 0.61
1.14 0.58
1.15 0.59
1.16 0.6
1.17 0.6
1.18 0.65
1.19 0.63
1.2 0.66
1.21 0.51
1.22 0.62
1.23 0.5
1.24 0.55
1.25 0.54
1.26 0.57
1.27 0.51
1.28 0.6
1.29 0.46
1.3 0.49
1.31 0.57
1.32 0.61
1.33 0.49
1.34 0.47
1.35 0.53
1.36 0.55
1.37 0.62
1.38 0.62
1.39 0.47
1.4 0.6
1.41 0.53
1.42 0.55
1.43 0.6
1.44 0.5
1.45 0.54
1.46 0.54
1.47 0.54
1.48 0.57
1.49 0.53
1.5 0.49
1.51 0.55
1.52 0.43
1.53 0.5
1.54 0.48
1.55 0.53
1.56 0.49
1.57 0.51
1.58 0.52
1.59 0.51
1.6 0.54
1.61 0.5
1.62 0.54
1.63 0.6
1.64 0.42
1.65 0.59
1.66 0.52
1.67 0.42
1.68 0.55
1.69 0.67
1.7 0.61
1.71 0.56
1.72 0.53
1.73 0.45
1.74 0.54
1.75 0.49
1.76 0.61
1.77 0.56
1.78 0.52
1.79 0.53
1.8 0.44
1.81 0.5
1.82 0.56
1.83 0.55
1.84 0.53
1.85 0.55
1.86 0.53
1.87 0.56
1.88 0.57
1.89 0.59
1.9 0.53
1.91 0.57
1.92 0.52
1.93 0.55
1.94 0.52
1.95 0.53
1.96 0.48
1.97 0.5
1.98 0.55
1.99 0.52
2 0.52
};
\addplot [semithick, black, opacity=1, dash pattern=on 1pt off 1pt]
table {%
-0.0895000000000001 0.5
2.0995 0.5
};
\addplot [semithick, black, opacity=1, dash pattern=on 1pt off 1pt]
table {%
-0.0895000000000001 1
2.0995 1
};
\addplot [semithick, black, opacity=1, dash pattern=on 1pt off 1pt]
table {%
1 0.2
1 1.05
};
\end{axis}

\end{tikzpicture}

%% file: plots/decoupled2/LAPxUNI.tex
% This file was created by tikzplotlib v0.9.6.
\begin{tikzpicture}

\definecolor{color0}{rgb}{0.866666666666667,0.494117647058824,0.164705882352941}
\definecolor{color1}{rgb}{0.164705882352941,0.643137254901961,0.866666666666667}
\definecolor{color2}{rgb}{0.584313725490196,0.866666666666667,0.164705882352941}
\definecolor{color3}{rgb}{0.109803921568627,0.337254901960784,0.129411764705882}
\definecolor{color4}{rgb}{0.529411764705882,0.305882352941176,0.858823529411765}
\definecolor{color5}{rgb}{0.858823529411765,0.305882352941176,0.435294117647059}
\definecolor{color6}{rgb}{0.937254901960784,0.929411764705882,0.392156862745098}
\definecolor{color7}{rgb}{0.0901960784313725,0.486274509803922,0.0980392156862745}
\definecolor{color8}{rgb}{0.156862745098039,0.188235294117647,0.827450980392157}
\definecolor{color9}{rgb}{0.937254901960784,0.392156862745098,0.894117647058824}
\definecolor{color10}{rgb}{0.2,0.184313725490196,0.184313725490196}
\definecolor{color11}{rgb}{0.0156862745098039,0.803921568627451,0.976470588235294}

\begin{axis}[
tick align=outside,
tick pos=left,
x grid style={white!69.0196078431373!black},
xmajorgrids,
xmin=-0.0895, xmax=2.0995,
xtick style={color=black},
xtick={0,0.1,0.2,0.3,0.4,0.5,0.6,0.7,0.8,0.9,1,1.1,1.2,1.3,1.4,1.5,1.6,1.7,1.8,1.9,2},
xticklabels={0,,.2,,.4,,.6,,.8,,1,,20,,40,,60,,80,,100},
height=4.8cm,
width=6.5cm,
y grid style={white!69.0196078431373!black},
ymajorgrids,
ymin=0.2, ymax=1.05,
ytick style={color=black}
]
\addplot [semithick, color0, opacity=0.75]
table {%
0.01 0.41
0.02 0.49
0.03 0.58
0.04 0.54
0.05 0.54
0.06 0.63
0.07 0.58
0.08 0.61
0.09 0.48
0.1 0.62
0.11 0.67
0.12 0.68
0.13 0.67
0.14 0.76
0.15 0.78
0.16 0.74
0.17 0.62
0.18 0.83
0.19 0.82
0.2 0.72
0.21 0.73
0.22 0.85
0.23 0.77
0.24 0.8
0.25 0.83
0.26 0.77
0.27 0.82
0.28 0.74
0.29 0.85
0.3 0.84
0.31 0.88
0.32 0.89
0.33 0.83
0.34 0.86
0.35 0.87
0.36 0.86
0.37 0.91
0.38 0.87
0.39 0.89
0.4 0.98
0.41 0.91
0.42 0.94
0.43 0.95
0.44 0.93
0.45 0.91
0.46 0.97
0.47 0.98
0.48 0.94
0.49 0.95
0.5 0.94
0.51 0.99
0.52 0.95
0.53 0.93
0.54 1
0.55 0.97
0.56 0.96
0.57 0.98
0.58 0.96
0.59 0.98
0.6 0.94
0.61 0.99
0.62 0.98
0.63 0.98
0.64 0.96
0.65 1
0.66 0.98
0.67 0.97
0.68 0.98
0.69 0.98
0.7 0.99
0.71 1
0.72 0.99
0.73 0.99
0.74 1
0.75 0.99
0.76 1
0.77 0.99
0.78 0.98
0.79 1
0.8 0.97
0.81 1
0.82 1
0.83 1
0.84 0.99
0.85 0.98
0.86 1
0.87 0.98
0.88 1
0.89 1
0.9 0.98
0.91 1
0.92 1
0.93 1
0.94 1
0.95 1
0.96 1
0.97 1
0.98 1
0.99 1
1 1
1.02 1
1.03 1
1.04 0.99
1.05 0.97
1.06 0.95
1.07 0.93
1.08 0.9
1.09 0.93
1.1 0.79
1.11 0.84
1.12 0.79
1.13 0.75
1.14 0.75
1.15 0.76
1.16 0.78
1.17 0.72
1.18 0.72
1.19 0.69
1.2 0.67
1.21 0.72
1.22 0.74
1.23 0.73
1.24 0.65
1.25 0.63
1.26 0.63
1.27 0.69
1.28 0.66
1.29 0.73
1.3 0.68
1.31 0.65
1.32 0.69
1.33 0.61
1.34 0.65
1.35 0.61
1.36 0.59
1.37 0.57
1.38 0.6
1.39 0.64
1.4 0.71
1.41 0.63
1.42 0.5
1.43 0.63
1.44 0.64
1.45 0.65
1.46 0.68
1.47 0.59
1.48 0.66
1.49 0.54
1.5 0.66
1.51 0.63
1.52 0.61
1.53 0.55
1.54 0.61
1.55 0.54
1.56 0.61
1.57 0.58
1.58 0.53
1.59 0.71
1.6 0.61
1.61 0.62
1.62 0.51
1.63 0.66
1.64 0.61
1.65 0.62
1.66 0.62
1.67 0.63
1.68 0.53
1.69 0.51
1.7 0.59
1.71 0.68
1.72 0.61
1.73 0.6
1.74 0.63
1.75 0.58
1.76 0.58
1.77 0.61
1.78 0.66
1.79 0.51
1.8 0.54
1.81 0.72
1.82 0.65
1.83 0.62
1.84 0.65
1.85 0.58
1.86 0.72
1.87 0.57
1.88 0.59
1.89 0.62
1.9 0.68
1.91 0.55
1.92 0.64
1.93 0.65
1.94 0.52
1.95 0.61
1.96 0.66
1.97 0.55
1.98 0.62
1.99 0.64
2 0.61
};
\addplot [semithick, color1, opacity=0.75]
table {%
0.01 0.46
0.02 0.51
0.03 0.46
0.04 0.54
0.05 0.49
0.06 0.51
0.07 0.56
0.08 0.57
0.09 0.47
0.1 0.49
0.11 0.67
0.12 0.58
0.13 0.57
0.14 0.65
0.15 0.65
0.16 0.67
0.17 0.59
0.18 0.72
0.19 0.72
0.2 0.63
0.21 0.65
0.22 0.78
0.23 0.61
0.24 0.7
0.25 0.71
0.26 0.63
0.27 0.69
0.28 0.6
0.29 0.73
0.3 0.77
0.31 0.72
0.32 0.78
0.33 0.74
0.34 0.74
0.35 0.78
0.36 0.79
0.37 0.81
0.38 0.74
0.39 0.75
0.4 0.85
0.41 0.84
0.42 0.79
0.43 0.83
0.44 0.76
0.45 0.79
0.46 0.83
0.47 0.85
0.48 0.8
0.49 0.84
0.5 0.81
0.51 0.86
0.52 0.83
0.53 0.82
0.54 0.93
0.55 0.91
0.56 0.85
0.57 0.89
0.58 0.84
0.59 0.85
0.6 0.82
0.61 0.9
0.62 0.92
0.63 0.9
0.64 0.9
0.65 0.93
0.66 0.87
0.67 0.88
0.68 0.9
0.69 0.94
0.7 0.95
0.71 0.97
0.72 0.94
0.73 0.95
0.74 0.95
0.75 0.92
0.76 0.97
0.77 0.9
0.78 0.9
0.79 0.95
0.8 0.95
0.81 0.92
0.82 0.97
0.83 0.93
0.84 0.95
0.85 0.95
0.86 0.95
0.87 0.95
0.88 0.93
0.89 0.97
0.9 0.97
0.91 0.99
0.92 0.97
0.93 0.98
0.94 0.99
0.95 0.99
0.96 0.99
0.97 0.98
0.98 0.97
0.99 0.98
1 1
1.02 1
1.03 0.99
1.04 0.99
1.05 0.98
1.06 0.96
1.07 0.94
1.08 0.91
1.09 0.97
1.1 0.88
1.11 0.84
1.12 0.82
1.13 0.83
1.14 0.81
1.15 0.84
1.16 0.73
1.17 0.76
1.18 0.68
1.19 0.65
1.2 0.67
1.21 0.77
1.22 0.64
1.23 0.66
1.24 0.67
1.25 0.61
1.26 0.63
1.27 0.52
1.28 0.55
1.29 0.53
1.3 0.67
1.31 0.67
1.32 0.63
1.33 0.59
1.34 0.63
1.35 0.55
1.36 0.63
1.37 0.63
1.38 0.54
1.39 0.57
1.4 0.56
1.41 0.63
1.42 0.52
1.43 0.51
1.44 0.53
1.45 0.46
1.46 0.54
1.47 0.56
1.48 0.47
1.49 0.57
1.5 0.48
1.51 0.53
1.52 0.54
1.53 0.53
1.54 0.54
1.55 0.51
1.56 0.44
1.57 0.49
1.58 0.48
1.59 0.48
1.6 0.53
1.61 0.48
1.62 0.52
1.63 0.57
1.64 0.52
1.65 0.56
1.66 0.52
1.67 0.57
1.68 0.57
1.69 0.42
1.7 0.51
1.71 0.53
1.72 0.51
1.73 0.43
1.74 0.58
1.75 0.52
1.76 0.5
1.77 0.53
1.78 0.51
1.79 0.53
1.8 0.42
1.81 0.61
1.82 0.47
1.83 0.5
1.84 0.56
1.85 0.54
1.86 0.5
1.87 0.57
1.88 0.52
1.89 0.54
1.9 0.52
1.91 0.46
1.92 0.48
1.93 0.55
1.94 0.6
1.95 0.51
1.96 0.48
1.97 0.49
1.98 0.47
1.99 0.56
2 0.56
};
\addplot [semithick, color2, opacity=0.75]
table {%
0.01 0.43
0.02 0.45
0.03 0.45
0.04 0.52
0.05 0.57
0.06 0.57
0.07 0.58
0.08 0.54
0.09 0.45
0.1 0.53
0.11 0.57
0.12 0.6
0.13 0.59
0.14 0.58
0.15 0.59
0.16 0.6
0.17 0.61
0.18 0.71
0.19 0.71
0.2 0.63
0.21 0.65
0.22 0.78
0.23 0.61
0.24 0.71
0.25 0.71
0.26 0.64
0.27 0.69
0.28 0.58
0.29 0.74
0.3 0.76
0.31 0.72
0.32 0.77
0.33 0.74
0.34 0.74
0.35 0.78
0.36 0.79
0.37 0.79
0.38 0.74
0.39 0.75
0.4 0.85
0.41 0.81
0.42 0.77
0.43 0.82
0.44 0.77
0.45 0.79
0.46 0.82
0.47 0.85
0.48 0.8
0.49 0.85
0.5 0.82
0.51 0.86
0.52 0.83
0.53 0.82
0.54 0.93
0.55 0.91
0.56 0.85
0.57 0.88
0.58 0.83
0.59 0.85
0.6 0.82
0.61 0.89
0.62 0.92
0.63 0.9
0.64 0.89
0.65 0.92
0.66 0.88
0.67 0.87
0.68 0.9
0.69 0.93
0.7 0.95
0.71 0.96
0.72 0.94
0.73 0.94
0.74 0.95
0.75 0.92
0.76 0.96
0.77 0.89
0.78 0.89
0.79 0.95
0.8 0.95
0.81 0.92
0.82 0.97
0.83 0.93
0.84 0.96
0.85 0.95
0.86 0.93
0.87 0.95
0.88 0.93
0.89 0.97
0.9 0.97
0.91 0.99
0.92 0.97
0.93 0.97
0.94 0.99
0.95 0.99
0.96 0.99
0.97 0.98
0.98 0.97
0.99 0.98
1 1
1.02 1
1.03 1
1.04 0.99
1.05 0.98
1.06 0.96
1.07 0.94
1.08 0.9
1.09 0.97
1.1 0.88
1.11 0.84
1.12 0.81
1.13 0.81
1.14 0.82
1.15 0.84
1.16 0.73
1.17 0.76
1.18 0.7
1.19 0.67
1.2 0.65
1.21 0.78
1.22 0.65
1.23 0.65
1.24 0.66
1.25 0.6
1.26 0.64
1.27 0.51
1.28 0.57
1.29 0.53
1.3 0.67
1.31 0.66
1.32 0.65
1.33 0.58
1.34 0.64
1.35 0.56
1.36 0.64
1.37 0.64
1.38 0.54
1.39 0.58
1.4 0.55
1.41 0.59
1.42 0.51
1.43 0.51
1.44 0.54
1.45 0.47
1.46 0.51
1.47 0.55
1.48 0.46
1.49 0.58
1.5 0.46
1.51 0.51
1.52 0.56
1.53 0.56
1.54 0.55
1.55 0.5
1.56 0.45
1.57 0.5
1.58 0.5
1.59 0.48
1.6 0.49
1.61 0.49
1.62 0.52
1.63 0.58
1.64 0.53
1.65 0.58
1.66 0.53
1.67 0.55
1.68 0.56
1.69 0.43
1.7 0.48
1.71 0.52
1.72 0.53
1.73 0.45
1.74 0.58
1.75 0.56
1.76 0.52
1.77 0.51
1.78 0.53
1.79 0.53
1.8 0.41
1.81 0.63
1.82 0.48
1.83 0.48
1.84 0.55
1.85 0.53
1.86 0.52
1.87 0.57
1.88 0.5
1.89 0.52
1.9 0.51
1.91 0.47
1.92 0.47
1.93 0.54
1.94 0.6
1.95 0.52
1.96 0.48
1.97 0.48
1.98 0.48
1.99 0.57
2 0.54
};
\addplot [semithick, color3, opacity=0.75]
table {%
0.01 0.39
0.02 0.55
0.03 0.5
0.04 0.51
0.05 0.54
0.06 0.56
0.07 0.55
0.08 0.59
0.09 0.51
0.1 0.62
0.11 0.69
0.12 0.64
0.13 0.58
0.14 0.69
0.15 0.74
0.16 0.7
0.17 0.59
0.18 0.82
0.19 0.76
0.2 0.67
0.21 0.73
0.22 0.82
0.23 0.73
0.24 0.78
0.25 0.83
0.26 0.77
0.27 0.75
0.28 0.68
0.29 0.8
0.3 0.83
0.31 0.84
0.32 0.86
0.33 0.81
0.34 0.84
0.35 0.86
0.36 0.81
0.37 0.9
0.38 0.83
0.39 0.86
0.4 0.96
0.41 0.89
0.42 0.91
0.43 0.93
0.44 0.86
0.45 0.86
0.46 0.89
0.47 0.92
0.48 0.89
0.49 0.92
0.5 0.92
0.51 0.96
0.52 0.91
0.53 0.89
0.54 0.96
0.55 0.94
0.56 0.9
0.57 0.97
0.58 0.91
0.59 0.94
0.6 0.91
0.61 0.99
0.62 0.94
0.63 0.95
0.64 0.95
0.65 0.96
0.66 0.96
0.67 0.93
0.68 0.94
0.69 0.98
0.7 0.97
0.71 0.97
0.72 0.97
0.73 0.97
0.74 0.98
0.75 0.97
0.76 0.96
0.77 0.97
0.78 0.97
0.79 0.98
0.8 0.96
0.81 0.98
0.82 1
0.83 0.97
0.84 0.96
0.85 0.97
0.86 1
0.87 0.96
0.88 0.95
0.89 0.98
0.9 0.98
0.91 1
0.92 0.99
0.93 0.99
0.94 1
0.95 0.99
0.96 0.99
0.97 0.99
0.98 0.98
0.99 1
1 1
1.02 1
1.03 0.99
1.04 0.95
1.05 0.92
1.06 0.89
1.07 0.88
1.08 0.85
1.09 0.93
1.1 0.8
1.11 0.8
1.12 0.75
1.13 0.77
1.14 0.7
1.15 0.73
1.16 0.75
1.17 0.75
1.18 0.73
1.19 0.65
1.2 0.67
1.21 0.65
1.22 0.63
1.23 0.74
1.24 0.68
1.25 0.65
1.26 0.65
1.27 0.71
1.28 0.65
1.29 0.69
1.3 0.67
1.31 0.66
1.32 0.7
1.33 0.59
1.34 0.68
1.35 0.61
1.36 0.61
1.37 0.58
1.38 0.59
1.39 0.63
1.4 0.7
1.41 0.65
1.42 0.58
1.43 0.63
1.44 0.59
1.45 0.68
1.46 0.65
1.47 0.58
1.48 0.65
1.49 0.58
1.5 0.65
1.51 0.67
1.52 0.58
1.53 0.61
1.54 0.62
1.55 0.56
1.56 0.65
1.57 0.64
1.58 0.56
1.59 0.65
1.6 0.62
1.61 0.6
1.62 0.53
1.63 0.63
1.64 0.64
1.65 0.61
1.66 0.62
1.67 0.64
1.68 0.59
1.69 0.54
1.7 0.56
1.71 0.66
1.72 0.59
1.73 0.56
1.74 0.63
1.75 0.65
1.76 0.64
1.77 0.64
1.78 0.65
1.79 0.5
1.8 0.62
1.81 0.74
1.82 0.65
1.83 0.61
1.84 0.63
1.85 0.61
1.86 0.67
1.87 0.53
1.88 0.64
1.89 0.62
1.9 0.62
1.91 0.55
1.92 0.6
1.93 0.68
1.94 0.55
1.95 0.62
1.96 0.64
1.97 0.6
1.98 0.63
1.99 0.59
2 0.61
};
\addplot [semithick, color4, opacity=0.75]
table {%
0.01 0.4
0.02 0.55
0.03 0.5
0.04 0.51
0.05 0.54
0.06 0.56
0.07 0.55
0.08 0.58
0.09 0.5
0.1 0.62
0.11 0.7
0.12 0.66
0.13 0.58
0.14 0.69
0.15 0.75
0.16 0.71
0.17 0.6
0.18 0.82
0.19 0.77
0.2 0.68
0.21 0.73
0.22 0.82
0.23 0.73
0.24 0.78
0.25 0.84
0.26 0.79
0.27 0.77
0.28 0.68
0.29 0.79
0.3 0.83
0.31 0.84
0.32 0.86
0.33 0.82
0.34 0.85
0.35 0.87
0.36 0.81
0.37 0.88
0.38 0.85
0.39 0.86
0.4 0.96
0.41 0.89
0.42 0.93
0.43 0.93
0.44 0.87
0.45 0.87
0.46 0.89
0.47 0.93
0.48 0.89
0.49 0.92
0.5 0.93
0.51 0.96
0.52 0.92
0.53 0.89
0.54 0.96
0.55 0.94
0.56 0.91
0.57 0.97
0.58 0.91
0.59 0.94
0.6 0.91
0.61 0.99
0.62 0.94
0.63 0.95
0.64 0.96
0.65 0.96
0.66 0.96
0.67 0.95
0.68 0.94
0.69 0.98
0.7 0.97
0.71 0.97
0.72 0.97
0.73 0.97
0.74 0.99
0.75 0.97
0.76 0.96
0.77 0.97
0.78 0.97
0.79 0.98
0.8 0.97
0.81 0.98
0.82 1
0.83 0.97
0.84 0.97
0.85 0.97
0.86 1
0.87 0.98
0.88 0.97
0.89 0.98
0.9 0.98
0.91 1
0.92 0.99
0.93 0.99
0.94 1
0.95 0.99
0.96 0.99
0.97 0.99
0.98 0.98
0.99 1
1 1
1.02 1
1.03 0.99
1.04 0.95
1.05 0.93
1.06 0.89
1.07 0.89
1.08 0.84
1.09 0.93
1.1 0.8
1.11 0.81
1.12 0.75
1.13 0.78
1.14 0.7
1.15 0.74
1.16 0.75
1.17 0.75
1.18 0.73
1.19 0.67
1.2 0.67
1.21 0.65
1.22 0.64
1.23 0.74
1.24 0.69
1.25 0.64
1.26 0.65
1.27 0.72
1.28 0.65
1.29 0.7
1.3 0.66
1.31 0.67
1.32 0.71
1.33 0.59
1.34 0.68
1.35 0.62
1.36 0.62
1.37 0.58
1.38 0.58
1.39 0.63
1.4 0.7
1.41 0.65
1.42 0.58
1.43 0.64
1.44 0.59
1.45 0.68
1.46 0.65
1.47 0.57
1.48 0.65
1.49 0.58
1.5 0.63
1.51 0.66
1.52 0.58
1.53 0.61
1.54 0.61
1.55 0.56
1.56 0.64
1.57 0.64
1.58 0.56
1.59 0.65
1.6 0.62
1.61 0.6
1.62 0.53
1.63 0.63
1.64 0.64
1.65 0.6
1.66 0.62
1.67 0.65
1.68 0.61
1.69 0.54
1.7 0.57
1.71 0.66
1.72 0.59
1.73 0.56
1.74 0.63
1.75 0.65
1.76 0.64
1.77 0.64
1.78 0.65
1.79 0.5
1.8 0.62
1.81 0.74
1.82 0.67
1.83 0.62
1.84 0.63
1.85 0.6
1.86 0.67
1.87 0.54
1.88 0.64
1.89 0.61
1.9 0.62
1.91 0.54
1.92 0.61
1.93 0.7
1.94 0.55
1.95 0.62
1.96 0.64
1.97 0.61
1.98 0.63
1.99 0.59
2 0.59
};
\addplot [semithick, color5, opacity=0.75]
table {%
0.01 0.41
0.02 0.55
0.03 0.48
0.04 0.56
0.05 0.51
0.06 0.57
0.07 0.59
0.08 0.56
0.09 0.47
0.1 0.63
0.11 0.63
0.12 0.63
0.13 0.59
0.14 0.62
0.15 0.71
0.16 0.65
0.17 0.59
0.18 0.81
0.19 0.79
0.2 0.69
0.21 0.72
0.22 0.83
0.23 0.68
0.24 0.75
0.25 0.85
0.26 0.74
0.27 0.75
0.28 0.65
0.29 0.79
0.3 0.81
0.31 0.79
0.32 0.81
0.33 0.77
0.34 0.78
0.35 0.84
0.36 0.78
0.37 0.78
0.38 0.83
0.39 0.83
0.4 0.93
0.41 0.83
0.42 0.88
0.43 0.9
0.44 0.83
0.45 0.84
0.46 0.9
0.47 0.91
0.48 0.86
0.49 0.88
0.5 0.89
0.51 0.94
0.52 0.88
0.53 0.85
0.54 0.93
0.55 0.9
0.56 0.87
0.57 0.95
0.58 0.9
0.59 0.94
0.6 0.89
0.61 0.93
0.62 0.94
0.63 0.93
0.64 0.92
0.65 0.95
0.66 0.94
0.67 0.91
0.68 0.94
0.69 0.96
0.7 0.95
0.71 0.96
0.72 0.95
0.73 0.97
0.74 0.98
0.75 0.96
0.76 0.92
0.77 0.9
0.78 0.95
0.79 0.95
0.8 0.95
0.81 0.95
0.82 0.99
0.83 0.95
0.84 0.93
0.85 0.96
0.86 0.98
0.87 0.93
0.88 0.93
0.89 0.96
0.9 0.97
0.91 0.98
0.92 0.98
0.93 0.95
0.94 0.99
0.95 0.98
0.96 0.96
0.97 0.99
0.98 0.95
0.99 0.96
1 0.99
1.02 1
1.03 0.97
1.04 0.95
1.05 0.92
1.06 0.86
1.07 0.87
1.08 0.85
1.09 0.88
1.1 0.8
1.11 0.8
1.12 0.72
1.13 0.72
1.14 0.68
1.15 0.71
1.16 0.76
1.17 0.7
1.18 0.71
1.19 0.64
1.2 0.66
1.21 0.65
1.22 0.61
1.23 0.71
1.24 0.65
1.25 0.64
1.26 0.62
1.27 0.66
1.28 0.65
1.29 0.65
1.3 0.68
1.31 0.63
1.32 0.64
1.33 0.58
1.34 0.69
1.35 0.57
1.36 0.6
1.37 0.61
1.38 0.6
1.39 0.64
1.4 0.63
1.41 0.59
1.42 0.55
1.43 0.64
1.44 0.65
1.45 0.65
1.46 0.72
1.47 0.61
1.48 0.67
1.49 0.58
1.5 0.73
1.51 0.69
1.52 0.57
1.53 0.68
1.54 0.6
1.55 0.54
1.56 0.59
1.57 0.67
1.58 0.59
1.59 0.69
1.6 0.6
1.61 0.6
1.62 0.54
1.63 0.65
1.64 0.63
1.65 0.68
1.66 0.65
1.67 0.64
1.68 0.56
1.69 0.6
1.7 0.6
1.71 0.68
1.72 0.58
1.73 0.61
1.74 0.68
1.75 0.62
1.76 0.66
1.77 0.65
1.78 0.7
1.79 0.49
1.8 0.57
1.81 0.76
1.82 0.68
1.83 0.64
1.84 0.65
1.85 0.61
1.86 0.66
1.87 0.57
1.88 0.62
1.89 0.61
1.9 0.64
1.91 0.57
1.92 0.64
1.93 0.66
1.94 0.53
1.95 0.66
1.96 0.63
1.97 0.56
1.98 0.66
1.99 0.6
2 0.64
};
\addplot [semithick, color6, opacity=0.75, dashed]
table {%
0.01 0.44
0.02 0.47
0.03 0.49
0.04 0.54
0.05 0.49
0.06 0.47
0.07 0.46
0.08 0.59
0.09 0.53
0.1 0.5
0.11 0.52
0.12 0.59
0.13 0.54
0.14 0.51
0.15 0.62
0.16 0.65
0.17 0.63
0.18 0.64
0.19 0.61
0.2 0.59
0.21 0.63
0.22 0.67
0.23 0.71
0.24 0.69
0.25 0.66
0.26 0.74
0.27 0.66
0.28 0.63
0.29 0.72
0.3 0.74
0.31 0.7
0.32 0.77
0.33 0.74
0.34 0.68
0.35 0.78
0.36 0.74
0.37 0.77
0.38 0.75
0.39 0.73
0.4 0.8
0.41 0.77
0.42 0.86
0.43 0.76
0.44 0.81
0.45 0.84
0.46 0.88
0.47 0.8
0.48 0.83
0.49 0.83
0.5 0.86
0.51 0.79
0.52 0.82
0.53 0.93
0.54 0.88
0.55 0.86
0.56 0.9
0.57 0.9
0.58 0.82
0.59 0.91
0.6 0.85
0.61 0.88
0.62 0.89
0.63 0.85
0.64 0.84
0.65 0.84
0.66 0.87
0.67 0.9
0.68 0.89
0.69 0.89
0.7 0.92
0.71 0.9
0.72 0.91
0.73 0.94
0.74 0.91
0.75 0.87
0.76 0.91
0.77 0.91
0.78 0.88
0.79 0.89
0.8 0.93
0.81 0.94
0.82 0.92
0.83 0.92
0.84 0.94
0.85 0.96
0.86 0.93
0.87 0.92
0.88 0.96
0.89 0.89
0.9 0.92
0.91 0.93
0.92 0.89
0.93 0.98
0.94 0.95
0.95 0.95
0.96 0.92
0.97 0.94
0.98 0.94
0.99 0.94
1 0.95
1.02 0.95
1.03 0.91
1.04 0.91
1.05 0.88
1.06 0.8
1.07 0.9
1.08 0.75
1.09 0.68
1.1 0.76
1.11 0.68
1.12 0.65
1.13 0.72
1.14 0.64
1.15 0.7
1.16 0.65
1.17 0.61
1.18 0.63
1.19 0.66
1.2 0.53
1.21 0.6
1.22 0.55
1.23 0.57
1.24 0.63
1.25 0.54
1.26 0.57
1.27 0.44
1.28 0.58
1.29 0.58
1.3 0.61
1.31 0.62
1.32 0.52
1.33 0.57
1.34 0.49
1.35 0.54
1.36 0.61
1.37 0.62
1.38 0.49
1.39 0.54
1.4 0.47
1.41 0.51
1.42 0.44
1.43 0.57
1.44 0.53
1.45 0.58
1.46 0.5
1.47 0.41
1.48 0.51
1.49 0.47
1.5 0.49
1.51 0.54
1.52 0.41
1.53 0.51
1.54 0.47
1.55 0.48
1.56 0.49
1.57 0.47
1.58 0.59
1.59 0.43
1.6 0.5
1.61 0.46
1.62 0.57
1.63 0.56
1.64 0.51
1.65 0.41
1.66 0.49
1.67 0.5
1.68 0.5
1.69 0.46
1.7 0.57
1.71 0.49
1.72 0.5
1.73 0.48
1.74 0.4
1.75 0.57
1.76 0.51
1.77 0.53
1.78 0.56
1.79 0.47
1.8 0.43
1.81 0.48
1.82 0.46
1.83 0.5
1.84 0.53
1.85 0.52
1.86 0.48
1.87 0.53
1.88 0.49
1.89 0.47
1.9 0.44
1.91 0.54
1.92 0.49
1.93 0.45
1.94 0.49
1.95 0.54
1.96 0.46
1.97 0.51
1.98 0.51
1.99 0.43
2 0.52
};
\addplot [semithick, color7, opacity=0.75, dashed]
table {%
0.01 0.44
0.02 0.47
0.03 0.49
0.04 0.54
0.05 0.49
0.06 0.47
0.07 0.46
0.08 0.59
0.09 0.53
0.1 0.5
0.11 0.52
0.12 0.59
0.13 0.54
0.14 0.51
0.15 0.62
0.16 0.65
0.17 0.63
0.18 0.64
0.19 0.61
0.2 0.59
0.21 0.63
0.22 0.67
0.23 0.71
0.24 0.69
0.25 0.66
0.26 0.74
0.27 0.66
0.28 0.63
0.29 0.72
0.3 0.74
0.31 0.7
0.32 0.77
0.33 0.74
0.34 0.68
0.35 0.78
0.36 0.74
0.37 0.77
0.38 0.75
0.39 0.73
0.4 0.8
0.41 0.77
0.42 0.86
0.43 0.76
0.44 0.81
0.45 0.84
0.46 0.88
0.47 0.8
0.48 0.83
0.49 0.83
0.5 0.86
0.51 0.79
0.52 0.82
0.53 0.93
0.54 0.88
0.55 0.86
0.56 0.9
0.57 0.9
0.58 0.82
0.59 0.91
0.6 0.85
0.61 0.88
0.62 0.89
0.63 0.85
0.64 0.84
0.65 0.84
0.66 0.87
0.67 0.9
0.68 0.89
0.69 0.89
0.7 0.92
0.71 0.9
0.72 0.91
0.73 0.94
0.74 0.91
0.75 0.87
0.76 0.91
0.77 0.91
0.78 0.88
0.79 0.89
0.8 0.93
0.81 0.94
0.82 0.92
0.83 0.92
0.84 0.94
0.85 0.96
0.86 0.93
0.87 0.92
0.88 0.96
0.89 0.89
0.9 0.92
0.91 0.93
0.92 0.89
0.93 0.98
0.94 0.95
0.95 0.95
0.96 0.92
0.97 0.94
0.98 0.94
0.99 0.94
1 0.95
1.02 0.95
1.03 0.91
1.04 0.91
1.05 0.88
1.06 0.8
1.07 0.9
1.08 0.75
1.09 0.68
1.1 0.76
1.11 0.68
1.12 0.65
1.13 0.72
1.14 0.64
1.15 0.7
1.16 0.65
1.17 0.61
1.18 0.63
1.19 0.66
1.2 0.53
1.21 0.6
1.22 0.55
1.23 0.57
1.24 0.63
1.25 0.54
1.26 0.57
1.27 0.44
1.28 0.58
1.29 0.58
1.3 0.61
1.31 0.62
1.32 0.52
1.33 0.57
1.34 0.49
1.35 0.54
1.36 0.61
1.37 0.62
1.38 0.49
1.39 0.54
1.4 0.47
1.41 0.51
1.42 0.44
1.43 0.57
1.44 0.53
1.45 0.58
1.46 0.5
1.47 0.41
1.48 0.51
1.49 0.47
1.5 0.49
1.51 0.54
1.52 0.41
1.53 0.51
1.54 0.47
1.55 0.48
1.56 0.49
1.57 0.47
1.58 0.59
1.59 0.43
1.6 0.5
1.61 0.46
1.62 0.57
1.63 0.56
1.64 0.51
1.65 0.41
1.66 0.49
1.67 0.5
1.68 0.5
1.69 0.46
1.7 0.57
1.71 0.49
1.72 0.5
1.73 0.48
1.74 0.4
1.75 0.57
1.76 0.51
1.77 0.53
1.78 0.56
1.79 0.47
1.8 0.43
1.81 0.48
1.82 0.46
1.83 0.5
1.84 0.53
1.85 0.52
1.86 0.48
1.87 0.53
1.88 0.49
1.89 0.47
1.9 0.44
1.91 0.54
1.92 0.49
1.93 0.45
1.94 0.49
1.95 0.54
1.96 0.46
1.97 0.51
1.98 0.51
1.99 0.43
2 0.52
};
\addplot [semithick, color8, opacity=0.75, dashed]
table {%
0.01 0.41
0.02 0.5
0.03 0.54
0.04 0.47
0.05 0.49
0.06 0.49
0.07 0.54
0.08 0.63
0.09 0.54
0.1 0.61
0.11 0.55
0.12 0.57
0.13 0.66
0.14 0.61
0.15 0.56
0.16 0.61
0.17 0.63
0.18 0.68
0.19 0.67
0.2 0.68
0.21 0.66
0.22 0.72
0.23 0.73
0.24 0.73
0.25 0.65
0.26 0.76
0.27 0.69
0.28 0.64
0.29 0.82
0.3 0.78
0.31 0.83
0.32 0.81
0.33 0.72
0.34 0.78
0.35 0.75
0.36 0.78
0.37 0.8
0.38 0.85
0.39 0.83
0.4 0.83
0.41 0.84
0.42 0.84
0.43 0.83
0.44 0.87
0.45 0.89
0.46 0.89
0.47 0.86
0.48 0.86
0.49 0.88
0.5 0.82
0.51 0.89
0.52 0.92
0.53 0.87
0.54 0.94
0.55 0.85
0.56 0.92
0.57 0.88
0.58 0.9
0.59 0.96
0.6 0.9
0.61 0.94
0.62 0.93
0.63 0.94
0.64 0.92
0.65 0.91
0.66 0.95
0.67 0.96
0.68 0.88
0.69 0.91
0.7 0.92
0.71 0.92
0.72 0.96
0.73 0.95
0.74 0.94
0.75 0.96
0.76 0.97
0.77 0.96
0.78 0.96
0.79 0.98
0.8 0.97
0.81 0.96
0.82 0.97
0.83 0.96
0.84 0.93
0.85 0.94
0.86 0.97
0.87 0.97
0.88 0.98
0.89 0.95
0.9 0.95
0.91 0.97
0.92 0.94
0.93 0.97
0.94 0.97
0.95 0.98
0.96 0.97
0.97 0.98
0.98 0.98
0.99 0.99
1 0.97
1.02 0.99
1.03 0.97
1.04 0.95
1.05 0.93
1.06 0.84
1.07 0.86
1.08 0.8
1.09 0.79
1.1 0.79
1.11 0.82
1.12 0.72
1.13 0.66
1.14 0.64
1.15 0.69
1.16 0.68
1.17 0.65
1.18 0.6
1.19 0.51
1.2 0.5
1.21 0.6
1.22 0.6
1.23 0.6
1.24 0.64
1.25 0.47
1.26 0.6
1.27 0.49
1.28 0.58
1.29 0.6
1.3 0.6
1.31 0.6
1.32 0.57
1.33 0.58
1.34 0.52
1.35 0.55
1.36 0.53
1.37 0.56
1.38 0.55
1.39 0.58
1.4 0.48
1.41 0.57
1.42 0.52
1.43 0.49
1.44 0.58
1.45 0.53
1.46 0.5
1.47 0.5
1.48 0.57
1.49 0.59
1.5 0.52
1.51 0.51
1.52 0.38
1.53 0.51
1.54 0.43
1.55 0.49
1.56 0.52
1.57 0.47
1.58 0.54
1.59 0.45
1.6 0.56
1.61 0.44
1.62 0.61
1.63 0.44
1.64 0.47
1.65 0.49
1.66 0.48
1.67 0.52
1.68 0.5
1.69 0.46
1.7 0.64
1.71 0.47
1.72 0.43
1.73 0.48
1.74 0.51
1.75 0.51
1.76 0.41
1.77 0.44
1.78 0.52
1.79 0.47
1.8 0.48
1.81 0.47
1.82 0.5
1.83 0.54
1.84 0.52
1.85 0.5
1.86 0.49
1.87 0.45
1.88 0.5
1.89 0.54
1.9 0.47
1.91 0.52
1.92 0.47
1.93 0.5
1.94 0.57
1.95 0.52
1.96 0.48
1.97 0.44
1.98 0.49
1.99 0.45
2 0.41
};
\addplot [semithick, color9, opacity=0.75, dashed]
table {%
0.01 0.51
0.02 0.55
0.03 0.57
0.04 0.56
0.05 0.51
0.06 0.62
0.07 0.52
0.08 0.68
0.09 0.52
0.1 0.62
0.11 0.62
0.12 0.74
0.13 0.64
0.14 0.74
0.15 0.77
0.16 0.77
0.17 0.66
0.18 0.77
0.19 0.85
0.2 0.77
0.21 0.75
0.22 0.81
0.23 0.81
0.24 0.87
0.25 0.88
0.26 0.86
0.27 0.81
0.28 0.83
0.29 0.87
0.3 0.89
0.31 0.87
0.32 0.92
0.33 0.93
0.34 0.9
0.35 0.93
0.36 0.88
0.37 0.94
0.38 0.91
0.39 0.95
0.4 0.97
0.41 0.93
0.42 0.94
0.43 0.96
0.44 0.98
0.45 0.94
0.46 0.91
0.47 0.97
0.48 0.96
0.49 0.98
0.5 1
0.51 0.96
0.52 0.99
0.53 0.94
0.54 0.98
0.55 0.98
0.56 0.98
0.57 0.99
0.58 0.97
0.59 1
0.6 0.96
0.61 1
0.62 1
0.63 0.99
0.64 0.99
0.65 1
0.66 1
0.67 1
0.68 0.99
0.69 1
0.7 1
0.71 1
0.72 1
0.73 0.99
0.74 1
0.75 1
0.76 1
0.77 1
0.78 1
0.79 1
0.8 1
0.81 1
0.82 1
0.83 1
0.84 1
0.85 1
0.86 1
0.87 1
0.88 0.99
0.89 1
0.9 1
0.91 1
0.92 1
0.93 0.99
0.94 0.99
0.95 1
0.96 1
0.97 1
0.98 1
0.99 1
1 1
1.02 1
1.03 1
1.04 1
1.05 0.98
1.06 1
1.07 0.96
1.08 0.93
1.09 0.9
1.1 0.9
1.11 0.88
1.12 0.81
1.13 0.84
1.14 0.83
1.15 0.79
1.16 0.82
1.17 0.77
1.18 0.75
1.19 0.67
1.2 0.74
1.21 0.74
1.22 0.71
1.23 0.74
1.24 0.63
1.25 0.68
1.26 0.67
1.27 0.67
1.28 0.72
1.29 0.78
1.3 0.72
1.31 0.7
1.32 0.69
1.33 0.66
1.34 0.7
1.35 0.65
1.36 0.56
1.37 0.56
1.38 0.6
1.39 0.71
1.4 0.66
1.41 0.6
1.42 0.54
1.43 0.61
1.44 0.59
1.45 0.6
1.46 0.53
1.47 0.62
1.48 0.61
1.49 0.55
1.5 0.63
1.51 0.68
1.52 0.58
1.53 0.56
1.54 0.62
1.55 0.55
1.56 0.57
1.57 0.56
1.58 0.5
1.59 0.66
1.6 0.58
1.61 0.64
1.62 0.59
1.63 0.59
1.64 0.49
1.65 0.51
1.66 0.58
1.67 0.54
1.68 0.63
1.69 0.58
1.7 0.58
1.71 0.51
1.72 0.53
1.73 0.51
1.74 0.56
1.75 0.66
1.76 0.54
1.77 0.5
1.78 0.57
1.79 0.49
1.8 0.52
1.81 0.62
1.82 0.58
1.83 0.51
1.84 0.6
1.85 0.51
1.86 0.61
1.87 0.49
1.88 0.52
1.89 0.52
1.9 0.55
1.91 0.62
1.92 0.54
1.93 0.65
1.94 0.56
1.95 0.57
1.96 0.59
1.97 0.55
1.98 0.59
1.99 0.63
2 0.59
};
\addplot [semithick, color10, opacity=0.75, dashed]
table {%
0.01 0.49
0.02 0.58
0.03 0.56
0.04 0.5
0.05 0.57
0.06 0.55
0.07 0.53
0.08 0.67
0.09 0.61
0.1 0.67
0.11 0.71
0.12 0.78
0.13 0.68
0.14 0.75
0.15 0.78
0.16 0.83
0.17 0.74
0.18 0.79
0.19 0.83
0.2 0.76
0.21 0.76
0.22 0.87
0.23 0.84
0.24 0.83
0.25 0.86
0.26 0.85
0.27 0.81
0.28 0.87
0.29 0.91
0.3 0.91
0.31 0.88
0.32 0.93
0.33 0.89
0.34 0.89
0.35 0.92
0.36 0.93
0.37 0.99
0.38 0.92
0.39 0.96
0.4 0.95
0.41 0.95
0.42 0.99
0.43 0.96
0.44 0.97
0.45 0.95
0.46 0.96
0.47 0.99
0.48 0.97
0.49 0.98
0.5 1
0.51 1
0.52 1
0.53 0.97
0.54 0.99
0.55 1
0.56 0.99
0.57 1
0.58 0.99
0.59 1
0.6 0.99
0.61 1
0.62 1
0.63 0.99
0.64 0.99
0.65 1
0.66 1
0.67 0.99
0.68 0.99
0.69 1
0.7 1
0.71 1
0.72 1
0.73 0.99
0.74 1
0.75 1
0.76 1
0.77 1
0.78 0.99
0.79 1
0.8 1
0.81 1
0.82 1
0.83 1
0.84 1
0.85 1
0.86 1
0.87 1
0.88 1
0.89 1
0.9 0.99
0.91 1
0.92 1
0.93 1
0.94 1
0.95 1
0.96 1
0.97 1
0.98 1
0.99 1
1 1
1.02 1
1.03 1
1.04 1
1.05 0.99
1.06 0.99
1.07 0.98
1.08 0.94
1.09 0.96
1.1 0.9
1.11 0.87
1.12 0.86
1.13 0.86
1.14 0.87
1.15 0.84
1.16 0.85
1.17 0.8
1.18 0.76
1.19 0.74
1.2 0.74
1.21 0.78
1.22 0.71
1.23 0.75
1.24 0.71
1.25 0.65
1.26 0.72
1.27 0.66
1.28 0.63
1.29 0.7
1.3 0.76
1.31 0.67
1.32 0.61
1.33 0.59
1.34 0.69
1.35 0.65
1.36 0.55
1.37 0.65
1.38 0.6
1.39 0.68
1.4 0.7
1.41 0.67
1.42 0.52
1.43 0.67
1.44 0.59
1.45 0.64
1.46 0.53
1.47 0.62
1.48 0.61
1.49 0.56
1.5 0.61
1.51 0.59
1.52 0.52
1.53 0.51
1.54 0.65
1.55 0.56
1.56 0.52
1.57 0.5
1.58 0.5
1.59 0.57
1.6 0.6
1.61 0.57
1.62 0.44
1.63 0.6
1.64 0.53
1.65 0.47
1.66 0.59
1.67 0.6
1.68 0.5
1.69 0.46
1.7 0.6
1.71 0.58
1.72 0.58
1.73 0.52
1.74 0.57
1.75 0.56
1.76 0.49
1.77 0.52
1.78 0.47
1.79 0.51
1.8 0.55
1.81 0.61
1.82 0.52
1.83 0.51
1.84 0.58
1.85 0.57
1.86 0.61
1.87 0.54
1.88 0.53
1.89 0.53
1.9 0.53
1.91 0.53
1.92 0.53
1.93 0.59
1.94 0.48
1.95 0.51
1.96 0.61
1.97 0.48
1.98 0.56
1.99 0.52
2 0.56
};
\addplot [semithick, color11, opacity=0.75, dashed]
table {%
0.01 0.52
0.02 0.6
0.03 0.6
0.04 0.61
0.05 0.56
0.06 0.64
0.07 0.68
0.08 0.76
0.09 0.75
0.1 0.86
0.11 0.84
0.12 0.83
0.13 0.8
0.14 0.89
0.15 0.82
0.16 0.91
0.17 0.87
0.18 0.88
0.19 0.88
0.2 0.91
0.21 0.93
0.22 0.94
0.23 0.93
0.24 0.97
0.25 0.95
0.26 0.97
0.27 0.95
0.28 0.97
0.29 0.99
0.3 0.99
0.31 0.98
0.32 0.98
0.33 0.98
0.34 0.97
0.35 1
0.36 0.99
0.37 0.99
0.38 1
0.39 1
0.4 0.99
0.41 1
0.42 1
0.43 0.99
0.44 1
0.45 0.98
0.46 1
0.47 0.99
0.48 1
0.49 1
0.5 1
0.51 1
0.52 1
0.53 1
0.54 1
0.55 1
0.56 1
0.57 1
0.58 1
0.59 1
0.6 1
0.61 1
0.62 1
0.63 1
0.64 1
0.65 1
0.66 1
0.67 1
0.68 1
0.69 1
0.7 1
0.71 1
0.72 1
0.73 1
0.74 1
0.75 1
0.76 1
0.77 1
0.78 1
0.79 1
0.8 1
0.81 1
0.82 1
0.83 1
0.84 1
0.85 1
0.86 1
0.87 1
0.88 1
0.89 1
0.9 1
0.91 1
0.92 1
0.93 1
0.94 1
0.95 1
0.96 1
0.97 1
0.98 1
0.99 1
1 1
1.02 1
1.03 1
1.04 1
1.05 1
1.06 1
1.07 1
1.08 0.99
1.09 1
1.1 0.98
1.11 0.99
1.12 1
1.13 0.99
1.14 0.96
1.15 0.99
1.16 0.98
1.17 0.95
1.18 0.98
1.19 0.94
1.2 0.93
1.21 0.95
1.22 0.89
1.23 0.88
1.24 0.87
1.25 0.86
1.26 0.88
1.27 0.9
1.28 0.84
1.29 0.8
1.3 0.77
1.31 0.81
1.32 0.78
1.33 0.74
1.34 0.8
1.35 0.76
1.36 0.75
1.37 0.78
1.38 0.74
1.39 0.73
1.4 0.74
1.41 0.73
1.42 0.61
1.43 0.66
1.44 0.66
1.45 0.64
1.46 0.6
1.47 0.67
1.48 0.65
1.49 0.6
1.5 0.64
1.51 0.57
1.52 0.56
1.53 0.63
1.54 0.6
1.55 0.56
1.56 0.5
1.57 0.55
1.58 0.6
1.59 0.53
1.6 0.59
1.61 0.55
1.62 0.5
1.63 0.56
1.64 0.55
1.65 0.52
1.66 0.57
1.67 0.49
1.68 0.46
1.69 0.52
1.7 0.51
1.71 0.44
1.72 0.39
1.73 0.48
1.74 0.47
1.75 0.48
1.76 0.42
1.77 0.39
1.78 0.43
1.79 0.48
1.8 0.46
1.81 0.49
1.82 0.46
1.83 0.34
1.84 0.44
1.85 0.41
1.86 0.5
1.87 0.36
1.88 0.44
1.89 0.43
1.9 0.48
1.91 0.39
1.92 0.44
1.93 0.51
1.94 0.44
1.95 0.42
1.96 0.36
1.97 0.36
1.98 0.49
1.99 0.4
2 0.43
};
\addplot [semithick, black, opacity=1, dash pattern=on 1pt off 1pt]
table {%
-0.0895000000000001 0.5
2.0995 0.5
};
\addplot [semithick, black, opacity=1, dash pattern=on 1pt off 1pt]
table {%
-0.0895000000000001 1
2.0995 1
};
\addplot [semithick, black, opacity=1, dash pattern=on 1pt off 1pt]
table {%
1 0.2
1 1.05
};
\end{axis}

\end{tikzpicture}

%% file: plots/decoupled2/LAP.tex
% This file was created by tikzplotlib v0.9.6.
\begin{tikzpicture}

\definecolor{color0}{rgb}{0.866666666666667,0.494117647058824,0.164705882352941}
\definecolor{color1}{rgb}{0.164705882352941,0.643137254901961,0.866666666666667}
\definecolor{color2}{rgb}{0.584313725490196,0.866666666666667,0.164705882352941}
\definecolor{color3}{rgb}{0.109803921568627,0.337254901960784,0.129411764705882}
\definecolor{color4}{rgb}{0.529411764705882,0.305882352941176,0.858823529411765}
\definecolor{color5}{rgb}{0.858823529411765,0.305882352941176,0.435294117647059}
\definecolor{color6}{rgb}{0.937254901960784,0.929411764705882,0.392156862745098}
\definecolor{color7}{rgb}{0.0901960784313725,0.486274509803922,0.0980392156862745}
\definecolor{color8}{rgb}{0.156862745098039,0.188235294117647,0.827450980392157}
\definecolor{color9}{rgb}{0.937254901960784,0.392156862745098,0.894117647058824}
\definecolor{color10}{rgb}{0.2,0.184313725490196,0.184313725490196}
\definecolor{color11}{rgb}{0.0156862745098039,0.803921568627451,0.976470588235294}

\begin{axis}[
tick align=outside,
tick pos=left,
x grid style={white!69.0196078431373!black},
xmajorgrids,
xmin=-0.0895, xmax=2.0995,
xtick style={color=black},
xtick={0,0.1,0.2,0.3,0.4,0.5,0.6,0.7,0.8,0.9,1,1.1,1.2,1.3,1.4,1.5,1.6,1.7,1.8,1.9,2},
xticklabels={0,,.2,,.4,,.6,,.8,,1,,20,,40,,60,,80,,100},
height=4.8cm,
width=6.5cm,
y grid style={white!69.0196078431373!black},
ymajorgrids,
ymin=0.2, ymax=1.05,
ytick style={color=black}
]
\addplot [semithick, color0, opacity=0.75]
table {%
0.01 0.54
0.02 0.5
0.03 0.48
0.04 0.6
0.05 0.51
0.06 0.58
0.07 0.55
0.08 0.7
0.09 0.73
0.1 0.62
0.11 0.71
0.12 0.76
0.13 0.72
0.14 0.78
0.15 0.73
0.16 0.77
0.17 0.81
0.18 0.88
0.19 0.82
0.2 0.86
0.21 0.88
0.22 0.8
0.23 0.83
0.24 0.93
0.25 0.86
0.26 0.93
0.27 0.89
0.28 0.91
0.29 0.87
0.3 0.91
0.31 0.9
0.32 0.93
0.33 0.95
0.34 0.95
0.35 0.94
0.36 0.93
0.37 0.94
0.38 0.93
0.39 0.98
0.4 0.98
0.41 0.99
0.42 0.99
0.43 0.98
0.44 0.97
0.45 0.98
0.46 0.98
0.47 0.99
0.48 0.99
0.49 0.99
0.5 0.99
0.51 0.99
0.52 0.97
0.53 0.99
0.54 0.99
0.55 0.97
0.56 0.99
0.57 0.99
0.58 0.98
0.59 1
0.6 1
0.61 1
0.62 0.98
0.63 0.99
0.64 0.99
0.65 0.97
0.66 1
0.67 0.99
0.68 1
0.69 0.99
0.7 1
0.71 1
0.72 1
0.73 0.99
0.74 0.99
0.75 0.99
0.76 0.99
0.77 1
0.78 1
0.79 0.99
0.8 1
0.81 1
0.82 0.99
0.83 1
0.84 0.99
0.85 0.97
0.86 0.99
0.87 1
0.88 1
0.89 1
0.9 1
0.91 1
0.92 0.99
0.93 0.99
0.94 1
0.95 1
0.96 0.98
0.97 1
0.98 1
0.99 0.99
1 0.99
1.02 0.99
1.03 0.95
1.04 0.9
1.05 0.86
1.06 0.79
1.07 0.77
1.08 0.79
1.09 0.68
1.1 0.66
1.11 0.65
1.12 0.66
1.13 0.52
1.14 0.6
1.15 0.59
1.16 0.67
1.17 0.53
1.18 0.62
1.19 0.5
1.2 0.59
1.21 0.51
1.22 0.53
1.23 0.61
1.24 0.55
1.25 0.7
1.26 0.55
1.27 0.55
1.28 0.49
1.29 0.54
1.3 0.53
1.31 0.55
1.32 0.59
1.33 0.41
1.34 0.48
1.35 0.44
1.36 0.6
1.37 0.46
1.38 0.53
1.39 0.45
1.4 0.6
1.41 0.59
1.42 0.56
1.43 0.45
1.44 0.56
1.45 0.62
1.46 0.47
1.47 0.55
1.48 0.46
1.49 0.41
1.5 0.54
1.51 0.55
1.52 0.55
1.53 0.47
1.54 0.45
1.55 0.47
1.56 0.54
1.57 0.54
1.58 0.59
1.59 0.59
1.6 0.5
1.61 0.53
1.62 0.52
1.63 0.54
1.64 0.44
1.65 0.51
1.66 0.57
1.67 0.51
1.68 0.54
1.69 0.56
1.7 0.53
1.71 0.53
1.72 0.46
1.73 0.47
1.74 0.54
1.75 0.48
1.76 0.64
1.77 0.51
1.78 0.48
1.79 0.55
1.8 0.48
1.81 0.49
1.82 0.5
1.83 0.49
1.84 0.49
1.85 0.5
1.86 0.54
1.87 0.49
1.88 0.48
1.89 0.45
1.9 0.41
1.91 0.53
1.92 0.49
1.93 0.49
1.94 0.43
1.95 0.5
1.96 0.56
1.97 0.53
1.98 0.57
1.99 0.51
2 0.47
};
\addplot [semithick, color1, opacity=0.75]
table {%
0.01 0.49
0.02 0.56
0.03 0.48
0.04 0.58
0.05 0.58
0.06 0.71
0.07 0.53
0.08 0.58
0.09 0.54
0.1 0.63
0.11 0.62
0.12 0.65
0.13 0.57
0.14 0.72
0.15 0.64
0.16 0.7
0.17 0.73
0.18 0.74
0.19 0.77
0.2 0.7
0.21 0.84
0.22 0.86
0.23 0.84
0.24 0.79
0.25 0.79
0.26 0.84
0.27 0.86
0.28 0.81
0.29 0.82
0.3 0.83
0.31 0.78
0.32 0.84
0.33 0.83
0.34 0.87
0.35 0.87
0.36 0.8
0.37 0.85
0.38 0.87
0.39 0.86
0.4 0.84
0.41 0.9
0.42 0.87
0.43 0.91
0.44 0.95
0.45 0.91
0.46 0.91
0.47 0.91
0.48 0.9
0.49 0.94
0.5 0.97
0.51 0.93
0.52 0.93
0.53 0.94
0.54 0.93
0.55 0.96
0.56 0.93
0.57 0.96
0.58 0.94
0.59 0.97
0.6 0.96
0.61 0.97
0.62 0.93
0.63 0.97
0.64 0.93
0.65 0.96
0.66 0.95
0.67 0.91
0.68 0.96
0.69 0.97
0.7 0.97
0.71 0.97
0.72 0.96
0.73 0.95
0.74 0.97
0.75 0.99
0.76 0.98
0.77 0.98
0.78 0.98
0.79 0.94
0.8 0.99
0.81 0.97
0.82 0.99
0.83 0.98
0.84 0.98
0.85 0.95
0.86 0.99
0.87 0.97
0.88 0.98
0.89 1
0.9 1
0.91 0.97
0.92 1
0.93 0.98
0.94 0.99
0.95 0.99
0.96 0.99
0.97 1
0.98 1
0.99 0.98
1 0.98
1.02 0.96
1.03 0.92
1.04 0.83
1.05 0.75
1.06 0.71
1.07 0.57
1.08 0.69
1.09 0.63
1.1 0.61
1.11 0.67
1.12 0.58
1.13 0.53
1.14 0.62
1.15 0.48
1.16 0.56
1.17 0.45
1.18 0.62
1.19 0.55
1.2 0.59
1.21 0.49
1.22 0.51
1.23 0.5
1.24 0.51
1.25 0.54
1.26 0.47
1.27 0.48
1.28 0.5
1.29 0.48
1.3 0.44
1.31 0.53
1.32 0.43
1.33 0.55
1.34 0.45
1.35 0.45
1.36 0.46
1.37 0.55
1.38 0.46
1.39 0.51
1.4 0.45
1.41 0.51
1.42 0.52
1.43 0.53
1.44 0.51
1.45 0.45
1.46 0.67
1.47 0.41
1.48 0.44
1.49 0.53
1.5 0.56
1.51 0.5
1.52 0.46
1.53 0.43
1.54 0.49
1.55 0.5
1.56 0.48
1.57 0.48
1.58 0.54
1.59 0.53
1.6 0.54
1.61 0.55
1.62 0.53
1.63 0.57
1.64 0.5
1.65 0.55
1.66 0.55
1.67 0.45
1.68 0.36
1.69 0.62
1.7 0.6
1.71 0.46
1.72 0.51
1.73 0.47
1.74 0.55
1.75 0.55
1.76 0.45
1.77 0.56
1.78 0.52
1.79 0.51
1.8 0.54
1.81 0.47
1.82 0.5
1.83 0.47
1.84 0.43
1.85 0.4
1.86 0.44
1.87 0.57
1.88 0.5
1.89 0.42
1.9 0.44
1.91 0.37
1.92 0.54
1.93 0.51
1.94 0.46
1.95 0.53
1.96 0.54
1.97 0.53
1.98 0.58
1.99 0.45
2 0.53
};
\addplot [semithick, color2, opacity=0.75]
table {%
0.01 0.39
0.02 0.53
0.03 0.52
0.04 0.6
0.05 0.52
0.06 0.63
0.07 0.53
0.08 0.64
0.09 0.59
0.1 0.62
0.11 0.61
0.12 0.63
0.13 0.56
0.14 0.69
0.15 0.66
0.16 0.72
0.17 0.7
0.18 0.76
0.19 0.75
0.2 0.7
0.21 0.83
0.22 0.86
0.23 0.82
0.24 0.79
0.25 0.78
0.26 0.83
0.27 0.85
0.28 0.8
0.29 0.82
0.3 0.83
0.31 0.78
0.32 0.84
0.33 0.82
0.34 0.87
0.35 0.88
0.36 0.8
0.37 0.86
0.38 0.87
0.39 0.87
0.4 0.84
0.41 0.91
0.42 0.87
0.43 0.91
0.44 0.94
0.45 0.9
0.46 0.91
0.47 0.9
0.48 0.9
0.49 0.93
0.5 0.96
0.51 0.91
0.52 0.93
0.53 0.93
0.54 0.93
0.55 0.96
0.56 0.93
0.57 0.96
0.58 0.94
0.59 0.97
0.6 0.96
0.61 0.97
0.62 0.93
0.63 0.96
0.64 0.93
0.65 0.96
0.66 0.95
0.67 0.91
0.68 0.95
0.69 0.97
0.7 0.97
0.71 0.97
0.72 0.96
0.73 0.95
0.74 0.97
0.75 0.99
0.76 0.98
0.77 0.97
0.78 0.97
0.79 0.94
0.8 0.99
0.81 0.97
0.82 0.99
0.83 0.97
0.84 0.98
0.85 0.95
0.86 0.99
0.87 0.97
0.88 0.98
0.89 1
0.9 1
0.91 0.97
0.92 0.99
0.93 0.98
0.94 0.99
0.95 1
0.96 0.99
0.97 1
0.98 1
0.99 0.98
1 0.97
1.02 0.96
1.03 0.92
1.04 0.83
1.05 0.76
1.06 0.72
1.07 0.56
1.08 0.68
1.09 0.65
1.1 0.6
1.11 0.67
1.12 0.58
1.13 0.54
1.14 0.61
1.15 0.46
1.16 0.55
1.17 0.46
1.18 0.6
1.19 0.55
1.2 0.6
1.21 0.5
1.22 0.51
1.23 0.48
1.24 0.5
1.25 0.53
1.26 0.46
1.27 0.48
1.28 0.49
1.29 0.5
1.3 0.42
1.31 0.54
1.32 0.41
1.33 0.54
1.34 0.48
1.35 0.45
1.36 0.46
1.37 0.53
1.38 0.45
1.39 0.5
1.4 0.47
1.41 0.48
1.42 0.51
1.43 0.52
1.44 0.51
1.45 0.45
1.46 0.65
1.47 0.43
1.48 0.46
1.49 0.52
1.5 0.55
1.51 0.48
1.52 0.48
1.53 0.42
1.54 0.49
1.55 0.5
1.56 0.5
1.57 0.5
1.58 0.53
1.59 0.54
1.6 0.54
1.61 0.59
1.62 0.54
1.63 0.58
1.64 0.55
1.65 0.54
1.66 0.54
1.67 0.46
1.68 0.37
1.69 0.62
1.7 0.6
1.71 0.44
1.72 0.52
1.73 0.48
1.74 0.54
1.75 0.56
1.76 0.46
1.77 0.56
1.78 0.51
1.79 0.54
1.8 0.53
1.81 0.51
1.82 0.51
1.83 0.48
1.84 0.41
1.85 0.41
1.86 0.43
1.87 0.55
1.88 0.47
1.89 0.41
1.9 0.47
1.91 0.37
1.92 0.55
1.93 0.48
1.94 0.44
1.95 0.57
1.96 0.48
1.97 0.55
1.98 0.59
1.99 0.43
2 0.53
};
\addplot [semithick, color3, opacity=0.75]
table {%
0.01 0.51
0.02 0.58
0.03 0.53
0.04 0.61
0.05 0.6
0.06 0.62
0.07 0.59
0.08 0.67
0.09 0.73
0.1 0.69
0.11 0.74
0.12 0.77
0.13 0.62
0.14 0.79
0.15 0.72
0.16 0.77
0.17 0.78
0.18 0.92
0.19 0.8
0.2 0.83
0.21 0.9
0.22 0.89
0.23 0.87
0.24 0.95
0.25 0.79
0.26 0.94
0.27 0.91
0.28 0.91
0.29 0.89
0.3 0.87
0.31 0.86
0.32 0.92
0.33 0.9
0.34 0.93
0.35 0.89
0.36 0.92
0.37 0.93
0.38 0.92
0.39 0.91
0.4 0.92
0.41 0.98
0.42 0.93
0.43 0.97
0.44 0.96
0.45 0.92
0.46 0.96
0.47 0.96
0.48 0.95
0.49 0.97
0.5 0.96
0.51 0.91
0.52 0.96
0.53 0.93
0.54 0.94
0.55 0.95
0.56 0.92
0.57 0.96
0.58 0.93
0.59 0.95
0.6 0.93
0.61 0.96
0.62 0.96
0.63 0.95
0.64 0.9
0.65 0.95
0.66 0.94
0.67 0.87
0.68 0.96
0.69 0.94
0.7 0.96
0.71 0.95
0.72 0.96
0.73 0.96
0.74 0.96
0.75 0.98
0.76 0.96
0.77 0.96
0.78 0.94
0.79 0.93
0.8 0.98
0.81 0.96
0.82 0.98
0.83 0.96
0.84 0.98
0.85 0.93
0.86 0.98
0.87 0.94
0.88 0.98
0.89 0.96
0.9 0.94
0.91 0.97
0.92 0.98
0.93 0.95
0.94 0.96
0.95 0.95
0.96 0.97
0.97 0.98
0.98 0.99
0.99 0.97
1 0.94
1.02 0.92
1.03 0.93
1.04 0.88
1.05 0.84
1.06 0.79
1.07 0.76
1.08 0.76
1.09 0.74
1.1 0.71
1.11 0.62
1.12 0.71
1.13 0.57
1.14 0.61
1.15 0.66
1.16 0.56
1.17 0.52
1.18 0.68
1.19 0.58
1.2 0.64
1.21 0.53
1.22 0.52
1.23 0.63
1.24 0.53
1.25 0.61
1.26 0.58
1.27 0.57
1.28 0.51
1.29 0.63
1.3 0.53
1.31 0.59
1.32 0.6
1.33 0.41
1.34 0.49
1.35 0.54
1.36 0.6
1.37 0.59
1.38 0.56
1.39 0.5
1.4 0.57
1.41 0.55
1.42 0.49
1.43 0.43
1.44 0.58
1.45 0.62
1.46 0.56
1.47 0.52
1.48 0.51
1.49 0.48
1.5 0.46
1.51 0.56
1.52 0.43
1.53 0.49
1.54 0.51
1.55 0.46
1.56 0.51
1.57 0.57
1.58 0.51
1.59 0.58
1.6 0.51
1.61 0.5
1.62 0.52
1.63 0.48
1.64 0.42
1.65 0.44
1.66 0.56
1.67 0.58
1.68 0.65
1.69 0.56
1.7 0.47
1.71 0.43
1.72 0.53
1.73 0.54
1.74 0.51
1.75 0.4
1.76 0.54
1.77 0.43
1.78 0.54
1.79 0.55
1.8 0.47
1.81 0.46
1.82 0.53
1.83 0.42
1.84 0.48
1.85 0.44
1.86 0.49
1.87 0.46
1.88 0.53
1.89 0.48
1.9 0.36
1.91 0.54
1.92 0.54
1.93 0.51
1.94 0.41
1.95 0.48
1.96 0.55
1.97 0.49
1.98 0.58
1.99 0.46
2 0.55
};
\addplot [semithick, color4, opacity=0.75]
table {%
0.01 0.51
0.02 0.57
0.03 0.53
0.04 0.61
0.05 0.59
0.06 0.63
0.07 0.6
0.08 0.66
0.09 0.72
0.1 0.69
0.11 0.73
0.12 0.77
0.13 0.62
0.14 0.77
0.15 0.71
0.16 0.76
0.17 0.77
0.18 0.91
0.19 0.78
0.2 0.83
0.21 0.87
0.22 0.89
0.23 0.84
0.24 0.95
0.25 0.77
0.26 0.89
0.27 0.91
0.28 0.89
0.29 0.87
0.3 0.86
0.31 0.83
0.32 0.89
0.33 0.9
0.34 0.92
0.35 0.88
0.36 0.89
0.37 0.91
0.38 0.91
0.39 0.89
0.4 0.91
0.41 0.97
0.42 0.93
0.43 0.94
0.44 0.95
0.45 0.92
0.46 0.94
0.47 0.96
0.48 0.95
0.49 0.96
0.5 0.96
0.51 0.9
0.52 0.95
0.53 0.93
0.54 0.94
0.55 0.94
0.56 0.89
0.57 0.94
0.58 0.92
0.59 0.95
0.6 0.88
0.61 0.95
0.62 0.93
0.63 0.94
0.64 0.88
0.65 0.94
0.66 0.93
0.67 0.86
0.68 0.96
0.69 0.92
0.7 0.93
0.71 0.92
0.72 0.94
0.73 0.96
0.74 0.92
0.75 0.97
0.76 0.93
0.77 0.96
0.78 0.93
0.79 0.91
0.8 0.95
0.81 0.95
0.82 0.96
0.83 0.94
0.84 0.94
0.85 0.91
0.86 0.95
0.87 0.92
0.88 0.96
0.89 0.93
0.9 0.93
0.91 0.96
0.92 0.96
0.93 0.93
0.94 0.94
0.95 0.92
0.96 0.97
0.97 0.97
0.98 0.96
0.99 0.95
1 0.91
1.02 0.91
1.03 0.92
1.04 0.85
1.05 0.82
1.06 0.78
1.07 0.75
1.08 0.76
1.09 0.74
1.1 0.71
1.11 0.6
1.12 0.72
1.13 0.56
1.14 0.61
1.15 0.66
1.16 0.58
1.17 0.52
1.18 0.66
1.19 0.58
1.2 0.64
1.21 0.53
1.22 0.52
1.23 0.62
1.24 0.53
1.25 0.6
1.26 0.57
1.27 0.56
1.28 0.5
1.29 0.63
1.3 0.52
1.31 0.6
1.32 0.6
1.33 0.38
1.34 0.47
1.35 0.54
1.36 0.59
1.37 0.58
1.38 0.53
1.39 0.49
1.4 0.56
1.41 0.55
1.42 0.5
1.43 0.43
1.44 0.57
1.45 0.62
1.46 0.55
1.47 0.52
1.48 0.52
1.49 0.49
1.5 0.47
1.51 0.56
1.52 0.43
1.53 0.49
1.54 0.52
1.55 0.45
1.56 0.49
1.57 0.57
1.58 0.52
1.59 0.57
1.6 0.51
1.61 0.5
1.62 0.54
1.63 0.47
1.64 0.42
1.65 0.44
1.66 0.56
1.67 0.58
1.68 0.65
1.69 0.56
1.7 0.47
1.71 0.43
1.72 0.54
1.73 0.54
1.74 0.52
1.75 0.4
1.76 0.53
1.77 0.43
1.78 0.56
1.79 0.55
1.8 0.48
1.81 0.45
1.82 0.55
1.83 0.44
1.84 0.48
1.85 0.44
1.86 0.49
1.87 0.46
1.88 0.53
1.89 0.48
1.9 0.37
1.91 0.53
1.92 0.54
1.93 0.51
1.94 0.42
1.95 0.49
1.96 0.55
1.97 0.5
1.98 0.58
1.99 0.45
2 0.53
};
\addplot [semithick, color5, opacity=0.75]
table {%
0.01 0.46
0.02 0.58
0.03 0.49
0.04 0.57
0.05 0.58
0.06 0.62
0.07 0.54
0.08 0.62
0.09 0.58
0.1 0.64
0.11 0.68
0.12 0.68
0.13 0.55
0.14 0.73
0.15 0.65
0.16 0.69
0.17 0.74
0.18 0.85
0.19 0.69
0.2 0.67
0.21 0.81
0.22 0.78
0.23 0.72
0.24 0.8
0.25 0.69
0.26 0.78
0.27 0.77
0.28 0.76
0.29 0.75
0.3 0.74
0.31 0.68
0.32 0.77
0.33 0.75
0.34 0.79
0.35 0.77
0.36 0.73
0.37 0.79
0.38 0.79
0.39 0.76
0.4 0.77
0.41 0.86
0.42 0.82
0.43 0.81
0.44 0.72
0.45 0.83
0.46 0.82
0.47 0.82
0.48 0.83
0.49 0.8
0.5 0.8
0.51 0.79
0.52 0.77
0.53 0.82
0.54 0.82
0.55 0.75
0.56 0.77
0.57 0.78
0.58 0.79
0.59 0.78
0.6 0.77
0.61 0.8
0.62 0.77
0.63 0.81
0.64 0.78
0.65 0.76
0.66 0.8
0.67 0.82
0.68 0.75
0.69 0.76
0.7 0.85
0.71 0.76
0.72 0.84
0.73 0.84
0.74 0.8
0.75 0.87
0.76 0.79
0.77 0.82
0.78 0.81
0.79 0.78
0.8 0.81
0.81 0.79
0.82 0.83
0.83 0.78
0.84 0.78
0.85 0.76
0.86 0.84
0.87 0.76
0.88 0.84
0.89 0.88
0.9 0.81
0.91 0.83
0.92 0.82
0.93 0.81
0.94 0.83
0.95 0.83
0.96 0.89
0.97 0.82
0.98 0.78
0.99 0.88
1 0.73
1.02 0.73
1.03 0.82
1.04 0.7
1.05 0.73
1.06 0.67
1.07 0.58
1.08 0.66
1.09 0.7
1.1 0.71
1.11 0.58
1.12 0.62
1.13 0.51
1.14 0.56
1.15 0.6
1.16 0.61
1.17 0.48
1.18 0.58
1.19 0.54
1.2 0.66
1.21 0.51
1.22 0.48
1.23 0.54
1.24 0.57
1.25 0.48
1.26 0.55
1.27 0.54
1.28 0.49
1.29 0.54
1.3 0.5
1.31 0.61
1.32 0.58
1.33 0.45
1.34 0.56
1.35 0.56
1.36 0.53
1.37 0.51
1.38 0.41
1.39 0.43
1.4 0.53
1.41 0.55
1.42 0.53
1.43 0.39
1.44 0.55
1.45 0.64
1.46 0.57
1.47 0.49
1.48 0.47
1.49 0.46
1.5 0.49
1.51 0.49
1.52 0.42
1.53 0.45
1.54 0.46
1.55 0.49
1.56 0.45
1.57 0.52
1.58 0.48
1.59 0.46
1.6 0.54
1.61 0.48
1.62 0.57
1.63 0.4
1.64 0.43
1.65 0.5
1.66 0.6
1.67 0.56
1.68 0.52
1.69 0.53
1.7 0.5
1.71 0.51
1.72 0.44
1.73 0.51
1.74 0.52
1.75 0.45
1.76 0.49
1.77 0.43
1.78 0.49
1.79 0.58
1.8 0.48
1.81 0.48
1.82 0.56
1.83 0.47
1.84 0.54
1.85 0.55
1.86 0.48
1.87 0.46
1.88 0.52
1.89 0.49
1.9 0.47
1.91 0.55
1.92 0.45
1.93 0.52
1.94 0.39
1.95 0.45
1.96 0.5
1.97 0.5
1.98 0.51
1.99 0.42
2 0.5
};
\addplot [semithick, color6, opacity=0.75, dashed]
table {%
0.01 0.51
0.02 0.63
0.03 0.55
0.04 0.56
0.05 0.49
0.06 0.53
0.07 0.51
0.08 0.46
0.09 0.51
0.1 0.52
0.11 0.57
0.12 0.51
0.13 0.5
0.14 0.57
0.15 0.59
0.16 0.57
0.17 0.51
0.18 0.56
0.19 0.62
0.2 0.6
0.21 0.62
0.22 0.59
0.23 0.61
0.24 0.62
0.25 0.62
0.26 0.64
0.27 0.55
0.28 0.6
0.29 0.64
0.3 0.69
0.31 0.63
0.32 0.75
0.33 0.63
0.34 0.71
0.35 0.68
0.36 0.71
0.37 0.67
0.38 0.75
0.39 0.72
0.4 0.68
0.41 0.83
0.42 0.73
0.43 0.71
0.44 0.69
0.45 0.68
0.46 0.71
0.47 0.76
0.48 0.78
0.49 0.73
0.5 0.74
0.51 0.75
0.52 0.72
0.53 0.74
0.54 0.76
0.55 0.73
0.56 0.75
0.57 0.79
0.58 0.78
0.59 0.76
0.6 0.74
0.61 0.78
0.62 0.78
0.63 0.77
0.64 0.81
0.65 0.69
0.66 0.81
0.67 0.76
0.68 0.8
0.69 0.69
0.7 0.82
0.71 0.86
0.72 0.81
0.73 0.75
0.74 0.86
0.75 0.81
0.76 0.75
0.77 0.81
0.78 0.84
0.79 0.78
0.8 0.81
0.81 0.71
0.82 0.79
0.83 0.75
0.84 0.83
0.85 0.84
0.86 0.87
0.87 0.8
0.88 0.83
0.89 0.86
0.9 0.79
0.91 0.89
0.92 0.76
0.93 0.82
0.94 0.8
0.95 0.83
0.96 0.8
0.97 0.77
0.98 0.8
0.99 0.79
1 0.86
1.02 0.76
1.03 0.73
1.04 0.6
1.05 0.54
1.06 0.51
1.07 0.52
1.08 0.48
1.09 0.52
1.1 0.46
1.11 0.52
1.12 0.55
1.13 0.48
1.14 0.51
1.15 0.55
1.16 0.51
1.17 0.56
1.18 0.58
1.19 0.47
1.2 0.36
1.21 0.49
1.22 0.5
1.23 0.5
1.24 0.57
1.25 0.56
1.26 0.48
1.27 0.49
1.28 0.45
1.29 0.57
1.3 0.59
1.31 0.51
1.32 0.51
1.33 0.49
1.34 0.45
1.35 0.44
1.36 0.53
1.37 0.5
1.38 0.49
1.39 0.59
1.4 0.54
1.41 0.62
1.42 0.48
1.43 0.48
1.44 0.49
1.45 0.45
1.46 0.58
1.47 0.52
1.48 0.41
1.49 0.54
1.5 0.55
1.51 0.61
1.52 0.46
1.53 0.43
1.54 0.5
1.55 0.48
1.56 0.5
1.57 0.51
1.58 0.49
1.59 0.51
1.6 0.51
1.61 0.5
1.62 0.55
1.63 0.58
1.64 0.56
1.65 0.48
1.66 0.53
1.67 0.45
1.68 0.52
1.69 0.64
1.7 0.49
1.71 0.48
1.72 0.59
1.73 0.53
1.74 0.53
1.75 0.5
1.76 0.38
1.77 0.47
1.78 0.44
1.79 0.5
1.8 0.41
1.81 0.53
1.82 0.53
1.83 0.47
1.84 0.47
1.85 0.48
1.86 0.45
1.87 0.62
1.88 0.47
1.89 0.45
1.9 0.46
1.91 0.52
1.92 0.57
1.93 0.54
1.94 0.56
1.95 0.49
1.96 0.54
1.97 0.47
1.98 0.49
1.99 0.48
2 0.47
};
\addplot [semithick, color7, opacity=0.75, dashed]
table {%
0.01 0.51
0.02 0.63
0.03 0.55
0.04 0.56
0.05 0.49
0.06 0.53
0.07 0.51
0.08 0.46
0.09 0.51
0.1 0.52
0.11 0.57
0.12 0.51
0.13 0.5
0.14 0.57
0.15 0.59
0.16 0.57
0.17 0.51
0.18 0.56
0.19 0.62
0.2 0.6
0.21 0.62
0.22 0.59
0.23 0.61
0.24 0.62
0.25 0.62
0.26 0.64
0.27 0.55
0.28 0.6
0.29 0.64
0.3 0.69
0.31 0.63
0.32 0.75
0.33 0.63
0.34 0.71
0.35 0.68
0.36 0.71
0.37 0.67
0.38 0.75
0.39 0.72
0.4 0.68
0.41 0.83
0.42 0.73
0.43 0.71
0.44 0.69
0.45 0.68
0.46 0.71
0.47 0.76
0.48 0.78
0.49 0.73
0.5 0.74
0.51 0.75
0.52 0.72
0.53 0.74
0.54 0.76
0.55 0.73
0.56 0.75
0.57 0.79
0.58 0.78
0.59 0.76
0.6 0.74
0.61 0.78
0.62 0.78
0.63 0.77
0.64 0.81
0.65 0.69
0.66 0.81
0.67 0.76
0.68 0.8
0.69 0.69
0.7 0.82
0.71 0.86
0.72 0.81
0.73 0.75
0.74 0.86
0.75 0.81
0.76 0.75
0.77 0.81
0.78 0.84
0.79 0.78
0.8 0.81
0.81 0.71
0.82 0.79
0.83 0.75
0.84 0.83
0.85 0.84
0.86 0.87
0.87 0.8
0.88 0.83
0.89 0.86
0.9 0.79
0.91 0.89
0.92 0.76
0.93 0.82
0.94 0.8
0.95 0.83
0.96 0.8
0.97 0.77
0.98 0.8
0.99 0.79
1 0.86
1.02 0.76
1.03 0.73
1.04 0.6
1.05 0.54
1.06 0.51
1.07 0.52
1.08 0.48
1.09 0.52
1.1 0.46
1.11 0.52
1.12 0.55
1.13 0.48
1.14 0.51
1.15 0.55
1.16 0.51
1.17 0.56
1.18 0.58
1.19 0.47
1.2 0.36
1.21 0.49
1.22 0.5
1.23 0.5
1.24 0.57
1.25 0.56
1.26 0.48
1.27 0.49
1.28 0.45
1.29 0.57
1.3 0.59
1.31 0.51
1.32 0.51
1.33 0.49
1.34 0.45
1.35 0.44
1.36 0.53
1.37 0.5
1.38 0.49
1.39 0.59
1.4 0.54
1.41 0.62
1.42 0.48
1.43 0.48
1.44 0.49
1.45 0.45
1.46 0.58
1.47 0.52
1.48 0.41
1.49 0.54
1.5 0.55
1.51 0.61
1.52 0.46
1.53 0.43
1.54 0.5
1.55 0.48
1.56 0.5
1.57 0.51
1.58 0.49
1.59 0.51
1.6 0.51
1.61 0.5
1.62 0.55
1.63 0.58
1.64 0.56
1.65 0.48
1.66 0.53
1.67 0.45
1.68 0.52
1.69 0.64
1.7 0.49
1.71 0.48
1.72 0.59
1.73 0.53
1.74 0.53
1.75 0.5
1.76 0.38
1.77 0.47
1.78 0.44
1.79 0.5
1.8 0.41
1.81 0.53
1.82 0.53
1.83 0.47
1.84 0.47
1.85 0.48
1.86 0.45
1.87 0.62
1.88 0.47
1.89 0.45
1.9 0.46
1.91 0.52
1.92 0.57
1.93 0.54
1.94 0.56
1.95 0.49
1.96 0.54
1.97 0.47
1.98 0.49
1.99 0.48
2 0.47
};
\addplot [semithick, color8, opacity=0.75, dashed]
table {%
0.01 0.44
0.02 0.46
0.03 0.54
0.04 0.58
0.05 0.52
0.06 0.55
0.07 0.54
0.08 0.49
0.09 0.65
0.1 0.57
0.11 0.49
0.12 0.6
0.13 0.55
0.14 0.55
0.15 0.65
0.16 0.67
0.17 0.6
0.18 0.6
0.19 0.65
0.2 0.62
0.21 0.61
0.22 0.6
0.23 0.64
0.24 0.7
0.25 0.63
0.26 0.62
0.27 0.58
0.28 0.62
0.29 0.72
0.3 0.67
0.31 0.63
0.32 0.71
0.33 0.69
0.34 0.68
0.35 0.7
0.36 0.73
0.37 0.71
0.38 0.82
0.39 0.82
0.4 0.76
0.41 0.8
0.42 0.78
0.43 0.75
0.44 0.88
0.45 0.79
0.46 0.77
0.47 0.78
0.48 0.76
0.49 0.74
0.5 0.82
0.51 0.8
0.52 0.77
0.53 0.8
0.54 0.8
0.55 0.79
0.56 0.78
0.57 0.85
0.58 0.84
0.59 0.82
0.6 0.8
0.61 0.92
0.62 0.83
0.63 0.82
0.64 0.9
0.65 0.82
0.66 0.84
0.67 0.82
0.68 0.83
0.69 0.83
0.7 0.89
0.71 0.82
0.72 0.85
0.73 0.81
0.74 0.9
0.75 0.9
0.76 0.77
0.77 0.87
0.78 0.88
0.79 0.85
0.8 0.88
0.81 0.81
0.82 0.86
0.83 0.88
0.84 0.9
0.85 0.88
0.86 0.86
0.87 0.84
0.88 0.91
0.89 0.94
0.9 0.81
0.91 0.87
0.92 0.8
0.93 0.87
0.94 0.89
0.95 0.84
0.96 0.85
0.97 0.85
0.98 0.81
0.99 0.91
1 0.86
1.02 0.79
1.03 0.76
1.04 0.65
1.05 0.56
1.06 0.56
1.07 0.64
1.08 0.45
1.09 0.61
1.1 0.54
1.11 0.51
1.12 0.48
1.13 0.44
1.14 0.53
1.15 0.49
1.16 0.55
1.17 0.48
1.18 0.55
1.19 0.55
1.2 0.42
1.21 0.47
1.22 0.55
1.23 0.55
1.24 0.54
1.25 0.6
1.26 0.52
1.27 0.49
1.28 0.35
1.29 0.53
1.3 0.61
1.31 0.52
1.32 0.48
1.33 0.47
1.34 0.45
1.35 0.56
1.36 0.55
1.37 0.54
1.38 0.53
1.39 0.44
1.4 0.55
1.41 0.61
1.42 0.55
1.43 0.54
1.44 0.41
1.45 0.47
1.46 0.54
1.47 0.5
1.48 0.49
1.49 0.47
1.5 0.49
1.51 0.58
1.52 0.54
1.53 0.46
1.54 0.58
1.55 0.49
1.56 0.52
1.57 0.51
1.58 0.48
1.59 0.53
1.6 0.54
1.61 0.53
1.62 0.5
1.63 0.52
1.64 0.6
1.65 0.52
1.66 0.43
1.67 0.47
1.68 0.51
1.69 0.51
1.7 0.5
1.71 0.45
1.72 0.54
1.73 0.58
1.74 0.5
1.75 0.43
1.76 0.46
1.77 0.48
1.78 0.51
1.79 0.51
1.8 0.56
1.81 0.52
1.82 0.55
1.83 0.45
1.84 0.44
1.85 0.56
1.86 0.55
1.87 0.49
1.88 0.42
1.89 0.46
1.9 0.51
1.91 0.55
1.92 0.5
1.93 0.49
1.94 0.41
1.95 0.47
1.96 0.51
1.97 0.47
1.98 0.52
1.99 0.57
2 0.47
};
\addplot [semithick, color9, opacity=0.75, dashed]
table {%
0.01 0.58
0.02 0.53
0.03 0.55
0.04 0.66
0.05 0.63
0.06 0.69
0.07 0.64
0.08 0.7
0.09 0.81
0.1 0.8
0.11 0.81
0.12 0.91
0.13 0.81
0.14 0.87
0.15 0.89
0.16 0.86
0.17 0.88
0.18 0.92
0.19 0.92
0.2 0.92
0.21 0.97
0.22 0.94
0.23 0.98
0.24 0.99
0.25 0.96
0.26 1
0.27 0.96
0.28 0.96
0.29 1
0.3 0.96
0.31 0.97
0.32 0.98
0.33 0.98
0.34 0.97
0.35 0.97
0.36 0.99
0.37 1
0.38 1
0.39 1
0.4 0.99
0.41 0.99
0.42 1
0.43 0.99
0.44 0.99
0.45 1
0.46 0.99
0.47 1
0.48 1
0.49 1
0.5 1
0.51 1
0.52 0.99
0.53 1
0.54 1
0.55 0.99
0.56 1
0.57 1
0.58 1
0.59 1
0.6 1
0.61 1
0.62 1
0.63 0.99
0.64 0.99
0.65 1
0.66 1
0.67 1
0.68 1
0.69 0.99
0.7 1
0.71 1
0.72 1
0.73 0.99
0.74 0.99
0.75 1
0.76 0.99
0.77 1
0.78 1
0.79 1
0.8 0.99
0.81 1
0.82 1
0.83 1
0.84 0.99
0.85 0.99
0.86 1
0.87 1
0.88 1
0.89 1
0.9 1
0.91 1
0.92 1
0.93 1
0.94 1
0.95 1
0.96 1
0.97 1
0.98 1
0.99 1
1 1
1.02 1
1.03 0.99
1.04 0.97
1.05 0.91
1.06 0.9
1.07 0.87
1.08 0.84
1.09 0.79
1.1 0.81
1.11 0.74
1.12 0.72
1.13 0.71
1.14 0.75
1.15 0.72
1.16 0.66
1.17 0.65
1.18 0.69
1.19 0.66
1.2 0.54
1.21 0.62
1.22 0.68
1.23 0.63
1.24 0.57
1.25 0.62
1.26 0.66
1.27 0.6
1.28 0.62
1.29 0.59
1.3 0.61
1.31 0.48
1.32 0.6
1.33 0.48
1.34 0.45
1.35 0.58
1.36 0.53
1.37 0.56
1.38 0.5
1.39 0.56
1.4 0.57
1.41 0.5
1.42 0.51
1.43 0.53
1.44 0.56
1.45 0.55
1.46 0.53
1.47 0.47
1.48 0.48
1.49 0.55
1.5 0.49
1.51 0.57
1.52 0.53
1.53 0.52
1.54 0.53
1.55 0.46
1.56 0.59
1.57 0.59
1.58 0.47
1.59 0.6
1.6 0.51
1.61 0.56
1.62 0.49
1.63 0.52
1.64 0.55
1.65 0.46
1.66 0.55
1.67 0.59
1.68 0.67
1.69 0.48
1.7 0.5
1.71 0.47
1.72 0.56
1.73 0.52
1.74 0.57
1.75 0.49
1.76 0.52
1.77 0.44
1.78 0.52
1.79 0.72
1.8 0.53
1.81 0.51
1.82 0.51
1.83 0.43
1.84 0.48
1.85 0.49
1.86 0.53
1.87 0.43
1.88 0.58
1.89 0.41
1.9 0.5
1.91 0.47
1.92 0.55
1.93 0.51
1.94 0.44
1.95 0.44
1.96 0.53
1.97 0.51
1.98 0.52
1.99 0.53
2 0.55
};
\addplot [semithick, color10, opacity=0.75, dashed]
table {%
0.01 0.53
0.02 0.53
0.03 0.58
0.04 0.62
0.05 0.63
0.06 0.67
0.07 0.61
0.08 0.64
0.09 0.76
0.1 0.76
0.11 0.76
0.12 0.85
0.13 0.72
0.14 0.82
0.15 0.76
0.16 0.86
0.17 0.86
0.18 0.93
0.19 0.9
0.2 0.91
0.21 0.93
0.22 0.94
0.23 0.96
0.24 0.98
0.25 0.91
0.26 0.96
0.27 0.95
0.28 0.95
0.29 0.99
0.3 0.95
0.31 0.96
0.32 0.99
0.33 0.98
0.34 0.97
0.35 0.96
0.36 0.98
0.37 0.98
0.38 1
0.39 1
0.4 0.98
0.41 1
0.42 0.99
0.43 0.99
0.44 0.98
0.45 0.99
0.46 1
0.47 1
0.48 1
0.49 0.98
0.5 1
0.51 1
0.52 0.99
0.53 0.99
0.54 0.99
0.55 1
0.56 0.99
0.57 0.99
0.58 1
0.59 1
0.6 1
0.61 1
0.62 1
0.63 0.99
0.64 0.99
0.65 1
0.66 1
0.67 0.98
0.68 1
0.69 0.99
0.7 1
0.71 1
0.72 1
0.73 0.99
0.74 1
0.75 1
0.76 0.99
0.77 1
0.78 1
0.79 1
0.8 0.99
0.81 1
0.82 1
0.83 1
0.84 1
0.85 0.99
0.86 1
0.87 1
0.88 0.99
0.89 1
0.9 1
0.91 1
0.92 1
0.93 1
0.94 1
0.95 1
0.96 1
0.97 1
0.98 1
0.99 1
1 0.98
1.02 0.99
1.03 0.97
1.04 0.95
1.05 0.9
1.06 0.88
1.07 0.84
1.08 0.77
1.09 0.78
1.1 0.79
1.11 0.73
1.12 0.72
1.13 0.61
1.14 0.68
1.15 0.6
1.16 0.61
1.17 0.65
1.18 0.63
1.19 0.64
1.2 0.56
1.21 0.63
1.22 0.5
1.23 0.67
1.24 0.55
1.25 0.61
1.26 0.62
1.27 0.53
1.28 0.61
1.29 0.63
1.3 0.58
1.31 0.52
1.32 0.6
1.33 0.45
1.34 0.45
1.35 0.55
1.36 0.56
1.37 0.6
1.38 0.56
1.39 0.6
1.4 0.63
1.41 0.54
1.42 0.58
1.43 0.51
1.44 0.61
1.45 0.54
1.46 0.48
1.47 0.52
1.48 0.5
1.49 0.46
1.5 0.5
1.51 0.59
1.52 0.54
1.53 0.45
1.54 0.5
1.55 0.51
1.56 0.53
1.57 0.57
1.58 0.52
1.59 0.53
1.6 0.56
1.61 0.54
1.62 0.5
1.63 0.55
1.64 0.45
1.65 0.43
1.66 0.55
1.67 0.52
1.68 0.58
1.69 0.48
1.7 0.52
1.71 0.46
1.72 0.54
1.73 0.48
1.74 0.55
1.75 0.45
1.76 0.49
1.77 0.45
1.78 0.52
1.79 0.53
1.8 0.5
1.81 0.5
1.82 0.48
1.83 0.51
1.84 0.39
1.85 0.45
1.86 0.56
1.87 0.46
1.88 0.54
1.89 0.46
1.9 0.43
1.91 0.54
1.92 0.54
1.93 0.48
1.94 0.52
1.95 0.51
1.96 0.57
1.97 0.56
1.98 0.52
1.99 0.49
2 0.5
};
\addplot [semithick, color11, opacity=0.75, dashed]
table {%
0.01 0.48
0.02 0.52
0.03 0.47
0.04 0.7
0.05 0.56
0.06 0.58
0.07 0.6
0.08 0.66
0.09 0.78
0.1 0.73
0.11 0.7
0.12 0.8
0.13 0.74
0.14 0.76
0.15 0.8
0.16 0.83
0.17 0.78
0.18 0.91
0.19 0.88
0.2 0.88
0.21 0.94
0.22 0.89
0.23 0.92
0.24 0.97
0.25 0.92
0.26 0.89
0.27 0.9
0.28 0.93
0.29 0.95
0.3 0.91
0.31 0.93
0.32 0.92
0.33 0.95
0.34 0.9
0.35 0.96
0.36 0.93
0.37 0.94
0.38 0.98
0.39 0.98
0.4 0.97
0.41 0.94
0.42 0.94
0.43 0.94
0.44 0.96
0.45 0.98
0.46 0.96
0.47 0.97
0.48 0.97
0.49 0.95
0.5 0.98
0.51 0.97
0.52 0.97
0.53 0.96
0.54 0.98
0.55 0.95
0.56 0.98
0.57 0.99
0.58 0.97
0.59 0.96
0.6 0.99
0.61 0.98
0.62 0.96
0.63 0.96
0.64 0.97
0.65 0.94
0.66 0.96
0.67 0.93
0.68 0.98
0.69 0.97
0.7 0.98
0.71 0.98
0.72 0.99
0.73 0.98
0.74 0.95
0.75 0.99
0.76 0.98
0.77 0.98
0.78 0.98
0.79 0.99
0.8 0.96
0.81 0.99
0.82 0.95
0.83 0.95
0.84 0.93
0.85 0.93
0.86 0.97
0.87 0.97
0.88 0.97
0.89 0.96
0.9 0.98
0.91 0.95
0.92 0.95
0.93 0.92
0.94 0.99
0.95 0.96
0.96 0.93
0.97 0.94
0.98 0.96
0.99 0.98
1 0.96
1.02 0.98
1.03 0.97
1.04 0.85
1.05 0.85
1.06 0.82
1.07 0.81
1.08 0.82
1.09 0.77
1.1 0.77
1.11 0.67
1.12 0.67
1.13 0.59
1.14 0.61
1.15 0.66
1.16 0.57
1.17 0.67
1.18 0.68
1.19 0.67
1.2 0.58
1.21 0.52
1.22 0.58
1.23 0.65
1.24 0.65
1.25 0.57
1.26 0.61
1.27 0.59
1.28 0.55
1.29 0.58
1.3 0.56
1.31 0.55
1.32 0.51
1.33 0.51
1.34 0.51
1.35 0.64
1.36 0.6
1.37 0.51
1.38 0.51
1.39 0.49
1.4 0.61
1.41 0.57
1.42 0.53
1.43 0.43
1.44 0.54
1.45 0.55
1.46 0.57
1.47 0.52
1.48 0.58
1.49 0.52
1.5 0.47
1.51 0.59
1.52 0.5
1.53 0.51
1.54 0.56
1.55 0.44
1.56 0.43
1.57 0.6
1.58 0.46
1.59 0.63
1.6 0.53
1.61 0.5
1.62 0.51
1.63 0.52
1.64 0.45
1.65 0.56
1.66 0.56
1.67 0.56
1.68 0.47
1.69 0.56
1.7 0.5
1.71 0.39
1.72 0.54
1.73 0.52
1.74 0.5
1.75 0.45
1.76 0.53
1.77 0.48
1.78 0.57
1.79 0.63
1.8 0.51
1.81 0.44
1.82 0.5
1.83 0.5
1.84 0.54
1.85 0.49
1.86 0.55
1.87 0.43
1.88 0.45
1.89 0.47
1.9 0.49
1.91 0.44
1.92 0.44
1.93 0.54
1.94 0.46
1.95 0.47
1.96 0.53
1.97 0.52
1.98 0.53
1.99 0.5
2 0.48
};
\addplot [semithick, black, opacity=1, dash pattern=on 1pt off 1pt]
table {%
-0.0895000000000001 0.5
2.0995 0.5
};
\addplot [semithick, black, opacity=1, dash pattern=on 1pt off 1pt]
table {%
-0.0895000000000001 1
2.0995 1
};
\addplot [semithick, black, opacity=1, dash pattern=on 1pt off 1pt]
table {%
1 0.2
1 1.05
};
\end{axis}

\end{tikzpicture}

%% file: sections/tex-plots-tables/tab_lin.tex
\begin{table*}[]
\caption{Summary for linear models. The numbers reflect the ranges of noise that allow identifiability with accuracy $\approx 90\%$.}
\label{tab:summary:linear}
\centering
\begin{tabular}{l|c|c|c|c|c|c|c|c|c}
Estimator             & $\mathcal{N}+\mathcal{N}$ & $\mathcal{N}+\mathcal{U}$ & $\mathcal{N}+\mathcal{L}$ & $\mathcal{U}+\mathcal{N}$ & $\mathcal{U}+\mathcal{U}$ & $\mathcal{U}+\mathcal{L}$ & $\mathcal{L}+\mathcal{N}$ & $\mathcal{L}+\mathcal{U}$ & $\mathcal{L}+\mathcal{L}$ \\
\hline
\textbf{HSIC}         &                & 0.60 -- 4       & 0.38   -- 2     & 0.21   -- 1     & 0.18   -- 6     & 0.15   -- 1     & 0.32   -- 3     & 0.40   -- 8     & 0.30   -- 5     \\
\textbf{HISC\_IC}      &                & 3 -- 7          & 0.75   -- 2     &                & 3              & 0.35   -- 1     & 0.36   -- 3     & 0.60   -- 7     & 0.40   -- 4     \\
\textbf{HSIC\_IC2}     &                & 3 -- 7          & 0.75   -- 2     &                & 3              & 0.35   -- 1     & 0.36   -- 3     & 0.60   -- 7     & 0.40   -- 4     \\
\textbf{DISTCOV}      &                & 1              & 0.80   -- 1     & 0.25 -- 1       & 0.20   -- 3     & 0.20   -- 1     & 0.33   -- 2     & 0.40   -- 5     & 0.23   -- 4     \\
\textbf{DISTCORR}     &                & 1              & 0.87   -- 1     & 0.25   -- 1     & 0.20   -- 3     & 0.20   -- 1     & 0.33   -- 2     & 0.40   -- 5     & 0.23   -- 4     \\
\textbf{HOEFFDING}    &                &                &                & 1              & 0.20   -- 3     & 0.15   -- 1     & 0.57   -- 1     & 0.40   -- 5     &                \\
\textbf{SH\_KNN}       &                &                &                & 0.26 -- 1       & 0.20   -- 3     & 0.21   -- 1     &                & 0.53 -- 4       &                \\
\textbf{SH\_KNN\_2}     &                &                &                & 0.26 -- 1       & 0.20   -- 2     & 0.21   -- 1     &                & 0.53 -- 4       &                \\
\textbf{SH\_KNN\_3}     &                & 0.85 -- 4       &                & 0.20 -- 1       & 0.20   -- 3     & 0.16   -- 1     &                & 0.51 -- 5       & 0.60 -- 1       \\
\textbf{SH\_MAXENT1}   &                & 0.65 -- 5       & 0.30 -- 3       & 0.20   -- 1     & 0.23   -- 3     & 0.12   -- 3     & 0.21 -- 4       & 0.32   -- 10    & 0.20   -- 6     \\
\textbf{SH\_MAXENT2}   &                & 0.40 -- 8       & 0.33   -- 3     & 0.10   -- 4     & 0.12   -- 8     & 0.10   -- 3     & 0.21   -- 4     & 0.32   -- 10    & 0.17   -- 5     \\
\textbf{SH\_SPACING\_V} &                & 0.20 -- 22      & 0.82   -- 1     & 0.04   -- 9     & 0.05   -- 21    & 0.03   -- 8     & 0.49   -- 4     & 0.16   -- 27    & 0.17   -- 4    
\end{tabular}
\end{table*}

%% file: sections/tex-plots-tables/fig_nonlin.tex
\begin{figure*}[]
\centering
\begin{subfigure}{.333\linewidth}
  \centering
  \input{plots/decoupled2/NLGAU}
  \vspace{-.5cm}
  \caption{$\mathcal{N}^3+\mathcal{N}$}
  \vspace{.5cm}
  \label{fig:acc-Nlin:NN}
\end{subfigure}%
\begin{subfigure}{.333\textwidth}
  \centering
  \input{plots/decoupled2/NL_GAUxUNI}
  \vspace{-.5cm}
  \caption{$\mathcal{N}^3+\mathcal{U}$}
  \vspace{.5cm}
  \label{fig:acc-Nlin:NU}
\end{subfigure}%
\begin{subfigure}{.333\textwidth}
  \centering
  \input{plots/decoupled2/NL_GAUxLAP}
  \vspace{-.5cm}
  \caption{$\mathcal{N}^3+\mathcal{L}$}
  \vspace{.5cm}
  \label{fig:acc-Nlin:NL}
\end{subfigure}
\begin{subfigure}{.333\textwidth}
  \centering
  \input{plots/decoupled2/NL_UNIxGAU}
  \vspace{-.5cm}
  \caption{$\mathcal{U}^3+\mathcal{N}$}
  \vspace{.5cm}
  \label{fig:acc-Nlin:UN}
\end{subfigure}%
\begin{subfigure}{.333\textwidth}
  \centering
  \input{plots/decoupled2/NLUNI}
  \vspace{-.5cm}
  \caption{$\mathcal{U}^3+\mathcal{U}$}
  \vspace{.5cm}
  \label{fig:acc-Nlin:UU}
\end{subfigure}%
\begin{subfigure}{.333\textwidth}
  \centering
  \input{plots/decoupled2/NL_UNIxLAP}
  \vspace{-.5cm}
  \caption{$\mathcal{U}^3+\mathcal{L}$}
  \vspace{.5cm}
  \label{fig:acc-Nlin:UL}
\end{subfigure}
\begin{subfigure}{.33\textwidth}
  \centering
  \input{plots/decoupled2/NL_LAPxGAU}
  \vspace{-.5cm}
  \caption{$\mathcal{L}^3+\mathcal{N}$}
  \vspace{.5cm}
  \label{fig:acc-Nlin:LN}
\end{subfigure}%
\begin{subfigure}{.33\textwidth}
  \centering
  \input{plots/decoupled2/NL_LAPxUNI}
  \vspace{-.5cm}
  \caption{$\mathcal{L}^3+\mathcal{U}$}
  \vspace{.5cm}
  \label{fig:acc-Nlin:LU}
\end{subfigure}
\begin{subfigure}{.33\textwidth}
  \centering
  \input{plots/decoupled2/NLLAP}
  \vspace{-.5cm}
  \caption{$\mathcal{L}^3+\mathcal{L}$}
  \vspace{.5cm}
  \label{fig:acc-Nlin:LL}
\end{subfigure}
\begin{subfigure}{.8\textwidth}
  \centering
  \includegraphics[scale=.6]{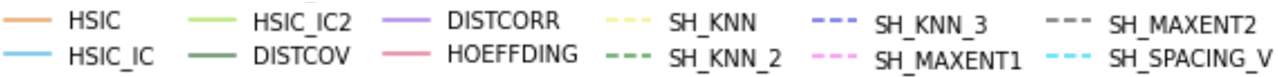}
  %\caption{A subfigure}
  %\label{fig:sub2}
\end{subfigure}
\caption{Accuracy of RESIT for nonlinear models as a function of $i$-factor.}
\label{fig:acc-Nlin}
\vspace{5mm}
\end{figure*}

%% file: plots/decoupled2/NLGAU.tex
% This file was created by tikzplotlib v0.9.6.
\begin{tikzpicture}

\definecolor{color0}{rgb}{0.866666666666667,0.494117647058824,0.164705882352941}
\definecolor{color1}{rgb}{0.164705882352941,0.643137254901961,0.866666666666667}
\definecolor{color2}{rgb}{0.584313725490196,0.866666666666667,0.164705882352941}
\definecolor{color3}{rgb}{0.109803921568627,0.337254901960784,0.129411764705882}
\definecolor{color4}{rgb}{0.529411764705882,0.305882352941176,0.858823529411765}
\definecolor{color5}{rgb}{0.858823529411765,0.305882352941176,0.435294117647059}
\definecolor{color6}{rgb}{0.937254901960784,0.929411764705882,0.392156862745098}
\definecolor{color7}{rgb}{0.0901960784313725,0.486274509803922,0.0980392156862745}
\definecolor{color8}{rgb}{0.156862745098039,0.188235294117647,0.827450980392157}
\definecolor{color9}{rgb}{0.937254901960784,0.392156862745098,0.894117647058824}
\definecolor{color10}{rgb}{0.2,0.184313725490196,0.184313725490196}
\definecolor{color11}{rgb}{0.0156862745098039,0.803921568627451,0.976470588235294}

\begin{axis}[
tick align=outside,
tick pos=left,
x grid style={white!69.0196078431373!black},
xmajorgrids,
xmin=-0.0895, xmax=2.0995,
xtick style={color=black},
xtick={0,0.1,0.2,0.3,0.4,0.5,0.6,0.7,0.8,0.9,1,1.1,1.2,1.3,1.4,1.5,1.6,1.7,1.8,1.9,2},
xticklabels={0,,.2,,.4,,.6,,.8,,1,,20,,40,,60,,80,,100},
height=4.8cm,
width=6.5cm,
y grid style={white!69.0196078431373!black},
ymajorgrids,
ymin=0.2, ymax=1.05,
ytick style={color=black}
]
\addplot [semithick, color0, opacity=0.75]
table {%
0.01 0.46
0.02 0.66
0.03 0.72
0.04 0.72
0.05 0.78
0.06 0.85
0.07 0.82
0.08 0.87
0.09 0.92
0.1 0.91
0.11 0.96
0.12 0.93
0.13 0.94
0.14 0.99
0.15 0.97
0.16 0.98
0.17 0.99
0.18 0.98
0.19 0.98
0.2 1
0.21 1
0.22 1
0.23 0.99
0.24 0.99
0.25 1
0.26 1
0.27 1
0.28 1
0.29 1
0.3 1
0.31 1
0.32 1
0.33 1
0.34 1
0.35 1
0.36 1
0.37 1
0.38 1
0.39 1
0.4 1
0.41 1
0.42 1
0.43 1
0.44 1
0.45 1
0.46 1
0.47 1
0.48 1
0.49 1
0.5 1
0.51 1
0.52 1
0.53 1
0.54 1
0.55 1
0.56 1
0.57 1
0.58 1
0.59 1
0.6 1
0.61 1
0.62 1
0.63 1
0.64 1
0.65 1
0.66 1
0.67 1
0.68 1
0.69 1
0.7 1
0.71 1
0.72 1
0.73 1
0.74 1
0.75 1
0.76 1
0.77 1
0.78 1
0.79 1
0.8 1
0.81 1
0.82 1
0.83 1
0.84 1
0.85 1
0.86 1
0.87 1
0.88 1
0.89 1
0.9 1
0.91 1
0.92 1
0.93 1
0.94 1
0.95 1
0.96 1
0.97 1
0.98 1
0.99 1
1 1
1.02 1
1.03 1
1.04 1
1.05 1
1.06 1
1.07 1
1.08 1
1.09 1
1.1 1
1.11 1
1.12 1
1.13 1
1.14 1
1.15 1
1.16 1
1.17 1
1.18 1
1.19 1
1.2 1
1.21 1
1.22 1
1.23 1
1.24 1
1.25 1
1.26 1
1.27 1
1.28 0.99
1.29 1
1.3 1
1.31 0.99
1.32 1
1.33 0.99
1.34 0.99
1.35 0.98
1.36 0.98
1.37 0.99
1.38 0.97
1.39 0.99
1.4 1
1.41 0.99
1.42 0.98
1.43 0.98
1.44 0.98
1.45 0.97
1.46 0.98
1.47 0.93
1.48 0.99
1.49 0.99
1.5 0.97
1.51 0.99
1.52 0.96
1.53 0.93
1.54 0.96
1.55 0.97
1.56 0.95
1.57 0.94
1.58 0.94
1.59 0.96
1.6 0.97
1.61 0.95
1.62 0.96
1.63 0.92
1.64 0.93
1.65 0.93
1.66 0.9
1.67 0.94
1.68 0.89
1.69 0.93
1.7 0.95
1.71 0.95
1.72 0.94
1.73 0.95
1.74 0.92
1.75 0.91
1.76 0.92
1.77 0.89
1.78 0.9
1.79 0.89
1.8 0.92
1.81 0.94
1.82 0.92
1.83 0.89
1.84 0.92
1.85 0.96
1.86 0.88
1.87 0.9
1.88 0.93
1.89 0.91
1.9 0.86
1.91 0.92
1.92 0.9
1.93 0.83
1.94 0.92
1.95 0.92
1.96 0.86
1.97 0.9
1.98 0.93
1.99 0.86
2 0.91
};
\addplot [semithick, color1, opacity=0.75]
table {%
0.01 0.53
0.02 0.66
0.03 0.68
0.04 0.74
0.05 0.77
0.06 0.77
0.07 0.79
0.08 0.86
0.09 0.89
0.1 0.9
0.11 0.93
0.12 0.92
0.13 0.92
0.14 0.96
0.15 0.95
0.16 0.96
0.17 0.97
0.18 0.98
0.19 0.97
0.2 0.99
0.21 0.99
0.22 1
0.23 0.99
0.24 0.99
0.25 1
0.26 1
0.27 1
0.28 1
0.29 1
0.3 1
0.31 1
0.32 1
0.33 1
0.34 1
0.35 1
0.36 1
0.37 1
0.38 1
0.39 1
0.4 1
0.41 1
0.42 1
0.43 1
0.44 1
0.45 1
0.46 1
0.47 1
0.48 1
0.49 1
0.5 1
0.51 1
0.52 1
0.53 1
0.54 1
0.55 1
0.56 1
0.57 1
0.58 1
0.59 1
0.6 1
0.61 1
0.62 1
0.63 1
0.64 1
0.65 1
0.66 1
0.67 1
0.68 1
0.69 1
0.7 1
0.71 1
0.72 1
0.73 1
0.74 1
0.75 1
0.76 1
0.77 1
0.78 1
0.79 1
0.8 1
0.81 1
0.82 1
0.83 1
0.84 1
0.85 1
0.86 1
0.87 1
0.88 1
0.89 1
0.9 1
0.91 1
0.92 1
0.93 1
0.94 1
0.95 1
0.96 1
0.97 1
0.98 1
0.99 1
1 1
1.02 1
1.03 1
1.04 1
1.05 1
1.06 1
1.07 1
1.08 1
1.09 1
1.1 1
1.11 1
1.12 1
1.13 1
1.14 1
1.15 1
1.16 1
1.17 1
1.18 1
1.19 1
1.2 0.99
1.21 1
1.22 1
1.23 1
1.24 1
1.25 1
1.26 0.99
1.27 0.99
1.28 0.96
1.29 1
1.3 0.99
1.31 0.93
1.32 0.97
1.33 0.98
1.34 0.94
1.35 0.94
1.36 0.96
1.37 0.94
1.38 0.96
1.39 0.94
1.4 0.93
1.41 0.83
1.42 0.92
1.43 0.93
1.44 0.9
1.45 0.88
1.46 0.93
1.47 0.79
1.48 0.88
1.49 0.87
1.5 0.86
1.51 0.88
1.52 0.9
1.53 0.86
1.54 0.86
1.55 0.86
1.56 0.75
1.57 0.8
1.58 0.81
1.59 0.86
1.6 0.89
1.61 0.72
1.62 0.77
1.63 0.8
1.64 0.81
1.65 0.79
1.66 0.83
1.67 0.74
1.68 0.74
1.69 0.82
1.7 0.8
1.71 0.74
1.72 0.76
1.73 0.86
1.74 0.78
1.75 0.77
1.76 0.82
1.77 0.73
1.78 0.71
1.79 0.78
1.8 0.83
1.81 0.71
1.82 0.8
1.83 0.7
1.84 0.76
1.85 0.77
1.86 0.76
1.87 0.71
1.88 0.73
1.89 0.81
1.9 0.73
1.91 0.69
1.92 0.74
1.93 0.69
1.94 0.76
1.95 0.81
1.96 0.64
1.97 0.7
1.98 0.66
1.99 0.7
2 0.84
};
\addplot [semithick, color2, opacity=0.75]
table {%
0.01 0.52
0.02 0.56
0.03 0.58
0.04 0.62
0.05 0.59
0.06 0.68
0.07 0.77
0.08 0.76
0.09 0.75
0.1 0.83
0.11 0.93
0.12 0.92
0.13 0.92
0.14 0.96
0.15 0.95
0.16 0.96
0.17 0.97
0.18 0.99
0.19 0.97
0.2 0.98
0.21 1
0.22 1
0.23 0.99
0.24 0.99
0.25 1
0.26 1
0.27 1
0.28 1
0.29 1
0.3 1
0.31 1
0.32 1
0.33 1
0.34 1
0.35 1
0.36 1
0.37 1
0.38 1
0.39 1
0.4 1
0.41 1
0.42 1
0.43 1
0.44 1
0.45 1
0.46 1
0.47 1
0.48 1
0.49 1
0.5 1
0.51 1
0.52 1
0.53 1
0.54 1
0.55 1
0.56 1
0.57 1
0.58 1
0.59 1
0.6 1
0.61 1
0.62 1
0.63 1
0.64 1
0.65 1
0.66 1
0.67 1
0.68 1
0.69 1
0.7 1
0.71 1
0.72 1
0.73 1
0.74 1
0.75 1
0.76 1
0.77 1
0.78 1
0.79 1
0.8 1
0.81 1
0.82 1
0.83 1
0.84 1
0.85 1
0.86 1
0.87 1
0.88 1
0.89 1
0.9 1
0.91 1
0.92 1
0.93 1
0.94 1
0.95 1
0.96 1
0.97 1
0.98 1
0.99 1
1 1
1.02 1
1.03 1
1.04 1
1.05 1
1.06 1
1.07 1
1.08 1
1.09 1
1.1 1
1.11 1
1.12 1
1.13 1
1.14 1
1.15 1
1.16 1
1.17 1
1.18 1
1.19 1
1.2 0.99
1.21 1
1.22 1
1.23 1
1.24 1
1.25 1
1.26 0.99
1.27 0.99
1.28 0.96
1.29 1
1.3 0.99
1.31 0.94
1.32 0.97
1.33 0.98
1.34 0.95
1.35 0.94
1.36 0.97
1.37 0.94
1.38 0.95
1.39 0.95
1.4 0.91
1.41 0.86
1.42 0.92
1.43 0.91
1.44 0.91
1.45 0.88
1.46 0.93
1.47 0.8
1.48 0.87
1.49 0.88
1.5 0.87
1.51 0.88
1.52 0.9
1.53 0.86
1.54 0.85
1.55 0.86
1.56 0.75
1.57 0.77
1.58 0.77
1.59 0.84
1.6 0.89
1.61 0.72
1.62 0.77
1.63 0.76
1.64 0.8
1.65 0.78
1.66 0.79
1.67 0.72
1.68 0.73
1.69 0.82
1.7 0.79
1.71 0.75
1.72 0.75
1.73 0.85
1.74 0.75
1.75 0.77
1.76 0.82
1.77 0.75
1.78 0.67
1.79 0.76
1.8 0.82
1.81 0.7
1.82 0.8
1.83 0.68
1.84 0.73
1.85 0.77
1.86 0.75
1.87 0.68
1.88 0.72
1.89 0.8
1.9 0.7
1.91 0.67
1.92 0.73
1.93 0.69
1.94 0.75
1.95 0.81
1.96 0.61
1.97 0.64
1.98 0.65
1.99 0.7
2 0.81
};
\addplot [semithick, color3, opacity=0.75]
table {%
0.01 0.67
0.02 0.75
0.03 0.83
0.04 0.84
0.05 0.88
0.06 0.94
0.07 0.87
0.08 0.92
0.09 0.96
0.1 0.95
0.11 0.96
0.12 0.94
0.13 0.98
0.14 0.99
0.15 0.99
0.16 0.97
0.17 0.97
0.18 0.98
0.19 1
0.2 1
0.21 1
0.22 0.99
0.23 1
0.24 0.99
0.25 0.99
0.26 1
0.27 1
0.28 0.99
0.29 1
0.3 1
0.31 1
0.32 1
0.33 1
0.34 1
0.35 1
0.36 1
0.37 1
0.38 0.99
0.39 1
0.4 1
0.41 1
0.42 1
0.43 1
0.44 1
0.45 1
0.46 1
0.47 0.99
0.48 1
0.49 1
0.5 1
0.51 1
0.52 1
0.53 1
0.54 1
0.55 1
0.56 1
0.57 1
0.58 1
0.59 1
0.6 1
0.61 1
0.62 1
0.63 1
0.64 1
0.65 1
0.66 1
0.67 1
0.68 1
0.69 1
0.7 1
0.71 1
0.72 1
0.73 1
0.74 1
0.75 1
0.76 1
0.77 1
0.78 1
0.79 1
0.8 1
0.81 1
0.82 1
0.83 1
0.84 1
0.85 1
0.86 1
0.87 1
0.88 1
0.89 1
0.9 1
0.91 1
0.92 1
0.93 1
0.94 1
0.95 1
0.96 1
0.97 1
0.98 1
0.99 1
1 1
1.02 1
1.03 1
1.04 1
1.05 1
1.06 1
1.07 1
1.08 1
1.09 1
1.1 1
1.11 1
1.12 1
1.13 1
1.14 1
1.15 1
1.16 1
1.17 1
1.18 1
1.19 1
1.2 1
1.21 1
1.22 1
1.23 1
1.24 1
1.25 1
1.26 1
1.27 1
1.28 0.99
1.29 1
1.3 1
1.31 1
1.32 1
1.33 1
1.34 1
1.35 0.99
1.36 1
1.37 0.99
1.38 0.98
1.39 1
1.4 1
1.41 0.99
1.42 0.98
1.43 0.99
1.44 1
1.45 0.99
1.46 0.98
1.47 0.97
1.48 1
1.49 0.99
1.5 0.97
1.51 0.99
1.52 1
1.53 0.96
1.54 0.97
1.55 0.97
1.56 0.98
1.57 0.94
1.58 0.96
1.59 1
1.6 0.98
1.61 0.97
1.62 0.99
1.63 0.96
1.64 0.92
1.65 0.97
1.66 0.94
1.67 0.96
1.68 0.96
1.69 0.96
1.7 0.95
1.71 0.99
1.72 0.97
1.73 0.95
1.74 0.98
1.75 0.9
1.76 0.97
1.77 0.93
1.78 0.93
1.79 0.93
1.8 0.96
1.81 0.95
1.82 0.97
1.83 0.97
1.84 0.97
1.85 0.98
1.86 0.94
1.87 0.9
1.88 0.92
1.89 0.91
1.9 0.94
1.91 0.99
1.92 0.93
1.93 0.93
1.94 0.95
1.95 0.99
1.96 0.93
1.97 0.96
1.98 0.97
1.99 0.94
2 0.94
};
\addplot [semithick, color4, opacity=0.75]
table {%
0.01 0.68
0.02 0.75
0.03 0.83
0.04 0.84
0.05 0.89
0.06 0.94
0.07 0.88
0.08 0.92
0.09 0.96
0.1 0.95
0.11 0.96
0.12 0.94
0.13 0.99
0.14 0.99
0.15 0.99
0.16 0.97
0.17 0.97
0.18 0.98
0.19 1
0.2 1
0.21 1
0.22 0.99
0.23 1
0.24 0.99
0.25 0.99
0.26 1
0.27 1
0.28 0.99
0.29 1
0.3 1
0.31 1
0.32 1
0.33 1
0.34 1
0.35 1
0.36 1
0.37 1
0.38 0.99
0.39 1
0.4 1
0.41 1
0.42 1
0.43 1
0.44 1
0.45 1
0.46 1
0.47 0.99
0.48 1
0.49 1
0.5 1
0.51 1
0.52 1
0.53 1
0.54 1
0.55 1
0.56 1
0.57 1
0.58 1
0.59 1
0.6 1
0.61 1
0.62 1
0.63 1
0.64 1
0.65 1
0.66 1
0.67 1
0.68 1
0.69 1
0.7 1
0.71 1
0.72 1
0.73 1
0.74 1
0.75 1
0.76 1
0.77 1
0.78 1
0.79 1
0.8 1
0.81 1
0.82 1
0.83 1
0.84 1
0.85 1
0.86 1
0.87 1
0.88 1
0.89 1
0.9 1
0.91 1
0.92 1
0.93 1
0.94 1
0.95 1
0.96 1
0.97 1
0.98 1
0.99 1
1 1
1.02 1
1.03 1
1.04 1
1.05 1
1.06 1
1.07 1
1.08 1
1.09 1
1.1 1
1.11 1
1.12 1
1.13 1
1.14 1
1.15 1
1.16 1
1.17 1
1.18 1
1.19 1
1.2 1
1.21 1
1.22 1
1.23 1
1.24 1
1.25 1
1.26 1
1.27 1
1.28 0.99
1.29 1
1.3 1
1.31 1
1.32 1
1.33 1
1.34 1
1.35 0.99
1.36 1
1.37 0.99
1.38 0.98
1.39 1
1.4 1
1.41 0.99
1.42 0.98
1.43 0.99
1.44 1
1.45 0.99
1.46 0.98
1.47 0.97
1.48 1
1.49 0.99
1.5 0.97
1.51 0.99
1.52 1
1.53 0.96
1.54 0.98
1.55 0.97
1.56 0.98
1.57 0.94
1.58 0.97
1.59 1
1.6 0.98
1.61 0.97
1.62 0.99
1.63 0.97
1.64 0.92
1.65 0.97
1.66 0.94
1.67 0.95
1.68 0.96
1.69 0.96
1.7 0.95
1.71 0.99
1.72 0.98
1.73 0.95
1.74 0.98
1.75 0.9
1.76 0.97
1.77 0.93
1.78 0.93
1.79 0.93
1.8 0.96
1.81 0.95
1.82 0.96
1.83 0.97
1.84 0.97
1.85 0.98
1.86 0.94
1.87 0.9
1.88 0.94
1.89 0.91
1.9 0.94
1.91 0.99
1.92 0.93
1.93 0.93
1.94 0.95
1.95 0.99
1.96 0.93
1.97 0.96
1.98 0.97
1.99 0.94
2 0.94
};
\addplot [semithick, color5, opacity=0.75]
table {%
0.01 0.81
0.02 0.83
0.03 0.95
0.04 0.92
0.05 0.92
0.06 0.94
0.07 0.93
0.08 0.96
0.09 0.97
0.1 0.99
0.11 0.97
0.12 0.95
0.13 0.99
0.14 0.98
0.15 0.98
0.16 0.98
0.17 0.99
0.18 1
0.19 0.99
0.2 1
0.21 1
0.22 0.99
0.23 1
0.24 0.99
0.25 1
0.26 1
0.27 1
0.28 0.99
0.29 1
0.3 1
0.31 0.99
0.32 1
0.33 1
0.34 0.99
0.35 1
0.36 0.99
0.37 1
0.38 1
0.39 1
0.4 1
0.41 1
0.42 1
0.43 1
0.44 1
0.45 1
0.46 1
0.47 1
0.48 1
0.49 1
0.5 1
0.51 1
0.52 1
0.53 1
0.54 1
0.55 1
0.56 1
0.57 1
0.58 1
0.59 1
0.6 0.99
0.61 1
0.62 1
0.63 1
0.64 1
0.65 1
0.66 1
0.67 1
0.68 1
0.69 1
0.7 1
0.71 1
0.72 1
0.73 1
0.74 1
0.75 1
0.76 1
0.77 1
0.78 1
0.79 1
0.8 1
0.81 1
0.82 1
0.83 1
0.84 1
0.85 1
0.86 1
0.87 1
0.88 1
0.89 1
0.9 1
0.91 1
0.92 1
0.93 1
0.94 1
0.95 1
0.96 1
0.97 1
0.98 1
0.99 1
1 1
1.02 1
1.03 1
1.04 1
1.05 1
1.06 1
1.07 1
1.08 1
1.09 1
1.1 1
1.11 1
1.12 1
1.13 1
1.14 1
1.15 1
1.16 1
1.17 1
1.18 1
1.19 1
1.2 0.99
1.21 1
1.22 1
1.23 1
1.24 1
1.25 1
1.26 1
1.27 1
1.28 0.99
1.29 1
1.3 1
1.31 1
1.32 0.99
1.33 1
1.34 1
1.35 0.99
1.36 0.99
1.37 0.99
1.38 0.97
1.39 0.99
1.4 1
1.41 0.98
1.42 0.97
1.43 0.99
1.44 1
1.45 1
1.46 0.98
1.47 0.97
1.48 1
1.49 1
1.5 0.97
1.51 0.99
1.52 0.98
1.53 0.99
1.54 0.96
1.55 0.97
1.56 0.99
1.57 0.96
1.58 0.98
1.59 1
1.6 0.98
1.61 0.99
1.62 0.99
1.63 0.98
1.64 0.97
1.65 0.98
1.66 0.96
1.67 0.96
1.68 0.94
1.69 0.96
1.7 0.98
1.71 0.97
1.72 0.99
1.73 0.98
1.74 0.99
1.75 0.92
1.76 0.96
1.77 0.96
1.78 0.95
1.79 0.98
1.8 0.97
1.81 0.96
1.82 0.99
1.83 0.99
1.84 0.97
1.85 0.99
1.86 0.95
1.87 0.96
1.88 0.97
1.89 0.96
1.9 0.96
1.91 1
1.92 0.97
1.93 0.94
1.94 0.98
1.95 0.98
1.96 0.95
1.97 0.97
1.98 0.97
1.99 0.94
2 0.97
};
\addplot [semithick, color6, opacity=0.75, dashed]
table {%
0.01 0.99
0.02 1
0.03 0.99
0.04 1
0.05 1
0.06 1
0.07 1
0.08 1
0.09 1
0.1 1
0.11 1
0.12 1
0.13 1
0.14 1
0.15 1
0.16 1
0.17 1
0.18 1
0.19 1
0.2 1
0.21 1
0.22 1
0.23 1
0.24 1
0.25 1
0.26 1
0.27 1
0.28 1
0.29 1
0.3 1
0.31 1
0.32 1
0.33 1
0.34 1
0.35 1
0.36 1
0.37 1
0.38 1
0.39 1
0.4 1
0.41 1
0.42 1
0.43 1
0.44 1
0.45 1
0.46 1
0.47 1
0.48 1
0.49 1
0.5 1
0.51 1
0.52 1
0.53 1
0.54 1
0.55 1
0.56 1
0.57 1
0.58 1
0.59 1
0.6 1
0.61 1
0.62 1
0.63 1
0.64 1
0.65 1
0.66 1
0.67 1
0.68 1
0.69 1
0.7 1
0.71 1
0.72 1
0.73 1
0.74 1
0.75 1
0.76 1
0.77 1
0.78 1
0.79 1
0.8 1
0.81 1
0.82 1
0.83 1
0.84 1
0.85 1
0.86 1
0.87 1
0.88 1
0.89 1
0.9 1
0.91 1
0.92 1
0.93 1
0.94 1
0.95 1
0.96 1
0.97 1
0.98 1
0.99 1
1 1
1.02 1
1.03 1
1.04 1
1.05 1
1.06 1
1.07 1
1.08 1
1.09 1
1.1 1
1.11 1
1.12 1
1.13 1
1.14 1
1.15 1
1.16 1
1.17 1
1.18 1
1.19 1
1.2 1
1.21 1
1.22 1
1.23 1
1.24 1
1.25 1
1.26 1
1.27 1
1.28 1
1.29 1
1.3 1
1.31 1
1.32 1
1.33 1
1.34 1
1.35 1
1.36 1
1.37 1
1.38 1
1.39 1
1.4 1
1.41 1
1.42 1
1.43 1
1.44 1
1.45 1
1.46 1
1.47 1
1.48 1
1.49 1
1.5 1
1.51 1
1.52 1
1.53 1
1.54 1
1.55 1
1.56 1
1.57 1
1.58 1
1.59 1
1.6 1
1.61 1
1.62 1
1.63 1
1.64 1
1.65 1
1.66 0.99
1.67 1
1.68 1
1.69 1
1.7 1
1.71 1
1.72 1
1.73 1
1.74 1
1.75 1
1.76 1
1.77 1
1.78 1
1.79 0.99
1.8 0.99
1.81 1
1.82 1
1.83 0.99
1.84 1
1.85 1
1.86 0.99
1.87 1
1.88 0.99
1.89 1
1.9 1
1.91 0.99
1.92 1
1.93 1
1.94 0.99
1.95 1
1.96 1
1.97 1
1.98 1
1.99 1
2 1
};
\addplot [semithick, color7, opacity=0.75, dashed]
table {%
0.01 0.99
0.02 1
0.03 0.99
0.04 1
0.05 1
0.06 1
0.07 1
0.08 1
0.09 1
0.1 1
0.11 1
0.12 1
0.13 1
0.14 1
0.15 1
0.16 1
0.17 1
0.18 1
0.19 1
0.2 1
0.21 1
0.22 1
0.23 1
0.24 1
0.25 1
0.26 1
0.27 1
0.28 1
0.29 1
0.3 1
0.31 1
0.32 1
0.33 1
0.34 1
0.35 1
0.36 1
0.37 1
0.38 1
0.39 1
0.4 1
0.41 1
0.42 1
0.43 1
0.44 1
0.45 1
0.46 1
0.47 1
0.48 1
0.49 1
0.5 1
0.51 1
0.52 1
0.53 1
0.54 1
0.55 1
0.56 1
0.57 1
0.58 1
0.59 1
0.6 1
0.61 1
0.62 1
0.63 1
0.64 1
0.65 1
0.66 1
0.67 1
0.68 1
0.69 1
0.7 1
0.71 1
0.72 1
0.73 1
0.74 1
0.75 1
0.76 1
0.77 1
0.78 1
0.79 1
0.8 1
0.81 1
0.82 1
0.83 1
0.84 1
0.85 1
0.86 1
0.87 1
0.88 1
0.89 1
0.9 1
0.91 1
0.92 1
0.93 1
0.94 1
0.95 1
0.96 1
0.97 1
0.98 1
0.99 1
1 1
1.02 1
1.03 1
1.04 1
1.05 1
1.06 1
1.07 1
1.08 1
1.09 1
1.1 1
1.11 1
1.12 1
1.13 1
1.14 1
1.15 1
1.16 1
1.17 1
1.18 1
1.19 1
1.2 1
1.21 1
1.22 1
1.23 1
1.24 1
1.25 1
1.26 1
1.27 1
1.28 1
1.29 1
1.3 1
1.31 1
1.32 1
1.33 1
1.34 1
1.35 1
1.36 1
1.37 1
1.38 1
1.39 1
1.4 1
1.41 1
1.42 1
1.43 1
1.44 1
1.45 1
1.46 1
1.47 1
1.48 1
1.49 1
1.5 1
1.51 1
1.52 1
1.53 1
1.54 1
1.55 1
1.56 1
1.57 1
1.58 1
1.59 1
1.6 1
1.61 1
1.62 1
1.63 1
1.64 1
1.65 1
1.66 0.99
1.67 1
1.68 1
1.69 1
1.7 1
1.71 1
1.72 1
1.73 1
1.74 1
1.75 1
1.76 1
1.77 1
1.78 1
1.79 0.99
1.8 0.99
1.81 1
1.82 1
1.83 0.99
1.84 1
1.85 1
1.86 0.99
1.87 1
1.88 0.99
1.89 1
1.9 1
1.91 0.99
1.92 1
1.93 1
1.94 0.99
1.95 1
1.96 1
1.97 1
1.98 1
1.99 1
2 1
};
\addplot [semithick, color8, opacity=0.75, dashed]
table {%
0.01 0.99
0.02 1
0.03 1
0.04 1
0.05 1
0.06 1
0.07 1
0.08 1
0.09 1
0.1 1
0.11 1
0.12 1
0.13 1
0.14 1
0.15 1
0.16 1
0.17 1
0.18 1
0.19 1
0.2 1
0.21 1
0.22 1
0.23 1
0.24 1
0.25 1
0.26 1
0.27 1
0.28 1
0.29 1
0.3 1
0.31 1
0.32 1
0.33 1
0.34 1
0.35 1
0.36 1
0.37 1
0.38 1
0.39 1
0.4 1
0.41 1
0.42 1
0.43 1
0.44 1
0.45 1
0.46 1
0.47 1
0.48 1
0.49 1
0.5 1
0.51 1
0.52 1
0.53 1
0.54 1
0.55 1
0.56 1
0.57 1
0.58 1
0.59 1
0.6 1
0.61 1
0.62 1
0.63 1
0.64 1
0.65 1
0.66 1
0.67 1
0.68 1
0.69 1
0.7 1
0.71 1
0.72 1
0.73 1
0.74 1
0.75 1
0.76 1
0.77 1
0.78 1
0.79 1
0.8 1
0.81 1
0.82 1
0.83 1
0.84 1
0.85 1
0.86 1
0.87 1
0.88 1
0.89 1
0.9 1
0.91 1
0.92 1
0.93 1
0.94 1
0.95 1
0.96 1
0.97 1
0.98 1
0.99 1
1 1
1.02 1
1.03 1
1.04 1
1.05 1
1.06 1
1.07 1
1.08 1
1.09 1
1.1 1
1.11 1
1.12 1
1.13 1
1.14 1
1.15 1
1.16 1
1.17 1
1.18 1
1.19 1
1.2 1
1.21 1
1.22 1
1.23 1
1.24 1
1.25 1
1.26 1
1.27 1
1.28 1
1.29 1
1.3 1
1.31 1
1.32 1
1.33 1
1.34 1
1.35 1
1.36 1
1.37 1
1.38 1
1.39 1
1.4 1
1.41 1
1.42 1
1.43 1
1.44 1
1.45 1
1.46 1
1.47 1
1.48 1
1.49 1
1.5 1
1.51 1
1.52 1
1.53 1
1.54 1
1.55 1
1.56 1
1.57 1
1.58 1
1.59 1
1.6 1
1.61 1
1.62 1
1.63 1
1.64 1
1.65 1
1.66 0.99
1.67 1
1.68 1
1.69 1
1.7 1
1.71 1
1.72 1
1.73 1
1.74 1
1.75 1
1.76 1
1.77 1
1.78 1
1.79 1
1.8 1
1.81 1
1.82 1
1.83 1
1.84 1
1.85 1
1.86 0.99
1.87 1
1.88 0.99
1.89 1
1.9 0.99
1.91 1
1.92 1
1.93 1
1.94 1
1.95 1
1.96 1
1.97 1
1.98 1
1.99 1
2 1
};
\addplot [semithick, color9, opacity=0.75, dashed]
table {%
0.01 0.52
0.02 0.64
0.03 0.62
0.04 0.63
0.05 0.68
0.06 0.69
0.07 0.73
0.08 0.78
0.09 0.71
0.1 0.78
0.11 0.87
0.12 0.89
0.13 0.88
0.14 0.86
0.15 0.85
0.16 0.91
0.17 0.9
0.18 0.94
0.19 0.88
0.2 0.95
0.21 0.91
0.22 0.95
0.23 0.96
0.24 0.95
0.25 0.92
0.26 0.96
0.27 0.95
0.28 0.96
0.29 0.97
0.3 0.98
0.31 0.99
0.32 0.99
0.33 0.99
0.34 0.98
0.35 1
0.36 1
0.37 1
0.38 0.98
0.39 1
0.4 0.99
0.41 1
0.42 1
0.43 1
0.44 1
0.45 1
0.46 0.99
0.47 1
0.48 1
0.49 1
0.5 1
0.51 1
0.52 0.99
0.53 1
0.54 1
0.55 1
0.56 1
0.57 1
0.58 1
0.59 1
0.6 1
0.61 1
0.62 1
0.63 0.99
0.64 1
0.65 1
0.66 1
0.67 1
0.68 1
0.69 1
0.7 1
0.71 1
0.72 1
0.73 1
0.74 1
0.75 1
0.76 1
0.77 1
0.78 1
0.79 1
0.8 1
0.81 1
0.82 1
0.83 1
0.84 1
0.85 1
0.86 1
0.87 1
0.88 1
0.89 1
0.9 1
0.91 1
0.92 1
0.93 1
0.94 1
0.95 1
0.96 1
0.97 1
0.98 1
0.99 1
1 1
1.02 1
1.03 1
1.04 1
1.05 1
1.06 1
1.07 1
1.08 1
1.09 1
1.1 1
1.11 1
1.12 1
1.13 1
1.14 1
1.15 1
1.16 1
1.17 1
1.18 1
1.19 1
1.2 1
1.21 1
1.22 1
1.23 1
1.24 1
1.25 0.99
1.26 1
1.27 1
1.28 0.98
1.29 1
1.3 0.99
1.31 0.98
1.32 0.99
1.33 1
1.34 0.97
1.35 0.97
1.36 0.98
1.37 0.96
1.38 1
1.39 0.96
1.4 0.99
1.41 0.92
1.42 0.97
1.43 0.96
1.44 0.95
1.45 0.97
1.46 0.98
1.47 0.88
1.48 0.96
1.49 0.95
1.5 0.95
1.51 0.95
1.52 0.94
1.53 0.92
1.54 0.92
1.55 0.94
1.56 0.93
1.57 0.94
1.58 0.92
1.59 0.91
1.6 0.93
1.61 0.91
1.62 0.9
1.63 0.9
1.64 0.9
1.65 0.93
1.66 0.85
1.67 0.85
1.68 0.88
1.69 0.91
1.7 0.9
1.71 0.89
1.72 0.85
1.73 0.88
1.74 0.88
1.75 0.78
1.76 0.86
1.77 0.82
1.78 0.83
1.79 0.83
1.8 0.91
1.81 0.89
1.82 0.87
1.83 0.86
1.84 0.88
1.85 0.92
1.86 0.84
1.87 0.84
1.88 0.89
1.89 0.89
1.9 0.91
1.91 0.87
1.92 0.75
1.93 0.78
1.94 0.82
1.95 0.82
1.96 0.8
1.97 0.85
1.98 0.84
1.99 0.79
2 0.85
};
\addplot [semithick, color10, opacity=0.75, dashed]
table {%
0.01 0.51
0.02 0.54
0.03 0.61
0.04 0.65
0.05 0.66
0.06 0.67
0.07 0.67
0.08 0.65
0.09 0.63
0.1 0.79
0.11 0.76
0.12 0.71
0.13 0.82
0.14 0.77
0.15 0.79
0.16 0.76
0.17 0.8
0.18 0.81
0.19 0.8
0.2 0.87
0.21 0.9
0.22 0.86
0.23 0.89
0.24 0.93
0.25 0.91
0.26 0.89
0.27 0.93
0.28 0.87
0.29 0.94
0.3 0.98
0.31 0.97
0.32 0.95
0.33 0.96
0.34 0.94
0.35 1
0.36 0.97
0.37 0.94
0.38 0.96
0.39 0.99
0.4 0.98
0.41 1
0.42 0.98
0.43 0.98
0.44 0.99
0.45 1
0.46 1
0.47 1
0.48 0.99
0.49 0.99
0.5 1
0.51 0.99
0.52 0.98
0.53 1
0.54 1
0.55 1
0.56 1
0.57 1
0.58 0.99
0.59 0.99
0.6 0.99
0.61 1
0.62 1
0.63 1
0.64 1
0.65 1
0.66 1
0.67 1
0.68 1
0.69 1
0.7 0.99
0.71 1
0.72 1
0.73 1
0.74 1
0.75 1
0.76 1
0.77 1
0.78 1
0.79 1
0.8 1
0.81 1
0.82 1
0.83 1
0.84 1
0.85 1
0.86 1
0.87 1
0.88 1
0.89 1
0.9 1
0.91 1
0.92 1
0.93 1
0.94 1
0.95 1
0.96 1
0.97 1
0.98 1
0.99 1
1 1
1.02 1
1.03 1
1.04 1
1.05 1
1.06 1
1.07 1
1.08 1
1.09 1
1.1 1
1.11 1
1.12 1
1.13 1
1.14 1
1.15 1
1.16 1
1.17 0.99
1.18 1
1.19 1
1.2 1
1.21 1
1.22 1
1.23 1
1.24 0.99
1.25 1
1.26 0.99
1.27 0.99
1.28 0.96
1.29 0.97
1.3 0.97
1.31 0.96
1.32 0.98
1.33 0.95
1.34 0.97
1.35 0.92
1.36 0.95
1.37 0.94
1.38 0.96
1.39 0.95
1.4 0.95
1.41 0.91
1.42 0.92
1.43 0.95
1.44 0.92
1.45 0.94
1.46 0.93
1.47 0.82
1.48 0.94
1.49 0.94
1.5 0.9
1.51 0.93
1.52 0.91
1.53 0.87
1.54 0.82
1.55 0.88
1.56 0.86
1.57 0.78
1.58 0.87
1.59 0.87
1.6 0.89
1.61 0.86
1.62 0.79
1.63 0.83
1.64 0.85
1.65 0.87
1.66 0.84
1.67 0.88
1.68 0.8
1.69 0.83
1.7 0.84
1.71 0.85
1.72 0.76
1.73 0.82
1.74 0.86
1.75 0.74
1.76 0.88
1.77 0.8
1.78 0.73
1.79 0.83
1.8 0.75
1.81 0.81
1.82 0.85
1.83 0.8
1.84 0.82
1.85 0.84
1.86 0.83
1.87 0.74
1.88 0.8
1.89 0.79
1.9 0.75
1.91 0.81
1.92 0.78
1.93 0.81
1.94 0.77
1.95 0.77
1.96 0.77
1.97 0.76
1.98 0.85
1.99 0.8
2 0.74
};
\addplot [semithick, color11, opacity=0.75, dashed]
table {%
0.01 0.99
0.02 1
0.03 1
0.04 1
0.05 1
0.06 1
0.07 1
0.08 1
0.09 1
0.1 1
0.11 1
0.12 1
0.13 1
0.14 1
0.15 1
0.16 1
0.17 1
0.18 1
0.19 1
0.2 1
0.21 1
0.22 1
0.23 1
0.24 1
0.25 1
0.26 1
0.27 1
0.28 1
0.29 1
0.3 1
0.31 1
0.32 1
0.33 1
0.34 1
0.35 1
0.36 1
0.37 1
0.38 1
0.39 1
0.4 1
0.41 1
0.42 1
0.43 1
0.44 1
0.45 1
0.46 1
0.47 1
0.48 1
0.49 1
0.5 1
0.51 1
0.52 1
0.53 1
0.54 1
0.55 1
0.56 1
0.57 1
0.58 1
0.59 1
0.6 1
0.61 1
0.62 1
0.63 1
0.64 1
0.65 1
0.66 1
0.67 1
0.68 1
0.69 1
0.7 1
0.71 1
0.72 1
0.73 1
0.74 1
0.75 1
0.76 1
0.77 1
0.78 1
0.79 1
0.8 1
0.81 1
0.82 1
0.83 1
0.84 1
0.85 1
0.86 1
0.87 1
0.88 1
0.89 1
0.9 1
0.91 1
0.92 1
0.93 1
0.94 1
0.95 1
0.96 1
0.97 1
0.98 1
0.99 1
1 1
1.02 1
1.03 1
1.04 1
1.05 1
1.06 1
1.07 1
1.08 1
1.09 1
1.1 1
1.11 1
1.12 1
1.13 1
1.14 1
1.15 1
1.16 1
1.17 1
1.18 1
1.19 1
1.2 1
1.21 1
1.22 1
1.23 1
1.24 1
1.25 1
1.26 1
1.27 1
1.28 1
1.29 1
1.3 1
1.31 1
1.32 1
1.33 1
1.34 1
1.35 1
1.36 1
1.37 1
1.38 1
1.39 1
1.4 1
1.41 1
1.42 1
1.43 1
1.44 1
1.45 1
1.46 1
1.47 1
1.48 1
1.49 1
1.5 1
1.51 1
1.52 1
1.53 1
1.54 1
1.55 1
1.56 1
1.57 1
1.58 1
1.59 1
1.6 1
1.61 1
1.62 1
1.63 1
1.64 1
1.65 1
1.66 0.99
1.67 1
1.68 1
1.69 1
1.7 1
1.71 1
1.72 1
1.73 1
1.74 0.99
1.75 1
1.76 1
1.77 1
1.78 1
1.79 1
1.8 1
1.81 1
1.82 1
1.83 1
1.84 1
1.85 1
1.86 1
1.87 0.99
1.88 0.99
1.89 1
1.9 0.99
1.91 0.99
1.92 1
1.93 1
1.94 1
1.95 1
1.96 1
1.97 1
1.98 1
1.99 1
2 1
};
\addplot [semithick, black, opacity=1, dash pattern=on 1pt off 1pt]
table {%
-0.0895000000000001 0.5
2.0995 0.5
};
\addplot [semithick, black, opacity=1, dash pattern=on 1pt off 1pt]
table {%
-0.0895000000000001 1
2.0995 1
};
\addplot [semithick, black, opacity=1, dash pattern=on 1pt off 1pt]
table {%
1 0.2
1 1.05
};
\end{axis}

\end{tikzpicture}

%% file: plots/decoupled2/NL_GAUxUNI.tex
% This file was created by tikzplotlib v0.9.6.
\begin{tikzpicture}

\definecolor{color0}{rgb}{0.866666666666667,0.494117647058824,0.164705882352941}
\definecolor{color1}{rgb}{0.164705882352941,0.643137254901961,0.866666666666667}
\definecolor{color2}{rgb}{0.584313725490196,0.866666666666667,0.164705882352941}
\definecolor{color3}{rgb}{0.109803921568627,0.337254901960784,0.129411764705882}
\definecolor{color4}{rgb}{0.529411764705882,0.305882352941176,0.858823529411765}
\definecolor{color5}{rgb}{0.858823529411765,0.305882352941176,0.435294117647059}
\definecolor{color6}{rgb}{0.937254901960784,0.929411764705882,0.392156862745098}
\definecolor{color7}{rgb}{0.0901960784313725,0.486274509803922,0.0980392156862745}
\definecolor{color8}{rgb}{0.156862745098039,0.188235294117647,0.827450980392157}
\definecolor{color9}{rgb}{0.937254901960784,0.392156862745098,0.894117647058824}
\definecolor{color10}{rgb}{0.2,0.184313725490196,0.184313725490196}
\definecolor{color11}{rgb}{0.0156862745098039,0.803921568627451,0.976470588235294}

\begin{axis}[
tick align=outside,
tick pos=left,
x grid style={white!69.0196078431373!black},
xmajorgrids,
xmin=-0.0895, xmax=2.0995,
xtick style={color=black},
xtick={0,0.1,0.2,0.3,0.4,0.5,0.6,0.7,0.8,0.9,1,1.1,1.2,1.3,1.4,1.5,1.6,1.7,1.8,1.9,2},
xticklabels={0,,.2,,.4,,.6,,.8,,1,,20,,40,,60,,80,,100},
height=4.8cm,
width=6.5cm,
y grid style={white!69.0196078431373!black},
ymajorgrids,
ymin=0.2, ymax=1.05,
ytick style={color=black}
]
\addplot [semithick, color0, opacity=0.75]
table {%
0.01 0.54
0.02 0.61
0.03 0.56
0.04 0.68
0.05 0.63
0.06 0.74
0.07 0.73
0.08 0.79
0.09 0.81
0.1 0.76
0.11 0.8
0.12 0.85
0.13 0.89
0.14 0.91
0.15 0.87
0.16 0.86
0.17 0.95
0.18 0.98
0.19 0.96
0.2 0.96
0.21 0.96
0.22 0.97
0.23 0.99
0.24 0.98
0.25 0.98
0.26 0.97
0.27 0.98
0.28 0.97
0.29 1
0.3 0.98
0.31 1
0.32 0.99
0.33 1
0.34 0.99
0.35 1
0.36 0.99
0.37 1
0.38 1
0.39 0.99
0.4 0.99
0.41 1
0.42 1
0.43 1
0.44 1
0.45 1
0.46 1
0.47 0.99
0.48 1
0.49 1
0.5 1
0.51 1
0.52 1
0.53 1
0.54 1
0.55 1
0.56 1
0.57 1
0.58 1
0.59 1
0.6 1
0.61 1
0.62 1
0.63 1
0.64 1
0.65 1
0.66 1
0.67 1
0.68 1
0.69 1
0.7 1
0.71 1
0.72 1
0.73 1
0.74 1
0.75 1
0.76 1
0.77 1
0.78 1
0.79 1
0.8 1
0.81 1
0.82 1
0.83 1
0.84 1
0.85 1
0.86 1
0.87 1
0.88 1
0.89 1
0.9 1
0.91 1
0.92 1
0.93 1
0.94 1
0.95 1
0.96 1
0.97 1
0.98 1
0.99 1
1 1
1.02 1
1.03 1
1.04 1
1.05 1
1.06 1
1.07 1
1.08 1
1.09 1
1.1 1
1.11 1
1.12 1
1.13 1
1.14 1
1.15 1
1.16 1
1.17 1
1.18 1
1.19 1
1.2 1
1.21 1
1.22 1
1.23 1
1.24 1
1.25 1
1.26 1
1.27 1
1.28 1
1.29 1
1.3 1
1.31 1
1.32 1
1.33 1
1.34 1
1.35 1
1.36 1
1.37 1
1.38 1
1.39 1
1.4 1
1.41 1
1.42 1
1.43 1
1.44 1
1.45 1
1.46 1
1.47 1
1.48 1
1.49 1
1.5 1
1.51 1
1.52 1
1.53 1
1.54 1
1.55 1
1.56 1
1.57 0.99
1.58 0.98
1.59 1
1.6 0.99
1.61 1
1.62 0.99
1.63 0.98
1.64 1
1.65 1
1.66 0.99
1.67 1
1.68 0.97
1.69 0.99
1.7 0.99
1.71 0.98
1.72 0.97
1.73 0.99
1.74 0.99
1.75 0.99
1.76 1
1.77 0.99
1.78 0.98
1.79 1
1.8 0.99
1.81 0.99
1.82 0.97
1.83 0.98
1.84 0.97
1.85 0.98
1.86 0.99
1.87 0.99
1.88 1
1.89 0.98
1.9 0.93
1.91 0.99
1.92 0.96
1.93 0.97
1.94 0.93
1.95 0.95
1.96 0.96
1.97 0.94
1.98 0.96
1.99 0.94
2 0.97
};
\addplot [semithick, color1, opacity=0.75]
table {%
0.01 0.53
0.02 0.62
0.03 0.53
0.04 0.67
0.05 0.63
0.06 0.72
0.07 0.73
0.08 0.78
0.09 0.78
0.1 0.7
0.11 0.79
0.12 0.84
0.13 0.84
0.14 0.9
0.15 0.84
0.16 0.79
0.17 0.91
0.18 0.92
0.19 0.95
0.2 0.92
0.21 0.93
0.22 0.96
0.23 0.99
0.24 0.98
0.25 0.96
0.26 0.97
0.27 0.96
0.28 1
0.29 1
0.3 0.98
0.31 1
0.32 0.99
0.33 0.98
0.34 0.97
0.35 0.98
0.36 0.99
0.37 0.99
0.38 1
0.39 0.99
0.4 1
0.41 1
0.42 1
0.43 1
0.44 1
0.45 1
0.46 1
0.47 1
0.48 1
0.49 1
0.5 1
0.51 1
0.52 1
0.53 1
0.54 1
0.55 1
0.56 1
0.57 1
0.58 1
0.59 1
0.6 1
0.61 1
0.62 1
0.63 1
0.64 1
0.65 1
0.66 1
0.67 1
0.68 1
0.69 1
0.7 1
0.71 1
0.72 1
0.73 1
0.74 1
0.75 1
0.76 1
0.77 1
0.78 1
0.79 1
0.8 1
0.81 1
0.82 1
0.83 1
0.84 1
0.85 1
0.86 1
0.87 1
0.88 1
0.89 1
0.9 1
0.91 1
0.92 1
0.93 1
0.94 1
0.95 1
0.96 1
0.97 1
0.98 1
0.99 1
1 1
1.02 1
1.03 1
1.04 1
1.05 1
1.06 1
1.07 1
1.08 1
1.09 1
1.1 1
1.11 1
1.12 1
1.13 1
1.14 1
1.15 1
1.16 1
1.17 1
1.18 1
1.19 1
1.2 1
1.21 1
1.22 1
1.23 1
1.24 1
1.25 1
1.26 1
1.27 1
1.28 1
1.29 1
1.3 1
1.31 1
1.32 1
1.33 1
1.34 1
1.35 1
1.36 1
1.37 1
1.38 1
1.39 1
1.4 1
1.41 1
1.42 1
1.43 0.99
1.44 1
1.45 1
1.46 1
1.47 1
1.48 0.99
1.49 1
1.5 0.99
1.51 0.99
1.52 1
1.53 0.99
1.54 0.99
1.55 0.99
1.56 1
1.57 0.99
1.58 1
1.59 0.99
1.6 0.98
1.61 1
1.62 0.99
1.63 0.97
1.64 0.99
1.65 1
1.66 0.93
1.67 0.98
1.68 0.95
1.69 0.97
1.7 0.97
1.71 0.96
1.72 0.94
1.73 0.94
1.74 0.95
1.75 0.96
1.76 0.97
1.77 0.95
1.78 0.93
1.79 0.95
1.8 0.97
1.81 0.94
1.82 0.94
1.83 0.96
1.84 0.95
1.85 0.92
1.86 0.93
1.87 0.94
1.88 0.87
1.89 0.91
1.9 0.89
1.91 0.96
1.92 0.91
1.93 0.89
1.94 0.9
1.95 0.89
1.96 0.86
1.97 0.91
1.98 0.88
1.99 0.84
2 0.88
};
\addplot [semithick, color2, opacity=0.75]
table {%
0.01 0.55
0.02 0.57
0.03 0.53
0.04 0.58
0.05 0.59
0.06 0.72
0.07 0.63
0.08 0.75
0.09 0.77
0.1 0.68
0.11 0.71
0.12 0.79
0.13 0.79
0.14 0.79
0.15 0.78
0.16 0.73
0.17 0.88
0.18 0.92
0.19 0.94
0.2 0.92
0.21 0.93
0.22 0.95
0.23 0.99
0.24 0.98
0.25 0.96
0.26 0.97
0.27 0.96
0.28 1
0.29 1
0.3 0.98
0.31 1
0.32 0.98
0.33 0.98
0.34 0.97
0.35 0.98
0.36 0.99
0.37 0.99
0.38 1
0.39 0.99
0.4 1
0.41 1
0.42 1
0.43 1
0.44 1
0.45 1
0.46 1
0.47 1
0.48 1
0.49 1
0.5 1
0.51 1
0.52 1
0.53 1
0.54 1
0.55 1
0.56 1
0.57 1
0.58 1
0.59 1
0.6 1
0.61 1
0.62 1
0.63 1
0.64 1
0.65 1
0.66 1
0.67 1
0.68 1
0.69 1
0.7 1
0.71 1
0.72 1
0.73 1
0.74 1
0.75 1
0.76 1
0.77 1
0.78 1
0.79 1
0.8 1
0.81 1
0.82 1
0.83 1
0.84 1
0.85 1
0.86 1
0.87 1
0.88 1
0.89 1
0.9 1
0.91 1
0.92 1
0.93 1
0.94 1
0.95 1
0.96 1
0.97 1
0.98 1
0.99 1
1 1
1.02 1
1.03 1
1.04 1
1.05 1
1.06 1
1.07 1
1.08 1
1.09 1
1.1 1
1.11 1
1.12 1
1.13 1
1.14 1
1.15 1
1.16 1
1.17 1
1.18 1
1.19 1
1.2 1
1.21 1
1.22 1
1.23 1
1.24 1
1.25 1
1.26 1
1.27 1
1.28 1
1.29 1
1.3 1
1.31 1
1.32 1
1.33 1
1.34 1
1.35 1
1.36 1
1.37 1
1.38 1
1.39 1
1.4 1
1.41 1
1.42 1
1.43 0.99
1.44 1
1.45 1
1.46 0.99
1.47 1
1.48 0.99
1.49 1
1.5 0.99
1.51 0.99
1.52 1
1.53 0.99
1.54 0.98
1.55 0.98
1.56 0.98
1.57 0.99
1.58 1
1.59 0.99
1.6 0.99
1.61 1
1.62 0.99
1.63 0.96
1.64 0.96
1.65 1
1.66 0.93
1.67 0.97
1.68 0.96
1.69 0.97
1.7 0.97
1.71 0.97
1.72 0.94
1.73 0.94
1.74 0.94
1.75 0.96
1.76 0.97
1.77 0.96
1.78 0.94
1.79 0.95
1.8 0.96
1.81 0.95
1.82 0.94
1.83 0.97
1.84 0.94
1.85 0.91
1.86 0.94
1.87 0.94
1.88 0.87
1.89 0.94
1.9 0.88
1.91 0.95
1.92 0.91
1.93 0.91
1.94 0.89
1.95 0.9
1.96 0.85
1.97 0.91
1.98 0.86
1.99 0.84
2 0.88
};
\addplot [semithick, color3, opacity=0.75]
table {%
0.01 0.62
0.02 0.75
0.03 0.66
0.04 0.78
0.05 0.72
0.06 0.88
0.07 0.84
0.08 0.93
0.09 0.93
0.1 0.85
0.11 0.9
0.12 0.95
0.13 0.88
0.14 0.94
0.15 0.94
0.16 0.94
0.17 0.94
0.18 0.96
0.19 0.98
0.2 0.95
0.21 0.96
0.22 0.99
0.23 0.97
0.24 0.99
0.25 0.98
0.26 0.99
0.27 0.99
0.28 0.98
0.29 1
0.3 0.98
0.31 1
0.32 0.99
0.33 0.99
0.34 1
0.35 0.99
0.36 1
0.37 1
0.38 0.98
0.39 1
0.4 1
0.41 1
0.42 1
0.43 0.99
0.44 1
0.45 1
0.46 1
0.47 1
0.48 0.99
0.49 1
0.5 0.99
0.51 1
0.52 1
0.53 1
0.54 1
0.55 0.99
0.56 1
0.57 1
0.58 1
0.59 1
0.6 1
0.61 1
0.62 1
0.63 1
0.64 1
0.65 1
0.66 1
0.67 1
0.68 1
0.69 1
0.7 1
0.71 1
0.72 1
0.73 1
0.74 1
0.75 1
0.76 1
0.77 1
0.78 1
0.79 1
0.8 1
0.81 1
0.82 1
0.83 1
0.84 1
0.85 1
0.86 1
0.87 1
0.88 1
0.89 1
0.9 1
0.91 1
0.92 1
0.93 1
0.94 1
0.95 1
0.96 1
0.97 1
0.98 1
0.99 1
1 1
1.02 1
1.03 1
1.04 1
1.05 1
1.06 1
1.07 1
1.08 1
1.09 1
1.1 1
1.11 1
1.12 1
1.13 1
1.14 1
1.15 1
1.16 1
1.17 1
1.18 1
1.19 1
1.2 1
1.21 1
1.22 1
1.23 1
1.24 1
1.25 1
1.26 1
1.27 1
1.28 1
1.29 1
1.3 1
1.31 1
1.32 1
1.33 1
1.34 1
1.35 1
1.36 1
1.37 1
1.38 1
1.39 1
1.4 1
1.41 1
1.42 1
1.43 1
1.44 1
1.45 1
1.46 1
1.47 1
1.48 1
1.49 1
1.5 1
1.51 1
1.52 1
1.53 1
1.54 1
1.55 1
1.56 1
1.57 0.99
1.58 0.99
1.59 0.99
1.6 1
1.61 1
1.62 0.99
1.63 0.99
1.64 1
1.65 1
1.66 0.99
1.67 1
1.68 0.99
1.69 0.99
1.7 1
1.71 0.98
1.72 0.97
1.73 1
1.74 0.99
1.75 1
1.76 1
1.77 0.99
1.78 0.97
1.79 0.99
1.8 1
1.81 0.97
1.82 0.99
1.83 0.99
1.84 0.98
1.85 0.99
1.86 0.99
1.87 0.99
1.88 0.98
1.89 0.99
1.9 0.97
1.91 0.98
1.92 0.97
1.93 0.97
1.94 0.97
1.95 0.96
1.96 0.97
1.97 0.97
1.98 0.98
1.99 0.96
2 0.95
};
\addplot [semithick, color4, opacity=0.75]
table {%
0.01 0.62
0.02 0.75
0.03 0.66
0.04 0.79
0.05 0.72
0.06 0.88
0.07 0.84
0.08 0.93
0.09 0.93
0.1 0.85
0.11 0.9
0.12 0.95
0.13 0.88
0.14 0.94
0.15 0.95
0.16 0.95
0.17 0.94
0.18 0.96
0.19 0.98
0.2 0.96
0.21 0.96
0.22 0.99
0.23 0.97
0.24 0.99
0.25 0.98
0.26 0.99
0.27 0.99
0.28 0.98
0.29 1
0.3 0.98
0.31 1
0.32 0.99
0.33 0.99
0.34 1
0.35 0.99
0.36 1
0.37 1
0.38 0.98
0.39 1
0.4 1
0.41 1
0.42 1
0.43 0.98
0.44 1
0.45 1
0.46 1
0.47 1
0.48 0.99
0.49 1
0.5 0.99
0.51 1
0.52 1
0.53 1
0.54 1
0.55 0.99
0.56 1
0.57 1
0.58 1
0.59 1
0.6 1
0.61 1
0.62 1
0.63 1
0.64 1
0.65 1
0.66 1
0.67 1
0.68 1
0.69 1
0.7 1
0.71 1
0.72 1
0.73 1
0.74 1
0.75 1
0.76 1
0.77 1
0.78 1
0.79 1
0.8 1
0.81 1
0.82 1
0.83 1
0.84 1
0.85 1
0.86 1
0.87 1
0.88 1
0.89 1
0.9 1
0.91 1
0.92 1
0.93 1
0.94 1
0.95 1
0.96 1
0.97 1
0.98 1
0.99 1
1 1
1.02 1
1.03 1
1.04 1
1.05 1
1.06 1
1.07 1
1.08 1
1.09 1
1.1 1
1.11 1
1.12 1
1.13 1
1.14 1
1.15 1
1.16 1
1.17 1
1.18 1
1.19 1
1.2 1
1.21 1
1.22 1
1.23 1
1.24 1
1.25 1
1.26 1
1.27 1
1.28 1
1.29 1
1.3 1
1.31 1
1.32 1
1.33 1
1.34 1
1.35 1
1.36 1
1.37 1
1.38 1
1.39 1
1.4 1
1.41 1
1.42 1
1.43 1
1.44 1
1.45 1
1.46 1
1.47 1
1.48 1
1.49 1
1.5 1
1.51 1
1.52 1
1.53 1
1.54 1
1.55 1
1.56 1
1.57 0.99
1.58 0.99
1.59 0.99
1.6 1
1.61 1
1.62 0.99
1.63 0.99
1.64 1
1.65 1
1.66 0.99
1.67 1
1.68 0.99
1.69 0.99
1.7 1
1.71 0.98
1.72 0.97
1.73 1
1.74 0.99
1.75 1
1.76 1
1.77 0.99
1.78 0.97
1.79 0.99
1.8 1
1.81 0.97
1.82 0.99
1.83 0.99
1.84 0.98
1.85 0.99
1.86 0.99
1.87 0.99
1.88 0.98
1.89 0.99
1.9 0.97
1.91 0.98
1.92 0.97
1.93 0.97
1.94 0.97
1.95 0.96
1.96 0.97
1.97 0.97
1.98 0.98
1.99 0.96
2 0.95
};
\addplot [semithick, color5, opacity=0.75]
table {%
0.01 0.7
0.02 0.88
0.03 0.85
0.04 0.86
0.05 0.86
0.06 0.93
0.07 0.9
0.08 0.96
0.09 0.97
0.1 0.97
0.11 0.91
0.12 0.96
0.13 0.95
0.14 0.98
0.15 0.94
0.16 0.94
0.17 0.96
0.18 1
0.19 0.98
0.2 0.97
0.21 0.99
0.22 1
0.23 1
0.24 1
0.25 0.99
0.26 1
0.27 0.98
0.28 0.99
0.29 1
0.3 0.99
0.31 1
0.32 0.99
0.33 1
0.34 0.99
0.35 0.99
0.36 0.99
0.37 1
0.38 0.98
0.39 1
0.4 1
0.41 1
0.42 1
0.43 1
0.44 1
0.45 1
0.46 1
0.47 1
0.48 0.99
0.49 1
0.5 1
0.51 1
0.52 1
0.53 1
0.54 1
0.55 1
0.56 1
0.57 1
0.58 1
0.59 1
0.6 1
0.61 1
0.62 1
0.63 1
0.64 1
0.65 1
0.66 1
0.67 1
0.68 1
0.69 1
0.7 1
0.71 1
0.72 1
0.73 1
0.74 1
0.75 1
0.76 1
0.77 1
0.78 1
0.79 1
0.8 1
0.81 1
0.82 1
0.83 1
0.84 1
0.85 1
0.86 1
0.87 1
0.88 1
0.89 1
0.9 1
0.91 1
0.92 1
0.93 1
0.94 1
0.95 1
0.96 1
0.97 1
0.98 1
0.99 1
1 1
1.02 1
1.03 1
1.04 1
1.05 1
1.06 1
1.07 1
1.08 1
1.09 1
1.1 1
1.11 1
1.12 1
1.13 1
1.14 1
1.15 1
1.16 1
1.17 1
1.18 1
1.19 1
1.2 1
1.21 1
1.22 1
1.23 1
1.24 1
1.25 1
1.26 1
1.27 1
1.28 1
1.29 1
1.3 1
1.31 1
1.32 1
1.33 1
1.34 1
1.35 0.99
1.36 1
1.37 1
1.38 1
1.39 1
1.4 1
1.41 1
1.42 1
1.43 1
1.44 1
1.45 1
1.46 1
1.47 1
1.48 1
1.49 1
1.5 1
1.51 1
1.52 1
1.53 1
1.54 1
1.55 1
1.56 1
1.57 0.99
1.58 0.99
1.59 0.99
1.6 1
1.61 1
1.62 0.99
1.63 0.99
1.64 1
1.65 1
1.66 0.99
1.67 1
1.68 1
1.69 0.99
1.7 1
1.71 0.99
1.72 0.98
1.73 1
1.74 0.99
1.75 1
1.76 1
1.77 0.99
1.78 0.98
1.79 0.99
1.8 1
1.81 0.98
1.82 1
1.83 1
1.84 0.99
1.85 0.99
1.86 1
1.87 0.99
1.88 0.98
1.89 1
1.9 0.98
1.91 1
1.92 0.99
1.93 0.98
1.94 0.99
1.95 0.96
1.96 1
1.97 0.99
1.98 0.98
1.99 0.98
2 0.96
};
\addplot [semithick, color6, opacity=0.75, dashed]
table {%
0.01 1
0.02 1
0.03 1
0.04 0.99
0.05 1
0.06 1
0.07 1
0.08 1
0.09 1
0.1 1
0.11 1
0.12 1
0.13 1
0.14 1
0.15 1
0.16 1
0.17 1
0.18 1
0.19 1
0.2 1
0.21 1
0.22 1
0.23 1
0.24 1
0.25 1
0.26 1
0.27 1
0.28 1
0.29 1
0.3 1
0.31 1
0.32 1
0.33 1
0.34 1
0.35 1
0.36 1
0.37 1
0.38 1
0.39 1
0.4 1
0.41 1
0.42 1
0.43 1
0.44 1
0.45 1
0.46 1
0.47 1
0.48 1
0.49 1
0.5 1
0.51 1
0.52 1
0.53 1
0.54 1
0.55 1
0.56 1
0.57 1
0.58 1
0.59 1
0.6 1
0.61 1
0.62 1
0.63 1
0.64 1
0.65 1
0.66 1
0.67 1
0.68 1
0.69 1
0.7 1
0.71 1
0.72 1
0.73 1
0.74 1
0.75 1
0.76 1
0.77 1
0.78 1
0.79 1
0.8 1
0.81 1
0.82 1
0.83 1
0.84 1
0.85 1
0.86 1
0.87 1
0.88 1
0.89 1
0.9 1
0.91 1
0.92 1
0.93 1
0.94 1
0.95 1
0.96 1
0.97 1
0.98 1
0.99 1
1 1
1.02 1
1.03 1
1.04 1
1.05 1
1.06 1
1.07 1
1.08 1
1.09 1
1.1 1
1.11 1
1.12 1
1.13 1
1.14 1
1.15 1
1.16 1
1.17 1
1.18 1
1.19 1
1.2 1
1.21 1
1.22 1
1.23 1
1.24 1
1.25 1
1.26 1
1.27 1
1.28 1
1.29 1
1.3 1
1.31 1
1.32 1
1.33 1
1.34 1
1.35 1
1.36 1
1.37 1
1.38 1
1.39 1
1.4 1
1.41 1
1.42 1
1.43 1
1.44 1
1.45 1
1.46 1
1.47 1
1.48 1
1.49 1
1.5 1
1.51 1
1.52 1
1.53 1
1.54 1
1.55 1
1.56 1
1.57 1
1.58 1
1.59 1
1.6 1
1.61 1
1.62 1
1.63 1
1.64 1
1.65 1
1.66 1
1.67 1
1.68 1
1.69 1
1.7 1
1.71 1
1.72 1
1.73 1
1.74 1
1.75 1
1.76 1
1.77 1
1.78 1
1.79 1
1.8 1
1.81 1
1.82 1
1.83 0.99
1.84 1
1.85 1
1.86 1
1.87 1
1.88 1
1.89 1
1.9 1
1.91 1
1.92 1
1.93 1
1.94 1
1.95 1
1.96 1
1.97 1
1.98 1
1.99 1
2 1
};
\addplot [semithick, color7, opacity=0.75, dashed]
table {%
0.01 1
0.02 1
0.03 1
0.04 0.99
0.05 1
0.06 1
0.07 1
0.08 1
0.09 1
0.1 1
0.11 1
0.12 1
0.13 1
0.14 1
0.15 1
0.16 1
0.17 1
0.18 1
0.19 1
0.2 1
0.21 1
0.22 1
0.23 1
0.24 1
0.25 1
0.26 1
0.27 1
0.28 1
0.29 1
0.3 1
0.31 1
0.32 1
0.33 1
0.34 1
0.35 1
0.36 1
0.37 1
0.38 1
0.39 1
0.4 1
0.41 1
0.42 1
0.43 1
0.44 1
0.45 1
0.46 1
0.47 1
0.48 1
0.49 1
0.5 1
0.51 1
0.52 1
0.53 1
0.54 1
0.55 1
0.56 1
0.57 1
0.58 1
0.59 1
0.6 1
0.61 1
0.62 1
0.63 1
0.64 1
0.65 1
0.66 1
0.67 1
0.68 1
0.69 1
0.7 1
0.71 1
0.72 1
0.73 1
0.74 1
0.75 1
0.76 1
0.77 1
0.78 1
0.79 1
0.8 1
0.81 1
0.82 1
0.83 1
0.84 1
0.85 1
0.86 1
0.87 1
0.88 1
0.89 1
0.9 1
0.91 1
0.92 1
0.93 1
0.94 1
0.95 1
0.96 1
0.97 1
0.98 1
0.99 1
1 1
1.02 1
1.03 1
1.04 1
1.05 1
1.06 1
1.07 1
1.08 1
1.09 1
1.1 1
1.11 1
1.12 1
1.13 1
1.14 1
1.15 1
1.16 1
1.17 1
1.18 1
1.19 1
1.2 1
1.21 1
1.22 1
1.23 1
1.24 1
1.25 1
1.26 1
1.27 1
1.28 1
1.29 1
1.3 1
1.31 1
1.32 1
1.33 1
1.34 1
1.35 1
1.36 1
1.37 1
1.38 1
1.39 1
1.4 1
1.41 1
1.42 1
1.43 1
1.44 1
1.45 1
1.46 1
1.47 1
1.48 1
1.49 1
1.5 1
1.51 1
1.52 1
1.53 1
1.54 1
1.55 1
1.56 1
1.57 1
1.58 1
1.59 1
1.6 1
1.61 1
1.62 1
1.63 1
1.64 1
1.65 1
1.66 1
1.67 1
1.68 1
1.69 1
1.7 1
1.71 1
1.72 1
1.73 1
1.74 1
1.75 1
1.76 1
1.77 1
1.78 1
1.79 1
1.8 1
1.81 1
1.82 1
1.83 0.99
1.84 1
1.85 1
1.86 1
1.87 1
1.88 1
1.89 1
1.9 1
1.91 1
1.92 1
1.93 1
1.94 1
1.95 1
1.96 1
1.97 1
1.98 1
1.99 1
2 1
};
\addplot [semithick, color8, opacity=0.75, dashed]
table {%
0.01 1
0.02 1
0.03 1
0.04 1
0.05 1
0.06 1
0.07 1
0.08 1
0.09 1
0.1 1
0.11 1
0.12 1
0.13 1
0.14 1
0.15 1
0.16 1
0.17 1
0.18 1
0.19 1
0.2 1
0.21 1
0.22 1
0.23 1
0.24 1
0.25 1
0.26 1
0.27 1
0.28 1
0.29 1
0.3 1
0.31 1
0.32 1
0.33 1
0.34 1
0.35 1
0.36 1
0.37 1
0.38 1
0.39 1
0.4 1
0.41 1
0.42 1
0.43 1
0.44 1
0.45 1
0.46 1
0.47 1
0.48 1
0.49 1
0.5 1
0.51 1
0.52 1
0.53 1
0.54 1
0.55 1
0.56 1
0.57 1
0.58 1
0.59 1
0.6 1
0.61 1
0.62 1
0.63 1
0.64 1
0.65 1
0.66 1
0.67 1
0.68 1
0.69 1
0.7 1
0.71 1
0.72 1
0.73 1
0.74 1
0.75 1
0.76 1
0.77 1
0.78 1
0.79 1
0.8 1
0.81 1
0.82 1
0.83 1
0.84 1
0.85 1
0.86 1
0.87 1
0.88 1
0.89 1
0.9 1
0.91 1
0.92 1
0.93 1
0.94 1
0.95 1
0.96 1
0.97 1
0.98 1
0.99 1
1 1
1.02 1
1.03 1
1.04 1
1.05 1
1.06 1
1.07 1
1.08 1
1.09 1
1.1 1
1.11 1
1.12 1
1.13 1
1.14 1
1.15 1
1.16 1
1.17 1
1.18 1
1.19 1
1.2 1
1.21 1
1.22 1
1.23 1
1.24 1
1.25 1
1.26 1
1.27 1
1.28 1
1.29 1
1.3 1
1.31 1
1.32 1
1.33 1
1.34 1
1.35 1
1.36 1
1.37 1
1.38 1
1.39 1
1.4 1
1.41 1
1.42 1
1.43 1
1.44 1
1.45 1
1.46 1
1.47 1
1.48 1
1.49 1
1.5 1
1.51 1
1.52 1
1.53 1
1.54 1
1.55 1
1.56 1
1.57 1
1.58 1
1.59 1
1.6 1
1.61 1
1.62 1
1.63 1
1.64 1
1.65 1
1.66 1
1.67 1
1.68 1
1.69 1
1.7 1
1.71 1
1.72 1
1.73 1
1.74 1
1.75 1
1.76 1
1.77 1
1.78 1
1.79 1
1.8 1
1.81 1
1.82 1
1.83 1
1.84 1
1.85 1
1.86 1
1.87 1
1.88 1
1.89 1
1.9 1
1.91 1
1.92 1
1.93 1
1.94 1
1.95 1
1.96 1
1.97 1
1.98 1
1.99 1
2 1
};
\addplot [semithick, color9, opacity=0.75, dashed]
table {%
0.01 0.54
0.02 0.49
0.03 0.52
0.04 0.66
0.05 0.58
0.06 0.72
0.07 0.63
0.08 0.68
0.09 0.75
0.1 0.71
0.11 0.71
0.12 0.71
0.13 0.83
0.14 0.75
0.15 0.76
0.16 0.75
0.17 0.79
0.18 0.79
0.19 0.83
0.2 0.77
0.21 0.84
0.22 0.82
0.23 0.84
0.24 0.88
0.25 0.88
0.26 0.87
0.27 0.9
0.28 0.86
0.29 0.93
0.3 0.86
0.31 0.88
0.32 0.9
0.33 0.91
0.34 0.9
0.35 0.95
0.36 0.93
0.37 0.94
0.38 0.94
0.39 0.92
0.4 0.91
0.41 0.93
0.42 0.91
0.43 0.96
0.44 0.95
0.45 0.95
0.46 0.96
0.47 0.97
0.48 0.98
0.49 0.97
0.5 0.98
0.51 0.97
0.52 0.97
0.53 0.95
0.54 0.98
0.55 0.97
0.56 0.97
0.57 0.96
0.58 1
0.59 0.99
0.6 1
0.61 0.96
0.62 0.99
0.63 0.97
0.64 1
0.65 0.93
0.66 0.98
0.67 0.99
0.68 1
0.69 1
0.7 1
0.71 0.99
0.72 0.99
0.73 1
0.74 1
0.75 1
0.76 1
0.77 1
0.78 0.97
0.79 1
0.8 1
0.81 0.99
0.82 0.98
0.83 0.99
0.84 0.99
0.85 1
0.86 1
0.87 0.99
0.88 1
0.89 0.99
0.9 1
0.91 1
0.92 0.99
0.93 1
0.94 1
0.95 1
0.96 1
0.97 1
0.98 1
0.99 1
1 1
1.02 1
1.03 1
1.04 1
1.05 1
1.06 1
1.07 1
1.08 1
1.09 1
1.1 1
1.11 1
1.12 1
1.13 1
1.14 1
1.15 1
1.16 1
1.17 1
1.18 1
1.19 1
1.2 1
1.21 1
1.22 1
1.23 1
1.24 1
1.25 1
1.26 1
1.27 1
1.28 1
1.29 1
1.3 1
1.31 1
1.32 1
1.33 1
1.34 0.99
1.35 0.99
1.36 1
1.37 1
1.38 0.98
1.39 1
1.4 1
1.41 0.99
1.42 0.98
1.43 0.98
1.44 1
1.45 1
1.46 0.99
1.47 1
1.48 1
1.49 1
1.5 1
1.51 0.99
1.52 0.99
1.53 1
1.54 1
1.55 0.98
1.56 0.97
1.57 0.99
1.58 0.98
1.59 0.98
1.6 1
1.61 0.97
1.62 0.99
1.63 0.94
1.64 0.99
1.65 0.97
1.66 0.95
1.67 0.99
1.68 0.97
1.69 0.96
1.7 0.97
1.71 0.95
1.72 0.95
1.73 0.96
1.74 0.94
1.75 0.93
1.76 0.97
1.77 0.97
1.78 0.91
1.79 0.97
1.8 0.95
1.81 0.96
1.82 0.95
1.83 0.93
1.84 0.96
1.85 0.86
1.86 0.95
1.87 0.95
1.88 0.92
1.89 0.96
1.9 0.92
1.91 0.9
1.92 0.91
1.93 0.9
1.94 0.9
1.95 0.93
1.96 0.91
1.97 0.95
1.98 0.93
1.99 0.88
2 0.87
};
\addplot [semithick, color10, opacity=0.75, dashed]
table {%
0.01 0.52
0.02 0.52
0.03 0.51
0.04 0.58
0.05 0.5
0.06 0.58
0.07 0.58
0.08 0.67
0.09 0.65
0.1 0.59
0.11 0.67
0.12 0.7
0.13 0.74
0.14 0.69
0.15 0.75
0.16 0.64
0.17 0.73
0.18 0.69
0.19 0.73
0.2 0.76
0.21 0.76
0.22 0.68
0.23 0.77
0.24 0.85
0.25 0.69
0.26 0.79
0.27 0.85
0.28 0.81
0.29 0.82
0.3 0.84
0.31 0.84
0.32 0.85
0.33 0.84
0.34 0.89
0.35 0.89
0.36 0.91
0.37 0.93
0.38 0.85
0.39 0.85
0.4 0.9
0.41 0.96
0.42 0.89
0.43 0.9
0.44 0.95
0.45 0.94
0.46 0.96
0.47 0.91
0.48 0.91
0.49 0.91
0.5 0.93
0.51 0.95
0.52 0.94
0.53 0.92
0.54 0.97
0.55 0.99
0.56 0.97
0.57 0.96
0.58 0.96
0.59 0.99
0.6 0.98
0.61 0.96
0.62 0.98
0.63 0.95
0.64 0.99
0.65 0.96
0.66 0.98
0.67 1
0.68 0.98
0.69 0.98
0.7 1
0.71 1
0.72 0.98
0.73 0.99
0.74 0.97
0.75 1
0.76 0.97
0.77 1
0.78 1
0.79 0.99
0.8 1
0.81 1
0.82 0.99
0.83 0.99
0.84 1
0.85 1
0.86 1
0.87 1
0.88 1
0.89 1
0.9 0.99
0.91 1
0.92 1
0.93 0.98
0.94 0.99
0.95 1
0.96 1
0.97 0.99
0.98 1
0.99 1
1 1
1.02 1
1.03 1
1.04 1
1.05 1
1.06 1
1.07 1
1.08 1
1.09 1
1.1 1
1.11 1
1.12 1
1.13 1
1.14 1
1.15 1
1.16 1
1.17 1
1.18 1
1.19 1
1.2 1
1.21 1
1.22 1
1.23 1
1.24 1
1.25 1
1.26 1
1.27 1
1.28 1
1.29 1
1.3 1
1.31 1
1.32 1
1.33 1
1.34 1
1.35 1
1.36 1
1.37 1
1.38 1
1.39 1
1.4 1
1.41 1
1.42 0.99
1.43 1
1.44 1
1.45 1
1.46 1
1.47 1
1.48 0.98
1.49 1
1.5 0.97
1.51 0.99
1.52 0.98
1.53 0.97
1.54 0.97
1.55 0.98
1.56 0.95
1.57 0.99
1.58 1
1.59 0.96
1.6 0.97
1.61 1
1.62 0.96
1.63 0.98
1.64 0.96
1.65 0.95
1.66 0.94
1.67 0.97
1.68 0.95
1.69 0.95
1.7 0.98
1.71 0.94
1.72 0.89
1.73 0.94
1.74 0.96
1.75 0.93
1.76 0.98
1.77 0.93
1.78 0.94
1.79 0.92
1.8 0.94
1.81 0.91
1.82 0.91
1.83 0.91
1.84 0.94
1.85 0.92
1.86 0.97
1.87 0.93
1.88 0.91
1.89 0.94
1.9 0.95
1.91 0.88
1.92 0.88
1.93 0.88
1.94 0.86
1.95 0.89
1.96 0.89
1.97 0.94
1.98 0.88
1.99 0.88
2 0.91
};
\addplot [semithick, color11, opacity=0.75, dashed]
table {%
0.01 1
0.02 1
0.03 0.99
0.04 1
0.05 1
0.06 1
0.07 1
0.08 1
0.09 1
0.1 1
0.11 1
0.12 1
0.13 1
0.14 1
0.15 1
0.16 1
0.17 1
0.18 1
0.19 1
0.2 1
0.21 1
0.22 1
0.23 1
0.24 1
0.25 1
0.26 1
0.27 1
0.28 1
0.29 1
0.3 1
0.31 1
0.32 1
0.33 1
0.34 1
0.35 1
0.36 1
0.37 1
0.38 1
0.39 1
0.4 1
0.41 1
0.42 1
0.43 1
0.44 1
0.45 1
0.46 1
0.47 1
0.48 1
0.49 1
0.5 1
0.51 1
0.52 1
0.53 1
0.54 1
0.55 1
0.56 1
0.57 1
0.58 1
0.59 1
0.6 1
0.61 1
0.62 1
0.63 1
0.64 1
0.65 1
0.66 1
0.67 1
0.68 1
0.69 1
0.7 1
0.71 1
0.72 1
0.73 1
0.74 1
0.75 1
0.76 1
0.77 1
0.78 1
0.79 1
0.8 1
0.81 1
0.82 1
0.83 1
0.84 1
0.85 1
0.86 1
0.87 1
0.88 1
0.89 1
0.9 1
0.91 1
0.92 1
0.93 1
0.94 1
0.95 1
0.96 1
0.97 1
0.98 1
0.99 1
1 1
1.02 1
1.03 1
1.04 1
1.05 1
1.06 1
1.07 1
1.08 1
1.09 1
1.1 1
1.11 1
1.12 1
1.13 1
1.14 1
1.15 1
1.16 1
1.17 1
1.18 1
1.19 1
1.2 1
1.21 1
1.22 1
1.23 1
1.24 1
1.25 1
1.26 1
1.27 1
1.28 1
1.29 1
1.3 1
1.31 1
1.32 1
1.33 1
1.34 1
1.35 1
1.36 1
1.37 1
1.38 1
1.39 1
1.4 1
1.41 1
1.42 1
1.43 1
1.44 1
1.45 1
1.46 1
1.47 1
1.48 1
1.49 1
1.5 1
1.51 1
1.52 1
1.53 1
1.54 1
1.55 1
1.56 1
1.57 1
1.58 1
1.59 1
1.6 1
1.61 1
1.62 1
1.63 1
1.64 1
1.65 1
1.66 1
1.67 1
1.68 1
1.69 1
1.7 1
1.71 1
1.72 1
1.73 1
1.74 1
1.75 1
1.76 1
1.77 1
1.78 1
1.79 1
1.8 1
1.81 1
1.82 1
1.83 1
1.84 1
1.85 1
1.86 1
1.87 1
1.88 1
1.89 0.99
1.9 1
1.91 1
1.92 1
1.93 0.99
1.94 1
1.95 1
1.96 1
1.97 1
1.98 1
1.99 0.99
2 1
};
\addplot [semithick, black, opacity=1, dash pattern=on 1pt off 1pt]
table {%
-0.0895000000000001 0.5
2.0995 0.5
};
\addplot [semithick, black, opacity=1, dash pattern=on 1pt off 1pt]
table {%
-0.0895000000000001 1
2.0995 1
};
\addplot [semithick, black, opacity=1, dash pattern=on 1pt off 1pt]
table {%
1 0.2
1 1.05
};
\end{axis}

\end{tikzpicture}

%% file: plots/decoupled2/NL_GAUxLAP.tex
% This file was created by tikzplotlib v0.9.6.
\begin{tikzpicture}

\definecolor{color0}{rgb}{0.866666666666667,0.494117647058824,0.164705882352941}
\definecolor{color1}{rgb}{0.164705882352941,0.643137254901961,0.866666666666667}
\definecolor{color2}{rgb}{0.584313725490196,0.866666666666667,0.164705882352941}
\definecolor{color3}{rgb}{0.109803921568627,0.337254901960784,0.129411764705882}
\definecolor{color4}{rgb}{0.529411764705882,0.305882352941176,0.858823529411765}
\definecolor{color5}{rgb}{0.858823529411765,0.305882352941176,0.435294117647059}
\definecolor{color6}{rgb}{0.937254901960784,0.929411764705882,0.392156862745098}
\definecolor{color7}{rgb}{0.0901960784313725,0.486274509803922,0.0980392156862745}
\definecolor{color8}{rgb}{0.156862745098039,0.188235294117647,0.827450980392157}
\definecolor{color9}{rgb}{0.937254901960784,0.392156862745098,0.894117647058824}
\definecolor{color10}{rgb}{0.2,0.184313725490196,0.184313725490196}
\definecolor{color11}{rgb}{0.0156862745098039,0.803921568627451,0.976470588235294}

\begin{axis}[
tick align=outside,
tick pos=left,
x grid style={white!69.0196078431373!black},
xmajorgrids,
xmin=-0.0895, xmax=2.0995,
xtick style={color=black},
xtick={0,0.1,0.2,0.3,0.4,0.5,0.6,0.7,0.8,0.9,1,1.1,1.2,1.3,1.4,1.5,1.6,1.7,1.8,1.9,2},
xticklabels={0,,.2,,.4,,.6,,.8,,1,,20,,40,,60,,80,,100},
height=4.8cm,
width=6.5cm,
y grid style={white!69.0196078431373!black},
ymajorgrids,
ymin=0.2, ymax=1.05,
ytick style={color=black}
]
\addplot [semithick, color0, opacity=0.75]
table {%
0.01 0.55
0.02 0.66
0.03 0.58
0.04 0.6
0.05 0.73
0.06 0.8
0.07 0.77
0.08 0.83
0.09 0.83
0.1 0.84
0.11 0.86
0.12 0.91
0.13 0.93
0.14 0.92
0.15 0.92
0.16 1
0.17 0.97
0.18 0.98
0.19 0.97
0.2 0.99
0.21 1
0.22 0.98
0.23 0.98
0.24 0.98
0.25 0.99
0.26 1
0.27 1
0.28 1
0.29 0.99
0.3 1
0.31 1
0.32 1
0.33 1
0.34 1
0.35 1
0.36 1
0.37 1
0.38 1
0.39 1
0.4 1
0.41 1
0.42 1
0.43 1
0.44 1
0.45 1
0.46 1
0.47 1
0.48 1
0.49 1
0.5 1
0.51 1
0.52 1
0.53 1
0.54 1
0.55 1
0.56 1
0.57 1
0.58 1
0.59 1
0.6 1
0.61 1
0.62 1
0.63 1
0.64 1
0.65 1
0.66 1
0.67 1
0.68 1
0.69 1
0.7 1
0.71 1
0.72 1
0.73 1
0.74 1
0.75 1
0.76 1
0.77 1
0.78 1
0.79 1
0.8 1
0.81 1
0.82 1
0.83 1
0.84 1
0.85 1
0.86 1
0.87 1
0.88 1
0.89 1
0.9 1
0.91 1
0.92 1
0.93 1
0.94 1
0.95 1
0.96 1
0.97 1
0.98 1
0.99 1
1 1
1.02 1
1.03 1
1.04 1
1.05 1
1.06 1
1.07 1
1.08 1
1.09 1
1.1 1
1.11 1
1.12 1
1.13 1
1.14 1
1.15 1
1.16 1
1.17 1
1.18 1
1.19 1
1.2 1
1.21 1
1.22 1
1.23 1
1.24 1
1.25 1
1.26 1
1.27 1
1.28 1
1.29 1
1.3 1
1.31 1
1.32 1
1.33 1
1.34 1
1.35 0.99
1.36 0.99
1.37 1
1.38 1
1.39 0.99
1.4 0.99
1.41 0.98
1.42 0.98
1.43 0.96
1.44 0.98
1.45 0.95
1.46 0.97
1.47 0.97
1.48 0.99
1.49 0.92
1.5 0.96
1.51 0.97
1.52 0.97
1.53 0.95
1.54 0.95
1.55 0.95
1.56 0.94
1.57 0.95
1.58 0.96
1.59 0.94
1.6 0.93
1.61 0.92
1.62 0.93
1.63 0.92
1.64 0.96
1.65 0.97
1.66 0.95
1.67 0.94
1.68 0.92
1.69 0.94
1.7 0.92
1.71 0.9
1.72 0.92
1.73 0.88
1.74 0.9
1.75 0.93
1.76 0.88
1.77 0.9
1.78 0.94
1.79 0.87
1.8 0.92
1.81 0.89
1.82 0.86
1.83 0.94
1.84 0.91
1.85 0.89
1.86 0.85
1.87 0.93
1.88 0.89
1.89 0.86
1.9 0.88
1.91 0.94
1.92 0.89
1.93 0.92
1.94 0.9
1.95 0.89
1.96 0.91
1.97 0.89
1.98 0.93
1.99 0.86
2 0.91
};
\addplot [semithick, color1, opacity=0.75]
table {%
0.01 0.58
0.02 0.71
0.03 0.58
0.04 0.61
0.05 0.7
0.06 0.82
0.07 0.82
0.08 0.83
0.09 0.94
0.1 0.88
0.11 0.91
0.12 0.93
0.13 0.96
0.14 0.94
0.15 0.99
0.16 0.99
0.17 0.96
0.18 1
0.19 0.98
0.2 1
0.21 1
0.22 0.98
0.23 0.97
0.24 1
0.25 0.99
0.26 0.98
0.27 1
0.28 1
0.29 1
0.3 1
0.31 1
0.32 0.99
0.33 1
0.34 1
0.35 1
0.36 1
0.37 0.99
0.38 1
0.39 1
0.4 1
0.41 1
0.42 1
0.43 1
0.44 1
0.45 1
0.46 1
0.47 1
0.48 1
0.49 1
0.5 1
0.51 1
0.52 1
0.53 1
0.54 1
0.55 1
0.56 1
0.57 1
0.58 1
0.59 1
0.6 1
0.61 1
0.62 1
0.63 1
0.64 1
0.65 1
0.66 1
0.67 1
0.68 1
0.69 1
0.7 1
0.71 1
0.72 1
0.73 1
0.74 1
0.75 1
0.76 1
0.77 1
0.78 1
0.79 1
0.8 1
0.81 1
0.82 1
0.83 1
0.84 1
0.85 1
0.86 1
0.87 1
0.88 1
0.89 1
0.9 1
0.91 1
0.92 1
0.93 1
0.94 1
0.95 1
0.96 1
0.97 1
0.98 1
0.99 1
1 1
1.02 1
1.03 1
1.04 1
1.05 1
1.06 1
1.07 1
1.08 1
1.09 1
1.1 1
1.11 1
1.12 1
1.13 1
1.14 1
1.15 1
1.16 1
1.17 1
1.18 1
1.19 1
1.2 1
1.21 0.99
1.22 1
1.23 1
1.24 0.99
1.25 1
1.26 0.97
1.27 0.97
1.28 0.99
1.29 0.97
1.3 0.96
1.31 0.96
1.32 0.96
1.33 0.96
1.34 0.92
1.35 0.89
1.36 0.93
1.37 0.95
1.38 0.89
1.39 0.9
1.4 0.93
1.41 0.91
1.42 0.87
1.43 0.84
1.44 0.89
1.45 0.88
1.46 0.89
1.47 0.84
1.48 0.86
1.49 0.81
1.5 0.77
1.51 0.81
1.52 0.81
1.53 0.81
1.54 0.84
1.55 0.79
1.56 0.8
1.57 0.79
1.58 0.82
1.59 0.82
1.6 0.83
1.61 0.81
1.62 0.77
1.63 0.69
1.64 0.81
1.65 0.77
1.66 0.7
1.67 0.75
1.68 0.79
1.69 0.82
1.7 0.75
1.71 0.79
1.72 0.7
1.73 0.7
1.74 0.76
1.75 0.77
1.76 0.73
1.77 0.68
1.78 0.73
1.79 0.75
1.8 0.73
1.81 0.73
1.82 0.76
1.83 0.73
1.84 0.72
1.85 0.67
1.86 0.71
1.87 0.75
1.88 0.77
1.89 0.66
1.9 0.72
1.91 0.73
1.92 0.67
1.93 0.66
1.94 0.76
1.95 0.73
1.96 0.65
1.97 0.78
1.98 0.76
1.99 0.66
2 0.73
};
\addplot [semithick, color2, opacity=0.75]
table {%
0.01 0.51
0.02 0.58
0.03 0.51
0.04 0.51
0.05 0.52
0.06 0.65
0.07 0.66
0.08 0.67
0.09 0.75
0.1 0.83
0.11 0.92
0.12 0.93
0.13 0.95
0.14 0.94
0.15 1
0.16 0.98
0.17 0.96
0.18 0.99
0.19 0.97
0.2 1
0.21 1
0.22 0.98
0.23 0.98
0.24 1
0.25 0.99
0.26 0.98
0.27 1
0.28 1
0.29 1
0.3 1
0.31 1
0.32 0.99
0.33 1
0.34 1
0.35 1
0.36 1
0.37 0.99
0.38 1
0.39 1
0.4 1
0.41 1
0.42 1
0.43 1
0.44 1
0.45 1
0.46 1
0.47 1
0.48 1
0.49 1
0.5 1
0.51 1
0.52 1
0.53 1
0.54 1
0.55 1
0.56 1
0.57 1
0.58 1
0.59 1
0.6 1
0.61 1
0.62 1
0.63 1
0.64 1
0.65 1
0.66 1
0.67 1
0.68 1
0.69 1
0.7 1
0.71 1
0.72 1
0.73 1
0.74 1
0.75 1
0.76 1
0.77 1
0.78 1
0.79 1
0.8 1
0.81 1
0.82 1
0.83 1
0.84 1
0.85 1
0.86 1
0.87 1
0.88 1
0.89 1
0.9 1
0.91 1
0.92 1
0.93 1
0.94 1
0.95 1
0.96 1
0.97 1
0.98 1
0.99 1
1 1
1.02 1
1.03 1
1.04 1
1.05 1
1.06 1
1.07 1
1.08 1
1.09 1
1.1 1
1.11 1
1.12 1
1.13 1
1.14 1
1.15 1
1.16 1
1.17 1
1.18 1
1.19 1
1.2 1
1.21 0.98
1.22 1
1.23 1
1.24 0.99
1.25 1
1.26 0.97
1.27 0.97
1.28 0.99
1.29 0.97
1.3 0.96
1.31 0.97
1.32 0.96
1.33 0.95
1.34 0.91
1.35 0.9
1.36 0.93
1.37 0.95
1.38 0.89
1.39 0.9
1.4 0.93
1.41 0.91
1.42 0.86
1.43 0.85
1.44 0.89
1.45 0.89
1.46 0.87
1.47 0.86
1.48 0.84
1.49 0.8
1.5 0.78
1.51 0.81
1.52 0.8
1.53 0.81
1.54 0.84
1.55 0.8
1.56 0.81
1.57 0.78
1.58 0.83
1.59 0.82
1.6 0.81
1.61 0.81
1.62 0.73
1.63 0.69
1.64 0.8
1.65 0.78
1.66 0.73
1.67 0.75
1.68 0.81
1.69 0.82
1.7 0.76
1.71 0.77
1.72 0.68
1.73 0.73
1.74 0.76
1.75 0.8
1.76 0.72
1.77 0.65
1.78 0.72
1.79 0.74
1.8 0.71
1.81 0.74
1.82 0.74
1.83 0.7
1.84 0.72
1.85 0.65
1.86 0.69
1.87 0.74
1.88 0.78
1.89 0.64
1.9 0.71
1.91 0.75
1.92 0.65
1.93 0.66
1.94 0.74
1.95 0.74
1.96 0.68
1.97 0.77
1.98 0.75
1.99 0.65
2 0.73
};
\addplot [semithick, color3, opacity=0.75]
table {%
0.01 0.68
0.02 0.79
0.03 0.79
0.04 0.74
0.05 0.84
0.06 0.9
0.07 0.91
0.08 0.89
0.09 0.99
0.1 0.97
0.11 0.9
0.12 0.96
0.13 0.97
0.14 0.97
0.15 0.97
0.16 1
0.17 0.99
0.18 0.99
0.19 0.99
0.2 1
0.21 1
0.22 1
0.23 0.98
0.24 1
0.25 1
0.26 1
0.27 1
0.28 1
0.29 1
0.3 1
0.31 1
0.32 0.99
0.33 1
0.34 1
0.35 1
0.36 1
0.37 0.99
0.38 1
0.39 1
0.4 1
0.41 1
0.42 1
0.43 1
0.44 1
0.45 0.99
0.46 1
0.47 1
0.48 1
0.49 1
0.5 1
0.51 1
0.52 1
0.53 1
0.54 1
0.55 1
0.56 1
0.57 1
0.58 1
0.59 1
0.6 1
0.61 1
0.62 1
0.63 1
0.64 1
0.65 1
0.66 1
0.67 1
0.68 1
0.69 1
0.7 1
0.71 1
0.72 1
0.73 1
0.74 1
0.75 1
0.76 1
0.77 1
0.78 1
0.79 1
0.8 1
0.81 1
0.82 1
0.83 1
0.84 1
0.85 1
0.86 1
0.87 1
0.88 1
0.89 1
0.9 1
0.91 1
0.92 1
0.93 1
0.94 1
0.95 1
0.96 1
0.97 1
0.98 1
0.99 1
1 1
1.02 1
1.03 1
1.04 1
1.05 1
1.06 1
1.07 1
1.08 1
1.09 1
1.1 1
1.11 1
1.12 1
1.13 1
1.14 1
1.15 1
1.16 1
1.17 1
1.18 1
1.19 1
1.2 1
1.21 1
1.22 1
1.23 1
1.24 1
1.25 1
1.26 1
1.27 1
1.28 0.99
1.29 0.99
1.3 1
1.31 1
1.32 1
1.33 0.99
1.34 0.99
1.35 1
1.36 0.98
1.37 1
1.38 0.98
1.39 1
1.4 1
1.41 0.99
1.42 0.99
1.43 0.97
1.44 0.99
1.45 0.96
1.46 0.96
1.47 0.99
1.48 0.98
1.49 0.95
1.5 0.99
1.51 0.98
1.52 0.98
1.53 0.96
1.54 0.98
1.55 0.98
1.56 0.99
1.57 0.94
1.58 1
1.59 0.97
1.6 0.98
1.61 0.97
1.62 0.96
1.63 0.95
1.64 0.97
1.65 0.99
1.66 0.97
1.67 0.97
1.68 0.94
1.69 0.95
1.7 0.96
1.71 0.96
1.72 0.95
1.73 0.94
1.74 0.96
1.75 0.97
1.76 0.93
1.77 0.97
1.78 0.93
1.79 0.95
1.8 0.93
1.81 0.94
1.82 0.9
1.83 0.97
1.84 0.93
1.85 0.95
1.86 0.95
1.87 0.97
1.88 0.91
1.89 0.89
1.9 0.92
1.91 0.96
1.92 0.94
1.93 0.94
1.94 0.91
1.95 0.89
1.96 0.93
1.97 0.94
1.98 0.92
1.99 0.93
2 0.97
};
\addplot [semithick, color4, opacity=0.75]
table {%
0.01 0.68
0.02 0.8
0.03 0.79
0.04 0.74
0.05 0.84
0.06 0.9
0.07 0.91
0.08 0.91
0.09 0.99
0.1 0.97
0.11 0.9
0.12 0.97
0.13 0.98
0.14 0.97
0.15 0.97
0.16 1
0.17 0.99
0.18 0.99
0.19 0.99
0.2 1
0.21 1
0.22 1
0.23 0.98
0.24 1
0.25 1
0.26 1
0.27 1
0.28 1
0.29 1
0.3 1
0.31 1
0.32 0.99
0.33 1
0.34 1
0.35 1
0.36 1
0.37 0.99
0.38 1
0.39 1
0.4 1
0.41 1
0.42 1
0.43 1
0.44 1
0.45 0.99
0.46 1
0.47 1
0.48 1
0.49 1
0.5 1
0.51 1
0.52 1
0.53 1
0.54 1
0.55 1
0.56 1
0.57 1
0.58 1
0.59 1
0.6 1
0.61 1
0.62 1
0.63 1
0.64 1
0.65 1
0.66 1
0.67 1
0.68 1
0.69 1
0.7 1
0.71 1
0.72 1
0.73 1
0.74 1
0.75 1
0.76 1
0.77 1
0.78 1
0.79 1
0.8 1
0.81 1
0.82 1
0.83 1
0.84 1
0.85 1
0.86 1
0.87 1
0.88 1
0.89 1
0.9 1
0.91 1
0.92 1
0.93 1
0.94 1
0.95 1
0.96 1
0.97 1
0.98 1
0.99 1
1 1
1.02 1
1.03 1
1.04 1
1.05 1
1.06 1
1.07 1
1.08 1
1.09 1
1.1 1
1.11 1
1.12 1
1.13 1
1.14 1
1.15 1
1.16 1
1.17 1
1.18 1
1.19 1
1.2 1
1.21 1
1.22 1
1.23 1
1.24 1
1.25 1
1.26 1
1.27 1
1.28 0.99
1.29 0.99
1.3 1
1.31 1
1.32 1
1.33 0.99
1.34 0.99
1.35 1
1.36 0.98
1.37 1
1.38 0.98
1.39 1
1.4 1
1.41 0.99
1.42 0.99
1.43 0.97
1.44 0.99
1.45 0.96
1.46 0.97
1.47 0.99
1.48 0.98
1.49 0.95
1.5 0.99
1.51 0.99
1.52 0.98
1.53 0.97
1.54 0.98
1.55 0.98
1.56 0.99
1.57 0.94
1.58 1
1.59 0.97
1.6 0.98
1.61 0.98
1.62 0.96
1.63 0.95
1.64 0.97
1.65 0.99
1.66 0.97
1.67 0.97
1.68 0.94
1.69 0.95
1.7 0.96
1.71 0.96
1.72 0.95
1.73 0.94
1.74 0.96
1.75 0.97
1.76 0.93
1.77 0.99
1.78 0.93
1.79 0.95
1.8 0.93
1.81 0.94
1.82 0.9
1.83 0.97
1.84 0.94
1.85 0.95
1.86 0.96
1.87 0.97
1.88 0.92
1.89 0.89
1.9 0.92
1.91 0.96
1.92 0.94
1.93 0.94
1.94 0.91
1.95 0.89
1.96 0.93
1.97 0.94
1.98 0.92
1.99 0.93
2 0.97
};
\addplot [semithick, color5, opacity=0.75]
table {%
0.01 0.78
0.02 0.85
0.03 0.88
0.04 0.92
0.05 0.91
0.06 0.94
0.07 0.95
0.08 0.94
0.09 0.96
0.1 0.96
0.11 0.96
0.12 0.98
0.13 0.98
0.14 0.96
0.15 0.98
0.16 1
0.17 0.97
0.18 0.98
0.19 0.99
0.2 0.99
0.21 1
0.22 0.98
0.23 0.99
0.24 0.99
0.25 1
0.26 1
0.27 1
0.28 1
0.29 0.99
0.3 1
0.31 1
0.32 0.99
0.33 1
0.34 1
0.35 1
0.36 1
0.37 1
0.38 1
0.39 1
0.4 1
0.41 1
0.42 1
0.43 1
0.44 1
0.45 1
0.46 0.99
0.47 0.99
0.48 1
0.49 1
0.5 1
0.51 1
0.52 1
0.53 1
0.54 1
0.55 1
0.56 1
0.57 1
0.58 1
0.59 1
0.6 1
0.61 1
0.62 1
0.63 1
0.64 1
0.65 1
0.66 1
0.67 1
0.68 1
0.69 1
0.7 1
0.71 1
0.72 1
0.73 1
0.74 1
0.75 1
0.76 1
0.77 1
0.78 1
0.79 1
0.8 0.99
0.81 1
0.82 1
0.83 1
0.84 1
0.85 1
0.86 1
0.87 1
0.88 1
0.89 1
0.9 1
0.91 1
0.92 1
0.93 1
0.94 1
0.95 1
0.96 1
0.97 1
0.98 1
0.99 1
1 1
1.02 1
1.03 1
1.04 1
1.05 1
1.06 1
1.07 1
1.08 1
1.09 1
1.1 1
1.11 1
1.12 1
1.13 1
1.14 1
1.15 1
1.16 1
1.17 1
1.18 1
1.19 1
1.2 0.99
1.21 1
1.22 1
1.23 1
1.24 1
1.25 1
1.26 1
1.27 0.99
1.28 1
1.29 0.99
1.3 1
1.31 0.99
1.32 0.99
1.33 0.99
1.34 0.99
1.35 0.99
1.36 0.98
1.37 1
1.38 0.98
1.39 0.99
1.4 0.98
1.41 0.97
1.42 0.98
1.43 0.97
1.44 0.98
1.45 0.96
1.46 0.97
1.47 0.99
1.48 1
1.49 0.98
1.5 0.99
1.51 0.99
1.52 0.96
1.53 0.97
1.54 1
1.55 0.96
1.56 0.97
1.57 0.95
1.58 0.98
1.59 0.97
1.6 0.99
1.61 0.93
1.62 0.96
1.63 0.98
1.64 0.98
1.65 0.96
1.66 0.97
1.67 0.98
1.68 0.98
1.69 0.97
1.7 0.96
1.71 0.95
1.72 0.96
1.73 0.92
1.74 0.96
1.75 0.99
1.76 0.97
1.77 0.97
1.78 0.97
1.79 0.96
1.8 0.93
1.81 0.94
1.82 0.95
1.83 0.98
1.84 0.98
1.85 0.97
1.86 0.94
1.87 0.97
1.88 0.99
1.89 0.95
1.9 0.95
1.91 0.97
1.92 0.98
1.93 0.95
1.94 0.92
1.95 0.94
1.96 0.93
1.97 0.94
1.98 0.96
1.99 0.98
2 0.97
};
\addplot [semithick, color6, opacity=0.75, dashed]
table {%
0.01 0.98
0.02 0.99
0.03 1
0.04 1
0.05 0.99
0.06 1
0.07 1
0.08 1
0.09 1
0.1 1
0.11 1
0.12 1
0.13 1
0.14 1
0.15 1
0.16 1
0.17 1
0.18 1
0.19 1
0.2 1
0.21 1
0.22 1
0.23 1
0.24 1
0.25 1
0.26 1
0.27 1
0.28 1
0.29 1
0.3 1
0.31 1
0.32 1
0.33 1
0.34 1
0.35 1
0.36 1
0.37 1
0.38 1
0.39 1
0.4 1
0.41 1
0.42 1
0.43 1
0.44 1
0.45 1
0.46 1
0.47 1
0.48 1
0.49 1
0.5 1
0.51 1
0.52 1
0.53 1
0.54 1
0.55 1
0.56 1
0.57 1
0.58 1
0.59 1
0.6 1
0.61 1
0.62 1
0.63 1
0.64 1
0.65 1
0.66 1
0.67 1
0.68 1
0.69 1
0.7 1
0.71 1
0.72 1
0.73 1
0.74 1
0.75 1
0.76 1
0.77 1
0.78 1
0.79 1
0.8 1
0.81 1
0.82 1
0.83 1
0.84 1
0.85 1
0.86 1
0.87 1
0.88 1
0.89 1
0.9 1
0.91 1
0.92 1
0.93 1
0.94 1
0.95 1
0.96 1
0.97 1
0.98 1
0.99 1
1 1
1.02 1
1.03 1
1.04 1
1.05 1
1.06 1
1.07 1
1.08 1
1.09 1
1.1 1
1.11 1
1.12 1
1.13 1
1.14 1
1.15 1
1.16 1
1.17 1
1.18 1
1.19 1
1.2 1
1.21 1
1.22 1
1.23 1
1.24 1
1.25 1
1.26 1
1.27 1
1.28 1
1.29 1
1.3 1
1.31 1
1.32 1
1.33 1
1.34 1
1.35 1
1.36 1
1.37 1
1.38 1
1.39 1
1.4 1
1.41 1
1.42 1
1.43 1
1.44 1
1.45 1
1.46 1
1.47 1
1.48 1
1.49 1
1.5 1
1.51 1
1.52 1
1.53 1
1.54 1
1.55 1
1.56 1
1.57 1
1.58 1
1.59 1
1.6 1
1.61 1
1.62 1
1.63 1
1.64 1
1.65 1
1.66 1
1.67 1
1.68 1
1.69 1
1.7 1
1.71 1
1.72 1
1.73 1
1.74 1
1.75 1
1.76 1
1.77 1
1.78 1
1.79 1
1.8 1
1.81 1
1.82 1
1.83 1
1.84 0.99
1.85 0.99
1.86 1
1.87 0.99
1.88 1
1.89 1
1.9 1
1.91 0.99
1.92 1
1.93 1
1.94 1
1.95 1
1.96 1
1.97 1
1.98 1
1.99 1
2 1
};
\addplot [semithick, color7, opacity=0.75, dashed]
table {%
0.01 0.98
0.02 0.99
0.03 1
0.04 1
0.05 0.99
0.06 1
0.07 1
0.08 1
0.09 1
0.1 1
0.11 1
0.12 1
0.13 1
0.14 1
0.15 1
0.16 1
0.17 1
0.18 1
0.19 1
0.2 1
0.21 1
0.22 1
0.23 1
0.24 1
0.25 1
0.26 1
0.27 1
0.28 1
0.29 1
0.3 1
0.31 1
0.32 1
0.33 1
0.34 1
0.35 1
0.36 1
0.37 1
0.38 1
0.39 1
0.4 1
0.41 1
0.42 1
0.43 1
0.44 1
0.45 1
0.46 1
0.47 1
0.48 1
0.49 1
0.5 1
0.51 1
0.52 1
0.53 1
0.54 1
0.55 1
0.56 1
0.57 1
0.58 1
0.59 1
0.6 1
0.61 1
0.62 1
0.63 1
0.64 1
0.65 1
0.66 1
0.67 1
0.68 1
0.69 1
0.7 1
0.71 1
0.72 1
0.73 1
0.74 1
0.75 1
0.76 1
0.77 1
0.78 1
0.79 1
0.8 1
0.81 1
0.82 1
0.83 1
0.84 1
0.85 1
0.86 1
0.87 1
0.88 1
0.89 1
0.9 1
0.91 1
0.92 1
0.93 1
0.94 1
0.95 1
0.96 1
0.97 1
0.98 1
0.99 1
1 1
1.02 1
1.03 1
1.04 1
1.05 1
1.06 1
1.07 1
1.08 1
1.09 1
1.1 1
1.11 1
1.12 1
1.13 1
1.14 1
1.15 1
1.16 1
1.17 1
1.18 1
1.19 1
1.2 1
1.21 1
1.22 1
1.23 1
1.24 1
1.25 1
1.26 1
1.27 1
1.28 1
1.29 1
1.3 1
1.31 1
1.32 1
1.33 1
1.34 1
1.35 1
1.36 1
1.37 1
1.38 1
1.39 1
1.4 1
1.41 1
1.42 1
1.43 1
1.44 1
1.45 1
1.46 1
1.47 1
1.48 1
1.49 1
1.5 1
1.51 1
1.52 1
1.53 1
1.54 1
1.55 1
1.56 1
1.57 1
1.58 1
1.59 1
1.6 1
1.61 1
1.62 1
1.63 1
1.64 1
1.65 1
1.66 1
1.67 1
1.68 1
1.69 1
1.7 1
1.71 1
1.72 1
1.73 1
1.74 1
1.75 1
1.76 1
1.77 1
1.78 1
1.79 1
1.8 1
1.81 1
1.82 1
1.83 1
1.84 0.99
1.85 0.99
1.86 1
1.87 0.99
1.88 1
1.89 1
1.9 1
1.91 0.99
1.92 1
1.93 1
1.94 1
1.95 1
1.96 1
1.97 1
1.98 1
1.99 1
2 1
};
\addplot [semithick, color8, opacity=0.75, dashed]
table {%
0.01 1
0.02 1
0.03 1
0.04 1
0.05 1
0.06 1
0.07 1
0.08 1
0.09 1
0.1 1
0.11 1
0.12 1
0.13 1
0.14 1
0.15 1
0.16 1
0.17 1
0.18 1
0.19 1
0.2 1
0.21 1
0.22 1
0.23 1
0.24 1
0.25 1
0.26 1
0.27 1
0.28 1
0.29 1
0.3 1
0.31 1
0.32 1
0.33 1
0.34 1
0.35 1
0.36 1
0.37 1
0.38 1
0.39 1
0.4 1
0.41 1
0.42 1
0.43 1
0.44 1
0.45 1
0.46 1
0.47 1
0.48 1
0.49 1
0.5 1
0.51 1
0.52 1
0.53 1
0.54 1
0.55 1
0.56 1
0.57 1
0.58 1
0.59 1
0.6 1
0.61 1
0.62 1
0.63 1
0.64 1
0.65 1
0.66 1
0.67 1
0.68 1
0.69 1
0.7 1
0.71 1
0.72 1
0.73 1
0.74 1
0.75 1
0.76 1
0.77 1
0.78 1
0.79 1
0.8 1
0.81 1
0.82 1
0.83 1
0.84 1
0.85 1
0.86 1
0.87 1
0.88 1
0.89 1
0.9 1
0.91 1
0.92 1
0.93 1
0.94 1
0.95 1
0.96 1
0.97 1
0.98 1
0.99 1
1 1
1.02 1
1.03 1
1.04 1
1.05 1
1.06 1
1.07 1
1.08 1
1.09 1
1.1 1
1.11 1
1.12 1
1.13 1
1.14 1
1.15 1
1.16 1
1.17 1
1.18 1
1.19 1
1.2 1
1.21 1
1.22 1
1.23 1
1.24 1
1.25 1
1.26 1
1.27 1
1.28 1
1.29 1
1.3 1
1.31 1
1.32 1
1.33 1
1.34 1
1.35 1
1.36 1
1.37 1
1.38 1
1.39 1
1.4 1
1.41 1
1.42 1
1.43 1
1.44 1
1.45 1
1.46 1
1.47 1
1.48 1
1.49 1
1.5 1
1.51 1
1.52 1
1.53 1
1.54 1
1.55 1
1.56 1
1.57 1
1.58 1
1.59 1
1.6 1
1.61 1
1.62 1
1.63 1
1.64 1
1.65 1
1.66 1
1.67 1
1.68 1
1.69 1
1.7 1
1.71 0.99
1.72 1
1.73 1
1.74 1
1.75 1
1.76 1
1.77 1
1.78 1
1.79 1
1.8 1
1.81 1
1.82 1
1.83 1
1.84 1
1.85 1
1.86 1
1.87 1
1.88 1
1.89 1
1.9 1
1.91 1
1.92 1
1.93 1
1.94 1
1.95 1
1.96 1
1.97 1
1.98 1
1.99 1
2 1
};
\addplot [semithick, color9, opacity=0.75, dashed]
table {%
0.01 0.53
0.02 0.66
0.03 0.57
0.04 0.64
0.05 0.65
0.06 0.73
0.07 0.65
0.08 0.74
0.09 0.84
0.1 0.85
0.11 0.83
0.12 0.88
0.13 0.86
0.14 0.93
0.15 0.85
0.16 0.87
0.17 0.97
0.18 0.9
0.19 0.96
0.2 0.94
0.21 0.98
0.22 1
0.23 0.93
0.24 0.93
0.25 0.94
0.26 0.99
0.27 0.99
0.28 1
0.29 0.99
0.3 0.98
0.31 0.99
0.32 0.97
0.33 1
0.34 0.99
0.35 0.98
0.36 0.99
0.37 0.98
0.38 0.98
0.39 1
0.4 1
0.41 1
0.42 1
0.43 1
0.44 1
0.45 0.99
0.46 1
0.47 1
0.48 1
0.49 1
0.5 0.99
0.51 1
0.52 1
0.53 1
0.54 1
0.55 1
0.56 1
0.57 1
0.58 0.99
0.59 1
0.6 1
0.61 1
0.62 1
0.63 1
0.64 1
0.65 1
0.66 1
0.67 1
0.68 1
0.69 1
0.7 1
0.71 1
0.72 1
0.73 1
0.74 1
0.75 1
0.76 1
0.77 1
0.78 1
0.79 1
0.8 1
0.81 1
0.82 1
0.83 1
0.84 1
0.85 1
0.86 1
0.87 1
0.88 1
0.89 1
0.9 1
0.91 1
0.92 1
0.93 1
0.94 1
0.95 1
0.96 1
0.97 1
0.98 1
0.99 1
1 1
1.02 1
1.03 1
1.04 1
1.05 1
1.06 1
1.07 1
1.08 1
1.09 1
1.1 1
1.11 1
1.12 1
1.13 1
1.14 1
1.15 1
1.16 0.99
1.17 1
1.18 1
1.19 1
1.2 1
1.21 1
1.22 1
1.23 1
1.24 1
1.25 1
1.26 1
1.27 1
1.28 1
1.29 0.97
1.3 1
1.31 1
1.32 0.99
1.33 1
1.34 0.99
1.35 0.99
1.36 0.98
1.37 0.99
1.38 0.98
1.39 1
1.4 0.97
1.41 0.94
1.42 0.94
1.43 0.94
1.44 0.97
1.45 0.98
1.46 0.96
1.47 0.96
1.48 0.97
1.49 0.95
1.5 0.94
1.51 0.95
1.52 0.94
1.53 0.93
1.54 0.93
1.55 0.9
1.56 0.92
1.57 0.9
1.58 0.97
1.59 0.92
1.6 0.92
1.61 0.92
1.62 0.92
1.63 0.92
1.64 0.92
1.65 0.97
1.66 0.91
1.67 0.92
1.68 0.86
1.69 0.9
1.7 0.89
1.71 0.92
1.72 0.87
1.73 0.87
1.74 0.88
1.75 0.92
1.76 0.81
1.77 0.87
1.78 0.84
1.79 0.85
1.8 0.84
1.81 0.89
1.82 0.84
1.83 0.93
1.84 0.84
1.85 0.83
1.86 0.86
1.87 0.88
1.88 0.86
1.89 0.76
1.9 0.86
1.91 0.9
1.92 0.89
1.93 0.78
1.94 0.86
1.95 0.83
1.96 0.85
1.97 0.8
1.98 0.79
1.99 0.82
2 0.89
};
\addplot [semithick, color10, opacity=0.75, dashed]
table {%
0.01 0.54
0.02 0.54
0.03 0.48
0.04 0.49
0.05 0.56
0.06 0.56
0.07 0.63
0.08 0.63
0.09 0.77
0.1 0.72
0.11 0.74
0.12 0.76
0.13 0.77
0.14 0.76
0.15 0.79
0.16 0.83
0.17 0.77
0.18 0.9
0.19 0.85
0.2 0.79
0.21 0.86
0.22 0.93
0.23 0.9
0.24 0.86
0.25 0.84
0.26 0.94
0.27 0.96
0.28 0.98
0.29 0.92
0.3 0.94
0.31 0.98
0.32 0.91
0.33 0.94
0.34 0.97
0.35 0.99
0.36 1
0.37 0.94
0.38 0.97
0.39 1
0.4 0.99
0.41 0.98
0.42 0.99
0.43 0.99
0.44 0.98
0.45 0.98
0.46 0.99
0.47 0.99
0.48 1
0.49 1
0.5 1
0.51 0.99
0.52 0.99
0.53 1
0.54 0.99
0.55 0.99
0.56 0.99
0.57 0.98
0.58 1
0.59 1
0.6 1
0.61 1
0.62 1
0.63 1
0.64 1
0.65 1
0.66 0.99
0.67 1
0.68 1
0.69 1
0.7 0.99
0.71 1
0.72 0.99
0.73 1
0.74 1
0.75 1
0.76 1
0.77 1
0.78 1
0.79 1
0.8 1
0.81 1
0.82 1
0.83 1
0.84 1
0.85 1
0.86 1
0.87 1
0.88 0.99
0.89 1
0.9 1
0.91 1
0.92 1
0.93 1
0.94 1
0.95 1
0.96 1
0.97 1
0.98 1
0.99 1
1 1
1.02 1
1.03 1
1.04 1
1.05 1
1.06 1
1.07 0.99
1.08 1
1.09 1
1.1 1
1.11 1
1.12 1
1.13 1
1.14 1
1.15 1
1.16 1
1.17 1
1.18 1
1.19 1
1.2 1
1.21 1
1.22 1
1.23 0.99
1.24 0.99
1.25 1
1.26 1
1.27 1
1.28 0.97
1.29 0.99
1.3 1
1.31 1
1.32 0.96
1.33 0.98
1.34 0.96
1.35 0.97
1.36 0.93
1.37 0.99
1.38 0.96
1.39 0.93
1.4 0.98
1.41 0.93
1.42 0.92
1.43 0.88
1.44 0.97
1.45 0.93
1.46 0.89
1.47 0.9
1.48 0.95
1.49 0.89
1.5 0.93
1.51 0.92
1.52 0.93
1.53 0.86
1.54 0.9
1.55 0.87
1.56 0.92
1.57 0.85
1.58 0.91
1.59 0.87
1.6 0.86
1.61 0.87
1.62 0.87
1.63 0.83
1.64 0.85
1.65 0.86
1.66 0.78
1.67 0.88
1.68 0.82
1.69 0.77
1.7 0.82
1.71 0.8
1.72 0.75
1.73 0.69
1.74 0.83
1.75 0.84
1.76 0.73
1.77 0.81
1.78 0.8
1.79 0.79
1.8 0.76
1.81 0.81
1.82 0.78
1.83 0.89
1.84 0.72
1.85 0.81
1.86 0.82
1.87 0.85
1.88 0.77
1.89 0.75
1.9 0.79
1.91 0.85
1.92 0.79
1.93 0.76
1.94 0.75
1.95 0.72
1.96 0.75
1.97 0.71
1.98 0.76
1.99 0.73
2 0.8
};
\addplot [semithick, color11, opacity=0.75, dashed]
table {%
0.01 0.99
0.02 0.99
0.03 1
0.04 1
0.05 1
0.06 1
0.07 1
0.08 1
0.09 1
0.1 1
0.11 1
0.12 1
0.13 1
0.14 1
0.15 1
0.16 1
0.17 1
0.18 1
0.19 1
0.2 1
0.21 1
0.22 1
0.23 1
0.24 1
0.25 1
0.26 1
0.27 1
0.28 1
0.29 1
0.3 1
0.31 1
0.32 1
0.33 1
0.34 1
0.35 1
0.36 1
0.37 1
0.38 1
0.39 1
0.4 1
0.41 1
0.42 1
0.43 1
0.44 1
0.45 1
0.46 1
0.47 1
0.48 1
0.49 1
0.5 1
0.51 1
0.52 1
0.53 1
0.54 1
0.55 1
0.56 1
0.57 1
0.58 1
0.59 1
0.6 1
0.61 1
0.62 1
0.63 1
0.64 1
0.65 1
0.66 1
0.67 1
0.68 1
0.69 1
0.7 1
0.71 1
0.72 1
0.73 1
0.74 1
0.75 1
0.76 1
0.77 1
0.78 1
0.79 1
0.8 1
0.81 1
0.82 1
0.83 1
0.84 1
0.85 1
0.86 1
0.87 1
0.88 1
0.89 1
0.9 1
0.91 1
0.92 1
0.93 1
0.94 1
0.95 1
0.96 1
0.97 1
0.98 1
0.99 1
1 1
1.02 1
1.03 1
1.04 1
1.05 1
1.06 1
1.07 1
1.08 1
1.09 1
1.1 1
1.11 1
1.12 1
1.13 1
1.14 1
1.15 1
1.16 1
1.17 1
1.18 1
1.19 1
1.2 1
1.21 1
1.22 1
1.23 1
1.24 1
1.25 1
1.26 1
1.27 1
1.28 1
1.29 1
1.3 1
1.31 1
1.32 1
1.33 1
1.34 1
1.35 1
1.36 1
1.37 1
1.38 1
1.39 1
1.4 1
1.41 1
1.42 1
1.43 1
1.44 1
1.45 1
1.46 1
1.47 1
1.48 1
1.49 1
1.5 1
1.51 1
1.52 1
1.53 1
1.54 1
1.55 0.99
1.56 1
1.57 1
1.58 1
1.59 1
1.6 1
1.61 1
1.62 1
1.63 1
1.64 1
1.65 1
1.66 1
1.67 1
1.68 1
1.69 1
1.7 1
1.71 1
1.72 1
1.73 1
1.74 1
1.75 1
1.76 1
1.77 1
1.78 1
1.79 1
1.8 1
1.81 0.99
1.82 1
1.83 1
1.84 0.99
1.85 1
1.86 1
1.87 0.99
1.88 1
1.89 1
1.9 1
1.91 1
1.92 1
1.93 1
1.94 0.99
1.95 1
1.96 1
1.97 0.99
1.98 1
1.99 1
2 1
};
\addplot [semithick, black, opacity=1, dash pattern=on 1pt off 1pt]
table {%
-0.0895000000000001 0.5
2.0995 0.5
};
\addplot [semithick, black, opacity=1, dash pattern=on 1pt off 1pt]
table {%
-0.0895000000000001 1
2.0995 1
};
\addplot [semithick, black, opacity=1, dash pattern=on 1pt off 1pt]
table {%
1 0.2
1 1.05
};
\end{axis}

\end{tikzpicture}

%% file: plots/decoupled2/NL_UNIxGAU.tex
% This file was created by tikzplotlib v0.9.6.
\begin{tikzpicture}

\definecolor{color0}{rgb}{0.866666666666667,0.494117647058824,0.164705882352941}
\definecolor{color1}{rgb}{0.164705882352941,0.643137254901961,0.866666666666667}
\definecolor{color2}{rgb}{0.584313725490196,0.866666666666667,0.164705882352941}
\definecolor{color3}{rgb}{0.109803921568627,0.337254901960784,0.129411764705882}
\definecolor{color4}{rgb}{0.529411764705882,0.305882352941176,0.858823529411765}
\definecolor{color5}{rgb}{0.858823529411765,0.305882352941176,0.435294117647059}
\definecolor{color6}{rgb}{0.937254901960784,0.929411764705882,0.392156862745098}
\definecolor{color7}{rgb}{0.0901960784313725,0.486274509803922,0.0980392156862745}
\definecolor{color8}{rgb}{0.156862745098039,0.188235294117647,0.827450980392157}
\definecolor{color9}{rgb}{0.937254901960784,0.392156862745098,0.894117647058824}
\definecolor{color10}{rgb}{0.2,0.184313725490196,0.184313725490196}
\definecolor{color11}{rgb}{0.0156862745098039,0.803921568627451,0.976470588235294}

\begin{axis}[
tick align=outside,
tick pos=left,
x grid style={white!69.0196078431373!black},
xmajorgrids,
xmin=-0.0895, xmax=2.0995,
xtick style={color=black},
xtick={0,0.1,0.2,0.3,0.4,0.5,0.6,0.7,0.8,0.9,1,1.1,1.2,1.3,1.4,1.5,1.6,1.7,1.8,1.9,2},
xticklabels={0,,.2,,.4,,.6,,.8,,1,,20,,40,,60,,80,,100},
height=4.8cm,
width=6.5cm,
y grid style={white!69.0196078431373!black},
ymajorgrids,
ymin=0.2, ymax=1.05,
ytick style={color=black}
]
\addplot [semithick, color0, opacity=0.75]
table {%
0.01 0.59
0.02 0.8
0.03 0.88
0.04 0.88
0.05 0.99
0.06 1
0.07 0.98
0.08 0.97
0.09 0.99
0.1 1
0.11 1
0.12 1
0.13 1
0.14 0.99
0.15 1
0.16 0.99
0.17 1
0.18 1
0.19 1
0.2 1
0.21 1
0.22 1
0.23 1
0.24 1
0.25 1
0.26 1
0.27 1
0.28 1
0.29 1
0.3 1
0.31 1
0.32 1
0.33 1
0.34 1
0.35 1
0.36 1
0.37 1
0.38 1
0.39 1
0.4 1
0.41 1
0.42 1
0.43 1
0.44 1
0.45 1
0.46 1
0.47 1
0.48 1
0.49 1
0.5 1
0.51 1
0.52 1
0.53 1
0.54 1
0.55 1
0.56 1
0.57 1
0.58 1
0.59 1
0.6 1
0.61 1
0.62 1
0.63 1
0.64 1
0.65 1
0.66 1
0.67 1
0.68 1
0.69 1
0.7 1
0.71 1
0.72 1
0.73 1
0.74 1
0.75 1
0.76 1
0.77 1
0.78 1
0.79 1
0.8 1
0.81 1
0.82 1
0.83 1
0.84 1
0.85 1
0.86 1
0.87 1
0.88 1
0.89 1
0.9 1
0.91 1
0.92 1
0.93 1
0.94 1
0.95 1
0.96 1
0.97 1
0.98 1
0.99 1
1 1
1.02 0.99
1.03 0.86
1.04 0.88
1.05 0.82
1.06 0.75
1.07 0.7
1.08 0.76
1.09 0.71
1.1 0.73
1.11 0.68
1.12 0.78
1.13 0.66
1.14 0.77
1.15 0.66
1.16 0.64
1.17 0.64
1.18 0.57
1.19 0.57
1.2 0.69
1.21 0.7
1.22 0.67
1.23 0.68
1.24 0.68
1.25 0.71
1.26 0.6
1.27 0.63
1.28 0.64
1.29 0.72
1.3 0.68
1.31 0.61
1.32 0.69
1.33 0.69
1.34 0.6
1.35 0.62
1.36 0.59
1.37 0.62
1.38 0.69
1.39 0.61
1.4 0.64
1.41 0.62
1.42 0.63
1.43 0.72
1.44 0.66
1.45 0.65
1.46 0.7
1.47 0.57
1.48 0.66
1.49 0.64
1.5 0.71
1.51 0.67
1.52 0.66
1.53 0.68
1.54 0.69
1.55 0.65
1.56 0.7
1.57 0.58
1.58 0.68
1.59 0.68
1.6 0.6
1.61 0.63
1.62 0.6
1.63 0.64
1.64 0.63
1.65 0.74
1.66 0.62
1.67 0.69
1.68 0.62
1.69 0.69
1.7 0.68
1.71 0.67
1.72 0.62
1.73 0.61
1.74 0.73
1.75 0.65
1.76 0.6
1.77 0.52
1.78 0.61
1.79 0.64
1.8 0.63
1.81 0.66
1.82 0.6
1.83 0.61
1.84 0.68
1.85 0.71
1.86 0.74
1.87 0.65
1.88 0.66
1.89 0.67
1.9 0.63
1.91 0.67
1.92 0.68
1.93 0.65
1.94 0.57
1.95 0.61
1.96 0.57
1.97 0.61
1.98 0.71
1.99 0.63
2 0.57
};
\addplot [semithick, color1, opacity=0.75]
table {%
0.01 0.51
0.02 0.53
0.03 0.49
0.04 0.46
0.05 0.52
0.06 0.56
0.07 0.47
0.08 0.51
0.09 0.58
0.1 0.52
0.11 0.51
0.12 0.62
0.13 0.59
0.14 0.63
0.15 0.57
0.16 0.54
0.17 0.53
0.18 0.51
0.19 0.46
0.2 0.46
0.21 0.57
0.22 0.6
0.23 0.57
0.24 0.6
0.25 0.57
0.26 0.58
0.27 0.6
0.28 0.65
0.29 0.64
0.3 0.56
0.31 0.61
0.32 0.59
0.33 0.61
0.34 0.55
0.35 0.57
0.36 0.68
0.37 0.55
0.38 0.54
0.39 0.64
0.4 0.67
0.41 0.67
0.42 0.61
0.43 0.65
0.44 0.63
0.45 0.62
0.46 0.64
0.47 0.61
0.48 0.66
0.49 0.6
0.5 0.52
0.51 0.65
0.52 0.56
0.53 0.58
0.54 0.65
0.55 0.63
0.56 0.62
0.57 0.63
0.58 0.7
0.59 0.63
0.6 0.61
0.61 0.56
0.62 0.66
0.63 0.59
0.64 0.63
0.65 0.6
0.66 0.6
0.67 0.66
0.68 0.58
0.69 0.55
0.7 0.54
0.71 0.61
0.72 0.57
0.73 0.59
0.74 0.56
0.75 0.6
0.76 0.63
0.77 0.5
0.78 0.64
0.79 0.66
0.8 0.56
0.81 0.56
0.82 0.52
0.83 0.62
0.84 0.57
0.85 0.59
0.86 0.55
0.87 0.55
0.88 0.53
0.89 0.55
0.9 0.6
0.91 0.57
0.92 0.53
0.93 0.67
0.94 0.61
0.95 0.58
0.96 0.61
0.97 0.51
0.98 0.55
0.99 0.44
1 0.59
1.02 0.44
1.03 0.41
1.04 0.44
1.05 0.5
1.06 0.49
1.07 0.54
1.08 0.58
1.09 0.5
1.1 0.49
1.11 0.51
1.12 0.46
1.13 0.46
1.14 0.62
1.15 0.42
1.16 0.53
1.17 0.53
1.18 0.58
1.19 0.42
1.2 0.5
1.21 0.53
1.22 0.49
1.23 0.48
1.24 0.4
1.25 0.51
1.26 0.5
1.27 0.48
1.28 0.59
1.29 0.52
1.3 0.51
1.31 0.55
1.32 0.48
1.33 0.43
1.34 0.49
1.35 0.61
1.36 0.54
1.37 0.45
1.38 0.48
1.39 0.52
1.4 0.55
1.41 0.54
1.42 0.45
1.43 0.42
1.44 0.55
1.45 0.52
1.46 0.5
1.47 0.44
1.48 0.43
1.49 0.52
1.5 0.5
1.51 0.47
1.52 0.62
1.53 0.5
1.54 0.42
1.55 0.52
1.56 0.52
1.57 0.48
1.58 0.54
1.59 0.53
1.6 0.59
1.61 0.47
1.62 0.43
1.63 0.49
1.64 0.57
1.65 0.51
1.66 0.44
1.67 0.5
1.68 0.47
1.69 0.48
1.7 0.47
1.71 0.62
1.72 0.54
1.73 0.45
1.74 0.44
1.75 0.51
1.76 0.49
1.77 0.45
1.78 0.54
1.79 0.5
1.8 0.56
1.81 0.52
1.82 0.43
1.83 0.5
1.84 0.51
1.85 0.49
1.86 0.5
1.87 0.6
1.88 0.44
1.89 0.47
1.9 0.53
1.91 0.46
1.92 0.54
1.93 0.6
1.94 0.43
1.95 0.43
1.96 0.5
1.97 0.47
1.98 0.52
1.99 0.6
2 0.44
};
\addplot [semithick, color2, opacity=0.75]
table {%
0.01 0.51
0.02 0.44
0.03 0.54
0.04 0.48
0.05 0.41
0.06 0.44
0.07 0.56
0.08 0.52
0.09 0.48
0.1 0.44
0.11 0.47
0.12 0.6
0.13 0.58
0.14 0.58
0.15 0.55
0.16 0.53
0.17 0.5
0.18 0.53
0.19 0.46
0.2 0.46
0.21 0.57
0.22 0.59
0.23 0.58
0.24 0.54
0.25 0.51
0.26 0.58
0.27 0.57
0.28 0.63
0.29 0.6
0.3 0.52
0.31 0.58
0.32 0.58
0.33 0.62
0.34 0.52
0.35 0.54
0.36 0.67
0.37 0.54
0.38 0.51
0.39 0.63
0.4 0.65
0.41 0.67
0.42 0.59
0.43 0.62
0.44 0.61
0.45 0.6
0.46 0.63
0.47 0.6
0.48 0.67
0.49 0.59
0.5 0.49
0.51 0.63
0.52 0.56
0.53 0.56
0.54 0.64
0.55 0.61
0.56 0.61
0.57 0.6
0.58 0.68
0.59 0.61
0.6 0.58
0.61 0.55
0.62 0.64
0.63 0.58
0.64 0.57
0.65 0.57
0.66 0.56
0.67 0.63
0.68 0.55
0.69 0.54
0.7 0.51
0.71 0.56
0.72 0.55
0.73 0.56
0.74 0.53
0.75 0.56
0.76 0.56
0.77 0.5
0.78 0.61
0.79 0.65
0.8 0.51
0.81 0.54
0.82 0.51
0.83 0.58
0.84 0.56
0.85 0.55
0.86 0.48
0.87 0.51
0.88 0.53
0.89 0.54
0.9 0.57
0.91 0.56
0.92 0.5
0.93 0.62
0.94 0.58
0.95 0.57
0.96 0.59
0.97 0.5
0.98 0.53
0.99 0.43
1 0.58
1.02 0.44
1.03 0.41
1.04 0.44
1.05 0.53
1.06 0.49
1.07 0.56
1.08 0.56
1.09 0.5
1.1 0.5
1.11 0.47
1.12 0.44
1.13 0.45
1.14 0.63
1.15 0.42
1.16 0.54
1.17 0.53
1.18 0.55
1.19 0.43
1.2 0.5
1.21 0.52
1.22 0.47
1.23 0.47
1.24 0.38
1.25 0.5
1.26 0.48
1.27 0.42
1.28 0.59
1.29 0.54
1.3 0.49
1.31 0.56
1.32 0.48
1.33 0.42
1.34 0.49
1.35 0.6
1.36 0.58
1.37 0.45
1.38 0.49
1.39 0.54
1.4 0.55
1.41 0.52
1.42 0.45
1.43 0.42
1.44 0.54
1.45 0.5
1.46 0.5
1.47 0.46
1.48 0.45
1.49 0.5
1.5 0.48
1.51 0.49
1.52 0.64
1.53 0.5
1.54 0.42
1.55 0.49
1.56 0.53
1.57 0.46
1.58 0.55
1.59 0.52
1.6 0.58
1.61 0.49
1.62 0.46
1.63 0.48
1.64 0.59
1.65 0.5
1.66 0.45
1.67 0.51
1.68 0.52
1.69 0.48
1.7 0.48
1.71 0.63
1.72 0.55
1.73 0.46
1.74 0.43
1.75 0.52
1.76 0.5
1.77 0.45
1.78 0.53
1.79 0.51
1.8 0.55
1.81 0.53
1.82 0.45
1.83 0.52
1.84 0.52
1.85 0.49
1.86 0.51
1.87 0.59
1.88 0.45
1.89 0.46
1.9 0.54
1.91 0.45
1.92 0.53
1.93 0.58
1.94 0.44
1.95 0.44
1.96 0.49
1.97 0.48
1.98 0.5
1.99 0.59
2 0.43
};
\addplot [semithick, color3, opacity=0.75]
table {%
0.01 0.82
0.02 0.94
0.03 0.98
0.04 0.97
0.05 0.98
0.06 0.99
0.07 0.99
0.08 0.98
0.09 0.99
0.1 0.98
0.11 0.96
0.12 0.99
0.13 0.99
0.14 0.98
0.15 0.98
0.16 0.99
0.17 0.99
0.18 1
0.19 1
0.2 0.99
0.21 0.99
0.22 1
0.23 0.99
0.24 1
0.25 0.99
0.26 1
0.27 0.97
0.28 0.99
0.29 0.99
0.3 0.97
0.31 0.98
0.32 1
0.33 0.98
0.34 0.93
0.35 0.98
0.36 0.98
0.37 1
0.38 0.97
0.39 0.99
0.4 1
0.41 0.99
0.42 0.99
0.43 0.99
0.44 0.98
0.45 0.95
0.46 0.97
0.47 0.97
0.48 1
0.49 0.93
0.5 0.99
0.51 0.97
0.52 0.97
0.53 1
0.54 0.99
0.55 0.99
0.56 0.99
0.57 1
0.58 1
0.59 0.99
0.6 0.99
0.61 0.99
0.62 0.99
0.63 0.99
0.64 0.99
0.65 0.99
0.66 0.99
0.67 1
0.68 1
0.69 0.99
0.7 1
0.71 0.99
0.72 0.96
0.73 0.99
0.74 0.99
0.75 0.99
0.76 0.99
0.77 0.99
0.78 0.98
0.79 0.98
0.8 0.97
0.81 1
0.82 1
0.83 0.98
0.84 1
0.85 1
0.86 1
0.87 0.99
0.88 1
0.89 0.99
0.9 0.99
0.91 0.99
0.92 0.99
0.93 1
0.94 1
0.95 1
0.96 0.98
0.97 1
0.98 0.99
0.99 0.99
1 0.99
1.02 1
1.03 0.96
1.04 0.95
1.05 0.95
1.06 0.91
1.07 0.91
1.08 0.93
1.09 0.89
1.1 0.84
1.11 0.82
1.12 0.88
1.13 0.89
1.14 0.88
1.15 0.84
1.16 0.82
1.17 0.88
1.18 0.79
1.19 0.76
1.2 0.87
1.21 0.86
1.22 0.88
1.23 0.86
1.24 0.81
1.25 0.81
1.26 0.77
1.27 0.78
1.28 0.82
1.29 0.85
1.3 0.86
1.31 0.8
1.32 0.83
1.33 0.84
1.34 0.75
1.35 0.78
1.36 0.8
1.37 0.87
1.38 0.85
1.39 0.83
1.4 0.8
1.41 0.86
1.42 0.83
1.43 0.88
1.44 0.85
1.45 0.79
1.46 0.84
1.47 0.79
1.48 0.85
1.49 0.78
1.5 0.84
1.51 0.81
1.52 0.84
1.53 0.85
1.54 0.83
1.55 0.82
1.56 0.93
1.57 0.78
1.58 0.86
1.59 0.8
1.6 0.79
1.61 0.83
1.62 0.75
1.63 0.78
1.64 0.78
1.65 0.81
1.66 0.82
1.67 0.86
1.68 0.83
1.69 0.86
1.7 0.86
1.71 0.84
1.72 0.82
1.73 0.8
1.74 0.86
1.75 0.84
1.76 0.8
1.77 0.75
1.78 0.83
1.79 0.81
1.8 0.84
1.81 0.91
1.82 0.82
1.83 0.84
1.84 0.81
1.85 0.92
1.86 0.84
1.87 0.88
1.88 0.81
1.89 0.82
1.9 0.77
1.91 0.87
1.92 0.84
1.93 0.77
1.94 0.79
1.95 0.79
1.96 0.83
1.97 0.82
1.98 0.84
1.99 0.82
2 0.79
};
\addplot [semithick, color4, opacity=0.75]
table {%
0.01 0.83
0.02 0.94
0.03 0.98
0.04 0.97
0.05 0.98
0.06 0.99
0.07 0.99
0.08 0.98
0.09 0.99
0.1 0.98
0.11 0.96
0.12 0.99
0.13 0.99
0.14 0.98
0.15 0.98
0.16 0.99
0.17 0.99
0.18 1
0.19 0.99
0.2 0.99
0.21 0.99
0.22 1
0.23 0.99
0.24 1
0.25 0.99
0.26 1
0.27 0.97
0.28 0.99
0.29 0.98
0.3 0.96
0.31 0.98
0.32 0.99
0.33 0.97
0.34 0.93
0.35 0.96
0.36 0.98
0.37 1
0.38 0.97
0.39 0.99
0.4 1
0.41 0.99
0.42 0.99
0.43 0.98
0.44 0.98
0.45 0.95
0.46 0.97
0.47 0.97
0.48 1
0.49 0.93
0.5 0.99
0.51 0.97
0.52 0.96
0.53 1
0.54 0.99
0.55 0.99
0.56 0.99
0.57 0.99
0.58 1
0.59 0.99
0.6 0.98
0.61 0.99
0.62 0.99
0.63 0.99
0.64 0.98
0.65 0.99
0.66 0.99
0.67 1
0.68 1
0.69 0.99
0.7 1
0.71 0.98
0.72 0.96
0.73 0.99
0.74 0.99
0.75 0.98
0.76 0.99
0.77 0.99
0.78 0.97
0.79 0.98
0.8 0.98
0.81 1
0.82 1
0.83 0.98
0.84 1
0.85 1
0.86 1
0.87 0.98
0.88 1
0.89 0.99
0.9 0.99
0.91 0.99
0.92 0.99
0.93 1
0.94 1
0.95 1
0.96 0.99
0.97 1
0.98 0.99
0.99 0.99
1 0.99
1.02 0.99
1.03 0.95
1.04 0.95
1.05 0.96
1.06 0.92
1.07 0.91
1.08 0.93
1.09 0.89
1.1 0.84
1.11 0.83
1.12 0.88
1.13 0.89
1.14 0.89
1.15 0.84
1.16 0.83
1.17 0.87
1.18 0.78
1.19 0.77
1.2 0.89
1.21 0.87
1.22 0.87
1.23 0.86
1.24 0.81
1.25 0.82
1.26 0.76
1.27 0.79
1.28 0.82
1.29 0.85
1.3 0.86
1.31 0.81
1.32 0.83
1.33 0.84
1.34 0.75
1.35 0.8
1.36 0.8
1.37 0.87
1.38 0.84
1.39 0.84
1.4 0.79
1.41 0.86
1.42 0.84
1.43 0.88
1.44 0.86
1.45 0.8
1.46 0.83
1.47 0.78
1.48 0.85
1.49 0.78
1.5 0.84
1.51 0.81
1.52 0.83
1.53 0.85
1.54 0.84
1.55 0.82
1.56 0.93
1.57 0.78
1.58 0.87
1.59 0.81
1.6 0.79
1.61 0.82
1.62 0.76
1.63 0.8
1.64 0.78
1.65 0.83
1.66 0.82
1.67 0.87
1.68 0.83
1.69 0.86
1.7 0.87
1.71 0.86
1.72 0.84
1.73 0.8
1.74 0.86
1.75 0.85
1.76 0.81
1.77 0.75
1.78 0.83
1.79 0.83
1.8 0.84
1.81 0.91
1.82 0.82
1.83 0.85
1.84 0.81
1.85 0.92
1.86 0.84
1.87 0.88
1.88 0.81
1.89 0.83
1.9 0.79
1.91 0.87
1.92 0.85
1.93 0.77
1.94 0.78
1.95 0.81
1.96 0.82
1.97 0.83
1.98 0.84
1.99 0.83
2 0.8
};
\addplot [semithick, color5, opacity=0.75]
table {%
0.01 0.86
0.02 0.93
0.03 0.97
0.04 0.95
0.05 0.98
0.06 1
0.07 0.95
0.08 0.95
0.09 0.99
0.1 0.97
0.11 0.96
0.12 1
0.13 0.97
0.14 0.96
0.15 0.98
0.16 0.99
0.17 0.98
0.18 1
0.19 0.99
0.2 0.98
0.21 0.96
0.22 0.99
0.23 0.99
0.24 0.98
0.25 0.98
0.26 1
0.27 0.97
0.28 0.98
0.29 0.98
0.3 0.96
0.31 0.96
0.32 0.96
0.33 0.97
0.34 0.93
0.35 0.97
0.36 0.98
0.37 0.99
0.38 0.98
0.39 0.98
0.4 0.99
0.41 0.99
0.42 0.99
0.43 0.99
0.44 0.98
0.45 0.94
0.46 0.96
0.47 0.96
0.48 1
0.49 0.94
0.5 0.99
0.51 0.96
0.52 0.95
0.53 0.97
0.54 0.99
0.55 0.98
0.56 0.99
0.57 1
0.58 0.97
0.59 0.98
0.6 0.97
0.61 0.99
0.62 0.98
0.63 0.97
0.64 0.99
0.65 0.95
0.66 0.98
0.67 1
0.68 1
0.69 0.99
0.7 1
0.71 0.94
0.72 0.97
0.73 0.99
0.74 0.99
0.75 0.99
0.76 0.98
0.77 0.96
0.78 0.95
0.79 0.98
0.8 0.99
0.81 0.98
0.82 1
0.83 0.96
0.84 0.98
0.85 1
0.86 0.98
0.87 0.98
0.88 0.99
0.89 0.95
0.9 0.98
0.91 0.98
0.92 0.97
0.93 0.99
0.94 0.97
0.95 0.99
0.96 0.98
0.97 0.96
0.98 0.99
0.99 0.97
1 0.96
1.02 0.98
1.03 0.89
1.04 0.94
1.05 0.92
1.06 0.86
1.07 0.91
1.08 0.91
1.09 0.83
1.1 0.84
1.11 0.9
1.12 0.87
1.13 0.9
1.14 0.88
1.15 0.87
1.16 0.84
1.17 0.85
1.18 0.88
1.19 0.88
1.2 0.88
1.21 0.9
1.22 0.89
1.23 0.84
1.24 0.87
1.25 0.8
1.26 0.78
1.27 0.87
1.28 0.9
1.29 0.91
1.3 0.9
1.31 0.92
1.32 0.86
1.33 0.9
1.34 0.89
1.35 0.85
1.36 0.91
1.37 0.89
1.38 0.87
1.39 0.88
1.4 0.85
1.41 0.89
1.42 0.87
1.43 0.87
1.44 0.9
1.45 0.8
1.46 0.85
1.47 0.81
1.48 0.85
1.49 0.83
1.5 0.88
1.51 0.91
1.52 0.85
1.53 0.9
1.54 0.94
1.55 0.86
1.56 0.89
1.57 0.82
1.58 0.87
1.59 0.83
1.6 0.79
1.61 0.9
1.62 0.84
1.63 0.79
1.64 0.86
1.65 0.85
1.66 0.88
1.67 0.89
1.68 0.88
1.69 0.92
1.7 0.93
1.71 0.87
1.72 0.86
1.73 0.87
1.74 0.88
1.75 0.87
1.76 0.86
1.77 0.8
1.78 0.85
1.79 0.89
1.8 0.84
1.81 0.88
1.82 0.83
1.83 0.88
1.84 0.91
1.85 0.89
1.86 0.84
1.87 0.92
1.88 0.86
1.89 0.9
1.9 0.86
1.91 0.82
1.92 0.85
1.93 0.86
1.94 0.88
1.95 0.83
1.96 0.91
1.97 0.85
1.98 0.87
1.99 0.86
2 0.82
};
\addplot [semithick, color6, opacity=0.75, dashed]
table {%
0.01 0.98
0.02 0.99
0.03 1
0.04 1
0.05 1
0.06 1
0.07 1
0.08 1
0.09 1
0.1 1
0.11 1
0.12 1
0.13 1
0.14 1
0.15 1
0.16 1
0.17 1
0.18 1
0.19 1
0.2 1
0.21 1
0.22 1
0.23 1
0.24 1
0.25 1
0.26 1
0.27 1
0.28 1
0.29 1
0.3 1
0.31 1
0.32 1
0.33 1
0.34 1
0.35 1
0.36 1
0.37 1
0.38 1
0.39 1
0.4 1
0.41 1
0.42 1
0.43 1
0.44 1
0.45 1
0.46 1
0.47 1
0.48 1
0.49 1
0.5 1
0.51 1
0.52 1
0.53 1
0.54 1
0.55 1
0.56 1
0.57 1
0.58 1
0.59 1
0.6 1
0.61 1
0.62 1
0.63 1
0.64 1
0.65 1
0.66 1
0.67 1
0.68 1
0.69 1
0.7 1
0.71 1
0.72 1
0.73 1
0.74 1
0.75 1
0.76 1
0.77 1
0.78 1
0.79 1
0.8 1
0.81 1
0.82 1
0.83 1
0.84 1
0.85 1
0.86 1
0.87 1
0.88 1
0.89 1
0.9 1
0.91 1
0.92 1
0.93 1
0.94 1
0.95 1
0.96 1
0.97 1
0.98 1
0.99 1
1 1
1.02 1
1.03 1
1.04 1
1.05 1
1.06 0.99
1.07 1
1.08 0.99
1.09 0.98
1.1 0.99
1.11 0.99
1.12 1
1.13 0.99
1.14 0.98
1.15 1
1.16 0.99
1.17 0.99
1.18 0.94
1.19 0.98
1.2 0.98
1.21 0.99
1.22 0.99
1.23 0.98
1.24 0.96
1.25 0.97
1.26 0.97
1.27 0.96
1.28 0.97
1.29 1
1.3 0.98
1.31 0.99
1.32 0.98
1.33 1
1.34 1
1.35 0.99
1.36 0.98
1.37 0.99
1.38 1
1.39 0.97
1.4 0.96
1.41 0.98
1.42 1
1.43 0.99
1.44 0.98
1.45 0.99
1.46 0.99
1.47 0.98
1.48 1
1.49 0.98
1.5 0.99
1.51 0.98
1.52 0.97
1.53 0.99
1.54 1
1.55 0.96
1.56 0.98
1.57 0.99
1.58 0.98
1.59 1
1.6 0.99
1.61 1
1.62 0.97
1.63 0.97
1.64 0.98
1.65 0.98
1.66 1
1.67 0.98
1.68 1
1.69 0.98
1.7 0.99
1.71 1
1.72 0.99
1.73 0.97
1.74 0.99
1.75 0.99
1.76 0.99
1.77 0.98
1.78 0.99
1.79 1
1.8 0.99
1.81 1
1.82 0.99
1.83 0.99
1.84 0.98
1.85 0.99
1.86 0.97
1.87 0.98
1.88 0.98
1.89 0.99
1.9 0.98
1.91 0.99
1.92 1
1.93 0.98
1.94 0.98
1.95 0.99
1.96 0.96
1.97 0.98
1.98 0.99
1.99 0.98
2 0.99
};
\addplot [semithick, color7, opacity=0.75, dashed]
table {%
0.01 0.98
0.02 0.99
0.03 1
0.04 1
0.05 1
0.06 1
0.07 1
0.08 1
0.09 1
0.1 1
0.11 1
0.12 1
0.13 1
0.14 1
0.15 1
0.16 1
0.17 1
0.18 1
0.19 1
0.2 1
0.21 1
0.22 1
0.23 1
0.24 1
0.25 1
0.26 1
0.27 1
0.28 1
0.29 1
0.3 1
0.31 1
0.32 1
0.33 1
0.34 1
0.35 1
0.36 1
0.37 1
0.38 1
0.39 1
0.4 1
0.41 1
0.42 1
0.43 1
0.44 1
0.45 1
0.46 1
0.47 1
0.48 1
0.49 1
0.5 1
0.51 1
0.52 1
0.53 1
0.54 1
0.55 1
0.56 1
0.57 1
0.58 1
0.59 1
0.6 1
0.61 1
0.62 1
0.63 1
0.64 1
0.65 1
0.66 1
0.67 1
0.68 1
0.69 1
0.7 1
0.71 1
0.72 1
0.73 1
0.74 1
0.75 1
0.76 1
0.77 1
0.78 1
0.79 1
0.8 1
0.81 1
0.82 1
0.83 1
0.84 1
0.85 1
0.86 1
0.87 1
0.88 1
0.89 1
0.9 1
0.91 1
0.92 1
0.93 1
0.94 1
0.95 1
0.96 1
0.97 1
0.98 1
0.99 1
1 1
1.02 1
1.03 1
1.04 1
1.05 1
1.06 0.99
1.07 1
1.08 0.99
1.09 0.98
1.1 0.99
1.11 0.99
1.12 1
1.13 0.99
1.14 0.98
1.15 1
1.16 0.99
1.17 0.99
1.18 0.94
1.19 0.98
1.2 0.98
1.21 0.99
1.22 0.99
1.23 0.98
1.24 0.96
1.25 0.97
1.26 0.97
1.27 0.96
1.28 0.97
1.29 1
1.3 0.98
1.31 0.99
1.32 0.98
1.33 1
1.34 1
1.35 0.99
1.36 0.98
1.37 0.99
1.38 1
1.39 0.97
1.4 0.96
1.41 0.98
1.42 1
1.43 0.99
1.44 0.98
1.45 0.99
1.46 0.99
1.47 0.98
1.48 1
1.49 0.98
1.5 0.99
1.51 0.98
1.52 0.97
1.53 0.99
1.54 1
1.55 0.96
1.56 0.98
1.57 0.99
1.58 0.98
1.59 1
1.6 0.99
1.61 1
1.62 0.97
1.63 0.97
1.64 0.98
1.65 0.98
1.66 1
1.67 0.98
1.68 1
1.69 0.98
1.7 0.99
1.71 1
1.72 0.99
1.73 0.97
1.74 0.99
1.75 0.99
1.76 0.99
1.77 0.98
1.78 0.99
1.79 1
1.8 0.99
1.81 1
1.82 0.99
1.83 0.99
1.84 0.98
1.85 0.99
1.86 0.97
1.87 0.98
1.88 0.98
1.89 0.99
1.9 0.98
1.91 0.99
1.92 1
1.93 0.98
1.94 0.98
1.95 0.99
1.96 0.96
1.97 0.98
1.98 0.99
1.99 0.98
2 0.99
};
\addplot [semithick, color8, opacity=0.75, dashed]
table {%
0.01 1
0.02 1
0.03 1
0.04 1
0.05 1
0.06 1
0.07 1
0.08 1
0.09 1
0.1 1
0.11 1
0.12 1
0.13 1
0.14 1
0.15 1
0.16 1
0.17 1
0.18 1
0.19 1
0.2 1
0.21 1
0.22 1
0.23 1
0.24 1
0.25 1
0.26 1
0.27 1
0.28 1
0.29 1
0.3 1
0.31 1
0.32 1
0.33 1
0.34 1
0.35 1
0.36 1
0.37 1
0.38 1
0.39 1
0.4 1
0.41 1
0.42 1
0.43 1
0.44 1
0.45 1
0.46 1
0.47 1
0.48 1
0.49 1
0.5 1
0.51 1
0.52 1
0.53 1
0.54 1
0.55 1
0.56 1
0.57 1
0.58 1
0.59 1
0.6 1
0.61 1
0.62 1
0.63 1
0.64 1
0.65 1
0.66 1
0.67 1
0.68 1
0.69 1
0.7 1
0.71 1
0.72 1
0.73 1
0.74 1
0.75 1
0.76 1
0.77 1
0.78 1
0.79 1
0.8 1
0.81 1
0.82 1
0.83 1
0.84 1
0.85 1
0.86 1
0.87 1
0.88 1
0.89 1
0.9 1
0.91 1
0.92 1
0.93 1
0.94 1
0.95 1
0.96 1
0.97 1
0.98 1
0.99 1
1 1
1.02 1
1.03 1
1.04 1
1.05 1
1.06 1
1.07 1
1.08 1
1.09 0.99
1.1 1
1.11 1
1.12 1
1.13 1
1.14 0.98
1.15 0.99
1.16 1
1.17 1
1.18 0.99
1.19 0.99
1.2 0.99
1.21 1
1.22 0.99
1.23 1
1.24 0.99
1.25 0.99
1.26 0.97
1.27 0.99
1.28 0.99
1.29 1
1.3 1
1.31 1
1.32 1
1.33 1
1.34 0.99
1.35 1
1.36 1
1.37 1
1.38 1
1.39 0.99
1.4 0.98
1.41 1
1.42 0.99
1.43 1
1.44 1
1.45 1
1.46 1
1.47 0.99
1.48 0.98
1.49 1
1.5 1
1.51 0.99
1.52 0.98
1.53 0.99
1.54 1
1.55 0.99
1.56 1
1.57 1
1.58 0.99
1.59 1
1.6 1
1.61 0.99
1.62 0.98
1.63 1
1.64 1
1.65 1
1.66 0.99
1.67 1
1.68 0.99
1.69 0.99
1.7 1
1.71 1
1.72 0.99
1.73 0.99
1.74 1
1.75 0.99
1.76 1
1.77 0.96
1.78 1
1.79 0.99
1.8 1
1.81 1
1.82 1
1.83 1
1.84 1
1.85 0.99
1.86 1
1.87 1
1.88 1
1.89 1
1.9 0.99
1.91 1
1.92 0.99
1.93 1
1.94 0.98
1.95 1
1.96 0.99
1.97 1
1.98 1
1.99 0.99
2 0.99
};
\addplot [semithick, color9, opacity=0.75, dashed]
table {%
0.01 0.78
0.02 0.92
0.03 0.92
0.04 0.97
0.05 0.99
0.06 1
0.07 0.99
0.08 1
0.09 1
0.1 1
0.11 1
0.12 1
0.13 1
0.14 1
0.15 1
0.16 1
0.17 1
0.18 1
0.19 1
0.2 1
0.21 1
0.22 1
0.23 1
0.24 1
0.25 1
0.26 1
0.27 1
0.28 1
0.29 1
0.3 1
0.31 1
0.32 1
0.33 1
0.34 1
0.35 1
0.36 1
0.37 1
0.38 1
0.39 1
0.4 1
0.41 1
0.42 1
0.43 1
0.44 1
0.45 1
0.46 1
0.47 1
0.48 1
0.49 1
0.5 1
0.51 1
0.52 1
0.53 1
0.54 1
0.55 1
0.56 1
0.57 1
0.58 1
0.59 1
0.6 1
0.61 1
0.62 1
0.63 1
0.64 1
0.65 1
0.66 1
0.67 1
0.68 1
0.69 1
0.7 1
0.71 1
0.72 1
0.73 1
0.74 1
0.75 1
0.76 1
0.77 1
0.78 1
0.79 1
0.8 1
0.81 1
0.82 1
0.83 1
0.84 1
0.85 1
0.86 1
0.87 1
0.88 1
0.89 1
0.9 1
0.91 1
0.92 1
0.93 1
0.94 1
0.95 1
0.96 1
0.97 1
0.98 1
0.99 1
1 1
1.02 1
1.03 0.99
1.04 0.91
1.05 0.88
1.06 0.82
1.07 0.8
1.08 0.88
1.09 0.74
1.1 0.76
1.11 0.81
1.12 0.76
1.13 0.81
1.14 0.75
1.15 0.74
1.16 0.79
1.17 0.77
1.18 0.74
1.19 0.7
1.2 0.7
1.21 0.7
1.22 0.72
1.23 0.72
1.24 0.68
1.25 0.72
1.26 0.71
1.27 0.74
1.28 0.66
1.29 0.8
1.3 0.74
1.31 0.79
1.32 0.72
1.33 0.72
1.34 0.66
1.35 0.73
1.36 0.63
1.37 0.78
1.38 0.7
1.39 0.66
1.4 0.69
1.41 0.71
1.42 0.64
1.43 0.75
1.44 0.73
1.45 0.66
1.46 0.73
1.47 0.73
1.48 0.73
1.49 0.65
1.5 0.74
1.51 0.67
1.52 0.66
1.53 0.7
1.54 0.71
1.55 0.73
1.56 0.78
1.57 0.7
1.58 0.71
1.59 0.65
1.6 0.74
1.61 0.64
1.62 0.71
1.63 0.69
1.64 0.67
1.65 0.73
1.66 0.73
1.67 0.73
1.68 0.72
1.69 0.68
1.7 0.69
1.71 0.76
1.72 0.69
1.73 0.67
1.74 0.71
1.75 0.69
1.76 0.73
1.77 0.6
1.78 0.68
1.79 0.75
1.8 0.74
1.81 0.7
1.82 0.71
1.83 0.69
1.84 0.73
1.85 0.74
1.86 0.72
1.87 0.64
1.88 0.7
1.89 0.74
1.9 0.7
1.91 0.72
1.92 0.74
1.93 0.68
1.94 0.66
1.95 0.71
1.96 0.69
1.97 0.73
1.98 0.71
1.99 0.7
2 0.66
};
\addplot [semithick, color10, opacity=0.75, dashed]
table {%
0.01 0.57
0.02 0.72
0.03 0.74
0.04 0.76
0.05 0.89
0.06 0.86
0.07 0.86
0.08 0.87
0.09 0.95
0.1 0.92
0.11 0.96
0.12 0.95
0.13 0.98
0.14 0.96
0.15 0.95
0.16 0.97
0.17 0.97
0.18 0.99
0.19 0.97
0.2 0.99
0.21 0.99
0.22 1
0.23 1
0.24 1
0.25 1
0.26 1
0.27 1
0.28 1
0.29 0.99
0.3 1
0.31 0.99
0.32 0.99
0.33 1
0.34 0.99
0.35 1
0.36 0.99
0.37 1
0.38 1
0.39 1
0.4 1
0.41 1
0.42 0.99
0.43 0.99
0.44 0.99
0.45 0.97
0.46 1
0.47 0.99
0.48 0.99
0.49 0.98
0.5 0.99
0.51 1
0.52 0.99
0.53 1
0.54 0.99
0.55 1
0.56 0.99
0.57 0.99
0.58 0.99
0.59 1
0.6 0.99
0.61 1
0.62 1
0.63 1
0.64 1
0.65 1
0.66 1
0.67 1
0.68 0.97
0.69 0.99
0.7 1
0.71 0.99
0.72 0.98
0.73 1
0.74 1
0.75 0.99
0.76 1
0.77 0.99
0.78 0.99
0.79 0.98
0.8 0.96
0.81 0.97
0.82 0.99
0.83 0.99
0.84 0.98
0.85 0.99
0.86 0.99
0.87 0.96
0.88 0.99
0.89 0.99
0.9 1
0.91 0.99
0.92 0.99
0.93 0.99
0.94 0.99
0.95 0.99
0.96 0.98
0.97 0.97
0.98 0.98
0.99 0.98
1 0.97
1.02 0.87
1.03 0.81
1.04 0.75
1.05 0.63
1.06 0.59
1.07 0.69
1.08 0.68
1.09 0.59
1.1 0.64
1.11 0.6
1.12 0.69
1.13 0.62
1.14 0.61
1.15 0.62
1.16 0.58
1.17 0.64
1.18 0.5
1.19 0.51
1.2 0.61
1.21 0.63
1.22 0.6
1.23 0.65
1.24 0.6
1.25 0.6
1.26 0.55
1.27 0.61
1.28 0.53
1.29 0.64
1.3 0.61
1.31 0.54
1.32 0.66
1.33 0.53
1.34 0.53
1.35 0.56
1.36 0.55
1.37 0.65
1.38 0.54
1.39 0.48
1.4 0.57
1.41 0.6
1.42 0.51
1.43 0.61
1.44 0.59
1.45 0.56
1.46 0.59
1.47 0.55
1.48 0.58
1.49 0.61
1.5 0.57
1.51 0.54
1.52 0.49
1.53 0.54
1.54 0.61
1.55 0.59
1.56 0.61
1.57 0.59
1.58 0.59
1.59 0.58
1.6 0.56
1.61 0.59
1.62 0.6
1.63 0.53
1.64 0.58
1.65 0.59
1.66 0.44
1.67 0.62
1.68 0.57
1.69 0.54
1.7 0.53
1.71 0.64
1.72 0.58
1.73 0.59
1.74 0.65
1.75 0.56
1.76 0.54
1.77 0.48
1.78 0.58
1.79 0.49
1.8 0.63
1.81 0.67
1.82 0.57
1.83 0.54
1.84 0.54
1.85 0.6
1.86 0.53
1.87 0.55
1.88 0.51
1.89 0.57
1.9 0.58
1.91 0.69
1.92 0.54
1.93 0.53
1.94 0.48
1.95 0.53
1.96 0.5
1.97 0.54
1.98 0.6
1.99 0.58
2 0.63
};
\addplot [semithick, color11, opacity=0.75, dashed]
table {%
0.01 1
0.02 1
0.03 1
0.04 1
0.05 1
0.06 1
0.07 1
0.08 1
0.09 1
0.1 1
0.11 1
0.12 1
0.13 1
0.14 1
0.15 1
0.16 1
0.17 1
0.18 1
0.19 1
0.2 1
0.21 1
0.22 1
0.23 1
0.24 1
0.25 1
0.26 1
0.27 1
0.28 1
0.29 1
0.3 1
0.31 1
0.32 1
0.33 1
0.34 1
0.35 1
0.36 1
0.37 1
0.38 1
0.39 1
0.4 1
0.41 1
0.42 1
0.43 1
0.44 1
0.45 1
0.46 1
0.47 1
0.48 1
0.49 1
0.5 1
0.51 1
0.52 1
0.53 1
0.54 1
0.55 1
0.56 1
0.57 1
0.58 1
0.59 1
0.6 1
0.61 1
0.62 1
0.63 1
0.64 1
0.65 1
0.66 1
0.67 1
0.68 1
0.69 1
0.7 1
0.71 1
0.72 1
0.73 1
0.74 1
0.75 1
0.76 1
0.77 1
0.78 1
0.79 1
0.8 1
0.81 1
0.82 1
0.83 1
0.84 1
0.85 1
0.86 1
0.87 1
0.88 1
0.89 1
0.9 1
0.91 1
0.92 1
0.93 1
0.94 1
0.95 1
0.96 1
0.97 1
0.98 1
0.99 1
1 1
1.02 1
1.03 1
1.04 1
1.05 1
1.06 0.99
1.07 1
1.08 0.99
1.09 0.98
1.1 0.99
1.11 0.98
1.12 0.98
1.13 1
1.14 0.98
1.15 0.98
1.16 0.99
1.17 0.98
1.18 0.99
1.19 0.98
1.2 1
1.21 0.99
1.22 0.99
1.23 0.98
1.24 0.98
1.25 0.99
1.26 0.98
1.27 0.97
1.28 0.99
1.29 0.99
1.3 0.99
1.31 0.98
1.32 0.99
1.33 1
1.34 0.98
1.35 0.99
1.36 0.99
1.37 0.99
1.38 0.98
1.39 1
1.4 0.97
1.41 0.98
1.42 0.98
1.43 0.99
1.44 0.98
1.45 0.94
1.46 1
1.47 0.99
1.48 0.95
1.49 0.96
1.5 0.98
1.51 0.97
1.52 0.96
1.53 0.98
1.54 0.97
1.55 0.99
1.56 0.97
1.57 0.96
1.58 0.99
1.59 0.99
1.6 0.97
1.61 0.99
1.62 0.95
1.63 0.96
1.64 0.98
1.65 0.97
1.66 0.98
1.67 0.99
1.68 0.99
1.69 0.99
1.7 1
1.71 0.97
1.72 0.98
1.73 0.99
1.74 0.97
1.75 0.98
1.76 0.98
1.77 0.94
1.78 0.97
1.79 0.97
1.8 0.98
1.81 0.97
1.82 0.95
1.83 0.99
1.84 0.96
1.85 1
1.86 0.99
1.87 1
1.88 0.98
1.89 0.99
1.9 0.97
1.91 1
1.92 1
1.93 0.99
1.94 0.97
1.95 0.99
1.96 0.99
1.97 1
1.98 1
1.99 0.99
2 0.99
};
\addplot [semithick, black, opacity=1, dash pattern=on 1pt off 1pt]
table {%
-0.0895000000000001 0.5
2.0995 0.5
};
\addplot [semithick, black, opacity=1, dash pattern=on 1pt off 1pt]
table {%
-0.0895000000000001 1
2.0995 1
};
\addplot [semithick, black, opacity=1, dash pattern=on 1pt off 1pt]
table {%
1 0.2
1 1.05
};
\end{axis}

\end{tikzpicture}

%% file: plots/decoupled2/NLUNI.tex
% This file was created by tikzplotlib v0.9.6.
\begin{tikzpicture}

\definecolor{color0}{rgb}{0.866666666666667,0.494117647058824,0.164705882352941}
\definecolor{color1}{rgb}{0.164705882352941,0.643137254901961,0.866666666666667}
\definecolor{color2}{rgb}{0.584313725490196,0.866666666666667,0.164705882352941}
\definecolor{color3}{rgb}{0.109803921568627,0.337254901960784,0.129411764705882}
\definecolor{color4}{rgb}{0.529411764705882,0.305882352941176,0.858823529411765}
\definecolor{color5}{rgb}{0.858823529411765,0.305882352941176,0.435294117647059}
\definecolor{color6}{rgb}{0.937254901960784,0.929411764705882,0.392156862745098}
\definecolor{color7}{rgb}{0.0901960784313725,0.486274509803922,0.0980392156862745}
\definecolor{color8}{rgb}{0.156862745098039,0.188235294117647,0.827450980392157}
\definecolor{color9}{rgb}{0.937254901960784,0.392156862745098,0.894117647058824}
\definecolor{color10}{rgb}{0.2,0.184313725490196,0.184313725490196}
\definecolor{color11}{rgb}{0.0156862745098039,0.803921568627451,0.976470588235294}

\begin{axis}[
tick align=outside,
tick pos=left,
x grid style={white!69.0196078431373!black},
xmajorgrids,
xmin=-0.0895, xmax=2.0995,
xtick style={color=black},
xtick={0,0.1,0.2,0.3,0.4,0.5,0.6,0.7,0.8,0.9,1,1.1,1.2,1.3,1.4,1.5,1.6,1.7,1.8,1.9,2},
xticklabels={0,,.2,,.4,,.6,,.8,,1,,20,,40,,60,,80,,100},
height=4.8cm,
width=6.5cm,
y grid style={white!69.0196078431373!black},
ymajorgrids,
ymin=0.2, ymax=1.05,
ytick style={color=black}
]
\addplot [semithick, color0, opacity=0.75]
table {%
0.01 0.58
0.02 0.64
0.03 0.75
0.04 0.86
0.05 0.86
0.06 0.91
0.07 0.97
0.08 0.95
0.09 0.96
0.1 1
0.11 1
0.12 0.99
0.13 1
0.14 1
0.15 1
0.16 0.99
0.17 1
0.18 1
0.19 1
0.2 1
0.21 1
0.22 1
0.23 1
0.24 1
0.25 1
0.26 1
0.27 1
0.28 1
0.29 1
0.3 1
0.31 1
0.32 1
0.33 1
0.34 1
0.35 1
0.36 1
0.37 1
0.38 1
0.39 1
0.4 1
0.41 1
0.42 1
0.43 1
0.44 1
0.45 1
0.46 1
0.47 1
0.48 1
0.49 1
0.5 1
0.51 1
0.52 1
0.53 1
0.54 1
0.55 1
0.56 1
0.57 1
0.58 1
0.59 1
0.6 1
0.61 1
0.62 1
0.63 1
0.64 1
0.65 1
0.66 1
0.67 1
0.68 1
0.69 1
0.7 1
0.71 1
0.72 1
0.73 1
0.74 1
0.75 1
0.76 1
0.77 1
0.78 1
0.79 1
0.8 1
0.81 1
0.82 1
0.83 1
0.84 1
0.85 1
0.86 1
0.87 1
0.88 1
0.89 1
0.9 1
0.91 1
0.92 1
0.93 1
0.94 1
0.95 1
0.96 1
0.97 1
0.98 1
0.99 1
1 1
1.02 1
1.03 0.99
1.04 0.95
1.05 0.93
1.06 0.9
1.07 0.87
1.08 0.86
1.09 0.75
1.1 0.79
1.11 0.79
1.12 0.83
1.13 0.71
1.14 0.78
1.15 0.79
1.16 0.83
1.17 0.78
1.18 0.69
1.19 0.71
1.2 0.69
1.21 0.75
1.22 0.72
1.23 0.71
1.24 0.74
1.25 0.69
1.26 0.7
1.27 0.63
1.28 0.62
1.29 0.64
1.3 0.71
1.31 0.7
1.32 0.64
1.33 0.69
1.34 0.66
1.35 0.66
1.36 0.64
1.37 0.69
1.38 0.76
1.39 0.62
1.4 0.66
1.41 0.66
1.42 0.71
1.43 0.6
1.44 0.57
1.45 0.72
1.46 0.72
1.47 0.68
1.48 0.68
1.49 0.63
1.5 0.66
1.51 0.69
1.52 0.73
1.53 0.67
1.54 0.68
1.55 0.64
1.56 0.65
1.57 0.7
1.58 0.64
1.59 0.61
1.6 0.68
1.61 0.71
1.62 0.7
1.63 0.67
1.64 0.6
1.65 0.62
1.66 0.5
1.67 0.64
1.68 0.7
1.69 0.66
1.7 0.72
1.71 0.62
1.72 0.65
1.73 0.71
1.74 0.68
1.75 0.65
1.76 0.74
1.77 0.71
1.78 0.68
1.79 0.6
1.8 0.7
1.81 0.66
1.82 0.65
1.83 0.76
1.84 0.58
1.85 0.79
1.86 0.7
1.87 0.69
1.88 0.62
1.89 0.65
1.9 0.66
1.91 0.67
1.92 0.59
1.93 0.71
1.94 0.69
1.95 0.77
1.96 0.7
1.97 0.7
1.98 0.67
1.99 0.64
2 0.73
};
\addplot [semithick, color1, opacity=0.75]
table {%
0.01 0.5
0.02 0.42
0.03 0.52
0.04 0.48
0.05 0.48
0.06 0.49
0.07 0.55
0.08 0.54
0.09 0.57
0.1 0.47
0.11 0.59
0.12 0.49
0.13 0.5
0.14 0.52
0.15 0.56
0.16 0.49
0.17 0.47
0.18 0.53
0.19 0.47
0.2 0.53
0.21 0.56
0.22 0.45
0.23 0.48
0.24 0.56
0.25 0.53
0.26 0.52
0.27 0.55
0.28 0.59
0.29 0.52
0.3 0.54
0.31 0.49
0.32 0.65
0.33 0.59
0.34 0.56
0.35 0.64
0.36 0.68
0.37 0.69
0.38 0.58
0.39 0.52
0.4 0.63
0.41 0.58
0.42 0.67
0.43 0.59
0.44 0.61
0.45 0.69
0.46 0.65
0.47 0.62
0.48 0.63
0.49 0.63
0.5 0.69
0.51 0.6
0.52 0.61
0.53 0.69
0.54 0.67
0.55 0.7
0.56 0.74
0.57 0.66
0.58 0.68
0.59 0.67
0.6 0.68
0.61 0.68
0.62 0.71
0.63 0.72
0.64 0.72
0.65 0.73
0.66 0.64
0.67 0.73
0.68 0.72
0.69 0.8
0.7 0.75
0.71 0.73
0.72 0.74
0.73 0.7
0.74 0.8
0.75 0.75
0.76 0.77
0.77 0.78
0.78 0.8
0.79 0.77
0.8 0.84
0.81 0.78
0.82 0.81
0.83 0.78
0.84 0.75
0.85 0.87
0.86 0.83
0.87 0.78
0.88 0.71
0.89 0.8
0.9 0.84
0.91 0.81
0.92 0.83
0.93 0.84
0.94 0.89
0.95 0.76
0.96 0.89
0.97 0.83
0.98 0.85
0.99 0.84
1 0.86
1.02 0.83
1.03 0.86
1.04 0.78
1.05 0.8
1.06 0.77
1.07 0.78
1.08 0.67
1.09 0.68
1.1 0.7
1.11 0.63
1.12 0.65
1.13 0.69
1.14 0.6
1.15 0.64
1.16 0.59
1.17 0.53
1.18 0.5
1.19 0.63
1.2 0.52
1.21 0.53
1.22 0.56
1.23 0.53
1.24 0.4
1.25 0.58
1.26 0.5
1.27 0.53
1.28 0.46
1.29 0.53
1.3 0.57
1.31 0.48
1.32 0.5
1.33 0.52
1.34 0.48
1.35 0.6
1.36 0.62
1.37 0.48
1.38 0.47
1.39 0.49
1.4 0.49
1.41 0.5
1.42 0.6
1.43 0.52
1.44 0.49
1.45 0.55
1.46 0.6
1.47 0.44
1.48 0.46
1.49 0.48
1.5 0.59
1.51 0.45
1.52 0.49
1.53 0.53
1.54 0.55
1.55 0.49
1.56 0.46
1.57 0.5
1.58 0.44
1.59 0.52
1.6 0.48
1.61 0.57
1.62 0.45
1.63 0.48
1.64 0.51
1.65 0.56
1.66 0.43
1.67 0.56
1.68 0.43
1.69 0.47
1.7 0.52
1.71 0.51
1.72 0.41
1.73 0.47
1.74 0.45
1.75 0.48
1.76 0.53
1.77 0.48
1.78 0.49
1.79 0.49
1.8 0.45
1.81 0.49
1.82 0.51
1.83 0.5
1.84 0.43
1.85 0.57
1.86 0.51
1.87 0.43
1.88 0.44
1.89 0.52
1.9 0.51
1.91 0.5
1.92 0.4
1.93 0.51
1.94 0.55
1.95 0.5
1.96 0.48
1.97 0.45
1.98 0.49
1.99 0.47
2 0.47
};
\addplot [semithick, color2, opacity=0.75]
table {%
0.01 0.6
0.02 0.55
0.03 0.45
0.04 0.48
0.05 0.55
0.06 0.51
0.07 0.54
0.08 0.58
0.09 0.51
0.1 0.49
0.11 0.45
0.12 0.54
0.13 0.42
0.14 0.48
0.15 0.54
0.16 0.47
0.17 0.48
0.18 0.48
0.19 0.44
0.2 0.51
0.21 0.52
0.22 0.43
0.23 0.47
0.24 0.57
0.25 0.53
0.26 0.51
0.27 0.55
0.28 0.55
0.29 0.46
0.3 0.55
0.31 0.46
0.32 0.65
0.33 0.58
0.34 0.54
0.35 0.6
0.36 0.65
0.37 0.69
0.38 0.58
0.39 0.49
0.4 0.59
0.41 0.54
0.42 0.63
0.43 0.58
0.44 0.6
0.45 0.68
0.46 0.6
0.47 0.62
0.48 0.63
0.49 0.57
0.5 0.67
0.51 0.56
0.52 0.57
0.53 0.66
0.54 0.64
0.55 0.68
0.56 0.7
0.57 0.62
0.58 0.66
0.59 0.66
0.6 0.67
0.61 0.66
0.62 0.7
0.63 0.71
0.64 0.69
0.65 0.72
0.66 0.62
0.67 0.71
0.68 0.7
0.69 0.77
0.7 0.73
0.71 0.72
0.72 0.68
0.73 0.68
0.74 0.78
0.75 0.75
0.76 0.76
0.77 0.76
0.78 0.79
0.79 0.76
0.8 0.82
0.81 0.74
0.82 0.77
0.83 0.75
0.84 0.73
0.85 0.86
0.86 0.8
0.87 0.77
0.88 0.69
0.89 0.8
0.9 0.85
0.91 0.78
0.92 0.78
0.93 0.82
0.94 0.85
0.95 0.72
0.96 0.87
0.97 0.81
0.98 0.82
0.99 0.83
1 0.83
1.02 0.83
1.03 0.85
1.04 0.76
1.05 0.79
1.06 0.75
1.07 0.78
1.08 0.68
1.09 0.68
1.1 0.67
1.11 0.64
1.12 0.65
1.13 0.68
1.14 0.62
1.15 0.65
1.16 0.6
1.17 0.53
1.18 0.5
1.19 0.62
1.2 0.54
1.21 0.54
1.22 0.52
1.23 0.55
1.24 0.41
1.25 0.57
1.26 0.52
1.27 0.54
1.28 0.44
1.29 0.52
1.3 0.57
1.31 0.44
1.32 0.48
1.33 0.53
1.34 0.49
1.35 0.6
1.36 0.62
1.37 0.5
1.38 0.49
1.39 0.49
1.4 0.48
1.41 0.52
1.42 0.58
1.43 0.52
1.44 0.47
1.45 0.54
1.46 0.59
1.47 0.43
1.48 0.46
1.49 0.46
1.5 0.57
1.51 0.44
1.52 0.49
1.53 0.54
1.54 0.57
1.55 0.49
1.56 0.48
1.57 0.5
1.58 0.44
1.59 0.52
1.6 0.48
1.61 0.58
1.62 0.44
1.63 0.46
1.64 0.51
1.65 0.54
1.66 0.42
1.67 0.55
1.68 0.43
1.69 0.51
1.7 0.51
1.71 0.5
1.72 0.42
1.73 0.47
1.74 0.45
1.75 0.48
1.76 0.51
1.77 0.48
1.78 0.49
1.79 0.49
1.8 0.46
1.81 0.47
1.82 0.51
1.83 0.49
1.84 0.45
1.85 0.56
1.86 0.48
1.87 0.41
1.88 0.45
1.89 0.53
1.9 0.46
1.91 0.52
1.92 0.39
1.93 0.53
1.94 0.55
1.95 0.51
1.96 0.46
1.97 0.44
1.98 0.51
1.99 0.47
2 0.48
};
\addplot [semithick, color3, opacity=0.75]
table {%
0.01 0.76
0.02 0.87
0.03 0.92
0.04 0.98
0.05 0.97
0.06 0.98
0.07 0.99
0.08 0.99
0.09 0.99
0.1 0.99
0.11 1
0.12 0.98
0.13 1
0.14 1
0.15 1
0.16 1
0.17 1
0.18 1
0.19 0.99
0.2 0.99
0.21 1
0.22 1
0.23 1
0.24 1
0.25 1
0.26 1
0.27 1
0.28 1
0.29 1
0.3 1
0.31 1
0.32 1
0.33 0.99
0.34 1
0.35 1
0.36 1
0.37 1
0.38 1
0.39 0.99
0.4 0.99
0.41 1
0.42 1
0.43 1
0.44 1
0.45 1
0.46 1
0.47 0.99
0.48 1
0.49 1
0.5 1
0.51 1
0.52 1
0.53 1
0.54 1
0.55 1
0.56 0.99
0.57 1
0.58 1
0.59 1
0.6 1
0.61 1
0.62 1
0.63 1
0.64 1
0.65 1
0.66 0.98
0.67 0.99
0.68 1
0.69 1
0.7 1
0.71 1
0.72 1
0.73 0.98
0.74 0.99
0.75 1
0.76 1
0.77 1
0.78 1
0.79 1
0.8 1
0.81 1
0.82 1
0.83 0.98
0.84 1
0.85 1
0.86 1
0.87 1
0.88 0.98
0.89 1
0.9 1
0.91 1
0.92 0.99
0.93 1
0.94 1
0.95 1
0.96 1
0.97 0.99
0.98 1
0.99 1
1 1
1.02 1
1.03 1
1.04 1
1.05 0.95
1.06 0.96
1.07 0.98
1.08 0.94
1.09 0.92
1.1 0.92
1.11 0.92
1.12 0.9
1.13 0.9
1.14 0.9
1.15 0.92
1.16 0.94
1.17 0.94
1.18 0.88
1.19 0.91
1.2 0.89
1.21 0.84
1.22 0.88
1.23 0.86
1.24 0.9
1.25 0.89
1.26 0.85
1.27 0.86
1.28 0.79
1.29 0.75
1.3 0.89
1.31 0.83
1.32 0.8
1.33 0.84
1.34 0.85
1.35 0.89
1.36 0.82
1.37 0.86
1.38 0.88
1.39 0.84
1.4 0.83
1.41 0.84
1.42 0.85
1.43 0.81
1.44 0.79
1.45 0.88
1.46 0.87
1.47 0.88
1.48 0.86
1.49 0.82
1.5 0.82
1.51 0.86
1.52 0.91
1.53 0.87
1.54 0.86
1.55 0.81
1.56 0.79
1.57 0.83
1.58 0.78
1.59 0.77
1.6 0.86
1.61 0.81
1.62 0.8
1.63 0.82
1.64 0.82
1.65 0.83
1.66 0.77
1.67 0.86
1.68 0.84
1.69 0.87
1.7 0.84
1.71 0.83
1.72 0.87
1.73 0.88
1.74 0.85
1.75 0.84
1.76 0.9
1.77 0.89
1.78 0.84
1.79 0.79
1.8 0.88
1.81 0.84
1.82 0.88
1.83 0.87
1.84 0.8
1.85 0.89
1.86 0.85
1.87 0.84
1.88 0.83
1.89 0.8
1.9 0.8
1.91 0.8
1.92 0.8
1.93 0.86
1.94 0.83
1.95 0.86
1.96 0.83
1.97 0.89
1.98 0.84
1.99 0.91
2 0.83
};
\addplot [semithick, color4, opacity=0.75]
table {%
0.01 0.76
0.02 0.87
0.03 0.92
0.04 0.98
0.05 0.97
0.06 0.98
0.07 0.99
0.08 0.99
0.09 0.99
0.1 0.99
0.11 1
0.12 0.98
0.13 1
0.14 1
0.15 1
0.16 1
0.17 1
0.18 1
0.19 0.99
0.2 0.99
0.21 1
0.22 1
0.23 1
0.24 1
0.25 1
0.26 1
0.27 1
0.28 1
0.29 1
0.3 1
0.31 1
0.32 1
0.33 0.99
0.34 1
0.35 1
0.36 1
0.37 1
0.38 1
0.39 0.99
0.4 1
0.41 1
0.42 1
0.43 1
0.44 1
0.45 1
0.46 1
0.47 1
0.48 1
0.49 1
0.5 1
0.51 1
0.52 1
0.53 1
0.54 1
0.55 1
0.56 0.99
0.57 1
0.58 1
0.59 1
0.6 0.99
0.61 1
0.62 1
0.63 1
0.64 1
0.65 1
0.66 0.99
0.67 0.99
0.68 1
0.69 1
0.7 1
0.71 1
0.72 1
0.73 0.98
0.74 0.99
0.75 1
0.76 1
0.77 1
0.78 1
0.79 1
0.8 1
0.81 1
0.82 1
0.83 0.99
0.84 1
0.85 1
0.86 1
0.87 1
0.88 0.98
0.89 1
0.9 1
0.91 1
0.92 1
0.93 1
0.94 1
0.95 1
0.96 1
0.97 1
0.98 1
0.99 1
1 1
1.02 1
1.03 1
1.04 1
1.05 0.95
1.06 0.96
1.07 0.98
1.08 0.94
1.09 0.92
1.1 0.93
1.11 0.93
1.12 0.9
1.13 0.9
1.14 0.91
1.15 0.92
1.16 0.94
1.17 0.94
1.18 0.88
1.19 0.91
1.2 0.9
1.21 0.85
1.22 0.87
1.23 0.86
1.24 0.89
1.25 0.9
1.26 0.85
1.27 0.85
1.28 0.79
1.29 0.76
1.3 0.89
1.31 0.83
1.32 0.8
1.33 0.84
1.34 0.84
1.35 0.89
1.36 0.82
1.37 0.86
1.38 0.88
1.39 0.84
1.4 0.83
1.41 0.86
1.42 0.85
1.43 0.81
1.44 0.79
1.45 0.88
1.46 0.86
1.47 0.88
1.48 0.86
1.49 0.82
1.5 0.81
1.51 0.86
1.52 0.91
1.53 0.87
1.54 0.86
1.55 0.81
1.56 0.79
1.57 0.82
1.58 0.77
1.59 0.77
1.6 0.86
1.61 0.81
1.62 0.8
1.63 0.82
1.64 0.82
1.65 0.83
1.66 0.79
1.67 0.87
1.68 0.85
1.69 0.87
1.7 0.84
1.71 0.84
1.72 0.87
1.73 0.89
1.74 0.85
1.75 0.84
1.76 0.9
1.77 0.89
1.78 0.84
1.79 0.79
1.8 0.88
1.81 0.84
1.82 0.88
1.83 0.87
1.84 0.81
1.85 0.89
1.86 0.85
1.87 0.84
1.88 0.83
1.89 0.8
1.9 0.8
1.91 0.8
1.92 0.8
1.93 0.86
1.94 0.83
1.95 0.86
1.96 0.83
1.97 0.88
1.98 0.84
1.99 0.92
2 0.83
};
\addplot [semithick, color5, opacity=0.75]
table {%
0.01 0.73
0.02 0.83
0.03 0.88
0.04 0.91
0.05 0.93
0.06 0.93
0.07 0.98
0.08 0.97
0.09 0.98
0.1 0.98
0.11 0.95
0.12 0.96
0.13 0.97
0.14 0.99
0.15 1
0.16 0.97
0.17 0.98
0.18 1
0.19 0.98
0.2 0.98
0.21 0.99
0.22 1
0.23 1
0.24 1
0.25 1
0.26 0.98
0.27 1
0.28 0.99
0.29 0.99
0.3 0.99
0.31 0.99
0.32 1
0.33 0.99
0.34 1
0.35 1
0.36 0.99
0.37 1
0.38 1
0.39 0.99
0.4 0.99
0.41 1
0.42 1
0.43 0.99
0.44 1
0.45 0.97
0.46 0.99
0.47 0.99
0.48 1
0.49 1
0.5 1
0.51 1
0.52 0.99
0.53 1
0.54 1
0.55 1
0.56 0.98
0.57 0.99
0.58 1
0.59 1
0.6 1
0.61 1
0.62 1
0.63 1
0.64 0.98
0.65 1
0.66 0.98
0.67 0.97
0.68 1
0.69 1
0.7 0.99
0.71 1
0.72 1
0.73 0.98
0.74 0.99
0.75 1
0.76 1
0.77 1
0.78 1
0.79 0.99
0.8 1
0.81 1
0.82 1
0.83 0.98
0.84 1
0.85 1
0.86 1
0.87 1
0.88 0.99
0.89 0.99
0.9 0.99
0.91 1
0.92 1
0.93 1
0.94 1
0.95 1
0.96 1
0.97 0.99
0.98 1
0.99 1
1 1
1.02 0.98
1.03 0.98
1.04 0.99
1.05 0.94
1.06 0.95
1.07 0.94
1.08 0.95
1.09 0.87
1.1 0.9
1.11 0.94
1.12 0.93
1.13 0.91
1.14 0.91
1.15 0.91
1.16 0.94
1.17 0.89
1.18 0.84
1.19 0.92
1.2 0.91
1.21 0.83
1.22 0.91
1.23 0.87
1.24 0.92
1.25 0.93
1.26 0.89
1.27 0.89
1.28 0.88
1.29 0.79
1.3 0.9
1.31 0.88
1.32 0.79
1.33 0.89
1.34 0.89
1.35 0.9
1.36 0.81
1.37 0.85
1.38 0.91
1.39 0.86
1.4 0.87
1.41 0.85
1.42 0.84
1.43 0.85
1.44 0.8
1.45 0.89
1.46 0.89
1.47 0.87
1.48 0.87
1.49 0.84
1.5 0.88
1.51 0.88
1.52 0.91
1.53 0.89
1.54 0.93
1.55 0.88
1.56 0.85
1.57 0.89
1.58 0.8
1.59 0.8
1.6 0.91
1.61 0.84
1.62 0.82
1.63 0.9
1.64 0.87
1.65 0.85
1.66 0.8
1.67 0.89
1.68 0.88
1.69 0.91
1.7 0.85
1.71 0.87
1.72 0.9
1.73 0.92
1.74 0.88
1.75 0.89
1.76 0.92
1.77 0.94
1.78 0.85
1.79 0.84
1.8 0.89
1.81 0.9
1.82 0.89
1.83 0.9
1.84 0.83
1.85 0.91
1.86 0.89
1.87 0.84
1.88 0.86
1.89 0.84
1.9 0.83
1.91 0.86
1.92 0.85
1.93 0.89
1.94 0.85
1.95 0.91
1.96 0.89
1.97 0.91
1.98 0.87
1.99 0.94
2 0.86
};
\addplot [semithick, color6, opacity=0.75, dashed]
table {%
0.01 0.98
0.02 1
0.03 0.98
0.04 1
0.05 1
0.06 0.98
0.07 0.99
0.08 1
0.09 1
0.1 1
0.11 1
0.12 1
0.13 1
0.14 1
0.15 1
0.16 1
0.17 1
0.18 1
0.19 1
0.2 1
0.21 1
0.22 1
0.23 1
0.24 1
0.25 1
0.26 1
0.27 1
0.28 1
0.29 1
0.3 1
0.31 1
0.32 1
0.33 1
0.34 1
0.35 1
0.36 1
0.37 1
0.38 1
0.39 1
0.4 1
0.41 1
0.42 1
0.43 1
0.44 1
0.45 1
0.46 1
0.47 1
0.48 1
0.49 1
0.5 1
0.51 1
0.52 1
0.53 1
0.54 1
0.55 1
0.56 1
0.57 1
0.58 1
0.59 1
0.6 1
0.61 1
0.62 1
0.63 1
0.64 1
0.65 1
0.66 1
0.67 1
0.68 1
0.69 1
0.7 1
0.71 1
0.72 1
0.73 1
0.74 1
0.75 1
0.76 1
0.77 1
0.78 1
0.79 1
0.8 1
0.81 1
0.82 1
0.83 1
0.84 1
0.85 1
0.86 1
0.87 1
0.88 1
0.89 1
0.9 1
0.91 1
0.92 1
0.93 1
0.94 1
0.95 1
0.96 1
0.97 1
0.98 1
0.99 1
1 1
1.02 1
1.03 1
1.04 1
1.05 1
1.06 1
1.07 1
1.08 1
1.09 1
1.1 1
1.11 1
1.12 1
1.13 1
1.14 0.98
1.15 0.99
1.16 0.99
1.17 0.99
1.18 0.98
1.19 1
1.2 0.98
1.21 0.97
1.22 0.98
1.23 0.98
1.24 0.97
1.25 0.98
1.26 0.97
1.27 0.99
1.28 0.99
1.29 1
1.3 0.99
1.31 0.98
1.32 0.98
1.33 0.95
1.34 1
1.35 0.99
1.36 0.99
1.37 0.95
1.38 1
1.39 0.97
1.4 0.95
1.41 0.96
1.42 0.99
1.43 0.96
1.44 0.99
1.45 0.97
1.46 0.99
1.47 0.95
1.48 0.97
1.49 0.98
1.5 0.99
1.51 0.98
1.52 0.99
1.53 0.98
1.54 1
1.55 0.98
1.56 0.97
1.57 0.99
1.58 0.98
1.59 0.98
1.6 0.96
1.61 0.99
1.62 0.99
1.63 0.95
1.64 0.97
1.65 0.96
1.66 0.99
1.67 1
1.68 1
1.69 0.98
1.7 0.99
1.71 0.97
1.72 0.97
1.73 0.96
1.74 1
1.75 0.99
1.76 0.99
1.77 0.98
1.78 0.97
1.79 0.96
1.8 0.97
1.81 0.98
1.82 1
1.83 0.98
1.84 0.97
1.85 0.97
1.86 0.94
1.87 1
1.88 0.98
1.89 0.97
1.9 0.99
1.91 0.97
1.92 0.98
1.93 0.99
1.94 0.97
1.95 1
1.96 0.95
1.97 0.99
1.98 0.95
1.99 0.98
2 0.97
};
\addplot [semithick, color7, opacity=0.75, dashed]
table {%
0.01 0.98
0.02 1
0.03 0.98
0.04 1
0.05 1
0.06 0.98
0.07 0.99
0.08 1
0.09 1
0.1 1
0.11 1
0.12 1
0.13 1
0.14 1
0.15 1
0.16 1
0.17 1
0.18 1
0.19 1
0.2 1
0.21 1
0.22 1
0.23 1
0.24 1
0.25 1
0.26 1
0.27 1
0.28 1
0.29 1
0.3 1
0.31 1
0.32 1
0.33 1
0.34 1
0.35 1
0.36 1
0.37 1
0.38 1
0.39 1
0.4 1
0.41 1
0.42 1
0.43 1
0.44 1
0.45 1
0.46 1
0.47 1
0.48 1
0.49 1
0.5 1
0.51 1
0.52 1
0.53 1
0.54 1
0.55 1
0.56 1
0.57 1
0.58 1
0.59 1
0.6 1
0.61 1
0.62 1
0.63 1
0.64 1
0.65 1
0.66 1
0.67 1
0.68 1
0.69 1
0.7 1
0.71 1
0.72 1
0.73 1
0.74 1
0.75 1
0.76 1
0.77 1
0.78 1
0.79 1
0.8 1
0.81 1
0.82 1
0.83 1
0.84 1
0.85 1
0.86 1
0.87 1
0.88 1
0.89 1
0.9 1
0.91 1
0.92 1
0.93 1
0.94 1
0.95 1
0.96 1
0.97 1
0.98 1
0.99 1
1 1
1.02 1
1.03 1
1.04 1
1.05 1
1.06 1
1.07 1
1.08 1
1.09 1
1.1 1
1.11 1
1.12 1
1.13 1
1.14 0.98
1.15 0.99
1.16 0.99
1.17 0.99
1.18 0.98
1.19 1
1.2 0.98
1.21 0.97
1.22 0.98
1.23 0.98
1.24 0.97
1.25 0.98
1.26 0.97
1.27 0.99
1.28 0.99
1.29 1
1.3 0.99
1.31 0.98
1.32 0.98
1.33 0.95
1.34 1
1.35 0.99
1.36 0.99
1.37 0.95
1.38 1
1.39 0.97
1.4 0.95
1.41 0.96
1.42 0.99
1.43 0.96
1.44 0.99
1.45 0.97
1.46 0.99
1.47 0.95
1.48 0.97
1.49 0.98
1.5 0.99
1.51 0.98
1.52 0.99
1.53 0.98
1.54 1
1.55 0.98
1.56 0.97
1.57 0.99
1.58 0.98
1.59 0.98
1.6 0.96
1.61 0.99
1.62 0.99
1.63 0.95
1.64 0.97
1.65 0.96
1.66 0.99
1.67 1
1.68 1
1.69 0.98
1.7 0.99
1.71 0.97
1.72 0.97
1.73 0.96
1.74 1
1.75 0.99
1.76 0.99
1.77 0.98
1.78 0.97
1.79 0.96
1.8 0.97
1.81 0.98
1.82 1
1.83 0.98
1.84 0.97
1.85 0.97
1.86 0.94
1.87 1
1.88 0.98
1.89 0.97
1.9 0.99
1.91 0.97
1.92 0.98
1.93 0.99
1.94 0.97
1.95 1
1.96 0.95
1.97 0.99
1.98 0.95
1.99 0.98
2 0.97
};
\addplot [semithick, color8, opacity=0.75, dashed]
table {%
0.01 0.99
0.02 1
0.03 1
0.04 0.99
0.05 1
0.06 1
0.07 1
0.08 1
0.09 1
0.1 1
0.11 1
0.12 1
0.13 1
0.14 1
0.15 1
0.16 1
0.17 1
0.18 1
0.19 1
0.2 1
0.21 1
0.22 1
0.23 1
0.24 1
0.25 1
0.26 1
0.27 1
0.28 1
0.29 1
0.3 1
0.31 1
0.32 1
0.33 1
0.34 1
0.35 1
0.36 1
0.37 1
0.38 1
0.39 1
0.4 1
0.41 1
0.42 1
0.43 1
0.44 1
0.45 1
0.46 1
0.47 1
0.48 1
0.49 1
0.5 1
0.51 1
0.52 1
0.53 1
0.54 1
0.55 1
0.56 1
0.57 1
0.58 1
0.59 1
0.6 1
0.61 1
0.62 1
0.63 1
0.64 1
0.65 1
0.66 1
0.67 1
0.68 1
0.69 1
0.7 1
0.71 1
0.72 1
0.73 1
0.74 1
0.75 1
0.76 1
0.77 1
0.78 1
0.79 1
0.8 1
0.81 1
0.82 1
0.83 1
0.84 1
0.85 1
0.86 1
0.87 1
0.88 1
0.89 1
0.9 1
0.91 1
0.92 1
0.93 1
0.94 1
0.95 1
0.96 1
0.97 1
0.98 1
0.99 1
1 1
1.02 1
1.03 1
1.04 1
1.05 1
1.06 1
1.07 1
1.08 1
1.09 1
1.1 1
1.11 1
1.12 0.99
1.13 1
1.14 1
1.15 1
1.16 1
1.17 1
1.18 1
1.19 0.99
1.2 1
1.21 0.99
1.22 1
1.23 0.99
1.24 1
1.25 1
1.26 0.99
1.27 1
1.28 0.99
1.29 0.98
1.3 1
1.31 0.99
1.32 1
1.33 1
1.34 1
1.35 1
1.36 0.99
1.37 1
1.38 1
1.39 1
1.4 0.98
1.41 0.99
1.42 1
1.43 1
1.44 1
1.45 0.99
1.46 1
1.47 1
1.48 0.99
1.49 1
1.5 0.99
1.51 1
1.52 1
1.53 0.99
1.54 1
1.55 1
1.56 1
1.57 1
1.58 0.99
1.59 0.98
1.6 0.99
1.61 1
1.62 1
1.63 1
1.64 1
1.65 0.98
1.66 0.99
1.67 0.99
1.68 0.99
1.69 0.98
1.7 0.99
1.71 1
1.72 0.99
1.73 1
1.74 0.99
1.75 0.99
1.76 1
1.77 0.99
1.78 1
1.79 0.98
1.8 1
1.81 1
1.82 0.98
1.83 0.98
1.84 1
1.85 1
1.86 0.98
1.87 0.98
1.88 0.99
1.89 1
1.9 0.99
1.91 1
1.92 0.99
1.93 1
1.94 1
1.95 1
1.96 0.99
1.97 0.98
1.98 1
1.99 1
2 1
};
\addplot [semithick, color9, opacity=0.75, dashed]
table {%
0.01 0.65
0.02 0.68
0.03 0.82
0.04 0.91
0.05 0.89
0.06 0.96
0.07 0.91
0.08 0.99
0.09 0.98
0.1 1
0.11 0.99
0.12 0.98
0.13 1
0.14 1
0.15 1
0.16 1
0.17 1
0.18 1
0.19 1
0.2 1
0.21 1
0.22 1
0.23 1
0.24 1
0.25 1
0.26 1
0.27 1
0.28 1
0.29 1
0.3 1
0.31 1
0.32 1
0.33 1
0.34 1
0.35 1
0.36 1
0.37 1
0.38 1
0.39 1
0.4 1
0.41 1
0.42 1
0.43 1
0.44 1
0.45 1
0.46 1
0.47 1
0.48 1
0.49 1
0.5 1
0.51 1
0.52 1
0.53 1
0.54 1
0.55 1
0.56 1
0.57 1
0.58 1
0.59 1
0.6 1
0.61 1
0.62 1
0.63 1
0.64 1
0.65 1
0.66 1
0.67 1
0.68 1
0.69 1
0.7 1
0.71 1
0.72 1
0.73 1
0.74 1
0.75 1
0.76 1
0.77 1
0.78 1
0.79 1
0.8 1
0.81 1
0.82 1
0.83 1
0.84 1
0.85 1
0.86 1
0.87 1
0.88 1
0.89 1
0.9 1
0.91 1
0.92 1
0.93 1
0.94 1
0.95 1
0.96 1
0.97 1
0.98 1
0.99 1
1 1
1.02 1
1.03 1
1.04 0.99
1.05 0.96
1.06 0.98
1.07 0.92
1.08 0.86
1.09 0.84
1.1 0.86
1.11 0.83
1.12 0.82
1.13 0.77
1.14 0.85
1.15 0.82
1.16 0.8
1.17 0.83
1.18 0.78
1.19 0.74
1.2 0.77
1.21 0.74
1.22 0.79
1.23 0.75
1.24 0.66
1.25 0.71
1.26 0.7
1.27 0.66
1.28 0.71
1.29 0.68
1.3 0.69
1.31 0.69
1.32 0.66
1.33 0.72
1.34 0.73
1.35 0.68
1.36 0.69
1.37 0.73
1.38 0.77
1.39 0.74
1.4 0.74
1.41 0.63
1.42 0.69
1.43 0.76
1.44 0.61
1.45 0.71
1.46 0.74
1.47 0.7
1.48 0.74
1.49 0.7
1.5 0.73
1.51 0.69
1.52 0.71
1.53 0.75
1.54 0.76
1.55 0.63
1.56 0.63
1.57 0.65
1.58 0.66
1.59 0.59
1.6 0.71
1.61 0.7
1.62 0.69
1.63 0.73
1.64 0.68
1.65 0.65
1.66 0.54
1.67 0.63
1.68 0.71
1.69 0.64
1.7 0.73
1.71 0.72
1.72 0.76
1.73 0.76
1.74 0.73
1.75 0.68
1.76 0.71
1.77 0.77
1.78 0.7
1.79 0.65
1.8 0.65
1.81 0.69
1.82 0.59
1.83 0.68
1.84 0.69
1.85 0.77
1.86 0.63
1.87 0.69
1.88 0.69
1.89 0.68
1.9 0.74
1.91 0.64
1.92 0.6
1.93 0.73
1.94 0.66
1.95 0.75
1.96 0.68
1.97 0.71
1.98 0.64
1.99 0.66
2 0.71
};
\addplot [semithick, color10, opacity=0.75, dashed]
table {%
0.01 0.53
0.02 0.58
0.03 0.66
0.04 0.75
0.05 0.83
0.06 0.77
0.07 0.88
0.08 0.9
0.09 0.9
0.1 0.98
0.11 0.94
0.12 0.95
0.13 0.99
0.14 0.99
0.15 0.99
0.16 0.99
0.17 0.98
0.18 1
0.19 1
0.2 1
0.21 1
0.22 1
0.23 1
0.24 1
0.25 1
0.26 1
0.27 1
0.28 1
0.29 1
0.3 1
0.31 1
0.32 1
0.33 1
0.34 1
0.35 1
0.36 1
0.37 1
0.38 1
0.39 1
0.4 1
0.41 1
0.42 1
0.43 1
0.44 1
0.45 1
0.46 1
0.47 1
0.48 1
0.49 1
0.5 1
0.51 1
0.52 1
0.53 1
0.54 1
0.55 1
0.56 1
0.57 1
0.58 1
0.59 1
0.6 1
0.61 1
0.62 1
0.63 1
0.64 1
0.65 1
0.66 1
0.67 1
0.68 1
0.69 1
0.7 1
0.71 1
0.72 1
0.73 1
0.74 1
0.75 1
0.76 1
0.77 1
0.78 1
0.79 1
0.8 1
0.81 1
0.82 1
0.83 0.99
0.84 1
0.85 1
0.86 1
0.87 1
0.88 1
0.89 1
0.9 1
0.91 1
0.92 1
0.93 1
0.94 1
0.95 1
0.96 1
0.97 1
0.98 1
0.99 1
1 1
1.02 1
1.03 1
1.04 0.92
1.05 0.91
1.06 0.86
1.07 0.81
1.08 0.73
1.09 0.78
1.1 0.76
1.11 0.65
1.12 0.67
1.13 0.65
1.14 0.69
1.15 0.63
1.16 0.74
1.17 0.66
1.18 0.56
1.19 0.72
1.2 0.62
1.21 0.66
1.22 0.56
1.23 0.56
1.24 0.54
1.25 0.55
1.26 0.64
1.27 0.52
1.28 0.57
1.29 0.49
1.3 0.6
1.31 0.5
1.32 0.54
1.33 0.56
1.34 0.47
1.35 0.56
1.36 0.54
1.37 0.5
1.38 0.56
1.39 0.49
1.4 0.56
1.41 0.46
1.42 0.62
1.43 0.44
1.44 0.44
1.45 0.66
1.46 0.61
1.47 0.6
1.48 0.57
1.49 0.58
1.5 0.58
1.51 0.61
1.52 0.57
1.53 0.52
1.54 0.55
1.55 0.51
1.56 0.53
1.57 0.48
1.58 0.48
1.59 0.43
1.6 0.55
1.61 0.52
1.62 0.49
1.63 0.55
1.64 0.48
1.65 0.48
1.66 0.44
1.67 0.52
1.68 0.56
1.69 0.48
1.7 0.56
1.71 0.49
1.72 0.48
1.73 0.61
1.74 0.51
1.75 0.44
1.76 0.63
1.77 0.65
1.78 0.5
1.79 0.51
1.8 0.52
1.81 0.54
1.82 0.43
1.83 0.63
1.84 0.51
1.85 0.52
1.86 0.44
1.87 0.54
1.88 0.53
1.89 0.54
1.9 0.54
1.91 0.61
1.92 0.46
1.93 0.53
1.94 0.53
1.95 0.47
1.96 0.45
1.97 0.52
1.98 0.44
1.99 0.46
2 0.6
};
\addplot [semithick, color11, opacity=0.75, dashed]
table {%
0.01 0.98
0.02 1
0.03 1
0.04 1
0.05 1
0.06 1
0.07 1
0.08 1
0.09 1
0.1 1
0.11 1
0.12 1
0.13 1
0.14 1
0.15 1
0.16 1
0.17 1
0.18 1
0.19 1
0.2 1
0.21 1
0.22 1
0.23 1
0.24 1
0.25 1
0.26 1
0.27 1
0.28 1
0.29 1
0.3 1
0.31 1
0.32 1
0.33 1
0.34 1
0.35 1
0.36 1
0.37 1
0.38 1
0.39 1
0.4 1
0.41 1
0.42 1
0.43 1
0.44 1
0.45 1
0.46 1
0.47 1
0.48 1
0.49 1
0.5 1
0.51 1
0.52 1
0.53 1
0.54 1
0.55 1
0.56 1
0.57 1
0.58 1
0.59 1
0.6 1
0.61 1
0.62 1
0.63 1
0.64 1
0.65 1
0.66 1
0.67 1
0.68 1
0.69 1
0.7 1
0.71 1
0.72 1
0.73 1
0.74 1
0.75 1
0.76 1
0.77 1
0.78 1
0.79 1
0.8 1
0.81 1
0.82 1
0.83 1
0.84 1
0.85 1
0.86 1
0.87 1
0.88 1
0.89 1
0.9 1
0.91 1
0.92 1
0.93 1
0.94 1
0.95 1
0.96 1
0.97 1
0.98 1
0.99 1
1 1
1.02 1
1.03 1
1.04 1
1.05 1
1.06 1
1.07 1
1.08 1
1.09 1
1.1 1
1.11 0.99
1.12 0.99
1.13 0.99
1.14 0.98
1.15 0.99
1.16 1
1.17 1
1.18 1
1.19 1
1.2 1
1.21 0.95
1.22 0.97
1.23 0.97
1.24 0.98
1.25 0.99
1.26 0.96
1.27 1
1.28 0.97
1.29 0.96
1.3 0.97
1.31 0.98
1.32 1
1.33 0.97
1.34 0.97
1.35 0.98
1.36 0.98
1.37 0.95
1.38 0.97
1.39 0.99
1.4 0.96
1.41 0.99
1.42 0.96
1.43 0.98
1.44 0.99
1.45 0.97
1.46 0.98
1.47 0.96
1.48 0.99
1.49 0.98
1.5 0.99
1.51 1
1.52 0.99
1.53 0.96
1.54 0.99
1.55 0.98
1.56 0.98
1.57 0.99
1.58 0.97
1.59 0.96
1.6 0.98
1.61 0.97
1.62 0.99
1.63 0.98
1.64 0.98
1.65 0.97
1.66 0.98
1.67 0.99
1.68 0.99
1.69 0.99
1.7 0.98
1.71 0.98
1.72 1
1.73 0.99
1.74 0.97
1.75 0.97
1.76 0.98
1.77 0.99
1.78 0.97
1.79 0.97
1.8 0.98
1.81 0.99
1.82 0.99
1.83 0.98
1.84 0.97
1.85 0.99
1.86 0.98
1.87 0.97
1.88 0.99
1.89 0.97
1.9 0.96
1.91 0.98
1.92 1
1.93 0.92
1.94 0.98
1.95 0.97
1.96 0.97
1.97 0.99
1.98 0.98
1.99 1
2 0.98
};
\addplot [semithick, black, opacity=1, dash pattern=on 1pt off 1pt]
table {%
-0.0895000000000001 0.5
2.0995 0.5
};
\addplot [semithick, black, opacity=1, dash pattern=on 1pt off 1pt]
table {%
-0.0895000000000001 1
2.0995 1
};
\addplot [semithick, black, opacity=1, dash pattern=on 1pt off 1pt]
table {%
1 0.2
1 1.05
};
\end{axis}

\end{tikzpicture}

%% file: plots/decoupled2/NL_UNIxLAP.tex
% This file was created by tikzplotlib v0.9.6.
\begin{tikzpicture}

\definecolor{color0}{rgb}{0.866666666666667,0.494117647058824,0.164705882352941}
\definecolor{color1}{rgb}{0.164705882352941,0.643137254901961,0.866666666666667}
\definecolor{color2}{rgb}{0.584313725490196,0.866666666666667,0.164705882352941}
\definecolor{color3}{rgb}{0.109803921568627,0.337254901960784,0.129411764705882}
\definecolor{color4}{rgb}{0.529411764705882,0.305882352941176,0.858823529411765}
\definecolor{color5}{rgb}{0.858823529411765,0.305882352941176,0.435294117647059}
\definecolor{color6}{rgb}{0.937254901960784,0.929411764705882,0.392156862745098}
\definecolor{color7}{rgb}{0.0901960784313725,0.486274509803922,0.0980392156862745}
\definecolor{color8}{rgb}{0.156862745098039,0.188235294117647,0.827450980392157}
\definecolor{color9}{rgb}{0.937254901960784,0.392156862745098,0.894117647058824}
\definecolor{color10}{rgb}{0.2,0.184313725490196,0.184313725490196}
\definecolor{color11}{rgb}{0.0156862745098039,0.803921568627451,0.976470588235294}

\begin{axis}[
tick align=outside,
tick pos=left,
x grid style={white!69.0196078431373!black},
xmajorgrids,
xmin=-0.0895, xmax=2.0995,
xtick style={color=black},
xtick={0,0.1,0.2,0.3,0.4,0.5,0.6,0.7,0.8,0.9,1,1.1,1.2,1.3,1.4,1.5,1.6,1.7,1.8,1.9,2},
xticklabels={0,,.2,,.4,,.6,,.8,,1,,20,,40,,60,,80,,100},
height=4.8cm,
width=6.5cm,
y grid style={white!69.0196078431373!black},
ymajorgrids,
ymin=0.2, ymax=1.05,
ytick style={color=black}
]
\addplot [semithick, color0, opacity=0.75]
table {%
0.01 0.54
0.02 0.78
0.03 0.87
0.04 0.85
0.05 0.96
0.06 0.99
0.07 0.96
0.08 0.94
0.09 0.97
0.1 1
0.11 1
0.12 1
0.13 1
0.14 1
0.15 1
0.16 1
0.17 1
0.18 1
0.19 1
0.2 1
0.21 1
0.22 1
0.23 1
0.24 1
0.25 1
0.26 1
0.27 1
0.28 1
0.29 1
0.3 1
0.31 1
0.32 1
0.33 1
0.34 1
0.35 1
0.36 1
0.37 1
0.38 1
0.39 1
0.4 1
0.41 1
0.42 1
0.43 1
0.44 1
0.45 1
0.46 1
0.47 1
0.48 1
0.49 1
0.5 1
0.51 1
0.52 1
0.53 1
0.54 1
0.55 1
0.56 1
0.57 1
0.58 1
0.59 1
0.6 1
0.61 1
0.62 1
0.63 1
0.64 1
0.65 1
0.66 1
0.67 1
0.68 1
0.69 1
0.7 1
0.71 1
0.72 1
0.73 1
0.74 1
0.75 1
0.76 1
0.77 1
0.78 1
0.79 1
0.8 1
0.81 1
0.82 1
0.83 1
0.84 1
0.85 1
0.86 1
0.87 1
0.88 1
0.89 1
0.9 1
0.91 1
0.92 1
0.93 1
0.94 1
0.95 1
0.96 1
0.97 1
0.98 1
0.99 1
1 1
1.02 0.99
1.03 0.9
1.04 0.85
1.05 0.84
1.06 0.86
1.07 0.71
1.08 0.7
1.09 0.7
1.1 0.62
1.11 0.69
1.12 0.69
1.13 0.72
1.14 0.66
1.15 0.57
1.16 0.59
1.17 0.6
1.18 0.57
1.19 0.54
1.2 0.67
1.21 0.59
1.22 0.55
1.23 0.56
1.24 0.6
1.25 0.57
1.26 0.53
1.27 0.54
1.28 0.72
1.29 0.54
1.3 0.56
1.31 0.6
1.32 0.64
1.33 0.59
1.34 0.55
1.35 0.53
1.36 0.52
1.37 0.54
1.38 0.49
1.39 0.57
1.4 0.54
1.41 0.59
1.42 0.59
1.43 0.57
1.44 0.51
1.45 0.57
1.46 0.53
1.47 0.64
1.48 0.6
1.49 0.62
1.5 0.46
1.51 0.64
1.52 0.61
1.53 0.58
1.54 0.53
1.55 0.54
1.56 0.54
1.57 0.52
1.58 0.64
1.59 0.61
1.6 0.7
1.61 0.6
1.62 0.53
1.63 0.52
1.64 0.63
1.65 0.6
1.66 0.59
1.67 0.49
1.68 0.54
1.69 0.51
1.7 0.64
1.71 0.57
1.72 0.56
1.73 0.54
1.74 0.62
1.75 0.58
1.76 0.57
1.77 0.6
1.78 0.6
1.79 0.59
1.8 0.52
1.81 0.63
1.82 0.56
1.83 0.55
1.84 0.68
1.85 0.64
1.86 0.57
1.87 0.55
1.88 0.52
1.89 0.43
1.9 0.6
1.91 0.55
1.92 0.52
1.93 0.58
1.94 0.54
1.95 0.62
1.96 0.52
1.97 0.62
1.98 0.57
1.99 0.57
2 0.57
};
\addplot [semithick, color1, opacity=0.75]
table {%
0.01 0.51
0.02 0.51
0.03 0.44
0.04 0.57
0.05 0.61
0.06 0.49
0.07 0.64
0.08 0.59
0.09 0.52
0.1 0.58
0.11 0.62
0.12 0.56
0.13 0.63
0.14 0.57
0.15 0.64
0.16 0.75
0.17 0.69
0.18 0.64
0.19 0.7
0.2 0.65
0.21 0.71
0.22 0.7
0.23 0.68
0.24 0.67
0.25 0.7
0.26 0.72
0.27 0.71
0.28 0.7
0.29 0.73
0.3 0.78
0.31 0.76
0.32 0.76
0.33 0.75
0.34 0.72
0.35 0.76
0.36 0.76
0.37 0.82
0.38 0.78
0.39 0.81
0.4 0.8
0.41 0.88
0.42 0.79
0.43 0.75
0.44 0.83
0.45 0.85
0.46 0.85
0.47 0.84
0.48 0.86
0.49 0.83
0.5 0.83
0.51 0.87
0.52 0.84
0.53 0.88
0.54 0.81
0.55 0.9
0.56 0.86
0.57 0.86
0.58 0.84
0.59 0.82
0.6 0.81
0.61 0.82
0.62 0.9
0.63 0.88
0.64 0.86
0.65 0.89
0.66 0.86
0.67 0.87
0.68 0.86
0.69 0.78
0.7 0.87
0.71 0.92
0.72 0.83
0.73 0.82
0.74 0.82
0.75 0.85
0.76 0.87
0.77 0.91
0.78 0.88
0.79 0.91
0.8 0.85
0.81 0.91
0.82 0.88
0.83 0.88
0.84 0.87
0.85 0.9
0.86 0.84
0.87 0.83
0.88 0.85
0.89 0.91
0.9 0.87
0.91 0.86
0.92 0.84
0.93 0.83
0.94 0.87
0.95 0.87
0.96 0.84
0.97 0.84
0.98 0.85
0.99 0.89
1 0.92
1.02 0.68
1.03 0.64
1.04 0.58
1.05 0.63
1.06 0.67
1.07 0.59
1.08 0.48
1.09 0.46
1.1 0.55
1.11 0.53
1.12 0.38
1.13 0.53
1.14 0.48
1.15 0.43
1.16 0.46
1.17 0.44
1.18 0.5
1.19 0.53
1.2 0.48
1.21 0.44
1.22 0.42
1.23 0.53
1.24 0.49
1.25 0.52
1.26 0.56
1.27 0.5
1.28 0.55
1.29 0.42
1.3 0.52
1.31 0.49
1.32 0.49
1.33 0.41
1.34 0.58
1.35 0.49
1.36 0.48
1.37 0.54
1.38 0.5
1.39 0.45
1.4 0.53
1.41 0.59
1.42 0.56
1.43 0.56
1.44 0.5
1.45 0.49
1.46 0.5
1.47 0.4
1.48 0.44
1.49 0.45
1.5 0.47
1.51 0.48
1.52 0.55
1.53 0.45
1.54 0.48
1.55 0.54
1.56 0.47
1.57 0.56
1.58 0.55
1.59 0.47
1.6 0.52
1.61 0.38
1.62 0.43
1.63 0.5
1.64 0.47
1.65 0.45
1.66 0.46
1.67 0.53
1.68 0.5
1.69 0.44
1.7 0.49
1.71 0.39
1.72 0.55
1.73 0.54
1.74 0.57
1.75 0.48
1.76 0.58
1.77 0.39
1.78 0.56
1.79 0.41
1.8 0.48
1.81 0.45
1.82 0.51
1.83 0.44
1.84 0.53
1.85 0.46
1.86 0.43
1.87 0.58
1.88 0.43
1.89 0.43
1.9 0.44
1.91 0.55
1.92 0.49
1.93 0.48
1.94 0.54
1.95 0.41
1.96 0.57
1.97 0.54
1.98 0.46
1.99 0.5
2 0.45
};
\addplot [semithick, color2, opacity=0.75]
table {%
0.01 0.47
0.02 0.55
0.03 0.51
0.04 0.43
0.05 0.53
0.06 0.58
0.07 0.52
0.08 0.62
0.09 0.5
0.1 0.52
0.11 0.55
0.12 0.55
0.13 0.64
0.14 0.54
0.15 0.64
0.16 0.72
0.17 0.63
0.18 0.59
0.19 0.65
0.2 0.6
0.21 0.66
0.22 0.65
0.23 0.65
0.24 0.64
0.25 0.67
0.26 0.69
0.27 0.68
0.28 0.68
0.29 0.73
0.3 0.75
0.31 0.77
0.32 0.74
0.33 0.76
0.34 0.7
0.35 0.74
0.36 0.73
0.37 0.83
0.38 0.78
0.39 0.82
0.4 0.78
0.41 0.87
0.42 0.77
0.43 0.75
0.44 0.86
0.45 0.85
0.46 0.86
0.47 0.84
0.48 0.85
0.49 0.84
0.5 0.83
0.51 0.85
0.52 0.84
0.53 0.88
0.54 0.82
0.55 0.9
0.56 0.83
0.57 0.86
0.58 0.84
0.59 0.82
0.6 0.82
0.61 0.83
0.62 0.89
0.63 0.87
0.64 0.86
0.65 0.88
0.66 0.84
0.67 0.88
0.68 0.85
0.69 0.77
0.7 0.87
0.71 0.92
0.72 0.83
0.73 0.84
0.74 0.82
0.75 0.84
0.76 0.86
0.77 0.91
0.78 0.88
0.79 0.91
0.8 0.85
0.81 0.91
0.82 0.88
0.83 0.89
0.84 0.84
0.85 0.89
0.86 0.84
0.87 0.82
0.88 0.85
0.89 0.89
0.9 0.87
0.91 0.87
0.92 0.85
0.93 0.84
0.94 0.87
0.95 0.87
0.96 0.85
0.97 0.84
0.98 0.85
0.99 0.89
1 0.91
1.02 0.68
1.03 0.63
1.04 0.57
1.05 0.6
1.06 0.67
1.07 0.58
1.08 0.47
1.09 0.43
1.1 0.56
1.11 0.52
1.12 0.39
1.13 0.53
1.14 0.48
1.15 0.43
1.16 0.45
1.17 0.45
1.18 0.5
1.19 0.54
1.2 0.47
1.21 0.47
1.22 0.44
1.23 0.52
1.24 0.51
1.25 0.5
1.26 0.57
1.27 0.49
1.28 0.54
1.29 0.43
1.3 0.5
1.31 0.48
1.32 0.47
1.33 0.41
1.34 0.57
1.35 0.49
1.36 0.47
1.37 0.55
1.38 0.52
1.39 0.44
1.4 0.52
1.41 0.58
1.42 0.53
1.43 0.57
1.44 0.5
1.45 0.49
1.46 0.5
1.47 0.4
1.48 0.46
1.49 0.44
1.5 0.48
1.51 0.49
1.52 0.57
1.53 0.45
1.54 0.47
1.55 0.56
1.56 0.49
1.57 0.55
1.58 0.53
1.59 0.49
1.6 0.5
1.61 0.4
1.62 0.44
1.63 0.5
1.64 0.47
1.65 0.45
1.66 0.46
1.67 0.54
1.68 0.51
1.69 0.44
1.7 0.5
1.71 0.38
1.72 0.56
1.73 0.53
1.74 0.58
1.75 0.48
1.76 0.59
1.77 0.39
1.78 0.54
1.79 0.4
1.8 0.48
1.81 0.45
1.82 0.52
1.83 0.46
1.84 0.56
1.85 0.45
1.86 0.41
1.87 0.58
1.88 0.43
1.89 0.43
1.9 0.45
1.91 0.55
1.92 0.49
1.93 0.5
1.94 0.54
1.95 0.41
1.96 0.57
1.97 0.53
1.98 0.46
1.99 0.51
2 0.43
};
\addplot [semithick, color3, opacity=0.75]
table {%
0.01 0.73
0.02 0.89
0.03 0.95
0.04 0.95
0.05 0.98
0.06 0.99
0.07 0.98
0.08 0.94
0.09 0.96
0.1 0.96
0.11 0.99
0.12 1
0.13 0.98
0.14 0.99
0.15 1
0.16 1
0.17 0.99
0.18 1
0.19 1
0.2 1
0.21 0.98
0.22 0.99
0.23 0.99
0.24 1
0.25 1
0.26 1
0.27 1
0.28 0.99
0.29 0.97
0.3 1
0.31 0.99
0.32 1
0.33 0.99
0.34 0.99
0.35 1
0.36 0.99
0.37 1
0.38 1
0.39 1
0.4 0.99
0.41 1
0.42 0.99
0.43 0.99
0.44 1
0.45 1
0.46 0.99
0.47 1
0.48 1
0.49 1
0.5 0.99
0.51 1
0.52 0.99
0.53 0.99
0.54 1
0.55 1
0.56 1
0.57 1
0.58 0.99
0.59 0.99
0.6 0.99
0.61 0.99
0.62 1
0.63 1
0.64 1
0.65 0.99
0.66 0.98
0.67 1
0.68 1
0.69 1
0.7 0.99
0.71 0.99
0.72 0.99
0.73 1
0.74 1
0.75 1
0.76 0.99
0.77 1
0.78 0.99
0.79 1
0.8 0.99
0.81 1
0.82 1
0.83 0.99
0.84 1
0.85 1
0.86 0.99
0.87 0.99
0.88 0.99
0.89 1
0.9 0.99
0.91 0.97
0.92 0.98
0.93 1
0.94 1
0.95 1
0.96 0.99
0.97 0.98
0.98 1
0.99 1
1 1
1.02 1
1.03 0.98
1.04 0.97
1.05 0.91
1.06 0.89
1.07 0.88
1.08 0.87
1.09 0.86
1.1 0.87
1.11 0.83
1.12 0.81
1.13 0.84
1.14 0.76
1.15 0.81
1.16 0.74
1.17 0.74
1.18 0.73
1.19 0.75
1.2 0.77
1.21 0.72
1.22 0.74
1.23 0.78
1.24 0.76
1.25 0.71
1.26 0.69
1.27 0.71
1.28 0.8
1.29 0.71
1.3 0.72
1.31 0.81
1.32 0.81
1.33 0.78
1.34 0.71
1.35 0.77
1.36 0.68
1.37 0.75
1.38 0.75
1.39 0.7
1.4 0.73
1.41 0.77
1.42 0.79
1.43 0.79
1.44 0.7
1.45 0.78
1.46 0.71
1.47 0.79
1.48 0.74
1.49 0.81
1.5 0.68
1.51 0.74
1.52 0.71
1.53 0.79
1.54 0.73
1.55 0.75
1.56 0.69
1.57 0.8
1.58 0.73
1.59 0.68
1.6 0.76
1.61 0.81
1.62 0.69
1.63 0.73
1.64 0.77
1.65 0.74
1.66 0.73
1.67 0.71
1.68 0.76
1.69 0.64
1.7 0.77
1.71 0.78
1.72 0.74
1.73 0.72
1.74 0.76
1.75 0.74
1.76 0.75
1.77 0.72
1.78 0.81
1.79 0.69
1.8 0.76
1.81 0.77
1.82 0.68
1.83 0.67
1.84 0.87
1.85 0.79
1.86 0.82
1.87 0.79
1.88 0.71
1.89 0.72
1.9 0.76
1.91 0.65
1.92 0.67
1.93 0.75
1.94 0.72
1.95 0.76
1.96 0.74
1.97 0.75
1.98 0.75
1.99 0.77
2 0.75
};
\addplot [semithick, color4, opacity=0.75]
table {%
0.01 0.75
0.02 0.89
0.03 0.95
0.04 0.95
0.05 0.98
0.06 0.99
0.07 0.98
0.08 0.94
0.09 0.96
0.1 0.97
0.11 0.99
0.12 1
0.13 0.98
0.14 0.99
0.15 0.99
0.16 1
0.17 0.99
0.18 1
0.19 1
0.2 1
0.21 0.98
0.22 0.99
0.23 0.99
0.24 1
0.25 1
0.26 1
0.27 1
0.28 0.99
0.29 0.97
0.3 1
0.31 0.99
0.32 1
0.33 0.99
0.34 0.97
0.35 1
0.36 0.99
0.37 1
0.38 1
0.39 1
0.4 0.99
0.41 1
0.42 0.99
0.43 0.99
0.44 1
0.45 1
0.46 0.99
0.47 1
0.48 1
0.49 1
0.5 0.99
0.51 1
0.52 0.99
0.53 0.99
0.54 1
0.55 1
0.56 1
0.57 1
0.58 0.97
0.59 0.99
0.6 0.98
0.61 0.99
0.62 1
0.63 1
0.64 1
0.65 0.99
0.66 0.98
0.67 1
0.68 1
0.69 0.99
0.7 0.99
0.71 0.99
0.72 0.99
0.73 1
0.74 0.99
0.75 1
0.76 0.99
0.77 1
0.78 0.99
0.79 1
0.8 0.99
0.81 1
0.82 1
0.83 0.99
0.84 1
0.85 1
0.86 0.99
0.87 0.99
0.88 0.99
0.89 1
0.9 0.99
0.91 0.97
0.92 0.98
0.93 1
0.94 1
0.95 1
0.96 0.99
0.97 0.98
0.98 1
0.99 1
1 1
1.02 1
1.03 0.98
1.04 0.97
1.05 0.91
1.06 0.9
1.07 0.89
1.08 0.87
1.09 0.87
1.1 0.87
1.11 0.83
1.12 0.81
1.13 0.84
1.14 0.76
1.15 0.83
1.16 0.75
1.17 0.76
1.18 0.72
1.19 0.75
1.2 0.76
1.21 0.73
1.22 0.77
1.23 0.77
1.24 0.76
1.25 0.72
1.26 0.69
1.27 0.71
1.28 0.79
1.29 0.72
1.3 0.72
1.31 0.81
1.32 0.82
1.33 0.78
1.34 0.71
1.35 0.78
1.36 0.68
1.37 0.74
1.38 0.77
1.39 0.7
1.4 0.74
1.41 0.76
1.42 0.78
1.43 0.81
1.44 0.7
1.45 0.78
1.46 0.74
1.47 0.78
1.48 0.74
1.49 0.82
1.5 0.7
1.51 0.75
1.52 0.72
1.53 0.78
1.54 0.72
1.55 0.76
1.56 0.69
1.57 0.8
1.58 0.74
1.59 0.67
1.6 0.77
1.61 0.81
1.62 0.69
1.63 0.74
1.64 0.77
1.65 0.74
1.66 0.73
1.67 0.71
1.68 0.76
1.69 0.64
1.7 0.77
1.71 0.78
1.72 0.74
1.73 0.73
1.74 0.76
1.75 0.74
1.76 0.75
1.77 0.72
1.78 0.82
1.79 0.7
1.8 0.77
1.81 0.79
1.82 0.69
1.83 0.68
1.84 0.87
1.85 0.81
1.86 0.8
1.87 0.8
1.88 0.7
1.89 0.73
1.9 0.75
1.91 0.66
1.92 0.67
1.93 0.75
1.94 0.71
1.95 0.75
1.96 0.74
1.97 0.76
1.98 0.75
1.99 0.77
2 0.75
};
\addplot [semithick, color5, opacity=0.75]
table {%
0.01 0.88
0.02 0.86
0.03 0.94
0.04 0.91
0.05 0.95
0.06 0.92
0.07 0.99
0.08 0.97
0.09 0.94
0.1 0.97
0.11 0.98
0.12 0.97
0.13 0.97
0.14 0.99
0.15 0.99
0.16 1
0.17 1
0.18 0.94
0.19 0.97
0.2 0.98
0.21 0.98
0.22 0.99
0.23 0.96
0.24 0.97
0.25 0.96
0.26 0.96
0.27 0.99
0.28 0.96
0.29 0.93
0.3 0.96
0.31 0.97
0.32 0.98
0.33 0.94
0.34 0.93
0.35 1
0.36 0.96
0.37 0.99
0.38 0.95
0.39 0.97
0.4 0.98
0.41 0.99
0.42 0.96
0.43 0.95
0.44 0.96
0.45 0.99
0.46 0.97
0.47 0.97
0.48 0.98
0.49 0.97
0.5 0.98
0.51 0.96
0.52 0.95
0.53 0.94
0.54 0.95
0.55 0.98
0.56 0.99
0.57 0.98
0.58 0.96
0.59 0.93
0.6 0.97
0.61 0.94
0.62 0.97
0.63 0.96
0.64 0.99
0.65 0.99
0.66 0.97
0.67 0.97
0.68 0.98
0.69 0.99
0.7 0.98
0.71 0.96
0.72 0.96
0.73 0.98
0.74 0.94
0.75 0.98
0.76 0.99
0.77 0.98
0.78 0.96
0.79 0.98
0.8 0.96
0.81 0.98
0.82 0.97
0.83 0.95
0.84 0.99
0.85 0.99
0.86 0.98
0.87 0.97
0.88 0.96
0.89 0.95
0.9 0.98
0.91 0.93
0.92 0.96
0.93 0.98
0.94 0.98
0.95 0.99
0.96 0.98
0.97 0.97
0.98 0.99
0.99 0.98
1 0.99
1.02 0.94
1.03 0.92
1.04 0.92
1.05 0.93
1.06 0.89
1.07 0.87
1.08 0.91
1.09 0.87
1.1 0.87
1.11 0.84
1.12 0.85
1.13 0.91
1.14 0.9
1.15 0.91
1.16 0.88
1.17 0.84
1.18 0.88
1.19 0.88
1.2 0.92
1.21 0.81
1.22 0.91
1.23 0.87
1.24 0.89
1.25 0.82
1.26 0.84
1.27 0.86
1.28 0.94
1.29 0.87
1.3 0.87
1.31 0.9
1.32 0.9
1.33 0.9
1.34 0.86
1.35 0.88
1.36 0.88
1.37 0.89
1.38 0.84
1.39 0.87
1.4 0.9
1.41 0.87
1.42 0.89
1.43 0.83
1.44 0.82
1.45 0.87
1.46 0.87
1.47 0.93
1.48 0.85
1.49 0.87
1.5 0.87
1.51 0.83
1.52 0.83
1.53 0.82
1.54 0.84
1.55 0.91
1.56 0.85
1.57 0.89
1.58 0.86
1.59 0.89
1.6 0.86
1.61 0.9
1.62 0.9
1.63 0.89
1.64 0.89
1.65 0.91
1.66 0.84
1.67 0.92
1.68 0.85
1.69 0.92
1.7 0.93
1.71 0.86
1.72 0.89
1.73 0.91
1.74 0.84
1.75 0.91
1.76 0.83
1.77 0.85
1.78 0.83
1.79 0.86
1.8 0.88
1.81 0.84
1.82 0.88
1.83 0.87
1.84 0.91
1.85 0.89
1.86 0.91
1.87 0.83
1.88 0.86
1.89 0.87
1.9 0.9
1.91 0.84
1.92 0.86
1.93 0.91
1.94 0.87
1.95 0.89
1.96 0.86
1.97 0.88
1.98 0.89
1.99 0.9
2 0.9
};
\addplot [semithick, color6, opacity=0.75, dashed]
table {%
0.01 0.99
0.02 0.99
0.03 1
0.04 1
0.05 1
0.06 1
0.07 1
0.08 1
0.09 0.99
0.1 1
0.11 1
0.12 1
0.13 1
0.14 1
0.15 1
0.16 1
0.17 1
0.18 1
0.19 1
0.2 1
0.21 1
0.22 1
0.23 1
0.24 1
0.25 1
0.26 1
0.27 1
0.28 1
0.29 1
0.3 1
0.31 1
0.32 1
0.33 1
0.34 1
0.35 1
0.36 1
0.37 1
0.38 1
0.39 1
0.4 1
0.41 1
0.42 1
0.43 1
0.44 1
0.45 1
0.46 1
0.47 1
0.48 1
0.49 1
0.5 1
0.51 1
0.52 1
0.53 1
0.54 1
0.55 1
0.56 1
0.57 1
0.58 1
0.59 1
0.6 1
0.61 1
0.62 1
0.63 1
0.64 1
0.65 1
0.66 1
0.67 1
0.68 1
0.69 1
0.7 1
0.71 1
0.72 1
0.73 1
0.74 1
0.75 1
0.76 1
0.77 1
0.78 1
0.79 1
0.8 1
0.81 1
0.82 1
0.83 1
0.84 1
0.85 1
0.86 1
0.87 1
0.88 1
0.89 1
0.9 1
0.91 1
0.92 1
0.93 1
0.94 1
0.95 1
0.96 1
0.97 1
0.98 1
0.99 1
1 1
1.02 1
1.03 1
1.04 0.98
1.05 1
1.06 0.99
1.07 1
1.08 0.99
1.09 1
1.1 0.98
1.11 0.99
1.12 0.97
1.13 0.99
1.14 0.95
1.15 0.98
1.16 0.97
1.17 0.99
1.18 0.98
1.19 0.97
1.2 0.99
1.21 0.98
1.22 1
1.23 0.98
1.24 0.99
1.25 0.98
1.26 0.98
1.27 0.96
1.28 0.99
1.29 0.99
1.3 0.97
1.31 0.95
1.32 0.98
1.33 1
1.34 1
1.35 0.98
1.36 0.98
1.37 0.97
1.38 0.99
1.39 0.97
1.4 0.99
1.41 0.99
1.42 0.94
1.43 0.99
1.44 0.99
1.45 0.98
1.46 0.99
1.47 1
1.48 0.99
1.49 0.97
1.5 0.98
1.51 0.97
1.52 0.98
1.53 0.97
1.54 0.96
1.55 0.98
1.56 0.99
1.57 0.98
1.58 0.98
1.59 1
1.6 0.97
1.61 0.99
1.62 1
1.63 0.96
1.64 0.98
1.65 0.95
1.66 1
1.67 0.99
1.68 0.96
1.69 1
1.7 0.99
1.71 0.97
1.72 0.98
1.73 0.98
1.74 0.97
1.75 0.99
1.76 0.99
1.77 0.97
1.78 0.99
1.79 1
1.8 0.98
1.81 0.99
1.82 0.95
1.83 0.97
1.84 0.98
1.85 0.97
1.86 1
1.87 0.97
1.88 0.98
1.89 1
1.9 0.99
1.91 0.99
1.92 0.99
1.93 0.98
1.94 0.95
1.95 0.98
1.96 0.97
1.97 0.97
1.98 0.98
1.99 0.98
2 0.99
};
\addplot [semithick, color7, opacity=0.75, dashed]
table {%
0.01 0.99
0.02 0.99
0.03 1
0.04 1
0.05 1
0.06 1
0.07 1
0.08 1
0.09 0.99
0.1 1
0.11 1
0.12 1
0.13 1
0.14 1
0.15 1
0.16 1
0.17 1
0.18 1
0.19 1
0.2 1
0.21 1
0.22 1
0.23 1
0.24 1
0.25 1
0.26 1
0.27 1
0.28 1
0.29 1
0.3 1
0.31 1
0.32 1
0.33 1
0.34 1
0.35 1
0.36 1
0.37 1
0.38 1
0.39 1
0.4 1
0.41 1
0.42 1
0.43 1
0.44 1
0.45 1
0.46 1
0.47 1
0.48 1
0.49 1
0.5 1
0.51 1
0.52 1
0.53 1
0.54 1
0.55 1
0.56 1
0.57 1
0.58 1
0.59 1
0.6 1
0.61 1
0.62 1
0.63 1
0.64 1
0.65 1
0.66 1
0.67 1
0.68 1
0.69 1
0.7 1
0.71 1
0.72 1
0.73 1
0.74 1
0.75 1
0.76 1
0.77 1
0.78 1
0.79 1
0.8 1
0.81 1
0.82 1
0.83 1
0.84 1
0.85 1
0.86 1
0.87 1
0.88 1
0.89 1
0.9 1
0.91 1
0.92 1
0.93 1
0.94 1
0.95 1
0.96 1
0.97 1
0.98 1
0.99 1
1 1
1.02 1
1.03 1
1.04 0.98
1.05 1
1.06 0.99
1.07 1
1.08 0.99
1.09 1
1.1 0.98
1.11 0.99
1.12 0.97
1.13 0.99
1.14 0.95
1.15 0.98
1.16 0.97
1.17 0.99
1.18 0.98
1.19 0.97
1.2 0.99
1.21 0.98
1.22 1
1.23 0.98
1.24 0.99
1.25 0.98
1.26 0.98
1.27 0.96
1.28 0.99
1.29 0.99
1.3 0.97
1.31 0.95
1.32 0.98
1.33 1
1.34 1
1.35 0.98
1.36 0.98
1.37 0.97
1.38 0.99
1.39 0.97
1.4 0.99
1.41 0.99
1.42 0.94
1.43 0.99
1.44 0.99
1.45 0.98
1.46 0.99
1.47 1
1.48 0.99
1.49 0.97
1.5 0.98
1.51 0.97
1.52 0.98
1.53 0.97
1.54 0.96
1.55 0.98
1.56 0.99
1.57 0.98
1.58 0.98
1.59 1
1.6 0.97
1.61 0.99
1.62 1
1.63 0.96
1.64 0.98
1.65 0.95
1.66 1
1.67 0.99
1.68 0.96
1.69 1
1.7 0.99
1.71 0.97
1.72 0.98
1.73 0.98
1.74 0.97
1.75 0.99
1.76 0.99
1.77 0.97
1.78 0.99
1.79 1
1.8 0.98
1.81 0.99
1.82 0.95
1.83 0.97
1.84 0.98
1.85 0.97
1.86 1
1.87 0.97
1.88 0.98
1.89 1
1.9 0.99
1.91 0.99
1.92 0.99
1.93 0.98
1.94 0.95
1.95 0.98
1.96 0.97
1.97 0.97
1.98 0.98
1.99 0.98
2 0.99
};
\addplot [semithick, color8, opacity=0.75, dashed]
table {%
0.01 1
0.02 1
0.03 1
0.04 1
0.05 1
0.06 1
0.07 1
0.08 1
0.09 1
0.1 1
0.11 1
0.12 1
0.13 1
0.14 1
0.15 1
0.16 1
0.17 1
0.18 1
0.19 1
0.2 1
0.21 1
0.22 1
0.23 1
0.24 1
0.25 1
0.26 1
0.27 1
0.28 1
0.29 1
0.3 1
0.31 1
0.32 1
0.33 1
0.34 1
0.35 1
0.36 1
0.37 1
0.38 1
0.39 1
0.4 1
0.41 1
0.42 1
0.43 1
0.44 1
0.45 1
0.46 1
0.47 1
0.48 1
0.49 1
0.5 1
0.51 1
0.52 1
0.53 1
0.54 1
0.55 1
0.56 1
0.57 1
0.58 1
0.59 1
0.6 1
0.61 1
0.62 1
0.63 1
0.64 1
0.65 1
0.66 1
0.67 1
0.68 1
0.69 1
0.7 1
0.71 1
0.72 1
0.73 1
0.74 1
0.75 1
0.76 1
0.77 1
0.78 1
0.79 1
0.8 1
0.81 1
0.82 1
0.83 1
0.84 1
0.85 1
0.86 1
0.87 1
0.88 1
0.89 1
0.9 1
0.91 1
0.92 1
0.93 1
0.94 1
0.95 1
0.96 1
0.97 1
0.98 1
0.99 1
1 1
1.02 1
1.03 1
1.04 1
1.05 1
1.06 1
1.07 0.99
1.08 1
1.09 0.99
1.1 0.99
1.11 1
1.12 1
1.13 0.99
1.14 0.99
1.15 1
1.16 1
1.17 1
1.18 0.98
1.19 0.98
1.2 1
1.21 0.99
1.22 1
1.23 1
1.24 0.97
1.25 0.98
1.26 0.97
1.27 0.99
1.28 1
1.29 0.99
1.3 0.97
1.31 1
1.32 0.99
1.33 1
1.34 1
1.35 0.99
1.36 0.99
1.37 1
1.38 0.98
1.39 1
1.4 0.98
1.41 1
1.42 0.99
1.43 0.99
1.44 0.99
1.45 1
1.46 0.99
1.47 0.99
1.48 1
1.49 1
1.5 0.98
1.51 0.97
1.52 0.99
1.53 0.98
1.54 1
1.55 0.98
1.56 0.98
1.57 0.99
1.58 0.98
1.59 1
1.6 1
1.61 1
1.62 1
1.63 0.96
1.64 1
1.65 1
1.66 1
1.67 0.99
1.68 0.98
1.69 1
1.7 1
1.71 0.99
1.72 1
1.73 1
1.74 0.99
1.75 0.97
1.76 1
1.77 0.99
1.78 0.98
1.79 1
1.8 0.98
1.81 0.98
1.82 0.98
1.83 0.96
1.84 1
1.85 0.98
1.86 1
1.87 1
1.88 1
1.89 0.99
1.9 0.99
1.91 0.99
1.92 0.99
1.93 1
1.94 0.99
1.95 1
1.96 0.99
1.97 0.99
1.98 1
1.99 1
2 1
};
\addplot [semithick, color9, opacity=0.75, dashed]
table {%
0.01 0.65
0.02 0.81
0.03 0.91
0.04 0.96
0.05 0.99
0.06 1
0.07 1
0.08 0.99
0.09 1
0.1 0.99
0.11 1
0.12 1
0.13 1
0.14 1
0.15 1
0.16 1
0.17 1
0.18 1
0.19 1
0.2 1
0.21 1
0.22 1
0.23 1
0.24 1
0.25 1
0.26 1
0.27 1
0.28 1
0.29 1
0.3 1
0.31 1
0.32 1
0.33 1
0.34 1
0.35 1
0.36 1
0.37 1
0.38 1
0.39 1
0.4 1
0.41 1
0.42 1
0.43 1
0.44 1
0.45 1
0.46 1
0.47 1
0.48 1
0.49 1
0.5 1
0.51 1
0.52 1
0.53 1
0.54 1
0.55 1
0.56 1
0.57 1
0.58 1
0.59 1
0.6 1
0.61 1
0.62 1
0.63 1
0.64 1
0.65 1
0.66 1
0.67 1
0.68 1
0.69 1
0.7 1
0.71 1
0.72 1
0.73 1
0.74 1
0.75 1
0.76 1
0.77 1
0.78 1
0.79 1
0.8 1
0.81 1
0.82 1
0.83 1
0.84 1
0.85 1
0.86 1
0.87 1
0.88 1
0.89 1
0.9 1
0.91 1
0.92 1
0.93 1
0.94 1
0.95 1
0.96 1
0.97 1
0.98 1
0.99 1
1 1
1.02 1
1.03 0.97
1.04 0.94
1.05 0.88
1.06 0.89
1.07 0.81
1.08 0.84
1.09 0.76
1.1 0.8
1.11 0.78
1.12 0.72
1.13 0.73
1.14 0.73
1.15 0.71
1.16 0.68
1.17 0.76
1.18 0.68
1.19 0.68
1.2 0.73
1.21 0.74
1.22 0.69
1.23 0.64
1.24 0.67
1.25 0.69
1.26 0.56
1.27 0.66
1.28 0.67
1.29 0.68
1.3 0.6
1.31 0.72
1.32 0.72
1.33 0.74
1.34 0.6
1.35 0.7
1.36 0.59
1.37 0.64
1.38 0.73
1.39 0.64
1.4 0.67
1.41 0.73
1.42 0.73
1.43 0.76
1.44 0.73
1.45 0.68
1.46 0.61
1.47 0.76
1.48 0.72
1.49 0.72
1.5 0.61
1.51 0.64
1.52 0.56
1.53 0.68
1.54 0.64
1.55 0.68
1.56 0.68
1.57 0.7
1.58 0.66
1.59 0.7
1.6 0.65
1.61 0.66
1.62 0.64
1.63 0.69
1.64 0.67
1.65 0.66
1.66 0.69
1.67 0.66
1.68 0.72
1.69 0.63
1.7 0.64
1.71 0.72
1.72 0.64
1.73 0.64
1.74 0.72
1.75 0.6
1.76 0.71
1.77 0.64
1.78 0.69
1.79 0.62
1.8 0.63
1.81 0.66
1.82 0.65
1.83 0.65
1.84 0.75
1.85 0.69
1.86 0.72
1.87 0.73
1.88 0.6
1.89 0.64
1.9 0.67
1.91 0.66
1.92 0.71
1.93 0.64
1.94 0.61
1.95 0.67
1.96 0.65
1.97 0.66
1.98 0.61
1.99 0.7
2 0.66
};
\addplot [semithick, color10, opacity=0.75, dashed]
table {%
0.01 0.45
0.02 0.69
0.03 0.71
0.04 0.84
0.05 0.85
0.06 0.89
0.07 0.91
0.08 0.95
0.09 0.94
0.1 0.97
0.11 0.95
0.12 0.98
0.13 1
0.14 0.98
0.15 0.98
0.16 0.96
0.17 0.99
0.18 0.99
0.19 1
0.2 1
0.21 1
0.22 1
0.23 1
0.24 1
0.25 1
0.26 1
0.27 1
0.28 1
0.29 0.99
0.3 1
0.31 1
0.32 1
0.33 0.99
0.34 1
0.35 1
0.36 1
0.37 1
0.38 1
0.39 1
0.4 1
0.41 1
0.42 1
0.43 1
0.44 1
0.45 1
0.46 1
0.47 1
0.48 1
0.49 1
0.5 0.99
0.51 1
0.52 1
0.53 1
0.54 1
0.55 1
0.56 1
0.57 1
0.58 0.99
0.59 1
0.6 1
0.61 1
0.62 1
0.63 1
0.64 1
0.65 1
0.66 0.99
0.67 1
0.68 1
0.69 1
0.7 0.99
0.71 1
0.72 1
0.73 1
0.74 1
0.75 0.99
0.76 0.99
0.77 1
0.78 1
0.79 0.99
0.8 0.99
0.81 1
0.82 1
0.83 0.99
0.84 1
0.85 1
0.86 0.99
0.87 1
0.88 1
0.89 1
0.9 0.98
0.91 0.99
0.92 1
0.93 1
0.94 1
0.95 1
0.96 1
0.97 1
0.98 1
0.99 1
1 0.99
1.02 0.98
1.03 0.81
1.04 0.77
1.05 0.74
1.06 0.79
1.07 0.66
1.08 0.69
1.09 0.63
1.1 0.63
1.11 0.64
1.12 0.55
1.13 0.63
1.14 0.61
1.15 0.58
1.16 0.54
1.17 0.58
1.18 0.56
1.19 0.44
1.2 0.53
1.21 0.5
1.22 0.49
1.23 0.51
1.24 0.51
1.25 0.51
1.26 0.45
1.27 0.5
1.28 0.57
1.29 0.46
1.3 0.51
1.31 0.48
1.32 0.54
1.33 0.54
1.34 0.51
1.35 0.56
1.36 0.48
1.37 0.51
1.38 0.5
1.39 0.56
1.4 0.53
1.41 0.44
1.42 0.56
1.43 0.53
1.44 0.56
1.45 0.49
1.46 0.43
1.47 0.47
1.48 0.52
1.49 0.5
1.5 0.42
1.51 0.53
1.52 0.5
1.53 0.56
1.54 0.49
1.55 0.55
1.56 0.51
1.57 0.55
1.58 0.52
1.59 0.59
1.6 0.54
1.61 0.58
1.62 0.43
1.63 0.43
1.64 0.55
1.65 0.52
1.66 0.53
1.67 0.51
1.68 0.56
1.69 0.39
1.7 0.47
1.71 0.49
1.72 0.51
1.73 0.54
1.74 0.57
1.75 0.51
1.76 0.54
1.77 0.51
1.78 0.48
1.79 0.44
1.8 0.47
1.81 0.61
1.82 0.46
1.83 0.48
1.84 0.57
1.85 0.54
1.86 0.55
1.87 0.56
1.88 0.47
1.89 0.42
1.9 0.48
1.91 0.44
1.92 0.44
1.93 0.55
1.94 0.45
1.95 0.57
1.96 0.49
1.97 0.57
1.98 0.48
1.99 0.54
2 0.61
};
\addplot [semithick, color11, opacity=0.75, dashed]
table {%
0.01 1
0.02 1
0.03 1
0.04 1
0.05 1
0.06 1
0.07 1
0.08 1
0.09 1
0.1 1
0.11 1
0.12 1
0.13 1
0.14 1
0.15 1
0.16 1
0.17 1
0.18 1
0.19 1
0.2 1
0.21 1
0.22 1
0.23 1
0.24 1
0.25 1
0.26 1
0.27 1
0.28 1
0.29 1
0.3 1
0.31 1
0.32 1
0.33 1
0.34 1
0.35 1
0.36 1
0.37 1
0.38 1
0.39 1
0.4 1
0.41 1
0.42 1
0.43 1
0.44 1
0.45 1
0.46 1
0.47 1
0.48 1
0.49 1
0.5 1
0.51 1
0.52 1
0.53 1
0.54 1
0.55 1
0.56 1
0.57 1
0.58 1
0.59 1
0.6 1
0.61 1
0.62 1
0.63 1
0.64 1
0.65 1
0.66 1
0.67 1
0.68 1
0.69 1
0.7 1
0.71 1
0.72 1
0.73 1
0.74 1
0.75 1
0.76 1
0.77 1
0.78 1
0.79 1
0.8 1
0.81 1
0.82 1
0.83 1
0.84 1
0.85 1
0.86 1
0.87 1
0.88 1
0.89 1
0.9 1
0.91 1
0.92 1
0.93 1
0.94 1
0.95 1
0.96 1
0.97 1
0.98 1
0.99 1
1 1
1.02 1
1.03 1
1.04 1
1.05 0.98
1.06 1
1.07 1
1.08 0.98
1.09 1
1.1 1
1.11 1
1.12 1
1.13 0.99
1.14 0.96
1.15 0.99
1.16 0.98
1.17 0.99
1.18 0.98
1.19 0.97
1.2 1
1.21 0.99
1.22 0.98
1.23 0.95
1.24 0.99
1.25 0.99
1.26 0.98
1.27 0.99
1.28 1
1.29 0.97
1.3 0.99
1.31 0.98
1.32 0.97
1.33 1
1.34 0.98
1.35 0.99
1.36 0.97
1.37 1
1.38 0.97
1.39 0.98
1.4 0.99
1.41 0.99
1.42 0.99
1.43 0.99
1.44 0.96
1.45 0.98
1.46 0.97
1.47 0.98
1.48 0.99
1.49 0.99
1.5 1
1.51 0.99
1.52 0.99
1.53 0.96
1.54 0.98
1.55 0.97
1.56 1
1.57 0.97
1.58 0.97
1.59 0.98
1.6 1
1.61 0.98
1.62 0.99
1.63 0.98
1.64 0.97
1.65 0.99
1.66 0.96
1.67 0.99
1.68 0.98
1.69 1
1.7 0.99
1.71 1
1.72 0.96
1.73 0.99
1.74 0.99
1.75 0.99
1.76 0.96
1.77 0.99
1.78 0.96
1.79 0.99
1.8 0.98
1.81 0.99
1.82 0.96
1.83 0.98
1.84 0.99
1.85 0.99
1.86 0.99
1.87 0.99
1.88 0.98
1.89 0.98
1.9 0.98
1.91 0.96
1.92 0.97
1.93 0.99
1.94 0.99
1.95 0.99
1.96 0.97
1.97 0.96
1.98 0.99
1.99 0.98
2 1
};
\addplot [semithick, black, opacity=1, dash pattern=on 1pt off 1pt]
table {%
-0.0895000000000001 0.5
2.0995 0.5
};
\addplot [semithick, black, opacity=1, dash pattern=on 1pt off 1pt]
table {%
-0.0895000000000001 1
2.0995 1
};
\addplot [semithick, black, opacity=1, dash pattern=on 1pt off 1pt]
table {%
1 0.2
1 1.05
};
\end{axis}

\end{tikzpicture}

%% file: plots/decoupled2/NL_LAPxGAU.tex
% This file was created by tikzplotlib v0.9.6.
\begin{tikzpicture}

\definecolor{color0}{rgb}{0.866666666666667,0.494117647058824,0.164705882352941}
\definecolor{color1}{rgb}{0.164705882352941,0.643137254901961,0.866666666666667}
\definecolor{color2}{rgb}{0.584313725490196,0.866666666666667,0.164705882352941}
\definecolor{color3}{rgb}{0.109803921568627,0.337254901960784,0.129411764705882}
\definecolor{color4}{rgb}{0.529411764705882,0.305882352941176,0.858823529411765}
\definecolor{color5}{rgb}{0.858823529411765,0.305882352941176,0.435294117647059}
\definecolor{color6}{rgb}{0.937254901960784,0.929411764705882,0.392156862745098}
\definecolor{color7}{rgb}{0.0901960784313725,0.486274509803922,0.0980392156862745}
\definecolor{color8}{rgb}{0.156862745098039,0.188235294117647,0.827450980392157}
\definecolor{color9}{rgb}{0.937254901960784,0.392156862745098,0.894117647058824}
\definecolor{color10}{rgb}{0.2,0.184313725490196,0.184313725490196}
\definecolor{color11}{rgb}{0.0156862745098039,0.803921568627451,0.976470588235294}

\begin{axis}[
tick align=outside,
tick pos=left,
x grid style={white!69.0196078431373!black},
xmajorgrids,
xmin=-0.0895, xmax=2.0995,
xtick style={color=black},
xtick={0,0.1,0.2,0.3,0.4,0.5,0.6,0.7,0.8,0.9,1,1.1,1.2,1.3,1.4,1.5,1.6,1.7,1.8,1.9,2},
xticklabels={0,,.2,,.4,,.6,,.8,,1,,20,,40,,60,,80,,100},
height=4.8cm,
width=6.5cm,
y grid style={white!69.0196078431373!black},
ymajorgrids,
ymin=0.2, ymax=1.05,
ytick style={color=black}
]
\addplot [semithick, color0, opacity=0.75]
table {%
0.01 0.61
0.02 0.7
0.03 0.84
0.04 0.91
0.05 0.95
0.06 0.95
0.07 0.95
0.08 0.97
0.09 0.96
0.1 0.98
0.11 1
0.12 1
0.13 1
0.14 1
0.15 1
0.16 1
0.17 1
0.18 0.99
0.19 1
0.2 1
0.21 1
0.22 1
0.23 1
0.24 1
0.25 1
0.26 1
0.27 1
0.28 1
0.29 1
0.3 1
0.31 1
0.32 1
0.33 1
0.34 1
0.35 1
0.36 1
0.37 1
0.38 1
0.39 1
0.4 1
0.41 1
0.42 1
0.43 1
0.44 1
0.45 1
0.46 1
0.47 1
0.48 1
0.49 1
0.5 1
0.51 1
0.52 1
0.53 1
0.54 1
0.55 1
0.56 1
0.57 1
0.58 1
0.59 1
0.6 1
0.61 1
0.62 1
0.63 1
0.64 1
0.65 1
0.66 1
0.67 1
0.68 1
0.69 1
0.7 1
0.71 1
0.72 1
0.73 1
0.74 1
0.75 1
0.76 1
0.77 1
0.78 1
0.79 1
0.8 1
0.81 1
0.82 1
0.83 1
0.84 1
0.85 1
0.86 1
0.87 1
0.88 1
0.89 1
0.9 1
0.91 1
0.92 1
0.93 1
0.94 1
0.95 1
0.96 1
0.97 1
0.98 1
0.99 1
1 1
1.02 1
1.03 1
1.04 1
1.05 1
1.06 1
1.07 1
1.08 1
1.09 1
1.1 1
1.11 1
1.12 1
1.13 1
1.14 1
1.15 1
1.16 1
1.17 1
1.18 1
1.19 1
1.2 1
1.21 1
1.22 1
1.23 1
1.24 1
1.25 1
1.26 1
1.27 1
1.28 1
1.29 1
1.3 1
1.31 1
1.32 1
1.33 1
1.34 1
1.35 1
1.36 1
1.37 1
1.38 1
1.39 1
1.4 1
1.41 1
1.42 1
1.43 1
1.44 1
1.45 1
1.46 1
1.47 1
1.48 1
1.49 1
1.5 1
1.51 1
1.52 1
1.53 1
1.54 1
1.55 1
1.56 1
1.57 1
1.58 1
1.59 0.99
1.6 1
1.61 1
1.62 1
1.63 1
1.64 1
1.65 1
1.66 1
1.67 1
1.68 1
1.69 1
1.7 1
1.71 1
1.72 1
1.73 1
1.74 1
1.75 1
1.76 1
1.77 1
1.78 1
1.79 1
1.8 1
1.81 1
1.82 1
1.83 1
1.84 1
1.85 1
1.86 0.99
1.87 1
1.88 1
1.89 1
1.9 0.99
1.91 0.99
1.92 1
1.93 1
1.94 1
1.95 1
1.96 1
1.97 0.98
1.98 0.97
1.99 0.99
2 0.99
};
\addplot [semithick, color1, opacity=0.75]
table {%
0.01 0.57
0.02 0.58
0.03 0.65
0.04 0.84
0.05 0.87
0.06 0.84
0.07 0.89
0.08 0.92
0.09 0.91
0.1 0.95
0.11 0.98
0.12 0.99
0.13 0.98
0.14 0.96
0.15 0.97
0.16 1
0.17 1
0.18 1
0.19 1
0.2 1
0.21 1
0.22 1
0.23 1
0.24 1
0.25 1
0.26 1
0.27 1
0.28 1
0.29 1
0.3 1
0.31 1
0.32 1
0.33 1
0.34 1
0.35 1
0.36 1
0.37 1
0.38 1
0.39 1
0.4 1
0.41 1
0.42 1
0.43 1
0.44 1
0.45 1
0.46 1
0.47 1
0.48 1
0.49 1
0.5 1
0.51 1
0.52 1
0.53 1
0.54 1
0.55 1
0.56 1
0.57 1
0.58 1
0.59 1
0.6 1
0.61 1
0.62 1
0.63 1
0.64 1
0.65 1
0.66 1
0.67 1
0.68 1
0.69 1
0.7 1
0.71 1
0.72 1
0.73 1
0.74 1
0.75 1
0.76 1
0.77 1
0.78 1
0.79 1
0.8 1
0.81 1
0.82 1
0.83 1
0.84 1
0.85 1
0.86 1
0.87 1
0.88 1
0.89 1
0.9 1
0.91 1
0.92 1
0.93 1
0.94 1
0.95 1
0.96 1
0.97 1
0.98 1
0.99 1
1 1
1.02 1
1.03 1
1.04 1
1.05 1
1.06 1
1.07 1
1.08 1
1.09 1
1.1 1
1.11 1
1.12 1
1.13 1
1.14 1
1.15 1
1.16 1
1.17 1
1.18 1
1.19 1
1.2 1
1.21 1
1.22 1
1.23 1
1.24 1
1.25 1
1.26 1
1.27 1
1.28 1
1.29 1
1.3 1
1.31 1
1.32 1
1.33 1
1.34 1
1.35 1
1.36 1
1.37 1
1.38 1
1.39 1
1.4 1
1.41 1
1.42 1
1.43 1
1.44 1
1.45 1
1.46 1
1.47 1
1.48 0.99
1.49 0.98
1.5 1
1.51 0.99
1.52 1
1.53 1
1.54 1
1.55 1
1.56 1
1.57 1
1.58 1
1.59 1
1.6 1
1.61 0.99
1.62 0.99
1.63 1
1.64 0.99
1.65 0.99
1.66 1
1.67 0.99
1.68 1
1.69 1
1.7 0.97
1.71 1
1.72 1
1.73 0.99
1.74 0.99
1.75 1
1.76 0.98
1.77 1
1.78 0.99
1.79 0.98
1.8 0.98
1.81 1
1.82 0.97
1.83 0.99
1.84 1
1.85 1
1.86 0.96
1.87 0.99
1.88 0.98
1.89 1
1.9 0.98
1.91 0.97
1.92 1
1.93 0.97
1.94 0.99
1.95 1
1.96 0.98
1.97 0.97
1.98 0.98
1.99 0.97
2 0.97
};
\addplot [semithick, color2, opacity=0.75]
table {%
0.01 0.53
0.02 0.55
0.03 0.5
0.04 0.7
0.05 0.7
0.06 0.78
0.07 0.8
0.08 0.79
0.09 0.87
0.1 0.91
0.11 0.98
0.12 0.99
0.13 0.98
0.14 0.96
0.15 0.97
0.16 0.99
0.17 1
0.18 1
0.19 1
0.2 1
0.21 1
0.22 1
0.23 1
0.24 1
0.25 1
0.26 1
0.27 1
0.28 1
0.29 1
0.3 1
0.31 1
0.32 1
0.33 1
0.34 1
0.35 1
0.36 1
0.37 1
0.38 1
0.39 1
0.4 1
0.41 1
0.42 1
0.43 1
0.44 1
0.45 1
0.46 1
0.47 1
0.48 1
0.49 1
0.5 1
0.51 1
0.52 1
0.53 1
0.54 1
0.55 1
0.56 1
0.57 1
0.58 1
0.59 1
0.6 1
0.61 1
0.62 1
0.63 1
0.64 1
0.65 1
0.66 1
0.67 1
0.68 1
0.69 1
0.7 1
0.71 1
0.72 1
0.73 1
0.74 1
0.75 1
0.76 1
0.77 1
0.78 1
0.79 1
0.8 1
0.81 1
0.82 1
0.83 1
0.84 1
0.85 1
0.86 1
0.87 1
0.88 1
0.89 1
0.9 1
0.91 1
0.92 1
0.93 1
0.94 1
0.95 1
0.96 1
0.97 1
0.98 1
0.99 1
1 1
1.02 1
1.03 1
1.04 1
1.05 1
1.06 1
1.07 1
1.08 1
1.09 1
1.1 1
1.11 1
1.12 1
1.13 1
1.14 1
1.15 1
1.16 1
1.17 1
1.18 1
1.19 1
1.2 1
1.21 1
1.22 1
1.23 1
1.24 1
1.25 1
1.26 1
1.27 1
1.28 1
1.29 1
1.3 1
1.31 1
1.32 1
1.33 1
1.34 1
1.35 1
1.36 1
1.37 1
1.38 1
1.39 1
1.4 1
1.41 1
1.42 1
1.43 1
1.44 1
1.45 1
1.46 1
1.47 0.99
1.48 0.99
1.49 0.98
1.5 1
1.51 0.99
1.52 1
1.53 1
1.54 1
1.55 1
1.56 1
1.57 1
1.58 1
1.59 1
1.6 1
1.61 0.99
1.62 0.99
1.63 1
1.64 1
1.65 0.99
1.66 1
1.67 0.99
1.68 1
1.69 1
1.7 0.97
1.71 0.99
1.72 1
1.73 0.99
1.74 0.99
1.75 1
1.76 0.99
1.77 1
1.78 0.99
1.79 0.99
1.8 0.99
1.81 1
1.82 0.98
1.83 0.99
1.84 1
1.85 0.99
1.86 0.96
1.87 0.99
1.88 0.97
1.89 1
1.9 0.98
1.91 0.96
1.92 1
1.93 0.98
1.94 0.99
1.95 1
1.96 0.97
1.97 0.98
1.98 0.98
1.99 0.97
2 0.96
};
\addplot [semithick, color3, opacity=0.75]
table {%
0.01 0.77
0.02 0.78
0.03 0.9
0.04 0.93
0.05 0.96
0.06 0.98
0.07 0.97
0.08 0.97
0.09 0.99
0.1 0.97
0.11 0.96
0.12 0.94
0.13 0.99
0.14 0.97
0.15 1
0.16 1
0.17 0.98
0.18 1
0.19 0.99
0.2 1
0.21 1
0.22 1
0.23 1
0.24 1
0.25 1
0.26 1
0.27 1
0.28 0.99
0.29 1
0.3 1
0.31 0.99
0.32 1
0.33 1
0.34 1
0.35 1
0.36 0.99
0.37 1
0.38 1
0.39 1
0.4 1
0.41 0.99
0.42 0.99
0.43 0.99
0.44 1
0.45 1
0.46 0.98
0.47 1
0.48 1
0.49 1
0.5 1
0.51 1
0.52 1
0.53 1
0.54 1
0.55 1
0.56 1
0.57 1
0.58 1
0.59 1
0.6 1
0.61 1
0.62 1
0.63 1
0.64 0.99
0.65 0.99
0.66 1
0.67 1
0.68 1
0.69 1
0.7 1
0.71 1
0.72 1
0.73 1
0.74 1
0.75 1
0.76 1
0.77 1
0.78 1
0.79 1
0.8 1
0.81 1
0.82 1
0.83 1
0.84 1
0.85 1
0.86 1
0.87 1
0.88 1
0.89 1
0.9 1
0.91 1
0.92 1
0.93 1
0.94 1
0.95 1
0.96 1
0.97 1
0.98 1
0.99 1
1 1
1.02 1
1.03 1
1.04 1
1.05 1
1.06 1
1.07 1
1.08 1
1.09 1
1.1 1
1.11 1
1.12 1
1.13 1
1.14 1
1.15 1
1.16 1
1.17 1
1.18 1
1.19 1
1.2 1
1.21 1
1.22 1
1.23 1
1.24 1
1.25 1
1.26 1
1.27 1
1.28 1
1.29 1
1.3 1
1.31 1
1.32 1
1.33 1
1.34 1
1.35 1
1.36 1
1.37 1
1.38 1
1.39 1
1.4 1
1.41 1
1.42 1
1.43 1
1.44 1
1.45 1
1.46 1
1.47 1
1.48 1
1.49 0.99
1.5 1
1.51 1
1.52 1
1.53 1
1.54 1
1.55 1
1.56 1
1.57 1
1.58 1
1.59 1
1.6 1
1.61 1
1.62 1
1.63 1
1.64 1
1.65 1
1.66 1
1.67 1
1.68 1
1.69 1
1.7 1
1.71 1
1.72 1
1.73 1
1.74 1
1.75 1
1.76 1
1.77 1
1.78 1
1.79 1
1.8 1
1.81 0.99
1.82 1
1.83 1
1.84 1
1.85 1
1.86 0.99
1.87 1
1.88 0.99
1.89 0.99
1.9 1
1.91 0.99
1.92 0.99
1.93 1
1.94 1
1.95 1
1.96 1
1.97 0.98
1.98 0.97
1.99 0.98
2 0.99
};
\addplot [semithick, color4, opacity=0.75]
table {%
0.01 0.77
0.02 0.78
0.03 0.91
0.04 0.93
0.05 0.96
0.06 0.98
0.07 0.97
0.08 0.97
0.09 0.99
0.1 0.98
0.11 0.96
0.12 0.94
0.13 0.99
0.14 0.97
0.15 1
0.16 1
0.17 0.98
0.18 1
0.19 0.99
0.2 1
0.21 1
0.22 1
0.23 1
0.24 1
0.25 1
0.26 1
0.27 1
0.28 0.99
0.29 1
0.3 1
0.31 0.99
0.32 1
0.33 1
0.34 1
0.35 1
0.36 0.99
0.37 1
0.38 1
0.39 1
0.4 1
0.41 0.99
0.42 0.99
0.43 0.99
0.44 1
0.45 1
0.46 0.98
0.47 1
0.48 1
0.49 1
0.5 1
0.51 1
0.52 1
0.53 1
0.54 1
0.55 1
0.56 1
0.57 1
0.58 1
0.59 1
0.6 1
0.61 1
0.62 1
0.63 1
0.64 0.99
0.65 0.99
0.66 1
0.67 1
0.68 1
0.69 1
0.7 1
0.71 1
0.72 1
0.73 1
0.74 1
0.75 1
0.76 1
0.77 1
0.78 1
0.79 1
0.8 1
0.81 1
0.82 1
0.83 1
0.84 1
0.85 1
0.86 1
0.87 1
0.88 1
0.89 1
0.9 1
0.91 1
0.92 1
0.93 1
0.94 1
0.95 1
0.96 1
0.97 1
0.98 1
0.99 1
1 1
1.02 1
1.03 1
1.04 1
1.05 1
1.06 1
1.07 1
1.08 1
1.09 1
1.1 1
1.11 1
1.12 1
1.13 1
1.14 1
1.15 1
1.16 1
1.17 1
1.18 1
1.19 1
1.2 1
1.21 1
1.22 1
1.23 1
1.24 1
1.25 1
1.26 1
1.27 1
1.28 1
1.29 1
1.3 1
1.31 1
1.32 1
1.33 1
1.34 1
1.35 1
1.36 1
1.37 1
1.38 1
1.39 1
1.4 1
1.41 1
1.42 1
1.43 1
1.44 1
1.45 1
1.46 1
1.47 1
1.48 1
1.49 0.99
1.5 1
1.51 1
1.52 1
1.53 1
1.54 1
1.55 1
1.56 1
1.57 1
1.58 1
1.59 1
1.6 1
1.61 1
1.62 1
1.63 1
1.64 1
1.65 1
1.66 1
1.67 1
1.68 1
1.69 1
1.7 1
1.71 1
1.72 1
1.73 1
1.74 1
1.75 1
1.76 1
1.77 1
1.78 1
1.79 1
1.8 1
1.81 0.99
1.82 1
1.83 1
1.84 1
1.85 1
1.86 0.99
1.87 1
1.88 0.99
1.89 0.99
1.9 1
1.91 0.99
1.92 0.99
1.93 1
1.94 1
1.95 1
1.96 1
1.97 0.98
1.98 0.97
1.99 0.99
2 0.99
};
\addplot [semithick, color5, opacity=0.75]
table {%
0.01 0.87
0.02 0.91
0.03 0.98
0.04 1
0.05 0.99
0.06 1
0.07 0.99
0.08 1
0.09 1
0.1 1
0.11 1
0.12 1
0.13 1
0.14 0.99
0.15 1
0.16 1
0.17 1
0.18 0.99
0.19 1
0.2 1
0.21 1
0.22 1
0.23 1
0.24 1
0.25 1
0.26 1
0.27 1
0.28 1
0.29 1
0.3 1
0.31 1
0.32 1
0.33 1
0.34 1
0.35 1
0.36 1
0.37 1
0.38 1
0.39 1
0.4 1
0.41 1
0.42 1
0.43 1
0.44 1
0.45 1
0.46 1
0.47 1
0.48 1
0.49 1
0.5 1
0.51 1
0.52 1
0.53 1
0.54 1
0.55 1
0.56 1
0.57 1
0.58 1
0.59 1
0.6 1
0.61 1
0.62 1
0.63 1
0.64 1
0.65 1
0.66 1
0.67 1
0.68 1
0.69 1
0.7 1
0.71 1
0.72 1
0.73 1
0.74 1
0.75 1
0.76 1
0.77 1
0.78 1
0.79 1
0.8 1
0.81 1
0.82 1
0.83 1
0.84 1
0.85 1
0.86 1
0.87 1
0.88 1
0.89 1
0.9 1
0.91 1
0.92 1
0.93 1
0.94 1
0.95 1
0.96 1
0.97 1
0.98 1
0.99 1
1 1
1.02 1
1.03 1
1.04 1
1.05 1
1.06 1
1.07 1
1.08 1
1.09 1
1.1 1
1.11 1
1.12 1
1.13 1
1.14 1
1.15 1
1.16 1
1.17 1
1.18 1
1.19 1
1.2 1
1.21 1
1.22 1
1.23 1
1.24 1
1.25 1
1.26 1
1.27 1
1.28 1
1.29 1
1.3 1
1.31 1
1.32 1
1.33 1
1.34 1
1.35 1
1.36 1
1.37 1
1.38 1
1.39 1
1.4 1
1.41 1
1.42 1
1.43 1
1.44 1
1.45 1
1.46 1
1.47 1
1.48 1
1.49 1
1.5 1
1.51 1
1.52 1
1.53 1
1.54 1
1.55 1
1.56 1
1.57 1
1.58 1
1.59 1
1.6 1
1.61 1
1.62 1
1.63 1
1.64 1
1.65 1
1.66 1
1.67 1
1.68 1
1.69 1
1.7 1
1.71 1
1.72 1
1.73 1
1.74 1
1.75 1
1.76 1
1.77 1
1.78 1
1.79 1
1.8 1
1.81 0.99
1.82 1
1.83 1
1.84 1
1.85 1
1.86 0.99
1.87 1
1.88 1
1.89 1
1.9 1
1.91 1
1.92 1
1.93 1
1.94 1
1.95 1
1.96 1
1.97 0.99
1.98 0.99
1.99 1
2 1
};
\addplot [semithick, color6, opacity=0.75, dashed]
table {%
0.01 1
0.02 1
0.03 1
0.04 1
0.05 1
0.06 1
0.07 1
0.08 1
0.09 1
0.1 1
0.11 1
0.12 1
0.13 1
0.14 1
0.15 1
0.16 1
0.17 1
0.18 1
0.19 1
0.2 1
0.21 1
0.22 1
0.23 1
0.24 1
0.25 1
0.26 1
0.27 1
0.28 1
0.29 1
0.3 1
0.31 1
0.32 1
0.33 1
0.34 1
0.35 1
0.36 1
0.37 1
0.38 1
0.39 1
0.4 1
0.41 1
0.42 1
0.43 1
0.44 1
0.45 1
0.46 1
0.47 1
0.48 1
0.49 1
0.5 1
0.51 1
0.52 1
0.53 1
0.54 1
0.55 1
0.56 1
0.57 1
0.58 1
0.59 1
0.6 1
0.61 1
0.62 1
0.63 1
0.64 1
0.65 1
0.66 1
0.67 1
0.68 1
0.69 1
0.7 1
0.71 1
0.72 1
0.73 1
0.74 1
0.75 1
0.76 1
0.77 1
0.78 1
0.79 1
0.8 1
0.81 1
0.82 1
0.83 1
0.84 1
0.85 1
0.86 1
0.87 1
0.88 1
0.89 1
0.9 1
0.91 1
0.92 1
0.93 1
0.94 1
0.95 1
0.96 1
0.97 1
0.98 1
0.99 1
1 1
1.02 1
1.03 1
1.04 1
1.05 1
1.06 1
1.07 1
1.08 1
1.09 1
1.1 1
1.11 1
1.12 1
1.13 1
1.14 1
1.15 1
1.16 1
1.17 1
1.18 1
1.19 1
1.2 1
1.21 1
1.22 1
1.23 1
1.24 1
1.25 1
1.26 1
1.27 1
1.28 1
1.29 1
1.3 1
1.31 1
1.32 1
1.33 1
1.34 1
1.35 1
1.36 1
1.37 1
1.38 1
1.39 1
1.4 1
1.41 1
1.42 1
1.43 1
1.44 1
1.45 1
1.46 1
1.47 1
1.48 1
1.49 1
1.5 1
1.51 1
1.52 1
1.53 1
1.54 1
1.55 1
1.56 1
1.57 1
1.58 1
1.59 1
1.6 1
1.61 1
1.62 1
1.63 1
1.64 1
1.65 1
1.66 1
1.67 1
1.68 1
1.69 1
1.7 1
1.71 1
1.72 1
1.73 1
1.74 1
1.75 1
1.76 1
1.77 1
1.78 1
1.79 1
1.8 1
1.81 1
1.82 1
1.83 1
1.84 1
1.85 1
1.86 1
1.87 1
1.88 1
1.89 1
1.9 1
1.91 1
1.92 1
1.93 1
1.94 1
1.95 1
1.96 1
1.97 1
1.98 1
1.99 1
2 1
};
\addplot [semithick, color7, opacity=0.75, dashed]
table {%
0.01 1
0.02 1
0.03 1
0.04 1
0.05 1
0.06 1
0.07 1
0.08 1
0.09 1
0.1 1
0.11 1
0.12 1
0.13 1
0.14 1
0.15 1
0.16 1
0.17 1
0.18 1
0.19 1
0.2 1
0.21 1
0.22 1
0.23 1
0.24 1
0.25 1
0.26 1
0.27 1
0.28 1
0.29 1
0.3 1
0.31 1
0.32 1
0.33 1
0.34 1
0.35 1
0.36 1
0.37 1
0.38 1
0.39 1
0.4 1
0.41 1
0.42 1
0.43 1
0.44 1
0.45 1
0.46 1
0.47 1
0.48 1
0.49 1
0.5 1
0.51 1
0.52 1
0.53 1
0.54 1
0.55 1
0.56 1
0.57 1
0.58 1
0.59 1
0.6 1
0.61 1
0.62 1
0.63 1
0.64 1
0.65 1
0.66 1
0.67 1
0.68 1
0.69 1
0.7 1
0.71 1
0.72 1
0.73 1
0.74 1
0.75 1
0.76 1
0.77 1
0.78 1
0.79 1
0.8 1
0.81 1
0.82 1
0.83 1
0.84 1
0.85 1
0.86 1
0.87 1
0.88 1
0.89 1
0.9 1
0.91 1
0.92 1
0.93 1
0.94 1
0.95 1
0.96 1
0.97 1
0.98 1
0.99 1
1 1
1.02 1
1.03 1
1.04 1
1.05 1
1.06 1
1.07 1
1.08 1
1.09 1
1.1 1
1.11 1
1.12 1
1.13 1
1.14 1
1.15 1
1.16 1
1.17 1
1.18 1
1.19 1
1.2 1
1.21 1
1.22 1
1.23 1
1.24 1
1.25 1
1.26 1
1.27 1
1.28 1
1.29 1
1.3 1
1.31 1
1.32 1
1.33 1
1.34 1
1.35 1
1.36 1
1.37 1
1.38 1
1.39 1
1.4 1
1.41 1
1.42 1
1.43 1
1.44 1
1.45 1
1.46 1
1.47 1
1.48 1
1.49 1
1.5 1
1.51 1
1.52 1
1.53 1
1.54 1
1.55 1
1.56 1
1.57 1
1.58 1
1.59 1
1.6 1
1.61 1
1.62 1
1.63 1
1.64 1
1.65 1
1.66 1
1.67 1
1.68 1
1.69 1
1.7 1
1.71 1
1.72 1
1.73 1
1.74 1
1.75 1
1.76 1
1.77 1
1.78 1
1.79 1
1.8 1
1.81 1
1.82 1
1.83 1
1.84 1
1.85 1
1.86 1
1.87 1
1.88 1
1.89 1
1.9 1
1.91 1
1.92 1
1.93 1
1.94 1
1.95 1
1.96 1
1.97 1
1.98 1
1.99 1
2 1
};
\addplot [semithick, color8, opacity=0.75, dashed]
table {%
0.01 1
0.02 1
0.03 1
0.04 1
0.05 1
0.06 1
0.07 1
0.08 1
0.09 1
0.1 1
0.11 1
0.12 1
0.13 1
0.14 1
0.15 1
0.16 1
0.17 1
0.18 1
0.19 1
0.2 1
0.21 1
0.22 1
0.23 1
0.24 1
0.25 1
0.26 1
0.27 1
0.28 1
0.29 1
0.3 1
0.31 1
0.32 1
0.33 1
0.34 1
0.35 1
0.36 1
0.37 1
0.38 1
0.39 1
0.4 1
0.41 1
0.42 1
0.43 1
0.44 1
0.45 1
0.46 1
0.47 1
0.48 1
0.49 1
0.5 1
0.51 1
0.52 1
0.53 1
0.54 1
0.55 1
0.56 1
0.57 1
0.58 1
0.59 1
0.6 1
0.61 1
0.62 1
0.63 1
0.64 1
0.65 1
0.66 1
0.67 1
0.68 1
0.69 1
0.7 1
0.71 1
0.72 1
0.73 1
0.74 1
0.75 1
0.76 1
0.77 1
0.78 1
0.79 1
0.8 1
0.81 1
0.82 1
0.83 1
0.84 1
0.85 1
0.86 1
0.87 1
0.88 1
0.89 1
0.9 1
0.91 1
0.92 1
0.93 1
0.94 1
0.95 1
0.96 1
0.97 1
0.98 1
0.99 1
1 1
1.02 1
1.03 1
1.04 1
1.05 1
1.06 1
1.07 1
1.08 1
1.09 1
1.1 1
1.11 1
1.12 1
1.13 1
1.14 1
1.15 1
1.16 1
1.17 1
1.18 1
1.19 1
1.2 1
1.21 1
1.22 1
1.23 1
1.24 1
1.25 1
1.26 1
1.27 1
1.28 1
1.29 1
1.3 1
1.31 1
1.32 1
1.33 1
1.34 1
1.35 1
1.36 1
1.37 1
1.38 1
1.39 1
1.4 1
1.41 1
1.42 1
1.43 1
1.44 1
1.45 1
1.46 1
1.47 1
1.48 1
1.49 1
1.5 1
1.51 1
1.52 1
1.53 1
1.54 1
1.55 1
1.56 1
1.57 1
1.58 1
1.59 1
1.6 1
1.61 1
1.62 1
1.63 1
1.64 1
1.65 1
1.66 1
1.67 1
1.68 1
1.69 1
1.7 1
1.71 1
1.72 1
1.73 1
1.74 1
1.75 1
1.76 1
1.77 1
1.78 1
1.79 1
1.8 1
1.81 1
1.82 1
1.83 1
1.84 1
1.85 1
1.86 1
1.87 1
1.88 1
1.89 1
1.9 1
1.91 1
1.92 1
1.93 1
1.94 1
1.95 1
1.96 1
1.97 1
1.98 1
1.99 1
2 1
};
\addplot [semithick, color9, opacity=0.75, dashed]
table {%
0.01 0.56
0.02 0.51
0.03 0.56
0.04 0.56
0.05 0.6
0.06 0.73
0.07 0.68
0.08 0.65
0.09 0.7
0.1 0.65
0.11 0.75
0.12 0.63
0.13 0.75
0.14 0.76
0.15 0.8
0.16 0.77
0.17 0.67
0.18 0.79
0.19 0.83
0.2 0.89
0.21 0.89
0.22 0.81
0.23 0.79
0.24 0.83
0.25 0.82
0.26 0.83
0.27 0.81
0.28 0.89
0.29 0.92
0.3 0.91
0.31 0.87
0.32 0.92
0.33 0.94
0.34 0.95
0.35 0.89
0.36 0.86
0.37 0.91
0.38 0.93
0.39 0.91
0.4 0.88
0.41 0.91
0.42 0.93
0.43 0.94
0.44 0.94
0.45 0.96
0.46 0.94
0.47 0.94
0.48 0.95
0.49 0.93
0.5 0.97
0.51 0.96
0.52 0.96
0.53 0.96
0.54 0.96
0.55 0.97
0.56 0.95
0.57 0.99
0.58 0.96
0.59 0.98
0.6 0.97
0.61 0.97
0.62 0.99
0.63 0.97
0.64 0.97
0.65 0.95
0.66 0.99
0.67 1
0.68 0.97
0.69 0.99
0.7 1
0.71 0.98
0.72 0.97
0.73 1
0.74 0.99
0.75 0.96
0.76 0.99
0.77 0.99
0.78 0.99
0.79 0.97
0.8 1
0.81 0.99
0.82 0.99
0.83 0.99
0.84 1
0.85 1
0.86 1
0.87 0.99
0.88 1
0.89 1
0.9 0.98
0.91 1
0.92 1
0.93 1
0.94 0.99
0.95 0.99
0.96 0.99
0.97 0.99
0.98 1
0.99 0.99
1 0.99
1.02 0.99
1.03 0.99
1.04 0.99
1.05 0.99
1.06 1
1.07 1
1.08 1
1.09 1
1.1 1
1.11 1
1.12 0.99
1.13 1
1.14 1
1.15 1
1.16 0.94
1.17 0.98
1.18 0.99
1.19 1
1.2 1
1.21 1
1.22 0.99
1.23 0.99
1.24 1
1.25 0.99
1.26 0.98
1.27 0.99
1.28 1
1.29 0.99
1.3 1
1.31 1
1.32 1
1.33 1
1.34 0.99
1.35 1
1.36 1
1.37 1
1.38 1
1.39 1
1.4 1
1.41 0.99
1.42 1
1.43 0.99
1.44 0.99
1.45 1
1.46 0.98
1.47 1
1.48 0.98
1.49 1
1.5 0.99
1.51 1
1.52 1
1.53 0.99
1.54 1
1.55 1
1.56 1
1.57 0.99
1.58 0.99
1.59 0.98
1.6 0.97
1.61 0.98
1.62 0.97
1.63 0.99
1.64 0.97
1.65 0.98
1.66 0.99
1.67 0.95
1.68 1
1.69 1
1.7 0.98
1.71 0.98
1.72 0.98
1.73 0.99
1.74 0.95
1.75 0.98
1.76 0.94
1.77 0.96
1.78 0.98
1.79 0.98
1.8 0.98
1.81 0.96
1.82 0.97
1.83 0.98
1.84 0.97
1.85 0.96
1.86 0.96
1.87 0.97
1.88 0.96
1.89 0.97
1.9 0.99
1.91 0.97
1.92 0.97
1.93 0.96
1.94 0.94
1.95 0.97
1.96 0.93
1.97 0.92
1.98 0.95
1.99 0.96
2 0.92
};
\addplot [semithick, color10, opacity=0.75, dashed]
table {%
0.01 0.62
0.02 0.46
0.03 0.56
0.04 0.59
0.05 0.61
0.06 0.6
0.07 0.63
0.08 0.65
0.09 0.65
0.1 0.61
0.11 0.71
0.12 0.65
0.13 0.75
0.14 0.76
0.15 0.78
0.16 0.83
0.17 0.72
0.18 0.77
0.19 0.8
0.2 0.88
0.21 0.77
0.22 0.75
0.23 0.8
0.24 0.83
0.25 0.84
0.26 0.85
0.27 0.86
0.28 0.89
0.29 0.89
0.3 0.91
0.31 0.88
0.32 0.94
0.33 0.9
0.34 0.89
0.35 0.94
0.36 0.91
0.37 0.92
0.38 0.96
0.39 0.96
0.4 0.94
0.41 0.93
0.42 0.97
0.43 0.96
0.44 0.95
0.45 0.95
0.46 0.93
0.47 0.95
0.48 0.97
0.49 0.94
0.5 0.96
0.51 0.98
0.52 0.96
0.53 0.98
0.54 0.98
0.55 1
0.56 0.97
0.57 0.99
0.58 0.98
0.59 0.99
0.6 1
0.61 0.99
0.62 1
0.63 0.98
0.64 0.97
0.65 0.98
0.66 0.98
0.67 1
0.68 0.98
0.69 1
0.7 1
0.71 0.99
0.72 0.98
0.73 1
0.74 1
0.75 0.97
0.76 1
0.77 0.99
0.78 0.99
0.79 1
0.8 1
0.81 1
0.82 0.99
0.83 0.98
0.84 0.99
0.85 0.99
0.86 0.99
0.87 1
0.88 1
0.89 1
0.9 0.99
0.91 1
0.92 1
0.93 1
0.94 0.99
0.95 0.99
0.96 0.99
0.97 0.99
0.98 1
0.99 0.99
1 1
1.02 0.98
1.03 0.98
1.04 0.98
1.05 0.96
1.06 0.99
1.07 0.98
1.08 0.99
1.09 1
1.1 0.99
1.11 0.98
1.12 0.95
1.13 0.97
1.14 0.96
1.15 1
1.16 0.92
1.17 0.96
1.18 0.98
1.19 0.98
1.2 0.98
1.21 0.99
1.22 0.99
1.23 0.98
1.24 1
1.25 0.98
1.26 0.97
1.27 0.99
1.28 1
1.29 0.98
1.3 1
1.31 1
1.32 1
1.33 0.99
1.34 0.99
1.35 1
1.36 1
1.37 1
1.38 1
1.39 1
1.4 0.99
1.41 1
1.42 1
1.43 1
1.44 1
1.45 1
1.46 0.97
1.47 1
1.48 1
1.49 0.99
1.5 1
1.51 0.99
1.52 1
1.53 0.98
1.54 1
1.55 1
1.56 1
1.57 1
1.58 0.98
1.59 0.98
1.6 0.98
1.61 1
1.62 0.98
1.63 0.99
1.64 0.98
1.65 0.98
1.66 1
1.67 0.99
1.68 0.99
1.69 0.99
1.7 0.97
1.71 1
1.72 0.98
1.73 1
1.74 0.99
1.75 1
1.76 0.97
1.77 0.98
1.78 0.99
1.79 0.98
1.8 0.99
1.81 0.98
1.82 0.98
1.83 0.98
1.84 0.97
1.85 0.99
1.86 0.94
1.87 0.93
1.88 0.95
1.89 0.96
1.9 0.96
1.91 0.96
1.92 0.98
1.93 0.96
1.94 0.97
1.95 0.98
1.96 0.96
1.97 0.97
1.98 0.9
1.99 0.94
2 0.95
};
\addplot [semithick, color11, opacity=0.75, dashed]
table {%
0.01 1
0.02 1
0.03 1
0.04 1
0.05 1
0.06 1
0.07 1
0.08 1
0.09 1
0.1 1
0.11 1
0.12 1
0.13 1
0.14 1
0.15 1
0.16 1
0.17 1
0.18 1
0.19 1
0.2 1
0.21 1
0.22 1
0.23 1
0.24 1
0.25 1
0.26 1
0.27 1
0.28 1
0.29 1
0.3 1
0.31 1
0.32 1
0.33 1
0.34 1
0.35 1
0.36 1
0.37 1
0.38 1
0.39 1
0.4 1
0.41 1
0.42 1
0.43 1
0.44 1
0.45 1
0.46 1
0.47 1
0.48 1
0.49 1
0.5 1
0.51 1
0.52 1
0.53 1
0.54 1
0.55 1
0.56 1
0.57 1
0.58 1
0.59 1
0.6 1
0.61 1
0.62 1
0.63 1
0.64 1
0.65 1
0.66 1
0.67 1
0.68 1
0.69 1
0.7 1
0.71 1
0.72 1
0.73 1
0.74 1
0.75 1
0.76 1
0.77 1
0.78 1
0.79 1
0.8 1
0.81 1
0.82 1
0.83 1
0.84 1
0.85 1
0.86 1
0.87 1
0.88 1
0.89 1
0.9 1
0.91 1
0.92 1
0.93 1
0.94 1
0.95 1
0.96 1
0.97 1
0.98 1
0.99 1
1 1
1.02 1
1.03 1
1.04 1
1.05 1
1.06 1
1.07 1
1.08 1
1.09 1
1.1 1
1.11 1
1.12 1
1.13 1
1.14 1
1.15 1
1.16 1
1.17 1
1.18 1
1.19 1
1.2 1
1.21 1
1.22 1
1.23 1
1.24 1
1.25 1
1.26 1
1.27 1
1.28 1
1.29 1
1.3 1
1.31 1
1.32 1
1.33 1
1.34 1
1.35 1
1.36 1
1.37 1
1.38 1
1.39 1
1.4 1
1.41 1
1.42 1
1.43 1
1.44 1
1.45 1
1.46 1
1.47 1
1.48 1
1.49 1
1.5 1
1.51 1
1.52 1
1.53 1
1.54 1
1.55 1
1.56 1
1.57 1
1.58 1
1.59 1
1.6 1
1.61 1
1.62 1
1.63 1
1.64 1
1.65 1
1.66 1
1.67 1
1.68 1
1.69 1
1.7 1
1.71 1
1.72 1
1.73 1
1.74 1
1.75 1
1.76 1
1.77 1
1.78 1
1.79 1
1.8 1
1.81 1
1.82 1
1.83 1
1.84 1
1.85 1
1.86 1
1.87 1
1.88 1
1.89 1
1.9 1
1.91 1
1.92 1
1.93 1
1.94 1
1.95 1
1.96 1
1.97 1
1.98 1
1.99 1
2 1
};
\addplot [semithick, black, opacity=1, dash pattern=on 1pt off 1pt]
table {%
-0.0895000000000001 0.5
2.0995 0.5
};
\addplot [semithick, black, opacity=1, dash pattern=on 1pt off 1pt]
table {%
-0.0895000000000001 1
2.0995 1
};
\addplot [semithick, black, opacity=1, dash pattern=on 1pt off 1pt]
table {%
1 0.2
1 1.05
};
\end{axis}

\end{tikzpicture}

%% file: plots/decoupled2/NL_LAPxUNI.tex
% This file was created by tikzplotlib v0.9.6.
\begin{tikzpicture}

\definecolor{color0}{rgb}{0.866666666666667,0.494117647058824,0.164705882352941}
\definecolor{color1}{rgb}{0.164705882352941,0.643137254901961,0.866666666666667}
\definecolor{color2}{rgb}{0.584313725490196,0.866666666666667,0.164705882352941}
\definecolor{color3}{rgb}{0.109803921568627,0.337254901960784,0.129411764705882}
\definecolor{color4}{rgb}{0.529411764705882,0.305882352941176,0.858823529411765}
\definecolor{color5}{rgb}{0.858823529411765,0.305882352941176,0.435294117647059}
\definecolor{color6}{rgb}{0.937254901960784,0.929411764705882,0.392156862745098}
\definecolor{color7}{rgb}{0.0901960784313725,0.486274509803922,0.0980392156862745}
\definecolor{color8}{rgb}{0.156862745098039,0.188235294117647,0.827450980392157}
\definecolor{color9}{rgb}{0.937254901960784,0.392156862745098,0.894117647058824}
\definecolor{color10}{rgb}{0.2,0.184313725490196,0.184313725490196}
\definecolor{color11}{rgb}{0.0156862745098039,0.803921568627451,0.976470588235294}

\begin{axis}[
tick align=outside,
tick pos=left,
x grid style={white!69.0196078431373!black},
xmajorgrids,
xmin=-0.0895, xmax=2.0995,
xtick style={color=black},
xtick={0,0.1,0.2,0.3,0.4,0.5,0.6,0.7,0.8,0.9,1,1.1,1.2,1.3,1.4,1.5,1.6,1.7,1.8,1.9,2},
xticklabels={0,,.2,,.4,,.6,,.8,,1,,20,,40,,60,,80,,100},
height=4.8cm,
width=6.5cm,
y grid style={white!69.0196078431373!black},
ymajorgrids,
ymin=0.2, ymax=1.05,
ytick style={color=black}
]
\addplot [semithick, color0, opacity=0.75]
table {%
0.01 0.55
0.02 0.61
0.03 0.7
0.04 0.88
0.05 0.81
0.06 0.88
0.07 0.87
0.08 0.91
0.09 0.98
0.1 0.92
0.11 0.97
0.12 0.98
0.13 0.98
0.14 0.99
0.15 1
0.16 1
0.17 0.99
0.18 0.99
0.19 1
0.2 1
0.21 1
0.22 1
0.23 0.99
0.24 1
0.25 1
0.26 1
0.27 1
0.28 1
0.29 1
0.3 1
0.31 1
0.32 1
0.33 1
0.34 1
0.35 1
0.36 1
0.37 1
0.38 1
0.39 1
0.4 1
0.41 1
0.42 1
0.43 1
0.44 1
0.45 1
0.46 1
0.47 1
0.48 1
0.49 1
0.5 1
0.51 1
0.52 1
0.53 1
0.54 1
0.55 1
0.56 1
0.57 1
0.58 1
0.59 1
0.6 1
0.61 1
0.62 1
0.63 1
0.64 1
0.65 1
0.66 1
0.67 1
0.68 1
0.69 1
0.7 1
0.71 1
0.72 1
0.73 1
0.74 1
0.75 1
0.76 1
0.77 1
0.78 1
0.79 1
0.8 1
0.81 1
0.82 1
0.83 1
0.84 1
0.85 1
0.86 1
0.87 1
0.88 1
0.89 1
0.9 1
0.91 1
0.92 1
0.93 1
0.94 1
0.95 1
0.96 1
0.97 1
0.98 1
0.99 1
1 1
1.02 1
1.03 1
1.04 1
1.05 1
1.06 1
1.07 1
1.08 1
1.09 1
1.1 1
1.11 1
1.12 1
1.13 1
1.14 1
1.15 1
1.16 1
1.17 1
1.18 1
1.19 1
1.2 1
1.21 1
1.22 1
1.23 1
1.24 1
1.25 1
1.26 1
1.27 1
1.28 1
1.29 1
1.3 1
1.31 1
1.32 1
1.33 1
1.34 1
1.35 1
1.36 1
1.37 1
1.38 1
1.39 1
1.4 1
1.41 1
1.42 1
1.43 1
1.44 1
1.45 1
1.46 1
1.47 1
1.48 1
1.49 1
1.5 1
1.51 1
1.52 1
1.53 1
1.54 1
1.55 1
1.56 1
1.57 1
1.58 1
1.59 1
1.6 1
1.61 1
1.62 1
1.63 1
1.64 1
1.65 1
1.66 1
1.67 1
1.68 1
1.69 1
1.7 1
1.71 1
1.72 1
1.73 1
1.74 1
1.75 1
1.76 1
1.77 1
1.78 1
1.79 1
1.8 1
1.81 1
1.82 1
1.83 1
1.84 1
1.85 1
1.86 1
1.87 1
1.88 1
1.89 1
1.9 1
1.91 1
1.92 1
1.93 1
1.94 0.99
1.95 1
1.96 1
1.97 1
1.98 0.99
1.99 1
2 1
};
\addplot [semithick, color1, opacity=0.75]
table {%
0.01 0.48
0.02 0.54
0.03 0.6
0.04 0.75
0.05 0.67
0.06 0.77
0.07 0.78
0.08 0.86
0.09 0.86
0.1 0.89
0.11 0.9
0.12 0.89
0.13 0.89
0.14 0.93
0.15 0.94
0.16 0.95
0.17 0.98
0.18 0.94
0.19 0.98
0.2 0.98
0.21 1
0.22 1
0.23 0.99
0.24 1
0.25 1
0.26 1
0.27 1
0.28 1
0.29 1
0.3 1
0.31 1
0.32 1
0.33 1
0.34 1
0.35 1
0.36 1
0.37 1
0.38 1
0.39 1
0.4 1
0.41 1
0.42 1
0.43 1
0.44 1
0.45 1
0.46 1
0.47 1
0.48 1
0.49 1
0.5 1
0.51 1
0.52 1
0.53 1
0.54 1
0.55 1
0.56 1
0.57 1
0.58 1
0.59 1
0.6 1
0.61 1
0.62 1
0.63 1
0.64 1
0.65 1
0.66 1
0.67 1
0.68 1
0.69 1
0.7 1
0.71 1
0.72 1
0.73 1
0.74 1
0.75 1
0.76 1
0.77 1
0.78 1
0.79 1
0.8 1
0.81 1
0.82 1
0.83 1
0.84 1
0.85 1
0.86 1
0.87 1
0.88 1
0.89 1
0.9 1
0.91 1
0.92 1
0.93 1
0.94 1
0.95 1
0.96 1
0.97 1
0.98 1
0.99 1
1 1
1.02 1
1.03 1
1.04 1
1.05 1
1.06 1
1.07 1
1.08 1
1.09 1
1.1 1
1.11 1
1.12 1
1.13 1
1.14 1
1.15 1
1.16 1
1.17 1
1.18 1
1.19 1
1.2 1
1.21 1
1.22 1
1.23 1
1.24 1
1.25 1
1.26 1
1.27 1
1.28 1
1.29 1
1.3 1
1.31 1
1.32 1
1.33 1
1.34 1
1.35 1
1.36 1
1.37 1
1.38 1
1.39 1
1.4 1
1.41 1
1.42 1
1.43 1
1.44 1
1.45 1
1.46 1
1.47 1
1.48 1
1.49 1
1.5 1
1.51 1
1.52 1
1.53 1
1.54 1
1.55 1
1.56 1
1.57 1
1.58 1
1.59 1
1.6 1
1.61 1
1.62 1
1.63 1
1.64 1
1.65 1
1.66 1
1.67 1
1.68 1
1.69 1
1.7 1
1.71 1
1.72 1
1.73 1
1.74 1
1.75 1
1.76 1
1.77 1
1.78 1
1.79 1
1.8 1
1.81 1
1.82 1
1.83 1
1.84 1
1.85 1
1.86 1
1.87 1
1.88 1
1.89 1
1.9 1
1.91 1
1.92 1
1.93 1
1.94 0.98
1.95 1
1.96 1
1.97 1
1.98 1
1.99 1
2 1
};
\addplot [semithick, color2, opacity=0.75]
table {%
0.01 0.49
0.02 0.54
0.03 0.52
0.04 0.71
0.05 0.65
0.06 0.67
0.07 0.69
0.08 0.74
0.09 0.65
0.1 0.84
0.11 0.8
0.12 0.81
0.13 0.79
0.14 0.88
0.15 0.88
0.16 0.91
0.17 0.94
0.18 0.93
0.19 0.98
0.2 0.98
0.21 1
0.22 1
0.23 0.99
0.24 0.99
0.25 0.99
0.26 1
0.27 1
0.28 1
0.29 0.99
0.3 0.99
0.31 1
0.32 1
0.33 1
0.34 1
0.35 1
0.36 1
0.37 1
0.38 1
0.39 1
0.4 1
0.41 1
0.42 1
0.43 1
0.44 1
0.45 1
0.46 1
0.47 1
0.48 1
0.49 1
0.5 1
0.51 1
0.52 1
0.53 1
0.54 1
0.55 1
0.56 1
0.57 1
0.58 1
0.59 1
0.6 1
0.61 1
0.62 1
0.63 1
0.64 1
0.65 1
0.66 1
0.67 1
0.68 1
0.69 1
0.7 1
0.71 1
0.72 1
0.73 1
0.74 1
0.75 1
0.76 1
0.77 1
0.78 1
0.79 1
0.8 1
0.81 1
0.82 1
0.83 1
0.84 1
0.85 1
0.86 1
0.87 1
0.88 1
0.89 1
0.9 1
0.91 1
0.92 1
0.93 1
0.94 1
0.95 1
0.96 1
0.97 1
0.98 1
0.99 1
1 1
1.02 1
1.03 1
1.04 1
1.05 1
1.06 1
1.07 1
1.08 1
1.09 1
1.1 1
1.11 1
1.12 1
1.13 1
1.14 1
1.15 1
1.16 1
1.17 1
1.18 1
1.19 1
1.2 1
1.21 1
1.22 1
1.23 1
1.24 1
1.25 1
1.26 1
1.27 1
1.28 1
1.29 1
1.3 1
1.31 1
1.32 1
1.33 1
1.34 1
1.35 1
1.36 1
1.37 1
1.38 1
1.39 1
1.4 1
1.41 1
1.42 1
1.43 1
1.44 1
1.45 1
1.46 1
1.47 1
1.48 1
1.49 1
1.5 1
1.51 1
1.52 1
1.53 1
1.54 1
1.55 1
1.56 1
1.57 1
1.58 1
1.59 1
1.6 1
1.61 1
1.62 1
1.63 1
1.64 1
1.65 1
1.66 1
1.67 1
1.68 1
1.69 1
1.7 1
1.71 1
1.72 1
1.73 1
1.74 1
1.75 1
1.76 1
1.77 1
1.78 1
1.79 1
1.8 1
1.81 1
1.82 1
1.83 1
1.84 1
1.85 1
1.86 1
1.87 1
1.88 1
1.89 1
1.9 1
1.91 1
1.92 1
1.93 1
1.94 0.98
1.95 1
1.96 1
1.97 1
1.98 1
1.99 1
2 1
};
\addplot [semithick, color3, opacity=0.75]
table {%
0.01 0.67
0.02 0.73
0.03 0.86
0.04 0.91
0.05 0.9
0.06 0.94
0.07 0.94
0.08 0.97
0.09 0.97
0.1 0.94
0.11 0.98
0.12 0.97
0.13 0.98
0.14 0.99
0.15 0.99
0.16 1
0.17 0.98
0.18 0.99
0.19 0.99
0.2 1
0.21 1
0.22 0.99
0.23 0.98
0.24 1
0.25 0.99
0.26 0.99
0.27 1
0.28 1
0.29 1
0.3 0.99
0.31 0.99
0.32 1
0.33 1
0.34 0.99
0.35 0.98
0.36 1
0.37 1
0.38 1
0.39 1
0.4 1
0.41 1
0.42 1
0.43 1
0.44 0.99
0.45 1
0.46 1
0.47 0.99
0.48 1
0.49 1
0.5 1
0.51 1
0.52 1
0.53 1
0.54 0.99
0.55 1
0.56 1
0.57 1
0.58 1
0.59 1
0.6 1
0.61 1
0.62 1
0.63 1
0.64 1
0.65 0.99
0.66 1
0.67 1
0.68 1
0.69 1
0.7 1
0.71 0.99
0.72 1
0.73 0.99
0.74 1
0.75 1
0.76 1
0.77 1
0.78 1
0.79 1
0.8 1
0.81 1
0.82 1
0.83 1
0.84 1
0.85 1
0.86 1
0.87 1
0.88 0.99
0.89 1
0.9 1
0.91 1
0.92 1
0.93 1
0.94 1
0.95 1
0.96 1
0.97 1
0.98 1
0.99 0.99
1 1
1.02 1
1.03 1
1.04 1
1.05 1
1.06 1
1.07 1
1.08 1
1.09 1
1.1 1
1.11 1
1.12 1
1.13 1
1.14 1
1.15 1
1.16 1
1.17 1
1.18 1
1.19 1
1.2 1
1.21 1
1.22 1
1.23 1
1.24 1
1.25 1
1.26 1
1.27 1
1.28 1
1.29 1
1.3 1
1.31 1
1.32 1
1.33 1
1.34 1
1.35 1
1.36 1
1.37 1
1.38 1
1.39 1
1.4 1
1.41 1
1.42 1
1.43 1
1.44 1
1.45 1
1.46 1
1.47 1
1.48 1
1.49 1
1.5 1
1.51 1
1.52 1
1.53 1
1.54 1
1.55 1
1.56 1
1.57 1
1.58 1
1.59 1
1.6 1
1.61 1
1.62 1
1.63 1
1.64 1
1.65 1
1.66 1
1.67 1
1.68 1
1.69 1
1.7 1
1.71 1
1.72 1
1.73 1
1.74 1
1.75 1
1.76 1
1.77 1
1.78 1
1.79 1
1.8 1
1.81 1
1.82 1
1.83 1
1.84 1
1.85 1
1.86 1
1.87 1
1.88 1
1.89 1
1.9 1
1.91 1
1.92 1
1.93 1
1.94 1
1.95 1
1.96 1
1.97 1
1.98 0.99
1.99 1
2 1
};
\addplot [semithick, color4, opacity=0.75]
table {%
0.01 0.67
0.02 0.73
0.03 0.86
0.04 0.91
0.05 0.9
0.06 0.94
0.07 0.94
0.08 0.97
0.09 0.97
0.1 0.95
0.11 0.98
0.12 0.98
0.13 0.98
0.14 0.99
0.15 0.99
0.16 1
0.17 0.98
0.18 0.99
0.19 0.99
0.2 1
0.21 1
0.22 0.99
0.23 0.98
0.24 1
0.25 0.99
0.26 0.99
0.27 1
0.28 1
0.29 1
0.3 0.99
0.31 0.99
0.32 1
0.33 1
0.34 0.99
0.35 0.99
0.36 1
0.37 1
0.38 1
0.39 1
0.4 1
0.41 1
0.42 1
0.43 1
0.44 0.99
0.45 1
0.46 1
0.47 0.99
0.48 1
0.49 1
0.5 1
0.51 1
0.52 1
0.53 1
0.54 0.99
0.55 1
0.56 1
0.57 1
0.58 1
0.59 1
0.6 1
0.61 1
0.62 1
0.63 1
0.64 1
0.65 0.99
0.66 1
0.67 1
0.68 1
0.69 1
0.7 1
0.71 0.99
0.72 1
0.73 0.99
0.74 1
0.75 1
0.76 1
0.77 1
0.78 1
0.79 1
0.8 1
0.81 1
0.82 1
0.83 1
0.84 1
0.85 1
0.86 1
0.87 1
0.88 0.99
0.89 1
0.9 1
0.91 1
0.92 1
0.93 1
0.94 1
0.95 1
0.96 1
0.97 1
0.98 1
0.99 0.99
1 1
1.02 1
1.03 1
1.04 1
1.05 1
1.06 1
1.07 1
1.08 1
1.09 1
1.1 1
1.11 1
1.12 1
1.13 1
1.14 1
1.15 1
1.16 1
1.17 1
1.18 1
1.19 1
1.2 1
1.21 1
1.22 1
1.23 1
1.24 1
1.25 1
1.26 1
1.27 1
1.28 1
1.29 1
1.3 1
1.31 1
1.32 1
1.33 1
1.34 1
1.35 1
1.36 1
1.37 1
1.38 1
1.39 1
1.4 1
1.41 1
1.42 1
1.43 1
1.44 1
1.45 1
1.46 1
1.47 1
1.48 1
1.49 1
1.5 1
1.51 1
1.52 1
1.53 1
1.54 1
1.55 1
1.56 1
1.57 1
1.58 1
1.59 1
1.6 1
1.61 1
1.62 1
1.63 1
1.64 1
1.65 1
1.66 1
1.67 1
1.68 1
1.69 1
1.7 1
1.71 1
1.72 1
1.73 1
1.74 1
1.75 1
1.76 1
1.77 1
1.78 1
1.79 1
1.8 1
1.81 1
1.82 1
1.83 1
1.84 1
1.85 1
1.86 1
1.87 1
1.88 1
1.89 1
1.9 1
1.91 1
1.92 1
1.93 1
1.94 1
1.95 1
1.96 1
1.97 1
1.98 0.99
1.99 1
2 1
};
\addplot [semithick, color5, opacity=0.75]
table {%
0.01 0.87
0.02 0.87
0.03 0.97
0.04 0.97
0.05 0.96
0.06 0.99
0.07 1
0.08 1
0.09 0.99
0.1 0.98
0.11 1
0.12 1
0.13 0.99
0.14 1
0.15 1
0.16 1
0.17 1
0.18 1
0.19 1
0.2 1
0.21 1
0.22 1
0.23 1
0.24 1
0.25 1
0.26 1
0.27 1
0.28 1
0.29 1
0.3 1
0.31 1
0.32 1
0.33 1
0.34 1
0.35 1
0.36 1
0.37 1
0.38 1
0.39 1
0.4 1
0.41 1
0.42 1
0.43 1
0.44 1
0.45 1
0.46 1
0.47 1
0.48 1
0.49 1
0.5 1
0.51 1
0.52 1
0.53 1
0.54 1
0.55 1
0.56 1
0.57 1
0.58 1
0.59 1
0.6 1
0.61 1
0.62 1
0.63 1
0.64 1
0.65 1
0.66 1
0.67 1
0.68 1
0.69 1
0.7 1
0.71 1
0.72 1
0.73 1
0.74 1
0.75 1
0.76 1
0.77 1
0.78 1
0.79 1
0.8 1
0.81 1
0.82 1
0.83 1
0.84 1
0.85 1
0.86 1
0.87 1
0.88 1
0.89 1
0.9 1
0.91 1
0.92 1
0.93 1
0.94 1
0.95 1
0.96 1
0.97 1
0.98 1
0.99 1
1 1
1.02 1
1.03 1
1.04 1
1.05 1
1.06 1
1.07 1
1.08 1
1.09 1
1.1 1
1.11 1
1.12 1
1.13 1
1.14 1
1.15 1
1.16 1
1.17 1
1.18 1
1.19 1
1.2 1
1.21 1
1.22 1
1.23 1
1.24 1
1.25 1
1.26 1
1.27 1
1.28 1
1.29 1
1.3 1
1.31 1
1.32 1
1.33 1
1.34 1
1.35 1
1.36 1
1.37 1
1.38 1
1.39 1
1.4 1
1.41 1
1.42 1
1.43 1
1.44 1
1.45 1
1.46 1
1.47 1
1.48 1
1.49 1
1.5 1
1.51 1
1.52 1
1.53 1
1.54 1
1.55 1
1.56 1
1.57 1
1.58 1
1.59 1
1.6 1
1.61 1
1.62 1
1.63 1
1.64 1
1.65 1
1.66 1
1.67 1
1.68 1
1.69 1
1.7 1
1.71 1
1.72 1
1.73 1
1.74 1
1.75 1
1.76 1
1.77 1
1.78 1
1.79 1
1.8 1
1.81 1
1.82 1
1.83 1
1.84 1
1.85 1
1.86 1
1.87 1
1.88 1
1.89 1
1.9 1
1.91 1
1.92 1
1.93 1
1.94 1
1.95 1
1.96 1
1.97 1
1.98 1
1.99 1
2 1
};
\addplot [semithick, color6, opacity=0.75, dashed]
table {%
0.01 1
0.02 1
0.03 1
0.04 1
0.05 1
0.06 1
0.07 1
0.08 1
0.09 1
0.1 1
0.11 1
0.12 1
0.13 1
0.14 1
0.15 1
0.16 1
0.17 1
0.18 1
0.19 1
0.2 1
0.21 1
0.22 1
0.23 1
0.24 1
0.25 1
0.26 1
0.27 1
0.28 1
0.29 1
0.3 1
0.31 1
0.32 1
0.33 1
0.34 1
0.35 1
0.36 1
0.37 1
0.38 1
0.39 1
0.4 1
0.41 1
0.42 1
0.43 1
0.44 1
0.45 1
0.46 1
0.47 1
0.48 1
0.49 1
0.5 1
0.51 1
0.52 1
0.53 1
0.54 1
0.55 1
0.56 1
0.57 1
0.58 1
0.59 1
0.6 1
0.61 1
0.62 1
0.63 1
0.64 1
0.65 1
0.66 1
0.67 1
0.68 1
0.69 1
0.7 1
0.71 1
0.72 1
0.73 1
0.74 1
0.75 1
0.76 1
0.77 1
0.78 1
0.79 1
0.8 1
0.81 1
0.82 1
0.83 1
0.84 1
0.85 1
0.86 1
0.87 1
0.88 1
0.89 1
0.9 1
0.91 1
0.92 1
0.93 1
0.94 1
0.95 1
0.96 1
0.97 1
0.98 1
0.99 1
1 1
1.02 1
1.03 1
1.04 1
1.05 1
1.06 1
1.07 1
1.08 1
1.09 1
1.1 1
1.11 1
1.12 1
1.13 1
1.14 1
1.15 1
1.16 1
1.17 1
1.18 1
1.19 1
1.2 1
1.21 1
1.22 1
1.23 1
1.24 1
1.25 1
1.26 1
1.27 1
1.28 1
1.29 1
1.3 1
1.31 1
1.32 1
1.33 1
1.34 1
1.35 1
1.36 1
1.37 1
1.38 1
1.39 1
1.4 1
1.41 1
1.42 1
1.43 1
1.44 1
1.45 1
1.46 1
1.47 1
1.48 1
1.49 1
1.5 1
1.51 1
1.52 1
1.53 1
1.54 1
1.55 1
1.56 1
1.57 1
1.58 1
1.59 1
1.6 1
1.61 1
1.62 1
1.63 1
1.64 1
1.65 1
1.66 1
1.67 1
1.68 1
1.69 1
1.7 1
1.71 1
1.72 1
1.73 1
1.74 1
1.75 1
1.76 1
1.77 1
1.78 1
1.79 1
1.8 1
1.81 1
1.82 1
1.83 1
1.84 1
1.85 1
1.86 1
1.87 1
1.88 1
1.89 1
1.9 1
1.91 1
1.92 1
1.93 1
1.94 1
1.95 1
1.96 1
1.97 1
1.98 1
1.99 1
2 1
};
\addplot [semithick, color7, opacity=0.75, dashed]
table {%
0.01 1
0.02 1
0.03 1
0.04 1
0.05 1
0.06 1
0.07 1
0.08 1
0.09 1
0.1 1
0.11 1
0.12 1
0.13 1
0.14 1
0.15 1
0.16 1
0.17 1
0.18 1
0.19 1
0.2 1
0.21 1
0.22 1
0.23 1
0.24 1
0.25 1
0.26 1
0.27 1
0.28 1
0.29 1
0.3 1
0.31 1
0.32 1
0.33 1
0.34 1
0.35 1
0.36 1
0.37 1
0.38 1
0.39 1
0.4 1
0.41 1
0.42 1
0.43 1
0.44 1
0.45 1
0.46 1
0.47 1
0.48 1
0.49 1
0.5 1
0.51 1
0.52 1
0.53 1
0.54 1
0.55 1
0.56 1
0.57 1
0.58 1
0.59 1
0.6 1
0.61 1
0.62 1
0.63 1
0.64 1
0.65 1
0.66 1
0.67 1
0.68 1
0.69 1
0.7 1
0.71 1
0.72 1
0.73 1
0.74 1
0.75 1
0.76 1
0.77 1
0.78 1
0.79 1
0.8 1
0.81 1
0.82 1
0.83 1
0.84 1
0.85 1
0.86 1
0.87 1
0.88 1
0.89 1
0.9 1
0.91 1
0.92 1
0.93 1
0.94 1
0.95 1
0.96 1
0.97 1
0.98 1
0.99 1
1 1
1.02 1
1.03 1
1.04 1
1.05 1
1.06 1
1.07 1
1.08 1
1.09 1
1.1 1
1.11 1
1.12 1
1.13 1
1.14 1
1.15 1
1.16 1
1.17 1
1.18 1
1.19 1
1.2 1
1.21 1
1.22 1
1.23 1
1.24 1
1.25 1
1.26 1
1.27 1
1.28 1
1.29 1
1.3 1
1.31 1
1.32 1
1.33 1
1.34 1
1.35 1
1.36 1
1.37 1
1.38 1
1.39 1
1.4 1
1.41 1
1.42 1
1.43 1
1.44 1
1.45 1
1.46 1
1.47 1
1.48 1
1.49 1
1.5 1
1.51 1
1.52 1
1.53 1
1.54 1
1.55 1
1.56 1
1.57 1
1.58 1
1.59 1
1.6 1
1.61 1
1.62 1
1.63 1
1.64 1
1.65 1
1.66 1
1.67 1
1.68 1
1.69 1
1.7 1
1.71 1
1.72 1
1.73 1
1.74 1
1.75 1
1.76 1
1.77 1
1.78 1
1.79 1
1.8 1
1.81 1
1.82 1
1.83 1
1.84 1
1.85 1
1.86 1
1.87 1
1.88 1
1.89 1
1.9 1
1.91 1
1.92 1
1.93 1
1.94 1
1.95 1
1.96 1
1.97 1
1.98 1
1.99 1
2 1
};
\addplot [semithick, color8, opacity=0.75, dashed]
table {%
0.01 1
0.02 1
0.03 1
0.04 1
0.05 1
0.06 1
0.07 1
0.08 1
0.09 1
0.1 1
0.11 1
0.12 1
0.13 1
0.14 1
0.15 1
0.16 1
0.17 1
0.18 1
0.19 1
0.2 1
0.21 1
0.22 1
0.23 1
0.24 1
0.25 1
0.26 1
0.27 1
0.28 1
0.29 1
0.3 1
0.31 1
0.32 1
0.33 1
0.34 1
0.35 1
0.36 1
0.37 1
0.38 1
0.39 1
0.4 1
0.41 1
0.42 1
0.43 1
0.44 1
0.45 1
0.46 1
0.47 1
0.48 1
0.49 1
0.5 1
0.51 1
0.52 1
0.53 1
0.54 1
0.55 1
0.56 1
0.57 1
0.58 1
0.59 1
0.6 1
0.61 1
0.62 1
0.63 1
0.64 1
0.65 1
0.66 1
0.67 1
0.68 1
0.69 1
0.7 1
0.71 1
0.72 1
0.73 1
0.74 1
0.75 1
0.76 1
0.77 1
0.78 1
0.79 1
0.8 1
0.81 1
0.82 1
0.83 1
0.84 1
0.85 1
0.86 1
0.87 1
0.88 1
0.89 1
0.9 1
0.91 1
0.92 1
0.93 1
0.94 1
0.95 1
0.96 1
0.97 1
0.98 1
0.99 1
1 1
1.02 1
1.03 1
1.04 1
1.05 1
1.06 1
1.07 1
1.08 1
1.09 1
1.1 1
1.11 1
1.12 1
1.13 1
1.14 1
1.15 1
1.16 1
1.17 1
1.18 1
1.19 1
1.2 1
1.21 1
1.22 1
1.23 1
1.24 1
1.25 1
1.26 1
1.27 1
1.28 1
1.29 1
1.3 1
1.31 1
1.32 1
1.33 1
1.34 1
1.35 1
1.36 1
1.37 1
1.38 1
1.39 1
1.4 1
1.41 1
1.42 1
1.43 1
1.44 1
1.45 1
1.46 1
1.47 1
1.48 1
1.49 1
1.5 1
1.51 1
1.52 1
1.53 1
1.54 1
1.55 1
1.56 1
1.57 1
1.58 1
1.59 1
1.6 1
1.61 1
1.62 1
1.63 1
1.64 1
1.65 1
1.66 1
1.67 1
1.68 1
1.69 1
1.7 1
1.71 1
1.72 1
1.73 1
1.74 1
1.75 1
1.76 1
1.77 1
1.78 1
1.79 1
1.8 1
1.81 1
1.82 1
1.83 1
1.84 1
1.85 1
1.86 1
1.87 1
1.88 1
1.89 1
1.9 1
1.91 1
1.92 1
1.93 1
1.94 1
1.95 1
1.96 1
1.97 1
1.98 1
1.99 1
2 1
};
\addplot [semithick, color9, opacity=0.75, dashed]
table {%
0.01 0.49
0.02 0.48
0.03 0.53
0.04 0.56
0.05 0.62
0.06 0.51
0.07 0.63
0.08 0.62
0.09 0.66
0.1 0.61
0.11 0.7
0.12 0.69
0.13 0.66
0.14 0.63
0.15 0.57
0.16 0.65
0.17 0.64
0.18 0.67
0.19 0.67
0.2 0.72
0.21 0.79
0.22 0.78
0.23 0.72
0.24 0.78
0.25 0.74
0.26 0.67
0.27 0.76
0.28 0.68
0.29 0.73
0.3 0.71
0.31 0.74
0.32 0.8
0.33 0.75
0.34 0.75
0.35 0.75
0.36 0.81
0.37 0.8
0.38 0.8
0.39 0.77
0.4 0.83
0.41 0.89
0.42 0.84
0.43 0.79
0.44 0.83
0.45 0.73
0.46 0.82
0.47 0.84
0.48 0.8
0.49 0.85
0.5 0.85
0.51 0.88
0.52 0.88
0.53 0.87
0.54 0.9
0.55 0.87
0.56 0.81
0.57 0.87
0.58 0.95
0.59 0.82
0.6 0.87
0.61 0.9
0.62 0.89
0.63 0.89
0.64 0.91
0.65 0.91
0.66 0.85
0.67 0.88
0.68 0.89
0.69 0.9
0.7 0.89
0.71 0.91
0.72 0.93
0.73 0.87
0.74 0.96
0.75 0.96
0.76 0.91
0.77 0.94
0.78 0.96
0.79 0.87
0.8 0.95
0.81 0.94
0.82 0.93
0.83 0.94
0.84 0.94
0.85 0.95
0.86 0.93
0.87 0.95
0.88 0.95
0.89 0.98
0.9 0.98
0.91 0.97
0.92 0.96
0.93 0.95
0.94 0.96
0.95 0.94
0.96 0.97
0.97 0.95
0.98 0.95
0.99 0.97
1 0.97
1.02 1
1.03 1
1.04 1
1.05 0.99
1.06 1
1.07 0.97
1.08 0.99
1.09 1
1.1 1
1.11 1
1.12 0.99
1.13 0.99
1.14 0.99
1.15 1
1.16 1
1.17 1
1.18 1
1.19 1
1.2 1
1.21 0.99
1.22 0.99
1.23 1
1.24 0.98
1.25 1
1.26 1
1.27 1
1.28 1
1.29 0.98
1.3 1
1.31 1
1.32 1
1.33 0.99
1.34 1
1.35 1
1.36 1
1.37 1
1.38 1
1.39 1
1.4 0.99
1.41 0.98
1.42 1
1.43 1
1.44 1
1.45 1
1.46 0.99
1.47 1
1.48 1
1.49 1
1.5 0.99
1.51 1
1.52 1
1.53 1
1.54 0.99
1.55 1
1.56 1
1.57 1
1.58 1
1.59 1
1.6 1
1.61 0.99
1.62 1
1.63 1
1.64 1
1.65 1
1.66 1
1.67 1
1.68 1
1.69 1
1.7 1
1.71 1
1.72 0.99
1.73 0.99
1.74 1
1.75 1
1.76 1
1.77 1
1.78 1
1.79 0.99
1.8 0.99
1.81 1
1.82 0.99
1.83 0.99
1.84 1
1.85 1
1.86 1
1.87 1
1.88 0.99
1.89 1
1.9 1
1.91 0.99
1.92 1
1.93 1
1.94 1
1.95 0.99
1.96 1
1.97 0.99
1.98 0.99
1.99 0.99
2 1
};
\addplot [semithick, color10, opacity=0.75, dashed]
table {%
0.01 0.47
0.02 0.46
0.03 0.47
0.04 0.59
0.05 0.55
0.06 0.57
0.07 0.61
0.08 0.51
0.09 0.55
0.1 0.65
0.11 0.68
0.12 0.66
0.13 0.59
0.14 0.69
0.15 0.61
0.16 0.68
0.17 0.68
0.18 0.69
0.19 0.57
0.2 0.69
0.21 0.71
0.22 0.76
0.23 0.72
0.24 0.7
0.25 0.82
0.26 0.83
0.27 0.81
0.28 0.72
0.29 0.69
0.3 0.75
0.31 0.74
0.32 0.83
0.33 0.84
0.34 0.78
0.35 0.8
0.36 0.86
0.37 0.85
0.38 0.86
0.39 0.81
0.4 0.83
0.41 0.86
0.42 0.89
0.43 0.91
0.44 0.88
0.45 0.87
0.46 0.88
0.47 0.84
0.48 0.88
0.49 0.87
0.5 0.91
0.51 0.88
0.52 0.94
0.53 0.92
0.54 0.97
0.55 0.89
0.56 0.95
0.57 0.89
0.58 0.92
0.59 0.93
0.6 0.91
0.61 0.95
0.62 0.92
0.63 0.95
0.64 0.94
0.65 0.89
0.66 0.98
0.67 0.93
0.68 0.93
0.69 0.94
0.7 0.93
0.71 0.96
0.72 0.97
0.73 0.91
0.74 0.93
0.75 0.98
0.76 0.98
0.77 0.94
0.78 0.98
0.79 0.96
0.8 0.97
0.81 0.97
0.82 0.99
0.83 0.98
0.84 0.97
0.85 0.98
0.86 0.95
0.87 0.99
0.88 0.96
0.89 0.98
0.9 0.99
0.91 0.97
0.92 0.97
0.93 0.99
0.94 0.98
0.95 0.99
0.96 0.99
0.97 0.99
0.98 0.95
0.99 0.98
1 0.98
1.02 1
1.03 1
1.04 0.99
1.05 0.99
1.06 0.98
1.07 0.96
1.08 0.97
1.09 0.98
1.1 0.97
1.11 0.99
1.12 0.98
1.13 0.95
1.14 0.97
1.15 0.98
1.16 0.97
1.17 0.99
1.18 1
1.19 0.98
1.2 1
1.21 0.97
1.22 0.97
1.23 1
1.24 0.97
1.25 1
1.26 1
1.27 0.99
1.28 1
1.29 0.98
1.3 0.98
1.31 1
1.32 0.99
1.33 0.99
1.34 1
1.35 1
1.36 1
1.37 1
1.38 1
1.39 0.99
1.4 0.99
1.41 0.98
1.42 1
1.43 1
1.44 1
1.45 1
1.46 0.99
1.47 1
1.48 1
1.49 1
1.5 0.99
1.51 1
1.52 1
1.53 1
1.54 0.99
1.55 1
1.56 1
1.57 1
1.58 1
1.59 1
1.6 0.99
1.61 1
1.62 1
1.63 1
1.64 1
1.65 1
1.66 1
1.67 1
1.68 1
1.69 1
1.7 1
1.71 1
1.72 0.99
1.73 1
1.74 1
1.75 1
1.76 1
1.77 1
1.78 1
1.79 1
1.8 1
1.81 1
1.82 1
1.83 1
1.84 1
1.85 1
1.86 1
1.87 0.99
1.88 1
1.89 0.99
1.9 1
1.91 1
1.92 0.99
1.93 0.98
1.94 1
1.95 0.99
1.96 1
1.97 1
1.98 0.99
1.99 1
2 1
};
\addplot [semithick, color11, opacity=0.75, dashed]
table {%
0.01 1
0.02 1
0.03 1
0.04 1
0.05 1
0.06 1
0.07 1
0.08 1
0.09 1
0.1 1
0.11 1
0.12 1
0.13 1
0.14 1
0.15 1
0.16 1
0.17 1
0.18 1
0.19 1
0.2 1
0.21 1
0.22 1
0.23 1
0.24 1
0.25 1
0.26 1
0.27 1
0.28 1
0.29 1
0.3 1
0.31 1
0.32 1
0.33 1
0.34 1
0.35 1
0.36 1
0.37 1
0.38 1
0.39 1
0.4 1
0.41 1
0.42 1
0.43 1
0.44 1
0.45 1
0.46 1
0.47 1
0.48 1
0.49 1
0.5 1
0.51 1
0.52 1
0.53 1
0.54 1
0.55 1
0.56 1
0.57 1
0.58 1
0.59 1
0.6 1
0.61 1
0.62 1
0.63 1
0.64 1
0.65 1
0.66 1
0.67 1
0.68 1
0.69 1
0.7 1
0.71 1
0.72 1
0.73 1
0.74 1
0.75 1
0.76 1
0.77 1
0.78 1
0.79 1
0.8 1
0.81 1
0.82 1
0.83 1
0.84 1
0.85 1
0.86 1
0.87 1
0.88 1
0.89 1
0.9 1
0.91 1
0.92 1
0.93 1
0.94 1
0.95 1
0.96 1
0.97 1
0.98 1
0.99 1
1 1
1.02 1
1.03 1
1.04 1
1.05 1
1.06 1
1.07 1
1.08 1
1.09 1
1.1 1
1.11 1
1.12 1
1.13 1
1.14 1
1.15 1
1.16 1
1.17 1
1.18 1
1.19 1
1.2 1
1.21 1
1.22 1
1.23 1
1.24 1
1.25 1
1.26 1
1.27 1
1.28 1
1.29 1
1.3 1
1.31 1
1.32 1
1.33 1
1.34 1
1.35 1
1.36 1
1.37 1
1.38 1
1.39 1
1.4 1
1.41 1
1.42 1
1.43 1
1.44 1
1.45 1
1.46 1
1.47 1
1.48 1
1.49 1
1.5 1
1.51 1
1.52 1
1.53 1
1.54 1
1.55 1
1.56 1
1.57 1
1.58 1
1.59 1
1.6 1
1.61 1
1.62 1
1.63 1
1.64 1
1.65 1
1.66 1
1.67 1
1.68 1
1.69 1
1.7 1
1.71 1
1.72 1
1.73 1
1.74 1
1.75 1
1.76 1
1.77 1
1.78 1
1.79 1
1.8 1
1.81 1
1.82 1
1.83 1
1.84 1
1.85 1
1.86 1
1.87 1
1.88 1
1.89 1
1.9 1
1.91 1
1.92 1
1.93 1
1.94 1
1.95 1
1.96 1
1.97 1
1.98 1
1.99 1
2 1
};
\addplot [semithick, black, opacity=1, dash pattern=on 1pt off 1pt]
table {%
-0.0895000000000001 0.5
2.0995 0.5
};
\addplot [semithick, black, opacity=1, dash pattern=on 1pt off 1pt]
table {%
-0.0895000000000001 1
2.0995 1
};
\addplot [semithick, black, opacity=1, dash pattern=on 1pt off 1pt]
table {%
1 0.2
1 1.05
};
\end{axis}

\end{tikzpicture}

%% file: plots/decoupled2/NLLAP.tex
% This file was created by tikzplotlib v0.9.6.
\begin{tikzpicture}

\definecolor{color0}{rgb}{0.866666666666667,0.494117647058824,0.164705882352941}
\definecolor{color1}{rgb}{0.164705882352941,0.643137254901961,0.866666666666667}
\definecolor{color2}{rgb}{0.584313725490196,0.866666666666667,0.164705882352941}
\definecolor{color3}{rgb}{0.109803921568627,0.337254901960784,0.129411764705882}
\definecolor{color4}{rgb}{0.529411764705882,0.305882352941176,0.858823529411765}
\definecolor{color5}{rgb}{0.858823529411765,0.305882352941176,0.435294117647059}
\definecolor{color6}{rgb}{0.937254901960784,0.929411764705882,0.392156862745098}
\definecolor{color7}{rgb}{0.0901960784313725,0.486274509803922,0.0980392156862745}
\definecolor{color8}{rgb}{0.156862745098039,0.188235294117647,0.827450980392157}
\definecolor{color9}{rgb}{0.937254901960784,0.392156862745098,0.894117647058824}
\definecolor{color10}{rgb}{0.2,0.184313725490196,0.184313725490196}
\definecolor{color11}{rgb}{0.0156862745098039,0.803921568627451,0.976470588235294}

\begin{axis}[
tick align=outside,
tick pos=left,
x grid style={white!69.0196078431373!black},
xmajorgrids,
xmin=-0.0895, xmax=2.0995,
xtick style={color=black},
xtick={0,0.1,0.2,0.3,0.4,0.5,0.6,0.7,0.8,0.9,1,1.1,1.2,1.3,1.4,1.5,1.6,1.7,1.8,1.9,2},
xticklabels={0,,.2,,.4,,.6,,.8,,1,,20,,40,,60,,80,,100},
height=4.8cm,
width=6.5cm,
y grid style={white!69.0196078431373!black},
ymajorgrids,
ymin=0.2, ymax=1.05,
ytick style={color=black}
]
\addplot [semithick, color0, opacity=0.75]
table {%
0.01 0.6
0.02 0.69
0.03 0.8
0.04 0.84
0.05 0.94
0.06 0.93
0.07 0.94
0.08 1
0.09 1
0.1 0.99
0.11 0.99
0.12 0.99
0.13 0.99
0.14 1
0.15 1
0.16 1
0.17 1
0.18 1
0.19 1
0.2 1
0.21 1
0.22 1
0.23 1
0.24 1
0.25 1
0.26 1
0.27 1
0.28 1
0.29 1
0.3 1
0.31 1
0.32 1
0.33 1
0.34 1
0.35 1
0.36 1
0.37 1
0.38 1
0.39 1
0.4 1
0.41 1
0.42 1
0.43 1
0.44 1
0.45 1
0.46 1
0.47 1
0.48 1
0.49 1
0.5 1
0.51 1
0.52 1
0.53 1
0.54 1
0.55 1
0.56 1
0.57 1
0.58 1
0.59 1
0.6 1
0.61 1
0.62 1
0.63 1
0.64 1
0.65 1
0.66 1
0.67 1
0.68 1
0.69 1
0.7 1
0.71 1
0.72 1
0.73 1
0.74 1
0.75 1
0.76 1
0.77 1
0.78 1
0.79 1
0.8 1
0.81 1
0.82 1
0.83 1
0.84 1
0.85 1
0.86 1
0.87 1
0.88 1
0.89 1
0.9 1
0.91 1
0.92 1
0.93 1
0.94 1
0.95 1
0.96 1
0.97 1
0.98 1
0.99 1
1 1
1.02 1
1.03 1
1.04 1
1.05 1
1.06 1
1.07 1
1.08 1
1.09 1
1.1 1
1.11 1
1.12 1
1.13 1
1.14 1
1.15 1
1.16 1
1.17 1
1.18 1
1.19 1
1.2 1
1.21 1
1.22 1
1.23 1
1.24 1
1.25 1
1.26 1
1.27 1
1.28 1
1.29 1
1.3 1
1.31 1
1.32 1
1.33 1
1.34 1
1.35 1
1.36 1
1.37 1
1.38 1
1.39 1
1.4 1
1.41 1
1.42 1
1.43 1
1.44 1
1.45 1
1.46 1
1.47 1
1.48 1
1.49 1
1.5 1
1.51 1
1.52 1
1.53 1
1.54 1
1.55 1
1.56 1
1.57 1
1.58 1
1.59 1
1.6 1
1.61 1
1.62 1
1.63 1
1.64 1
1.65 1
1.66 1
1.67 1
1.68 0.99
1.69 1
1.7 1
1.71 1
1.72 1
1.73 1
1.74 1
1.75 1
1.76 1
1.77 1
1.78 1
1.79 1
1.8 1
1.81 1
1.82 0.99
1.83 1
1.84 1
1.85 1
1.86 1
1.87 1
1.88 0.99
1.89 1
1.9 0.98
1.91 0.99
1.92 1
1.93 0.99
1.94 0.99
1.95 0.99
1.96 0.99
1.97 1
1.98 1
1.99 1
2 0.99
};
\addplot [semithick, color1, opacity=0.75]
table {%
0.01 0.52
0.02 0.72
0.03 0.8
0.04 0.81
0.05 0.78
0.06 0.83
0.07 0.94
0.08 0.91
0.09 0.97
0.1 0.96
0.11 0.97
0.12 0.99
0.13 0.99
0.14 0.99
0.15 0.99
0.16 0.99
0.17 0.98
0.18 0.99
0.19 1
0.2 1
0.21 1
0.22 1
0.23 1
0.24 1
0.25 1
0.26 1
0.27 1
0.28 1
0.29 1
0.3 1
0.31 1
0.32 1
0.33 1
0.34 1
0.35 1
0.36 1
0.37 1
0.38 1
0.39 1
0.4 1
0.41 1
0.42 1
0.43 1
0.44 1
0.45 1
0.46 1
0.47 1
0.48 1
0.49 1
0.5 1
0.51 1
0.52 1
0.53 1
0.54 1
0.55 1
0.56 1
0.57 1
0.58 1
0.59 1
0.6 1
0.61 1
0.62 1
0.63 1
0.64 1
0.65 1
0.66 1
0.67 1
0.68 1
0.69 1
0.7 1
0.71 1
0.72 1
0.73 1
0.74 1
0.75 1
0.76 1
0.77 1
0.78 1
0.79 1
0.8 1
0.81 1
0.82 1
0.83 1
0.84 1
0.85 1
0.86 1
0.87 1
0.88 1
0.89 1
0.9 1
0.91 1
0.92 1
0.93 1
0.94 1
0.95 1
0.96 1
0.97 1
0.98 1
0.99 1
1 1
1.02 1
1.03 1
1.04 1
1.05 1
1.06 1
1.07 1
1.08 1
1.09 1
1.1 1
1.11 1
1.12 1
1.13 1
1.14 1
1.15 1
1.16 1
1.17 1
1.18 1
1.19 1
1.2 1
1.21 1
1.22 1
1.23 1
1.24 1
1.25 1
1.26 1
1.27 1
1.28 1
1.29 1
1.3 1
1.31 1
1.32 1
1.33 1
1.34 1
1.35 1
1.36 1
1.37 1
1.38 1
1.39 1
1.4 1
1.41 1
1.42 1
1.43 1
1.44 1
1.45 1
1.46 1
1.47 1
1.48 1
1.49 1
1.5 0.99
1.51 1
1.52 0.99
1.53 1
1.54 1
1.55 1
1.56 0.98
1.57 0.98
1.58 1
1.59 1
1.6 0.99
1.61 0.99
1.62 0.99
1.63 1
1.64 1
1.65 1
1.66 0.99
1.67 0.98
1.68 0.96
1.69 1
1.7 0.99
1.71 1
1.72 0.99
1.73 0.98
1.74 1
1.75 0.98
1.76 0.98
1.77 0.99
1.78 0.99
1.79 1
1.8 1
1.81 0.99
1.82 0.98
1.83 0.97
1.84 0.99
1.85 0.98
1.86 1
1.87 1
1.88 0.99
1.89 0.99
1.9 0.99
1.91 0.99
1.92 0.98
1.93 0.96
1.94 0.97
1.95 0.98
1.96 0.97
1.97 0.96
1.98 0.99
1.99 0.99
2 0.97
};
\addplot [semithick, color2, opacity=0.75]
table {%
0.01 0.48
0.02 0.49
0.03 0.62
0.04 0.61
0.05 0.61
0.06 0.75
0.07 0.78
0.08 0.76
0.09 0.81
0.1 0.89
0.11 0.93
0.12 0.99
0.13 0.98
0.14 0.99
0.15 0.99
0.16 1
0.17 0.98
0.18 0.99
0.19 1
0.2 1
0.21 1
0.22 1
0.23 1
0.24 1
0.25 1
0.26 1
0.27 1
0.28 1
0.29 1
0.3 1
0.31 1
0.32 1
0.33 1
0.34 1
0.35 1
0.36 1
0.37 1
0.38 1
0.39 1
0.4 1
0.41 1
0.42 1
0.43 1
0.44 1
0.45 1
0.46 1
0.47 1
0.48 1
0.49 1
0.5 1
0.51 1
0.52 1
0.53 1
0.54 1
0.55 1
0.56 1
0.57 1
0.58 1
0.59 1
0.6 1
0.61 1
0.62 1
0.63 1
0.64 1
0.65 1
0.66 1
0.67 1
0.68 1
0.69 1
0.7 1
0.71 1
0.72 1
0.73 1
0.74 1
0.75 1
0.76 1
0.77 1
0.78 1
0.79 1
0.8 1
0.81 1
0.82 1
0.83 1
0.84 1
0.85 1
0.86 1
0.87 1
0.88 1
0.89 1
0.9 1
0.91 1
0.92 1
0.93 1
0.94 1
0.95 1
0.96 1
0.97 1
0.98 1
0.99 1
1 1
1.02 1
1.03 1
1.04 1
1.05 1
1.06 1
1.07 1
1.08 1
1.09 1
1.1 1
1.11 1
1.12 1
1.13 1
1.14 1
1.15 1
1.16 1
1.17 1
1.18 1
1.19 1
1.2 1
1.21 1
1.22 1
1.23 1
1.24 1
1.25 1
1.26 1
1.27 1
1.28 1
1.29 1
1.3 1
1.31 1
1.32 1
1.33 1
1.34 1
1.35 1
1.36 1
1.37 1
1.38 1
1.39 1
1.4 1
1.41 1
1.42 1
1.43 1
1.44 1
1.45 1
1.46 1
1.47 1
1.48 1
1.49 1
1.5 0.99
1.51 1
1.52 0.99
1.53 1
1.54 1
1.55 1
1.56 0.98
1.57 0.99
1.58 1
1.59 1
1.6 0.99
1.61 0.99
1.62 0.99
1.63 1
1.64 1
1.65 1
1.66 0.99
1.67 0.99
1.68 0.96
1.69 1
1.7 0.99
1.71 1
1.72 0.99
1.73 0.98
1.74 1
1.75 0.98
1.76 0.99
1.77 0.99
1.78 0.99
1.79 1
1.8 1
1.81 0.99
1.82 0.99
1.83 0.99
1.84 0.98
1.85 0.96
1.86 1
1.87 1
1.88 0.99
1.89 0.99
1.9 0.99
1.91 0.98
1.92 0.99
1.93 0.97
1.94 0.97
1.95 0.98
1.96 0.97
1.97 0.94
1.98 0.99
1.99 0.99
2 0.97
};
\addplot [semithick, color3, opacity=0.75]
table {%
0.01 0.72
0.02 0.88
0.03 0.88
0.04 0.93
0.05 0.95
0.06 0.94
0.07 0.94
0.08 0.99
0.09 1
0.1 0.99
0.11 0.99
0.12 1
0.13 0.98
0.14 1
0.15 0.98
0.16 0.97
0.17 0.99
0.18 0.99
0.19 0.97
0.2 0.99
0.21 0.96
0.22 0.98
0.23 0.99
0.24 0.98
0.25 1
0.26 0.98
0.27 0.98
0.28 1
0.29 1
0.3 1
0.31 1
0.32 0.99
0.33 1
0.34 1
0.35 0.99
0.36 1
0.37 1
0.38 1
0.39 1
0.4 0.99
0.41 1
0.42 1
0.43 1
0.44 1
0.45 1
0.46 0.99
0.47 1
0.48 1
0.49 0.98
0.5 1
0.51 1
0.52 1
0.53 1
0.54 1
0.55 1
0.56 1
0.57 1
0.58 1
0.59 1
0.6 1
0.61 0.99
0.62 1
0.63 1
0.64 1
0.65 1
0.66 0.99
0.67 1
0.68 1
0.69 1
0.7 1
0.71 1
0.72 1
0.73 1
0.74 1
0.75 1
0.76 1
0.77 0.99
0.78 1
0.79 1
0.8 1
0.81 1
0.82 1
0.83 1
0.84 1
0.85 1
0.86 1
0.87 1
0.88 1
0.89 1
0.9 1
0.91 1
0.92 1
0.93 1
0.94 1
0.95 0.99
0.96 1
0.97 1
0.98 1
0.99 1
1 1
1.02 1
1.03 1
1.04 1
1.05 1
1.06 1
1.07 1
1.08 1
1.09 1
1.1 1
1.11 1
1.12 1
1.13 1
1.14 1
1.15 1
1.16 1
1.17 1
1.18 1
1.19 1
1.2 1
1.21 1
1.22 1
1.23 1
1.24 1
1.25 1
1.26 1
1.27 1
1.28 1
1.29 1
1.3 1
1.31 1
1.32 1
1.33 1
1.34 1
1.35 1
1.36 1
1.37 1
1.38 1
1.39 1
1.4 1
1.41 1
1.42 1
1.43 1
1.44 1
1.45 1
1.46 1
1.47 1
1.48 1
1.49 1
1.5 1
1.51 1
1.52 1
1.53 1
1.54 1
1.55 1
1.56 1
1.57 1
1.58 1
1.59 1
1.6 1
1.61 1
1.62 1
1.63 1
1.64 1
1.65 1
1.66 1
1.67 1
1.68 0.99
1.69 1
1.7 1
1.71 1
1.72 1
1.73 1
1.74 0.99
1.75 1
1.76 1
1.77 1
1.78 0.99
1.79 1
1.8 1
1.81 1
1.82 0.99
1.83 1
1.84 0.99
1.85 1
1.86 1
1.87 0.99
1.88 1
1.89 1
1.9 1
1.91 0.99
1.92 0.98
1.93 0.99
1.94 1
1.95 0.99
1.96 0.99
1.97 0.99
1.98 0.99
1.99 1
2 0.98
};
\addplot [semithick, color4, opacity=0.75]
table {%
0.01 0.73
0.02 0.88
0.03 0.89
0.04 0.93
0.05 0.95
0.06 0.94
0.07 0.94
0.08 0.99
0.09 1
0.1 0.99
0.11 0.99
0.12 1
0.13 0.98
0.14 1
0.15 0.98
0.16 0.97
0.17 0.99
0.18 0.99
0.19 0.97
0.2 1
0.21 0.96
0.22 0.98
0.23 0.99
0.24 0.98
0.25 1
0.26 0.98
0.27 0.98
0.28 1
0.29 1
0.3 1
0.31 1
0.32 0.99
0.33 1
0.34 1
0.35 0.99
0.36 1
0.37 1
0.38 1
0.39 1
0.4 0.99
0.41 1
0.42 1
0.43 1
0.44 1
0.45 1
0.46 0.99
0.47 1
0.48 1
0.49 0.98
0.5 1
0.51 1
0.52 1
0.53 1
0.54 1
0.55 1
0.56 1
0.57 1
0.58 1
0.59 1
0.6 1
0.61 0.99
0.62 1
0.63 1
0.64 1
0.65 1
0.66 0.99
0.67 1
0.68 1
0.69 1
0.7 1
0.71 1
0.72 1
0.73 1
0.74 1
0.75 1
0.76 1
0.77 0.99
0.78 1
0.79 1
0.8 1
0.81 1
0.82 1
0.83 1
0.84 1
0.85 1
0.86 1
0.87 1
0.88 1
0.89 1
0.9 1
0.91 1
0.92 1
0.93 1
0.94 1
0.95 0.99
0.96 1
0.97 1
0.98 1
0.99 1
1 1
1.02 1
1.03 1
1.04 1
1.05 1
1.06 1
1.07 1
1.08 1
1.09 1
1.1 1
1.11 1
1.12 1
1.13 1
1.14 1
1.15 1
1.16 1
1.17 1
1.18 1
1.19 1
1.2 1
1.21 1
1.22 1
1.23 1
1.24 1
1.25 1
1.26 1
1.27 1
1.28 1
1.29 1
1.3 1
1.31 1
1.32 1
1.33 1
1.34 1
1.35 1
1.36 1
1.37 1
1.38 1
1.39 1
1.4 1
1.41 1
1.42 1
1.43 1
1.44 1
1.45 1
1.46 1
1.47 1
1.48 1
1.49 1
1.5 1
1.51 1
1.52 1
1.53 1
1.54 1
1.55 1
1.56 1
1.57 1
1.58 1
1.59 1
1.6 1
1.61 1
1.62 1
1.63 1
1.64 1
1.65 1
1.66 1
1.67 1
1.68 0.99
1.69 1
1.7 1
1.71 1
1.72 1
1.73 1
1.74 0.99
1.75 1
1.76 1
1.77 1
1.78 0.99
1.79 1
1.8 1
1.81 1
1.82 0.99
1.83 1
1.84 0.99
1.85 1
1.86 1
1.87 0.99
1.88 1
1.89 1
1.9 1
1.91 1
1.92 0.99
1.93 0.99
1.94 1
1.95 0.99
1.96 0.99
1.97 0.99
1.98 0.99
1.99 1
2 0.98
};
\addplot [semithick, color5, opacity=0.75]
table {%
0.01 0.88
0.02 0.96
0.03 0.99
0.04 0.99
0.05 1
0.06 0.98
0.07 1
0.08 1
0.09 1
0.1 1
0.11 1
0.12 1
0.13 0.99
0.14 1
0.15 1
0.16 1
0.17 1
0.18 1
0.19 1
0.2 1
0.21 1
0.22 1
0.23 1
0.24 1
0.25 1
0.26 1
0.27 1
0.28 1
0.29 1
0.3 1
0.31 1
0.32 1
0.33 1
0.34 1
0.35 1
0.36 1
0.37 1
0.38 1
0.39 1
0.4 1
0.41 1
0.42 1
0.43 1
0.44 1
0.45 1
0.46 1
0.47 1
0.48 1
0.49 1
0.5 1
0.51 1
0.52 1
0.53 1
0.54 1
0.55 1
0.56 1
0.57 1
0.58 1
0.59 1
0.6 1
0.61 1
0.62 1
0.63 1
0.64 1
0.65 1
0.66 1
0.67 1
0.68 1
0.69 1
0.7 1
0.71 1
0.72 1
0.73 1
0.74 1
0.75 1
0.76 1
0.77 1
0.78 1
0.79 1
0.8 1
0.81 1
0.82 1
0.83 1
0.84 1
0.85 1
0.86 1
0.87 1
0.88 1
0.89 1
0.9 1
0.91 1
0.92 1
0.93 1
0.94 1
0.95 1
0.96 1
0.97 1
0.98 1
0.99 1
1 1
1.02 1
1.03 1
1.04 1
1.05 1
1.06 1
1.07 1
1.08 1
1.09 1
1.1 1
1.11 1
1.12 1
1.13 1
1.14 1
1.15 1
1.16 1
1.17 1
1.18 1
1.19 1
1.2 1
1.21 1
1.22 1
1.23 1
1.24 1
1.25 1
1.26 1
1.27 1
1.28 1
1.29 1
1.3 1
1.31 1
1.32 1
1.33 1
1.34 1
1.35 1
1.36 1
1.37 1
1.38 1
1.39 1
1.4 1
1.41 1
1.42 1
1.43 1
1.44 1
1.45 1
1.46 1
1.47 1
1.48 1
1.49 1
1.5 1
1.51 1
1.52 1
1.53 1
1.54 1
1.55 1
1.56 1
1.57 1
1.58 1
1.59 1
1.6 1
1.61 1
1.62 1
1.63 1
1.64 1
1.65 1
1.66 1
1.67 1
1.68 0.99
1.69 1
1.7 1
1.71 1
1.72 1
1.73 1
1.74 1
1.75 1
1.76 1
1.77 1
1.78 1
1.79 1
1.8 1
1.81 1
1.82 1
1.83 1
1.84 1
1.85 1
1.86 1
1.87 1
1.88 1
1.89 1
1.9 1
1.91 1
1.92 1
1.93 1
1.94 1
1.95 0.99
1.96 1
1.97 1
1.98 1
1.99 1
2 0.99
};
\addplot [semithick, color6, opacity=0.75, dashed]
table {%
0.01 1
0.02 1
0.03 1
0.04 1
0.05 1
0.06 1
0.07 1
0.08 1
0.09 1
0.1 1
0.11 1
0.12 1
0.13 1
0.14 1
0.15 1
0.16 1
0.17 1
0.18 1
0.19 1
0.2 1
0.21 1
0.22 1
0.23 1
0.24 1
0.25 1
0.26 1
0.27 1
0.28 1
0.29 1
0.3 1
0.31 1
0.32 1
0.33 1
0.34 1
0.35 1
0.36 1
0.37 1
0.38 1
0.39 1
0.4 1
0.41 1
0.42 1
0.43 1
0.44 1
0.45 1
0.46 1
0.47 1
0.48 1
0.49 1
0.5 1
0.51 1
0.52 1
0.53 1
0.54 1
0.55 1
0.56 1
0.57 1
0.58 1
0.59 1
0.6 1
0.61 1
0.62 1
0.63 1
0.64 1
0.65 1
0.66 1
0.67 1
0.68 1
0.69 1
0.7 1
0.71 1
0.72 1
0.73 1
0.74 1
0.75 1
0.76 1
0.77 1
0.78 1
0.79 1
0.8 1
0.81 1
0.82 1
0.83 1
0.84 1
0.85 1
0.86 1
0.87 1
0.88 1
0.89 1
0.9 1
0.91 1
0.92 1
0.93 1
0.94 1
0.95 1
0.96 1
0.97 1
0.98 1
0.99 1
1 1
1.02 1
1.03 1
1.04 1
1.05 1
1.06 1
1.07 1
1.08 1
1.09 1
1.1 1
1.11 1
1.12 1
1.13 1
1.14 1
1.15 1
1.16 1
1.17 1
1.18 1
1.19 1
1.2 1
1.21 1
1.22 1
1.23 1
1.24 1
1.25 1
1.26 1
1.27 1
1.28 1
1.29 1
1.3 1
1.31 1
1.32 1
1.33 1
1.34 1
1.35 1
1.36 1
1.37 1
1.38 1
1.39 1
1.4 1
1.41 1
1.42 1
1.43 1
1.44 1
1.45 1
1.46 1
1.47 1
1.48 1
1.49 1
1.5 1
1.51 1
1.52 1
1.53 1
1.54 1
1.55 1
1.56 1
1.57 1
1.58 1
1.59 1
1.6 1
1.61 1
1.62 1
1.63 1
1.64 1
1.65 1
1.66 1
1.67 1
1.68 1
1.69 1
1.7 1
1.71 1
1.72 1
1.73 1
1.74 1
1.75 1
1.76 1
1.77 1
1.78 1
1.79 1
1.8 1
1.81 1
1.82 1
1.83 1
1.84 1
1.85 1
1.86 1
1.87 1
1.88 1
1.89 1
1.9 1
1.91 1
1.92 1
1.93 1
1.94 1
1.95 1
1.96 1
1.97 1
1.98 1
1.99 1
2 1
};
\addplot [semithick, color7, opacity=0.75, dashed]
table {%
0.01 1
0.02 1
0.03 1
0.04 1
0.05 1
0.06 1
0.07 1
0.08 1
0.09 1
0.1 1
0.11 1
0.12 1
0.13 1
0.14 1
0.15 1
0.16 1
0.17 1
0.18 1
0.19 1
0.2 1
0.21 1
0.22 1
0.23 1
0.24 1
0.25 1
0.26 1
0.27 1
0.28 1
0.29 1
0.3 1
0.31 1
0.32 1
0.33 1
0.34 1
0.35 1
0.36 1
0.37 1
0.38 1
0.39 1
0.4 1
0.41 1
0.42 1
0.43 1
0.44 1
0.45 1
0.46 1
0.47 1
0.48 1
0.49 1
0.5 1
0.51 1
0.52 1
0.53 1
0.54 1
0.55 1
0.56 1
0.57 1
0.58 1
0.59 1
0.6 1
0.61 1
0.62 1
0.63 1
0.64 1
0.65 1
0.66 1
0.67 1
0.68 1
0.69 1
0.7 1
0.71 1
0.72 1
0.73 1
0.74 1
0.75 1
0.76 1
0.77 1
0.78 1
0.79 1
0.8 1
0.81 1
0.82 1
0.83 1
0.84 1
0.85 1
0.86 1
0.87 1
0.88 1
0.89 1
0.9 1
0.91 1
0.92 1
0.93 1
0.94 1
0.95 1
0.96 1
0.97 1
0.98 1
0.99 1
1 1
1.02 1
1.03 1
1.04 1
1.05 1
1.06 1
1.07 1
1.08 1
1.09 1
1.1 1
1.11 1
1.12 1
1.13 1
1.14 1
1.15 1
1.16 1
1.17 1
1.18 1
1.19 1
1.2 1
1.21 1
1.22 1
1.23 1
1.24 1
1.25 1
1.26 1
1.27 1
1.28 1
1.29 1
1.3 1
1.31 1
1.32 1
1.33 1
1.34 1
1.35 1
1.36 1
1.37 1
1.38 1
1.39 1
1.4 1
1.41 1
1.42 1
1.43 1
1.44 1
1.45 1
1.46 1
1.47 1
1.48 1
1.49 1
1.5 1
1.51 1
1.52 1
1.53 1
1.54 1
1.55 1
1.56 1
1.57 1
1.58 1
1.59 1
1.6 1
1.61 1
1.62 1
1.63 1
1.64 1
1.65 1
1.66 1
1.67 1
1.68 1
1.69 1
1.7 1
1.71 1
1.72 1
1.73 1
1.74 1
1.75 1
1.76 1
1.77 1
1.78 1
1.79 1
1.8 1
1.81 1
1.82 1
1.83 1
1.84 1
1.85 1
1.86 1
1.87 1
1.88 1
1.89 1
1.9 1
1.91 1
1.92 1
1.93 1
1.94 1
1.95 1
1.96 1
1.97 1
1.98 1
1.99 1
2 1
};
\addplot [semithick, color8, opacity=0.75, dashed]
table {%
0.01 1
0.02 1
0.03 1
0.04 1
0.05 1
0.06 1
0.07 1
0.08 1
0.09 1
0.1 1
0.11 1
0.12 1
0.13 1
0.14 1
0.15 1
0.16 1
0.17 1
0.18 1
0.19 1
0.2 1
0.21 1
0.22 1
0.23 1
0.24 1
0.25 1
0.26 1
0.27 1
0.28 1
0.29 1
0.3 1
0.31 1
0.32 1
0.33 1
0.34 1
0.35 1
0.36 1
0.37 1
0.38 1
0.39 1
0.4 1
0.41 1
0.42 1
0.43 1
0.44 1
0.45 1
0.46 1
0.47 1
0.48 1
0.49 1
0.5 1
0.51 1
0.52 1
0.53 1
0.54 1
0.55 1
0.56 1
0.57 1
0.58 1
0.59 1
0.6 1
0.61 1
0.62 1
0.63 1
0.64 1
0.65 1
0.66 1
0.67 1
0.68 1
0.69 1
0.7 1
0.71 1
0.72 1
0.73 1
0.74 1
0.75 1
0.76 1
0.77 1
0.78 1
0.79 1
0.8 1
0.81 1
0.82 1
0.83 1
0.84 1
0.85 1
0.86 1
0.87 1
0.88 1
0.89 1
0.9 1
0.91 1
0.92 1
0.93 1
0.94 1
0.95 1
0.96 1
0.97 1
0.98 1
0.99 1
1 1
1.02 1
1.03 1
1.04 1
1.05 1
1.06 1
1.07 1
1.08 1
1.09 1
1.1 1
1.11 1
1.12 1
1.13 1
1.14 1
1.15 1
1.16 1
1.17 1
1.18 1
1.19 1
1.2 1
1.21 1
1.22 1
1.23 1
1.24 1
1.25 1
1.26 1
1.27 1
1.28 1
1.29 1
1.3 1
1.31 1
1.32 1
1.33 1
1.34 1
1.35 1
1.36 1
1.37 1
1.38 1
1.39 1
1.4 1
1.41 1
1.42 1
1.43 1
1.44 1
1.45 1
1.46 1
1.47 1
1.48 1
1.49 1
1.5 1
1.51 1
1.52 1
1.53 1
1.54 1
1.55 1
1.56 1
1.57 1
1.58 1
1.59 1
1.6 1
1.61 1
1.62 1
1.63 1
1.64 1
1.65 1
1.66 1
1.67 1
1.68 1
1.69 1
1.7 1
1.71 1
1.72 1
1.73 1
1.74 1
1.75 1
1.76 1
1.77 1
1.78 1
1.79 1
1.8 1
1.81 1
1.82 1
1.83 1
1.84 1
1.85 1
1.86 1
1.87 1
1.88 1
1.89 1
1.9 1
1.91 1
1.92 1
1.93 1
1.94 1
1.95 1
1.96 1
1.97 1
1.98 1
1.99 1
2 1
};
\addplot [semithick, color9, opacity=0.75, dashed]
table {%
0.01 0.47
0.02 0.64
0.03 0.52
0.04 0.61
0.05 0.66
0.06 0.6
0.07 0.65
0.08 0.67
0.09 0.65
0.1 0.75
0.11 0.68
0.12 0.72
0.13 0.64
0.14 0.66
0.15 0.8
0.16 0.8
0.17 0.87
0.18 0.88
0.19 0.78
0.2 0.88
0.21 0.83
0.22 0.79
0.23 0.85
0.24 0.87
0.25 0.88
0.26 0.86
0.27 0.94
0.28 0.91
0.29 0.92
0.3 0.91
0.31 0.88
0.32 0.9
0.33 0.92
0.34 0.94
0.35 0.9
0.36 0.93
0.37 0.91
0.38 0.96
0.39 0.93
0.4 0.94
0.41 0.94
0.42 0.92
0.43 0.97
0.44 0.95
0.45 0.98
0.46 0.94
0.47 0.98
0.48 0.95
0.49 0.95
0.5 0.93
0.51 0.96
0.52 0.98
0.53 0.99
0.54 0.94
0.55 0.98
0.56 1
0.57 0.99
0.58 0.97
0.59 0.99
0.6 0.97
0.61 0.98
0.62 0.98
0.63 0.99
0.64 0.98
0.65 0.97
0.66 0.97
0.67 0.99
0.68 1
0.69 1
0.7 1
0.71 1
0.72 0.99
0.73 1
0.74 0.99
0.75 1
0.76 1
0.77 0.98
0.78 0.98
0.79 0.98
0.8 1
0.81 1
0.82 1
0.83 0.99
0.84 0.99
0.85 1
0.86 1
0.87 1
0.88 1
0.89 0.97
0.9 0.98
0.91 0.99
0.92 1
0.93 0.98
0.94 0.99
0.95 1
0.96 1
0.97 1
0.98 0.99
0.99 1
1 0.99
1.02 0.99
1.03 0.98
1.04 0.99
1.05 0.97
1.06 0.98
1.07 0.99
1.08 0.97
1.09 0.98
1.1 1
1.11 0.97
1.12 0.98
1.13 0.99
1.14 0.97
1.15 1
1.16 1
1.17 1
1.18 0.97
1.19 0.99
1.2 1
1.21 0.99
1.22 0.97
1.23 1
1.24 0.99
1.25 1
1.26 1
1.27 1
1.28 0.99
1.29 1
1.3 1
1.31 1
1.32 0.99
1.33 0.99
1.34 1
1.35 0.99
1.36 0.99
1.37 0.98
1.38 0.99
1.39 1
1.4 1
1.41 1
1.42 0.99
1.43 1
1.44 1
1.45 0.99
1.46 1
1.47 1
1.48 1
1.49 0.99
1.5 0.99
1.51 1
1.52 1
1.53 1
1.54 1
1.55 0.98
1.56 1
1.57 0.99
1.58 1
1.59 0.99
1.6 1
1.61 0.96
1.62 0.99
1.63 0.98
1.64 0.99
1.65 1
1.66 1
1.67 0.99
1.68 0.98
1.69 0.99
1.7 0.99
1.71 1
1.72 0.98
1.73 0.99
1.74 0.94
1.75 0.98
1.76 0.97
1.77 0.99
1.78 0.97
1.79 0.98
1.8 1
1.81 0.98
1.82 0.96
1.83 1
1.84 0.94
1.85 0.98
1.86 0.98
1.87 0.98
1.88 1
1.89 0.99
1.9 0.94
1.91 0.95
1.92 0.98
1.93 0.94
1.94 0.95
1.95 0.97
1.96 0.94
1.97 0.96
1.98 0.98
1.99 0.96
2 0.95
};
\addplot [semithick, color10, opacity=0.75, dashed]
table {%
0.01 0.48
0.02 0.55
0.03 0.49
0.04 0.61
0.05 0.59
0.06 0.56
0.07 0.66
0.08 0.56
0.09 0.58
0.1 0.68
0.11 0.67
0.12 0.75
0.13 0.64
0.14 0.73
0.15 0.76
0.16 0.75
0.17 0.77
0.18 0.88
0.19 0.79
0.2 0.85
0.21 0.82
0.22 0.83
0.23 0.86
0.24 0.86
0.25 0.88
0.26 0.83
0.27 0.85
0.28 0.84
0.29 0.96
0.3 0.88
0.31 0.84
0.32 0.87
0.33 0.92
0.34 0.92
0.35 0.91
0.36 0.94
0.37 0.95
0.38 0.94
0.39 0.92
0.4 0.92
0.41 0.94
0.42 0.95
0.43 0.98
0.44 0.95
0.45 0.94
0.46 0.92
0.47 0.94
0.48 0.94
0.49 0.92
0.5 0.96
0.51 0.95
0.52 0.96
0.53 0.97
0.54 0.96
0.55 0.97
0.56 0.98
0.57 0.98
0.58 0.95
0.59 0.98
0.6 0.96
0.61 0.98
0.62 0.98
0.63 0.99
0.64 0.93
0.65 0.96
0.66 0.96
0.67 1
0.68 1
0.69 0.98
0.7 0.99
0.71 0.98
0.72 0.99
0.73 1
0.74 1
0.75 0.98
0.76 0.99
0.77 0.98
0.78 1
0.79 0.99
0.8 1
0.81 0.99
0.82 1
0.83 1
0.84 0.99
0.85 1
0.86 1
0.87 0.99
0.88 0.98
0.89 0.96
0.9 0.98
0.91 0.98
0.92 1
0.93 0.98
0.94 0.99
0.95 0.99
0.96 0.98
0.97 1
0.98 0.99
0.99 0.99
1 0.97
1.02 0.98
1.03 0.95
1.04 0.98
1.05 0.93
1.06 0.95
1.07 0.99
1.08 0.96
1.09 0.95
1.1 0.96
1.11 0.95
1.12 0.94
1.13 0.96
1.14 0.96
1.15 0.99
1.16 0.98
1.17 0.96
1.18 0.97
1.19 0.99
1.2 0.99
1.21 0.99
1.22 0.97
1.23 0.99
1.24 0.98
1.25 0.98
1.26 1
1.27 0.98
1.28 0.99
1.29 1
1.3 0.98
1.31 1
1.32 0.99
1.33 0.99
1.34 1
1.35 0.99
1.36 0.99
1.37 0.97
1.38 0.98
1.39 1
1.4 0.99
1.41 1
1.42 0.98
1.43 0.98
1.44 0.99
1.45 0.98
1.46 0.99
1.47 1
1.48 0.99
1.49 0.99
1.5 0.99
1.51 0.99
1.52 1
1.53 1
1.54 0.99
1.55 0.99
1.56 0.99
1.57 0.99
1.58 0.99
1.59 1
1.6 1
1.61 0.97
1.62 1
1.63 0.97
1.64 0.97
1.65 1
1.66 1
1.67 0.99
1.68 0.97
1.69 0.97
1.7 0.99
1.71 0.99
1.72 0.98
1.73 0.98
1.74 0.96
1.75 0.97
1.76 0.95
1.77 0.99
1.78 0.98
1.79 0.98
1.8 1
1.81 0.97
1.82 0.96
1.83 0.98
1.84 0.96
1.85 0.94
1.86 0.98
1.87 0.98
1.88 0.99
1.89 0.99
1.9 0.94
1.91 0.97
1.92 0.96
1.93 0.95
1.94 0.97
1.95 0.97
1.96 0.91
1.97 0.94
1.98 0.96
1.99 0.98
2 0.94
};
\addplot [semithick, color11, opacity=0.75, dashed]
table {%
0.01 1
0.02 1
0.03 1
0.04 1
0.05 1
0.06 1
0.07 1
0.08 1
0.09 1
0.1 1
0.11 1
0.12 1
0.13 1
0.14 1
0.15 1
0.16 1
0.17 1
0.18 1
0.19 1
0.2 1
0.21 1
0.22 1
0.23 1
0.24 1
0.25 1
0.26 1
0.27 1
0.28 1
0.29 1
0.3 1
0.31 1
0.32 1
0.33 1
0.34 1
0.35 1
0.36 1
0.37 1
0.38 1
0.39 1
0.4 1
0.41 1
0.42 1
0.43 1
0.44 1
0.45 1
0.46 1
0.47 1
0.48 1
0.49 1
0.5 1
0.51 1
0.52 1
0.53 1
0.54 1
0.55 1
0.56 1
0.57 1
0.58 1
0.59 1
0.6 1
0.61 1
0.62 1
0.63 1
0.64 1
0.65 1
0.66 1
0.67 1
0.68 1
0.69 1
0.7 1
0.71 1
0.72 1
0.73 1
0.74 1
0.75 1
0.76 1
0.77 1
0.78 1
0.79 1
0.8 1
0.81 1
0.82 1
0.83 1
0.84 1
0.85 1
0.86 1
0.87 1
0.88 1
0.89 1
0.9 1
0.91 1
0.92 1
0.93 1
0.94 1
0.95 1
0.96 1
0.97 1
0.98 1
0.99 1
1 1
1.02 1
1.03 1
1.04 1
1.05 1
1.06 1
1.07 1
1.08 1
1.09 1
1.1 1
1.11 1
1.12 1
1.13 1
1.14 1
1.15 1
1.16 1
1.17 1
1.18 1
1.19 1
1.2 1
1.21 1
1.22 1
1.23 1
1.24 1
1.25 1
1.26 1
1.27 1
1.28 1
1.29 1
1.3 1
1.31 1
1.32 1
1.33 1
1.34 1
1.35 1
1.36 1
1.37 1
1.38 1
1.39 1
1.4 1
1.41 1
1.42 1
1.43 1
1.44 1
1.45 1
1.46 1
1.47 1
1.48 1
1.49 1
1.5 1
1.51 1
1.52 1
1.53 1
1.54 1
1.55 1
1.56 1
1.57 1
1.58 1
1.59 1
1.6 1
1.61 1
1.62 1
1.63 1
1.64 1
1.65 1
1.66 1
1.67 1
1.68 1
1.69 1
1.7 1
1.71 1
1.72 1
1.73 1
1.74 1
1.75 1
1.76 1
1.77 1
1.78 1
1.79 1
1.8 1
1.81 1
1.82 1
1.83 1
1.84 1
1.85 1
1.86 1
1.87 1
1.88 1
1.89 1
1.9 1
1.91 1
1.92 1
1.93 1
1.94 1
1.95 1
1.96 1
1.97 1
1.98 1
1.99 1
2 1
};
\addplot [semithick, black, opacity=1, dash pattern=on 1pt off 1pt]
table {%
-0.0895000000000001 0.5
2.0995 0.5
};
\addplot [semithick, black, opacity=1, dash pattern=on 1pt off 1pt]
table {%
-0.0895000000000001 1
2.0995 1
};
\addplot [semithick, black, opacity=1, dash pattern=on 1pt off 1pt]
table {%
1 0.2
1 1.05
};
\end{axis}

\end{tikzpicture}

%% file: sections/tex-plots-tables/tab_nonlin.tex
\begin{table*}[]
\caption{Summary for nonlinear models. The numbers reflect the ranges of noise that allow identifiability with accuracy $\approx 90\%$.}
\label{tab:summary:nonlinear}
\begin{tabular}{l|c|c|c|c|c|c|c|c|c}
Estimator             & $\mathcal{N}^3+\mathcal{N}$ & $\mathcal{N}^3+\mathcal{U}$ & $\mathcal{N}^3+\mathcal{L}$ & $\mathcal{U}^3+\mathcal{N}$ & $\mathcal{U}^3+\mathcal{U}$ & $\mathcal{U}^3+\mathcal{L}$ & $\mathcal{L}^3+\mathcal{N}$ & $\mathcal{L}^3+\mathcal{U}$ & $\mathcal{L}^3+\mathcal{L}$ \\
\hline
\textbf{HSIC}         & 0.09   -- 68     & 0.16   --        & 0.11   -- 70     & 0.04   -- 3      & 0.05   -- 6      & 0.05   -- 3      & 0.04   --        & 0.07   --        & 0.05   --        \\
\textbf{HISC\_IC}      & 0.10   -- 40     & 0.16   -- 88     & 0.10   -- 35     &                 &                 & 0.70 -- 1        & 0.07   --        & 0.12   --        & 0.06   --        \\
\textbf{HSIC\_IC2}     & 0.10   -- 40     & 0.17   -- 88     & 0.10   -- 35     &                 &                 & 0.70 -- 1        & 0.10   --        & 0.15   --        & 0.10   --        \\
\textbf{DISTCOV}      & 0.07   --        & 0.14   --        & 0.08   -- 88     & --85             & 0.03 -- 20       & 0.02   -- 5      & 0.03   --        & 0.04   --        & 0.03   --        \\
\textbf{DISTCORR}     & 0.07   --        & 0.14   --        & 0.05   -- 88     & --85             & 0.03   -- 20     & 0.02   -- 5      & 0.03   --        & 0.04   --        & 0.03   --        \\
\textbf{HOEFFDING}    & 0.02   --        & 0.05   --        & 0.04   --        & --95             & 0.03   --        & 0.02   --        & 0.02   --        & 0.02   --        & --               \\
\textbf{SH\_KNN}       & --               & --               & --               & --               & --               & --               & --               & --               & --               \\
\textbf{SH\_KNN\_2}     & --               & --               & --               & --               & --               & --               & --               & --               & --               \\
\textbf{SH\_KNN\_3}     & --               & --               & --               & --               & --               & --               & --               & --               & --               \\
\textbf{SH\_MAXENT1}   & 0.20   -- 45     & 0.33   -- 85     & 0.18   -- 67     & 0.02   -- 4      & 0.05   -- 7      & 0.03   -- 4      & 0.28   --        & 0.53   --        & 0.26   --        \\
\textbf{SH\_MAXENT2}   & 0.28   -- 45     & 0.43   -- 91     & 0.25   -- 65     & 0.08   -- 2      & 0.09   -- 5      & 0.06   -- 3      & 0.30   --        & 0.43   --        & 0.28   --        \\
\textbf{SH\_SPACING\_V} & --               & --               & --               & --               & --               & --               & --               & --               & --              
\end{tabular}
\end{table*}

%% file: sections/06-conclusions.tex
\section{Conclusions}
\label{sec:conclusions}

In this paper, we study the performance of a well-known causal discovery method
RESIT which falls in a group of additive noise models. While RESIT was widely studied in the literature before, previous research 
paid little attention to the effect of noise level on the accuracy of this approach. 
This work aims to fill this gap by means of an empirical study.
In our experiments, we tested a liner model $Y = X + N_Y$ and a nonlinear model
$Y = X^3 + N_Y$ with $X$ and $N_Y$ being drawn from one of the following distributions: Normal $\mathcal{N}$, Uniform $\mathcal{U}$ or Laplace $\mathcal{L}$.
We also used 12 different estimators
(6 independence estimators and 6 entropy estimators). 
The results from our experiments show that the effect of noise is not negligible
and can impact the model's identifiability.
For significantly small noise levels in the disturbance term $N_Y$
%(almost deterministic data), 
or significantly
large noise levels, this causal discovery method fails to capture
the true causal relationship of the given structural equation model.
\textit{Significantly} here depends on the model. For example, on some
models if the noise level is already twice larger than the variation of the 
independent variable, then the model becomes unidentifiable. 
Other models remained identifiable with 100 times larger noise levels, see 
\cref{sec:experimental-results} for details.

The range of different noise levels in our experiments
is quite exhaustive, changing from 100 times less to 100 times larger than the 
variance of the causal variable $X$. Some of these cases can be very rare in
practice, however, the discovered relationships can be useful for the practitioners
and researchers. In general, if the standard deviation
of the noise term is smaller than the standard deviation of the cause, then models remained identifiable more often
as opposed to the case when the standard deviation of the noise term is larger. For example, often when the standard deviation
of the noise term was only half of that of the cause, the model was still identifiable. However, in several
cases, if the standard deviation of the noise term was already twice larger than the standard deviation of the cause, then the model became unidentifiable. 

Our results also show differences in terms of the performance of the analyzed
estimators. In our experiments, Hilbert-Schmidt Independence Criterion with RBF 
Kernel is the best independence estimator, and Shannon entropy with Vasicek's spacing method is the best entropy estimator.
Comparing the performance on linear and non-linear models, our results show that
non-linear models are still identifiable in situations where linear models are not. 
For example, some non-linear models with the noise term's standard deviation of 100 
times higher than that of the cause, are perfectly identifiable while their linear
counterparts are not. 
Finally, our experiments show different behavior for different distribution types
(e.g., Gaussian, Uniform. or Laplace). Generally, models with the causal variable drawn from Laplace distribution $X \sim \mathcal{L}$ allow better identifiability.

In our experiments, we tested only two particular models and three different
distribution types. Similar results are expected with other
methods for causal discovery with additive noise models, as the failing point
is the independence estimation (or entropy estimation).
Therefore, methods relying on these estimations are generally prone to errors for some levels
of noise. This work also does not formalize the effect of different
noise levels in ANM causal discovery methods but it could be done in future work.
In reality, observed data does not always strictly follow a certain distribution type. 
As there are many different possible combinations, it would be interesting
to generalize the impact of different noise levels on any distribution by using the properties exhibited by an observed distribution.

%% file: sections/07-acknowledgement.tex
\section{Acknowledgments}

This work was partially supported by the European Union 
Horizon 2020 research programme  within
the project CITIES2030 ``Co-creating resilient and sustainable food towards 
FOOD2030'', grant 101000640.